\documentclass[11pt]{report}
\pdfoutput=1
\usepackage{amsmath}
\usepackage{amssymb}
\usepackage{amsfonts}
\usepackage{txfonts}
\usepackage{alltt}
\usepackage{caption}
\usepackage{subfigure}
\usepackage{float}

\usepackage[authoryear]{natbib}

\usepackage{boxedminipage}
\usepackage{comment}
\usepackage{graphicx}

\usepackage{longtable}
\usepackage{rotating}
\usepackage[usenames,dvipsnames,svgnames,table]{xcolor}
\usepackage{listings}
\definecolor{lightgray}{gray}{0.97}
\usepackage{color} %
\usepackage{hyperref}
\allowdisplaybreaks

\usepackage{dsfont}
\usepackage{listings}
\usepackage{boxedminipage}
\usepackage{float}
\usepackage{paralist}
\allowdisplaybreaks

\newtheorem{lemma}{Lemma}
\newtheorem{proposition}{Proposition}
\newtheorem{definition}{Definition}
\newtheorem{question}{Question}

\newcounter{count_a} %
\newcounter{count_f} %
\newcounter{count_i} %
\newcounter{count_e} %
\newcounter{count_r} %
\newcounter{count_x} %
\newcommand{\nexta}{\refstepcounter{count_a}a\arabic{count_a}}
\newcommand{\nextf}{\refstepcounter{count_f}f\arabic{count_f}}
\newcommand{\nexti}{\refstepcounter{count_i}i\arabic{count_i}}
\newcommand{\nexte}{\refstepcounter{count_e}e\arabic{count_e}}
\newcommand{\nextr}{\refstepcounter{count_r}r\arabic{count_r}}

\newtheorem{solution}{Solution}

\newcommand{\N}{\mathds{N}}

\begin{document}

\title{Representing, Reasoning and Answering Questions About Biological Pathways - Various Applications}

\author{Saadat Anwar\\
	Arizona State University,\\
	Tempe, AZ 85233\\
	USA\\
	\texttt{sanwar@asu.edu}
}

\date{\today}

\pagestyle{headings}
\maketitle

\begin{abstract}
Biological organisms are made up of cells containing numerous interconnected biochemical processes. Diseases occur when normal functionality of these processes is disrupted, manifesting as disease symptoms. Thus, understanding these biochemical processes and their interrelationships is a primary task in biomedical research and a prerequisite for activities including diagnosing diseases, and drug development. Scientists studying these interconnected processes have identified various pathways involved in drug metabolism, diseases, and signal transduction, etc. 

Over the last decade high-throughput technologies, new algorithms and speed improvements have resulted in deeper knowledge about biological systems and pathways, resulting in more refined models. These refined models tend to be large and complex, making it difficult for a person to remember all aspects of it. Thus, computer models are needed to represent and analyze them. The refinement activity itself requires reasoning with a pathway model by posing queries against it and comparing the results against a real biological system. We want to model biological systems and pathways in such a way that we can answer questions about them.

Many existing models focus on structural and/or factoid questions, relying on surface-level information that does not require understanding the underlying model. We believe these are not the kind of questions that a biologist may ask someone to test their understanding of the biological processes. We want our system to be able to answer the kind of questions a biologist may ask. So, we turned to early college level text books on biology for such questions. %

Thus the main goal of our thesis is to develop a system that allows us to encode knowledge about biological pathways and answer such questions about them that demonstrate understanding of the pathway. To that end, we develop a language that  will allow posing such questions  and illustrate the utility of our framework with various applications in the biological domain. We use some existing tools with modifications to accomplish our goal.

Finally, we use our question answering system in real world applications by extracting pathway knowledge from text and answering questions related to drug development.

\end{abstract}

\tableofcontents
\listoffigures

\chapter{Introduction}
\label{ch:intro}

Biological organisms are composed of cells that contain numerous interconnected and interacting biochemical processes occurring simultaneously. Disruptions in the normal functionality of these processes causes diseases, which appear as symptoms (of these diseases).  As a result understanding these processes is a fundamental activity in the biological domain and is prerequisite for activities such as disease diagnosis and drug discovery. One aspect of understanding the biological systems is the identification of pathways responsible for drug metabolism, diseases, and signal transduction, etc. 
The availability of high throughput approaches like micro-arrays, improvements in algorithms, and hardware that have come online during the last decade has resulted in significant refinement in these pathways. The pathways have become much larger in size and complexity to the degree that it is not reasonable for one person to fully retain all aspects of the pathway. As a result, computer based models of pathways are needed that allow the biologists to ask questions against them and compare them with real-world knowledge. The model should be such that it has an understanding of the pathway. Such a system would be considered intelligent and would assist the biologists in expanding the breadth of their search for new drugs and diagnoses. Source knowledge for these pathways comes from volumes of research papers published every year. Though there are a number of curated pathway resources available, they significantly lag behind the current state of the research in biology. As a result, we need a way to extract this pathway information from published text.

\section{Choosing the right questions}
A large body of research exists on computer modeling of biological processes and it continues to be an active area of research. However, many such models focus on surface properties, like structure; or factoid questions. Though important, we feel these systems do not test the understanding of the underlying system being modeled.  We want to go beyond this surface level information and answer questions requiring deeper reasoning. We want our system to answer questions that a biology teacher expects his / her students to answer after reading the required text. So, we turned to college level biological text books for the questions that we feel are more indicative of such understanding. Following questions from \cite{CampbellBook} illustrate the kind of questions we are interested in answering:
\begin{itemize}
\item ``What would happen to the rate of glycolysis if DHAP were removed from the process of glycolysis as quickly as it was produced?''
\item ``A muscle cell had used up its supply of oxygen and ATP. Explain what affect would this have on the rate of cellular respiration and glycolysis?''
\end{itemize}
These questions and others like it were the subject of a recent deep knowledge representation challenge\footnote{https://sites.google.com/site/2nddeepkrchallenge/}. In this thesis, we focus on questions that require reasoning over simulations.

\section{Choosing the right tools}
Data about biological systems can be qualitative or quantitative in nature. The fully quantitative data about reaction dynamics is based on ordinary differential equations (ODEs) of reaction kinetics, which are often lacking~\cite{chaouiya2007petri}. Qualitative data is more prevalent. It is less precise, but tends to capture general relationships between various components of a biological pathway. Adding quantitative information to a qualitative model provides the next step in refinement of the biological pathways~\cite{heiner2004model}, providing better coverage of biological systems and processes.
We want to use this qualitative+quantitative data for our modeling.

To simulate and reason with the pathways, we need tools that can model a biological pathway that contains qualitative+quantitative information, simulate the pathway and reason with the results.

\section{Need for a pathway a specification and a query language} 
Pathway information comes in various formats, such as cartoon drawings, formal graphical representations like Kohn's Maps~\cite{kohn2006molecular}, curated databases of pathways~\cite{kanehisa2000kegg,karp2002metacyc,croft2011reactome} and free text. The depth of this knowledge as well as its taxonomy varies with the source. Thus, a common specification language is needed. Such a language must be easy to understand and must have a well defined semantics.

Queries are normally specified in natural language, which is vague. So, a more precise query language is needed. One could ask queries in one of the existing formal languages~\cite{gelfond1998action}, but that will be burdensome for a user to become fluent. As a result, we need a language that has a simple English-like syntax, but a well defined semantics, so that it does not have the vagaries of the natural language.

\section{Text extraction}
Knowledge about biological pathways is spread over collections of published papers as nuggets of information, such as relationships between proteins; between proteins and drugs; genetic variation; and association of population groups with genetic variation; to name a few. Published research may also contain contradictory information, e.g. an earlier conjecture that was proven to be untrue in later research, or knowledge with limited amount of certainty. To extract these nuggets and to assemble them into a coherent pathway requires background knowledge, similar to other technical fields. Portions of this knowledge are published in books and online repositories. Thus, we need a method of text extraction that allows one to extract nuggets of information, consult available databases and produce knowledge about pathway that is self-consistent.

\section{Overview}
In this thesis, we propose to build a system, called BioPathQA, to answer deeper reasoning questions using existing tools with modifications.
To that end, we develop a language to specify pathways and queries. Our system is designed to answer reasoning questions requiring simulation. We demonstrate the applicability of our system with applications to drug development on knowledge obtained from text extraction.

To implement an answering system that can answer simulation based reasoning questions, we first looked for available tools that could help in this task and we found Petri Nets as providing the right level of formalism for our application. Petri Nets~\cite{PetersonPetriNets}
are a popular representation formalism used to model biological systems and to simulate them. They have been used to model and analyze the dynamic behavior as well as structural properties of biological systems. However, such analysis is usually limited to invariant determination, liveness, boundedness and reachability. To our knowledge they have not been used to answer questions similar to the aforementioned. 

In order to represent deeper reasoning questions, we have to make extensions to the Petri Net model as the basic model lacks sufficient richness. For example, we may want to change the firing semantics to limit the state space or maximize parallel activity. Although numerous Petri Net modeling, simulation and analysis systems exist~\cite{jensen2007coloured,rohr2010snoopy,kounev2006qpme,berthomieu2004tool,nagasaki2010cell,kummer2000renew}, we found certain limitations in the default implementation of these systems as well that prevented us from using them as is. For example, the Colored Petri Net implementation CPNtools~\footnote{http://cpntools.org} does not allow inhibitor arcs (we use to model protein inhibition); Cell Illustrator~\cite{nagasaki2010cell} is closed source and does not support colored tokens (we use to model locations); Snoopy~\cite{rohr2010snoopy} supports a large number of extensions, but it is unclear how one exports the simulation results for further reasoning; and most did not allow exploring all possible state evolutions of a pathway or using different firing semantics.

To make these extensions in an easy manner we use Answer Set Programming (ASP)~\cite{ASP} as the language to represent and simulate Petri Nets. It allows a simple encoding of the Petri Net and can be easily extended to incorporate extensions~\footnote{Certain commercial tools, like Cell Illustrator (http://www.cellillustrator.com) do allow exporting their model into a high level language, but we believe that a declarative language is more suited to succinctly describe the problem.}. In addition, ASP allows powerful reasoning capability and the possibility of implementing additional constructs not supported by Petri Nets directly, such as the ability to filter trajectories.

Petri Net to ASP translation has been studied before~\cite{LogicPetriNets,HeljankoNMR}. However, these implementations have been limited to specific classes of Petri Nets and have different focus. For example, the Simple Logic Petri Nets~\cite{LogicPetriNets} do not allow numerical accumulation of the same tokens from multiple transitions to a single place and the Binary Petri Nets~\cite{HeljankoNMR} do not allow more than one tokens at any place.

\section{Specific contributions}
The research contribution of this thesis can be divided into four major parts. The first part gives a general encoding of Petri Nets in ASP, which allows easy extension by making local changes. The second part shows how the ASP encoding of Petri Nets can be used to answer simulation based reasoning questions. The third part describes the high-level language for pathway and query specification; and the system that we have developed to answer deep reasoning questions. The fourth part shows how knowledge is extracted from text of research papers, cleaned and assembled into a pathway to answer simulation based reasoning questions using our system.

\subsection{General ASP encoding of Petri Net for simulation} %
Although previous work on encoding Petri Nets in ASP exists, it is limited to specific classes of Petri Nets. We present an encoding of a basic Petri Net in ASP to show it is done in an intuitive yet concise way. The default execution semantics of a Petri Net is the {\em set}-semantics, which allows a subset of transitions to fire simultaneously when ready. This can result in far too many combinations of transition firing arrangements. A simpler approach is to use the so called {\em interleaved} execution semantics, in which at most one transition fires when ready. This too can generate many firing arrangements.
Biological systems are highly parallel in nature, as a result it is beneficial to model maximum parallel activity. So, we introduce a new firing semantics, called the {\em maximal firing set} semantics by extending the {\em set semantics}. In this semantics, a maximal subset of non-conflicting enabled transitions fire simultaneously when ready.

Then, we extend the basic ASP encoding to include Petri Net extensions like reset-arcs (to model immediate consumption of any amount of substrate), inhibit-arcs (to model gene/protein inhibition), read-arcs (to model additional pre-conditions of a reaction, such as different start vs. maintenance quantity of a reactant), colored-tokens (to model quantities of different types of substances at the same location), priority-transitions (to select between alternate metabolic paths), and timed-transitions (to model slow reactions) that allow modeling of various concepts in biological systems. We show how ASP allows us to make these extensions with small amount of local changes.

This component is one of the major focuses of our research. It is described in Chapter \ref{ch:asp_enc} and is the basis for implementation of our system to model pathways and answer questions about them.

\subsection{Answering simulation based reasoning questions} %
We use the encoding developed in Chapter~\ref{ch:asp_enc} to questions in \cite[Chapter 9]{CampbellBook} that were a part of the Second Deep Knowledge Representation Challenge~\footnote{\texttt{https://sites.google.com/site/2nddeepkrchallenge/}}. These questions are focused on the mechanism of cellular respiration and test the understanding of the student studying the material; and appear in two main forms: 
\begin{inparaenum}[(i)]
\item inquiry about change of rate of a process due to a change in the modeled system, and 
\item explanation of a change due to a change in the modeled system.
\end{inparaenum}

We built Petri Net models for the situations specified in the questions, encoded them in ASP and simulated them over a period of time. For change of rate questions, we computed the rate for both nominal and modified cases and observed that they matched the responses provided with the challenge questions.  For the explanation of change questions, we collected the summary of firing transitions as well as substance quantities produced at various times. This information formed the basis of our answer. We compared our results with the answers provided with the challenge questions. 

A novel aspect of our approach is that we apply the initial conditions and interventions mentioned in the questions as modifications to the pathway representation. These interventions can be considered as a generalized form of actions.

For certain questions, additional domain knowledge outside the source material was required. We filled this gap in knowledge as necessary. We also kept the models to a subset of the pathway for performance as well as to reduce clutter in the output that can bury the results with unnecessary details.

This component of our research is described in Chapter \ref{ch:modeling_qa}.

\subsection{BioPathQA: a system and a language to represent pathways and query them}
We combined the techniques learned from Chapter~\ref{ch:asp_enc}, action languages, and biological modeling languages to build a question answering system that takes a pathway and a query as input. Both the pathway specification language and the query language have strict formal semantics, which allow them to be free of the vagaries of natural language, the language of the research papers as well as the query statements. 

\subsubsection{Guarded-Arc Petri Net}
Since the biological pathways are constructed of biochemical reactions, they are effected by environmental changes. Mutations within the cell can also result in conditional change in behavior of certain processes. As a result, we needed actions with conditional effects. Our Petri Net model wasn't rich enough to model conditional actions, so we extended the Petri Nets with conditional arcs. We call this extension, the Guarded-Arc Petri Net, where a guard is a condition on an arc, which must be true for that arc to be traversed. With this extension, a Petri Net transition can have different outcomes for different markings. Our model is similar to the model in \cite{jensen2007coloured} in many aspects.

This component of our research is described in Chapter~\ref{ch:deepqa}.

\subsection{Text Extraction to Answer Questions about Real World Applications} %
To apply our system to real world applications, we have to extract pathway knowledge from published papers, which are published in natural language text. For this, we use a system called the Parse Tree Query Language (PTQL)~\cite{PTQL} to nuggets of information from the abstracts published on PubMed~\footnote{http://www.ncbi.nlm.nih.gov/pubmed}. Sentences are parsed using the Link Grammar~\cite{LinkGrammar} or Stanford Parser~\cite{StanfordParser}; with various object-classes identified within the sentence. Unlike Information Retrieval (IR) approaches that tend to treat documents as unstructured bags-of-words, PTQL treats words (or word-groups) as sentence elements with syntactic as well as dependency relationships between them. PTQL queries combine lexical, syntactic and semantic features of sentence elements. Thus with PTQL, one can ask for the first-noun of a noun-phrase that is the direct-object of a verb-phrase for some specific verb string. To accomplish its task, PTQL performs a number of pre-processing steps on its input useful for text extraction and leverages on various existing databases. These include sentence splitting, tokenization, part-of-speech (POS) tagging, named entity recognition, entity-mention normalization, cross-linking with concepts from external databases, such as Gene Ontology~\cite{GOA} and UniProt~\cite{UniProt}. 
We extract gene-gene, gene-drug, and gene-disease relationships using PTQL, assemble them into a pathway specification and reason with the extracted knowledge to determine possible drug interactions. 

Facts and relationships extracted using PTQL are further subject to filtering to remove inconsistent information. A pathway specification is then constructed from the extracted facts, which can be queried using the query specification language. We illustrate the use of our deep reasoning system by an example from the drug-drug interaction domain.

This component is described in Chapter~\ref{ch:prev_work}.

\section{Summary}
The main contributions of this thesis can be summarized as follows:
\begin{enumerate}
\item Generalized Petri Net encoding in ASP, including a new maximal firing set semantics (Chapter~\ref{ch:asp_enc})
\begin{itemize}
\item An easy to extend encoding is developed, that allows adding extensions using local changes
\item A new Petri Net firing semantics, the so called maximal firing set semantics is defined, which ensures maximum possible parallel activity at any given point 
\end{itemize}
\item Answering simulation based deep reasoning questions using our ASP encoding (Chapter~\ref{ch:modeling_qa})
\begin{itemize}
\item It is shown, how deep reasoning questions requiring simulation based reasoning can be answered.
\end{itemize}
\item Developed a system called BioPathQA and a language to specify biological pathways and answer deep reasoning questions about it (Chapter~\ref{ch:deepqa})
\begin{itemize}
\item A pathway specification language is developed, combining concepts from Petri Nets, Action Languages, and Biological Pathways
\item A query specification language is developed, which is english like, with well defined semantics, avoiding the vagaries of Natural Language
\item A description of our implementation using ASP and Python is given; and an execution trace is shown
\end{itemize}
\item Performed text extraction to extract pathway knowledge (Chapter~\ref{ch:prev_work})
\begin{itemize}
\item It is shown pathway knowledge is extracted and used to answer questions in the drug-drug interaction domain
\end{itemize}
\end{enumerate}

\chapter{Petri Net Encoding in ASP for Biological Domain}\label{ch:asp_enc}

\section{Introduction}
Petri Net~\cite{PetersonPetriNets} is a graphical modeling language with formal semantics used for description of distributed systems. It is named after Carl Adam Petri, who formally defined Petri Nets in his PhD thesis in the 1960's \cite{Brauer2006}. Petri nets have been widely used to model a wide range of systems, from distributed systems to biological pathways. The main advantages of Petri Net representation include its simplicity and the ability to model concurrent and asynchronous systems and processes.

A variety of Petri Net extensions have been proposed in the literature, e.g. inhibitor arcs, reset transitions, timed transitions, stochastic transitions, prioritized transitions, colored petri nets, logic petri nets, hierarchical petri nets, hybrid petri nets and functional petri nets to a name a few \cite{LogicPetriNets,music2012schedule,Hardy2004}. %

Our interest in Petri Nets is for representing biological pathways and simulating them in order to answer simulation based reasoning questions. We show how Petri nets can be represented in ASP. We also demonstrate how various extensions of basic Petri nets can be easily expressed and implemented by making small changes to the initial encoding. During this process we will relate the extensions to their use in the biological domain. Later chapters will show how this representation and simulation is used to answer biologically relevant questions.

The rest of this chapter is organized as follows: We present some background material on Answer Set Programming (ASP) and Petri Nets. Following that, we present the Answer Set encoding of a basic Petri Net. After that we will introduce various Petri Nets extensions and the relevant ASP code changes to implement such extensions.

\section{Background}

\subsection{Answer Set Programming}\label{sec:asp}
Answer Set Programming (ASP) is a declarative logic programming language that is based on the Stable Model Semantics \cite{StableModels}. It has been applied to a problems ranging from spacecrafts, work flows, natural language processing and biological systems modeling. 

Although ASP language is quite general, we limit ourselves to 
language and extensions relevant to our work.

\begin{definition}[Term]\label{def:term}
A {\em term} is a term in the propositional logic sense.
\end{definition}

\begin{definition}[Literal]\label{def:literal}
A {\em literal} is an atom in the propositional logic sense.  A literal prefixed with $\mathbf{not}$ is referred to as a negation-as-failure literal or a {\em naf-literal}, with $\mathbf{not}$ representing negation-as-failure.
\end{definition}

We will refer to propositional atoms as {\em basic atoms} to differentiate them from other atoms, such as the aggregate atoms defined below.

\begin{definition}[Aggregate Atom]\label{def:agg:atom}
A {\em sum aggregate atom} is of the form:
\begin{align}\label{form:agg:atom}
L \; [ B_0=w_0,\dots,B_m=w_m ] \; U
\end{align}
where, $B_i$ are basic atoms, $w_i$ are positive integer weight terms, $L,U$ are integer terms specifying the lower and upper limits of aggregate weights. The lower and upper limits are assumed to be $-\infty$ and $\infty$, if not specified.
\end{definition}
A {\em count aggregate atom} is a special case of the {\em sum aggregate atom} in which all weights are $1$, i.e. 
$L \; [ B_0=1,\dots,B_m=1] \; U$ 
and it is represented by:
\begin{align}\label{form:count:agg:atom}
L \; \{ B_0,\dots,B_m \} \; U
\end{align}

A {\em choice atom} is a special case of the {\em count aggregate atom}~\eqref{form:count:agg:atom} in which $n=m$.

\begin{definition}[ASP Program]\label{def:asp:program}
An {\em ASP program} $\Pi$ is a finite set of rules of the following form:
\begin{align}\label{form:asp:rule}
A_0 \leftarrow A_1,\dots,A_m,\mathbf{not~} B_{1},\dots,\mathbf{not~} B_n, C_1,\dots,C_k.
\end{align}
where each $A_0$ is either a basic atom or a choice atom, $A_i$ and $B_i$ are basic atoms, $C_i$ are aggregate atoms and $\mathbf{not}$ is negation-as-failure.
\end{definition}

In rule~\eqref{form:asp:rule}, $\{ A_0 \}$ is called the {\em head} of the rule, and $\{A_1,\dots,$ $A_m,$ $\mathbf{not~} B_{1},\dots,$ $\mathbf{not~} B_n, $ $C_1,\dots,$ $C_k\}$ is called its {\em tail}. A rule in which $A_0$ is a choice atom is called a {\em choice rule}. A rule without a head is called a {\em constraint}. A rule with a basic atom as its head and empty tail is called a {\em fact} in which case the ``$\leftarrow$'' is dropped.

Let $R$ be an ASP rule of the form~\eqref{form:asp:rule} and let $pos(R) = \{ A_1,\dots,A_m \}$ represent the positive atoms, $neg(R) = \{ B_1,\dots,B_n \}$ the negation-as-failure atoms, and $agg(R) = \{ C_1,\dots,C_k \}$ represent the aggregate atoms in the body of a rule $R$. Let $lit(A)$ represent the set of basic literals in atom $A$, i.e. $lit(A)=\{A\}$ if $A$ is a basic atom; $lit(A)=\{B_0,\dots,B_n\}$ if $A$ is an aggregate atom. Let $C$ be an aggregate atom of the form~\eqref{form:agg:atom} and let $pos(C) = \{ B_0,\dots,B_m \}$ 
be the sets of basic positive 
literals such that 
$lit(C) = pos(C)$. 

Let $lit(R) = lit(head(R)) \cup pos(R) \cup neg(R) \cup \bigcup_{C \in agg(R)}{lit(C)} $ for a rule $R \in \Pi$ and $lit(\Pi) = \bigcup_{R \in \Pi}{lit(R)}$ be the set of basic literals in ASP program $\Pi$.

\begin{definition}[Aggregate Atom Satisfaction]\label{def:agg:sat}
A ground aggregate atom $C$ of the form~\eqref{form:agg:atom} is satisfied by a set of basic ground atoms $S$, if 
$L \leq \sum_{0 \leq i \leq m, B_i \in S}{w_i} \leq U$ 
and we write $S \models C$.
\end{definition}

Given a set of basic ground literals $S$ and a basic ground atom $A$, we say $S \models A$ if $A \in S$, $S \models \mathbf{not~} A$ if $A \not\in S$. For a rule $R$ of the form~\eqref{form:asp:rule} $S \models body(R)$ if $\forall A \in \{A_1,\dots,A_m\}, S \models A$, $\forall B \in \{ B_{1},\dots,B_n \}, S \models \mathbf{not~} B$, and $\forall C \in \{ C_1,\dots,C_k \}, S \models C$; $S \models head(R)$ if $S \models A_0$. 

\begin{definition}[Rule Satisfaction]\label{def:rule:sat}
A ground rule $R \in \Pi$ is satisfied by a set of basic ground atoms $S$, iff, $S \models body(R)$ implies $S \models head(R)$. A constraint rule $R \in \Pi$ is satisfied by set $S$ if $S \not\models body(R)$.

\end{definition}

We define reduct of an ASP program by treating aggregate atoms in a similar way as negation-as-failure literals, since our code does not contain recursion through aggregation (which can yield non-intuitive answer-sets~\cite{son2007constructive}).
\begin{definition}[Reduct]\label{def:reduct}
Let $S$ be a set of ground basic atoms, the reduct of ground ASP program $\Pi$ w.r.t. $S$, written $\Pi^S$ is the set of rules: $\{ p \leftarrow A_1,\dots,A_m. \; | \; A_0 \leftarrow A_1, \dots, $ $A_m, $ $\mathbf{not~} B_{1}, \dots, $ $\mathbf{not~} B_n,  $ $C_1,\dots,$ $C_k. \in \Pi, p \in lit(A_0) \cap S, \{A_1,\dots,A_m\} \subseteq S, \{ B_1,\dots,B_n \} \cap S = \emptyset, $ $\nexists C \in \{ C_1,\dots,C_k \}, S \not\models C \}$. 
\end{definition}

Intuitively, this definition of reduct removes all rules which contain a naf-literal or an aggregate atom in their bodies that does not hold in $S$, and it removes aggregate atoms as well as naf-literals from the body of the remaining rules.

Heads of choice-rules are split into multiple rules containing at most one atom in their heads. The resulting reduct is a program that does not contain any aggregate atoms or negative literals. The rules of such a program are monotonic, such that if it satisfied by a set $S$ of atoms, it is also satisfied by any superset of $S$. 

A {\bf deductive closure} of such a (positive) monotonic program is defined as the unique smallest set of atoms $S$ such that whenever all body atoms of a rule hold in $S$, the head also holds in $S$. The deductive closure can be iteratively computed by starting with an empty set and adding heads of rules for which the bodies are satisfied, until a fix point is reached, where no additional rules can be satisfied. (adopted from ~\cite{Baral2003})

\begin{definition}[Answer Set]\label{def:ans:set}
A set of basic ground atoms $S$ is an answer set of a ground ASP program $\Pi$, iff $S$ is equal to the deductive closure of $\Pi^S$ and $S$ satisfies each rule of $\Pi$. (adopted from~\cite{Baral2003})
\end{definition}

\subsubsection{Clingo Specific Syntactic Elements}

The ASP code in this thesis is in the syntax of ASP solver called clingo \cite{Clingo}. The ``$\leftarrow$'' in ASP rules is replaced by the symbol ``\texttt{:-}''. Though the semantics of ASP are defined on ground programs, Clingo allows variables and other constructs for compact representation. We intuitively describe specific syntactic elements and their meanings below:

{\bf Comments:} Text following ``\%'' to the end of the line is treated as a comment.

{\bf Interval:} Atoms defined over an contiguous range of integer values can be compactly written as intervals, e.g. $p(1\;..\;5)$ represents atoms $p(1), p(2), p(3), p(4), p(5)$.

{\bf Pooling:} Symbol ``;'' allows for pooling alternative terms. For example, an atom $p(\dots,X,\dots)$ and $p(\dots,Y,\dots)$ can be pooled together into a single atom as $p(\dots,X;Y,\dots)$.

{\bf Aggregate assignment atom:} An aggregate assignment atom $Q=\#sum[A_0=w_0,$ $\dots,$ $A_m=w_m,$ $\mathbf{not~} A_{m+1}=w_{m+1},\dots,$ $\mathbf{not~} A_n=w_n]$ assigns the sum \\ $\sum_{A_i \in S, 0 \leq i \leq m}{w_i} + $ $\sum_{A_j \notin S, m+1 \leq n}{w_j}$ to $Q$ w.r.t. a consistent set of basic ground atoms $S$.

{\bf Condition:} Conditions allow instantiating variables to collections of terms within aggregates, e.g. $\{ p(X) : q(X) \}$ instantiates $p(X)$ for only those $X$ that $q(X)$ satisfies. For example, if we have $p(1..5)$ but only $q(3;5)$, then $\{ p(X) : q(X) \}$ is expanded to $\{ p(3), p(5) \}$.

\subsubsection{Grounding}
Grounding makes a program variable free by replacing variables with the possible values they can take. Clingo uses the grounder Gringo~\cite{gebser2007gringo} for ``smart'' grounding, which results in substantial reduction in the size of the program. Details of this grounding are implementation specific. We present the intuitive process of grounding below.

\begin{enumerate}
\item A set of ground terms is constructed, where a ground term is a term that contains no variables.
\item The variables are split into two categories: local and global. Local variables are the ones that appear only within an aggregate atom (minus the limits) and nowhere else in a rule. Such variables are considered local writ. the aggregate atom. All other variables are considered global.
\item First the global variables are eliminated in the rules as follows:
\begin{itemize}
\item Each rule $r$ containing an aggregate assignment atom of the form~\eqref{form:agg:atom} is replaced with set of rules $r'$ in which the aggregate assignment atom is is replaced with an aggregate atom with lower and upper bounds of $Q$ for all possible substitutions of $Q$. This is generalized to multiple aggregate assignment atoms by repeating this step for each such atom, where output of previous iteration forms the input of the next iteration.
\item Each rule $r'$, is replaced with the set of all rules $r''$ obtained by all possible substitutions of ground terms for global variables in $r$.
\end{itemize}
\item Then the local variables are eliminated in the rules by expanding conditions, such that $p(\dots,X,\dots) : d(X)$ are replaced by $p(\dots,d_1,\dots), \dots, p(\dots,d_k,\dots)$ for the extent $\{d_1,\dots,d_k\}$ of $d(X)$. This is generalized to multiple conditions in the obvious way.
\end{enumerate}

Following the convention of the Clingo system, Variables in rules presented in this thesis start with capital letters while lower-case text and  numbers are constants. Italicized text represents a constant term from a definition in context.

A recent work \cite{harrison2013semantics} gives the semantics of Gringo with ASP Core 2 syntax~\cite{calimeri2013asp} using Infintary Propositional Formulas, which translate Gringo to propositional formulas with infinitely long conjunctions and disjunctions. Their approach removes the safety requirement, but the subset of Gringo presented does appear to cover assignments. Although their approach provides a way to improve our ASP encoding by removing the requirement of specifying the maximum number of tokens or running simulations until a condition holds, our simpler (limited) semantics is sufficient for the limited syntax and semantics we use.

\subsection{Multiset}
A {\bf multiset} $A$ over a domain set $D$ is a pair $\langle D,m \rangle$, where $m: D \rightarrow \mathds{N}$ is a function giving the multiplicity of $d \in D$ in $A$. Given two multsets $A = \langle D,m_A \rangle, B = \langle D,m_B \rangle$, $A \odot B$ if $\forall d \in D: m_A(d) \odot m_B(d)$, where $\odot \in \{<,>,\leq,\geq,=\}$, and $A \neq B$ if $\exists d \in D : m_A(d) \neq m_B(d)$. Multiset sum/difference is defined in the usual way.  We use the short-hands $d \in A$ to represent $m_A(d) > 0$, $A = \emptyset$ to represent $\forall d \in D, m(d) = 0$, $A \otimes n$ to represent $\forall d \in D, m(d) \otimes n$, where $n \in \mathds{N}$, $\otimes \in \{<,>,\leq,\geq,=,\neq\}$. We use the notation $d/n \in A$ to represent that $d$ appears $n$-times in $A$; we drop $A$ when clear from context. The reader is referred to \cite{syropoulos2001mathematics} for details.

\subsection{Petri Net}
A Petri Net is a graph of a finite set of nodes and directed arcs, where nodes are split between places and transitions, and each arc either connects a place to a transition or a transition to a place. Each place has a number of tokens (called the its marking)~\footnote{Standard convention is to use dots in place nodes to represent the marking of the place. We use numbers for compact representation.}. Collective marking of all places in a Petri Net is called its \textit{marking} (or \textit{state}). Arc labels represent arc weights. When missing, arc-weight is assumed as one, and place marking is assumed as zero.

\begin{figure}[htbp]
\centering
\vspace{-30pt}
\includegraphics[width=9cm]{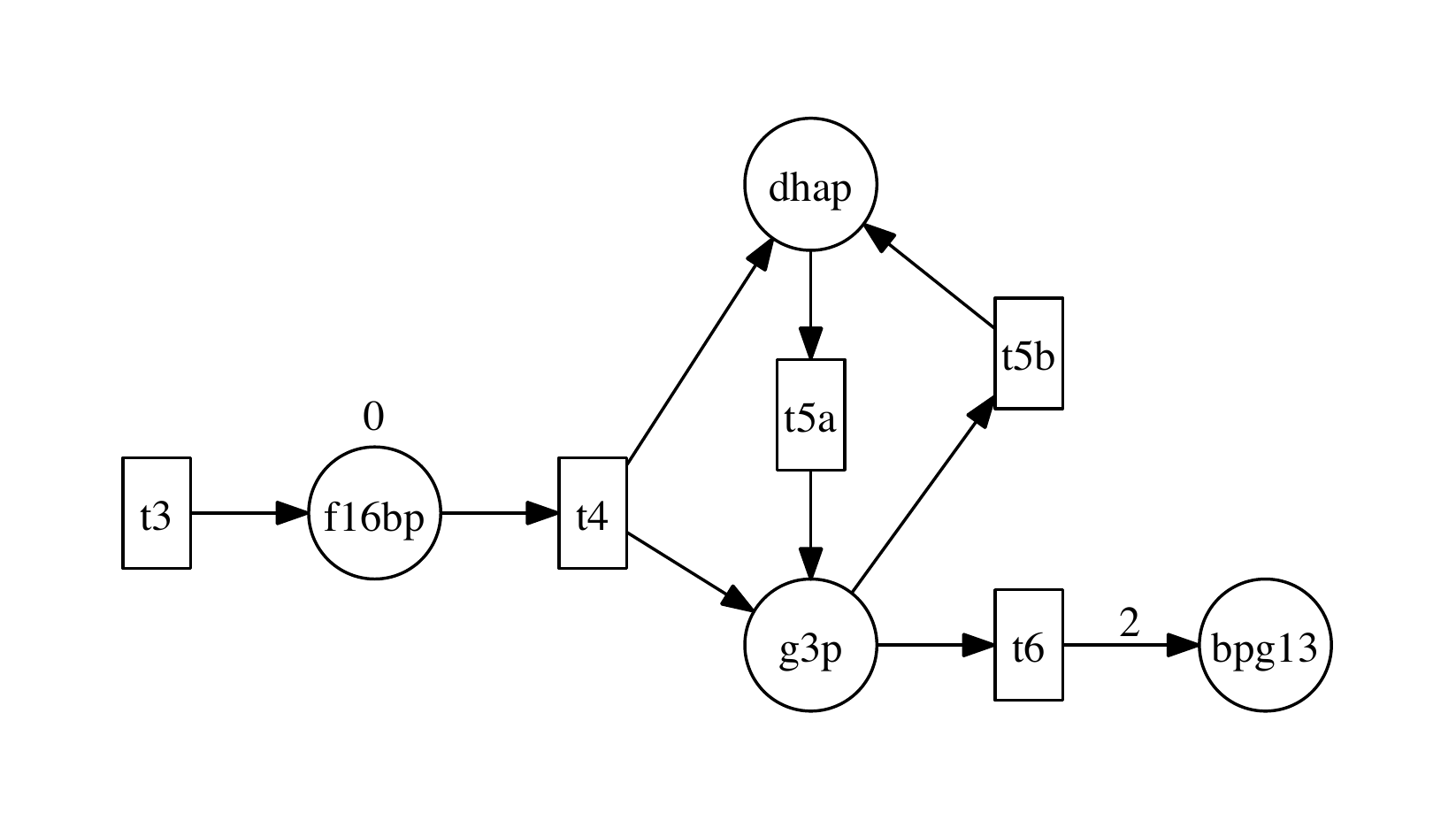}
\caption{Petri Net graph (of sub-section of glycolysis pathway) showing places as circles, transitions as boxes and arcs as directed arrows. Places have token count (or marking) written above them, assumed 0 when missing. Arcs labels represent arc-weights, assumed 1 when missing.}
\label{fig:q1:a}
\end{figure}

The set of place nodes on incoming and outgoing arcs of a transition are called its pre-set (input place set or input-set) and post-set (output place set or output-set), respectively. A transition $t$ is enabled when each of its pre-set place $p$ has at least the number of tokens equal to the arc-weight from $p$ to $t$. An enabled transition may fire, consuming tokens equal to arc-weight from place $p$ to transition $t$ from each pre-set place $p$, producing tokens equal to arc-weight from transition $t$ to place $p$ to each post-set place $p$. 

Multiple transitions may fire as long as they consume no more than the available tokens, with the assumption that tokens cannot be shared. Fig.~\ref{fig:q1:a} shows a representation of a portion of the glycolysis pathway as given in~\cite{CampbellBook}. In this figure, places represent reactants and products, transitions represent reactions, and arc weights represent reactant quantity consumed or the product quantity produced by the reaction. When unspecified, arc-weight is assumed to be $1$ and place-marking is assumed to be $0$.

\begin{definition}[Petri Net]
A Petri Net is a tuple $PN=(P,T,E,W)$, where, 
$P=\{p_1, \dots, p_n\}$ is a finite set of places;
$T=\{t_1, \dots, t_m\}$ is a finite set of transitions, $P \cap T = \emptyset$;
$E^+ \subseteq T \times P$ is a set of arcs from transitions to places;
$E^- \subseteq P \times T$ is a set of arcs from places to transitions;
$E= E^+ \cup E^- $; and
$W: E \rightarrow \N \setminus \{0\}$ is the arc-weight function
\end{definition}

\begin{definition}[Marking]\label{def:pn:marking} A marking $M=(M(p_1),\dots,M(p_{n}))$ is the token assignment of each place node $p_i \in P$ of $PN$, where $M(p_i) \in \N$. Initial token assignment $M_0: P \rightarrow \N$ is called the initial marking. Marking at step $k$ is written as $M_k$.
\end{definition}

\begin{definition}[Pre-set \& post-set of a transition]\label{def:pn:preset}
Pre-set / input-set of a transition $t \in T$ of $PN$ is $\bullet t = \{ p \in P : (p,t) \in E^- \}$, while the post-set / output-set is $t \bullet = \{ p \in P : (t,p) \in E^+ \}$
\end{definition}

\begin{definition}[Enabled Transition]\label{def:pn:enable}
A transition $t \in T$ of $PN$ is enabled with respect to marking $M$, $enabled_M(t)$, if $\forall p \in \bullet t, W(p,t) \leq M(p)$. An enabled transition may fire. 
\end{definition}

\begin{definition}[Transition Execution]\label{def:pn:texec}
A transition execution is the simulation of change of marking from $M_k$ to $M_{k+1}$ due to firing of a transition $t \in T$ of $PN$.
$M_{k+1}$ is computed as follows:
\[ \forall p_i \in \bullet t, M_{k+1}(p_i) = M_{k}(p_i) - W(p_i,t) \]
\vspace{-20pt}
\[ \forall p_j \in t \bullet, M_{k+1}(p_j) = M_{k}(p_j)+ W(t,p_j) \]
\end{definition}

Petri Nets allow simultaneous firing of a set of enabled transitions w.r.t. a marking as long as they do not conflict.

\begin{definition}[Conflicting Transitions]\label{def:pn:conflict}
Given $PN$ with marking $M$. A set of enabled transitions $T_e = \{ t \in T : enabled_M(t) \}$ of $PN$ conflict if their simultaneous firing will consume more tokens than are available at an input place:
\[
\exists p \in P : M(p) < \displaystyle\sum_{\substack{t \in T_e \wedge p \in \bullet t}}{W(p,t)}
\]
\end{definition}

\begin{definition}[Firing Set]\label{def:pn:firing_set}
A firing set is a set $T_k=\{t_{1},\dots,t_{m}\} \subseteq T$ of simultaneously firing transitions that are enabled and do not conflict w.r.t. to the current marking $M_k$ of $PN$. 
\end{definition}

\begin{definition}[Firing Set Execution]\label{def:pn:exec}
Execution of a firing set $T_k$ of $PN$ on a marking $M_{k}$ computes the new marking $M_{k+1}$ as follows:
\[
\forall p \in P, M_{k+1}(p) = M_k(p) 
- \sum_{\substack{t \in T_k \wedge p \in \bullet t}} W(p,t)
+ \sum_{\substack{t \in T_k \wedge p \in t \bullet}} W(t,p)
\]
where $\sum_{\substack{t \in T_k \wedge p \in \bullet t}} W(p,t)$ is the total consumption from place $p$ and $\sum_{\substack{t \in T_k \wedge p \in t \bullet}} W(t,p)$ is the total production at place $p$.
\end{definition}

\begin{definition}[Execution Sequence]\label{def:pn:exec_seq}
An execution sequence $X = M_0, T_0, M_1, T_1, \dots, $ $M_k, T_k, M_{k+1}$ of $PN$ is the simulation of a firing set sequence $\sigma = T_1,T_2,\dots,T_k$ w.r.t. an initial marking $M_0$, producing the final marking $M_{k+1}$. $M_{k+1}$ is the transitive closure of firing set executions, where subsequent marking become the initial marking for the next firing set.
\end{definition}
For an execution sequence $X = M_0, T_0, M_1, T_1, \dots, M_k, T_k, M_{k+1}$, the firing of $T_0$ with respect to marking $M_0$ produces the marking $M_1$ which becomes the initial marking for $T_1$, which produces $M_2$ and so on.
 
\section{Translating Basic Petri Net Into ASP}\label{sec:enc_basic}
In this section we present ASP encoding of simple Petri Nets. We describe, how a given Petri Net $PN$, and an initial marking $M_0$ are encoded into ASP for a simulation length $k$. Following sections will show how Petri Net extensions can be easily added to it.
We represent a {\bf Petri Net} with the following facts:
{\small
\begin{sloppypar}
\begin{description}
\item[\nextf:\label{f:place}\label{f:c:place}] Facts \texttt{\small place($p_i$).} where $p_i \in P$ is a place. 
\item[\nextf:\label{f:trans}\label{f:c:trans}] Facts \texttt{\small trans($t_j$).} where $t_j \in T$ is a transition. 
\item[\nextf:\label{f:ptarc}] Facts \texttt{\small ptarc($p_i,t_j,W(p_i,t_j)$).} where $(p_i,t_j) \in E^-$ with weight $W(p_i,t_j)$. %
\item[\nextf:\label{f:tparc}] Facts \texttt{\small tparc($t_i,p_j,W(t_i,p_j)$).} where $(t_i,p_j) \in E^+$ with weight $W(t_i,p_j)$. %
\end{description}
\end{sloppypar}
}

Petri Net {\bf execution simulation} proceeds in discrete time-steps, these time steps are encoded by the following facts:
{\small
\begin{sloppypar}
\begin{description}
\item[\nextf:\label{f:time}\label{f:c:time}] Facts \texttt{\small time($ts_i$)} where $0 \leq ts_i \leq k$.
\end{description}
\end{sloppypar}
}

The {\bf initial marking} (or initial state) of the Petri Net is represented by the following facts:
{\small
\begin{sloppypar}
\begin{description}
\item[\nexti:\label{i:holds}] Facts \texttt{\small holds($p_i,M_0(p_i),0$)} for every place $p_i \in P$ with initial marking $M_0(p_i)$.
\end{description}
\end{sloppypar}
}

ASP requires all variables in rule bodies be domain restricted. So, we add the following facts to capture the possible token quantities produced during the simulation~\footnote{Note that $ntok$ can be arbitrarily chosen to be larger than the maximum expected token quantity produced during the simulation.}:
{\small
\begin{sloppypar}
\begin{description}
\item[\nextf:\label{f:num}\label{f:c:num}] Facts \texttt{\small num($n$).}, where $0 \leq n \leq ntok$ 
\end{description}
\end{sloppypar}
}

A transition $t_i$ is enabled if each of its input places $p_j \in \bullet t_i$ has at least arc-weight $W(p_j, t_i)$ tokens. Conversely, $t_i$ is not enabled if $\exists p_j \in \bullet t_i : M(p_j) < W(p_j,t_i)$, and is only enabled when no such place $p_j$ exists. These are captured in $e\ref{e:ne:ptarc}$ and $e\ref{e:enabled}$ respectively:

{
\small
\begin{sloppypar}
\begin{description}
\item[\nexte:\label{e:ne:ptarc}] \texttt{\small notenabled(T,TS):-ptarc(P,T,N),holds(P,Q,TS),Q<N, place(P), \\trans(T), time(TS),num(N),num(Q).}
\item[\nexte:\label{e:enabled}] \texttt{\small enabled(T,TS) :- trans(T), time(TS), not notenabled(T, TS).}
\end{description}
\end{sloppypar}
}

Rule $e\ref{e:ne:ptarc}$ encodes \texttt{\small notenabled(T,TS)} which captures the existence of an input place $P$ of transition $T$ that violates the minimum token requirement $N$ at time-step $TS$. Where, the predicate \texttt{\small holds(P,Q,TS)} encodes the marking $Q$ of place $P$ at $TS$.

Rule $e\ref{e:enabled}$ encodes \texttt{\small enabled(T,TS)} which captures that transition $T$ is enabled at $TS$ since there is no input place $P$ of transition $T$ that violates the minimum input token requirement at $TS$. In biological context, $e\ref{e:enabled}$ captures the conditions when a reaction (represented by $T$) is ready to proceed. A subset of enabled transitions may fire simultaneously at a given time-step. This is encoded as:

{\small
\begin{sloppypar}
\begin{description}
\item[\nexta:\label{a:fires}\label{a:c:fires}] \texttt{\small  \{fires(T,TS)\} :- enabled(T,TS), trans(T), time(TS). }
\end{description}
\end{sloppypar}
}

Rule $a\ref{a:fires}$ encodes \texttt{\small fires(T,TS)}, which captures the firing of transition $T$ at $TS$. The rule is encoded with a count atom as its head, which makes it a choice rule. This rule either picks the enabled transition $T$ for firing at $TS$ or not, effectively enumerating a subset of enabled transitions to fire. Whether this set can fire or not in an answer set is subject to conflict checking, which is done by rules $a\ref{a:overc:place},a\ref{a:overc:gen},a\ref{a:overc:elim}$ shown later.

In biological context, the selected transition-set models simultaneously occurring reactions and the conflict models limited reactant supply that cannot be shared. Such a conflict can lead to multiple choices in parallel reaction evolutions and different outcomes. 

The next set of rules captures the consumption and production of tokens due to the firing of individual transitions in a firing-set as well as their aggregate effect, which computes the marking for the next time step:

{\small
\begin{sloppypar}
\begin{description}
\item[\nextr:\label{r:add}] \texttt{\small add(P,Q,T,TS) :- fires(T,TS), tparc(T,P,Q), time(TS).}
\item[\nextr:\label{r:del}] \texttt{\small del(P,Q,T,TS) :- fires(T,TS), ptarc(P,T,Q), time(TS).} 

\item[\nextr:\label{r:totincr}] \texttt{\small tot\_incr(P,QQ,TS) :- 
   QQ=\#sum[add(P,Q,T,TS)=Q:num(Q):trans(T)], 
   time(TS), num(QQ), place(P).}

\item[\nextr:\label{r:totdecr}] \texttt{\small tot\_decr(P,QQ,TS) :- 
   QQ=\#sum[del(P,Q,T,TS)=Q:num(Q):trans(T)], 
   time(TS), num(QQ), place(P).}

\item[\nextr:\label{r:nextstate}] \texttt{\small holds(P,Q,TS+1) :-holds(P,Q1,TS),tot\_incr(P,Q2,TS),time(TS+1),
   tot\_decr(P,Q3,TS),Q=Q1+Q2-Q3,place(P),num(Q;Q1;Q2;Q3),time(TS).}
\end{description}
\end{sloppypar}
}

Rule $r\ref{r:add}$ encodes \texttt{\small add(P,Q,T,TS)} and captures the addition of $Q$ tokens to place $P$ due to firing of transition $T$ at time-step $TS$. Rule $r\ref{r:del}$ encodes \texttt{\small del(P,Q,T,TS)} and captures the deletion of $Q$ tokens from place $P$ due to firing of transition $T$ at $TS$. 

Rules $r\ref{r:totincr}$ and $r\ref{r:totdecr}$ aggregate all \texttt{\small add}'s and \texttt{\small del}'s for place $P$ due to $r\ref{r:add}$ and $r\ref{r:del}$ at time-step $TS$, respectively, by using the \texttt{\small QQ=\#sum[]} construct to sum the $Q$ values into $QQ$.  Rule $r\ref{r:nextstate}$ which encodes \texttt{\small holds(P,Q,TS+1)} uses these aggregate adds and removes and updates $P$'s marking for the next time-step $TS+1$. In biological context, these rules capture the effect of a reaction on reactant and product quantities available in the next simulation step. To prevent overconsumption at a place following rules are added:

{\small
\begin{sloppypar}
\begin{description}
\item[\nexta:\label{a:overc:place}] \texttt{\small consumesmore(P,TS) :- holds(P,Q,TS), tot\_decr(P,Q1,TS), Q1 > Q.} 
\item[\nexta:\label{a:overc:gen}\label{a:c:overc:gen}] \texttt{\small consumesmore :- consumesmore(P,TS).}
\item[\nexta:\label{a:overc:elim}\label{a:c:overc:elim}] \texttt{\small :- consumesmore.}
\end{description}
\end{sloppypar}
}

Rule $a\ref{a:overc:place}$ encodes \texttt{\small consumesmore(P,TS)} which captures overconsumption of tokens at input place $P$ at time $TS$ due to the firing set selected by $a\ref{a:fires}$. Overconsumption (and hence conflict) occurs when tokens $Q1$ consumed by the firing set are greater than the tokens $Q$ available at $P$. Rule $a\ref{a:overc:gen}$ generalizes this notion of overconsumption and constraint $a\ref{a:overc:elim}$ eliminates answers where overconsumption is possible. 

\begin{definition}\label{def:11cor}
Given a Petri Net $PN$, its initial marking $M_0$ and its encoding $\Pi(PN,$ $M_0,$ $k,$ $ntok)$ for $k$-steps and maximum $ntok$ tokens at any place. We say that there is a 1-1 correspondence between the answer sets of $\Pi(PN,M_0,k,ntok)$ and the execution sequences of $PN$ iff for each answer set $A$ of $\Pi(PN,M_0,k,ntok)$, there is a corresponding execution sequence $X=M_0,T_0,M_1,\dots,M_k,T_k,M_{k+1}$ of $PN$ and for each execution sequence $X$ of $PN$ there is an answer-set $A$ of $\Pi(PN,M_0,k,ntok)$ such that 
\begin{equation*}
\{ fires(t,ts) : t \in T_{ts}, 0 \leq ts \leq k \} = \{ fires(t,ts) : fires(t,ts) \in A \} 
\end{equation*}
\begin{equation*}
\begin{split}
\{ holds(p,q,ts) &: p \in P, q=M_{ts}(p), 0 \leq ts \leq k+1 \} \\
&= \{ holds(p,q,ts) : holds(p,q,ts) \in A\}
\end{split}
\end{equation*}
\end{definition}

\begin{proposition}\label{prop:basic_enc}
There is a 1-1 correspondence between the answer sets of $\Pi^0(PN,$ $M_0,$ $k,$ $ntok)$ and the execution sequences of $PN$.

\end{proposition}

\subsection{An example execution}\label{sec:basic_pn_exec}

Next we look at an example execution of the Petri Net shown in Figure~\ref{fig:q1:a}. The Petri Net and its initial marking are encoded as follows\footnote{\texttt{\{holds(p1,0,0),\dots,holds(pN,0,0)\}}, \texttt{\{num(0),\dots,num(60)\}}, \texttt{\{time(0),\dots,time(5)\}} have been written as \texttt{holds(p1;\dots;pN,0,0)}, \texttt{num(0..60)}, \texttt{time(0..5)},  respectively, to save space.}:
{\small
\begin{verbatim}
num(0..60).time(0..5).place(f16bp;dhap;g3p;bpg13).
trans(t3;t4;t5a;t5b;t6).tparc(t3,f16bp,1).ptarc(f16bp,t4,1).
tparc(t4,dhap,1).tparc(t4,g3p,1).ptarc(dhap,t5a,1).
tparc(t5a,g3p,1).ptarc(g3p,t5b,1).tparc(t5b,dhap,1).
ptarc(g3p,t6,1).tparc(t6,bpg13,2).holds(f16bp;dhap;g3p;bgp13,0,0).
\end{verbatim}
}
we get thousands of answer-sets, for example\footnote{\texttt{\{fires(t1,ts1),\dots,fires(tN,ts1)\}} have been written as \texttt{fires(t1;\dots;tN;ts1)} to save space.}: %
{\small
\begin{verbatim}
holds(bpg13,0,0) holds(dhap,0,0) holds(f16bp,0,0) holds(g3p,0,0) 
holds(bpg13,0,1) holds(dhap,0,1) holds(f16bp,1,1) holds(g3p,0,1) 
holds(bpg13,0,2) holds(dhap,1,2) holds(f16bp,1,2) holds(g3p,1,2) 
holds(bpg13,0,3) holds(dhap,2,3) holds(f16bp,1,3) holds(g3p,2,3) 
holds(bpg13,2,4) holds(dhap,3,4) holds(f16bp,1,4) holds(g3p,2,4) 
holds(bpg13,4,5) holds(dhap,4,5) holds(f16bp,1,5) holds(g3p,2,5) 
fires(t3,0) fires(t3;t4,1) fires(t3;t4;t5a;t5b,2) 
fires(t3;t4;t5a;t5b;t6,3) fires(t3;t4;t5a;t5b;t6,4) 
fires(t3;t4;t5a;t5b;t6,5)
\end{verbatim}
}

\section{Changing Firing Semantics}\label{sec:enc_max}
The ASP code above implements the \textit{set firing} semantics. It can produce a large number of answer-sets, since any subset of a firing set will also be fired as a firing set. For our biological system modeling, it is often beneficial to simulate only the maximum activity at any given time-step. We accomplish this by defining the \textit{maximal firing set} semantics, which requires that a maximal subset of non-conflicting transitions fires at a single time step\footnote{Such a semantics reduces the reachable markings. See \cite{Burkhard1980} for the analysis of its computational power.}. Our semantics is different from the firing multiplier approach used by \cite{PetriNetMaxParallelism}, in which a transition can fire as many times as allowed by the tokens available in its source places. Their approach requires an exponential time firing algorithm in the number of transitions. Our maximal firing set semantics is implemented by adding the following rules to the encoding in Section~\ref{sec:enc_basic}:

{\small
\begin{sloppypar}
\begin{description}
\item[\nexta:\label{a:maxfire:cnh}] \texttt{ could\_not\_have(T,TS) :- enabled(T,TS), not fires(T,TS), 
   ptarc(S,T,Q), holds(S,QQ,TS), tot\_decr(S,QQQ,TS), Q > QQ - QQQ.}
\item[\nexta:\label{a:maxfire:elim}\label{a:c:maxfire:elim}] \texttt{:- not could\_not\_have(T,TS), enabled(T,TS),
   not fires(T,TS), trans(T), time(TS).}
\end{description}
\end{sloppypar}
}

Rule $a\ref{a:maxfire:cnh}$ encodes \texttt{\small could\_not\_have(T,TS)} which means that an enabled transition $T$ that did not fire at time $TS$, could not have fired because its firing would have resulted in overconsumption. Rule $a\ref{a:maxfire:elim}$ eliminates any answer-sets in which an enabled transition did not fire, that could not have caused overconsumption. Intuitively, these two rules guarantee that the only reason for an enabled transition to not fire is conflict avoidance (due to overconsumption). With this firing semantics, the number of answer-sets produced for Petri Net in Figure~\ref{fig:q1:a} reduces to 2.

\begin{proposition}\label{prop:max}
There is 1-1 correspondence between the answer sets of $\Pi^1(PN,$ $M_0,$ $k,$ $ntok)$ and the execution sequences of $PN$.
\end{proposition}

Other firing semantics can be encoded with similar ease. For example, if \textit{interleaved} firing semantics is desired, replace rules $a\ref{a:maxfire:cnh},a\ref{a:maxfire:elim}$ with the following:
{\small
\begin{sloppypar}
\begin{description}
\item[a\ref{a:maxfire:cnh}':] \texttt{more\_than\_one\_fires :- fires(T1,TS), fires(T2, TS), T1 != T2, time(TS).}
\item[a\ref{a:maxfire:elim}':] \texttt{:- more\_than\_one\_fires.}
\end{description}
\end{sloppypar}
}
We now look at Petri Net extensions and show how they can be easily encoded in ASP.

\section{Extension - Reset Arcs}\label{sec:enc_reset}

\begin{definition}[Reset Arc]
A Reset Arc in a Petri Net $PN^R$ is an arc from place $p$ to transition $t$ that consumes all tokens from its input place $p$ upon firing of $t$. A Reset Petri Net is a tuple $PN^R = (P,T,E,W,R)$ where, $P, T, E, W$ are the same as for PN; and $R: T \rightarrow 2^P$ defines reset arcs
\end{definition}

\begin{figure}[htbp]
\centering
\vspace{-30pt}
\includegraphics[width=9cm]{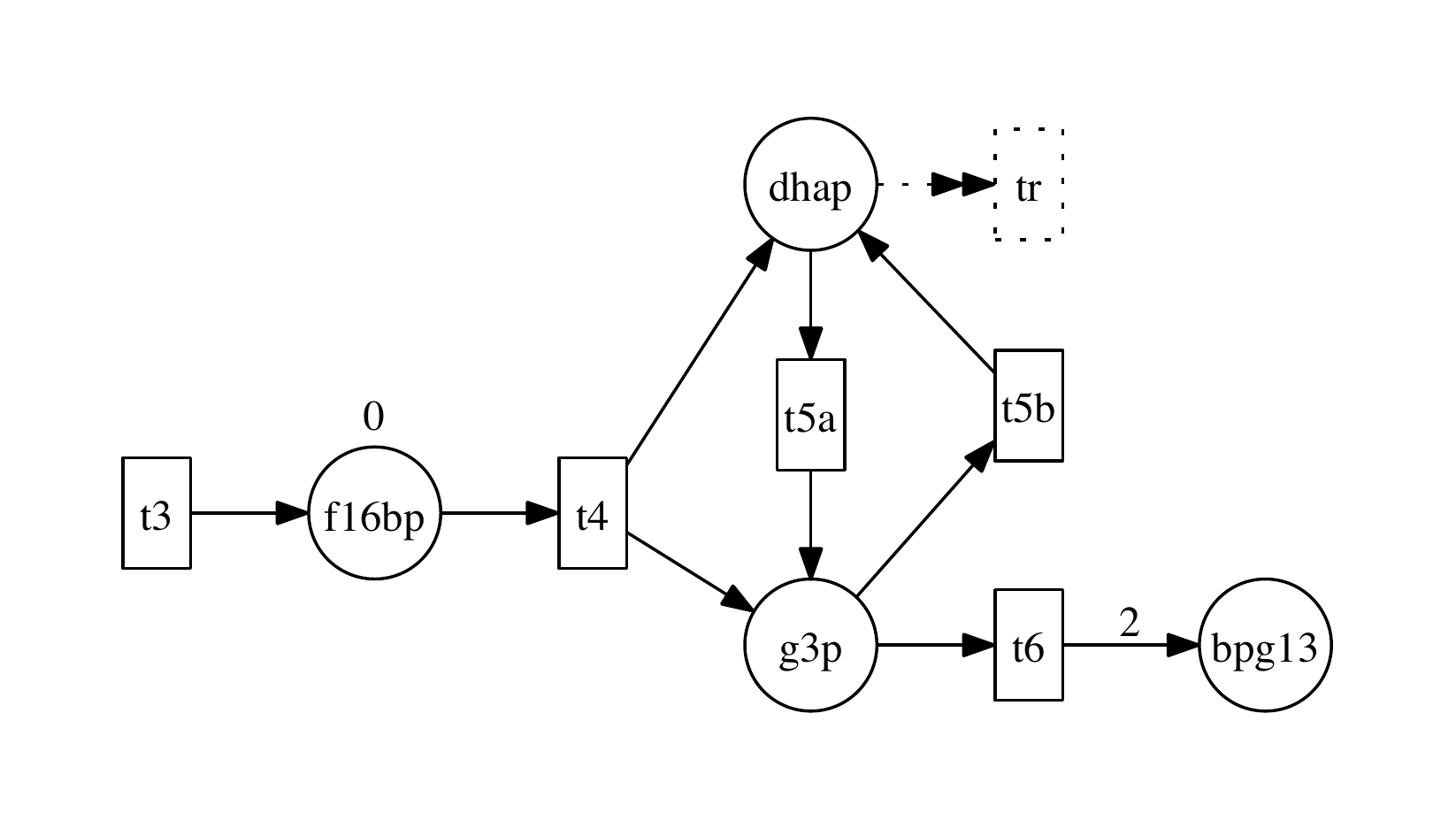}
\caption{Petri Net of Fig~\ref{fig:q1:a} extended with a reset arc from $dhap$ to $tr$ shown with double arrowhead.}
\label{fig:q1:b}
\end{figure}

Figure~\ref{fig:q1:b} shows an extended version of the Petri Net in Figure~\ref{fig:q1:a} with a reset arc from $dhap$ to $tr$ (shown with double arrowhead). In biological context it models the removal of all quantity of compound $dhap$. Petri Net execution semantics with reset arcs is modified for conflict detection and execution as follows:

\begin{definition}[Reset Transition]\label{def:pnr:reset_trans}
A transition $t \in T$ of $PN^R$ is called a reset-transition if it has a reset arc incident on it, i.e. $R(t) \neq \emptyset$.
\end{definition}

\begin{definition}[Firing Set]\label{def:pnr:firing_set}\label{def:pnri:firing_set}\label{def:pnriq:firing_set}
A firing set is a set $T_k=\{t_{1},\dots,t_{m}\} \subseteq T$ of simultaneously firing transitions that are enabled and do not conflict w.r.t. to the current marking $M_k$ of $PN^R$. $T_k$ is not a firing set if there is an enabled reset-transition that is not in $T_k$, i.e. $\exists t : enabled_{M_k}(t), R(t) \neq \emptyset, t \not\in T_k$.~\footnote{\label{fn:rptarc:conflict}The reset arc is involved here because we use a modified execution semantics of reset arcs compared to the standard definition~\cite{araki1976some}. Even though both capture similar operation, our definition allows us to model elimination of all quantity of a substance as soon as it is produced, even in a maximal firing set semantics. Our semantics considers reset arc's token consumption in contention with other arcs, while the standard definition does not.}.
\end{definition}

\begin{definition}[Transition Execution in $PN^R$]\label{def:pnr:texec}\label{def:pnri:texec}\label{def:pnriq:texec}
A transition execution is the simulation of change of marking from $M_k$ to $M_{k+1}$ due to firing of a transition $t \in T$ of $PN^R$.
$M_{k+1}$ is computed as follows:
\[ \forall p_i \in \bullet t, M_{k+1}(p_i) = M_{k}(p_i) - W(p_i,t) \]
\vspace{-20pt}
\[ \forall p_j \in t \bullet, M_{k+1}(p_j) = M_{k}(p_j) + W(t,p_j) \]
\vspace{-20pt}
\[ \forall p_r \in R(t), M_{k+1}(p_r) = M_{k}(p_r) - M_{k}(p_r) \]
\end{definition}

\begin{definition}[Conflicting Transitions in $PN^R$]\label{def:pnr:conflict}\label{def:pnri:conflict}\label{def:pnriq:conflict}
A set of enabled transitions conflict in $PN^R$ w.r.t. $M_k$ if firing them simultaneously will consume more tokens than are available at any one of their common input-places. $T_e = \{ t \in T : enabled_{M_k}(t) \}$ conflict if:
\[
\exists p \in P : M_k(p) < (\displaystyle\sum_{t \in T_e \wedge (p,t) \in E^-}{W(p,t)} + \displaystyle\sum_{t \in T_e \wedge p \in R(t)}{M_k(p)})
\]
\end{definition}

\begin{definition}[Firing Set Execution in $PN^R$]\label{def:pnr:exec}\label{def:pnri:exec}\label{def:pnriq:exec}
Execution of a transition set $T_i$ in $PN^R$ has the following effect:
\[
\forall p \in P \setminus R(T_i), M_{k+1}(p) = M_k(p) -  \sum_{\substack{t \in T_i \wedge p \in \bullet t}} W(p,t) + \sum_{\substack{t \in T_i \wedge p \in t \bullet}} W(t,p)
\]
\[
\forall p \in R(T_i), M_{k+1}(p) = \sum_{t \in T_i \wedge p \in t \bullet} W(t,p)
\]
where $R(T_i)=\displaystyle\bigcup_{\substack{t \in T_i}} R(t)$ and represents the places emptied by $T_i$ due to reset arcs~\footnote{\label{fn:rptarc:exec}Our definition of conflicting transitions allows at most one transition with a reset arc from a place to fire, any more create a conflict. Thus, the new marking computation is equivalent to $\forall p \in P, M_{k+1}(p) = M_k(p) -  (\sum_{\substack{t \in T_k  \wedge p \in \bullet t}}{W(p,t)} + \sum_{\substack{t \in T_k \wedge p \in R(t)}}{M_k(p)})+ \sum_{\substack{t \in T_k  \wedge p \in t \bullet}} W(t,p)$}.
\end{definition}

Since a reset arc from $p$ to $t$, $p \in R(t)$ consumes current marking dependent tokens, we extend \texttt{\small ptarc} to include time-step and replace $f\ref{f:ptarc},f\ref{f:tparc},e\ref{e:ne:ptarc},r\ref{r:add},r2,a\ref{a:maxfire:cnh}$ with $f\ref{f:r:ptarc},f\ref{f:r:tparc},e\ref{e:r:ne:ptarc},r\ref{r:r:add},r\ref{r:r:del},a\ref{a:r:maxfire:cnh}$, respectively in the Section~\ref{sec:enc_max} encoding and add rule $f\ref{f:rptarc}$ for each reset arc:
{\small
\begin{sloppypar}
\begin{description}
\item[\nextf:\label{f:r:ptarc}] Rules \texttt{\small ptarc($p_i,t_j,W(p_i,t_j),ts_k$):-time($ts_k$).} for each non-reset arc $(p_i,t_j) \in E^-$
\item[\nextf:\label{f:r:tparc}] Rules \texttt{\small tparc($t_i,p_j,W(t_i,p_j),ts_k$):-time($ts_k$).} for each non-reset arc $(t_i,p_j) \in E^+$
\item[\nexte:\label{e:r:ne:ptarc}] \texttt{\small notenabled(T,TS) :- ptarc(P,T,N,TS), holds(P,Q,TS), Q < N,\\
   place(P), trans(T), time(TS), num(N), num(Q).}
\item[\nextr:\label{r:r:add}] \texttt{\small add(P,Q,T,TS) :- fires(T,TS), tparc(T,P,Q,TS), time(TS).}
\item[\nextr:\label{r:r:del}] \texttt{\small del(P,Q,T,TS) :- fires(T,TS), ptarc(P,T,Q,TS), time(TS).}
\item[\nextf:\label{f:rptarc}] Rules \texttt{\small ptarc($p_i,t_j,X,ts_k$) :- holds($p_i,X,ts_k$), num($X$), $X>0$.} for each reset arc between $p_i$ and $t_j$ using $X=M_k(p_i)$ as arc-weight at time step $ts_k$.
\item[\nextf:\label{f:rptarc:elim}\label{f:c:rptarc:elim}] Rules \texttt{\small :- enabled($t_j,ts_k$),not fires($t_j,ts_k$), time($ts_k$).} for each transition $t_j$ with an incoming reset arc, i.e. $R(t_j) \neq \emptyset$.
\item[\nexta:\label{a:r:maxfire:cnh}] \texttt{\small could\_not\_have(T,TS) :- enabled(T,TS), not fires(T,TS), \\ ptarc(S,T,Q,TS), holds(S,QQ,TS), tot\_decr(S,QQQ,TS), Q>QQ-QQQ.}
\end{description}
\end{sloppypar}
}

Rule $f\ref{f:rptarc}$ encodes place-transition arc with marking dependent weight to capture the notion of a reset arc, while rule $f\ref{f:rptarc:elim}$ ensures that the reset-transition (i.e. the transition on which the reset arc terminates) always fires when enabled.

\begin{proposition}\label{prop:reset}
There is 1-1 correspondence between the answer sets of $\Pi^2(PN^R,M_0,$ $k,ntok)$ and the execution sequences of $PN^R$.
\end{proposition}

The execution semantics of our definition are slightly different from the standard definition in~\cite{araki1976some}, even though both capture similar operations. Our implementation considers token consumption by reset arc in contention with other token consuming arcs from the same place, while the standard definition considers token consumption as a side effect, not in contention with other arcs.

We chose our definition to allow modeling of biological process that removes all available quantity of a substance in a maximal firing set. Consider Figure~\ref{fig:q1:b}, if $dhap$ has $1$ or more tokens, our semantics would only permit either $t5a$ or $tr$ to fire in a single time-step, while the standard semantics can allow both $t5a$ and $tr$ to fire simultaneously, such that the reset arc removes left over tokens after $(dhap,t5a)$ consumes one token. 

We could have, instead, extended our encoding to include self-modifying nets~\cite{SelfModNets}, but our modified-definition provides a simpler solution. Standard semantics, however, can be easily encoded by replacing $r\ref{r:nextstate}$ by $r\ref{r:nextstate}a', r\ref{r:nextstate}b'$; replacing $f\ref{f:rptarc},f\ref{f:rptarc:elim}$ with $f\ref{f:rptarc}'$; and adding $a\ref{a:reset:std}$ as follows:
{\small
\begin{sloppypar}
\begin{description}
\item[f\ref{f:rptarc}':] \texttt{rptarc($p_i$,$t_j$).} - for each reset arc between $p_i \in R(t_j)$ and $t_j$.
\item[\nexta:\label{a:reset:std}] \texttt{reset(P,TS) :- rptarc(P,T), place(P), trans(T), fires(T,TS), time(TS).}
\item[r\ref{r:nextstate}a':] \texttt{holds(P,Q,TS+1) :- holds(P,Q1,TS), tot\_incr(P,Q2,TS), tot\_decr(P,Q3,TS), Q=Q1+Q2-Q3, place(P),  num(Q;Q1;Q2;Q3), time(TS), time(TS+1), not reset(P,TS).}
\item[r\ref{r:nextstate}b':] \texttt{holds(P,Q,TS+1) :- tot\_incr(P,Q,TS), place(P),  num(Q), time(TS), time(TS+1), reset(P,TS).}
\end{description}
\end{sloppypar}
}

where, the fact $f\ref{f:rptarc}'$ encodes the reset arc; rule $a\ref{a:reset:std}$ encodes if place $P$ will be reset at time $TS$ due to firing of transition $T$ that has a reset arc on it from $P$ to $T$; rule $r\ref{r:nextstate}a'$ computes marking at $TS+1$ when place $P$ is not being reset; and rule $r\ref{r:nextstate}b'$ computes marking at $TS+1$ when $P$ is being reset.

\section{Extension - Inhibitor Arcs}\label{sec:enc_inhibit}
\begin{definition}[Inhibitor Arc]
An inhibitor arc~\cite{PetersonPetriNets} is a place-transition arc that inhibits its transition from firing as long as the place has any tokens in it. An inhibitor arc does not consume any tokens from its input place. A Petri Net with reset and inhibitor arcs is a tuple $PN^{RI}=(P,T,E,W,R,I)$, where, $P, T, E, W, R$ are the same as for $PN^R$; and $I: T \rightarrow 2^P$ defines inhibitor arcs.
\end{definition}

\begin{figure}
\centering
\vspace{-30pt}
\includegraphics[width=6cm]{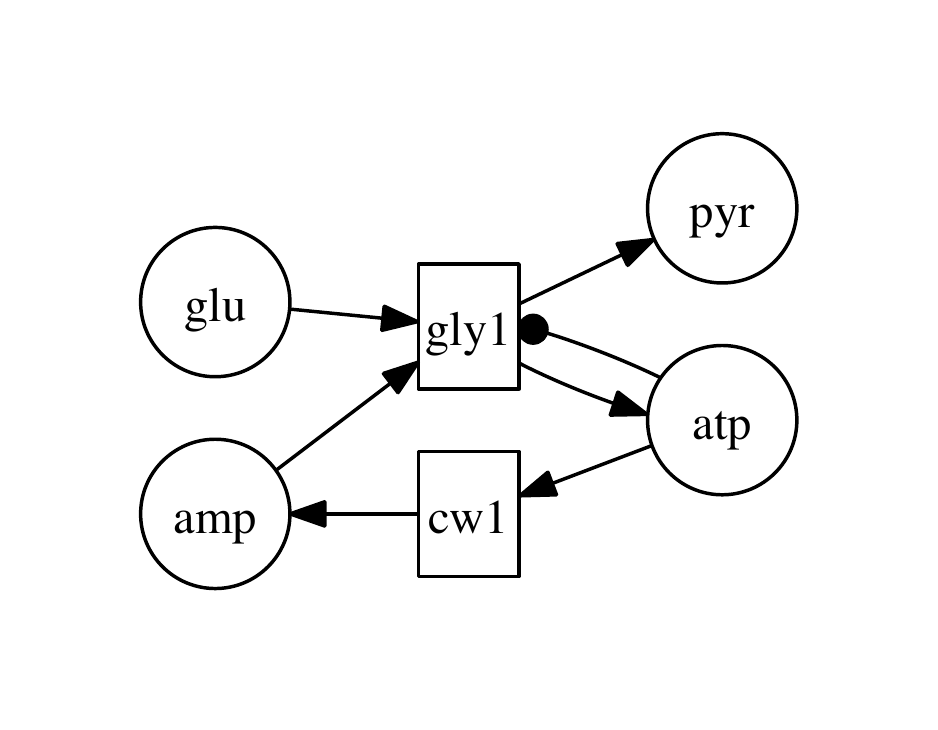}
\caption{Petri Net showing feedback inhibition arc from $atp$ to $gly1$ with a bullet arrowhead. Inhibitor arc weight is assumed $1$ when not specified.} 
\label{fig:q:q3}
\end{figure}

Figure~\ref{fig:q:q3} shows a Petri Net with inhibition arc from $atp$ to $gly1$ with a bulleted arrowhead. It models biological feedback regulation in simplistic terms, where excess $atp$ downstream causes the upstream $atp$ production by glycolysis $gly$ to be inhibited until the excess quantity is consumed~\cite{CampbellBook}. Petri Net execution semantics with inhibit arcs is modified for determining enabled transitions as follows:

\begin{definition}[Enabled Transition in $PN^{RI}$]\label{def:pnri:enable}
A transition $t$ is enabled with respect to marking $M$, $enabled_M(t)$, if all its input places $p$ have at least the number of tokens as the arc-weight $W(p,t)$ and all $p \in I(t)$ have zero tokens, i.e. $(\forall p \in \bullet t, W(p,t) \leq M(p)) \wedge (\forall p \in I(t), M(p) = 0)$ 
\end{definition}

We add inhibitor arcs to our encoding in Section~\ref{sec:enc_reset} as follows:

{\small
\begin{sloppypar}
\begin{description}
\item[\nextf:\label{f:iptarc}] Rules \texttt{\small iptarc($p_i,t_j,1,ts_k$):-time($ts_k$).} for each inhibitor arc between $p_i \in I(t_j)$ and $t_j$.
\item[\nexte:\label{e:ne:iptarc}] \texttt{\small notenabled(T,TS) :- iptarc(P,T,N,TS), holds(P,Q,TS),
   place(P), \\trans(T), time(TS), num(N), num(Q), Q >= N.}
\end{description}
\end{sloppypar}
}

The new rule $e\ref{e:ne:iptarc}$ encodes another reason for a transition to be disabled (or not enabled). An inhibitor arc from $p$ to $t$ with arc weight $N$ will cause its target transition $t$ to not enable when the number of tokens at its source place $p$ is greater than or equal to $N$, where $N$ is always $1$ per rule $f\ref{f:iptarc}$. 

\begin{proposition}\label{prop:inhibit}
There is 1-1 correspondence between the answer sets of $\Pi^3(PN^{RI},M_0,k,ntok)$ and the execution sequences of $PN$.
\end{proposition}

\section{Extension - Read Arcs}\label{sec:enc_query}
\begin{definition}[Read Arc]
A read arc (a test arc or a query arc)~\cite{christensen1993coloured} is an arc from place to transition, which enables its transition only when its source place has at least the number of tokens as its arc weight. It does not consume any tokens from its input place. A Petri Net with reset, inhibitor and read arcs is a tuple $PN^{RIQ}=(P,T,W,R,I,Q,QW)$, where, $P,T,E,W,R,I$ are the same as for $PN^{RI}$; $Q \subseteq P \times T $ defines read arcs; and $QW: Q \rightarrow \mathds{N} \setminus \{0\}$ defines read arc weight.
\end{definition}

\begin{figure}[htbp]
\centering
\vspace{-25pt}
\includegraphics[width=6cm]{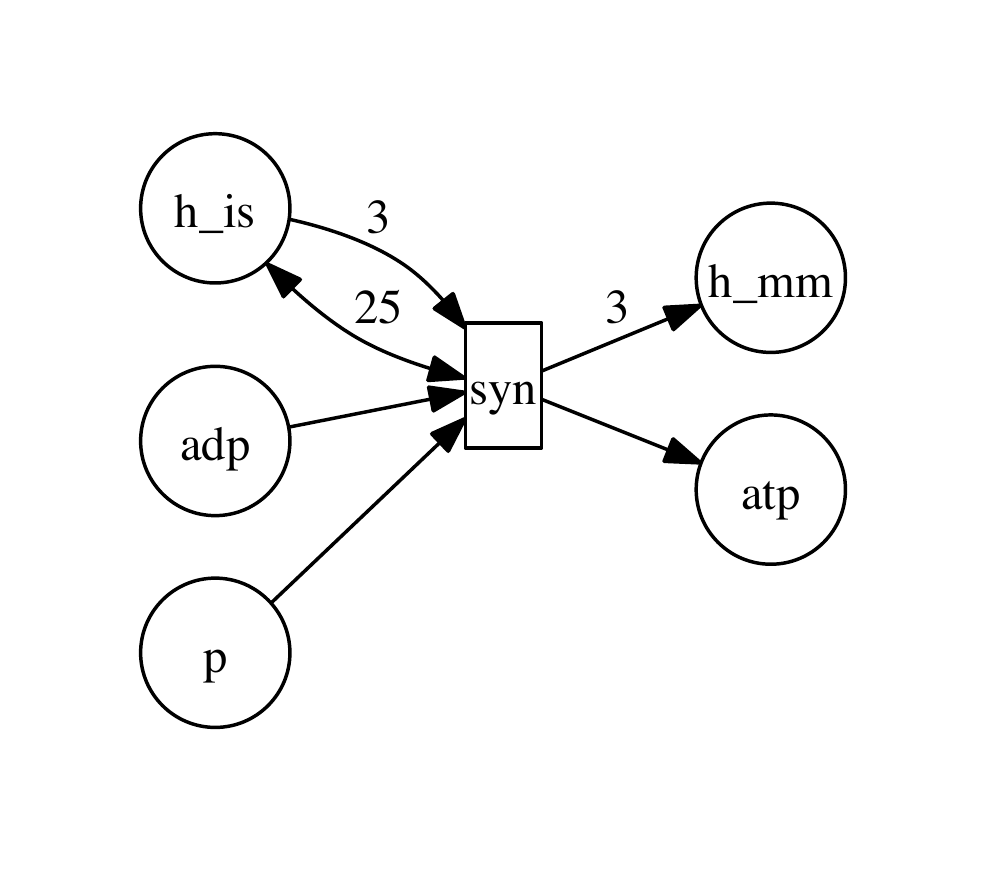}
\caption{Petri Net with read arc from $h\_is$ to $syn$ shown with arrowhead on both ends. The transition $syn$ will not fire unless there are at least $25$ tokens in $h\_is$, but when it executes, it only consumes $3$ tokens.}
\label{fig:atp_synth}
\end{figure}

Figure~\ref{fig:atp_synth} shows a Petri Net with read arc from $h\_is$ to $syn$ shown with arrowhead on both ends. It models the ATP synthase $syn$ activation requiring a higher concentration of $H+$ ions  $h\_is$ in the intermembrane space~\footnote{This is an oversimplified model of $syn$ (ATP synthase) activation, since the actual model requires an $H+$ concentration differential across membrane.}. The reaction itself consumes a lower quantity of $H+$ ions represented by the regular place-transition arc~\cite{CampbellBook,berg2002proton}. 
Petri Net execution semantics with read arcs is modified for determining enabled transitions as follows:

\begin{definition}[Enabled Transition in $PN^{RIQ}$]\label{def:pnriq:enable}
A transition $t$ is enabled with respect to marking $M$, $enabled_M(t)$, if all its input places $p$ have at least the number of tokens as the arc-weight $W(p,t)$, all $p_i \in I(t)$ have zero tokens and all $p_q : (p_q,t) \in Q$ have at least the number of tokens as the arc-weight $W(p,t)$, i.e. $(\forall p \in \bullet t, W(p,t) \leq M(p)) \wedge (\forall p \in I(t), M(p) = 0) \wedge (\forall (p,t) \in Q, M(p) \geq QW(p,t))$
\end{definition}

We add read arcs to our encoding of Section~\ref{sec:enc_inhibit} as follows:
{\small
\begin{sloppypar}
\begin{description}
\item[\nextf:\label{f:tptarc}] Rules \texttt{\small tptarc($p_i,t_j,QW(p_i,t_j),ts_k$):-time($ts_k$).} for each read arc $(p_i,t_j) \in Q$.
\item[\nexte:\label{e:ne:tptarc}] \texttt{\small notenabled(T,TS):-tptarc(P,T,N,TS),holds(P,Q,TS),\\
    place(P),trans(T), time(TS), num(N), num(Q), Q < N.}
\end{description}
\end{sloppypar}
}

The new rule $f\ref{f:tptarc}$ captures the read arc and its arc-weight; and the new rule $e\ref{e:ne:tptarc}$ encodes another reason for a transition to not be enabled. A read arc from $p$ to $t$ with arc weight $N$ will cause its target transition $t$ to not enable when the number of tokens at its source place $p$ is less than the arc weight $N$.

\begin{proposition}\label{prop:query}
There is a 1-1 correspondence between the answer sets of $\Pi^4(PN^{RIQ},M_0,$ $k,ntok)$ and the execution sequences of $PN^{RIQ}$.
\end{proposition} 

\section{Extension - Colored Tokens}\label{sec:pnc}
Higher level Petri Nets extend the notion of tokens to typed (or colored) tokens. This allows a more compact representation of complicated networks~\cite{peterson1980note}.
\begin{definition}[Petri Net with Colored Tokens]\label{def:pnc}
A Petri Net with Colored Tokens (with reset, inhibit and read arcs) is a tuple $PN^C=(P,T,E,C,W,R,I,Q,QW)$, where $P,T,E,R,I,Q$ are the same as for basic Petri Nets, $C=\{c_1,\dots,c_l\}$ is a finite set of colors (or types), and arc weights $W : E \rightarrow \langle C,m \rangle$, $QW : Q \rightarrow \langle C,m \rangle$ are specified as multi-sets of colored tokens over color set $C$. The state (or marking) of place nodes $M(p_i) = \langle C,m \rangle, p_i \in P$ is specified as a multiset of colored tokens over set $C$.
\end{definition}

We will now update some definitions related to Petri Nets to include colored tokens.
\begin{definition}[Marking]\label{def:pnc:marking}
A marking $M=(M(p_1),\dots,M(p_n))$ assigns a colored multi-set of tokens over the domain of colors $C$ to each place $\{p_1,\dots,p_n\} \in P$ of $PN^C$. The initial marking is the initial token assignment of place nodes and is represented by $M_0$. The marking at time-step $k$ is written as $M_k$. 
\end{definition}

\begin{definition}[Pre-set and post-set of a transition]\label{def:pnc:preset}\label{def:pnc:postset}\label{def:pnpri:preset}\label{def:pnpri:postset}\label{def:pndur:preset}\label{def:pndur:postset}
The pre-set (or input-set) of a transition $t$ is $\bullet t = \{ p \in P | (p,t) \in E^- \}$, while the post-set (or output-set) is $t \bullet = \{ p \in P | (t,p) \in E^+ \}$.
\end{definition}

\begin{definition}[Enabled Transition]\label{def:pnc:enable}
A transition $t$ is enabled with respect to marking $M$, $enabled_M(t)$, if each of its input places $p$ has at least the number of colored-tokens as the arc-weight $W(p,t)$\footnote{In the following text, for simplicity, we will use $W(p,t)$ to mean $W(\langle p,t \rangle)$. We use similar simpler notation for $QW$.}, each of its inhibiting places $p_i \in I(t)$ have zero tokens and each of its read places $p_q : (p_q,t) \in Q$ have at least the number of colored-tokens as the read-arc-weight $QW(p_q,t)$, i.e. $(\forall p \in \bullet t, W(p,t) \leq M(p)) \wedge (\forall p \in I(t), M(p) = \emptyset) \wedge (\forall (p,t) \in Q, M(p) \geq QW(p,t))$ for a given $t$.\footnote{This is equivalent to $\forall c \in C, (\forall p \in \bullet t, m_{W(p,t)}(c) \leq m_{M(p)}(c)) \wedge (\forall p \in I(t), m_{M(p)}(c) = 0) \wedge (\forall (p,t) \in Q, m_{M(p)}(c) \geq m_{QW(p,t)}(c))$.}
\end{definition}

\begin{definition}[Transition Execution]\label{def:pnc:texec}\label{def:pnpri:texec}
Execution of a transition $t$ of $PN^C$ on a marking $M_k$ computes a new marking $M_{k+1}$ as: 
\[
\forall p \in \bullet t M_{k+1}(p) = M_k(p) -  W(p,t) 
\]
\[
\forall p \in t\bullet M_{k+1}(p) = M_k(p) + W(t,p)
\]
\[
\forall p \in R(t)  M_{k+1}(p) = M_k(p) - M_k(p)
\] 
\end{definition}

Any number of enabled transitions may fire simultaneously as long as they don't conflict. A transition when fired consumed tokens from its pre-set places equivalent to the (place,transition) arc-weight.

\begin{definition}[Conflicting Transitions]\label{def:pnc:conflict}\label{def:pnpri:conflict}\label{def:pndur:conflict}
A set of transitions $T_c \subseteq \{ t : $ $enabled_{M_k}(t) \}$ is in conflict in $PN^C$ with respect to $M_k$ if firing them will consume more tokens than are available at one of their common input places, i.e., 
$\exists p \in P : M_k(p) < (\sum_{t \in T_c \wedge p \in \bullet t}{W(p,t)} + \sum_{\substack{t \in T_c  \wedge p \in R(t)}}{M_k(p)})$
\end{definition}

\begin{definition}[Firing Set]\label{def:pnc:firing_set}
A firing set is a set $T_k=\{t_{k_1},\dots,t_{k_n}\} \subseteq T$ of simultaneously firing transitions of $PN^C$ that are enabled and do not conflict w.r.t. to the current marking $M_k$ of $PN$. A set $T_k$ is not a firing set if there is an enabled reset-transition that is not in $T_k$, i.e. $\exists t \in enabled_{M_k}, R(t) \neq \emptyset, t \not\in T_k$.~\footnote{See footnote~\ref{fn:rptarc:conflict}} %
\end{definition}

\begin{definition}[Firing Set Execution]\label{def:pnc:exec}\label{def:pnpri:exec}
Execution of a firing set $T_k$ of $PN^C$ on a marking $M_k$ computes a new marking $M_{k+1}$ as: 
\[
\forall p \in P \setminus R(T_k), M_{k+1}(p) = M_k(p) 
-  \sum_{\substack{t \in T_k  \wedge p \in \bullet t}} W(p,t) 
+ \sum_{\substack{t \in T_k  \wedge p \in t \bullet}} W(t,p)
\] 
\[
\forall p \in R(T_k), M_{k+1}(p) = \sum_{\substack{t \in T_k \wedge p \in t \bullet}} W(t,p)
\] 
where $R(T_k)=\bigcup_{t \in T_k} R(t)$~\footnote{See footnote~\ref{fn:rptarc:exec}} %
\end{definition}

\begin{definition}[Execution Sequence]\label{def:pnc:execseq} 
An execution sequence $X = M_0, T_0, M_1, $ $T_1, \dots, $ $M_k, $ $T_k, M_{k+1}$ of $PN$ is the simulation of a firing set sequence $\sigma = T_1,T_2,\dots,T_k$ w.r.t. an initial marking $M_0$, producing the final marking $M_{k+1}$. $M_{k+1}$ is the transitive closure of firing set executions, where subsequent marking become the initial marking for the next firing set.
\end{definition}
For an execution sequence $X = M_0, T_0, M_1, T_1, \dots, $ $M_k, T_k, M_{k+1}$, the firing of $T_0$ with respect to marking $M_0$ produces the marking $M_1$ which becomes the initial marking for $T_1$, which produces $M_2$ and so on.

\begin{figure}[htbp]
\centering
\vspace{-30pt}
\includegraphics[width=\linewidth]{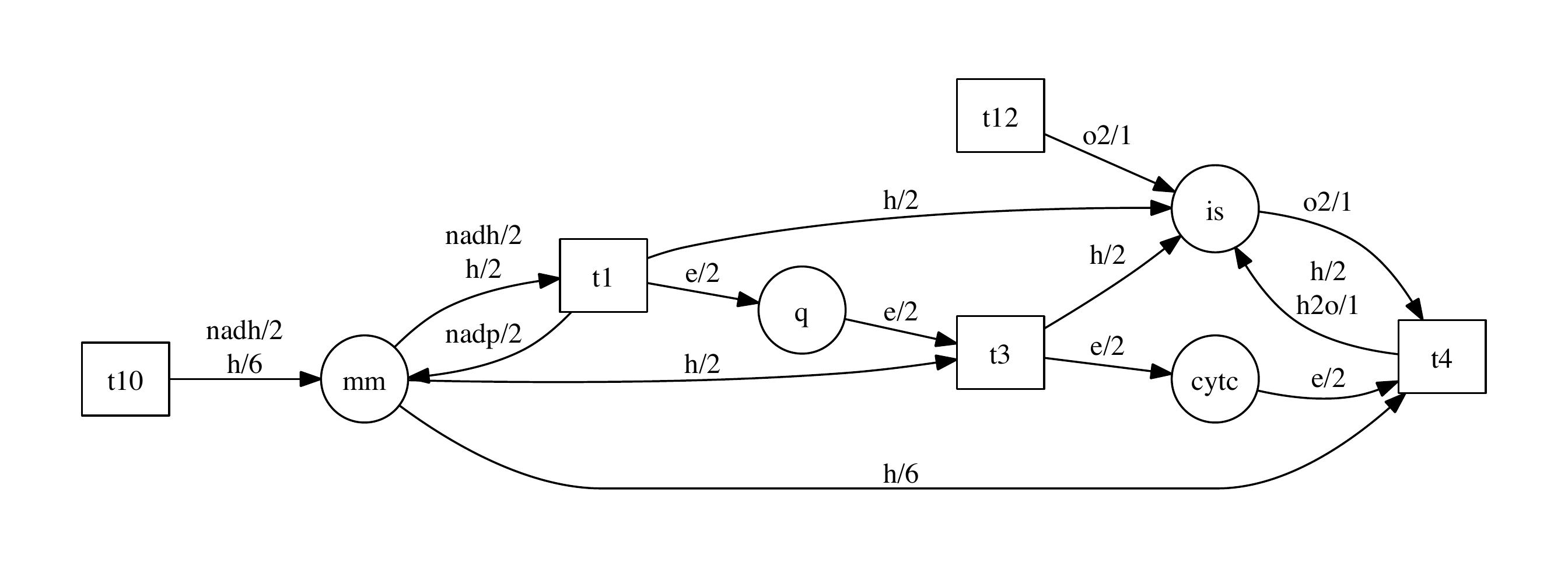}
\caption{Petri Net with tokens of colors $\{e,h,h2o,nadh,nadp,o2\}$. Circles represent places, and rectangles represent transitions. Arc weights such as ``$nadh/2,h/2$'', ``$h/2,h2o/1$'' specify the number of tokens consumed and produced during the execution of their respective transitions, where ``$nadh/2,h/2$'' means 2 tokens of color $nadh$ and 2 tokens of $h$. Similar notation is used to specify marking on places, when not present, the place is assumed to be empty of tokens.}
\label{fig:echain}
\vspace{-10pt}
\end{figure}

If the Figure~\ref{fig:echain}  Petri Net has the marking: $M_0(mm)=[nadh/2,h/6]$, $M_0(q)=[e/2]$, $M_0(cytc)=[e/2]$, $M_0(is)=[o2/1]$, then transitions $t1,t3,t4$ are enabled. However, either $\{t1,t3\}$ or $\{t4\}$ can fire simultaneously in a single firing at time 0 due to limited $h$ tokens in $mm$. $t4$ is said to be in conflict with $t1,t3$.

\section{Translating Petri Nets with Colored Tokens to ASP}\label{sec:enc_cpn}
In order to represent the Petri Net $PN^C$ with colored tokens, initial marking $M_0$, and simulation length $k$, we modify our encoding in Section~\ref{sec:enc_query} to add a new color parameter to all rules and facts containing token counts in them.
We keep rules $f\ref{f:c:place}, f\ref{f:c:trans}, f\ref{f:c:time}, f\ref{f:c:num}, f\ref{f:rptarc:elim}$ remain as they were for basic Petri Nets. We add a new rule $f\ref{f:c:col}$ for possible set of token colors and replace rules $f\ref{f:r:ptarc}, f\ref{f:r:tparc}, f\ref{f:rptarc}, f\ref{f:iptarc}, f\ref{f:tptarc}, i\ref{i:holds}$ with $f\ref{f:c:ptarc}, f\ref{f:c:tparc}, f\ref{f:c:rptarc}, f\ref{f:c:iptarc}, f\ref{f:c:tptarc}, i\ref{i:c:init}$ to add the color parameter as follows:
\begin{sloppypar}
\begin{description}
\item[\nextf:\label{f:c:col}] Facts \texttt{\small col($c_k$)} where $c_k \in C$ is a color. %
\item[\nextf:\label{f:c:ptarc}] Rules \texttt{\small ptarc($p_i,t_j,n_c,c,ts_k$) :- time($ts_k$).} for each $(p_i,t_j) \in E^-$, $c \in C$, $n_c=m_{W(p_i,t_j)}(c) : n_c > 0$.\footnote{The time parameter $ts_k$ allows us to capture reset arcs, which consume tokens equal to the current (time-step based) marking of their source nodes.} %
\item[\nextf:\label{f:c:tparc}] Rules \texttt{\small tparc($t_i,p_j,n_c,c,ts_k$) :- time($ts_k$).} for each $(t_i,p_j) \in E^+$ , $c \in C$, $n_c=m_{W(t_i,p_j)}(c) : n_c > 0$. %
\item[\nextf:\label{f:c:rptarc}] Rules \texttt{\small ptarc($p_i,t_j,n_c,c,ts_k$) :- holds($p_i,n_c,c,ts_k$),  num($n_c$), $n_c$>0, time($ts_k$).}  for each $(p_i,t_j):$ $p_i \in R(t_j)$, $c \in C$, $n_c=m_{M_k(p_i)}(c)$. %
\item[\nextf:\label{f:c:iptarc}] Rules \texttt{\small iptarc($p_i,t_j,1,c,ts_k$) :- time($ts_k$).} for each $(p_i,t_j): p_i \in I(t_j)$, $c \in C$. %
\item[\nextf:\label{f:c:tptarc}] Rules \texttt{\small tptarc($p_i,t_j,n_c,c,ts_k$) :- time($ts_k$).} for each $(p_i,t_j) \in Q$, $c \in C$, $n_c=m_{QW(p_i,t_j)}(c) : n_c > 0 $. %
\item[\nexti:\label{i:c:init}] Facts \texttt{\small holds($p_i,n_c,c,0$).} for each place $p_i \in P, c \in C, n_c=m_{M_0(p_i)}(c)$.
\end{description}
\end{sloppypar}

Next, we encode Petri Net's execution behavior, which proceeds in discrete time steps. Rules $e\ref{e:r:ne:ptarc}, e\ref{e:ne:iptarc}, e\ref{e:ne:tptarc}, e\ref{e:enabled}$ are replaced by $e\ref{e:c:ne:ptarc}, e\ref{e:c:ne:iptarc}, e\ref{e:c:ne:tptarc}, e\ref{e:c:enabled}$. For a transition $t_i$ to be enabled, it must satisfy the following conditions:
\begin{inparaenum}[(i)]
\item $\nexists p_j \in \bullet t_i : M(p_j) < W(p_j,t_i)$,
\item $\nexists p_j \in I(t_i) : M(p_j) > 0$, and
\item $\nexists (p_j,t_i) \in Q :  M(p_j) < QW(p_j,t_i)$
\end{inparaenum}.
These three conditions are encoded as $e\ref{e:c:ne:ptarc},e\ref{e:c:ne:iptarc},e\ref{e:c:ne:tptarc}$, respectively and we encode the absence of any of these conditions for a transition as $e\ref{e:c:enabled}$:
\begin{sloppypar}
\begin{description}
\item[\nexte:\label{e:c:ne:ptarc}] \texttt{\small notenabled(T,TS) :- ptarc(P,T, N,C,TS), holds(P,Q,C,TS),  
   place(P),  \\ trans(T), time(TS), num(N), num(Q), col(C), Q<N.} %

\item[\nexte:\label{e:c:ne:iptarc}] \texttt{\small notenabled(T,TS) :- iptarc(P,T,N,C,TS), holds(P,Q,C,TS), 
   place(P), \\ trans(T), time(TS), num(N), num(Q), col(C), Q>=N.} %

\item[\nexte:\label{e:c:ne:tptarc}] \texttt{\small notenabled(T,TS) :- tptarc(P,T,N,C,TS), holds(P,Q,C,TS),
    place(P), \\ trans(T), time(TS), num(N), num(Q), col(C), Q<N.} %

\item[\nexte:\label{e:c:enabled}] \texttt{\small enabled(T,TS) :- trans(T), time(TS), not notenabled(T,TS).} %

\end{description}
\end{sloppypar}
Rule $e\ref{e:c:ne:ptarc}$ captures the existence of an input place $P$ with insufficient number of tokens for transition $T$ to fire. Rule $e\ref{e:c:ne:iptarc}$ captures existence of a non-empty source place $P$ of an inhibitor arc to $T$ preventing $T$ from firing. Rule $e\ref{e:c:ne:tptarc}$ captures existence of a source place $P$ with less than arc-weight tokens required by the read arc to transition $T$ for $T$ to be enabled. The, \texttt{\small holds(P,Q,C,TS)} predicate captures the marking of place $P$ at time $TS$ as $Q$ tokens of color $C$. Rule $e\ref{e:c:enabled}$ captures enabling of transition $T$ when no reason for it to be not enabled is determined by $e\ref{e:c:ne:ptarc},e\ref{e:c:ne:iptarc},e\ref{e:c:ne:tptarc}$. In a biological context, this enabling is equivalent to a reaction's pre-conditions being satisfied. A reaction can proceed when its input substances are available in the required quantities, it is not inhibited, and any required activation quantity of activating substances is available.

Any subset of enabled transitions can fire simultaneously at a given time-step. We select a subset of fireable transitions using the choice rule $a\ref{a:c:fires}$ %
The choice rule $a\ref{a:c:fires}$ either picks an enabled transition $T$ for firing at time $TS$ or not. The combined effect over all transitions is to pick a subset of enabled transitions to fire. Rule $f\ref{f:c:rptarc:elim}$ ensures that enabled reset-transitions will be a part of this firing set. Whether these transitions are in conflict are checked by later rules $a\ref{a:c:overc:place},a\ref{a:c:overc:gen},a\ref{a:c:overc:elim}$. In a biological context, the multiple firing models parallel processes occurring simultaneously. The marking is updated according to the firing set using rules $r\ref{r:c:add}, r\ref{r:c:del}, r\ref{r:c:totincr}, r\ref{r:c:totdecr}, r\ref{r:c:nextstate}$ which replaced $r\ref{r:r:add}, r\ref{r:r:del}, r\ref{r:totincr}, r\ref{r:totdecr}, r\ref{r:nextstate}$ as follows:
\begin{sloppypar}
\begin{description}
\item[\nextr:\label{r:c:add}] \texttt{\small add(P,Q,T,C,TS) :- fires(T,TS), tparc(T,P,Q,C,TS), time(TS).} %
\item[\nextr:\label{r:c:del}] \texttt{\small del(P,Q,T,C,TS) :- fires(T,TS), ptarc(P,T,Q,C,TS), time(TS). } %

\item[\nextr:\label{r:c:totincr}] \texttt{\small tot\_incr(P,QQ,C,TS) :- col(C), 
   QQ = \#sum[add(P,Q,T,C,TS) = Q : num(Q) : trans(T)], 
   time(TS), num(QQ), place(P).} %

\item[\nextr:\label{r:c:totdecr}] \texttt{\small tot\_decr(P,QQ,C,TS) :- col(C),
   QQ = \#sum[del(P,Q,T,C,TS) = Q : num(Q) : trans(T)], 
   time(TS), num(QQ), place(P).} %

\item[\nextr:\label{r:c:nextstate}] \texttt{\small holds(P,Q,C,TS+1):-place(P),num(Q;Q1;Q2;Q3),time(TS),time(TS+1),col(C),
  holds(P,Q1,C,TS), tot\_incr(P,Q2,C,TS), 
    tot\_decr(P,Q3,C,TS), Q=Q1+Q2-Q3.} %
\end{description}
\end{sloppypar}

Rules $r\ref{r:c:add}$ and $r\ref{r:c:del}$ capture that $Q$ tokens of color $C$ will be added or removed to/from place $P$ due to firing of transition $T$ at the respective time-step $TS$. Rules $r\ref{r:c:totincr}$ and $r\ref{r:c:totdecr}$ aggregate these tokens for each $C$ for each place $P$ (using aggregate assignment \texttt{\small QQ = \#sum[\dots ]}) at the respective time-step $TS$. Rule $r\ref{r:c:nextstate}$ uses the aggregates to compute the next marking of $P$ for color $C$ at the time-step ($TS+1$) by subtracting removed tokens and adding added tokens to the current marking. In a biological context, this captures the effect of a process / reaction, which consumes its inputs and produces outputs for the downstream processes. We capture token  overconsumption using the rules $a\ref{a:c:overc:place}, a\ref{a:c:overc:gen}, a\ref{a:c:overc:elim}$ of which $a\ref{a:c:overc:place}$ is a colored replacement for $a\ref{a:overc:place}$ and is encoded as follows:
\begin{sloppypar}
\begin{description}
\item[\nexta:\label{a:c:overc:place}] \texttt{\small consumesmore(P,TS) :- holds(P,Q,C,TS), tot\_decr(P,Q1,C,TS), Q1 > Q. } %
\end{description} 
\end{sloppypar}

Rule $a\ref{a:c:overc:place}$ determines whether firing set selected by $a\ref{a:c:fires}$ will cause overconsumption of tokens at $P$ at time $TS$ by comparing available tokens to aggregate tokens removed as determined by $r\ref{r:c:totdecr}$. Rule $a\ref{a:c:overc:gen}$ generalizes the notion of overconsumption, while rule $a\ref{a:c:overc:elim}$ eliminates answer with such overconsumption.

In a biological context, conflict (through overconsumption) models the limitation of input substances, which dictate which downstream processes can occur simultaneously.

We remove rules $a\ref{a:maxfire:cnh}, a\ref{a:maxfire:elim}$ from previous encoding to get the {\em set firing} semantics. Now, we extend the definition \eqref{def:11cor} of $\text{1-1}$ correspondence between the execution sequence of Petri Net and the answer-sets of its ASP encoding to Petri Nets with colored tokens as follows.
\begin{definition}\label{def:c:11cor}
Given a Petri Net $PN$ with colored tokens, its initial marking $M_0$ and its encoding $\Pi(PN,M_0,k,ntok)$ for $k$-steps and maximum $ntok$ tokens at any place. We say that there is a 1-1 correspondence between the answer sets of $\Pi(PN,M_0,k,ntok)$ and the execution sequences of $PN$ iff for each answer set $A$ of $\Pi(PN,M_0,k,ntok)$, there is a corresponding execution sequence $X=M_0,T_0,M_1,\dots,M_k,T_k,M_{k+1}$ of $PN$ and for each execution sequence $X$ of $PN$ there is an answer-set $A$ of $\Pi(PN,M_0,$ $k,ntok)$ such that 
\begin{equation*}
\{ fires(t,ts) : t \in T_{ts}, 0 \leq ts \leq k \} = \{ fires(t,ts) : fires(t,ts) \in A \}
\end{equation*}
\begin{equation*}
\begin{split}
\{ holds(p,q,c,ts) &: p \in P, c/q=M_{ts}(p), 0 \leq ts \leq k+1 \} \\
&= \{ holds(p,q,c,ts) : holds(p,q,c,ts) \in A\} 
\end{split}
\end{equation*}
\end{definition}

\begin{proposition}\label{prop:pnc}
There is 1-1 correspondence between the answer sets of $\Pi^5(PN^C,M_0,$ $k,ntok)$ and the execution sequences of $PN$.
\end{proposition}

To add {\em maximal firing semantics}, we add $a\ref{a:c:maxfire:elim}$ as it is and replace $a\ref{a:maxfire:cnh}$ with $a\ref{a:c:maxfire:cnh}$ as follows:
{\small
\begin{sloppypar}
\begin{description}
\item[\nexta:\label{a:c:maxfire:cnh}] \texttt{\small could\_not\_have(T,TS):-enabled(T,TS),not fires(T,TS),
   ptarc(S,T,Q,C,TS), holds(S,QQ,C,TS), tot\_decr(S,QQQ,C,TS),
   Q > QQ - QQQ.} %
\end{description} %
\end{sloppypar}
}
Rule $a\ref{a:c:maxfire:cnh}$ captures the fact that transition $T$, though enabled, could not have fired at $TS$, as its firing would have caused overconsumption. Rule $a\ref{a:c:maxfire:elim}$ eliminates any answers where an enabled transition could have fired without causing overconsumption but did not. This modification reduces the number of answers produced for the Petri Net in Figure~\ref{fig:echain} to 4. We can encode other firing semantics with similar ease\footnote{For example, if \textit{interleaved} semantics is desired, rules $a\ref{a:c:maxfire:cnh},a\ref{a:c:maxfire:elim}$ can changed to capture and eliminate answer-sets in which more than one transition fires in a firing set as:
\vspace{-5pt}
\begin{description}
\item[a\ref{a:c:maxfire:cnh}':] \texttt{\small more\_than\_one\_fires :- fires(T1,TS),fires(T2,TS),T1!=T2,time(TS).}
\item[a\ref{a:c:maxfire:elim}':] \texttt{\small :-more\_than\_one\_fires.}
\end{description}}.
We now look at how additional extensions can be easily encoded by making small code changes.

\section{Extension - Priority Transitions}\label{sec:enc_priority}
Priority transitions enable ordering of Petri Net transitions, favoring high priority transitions over lower priority ones~\cite{best1992petri}. In a biological context, this is used to model primary (or dominant) vs. secondary pathways / processes in a biological system. This prioritization may be due to an intervention (such as prioritizing elimination of a metabolite over recycling it).

\begin{definition}[Priority Colored Petri Net]\label{def:pnri}
A {\em Priority Colored Petri Net} with reset, inhibit, and read arcs is a tuple $PN^{pri} = (P,T,E,C,W,R,I,Q,QW,Z)$, where:
$P,T,E,C,W,R,I,Q,QW$ are the same as for $PN^C$, and 
$Z : T \rightarrow \mathds{N}$ is a priority function that assigns priorities to transitions. Lower number signifies higher priority.
\end{definition}

\begin{definition}[Enabled Transition]\label{def:pnpri:enable}\label{def:pndur:enable}
A transition $t_i$ is enabled in $PN^{pri}$ w.r.t. a marking $M$ (prenabled$_{M}(t)$) if it would be enabled in $PN^C$ w.r.t. $M$ and there isn't another transition $t_j$ that would be enabled in $PN^C$ (with respect to M) s.t. $Z(t_j) < Z(t_i)$.
\end{definition}

\begin{definition}[Firing Set]\label{def:pnpri:firing_set}\label{def:pndur:firing_set}
A firing set is a set $T_k=\{t_{k_1},\dots,t_{k_n}\} \subseteq T$ of simultaneously firing transitions of $PN^{pri}$ that are priority enabled and do not conflict w.r.t. to the current marking $M_k$ of $PN$. A set $T_k$ is not a firing set if there is an priority enabled reset-transition that is not in $T_k$, i.e. $\exists t : prenabled_{M_k}(t), R(t) \neq \emptyset, t \not\in T_k$.~\footnote{See footnote~\ref{fn:rptarc:conflict}} %
\end{definition}

We add the following facts and rules to encode transition priority and enabled priority transitions:
{\small
\begin{sloppypar}
\begin{description}
\item[\nextf:\label{f:c:pr}] Facts \texttt{\small transpr($t_i$,$pr_i$)} where $pr_i = Z(t_i)$ is $t_i's$ priority. %
\item[\nexta:\label{a:c:prne}] \texttt{\small notprenabled(T,TS) :- enabled(T,TS), transpr(T,P), 
   enabled(TT,TS), \\ transpr(TT,PP), PP < P.} %
\item[\nexta:\label{a:c:prenabled}] \texttt{\small prenabled(T,TS) :- enabled(T,TS), not notprenabled(T,TS).} %
\end{description}
\end{sloppypar}
}

Rule $a\ref{a:c:prne}$ captures that an enabled transition $T$ is not priority-enabled, if there is another enabled transition with higher priority at $TS$. Rule $a\ref{a:c:prenabled}$ captures that transition $T$ is priority-enabled at $TS$ since there is no enabled transition with higher priority. We replace rules $a\ref{a:c:fires},f\ref{f:c:rptarc:elim},a\ref{a:c:maxfire:cnh},a\ref{a:c:maxfire:elim}$ with $a\ref{a:c:prfires},f\ref{f:c:pr:rptarc:elim}, a\ref{a:c:prmaxfire:cnh},a\ref{a:c:prmaxfire:elim}$ respectively to propagate priority as follows:

{\small
\begin{sloppypar}
\begin{description}
\item[\nexta:\label{a:c:prfires}] \texttt{\small \{fires(T,TS)\} :- prenabled(T,TS), trans(T), time(TS).} %

\item[\nextf:\label{f:c:pr:rptarc:elim}] Rules \texttt{\small :- prenabled($t_j,ts_k$),not fires($t_j,ts_k$), time($ts_k$).} for each transition $t_j$ with an incoming reset arc.

\item[\nexta:\label{a:c:prmaxfire:cnh}] \texttt{\small could\_not\_have(T,TS) :- prenabled(T,TS), not fires(T,TS), 
    ptarc(S,T,Q,C,TS), holds(S,QQ,C,TS), tot\_decr(S,QQQ,C,TS),
   Q > QQ - QQQ.} %

\item[\nexta:\label{a:c:prmaxfire:elim}] \texttt{\small :- not could\_not\_have(T,TS), time(TS), prenabled(T,TS), not fires(T,TS), trans(T).}

\end{description}
\end{sloppypar}
}

Rules $a\ref{a:c:prfires},f\ref{f:c:rptarc:elim},a\ref{a:c:prmaxfire:cnh},a\ref{a:c:prmaxfire:elim}$ perform the same function as $a\ref{a:c:fires},f\ref{f:c:pr:rptarc:elim},a\ref{a:c:maxfire:cnh},a\ref{a:c:maxfire:elim}$, except that they consider only priority-enabled transitions as compared all enabled transitions.

\begin{proposition}\label{prop:pri}
There is 1-1 correspondence between the answer sets of $\Pi^6(PN^{pri},M_0,$ $k,ntok)$ and the execution sequences of $PN^{pri}$.
\end{proposition}

\section{Extension - Timed Transitions}\label{sec:enc_dur}
Biological processes vary in time required for them to complete. Timed transitions~\cite{ramchandani1974analysis} model this variation of duration. The timed transitions can be reentrant or non-reentrant\footnote{A {\bf reentrant} transition is like a vehicle assembly line, which accepts new parts while working on multiple vehicles at various stages of completion; whereas a {\bf non-reentrant} transition only accepts new input when the current processing is finished.}. We extend our encoding to allow reentrant timed transitions.

\begin{figure}[htbp]
\centering
\vspace{-20pt}
\includegraphics[width=\linewidth]{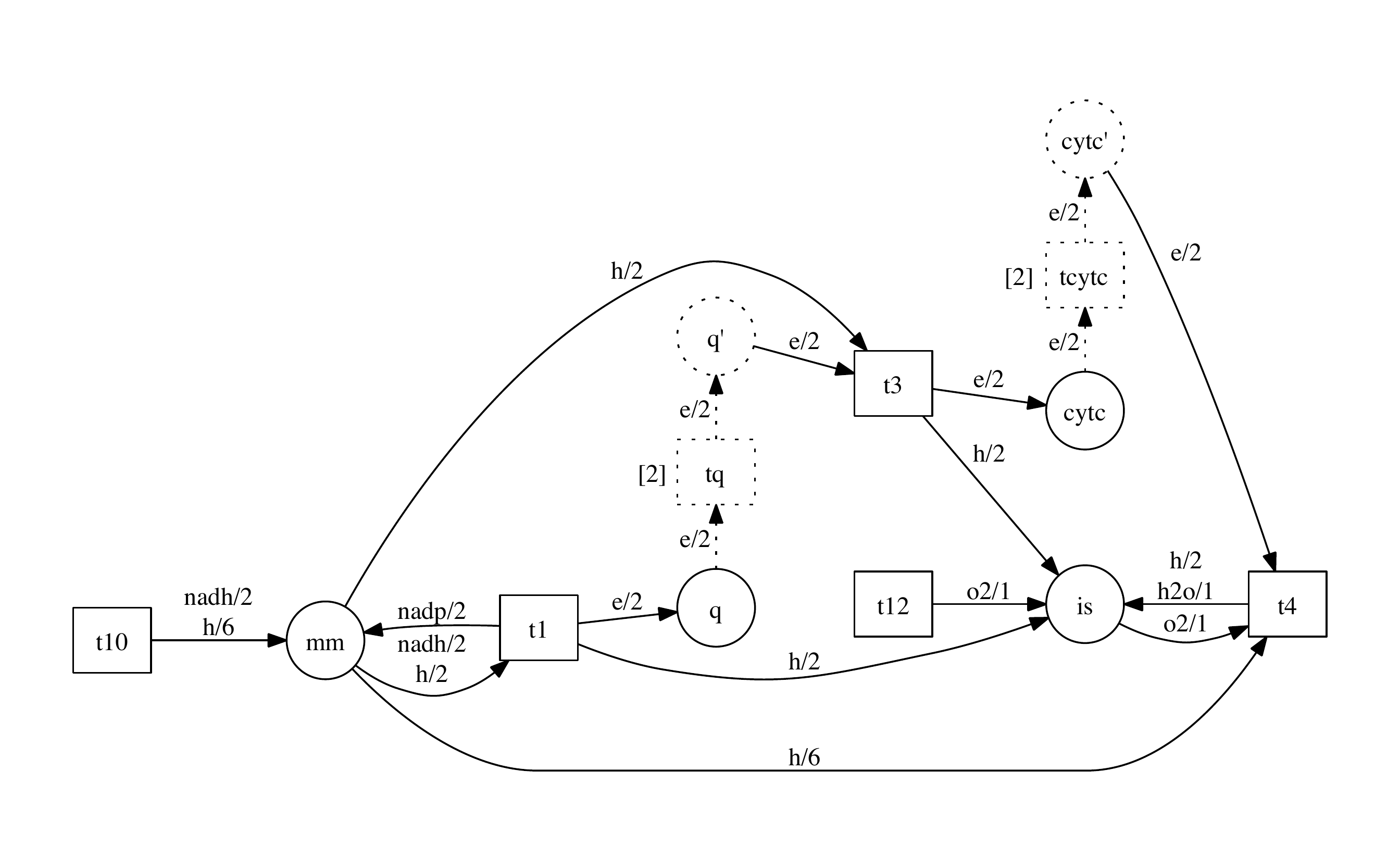}
\caption{An extended version of the Petri Net model from Fig.~\ref{fig:echain}. The new transitions $tq,tcytc$ have a duration of 2 each (shown in square brackets (``[ ]'') next to the transition). When missing, transition duration is assumed to be 1.}
\label{fig:echaintm}
\end{figure}

\begin{definition}[Priority Colored Petri Net with Timed Transitions]\label{def:pndur}
A {\em Priority Colored Petri Net with Timed Transitions}, reset, inhibit, and query arcs is a tuple $PN^D=(P,T,E,C,W,R,I,Q,QW,Z,D)$, where
$P,T,E,C,W,R,I,Q,QW,Z$ are the same as for $PN^{pri}$, and 
$D : T \rightarrow \mathds{N} \setminus \{0\}$ is a duration function that assigns positive integer durations to transitions.
\end{definition}

Figure~\ref{fig:echaintm} shows an extended version of Petri Net model of the Electron Transport Chain~\cite{CampbellBook} shown in Figure~\ref{fig:echain}. The new transitions $tq$ and $tcytc$ (shown in dotted outline) are timed transitions modeling the speed of the small carrier molecules, Coenzyme Q ($q$) and Cytochrome C ($cytc$) as an effect of membrane fluidity. Higher numbers for transition duration represent slower movement of the carrier molecules due to lower fluidity. 

\begin{definition}[Transition Execution]\label{def:pndur:texec}
A transition $t$ in $PN^D$ consumes tokens from its input places and reset places immediately, while it produces tokens in its output places at the end of transition duration $D(t)$, as follows:
\[
\forall p \in \bullet t, M_{k+1}(p)  = M_k(p) - W(p,t)
\]
\[
\forall p \in t\bullet, M_{k+D(t)}(p)  = M_{k+D(t)-1}(p) + W(p,t)
\]
\[
\forall p \in R(t), M_{k+1}(p)  = M_k(p) - M_k(p)
\]
\end{definition}

{\em Execution in $PN^D$} changes, since the token update from $M_k$ to $M_{k+1}$ can involve transitions that started at some time $l$ before time $k$, but finish at $k+1$.
\begin{definition}[Firing Set Execution]\label{def:pndur:exec}
New marking due to firing set execution is computed as follows:
\[
\forall p \in P \setminus R(T_k), M_{k+1}(p)  = M_k(p)  
-  \sum_{\substack{t \in T_k, p \in \bullet t}} W(p,t) 
+ \sum_{\substack{t \in T_l, p \in t \bullet: 0 \leq l \leq k, l+D(t) = k+1}} W(t,p)
\]
\[ 
\forall p \in R(T_k), M_{k+1}(p) = \sum_{\substack{t \in T_l , p \in t \bullet : l \leq k, l+D(t) = k+1 }} W(t,p)
\] 
where $R(T_i)=\displaystyle\cup_{\substack{t \in T_i}} R(t)$.
\end{definition}

A timed transition $t$ produces its output $D(t)$ time units after being fired. We replace $f\ref{f:c:tparc}$ with $f\ref{f:c:dur:tparc}$ adding transition duration and replace rule $r\ref{r:c:add}$ with $r\ref{r:c:dur:add}$ that produces tokens at the end of transition duration:
{\small
\begin{sloppypar}
\begin{description}
\item[\nextf:\label{f:c:dur:tparc}] Rules \texttt{\small tparc($t_i,p_j,n_c,c,ts_k,D(t_i)$):-time($ts_k$).} for each $(t_i,p_j) \in E^+$, $c \in C$, $n_c=m_{W(t_i,p_j)}(c) : n_c > 0$. %
\item[\nextr:\label{r:c:dur:add}] \texttt{\small add(P,Q,T,C,TS):-fires(T,TS0),time(TS0;TS), tparc(T,P,Q,C,TS0,D), \\ TS=TS0+D-1.} %
\end{description}
\end{sloppypar}
}

\begin{proposition}\label{prop:dur}
There is 1-1 correspondence between the answer sets of $\Pi^7(PN^D,M_0,$ $k,ntok)$ and the execution sequences of $PN^D$.
\end{proposition}

Above implementation of timed-transition is reentrant, however, we can easily make these timed transitions non-reentrant by adding rule $e\ref{e:c:ne:dur}$ that disallows a transition from being enabled if it is already in progress:
{\small
\begin{sloppypar}
\begin{description}\label{fn:c:enc_dur_nre} 
\item[\nexte:\label{e:c:ne:dur}] \texttt{\small notenabled(T,TS):-fires(T,TS0), num(N), TS>TS0, 
    tparc(T,P,N,C,TS0,D), 
    col(C), time(TS0), time(TS), TS<(TS0+D).} %
\end{description}
\end{sloppypar}
}

\section{Other Extensions}
Other Petri Net extensions can be implemented with similar ease. For example, \textit{Guard Conditions} on transitions can be trivially implemented as a \texttt{\small notenabled/2} rules. \textit{Self Modifying Petri Nets}~\cite{SelfModNets}, which allow marking-dependent arc-weights can be implemented in a similar manner as the \textit{Reset Arc} extension in section~\ref{sec:enc_reset}. \textit{Object Petri Nets}~\cite{ValkObjectPetriNets}, in which each token is a Petri Net itself can be implemented (using token reference semantics) by adding an additional ``network-id'' parameter to our encoding, where ``id=0'' is reserved for system net and higher numbers are used for token nets. Transition coordination between system \& token nets is enforced through constraints on transition labels, where transition labels are added as additional facts about transitions.

\section{Related Work}
Petri Nets have been previously encoded in ASP, but the previous implementations have been limited to restricted classes of Petri Nets. For example, 1-safe Petri Net to ASP translation has been presented in \cite{HeljankoNMR}, which is limited to binary Petri Nets. Translation of Logic Petri Nets to ASP has been presented in \cite{LogicPetriNets}, but their model cannot handle numerical aggregation of tokens from multiple input transitions to the same place. Our work focused on problems in the biological domain and is more generalized. We can represent reset arcs, inhibition arcs, priority arcs as well as durative transitions.

\section{Conclusion}
We have presented an encoding of basic Petri Nets in ASP and showed how it can be easily extended to include extension to model various biological constructs. Portions of this work were published in ~\cite{anwar2013encoding} and ~\cite{anwar2013encodinghl}. In the next chapter we will use Petri Nets and their ASP encoding to model biological pathways to answer questions about them.

\chapter{Answering Questions using Petri Nets and ASP}
\label{ch:modeling_qa}

\section{Introduction}
In this chapter we use various Petri Net extensions presented in Chapter \ref{ch:asp_enc} and their ASP encoding to answer question from \cite{CampbellBook} that were a part of the Second Deep Knowledge Representation Challenge\footnote{https://sites.google.com/site/2nddeepkrchallenge/}. 

\begin{definition}[Rate]
Rate of product P is defined as the quantity of P produced per unit-time. Rate of an action A is defined as the number of time A occurs per unit-time.
\end{definition}

\section{Comparing Altered Trajectories due to Reset Intervention}
\begin{question}\label{q1}
At one point in the process of glycolysis, both dihydroxyacetone phosphate (DHAP) and glyceraldehyde 3-phosphate (G3P) are produced. Isomerase catalyzes the reversible conversion between these two isomers. The conversion of DHAP to G3P never reaches equilibrium and G3P is used in the next step of glycolysis. What would happen to the rate of glycolysis if DHAP were removed from the process of glycolysis as quickly as it was produced?
\end{question}

Provided Answer: 
\begin{quote}
``Glycolysis is likely to stop, or at least slow it down. The conversion of the two isomers is reversible, and the removal of DHAP will cause the reaction to shift in that direction so more G3P is converted to DHAP. If less (or no) G3P were available, the conversion of G3P into DHAP would slow down (or be unable to occur).''
\end{quote}

\begin{solution}
The process of glycolysis is shown in Fig 9.9 of Campbell's book. Glycolysis splits Glucose into Pyruvate. In the process it produces ATP and NADH. Any one of these can be used to gauge the glycolysis rate, since they will be produced in proportion to the input Glucose. The amount of pyruvate produced is the best choice since it is the direct end product of glycolysis. The ratio of the quantity of pyruvate produced over a representative span of time gives us the glycolysis rate. We assume a steady supply of Glucose is available and also assume that sufficient quantity of various enzymes used in glycolysis is available, since the question does not place any restriction on these substances.

We narrow our focus to a subsection from Fructose 1,6-bisphosphate (F16BP) to 1,3-Bisphosphoglycerate (BPG13) as shown in Figure \ref{fig:q1} since that is the part the question is concerned with. We can ignore the linear chain up-stream of F16BP as well as the linear chain down-stream of BPG13 since the amount of F16BP available will be equal to Glucose and the amount of BPG13 will be equal to the amount of Pyruvate given our steady supply assumption.

\begin{figure}[H]
\begin{center}
\includegraphics[width=9cm]{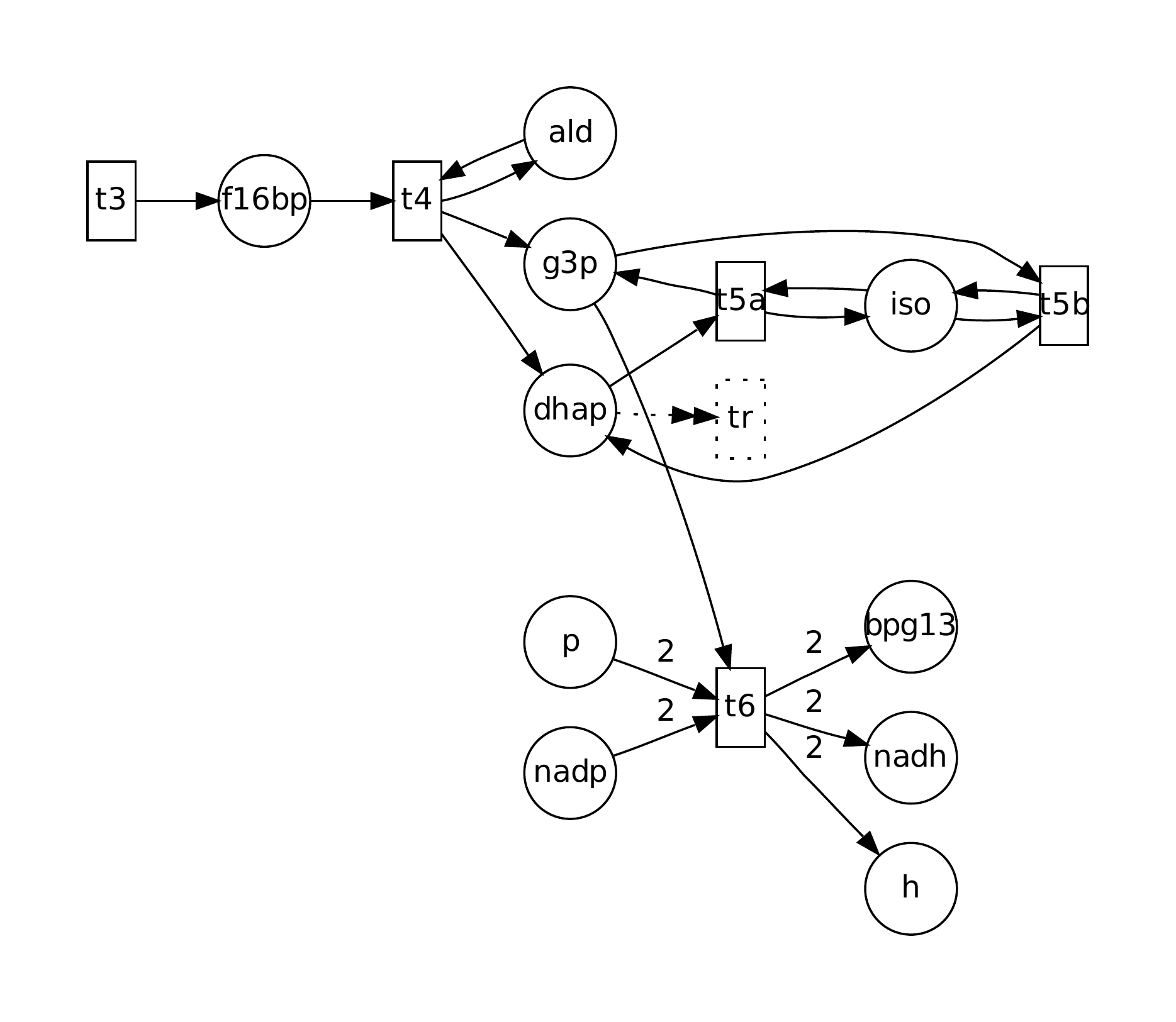}
\caption{Petri Net graph relevant to question \ref{q1}. ``f16bp'' is the compound Fructose 1,6-biphosphate, ``bpg13'' is 1,3-Bisphosphoglycerate. Transition $tr$ shown in dotted lines is added to model the elimination of $dhap$ as soon as it is produced.}
\label{fig:q1}
\end{center}
\end{figure}

We fulfill the steady supply requirement of Glucose by a source transition-node $t3$. We fulfill sufficient enzyme supply by a fixed quantity for each enzyme such that this quantity is in excess of  what can be consumed during our simulation interval. Where the simulation interval is the number of time-steps over which we will measure the rate of glycolysis.

We model the elimination of DHAP as soon as it is produced with a reset arc, shown with a dotted style in Figures \ref{fig:q1}. Such an arc removes all tokens from its source place when it fires. Since we have added it as an unconditional arc, it is always enabled for firing. We encode both situations in ASP with the maximal firing set policy. Both situations (without and with reset arc) are encoded in ASP and run for 10 steps. At the end of those 10 steps the amount of BPG13 is compared to determine the difference in the rate of glycolysis.

In normal situation (without $(dhap,tr)$ reset arc), unique quantities of ``bpg13'' from all (2) answer-sets after 10 steps were as follows:
{\footnotesize
\begin{verbatim}
holds(bpg13,14,10)
holds(bpg13,16,10)
\end{verbatim}
}

with reset arc $tr$, unique quantities of ``bpg13'' from all (512) answer-sets after 10 steps were as follows:
{\footnotesize
\begin{verbatim}
holds(bpg13,0,10)
holds(bpg13,10,10)
holds(bpg13,12,10)
holds(bpg13,14,10)
holds(bpg13,16,10)
holds(bpg13,2,10)
holds(bpg13,4,10)
holds(bpg13,6,10)
holds(bpg13,8,10)
\end{verbatim}
}

\begin{figure}[H]
\begin{center}
\includegraphics[width=12cm]{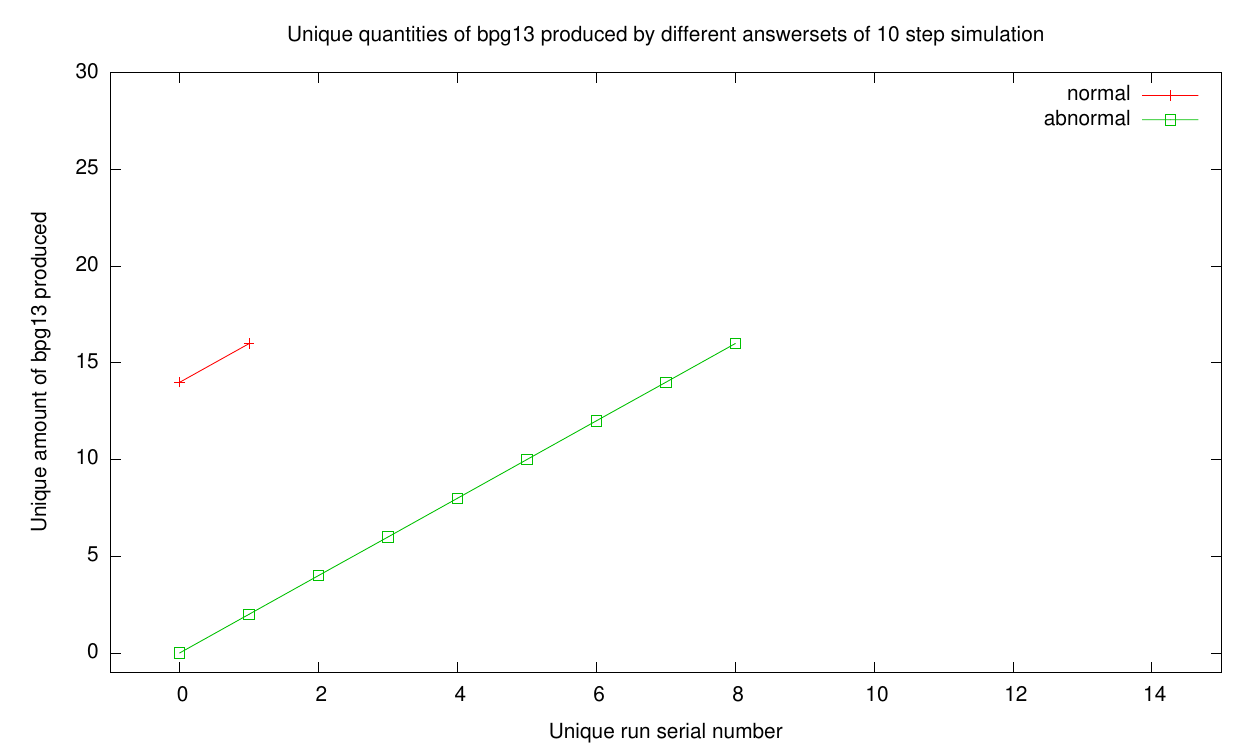}
\caption{Amount of ``bpg13'' produced in unique answer-sets produced by a 10 step simulation. The graph shows two situations, without the $(dhap,tr)$ reset arc  (normal situation) and with the reset arc (abnormal situation). The purpose of this graph is to depict the variation in the amounts of glycolysis produced in various answer sets.}
\label{fig:q1:result}
\end{center}
\end{figure}

Note that the rate of glycolysis is generally lower when DHAP is immediately consumed. It is as low as zero essentially stopping glycolysis. The range of values are due to the choice between G3P being converted to DHAP or BPG13. If more G3P is converted to DHAP, then less BPG13 is produced and vice versa. Also, note that if G3P is not converted to BPG13, no NADH or ATP is produced either due to the liner chain from G3P to Pyruvate. The unique quantities of BPG13 are shown in a graphical format in Figure \ref{fig:q1:result}, while a trend of average quantity of BPG13 produced is shown in Figure \ref{fig:q1:result:avg}.

\begin{figure}[H]
\begin{center}
\includegraphics[width=12cm]{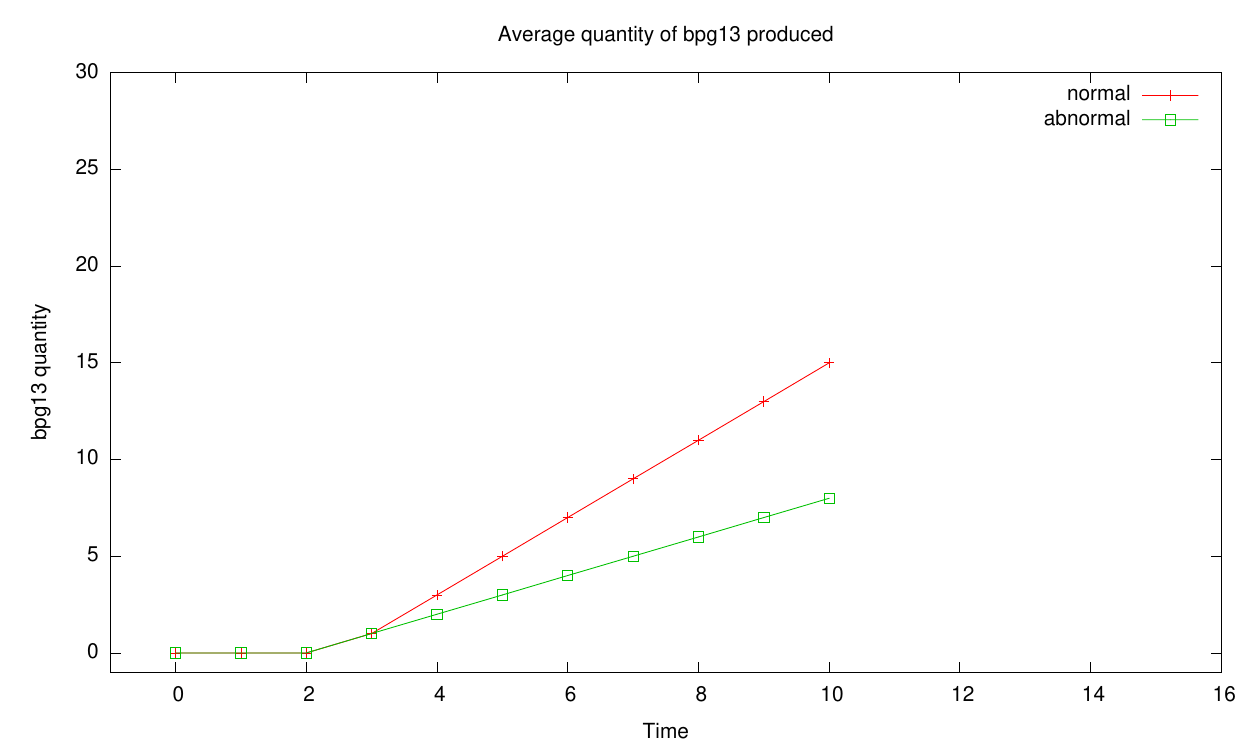}
\caption{Average amount of ``bpg13'' produced during the 10-step simulation at various time steps. The average is over all answer-sets. The graph shows two situations, without the $(dhap,tr)$ reset arc  (normal situation) and with the reset arc (abnormal situation). The divergence in ``bpg13'' production is clearly shown.}
\label{fig:q1:result:avg}
\end{center}
\end{figure}

We created a minimal model of the Petri Net in Figure \ref{fig:q1} by removing enzymes and reactants that were not relevant to the question and did not contribute to the estimation of glycolysis. This is shown in Figure \ref{fig:q1:min}.

\begin{figure}[H]
\begin{center}
\includegraphics[width=7cm]{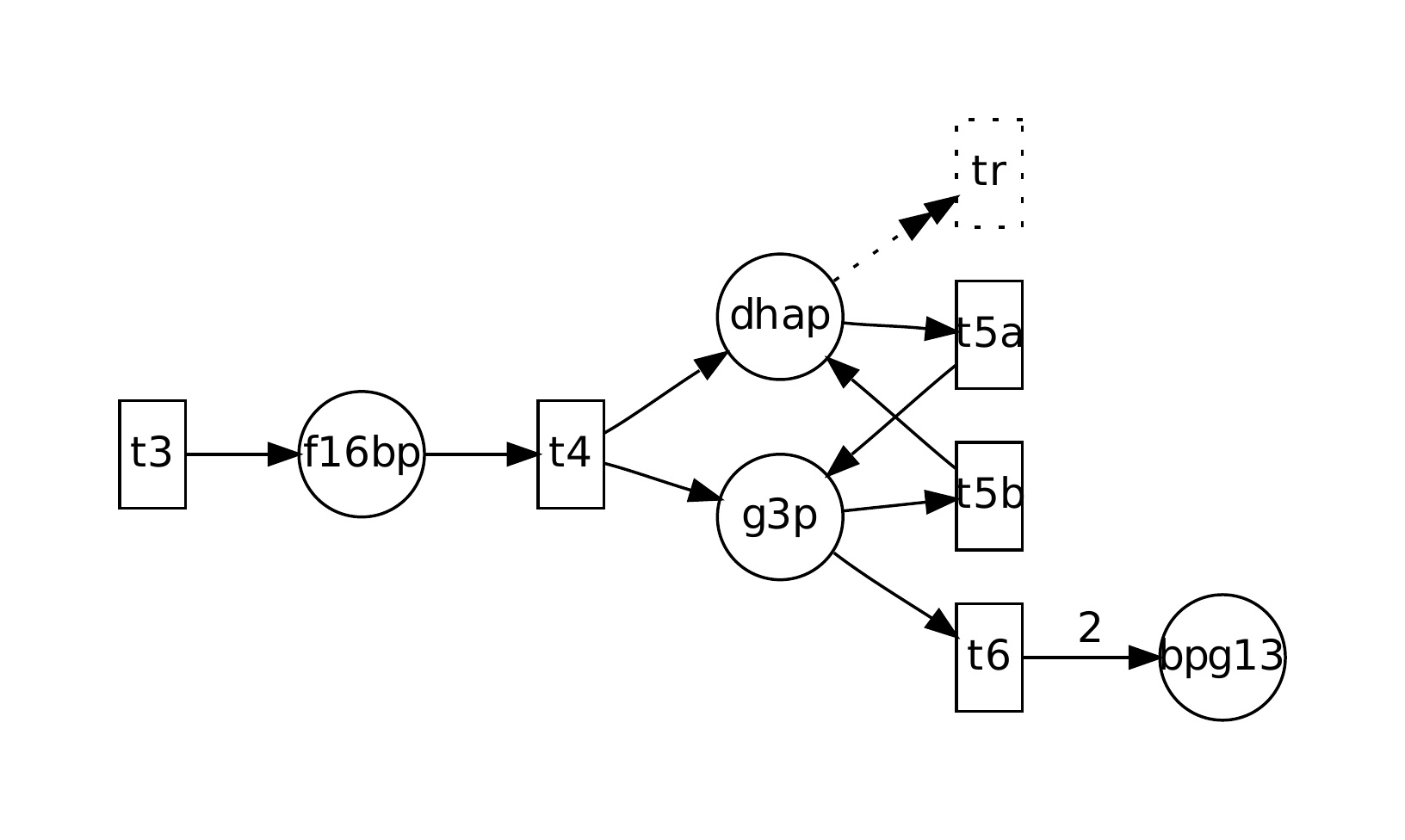}
\caption{Minimal version of the Petri Net graph in Figure \ref{fig:q1}. All reactants that do not contribute to the estimation of the rate of glycolysis have been removed.}
\label{fig:q1:min}
\end{center}
\end{figure}

Simulating it for 10 steps with the same initial marking as the Petri Net in Figure \ref{fig:q1} produced the same results as for Figure \ref{fig:q1}.

\end{solution}

\section{Determining Conditions Leading to an Observation}
\begin{question}\label{q2}
When and how does the body switch to B oxidation versus glycolysis as the major way of burning fuel?
\end{question}

Provided Answer: 
\begin{quote}
``The relative volumes of the raw materials for B oxidation and glycolysis indicate which of these two processes will occur. Glycolysis uses the raw material glucose, and B oxidation uses Acyl CoA from fatty acids. When the blood sugar level decreases below its homeostatic level, then B oxidation will occur with available fatty acids. If no fatty acids are immediately available, glucagon and other hormones regulate release of stored sugar and fat, or even catabolism of proteins and nucleic acids, to be used as energy sources.''
\end{quote}

\begin{solution}
The answer provided requires background knowledge about the mechanism that regulates which source of energy will be used. This information is not presented in Chapter 9 of Campbell's book, which is the source material of this exercise. However, we can model it based on background information combined with Figure 9.19 of Campbell's book. Our model is presented in Figure \ref{fig:q2}\footnote{We can extend this model by adding expressions to inhibition arcs that compare available substances.}.

\begin{figure}[H]
\begin{center}
\includegraphics[width=7cm]{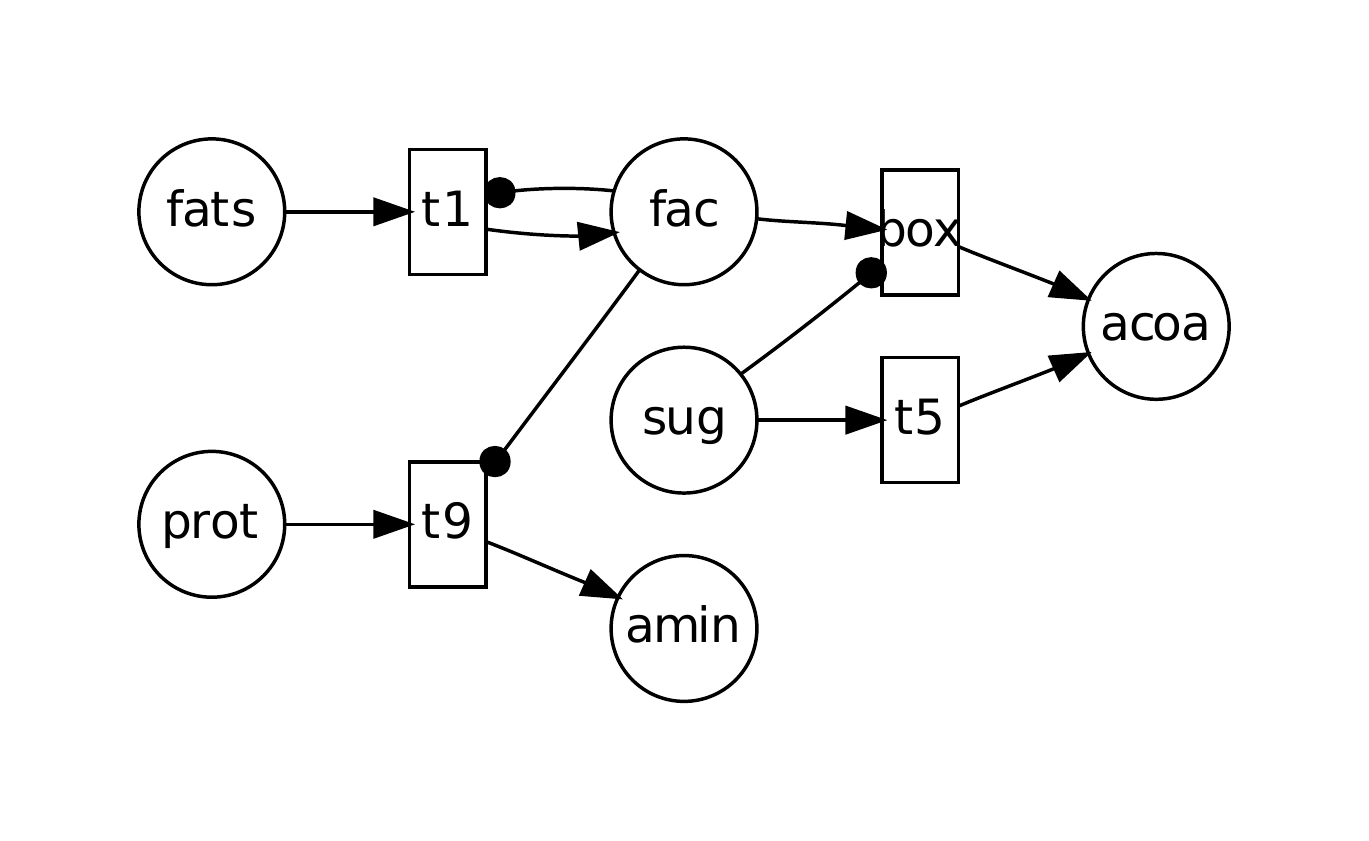}
\caption{Petri Net graph relevant to question \ref{q2}. ``fats'' are fats, ``prot'' are proteins, ``fac'' are fatty acids, ``sug'' are sugars, ``amin'' are amino acids and ``acoa'' is ACoA. Transition ``box'' is the beta oxidation, ``t5'' is glycolysis, ``t1'' is fat digestion into fatty acids, and ``t9'' is protein deamination.}
\label{fig:q2}
\end{center}
\end{figure}

We can test the model by simulating it for a time period and testing whether beta oxidation (box) is started when sugar (sug) is finished. We do not need a steady supply of sugar in this case, just enough to be consumed in a few time steps to capture the switch over. Fats and proteins may or may not modeled as a steady supply, since their bioavailability is dependent upon a number of external factors. We assume a steady supply of both and model it with large enough initial quantity that will last beyond the simulation period.

We translate the petri net model into ASP and run it for 10 iterations. Following are the results:
{\footnotesize
\begin{verbatim}
holds(acoa,0,0) holds(amin,0,0) holds(fac,0,0) holds(fats,5,0) 
holds(prot,3,0) holds(sug,4,0)

fires(t1,0) fires(t5,0) fires(t9,0)

holds(acoa,1,1) holds(amin,1,1) holds(fac,1,1) holds(fats,4,1) 
holds(prot,3,1) holds(sug,3,1)

fires(t5,1)

holds(acoa,2,2) holds(amin,1,2) holds(fac,1,2) holds(fats,4,2) 
holds(prot,3,2) holds(sug,2,2)

fires(t5,2)

holds(acoa,3,3) holds(amin,1,3) holds(fac,1,3) holds(fats,4,3) 
holds(prot,3,3) holds(sug,1,3)

fires(t5,3)

holds(acoa,4,4) holds(amin,1,4) holds(fac,1,4) holds(fats,4,4) 
holds(prot,3,4) holds(sug,0,4)

fires(box,4)

holds(acoa,5,5) holds(amin,1,5) holds(fac,0,5) holds(fats,4,5) 
holds(prot,3,5) holds(sug,0,5)

fires(t1,5) fires(t9,5)

holds(acoa,5,6) holds(amin,2,6) holds(fac,1,6) holds(fats,3,6) 
holds(prot,3,6) holds(sug,0,6)

fires(box,6)c

holds(acoa,6,7) holds(amin,2,7) holds(fac,0,7) holds(fats,3,7) 
holds(prot,3,7) holds(sug,0,7)

fires(t1,7) fires(t9,7)

holds(acoa,6,8) holds(amin,3,8) holds(fac,1,8) holds(fats,2,8) 
holds(prot,3,8) holds(sug,0,8)

fires(box,8)

holds(acoa,7,9) holds(amin,3,9) holds(fac,0,9) holds(fats,2,9) 
holds(prot,3,9) holds(sug,0,9)

fires(t1,9) fires(t9,9)

holds(acoa,7,10) holds(amin,4,10) holds(fac,1,10) holds(fats,1,10) 
holds(prot,3,10) holds(sug,0,10)

fires(box,10)
\end{verbatim}
}

We can see that by time-step 4, the sugar supply is depleted and beta oxidation starts occurring.

\end{solution}

\section{Comparing Altered Trajectories due to Accumulation Intervention}
\begin{question}\label{q3}
ATP is accumulating in the cell. What affect would this have on the rate of glycolysis? Explain.
\end{question}

Provided Answer:
\begin{quote}
``ATP and AMP regulate the activity of phosphofructokinase. When there is an abundance of AMP in the cell, this indicates that the rate of ATP consumption is high. The cell is in need for more ATP. If ATP is accumulating in the cell, this indicates that the cell's demand for ATP had decreased. The cell can decrease its production of ATP. Therefore, the rate of glycolysis will decrease.''
\end{quote}

\begin{solution}
Control of cellular respiration is summarized in Fig 9.20 of Campbell's book. We can gauge the rate of glycolysis by the amount of Pyruvate produced, which is the end product of glycolysis. We assume a steady supply of glucose is available. Its availability is not impacted by any of the feedback mechanism depicted in Fig 9.20 of Campbell's book or restricted by the question. We can ignore the respiration steps after glycolysis, since they are directly dependent upon the end product of glycolysis, i.e. Pyruvate. These steps only reinforce the negative effect of ATP. The Citrate feed-back shown in Campbell's Fig 9.20 is also not relevant to the question, so we can assume a constant level of it and leave it out of the picture. Another simplification that we do is to treat the inhibition of Phosphofructokinase (PFK) by ATP as the inhibition of glycolysis itself. This is justified, since PFK is on a linear path from Glucose to Fructose 1,6-bisphosphate (F16BP), and all downstream product quantities are directly dependent upon the amount of F16BP (as shown in Campbell's Fig 9.9), given steady supply of substances involved in glycolysis. Our assumption also applies to ATP consumed in Fig 9.9. Our simplified picture is shown in Figure \ref{fig:q3} as a Petri Net.

\begin{figure}[H]
\begin{center}
\includegraphics[width=7cm]{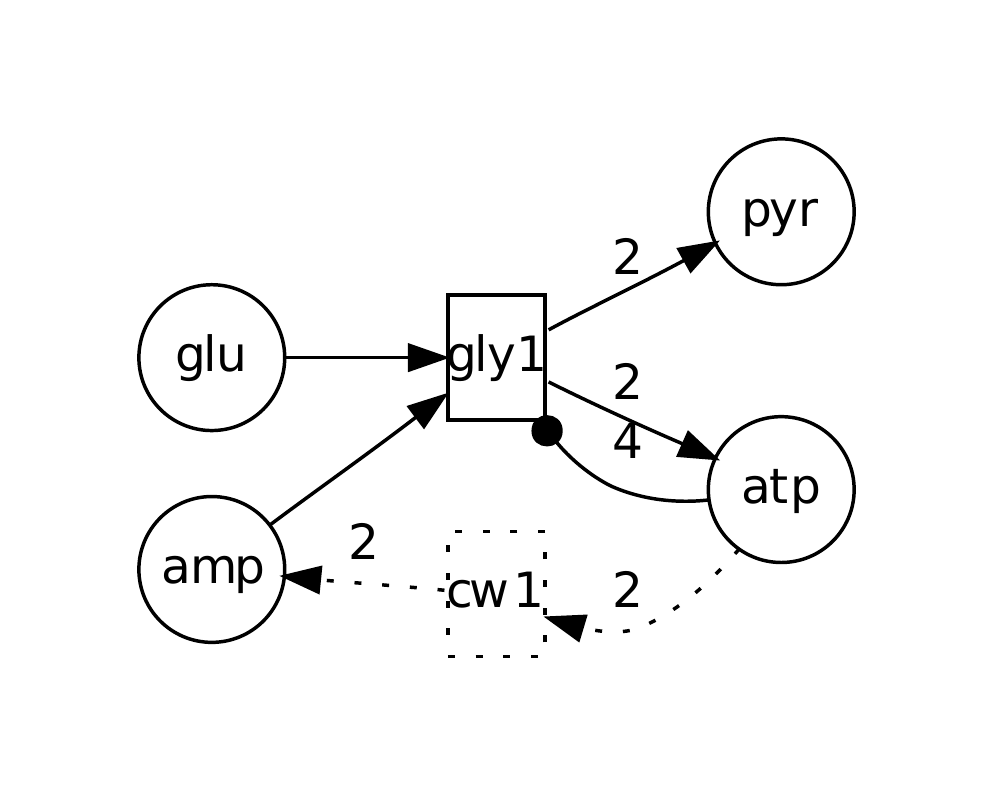}
\caption{Petri Net graph relevant to question \ref{q3}. ``glu'' is Glucose, ``pyr'' is Pyruvate. Transitions ``gly1'' represents glycolysis and ``cw1'' is cellular work that consumes ATP and produces AMP. Transition ``gly1'' is inhibited only when the number of $atp$ tokens is greater than 4.}
\label{fig:q3}
\end{center}
\end{figure}

We model cellular work that recycles ATP to AMP (see p/181 of Campbell's book) by the $cw1$ transition, shown in dotted style. In normal circumstances, this arc does not let ATP to collect. If we reduce the arc-weights incident on $cw1$ to 1, we get the situation where less work is being done and some ATP will collect, as a result glycolysis will pause and resume. If we remove $cw1$ (representing no cellular work), ATP will start accumulating and glycolysis will stop. We use an arbitrary arc-weight of $4$ on the inhibition arc $(atp,gly1)$ to model an elevated level of ATP beyond normal that would cause inhibition \footnote{An alternate modeling would be compare the number of tokens on the $amp$ node and the $atp$ node and set a level-threshold that inhibits $gly1$. Such technique is common in colored-peri nets.}. We encode all three situations in ASP with maximal firing set policy. We run them for 10 steps and compare the quantity of pyruvate produced to determine the difference in the rate of glycolysis.

In normal situation when cellular work is being performed ($cw1$ arc is present), unique quantities of ``pyr'' after 10 step are as follows:
{\footnotesize
\begin{verbatim}
holds(pyr,20,10)
\end{verbatim}
}
when the cellular work is reduced, i.e. ($(atp,cw1)$, $(cw1,amp)$ arc weights changed to 1), unique quantities of ``pyr'' after 10 steps are as follows:
{\footnotesize
\begin{verbatim}
holds(pyr,14,10)
\end{verbatim}
}
with no cellular work ($cw1$ arc removed), unique quantities of ``pyr'' after 10 steps are as follows:
{\footnotesize
\begin{verbatim}
holds(pyr,6,10)
\end{verbatim}
}

The results show the rate of glycolysis reducing as the cellular work decreases to the point where it stops once ATP reaches the inhibition threshold. Higher numbers of ATP produced in later steps of cellular respiration will reinforce this inhibition even more quickly. Trend of answers from various runs is shown in Figure \ref{fig:q3:runs}.

\begin{figure}[H]
\begin{center}
\includegraphics[width=10cm]{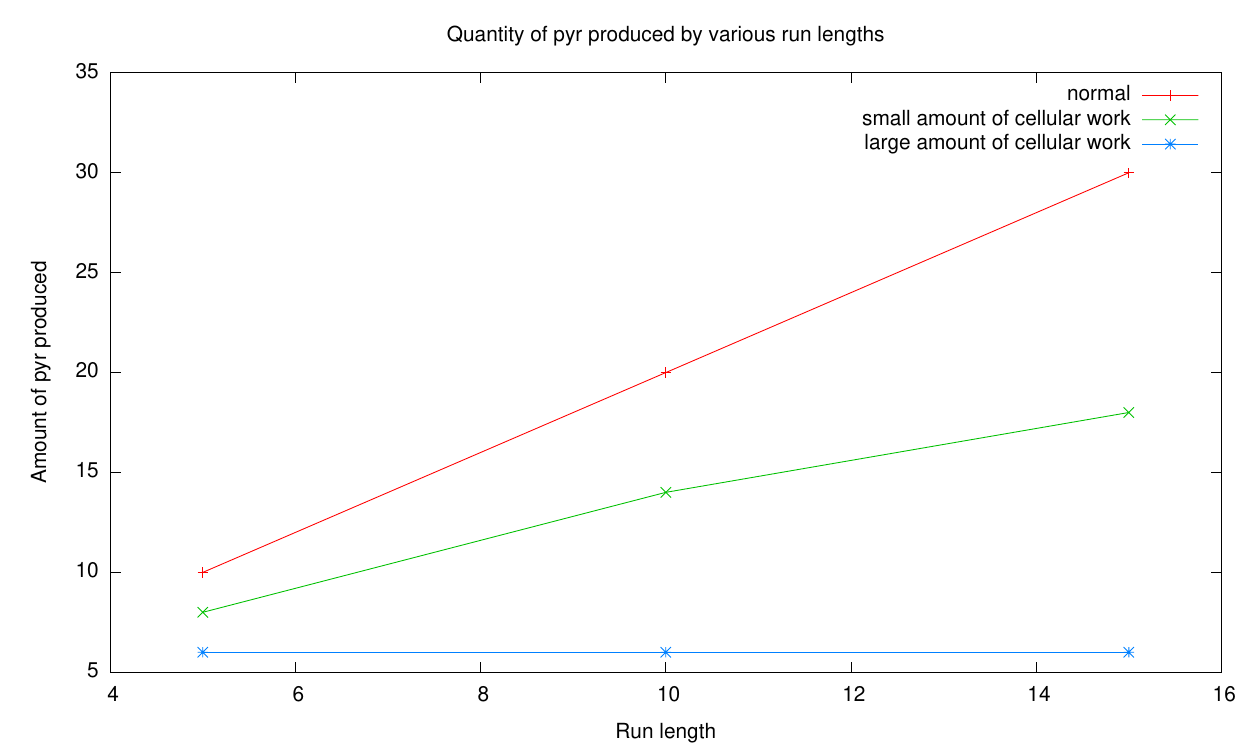}
\caption{Amount of pyruvate produced from various lengths of runs.}
\label{fig:q3:runs}
\end{center}
\end{figure}

\end{solution}

\section{Comparing Altered Trajectories due to Initial Value Intervention}
\begin{question}\label{q4}
A muscle cell had used up its supply of oxygen and ATP. Explain what affect would this have on the rate of cellular respiration and glycolysis?
\end{question}

Provided Answer:
\begin{quote}
``Oxygen is needed for cellular respiration to occur. Therefore, cellular respiration would stop. The cell would generate ATP by glycolysis only. Decrease in the concentration of ATP in the cell would stimulate an increased rate of glycolysis in order to produce more ATP.''
\end{quote}

\begin{solution}
Figure 9.18 of Campbell's book gives the general idea of what happens when oxygen is not present. Figure 9.20 of Campbell's book shows the control of glycolysis by ATP. To formulate the answer, we need pieces from both. 

ATP inhibits Phosphofructokinase (Fig 9.20 of Campbell), which is an enzyme used in glycolysis. No ATP means that enzyme is no longer inhibited and glycolysis can proceed at full throttle. Pyruvate either goes through aerobic respiration when oxygen is present or it goes through fermentation when oxygen is absent (Fig 9.18 of Campbell). We can monitor the rate of glycolysis and cellular respiration by observing these operations occurring (by looking at corresponding transition firing) over a simulation time period. Our simplified Petri Net model is shown in Figure \ref{fig:q4}.

We ignore the details of processes following glycolysis, except that these steps produce additional ATP. We do not need an exact number of ATP produced as long as we keep it higher than the ATP produced by glycolysis. Higher numbers will just have a higher negative feed-back (or inhibition) effect on glycolysis. We ignore citrate's inhibition of glycolysis since that is not relevant to the question and since it gets recycled by the citric acid cycle (see Fig 9.12 of Campbell). We also ignore AMP, since it is not relevant to the question, by assuming sufficient supply to maintain glycolysis. We also assume continuous cellular work consuming ATP, without that ATP will accumulate almost immediately and stop glycolysis.

We assume a steady supply of glucose is available to carry out glycolysis and fulfill this requirement by having a quantity in excess of the consumption during our simulation interval. All other substances participating in glycolysis are assumed to be available in a steady supply so that glycolysis can continue.

\begin{figure}[H]
\begin{center}
\includegraphics[width=10cm]{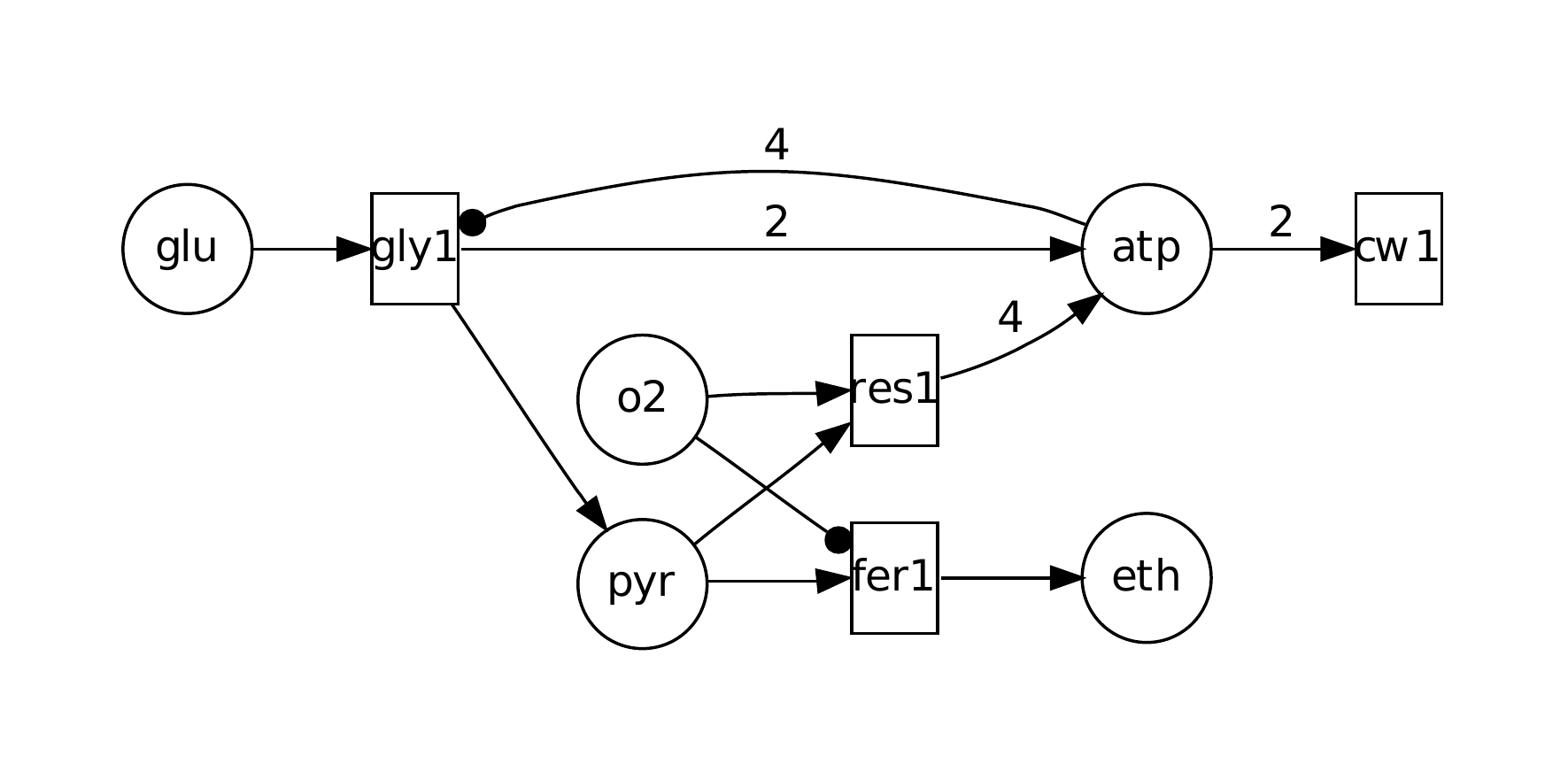}
\caption{Petri Net graph relevant to question \ref{q4}. ``glu'' is Glucose, ``pyr'' is Pyruvate, ``atp'' is ATP, ``eth'' is ethenol or other products of fermentation, and ``o2'' is Oxygen. Transitions ``gly1'' represents glycolysis, ``res1'' is respiration in presence of oxygen, ``fer1'' is fermentation when no oxygen is present, and ``cw1'' is cellular work that consumes ATP. Transition ``gly1'' is inhibited only when the number of $atp$ tokens is greater than 4.}
\label{fig:q4}
\end{center}
\end{figure}

We then consider two scenarios, one where oxygen is present and where oxygen is absent and determine the change in rate of glycolysis and respiration by counting the firings of their respective transitions. We encode both situations in ASP with maximal firing set policy. Both situations are executed for 10 steps. At the end of those steps the firing count of ``gly1'' and ``res1'' is computed and compared to determine the difference in the rates of glycolysis and respiration respectively.

In the normal situation (when oxygen is present), we get the following answer sets:
{\footnotesize
\begin{verbatim}
fires(gly1,0)
fires(cw1,1) fires(gly1,1) fires(res1,1)
fires(cw1,2) fires(res1,2)
fires(cw1,3)
fires(cw1,4)
fires(cw1,5) fires(gly1,5)
fires(cw1,6) fires(gly1,6) fires(res1,6)
fires(cw1,7) fires(res1,7)
fires(cw1,8)
fires(cw1,9)
fires(cw1,10)
\end{verbatim}
}

while in the abnormal situation (when oxygen is absent), we get the following firings:
{\footnotesize
\begin{verbatim}
fires(gly1,0)
fires(cw1,1) fires(fer1,1) fires(gly1,1)
fires(cw1,2) fires(fer1,2) fires(gly1,2)
fires(cw1,3) fires(fer1,3) fires(gly1,3)
fires(cw1,4) fires(fer1,4) fires(gly1,4)
fires(cw1,5) fires(fer1,5) fires(gly1,5)
fires(cw1,6) fires(fer1,6) fires(gly1,6)
fires(cw1,7) fires(fer1,7) fires(gly1,7)
fires(cw1,8) fires(fer1,8) fires(gly1,8)
fires(cw1,9) fires(fer1,9) fires(gly1,9)
fires(cw1,10) fires(fer1,10) fires(gly1,10)
\end{verbatim}
}

Note that the number of firings of glycolysis for normal situation is lower when oxygen is present and higher when oxygen is absent. While, the number of firings is zero when no oxygen is present. Thus, respiration stops when no oxygen is present and the need of ATP by cellular work is fulfilled by a higher amount of glycolysis. Trend from various runs is shown in Figure \ref{fig:q4:runs}.

\begin{figure}[H]
\begin{center}
\includegraphics[width=12cm]{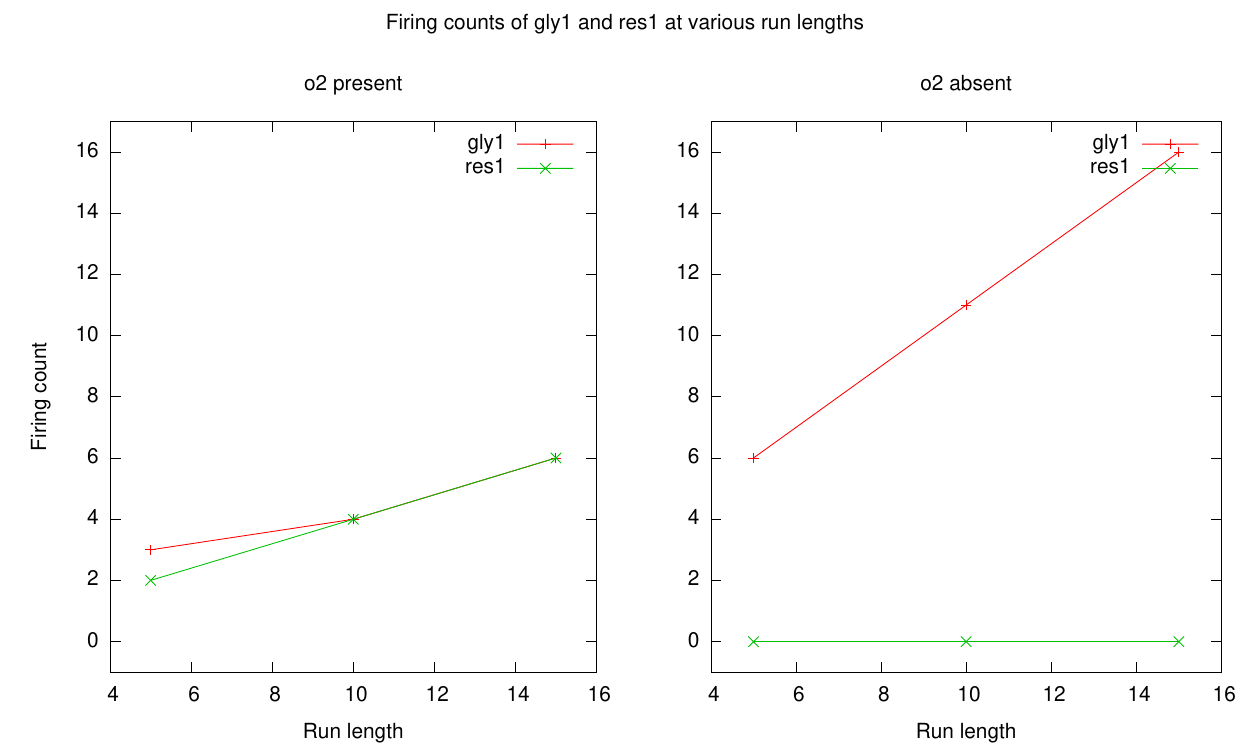}
\caption{Firing counts of glycolysis (gly1) and respiration (res1) for different simulation lengths for the petri net in Figure \ref{fig:q4}}
\label{fig:q4:runs}
\end{center}
\end{figure}

\end{solution}

\section{Comparing Altered Trajectories due to Inhibition Intervention}
\begin{question}\label{q5}
The final protein complex in the electron transport chain of the mitochondria is non-functional. Explain the effect of this on pH of the intermembrane space of the mitochondria.
\end{question}

Provided Answer:
\begin{quote}
``The H+ ion gradient would gradually decrease and the pH would gradually increase. The other proteins in the chain are still able to produce the H+ ion gradient. However, a non-functional, final protein in the electron transport chain would mean that oxygen is not shuttling electrons away from the electron transport chain. This would cause a backup in the chain, and the other proteins in the electron transport chain would no longer be able to accept electrons and pump H+ ions into the intermembrane space. A concentration decrease in the H+ ions means an increase in the pH.''
\end{quote}

\begin{solution}
The electron transport chain is shown in Fig 9.15 (1) of Campbell's book. In order to explain the effect on pH, we will show the change in the execution of the electron transport chain with both a functioning and non-functioning final protein. Since pH depends upon the concentration of H+ ions, we will quantify the difference its quantity in the intermembrane space in both scenarios as well. We assume that a steady input of NADH, FADH2, H+ and O2 are available in the mitochondrial matrix. We also assume an electron carrying capacity of 2 for both ubiquinone (Q)\footnote{http://www.benbest.com/nutrceut/CoEnzymeQ.html} and cytochrome c (Cyt c). This carrying capacity is background information not provided in Campbell's Chapter 9. As with previous questions, we fulfill the steady supply requirement of substances by having input quantities in excess of what would be consumed during our simulation interval.

\begin{figure}[H]
\begin{center}
\includegraphics[width=12cm]{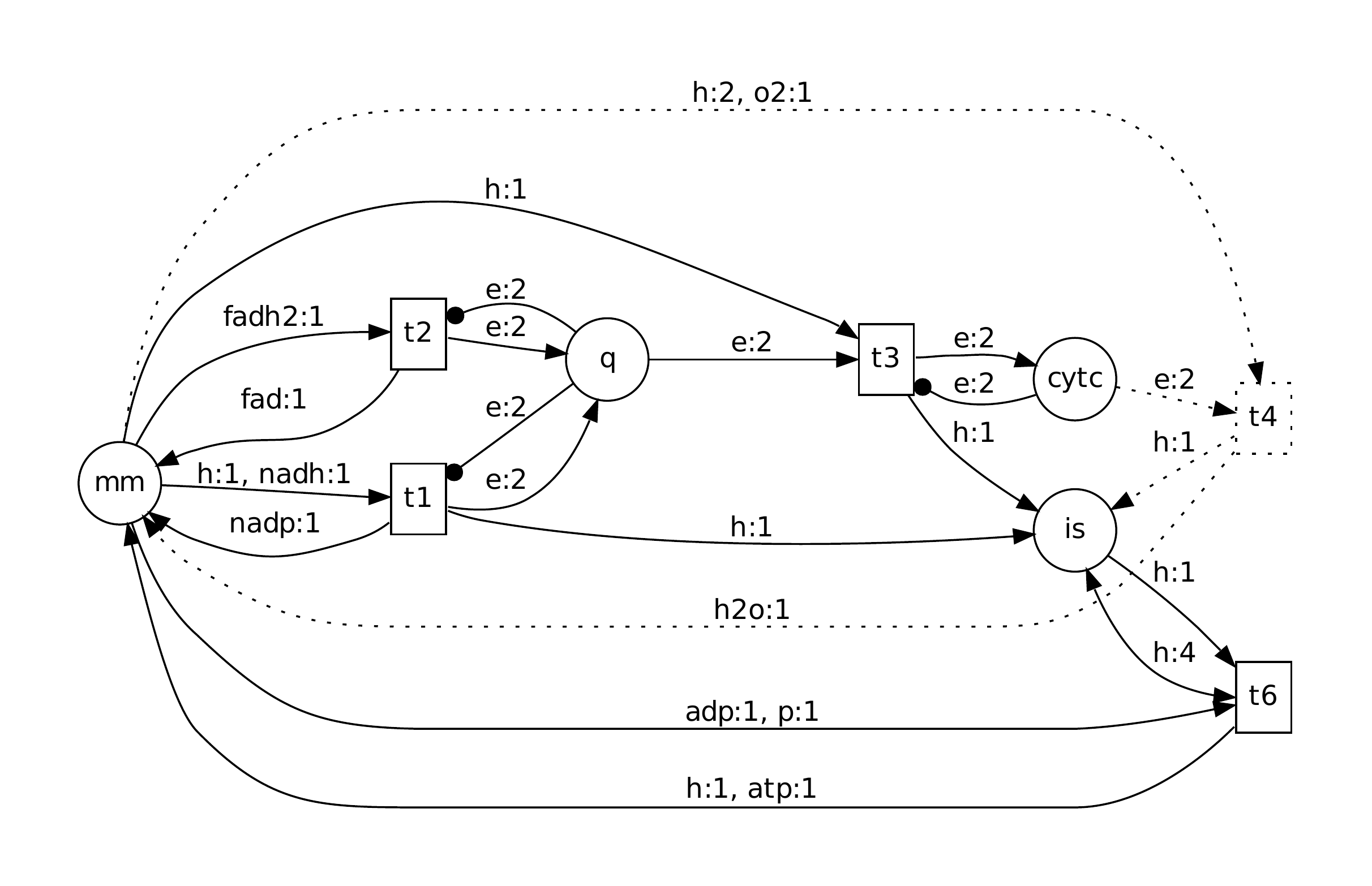}
\caption{Petri Net graph relevant to question \ref{q5}. ``is'' is the intermembrane space, ``mm'' is mitochondrial matrix, ``q'' is ubiquinone and ``cytc'' is cytochrome c. The inhibition arcs $(q,t1)$,  $(q,t2)$ and $(cytc,t3)$ capture the electron carrying capacities of $q$ and $cytc$. Over capacity will cause backup in electron transport chain. Tokens are colored, e.g. ``h:1, nadh:1'' specify one token of $h$ and $nadh$ each. Token types are ``h'' for H+, ``nadh'' for NADH, ``nadp'' for NADP, ``fadh2'' for FADH2, ``fad'' for FAD, ``e'' for electrons, ``o2'' for oxygen and ``h2o'' for water. We remove $t4$ to model non-functioning protein complex $IV$.}
\label{fig:q5}
\end{center}
\end{figure}

We model this problem as a colored petri net shown in Figure \ref{fig:q5}. The normal situation is made up of the entire graph. The abnormal situation (with non-functional final protein complex) is modeled by removing transition $t4$ from the graph\footnote{Alternatively, we can model a non-functioning transition by attaching an inhibition arc to it with one token at its source place}. We encode both situations in ASP with maximal firing set policy. Both are run for 10 steps. The amount of $h$ (H+) is compared in the $is$ (intermembrane space) to determine change in pH and the firing sequence is compared to explain the effect.

In normal situation (entire graph), we get the following:
{\footnotesize
\begin{verbatim}
fires(t1,0) fires(t2,0)
fires(t3,1)
fires(t1,2) fires(t2,2) fires(t3,2) fires(t4,2)
fires(t3,3) fires(t4,3) fires(t6,3)
fires(t1,4) fires(t2,4) fires(t3,4) fires(t4,4) fires(t6,4)
fires(t3,5) fires(t4,5) fires(t6,5)
fires(t1,6) fires(t2,6) fires(t3,6) fires(t4,6) fires(t6,6)
fires(t3,7) fires(t4,7) fires(t6,7)
fires(t1,8) fires(t2,8) fires(t3,8) fires(t4,8) fires(t6,8)
fires(t3,9) fires(t4,9) fires(t6,9)
fires(t1,10) fires(t2,10) fires(t3,10) fires(t4,10) fires(t6,10)

holds(is,15,h,10)
\end{verbatim}
}

with $t4$ removed, we get the following:
{\footnotesize
\begin{verbatim}
fires(t1,0) fires(t2,0)
fires(t3,1)
fires(t1,2) fires(t2,2) fires(t3,2)
fires(t6,3)
fires(t6,4)

holds(is,2,h,10)
\end{verbatim}
}

Note that the amount of H+ ($h$) produced in the intermembrane space ($is$) is much smaller when the final protein complex is non-functional ($t4$ removed). Lower H+ translates to higher pH. Thus, the pH of intermembrane space will increase as a result of nonfunctional final protein. Also, note that the firing of $t3$, $t1$ and $t2$ responsible for shuttling electrons also stop very quickly when $t4$ no longer removes the electrons ($e$) from Cyt c ($cytc$) to produce $H_2O$. This is because $cytc$ and $q$ are at their capacity on electrons that they can carry and stop the electron transport chain by inhibiting transitions $t3$, $t2$ and $t1$. Trend for various runs is shown in Figure \ref{fig:q5:runs}.

\begin{figure}[H]
\begin{center}
\includegraphics[width=12cm]{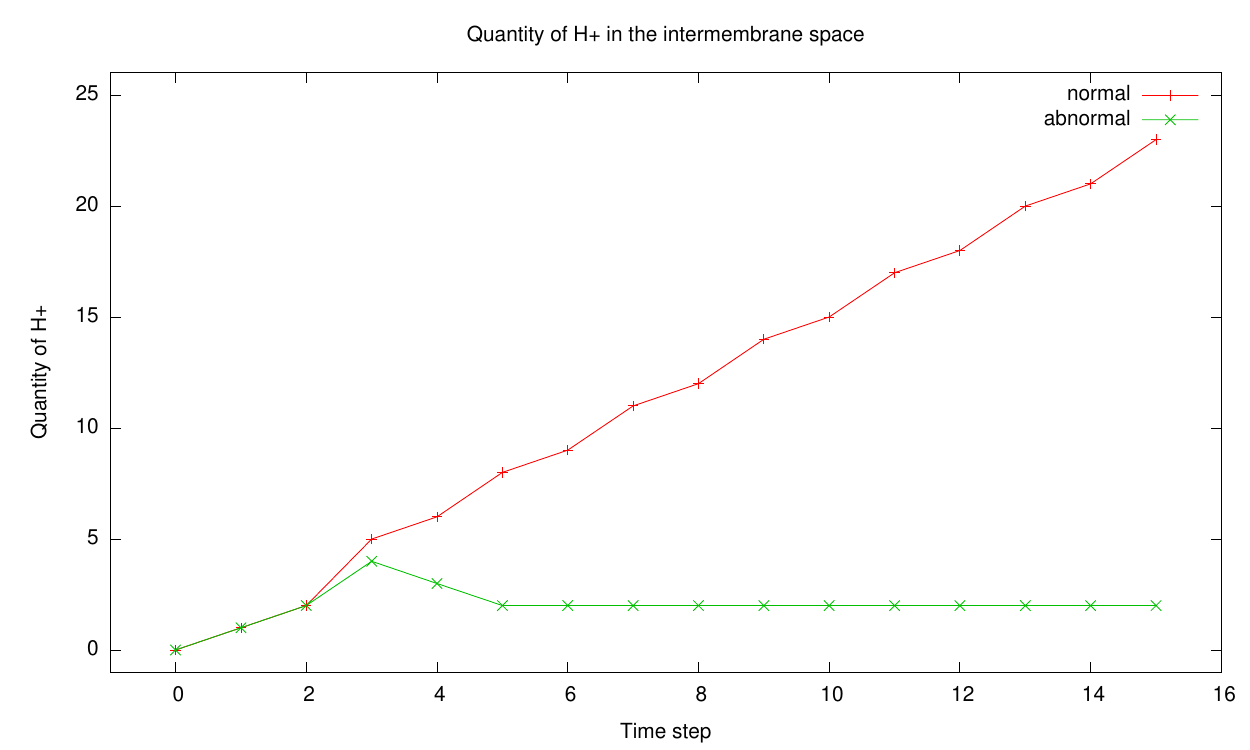}
\caption{Simulation of Petri Net in Figure \ref{fig:q5}. In a complete model of the biological system, there will be a mechanism that keeps the quantity of H+ in check in the intermembrane space and will plateau at some point.}
\label{fig:q5:runs}
\end{center}
\end{figure}

\end{solution}

\section{Comparing Altered Trajectories due to Gradient Equilization Intervention}
\begin{question}\label{q6}
Exposure to a toxin caused the membranes to become permeable to ions. In a mitochondrion, how would this affect the pH in the intermembrane space and also ATP production?
\end{question}

Provided Answer: 
\begin{quote}
``The pH of the intermembrane space would decrease as H+ ions diffuse through the membrane, and the H+ ion gradient is lost. The H+ gradient is essential in ATP production b/c facilitated diffusion of H+ through ATP synthase drives ATP synthesis. Decreasing the pH would lead to a decrease in the rate of diffusion through ATP synthase and therefore a decrease in the production of ATP.''
\end{quote}

\begin{solution}
Oxidative phosphorylation is shown in Fig 9.15 of Campbell's book. In order to explain the effect on pH in the intermembrane space and the ATP production we will show the change in the amount of H+ ions in the intermembrane space as well as the amount of ATP produced when the inner mitochondrial membrane is impermeable and permeable. Note that the concentration of H+ determines the pH. we have chosen to simplify the diagram by not having FADH2 in the picture. Its removal does not change the response, since it provides an alternate input mechanism to electron transport chain. We will assume that a steady input of NADH, H+, O2, ADP and P is available in the mitochondrial matrix. We also assume an electron carrying capacity of 2 for both ubiquinone (Q) and cytochrome c (Cyt c). We fulfill the steady supply requirement of substances by having input quantities in excess of what would be consumed during our simulation interval.

We model this problem as a colored petri net shown in Figure \ref{fig:q6}. Transition $t6,t7$ shown in dotted style are added to model the abnormal situation\footnote{If reverse permeability is also desired additional arcs may be added from mm to is}. They capture the diffusion of H+ ions back from Intermembrane Space to the Mitochondrial matrix. One or both may be enabled to capture degrees of permeability. we have added a condition on the firing of $t5$ (ATP Synthase activation) to enforce gradient to pump ATP Synthase.

\begin{figure}[H]
\begin{center}
\includegraphics[width=12cm]{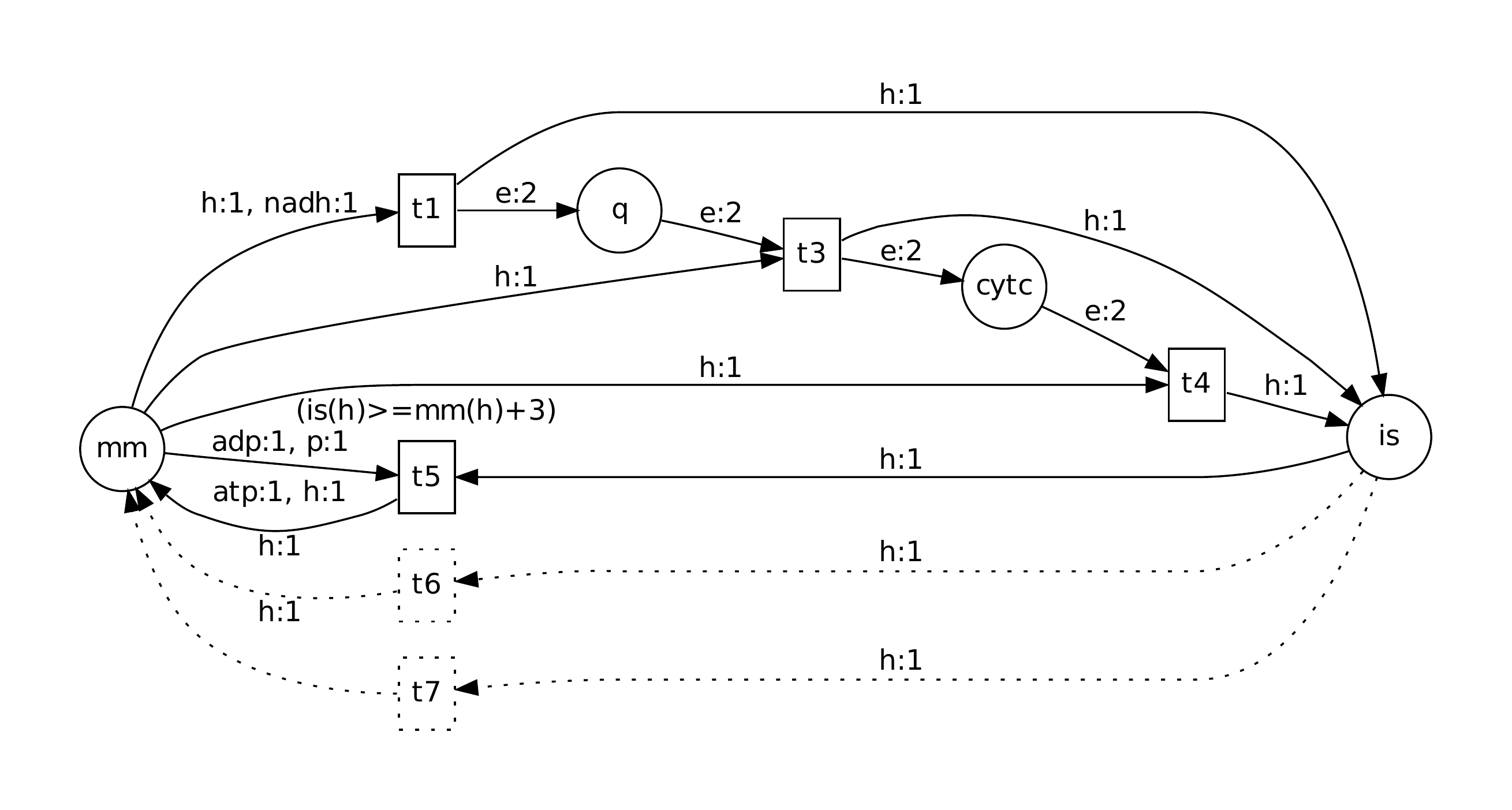}
\caption{Petri Net graph relevant to question \ref{q6}. ``is'' is the intermembrane space, ``mm'' is mitochondrial matrix, ``q'' is ubiquinone and ``cytc'' is cytochrome c. Tokens are colored, e.g. ``h:1, nadh:1'' specify one token of $h$ and $nadh$ each. Token types are ``h'' for H+, ``nadh'' for NADH, ``e'' for electrons, ``o2'' for oxygen, ``h2o'' for water, ``atp'' for ATP and ``adp'' for ADP. We add $t6,t7$ to model cross domain diffusion from intermembrane space to mitochondrial matrix. One or both of $t6,t7$ may be enabled at a time to control the degree of permeability. The text above ``t5'' is an additional condition which must be satisfied for ``t5'' to be enabled.}
\label{fig:q6}
\end{center}
\end{figure}

We encode both situations in ASP with maximal firing set policy. Both are run for 10 steps each and the amount of $h$ and $atp$ is compared to determine the effect of pH and ATP production. We capture the gradient requirement as the following ASP code\footnote{We can alternatively model this by having a threshold arc from ``is'' to ``t5'' if only a minimum trigger quantity is required in the intermembrane space.}:
\begin{verbatim}
notenabled(T,TS) :- 
  T==t5, C==h, trans(T), col(C), holds(is,Qis,C,TS), 
  holds(mm,Qmm,C,TS), Qmm+3 > Qis, 
  num(Qis;Qmm), time(TS).
\end{verbatim}

In the normal situation, we get the following $h$ token distribution after 10 steps:
{\footnotesize
\begin{verbatim}
holds(is,11,h,10) holds(mm,1,h,10) holds(mm,6,atp,10)
\end{verbatim}
}

we change the permeability to 1 ($t6$ enabled), we get the following token distribution instead:
{\footnotesize
\begin{verbatim}
holds(is,10,h,10) holds(mm,2,h,10) holds(mm,5,atp,10)
\end{verbatim}
}

we change the permeability to 2 ($t6,t7$ enabled), the distribution changes as follows:
{\footnotesize
\begin{verbatim}
holds(is,8,h,10) holds(mm,4,h,10) holds(mm,2,atp,10)
\end{verbatim}
}

Note that as the permeability increases, the amount of H+ ($h$) in intermembrane space ($is$) decreases and so does the amount of ATP ($h$) in mitochondrial matrix. Thus, an increase in permeability will increase the pH. If the permeability increases even beyond 2, no ATP will be produced from ADP due to insufficient H+ gradient. Trend from various runs is shown in Figure \ref{fig:q6:runs}.

\begin{figure}[H]
\begin{center}
\includegraphics[width=12cm]{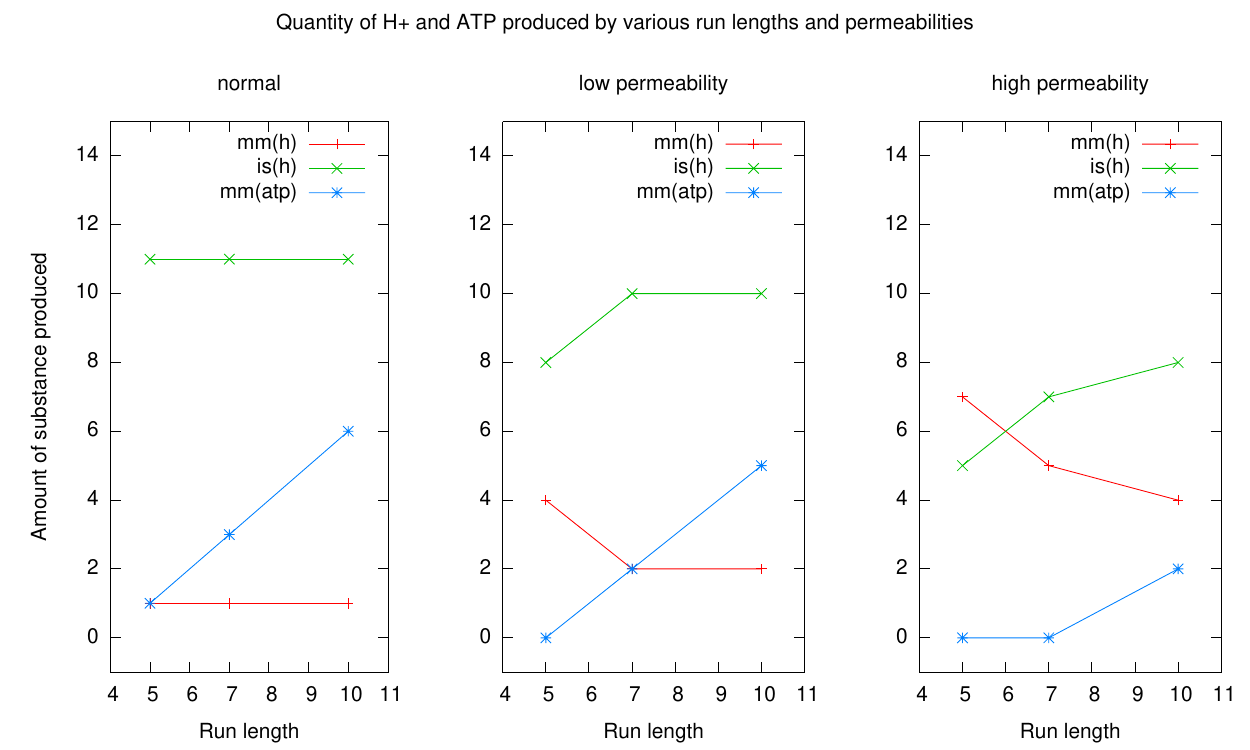}
\caption{Quantities of H+ and ATP at various run lengths and permeabilities for the Petri Net model in Figure \ref{fig:q6}.}
\label{fig:q6:runs}
\end{center}
\end{figure}

\end{solution}

\section{Comparing Altered Trajectories due to Delay Intervention}
\begin{question}\label{q7}
Membranes must be fluid to function properly. How would decreased fluidity of the membrane affect the efficiency of the electron transport chain?
\end{question}

Provided Answer: 
\begin{quote}
``Some of the components of the electron transport chain are mobile electron carriers, which means they must be able to move within the membrane. If fluidity decreases, these movable components would be encumbered and move more slowly. This would cause decreased efficiency of the electron transport chain.''
\end{quote}

\begin{solution}
The answer deals requires background knowledge about fluidity and how it relates to mobile carriers not presented in the source chapter. From background knowledge we find that the higher the fluidity, higher the mobility. The electron transport chain is presented in Fig 9.15 of Campbell's book. From background knowledge, we know that the efficiency of the electron transport chain is measured by the amount of ATP produced per NADH/FADH2. The ATP production happens due to the gradient of H+ ions across the mitochondrial membrane. The higher the number of H+ ions in the intermembrane space, the higher would be the gradient and the resulting efficiency. So we measure the efficiency of the chain by the amount of H+ transported to intermembrane space, assuming all other (fixed) molecules behave normally. This is a valid assumption since H+ transported from mitochondrial matrix is directly proportional to the amount of electrons shuttled through the non-mobile complexes and there is a linear chain from the electron carrier to oxygen.

We model this chain using a Petri Net with durative transitions shown in Figure \ref{fig:q7}. Higher the duration of transitions, lower the fluidity of the membrane. We assume that a steady supply of NADH and H+ is available in the mitochondrial membrane. We fulfill this requirement by having quantities in excess of what will be consumed during the simulation. We ignore FADH2 from the diagram, since it is just an alternate path to the electron chain. Using it by itself will produce a lower number of H+ transporter to intermembrane space, but it will not change the result. We compare the amount of H+ transported into the intermembrane space to gauge the efficiency of the electron transport chain. More efficient the chain is, more H+ will it transport.

We model three scenarios: normal fluidity, low fluidity with transitions $t3$ and $t4$ having an execution time of 2 and an lower fluidity with transitions $t3,t4$ having execution time of 4.

\begin{figure}[H]
\begin{center}
\includegraphics[width=12cm]{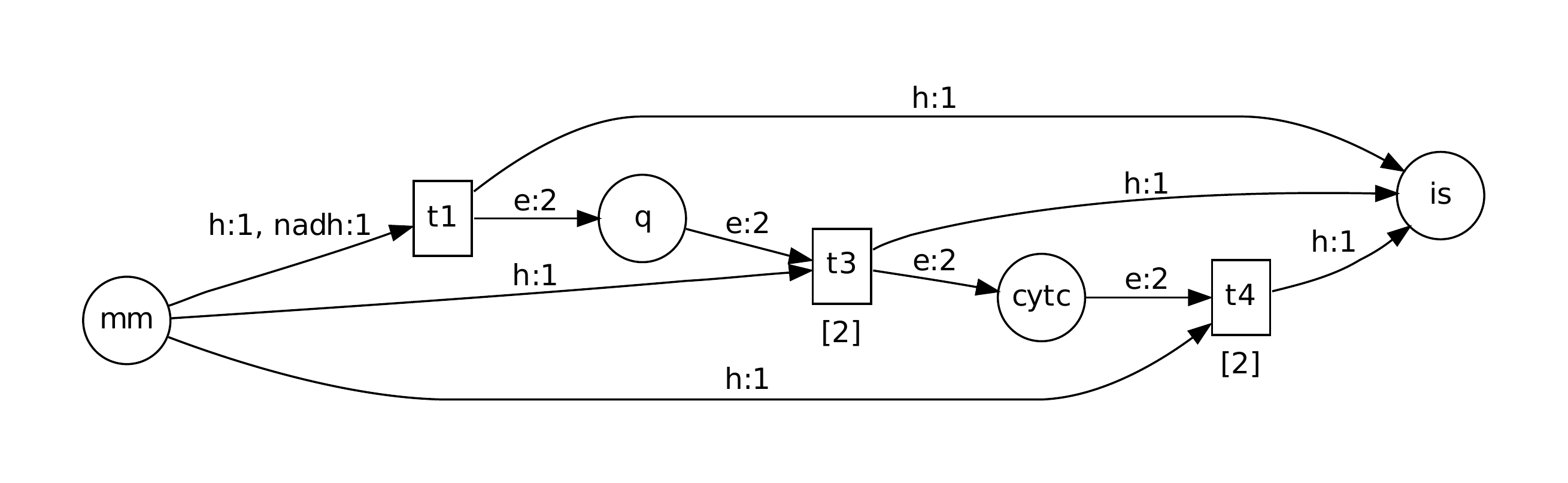}
\caption{Petri Net graph relevant to question \ref{q7}. ``is'' is the intermembrane space, ``mm'' is mitochondrial matrix, ``q'' is ubiquinone and ``cytc'' is cytochrome c. Tokens are colored, e.g. ``h:1, nadh:1'' specify one token of $h$ and $nadh$ each. Token types are ``h'' for H+, ``nadh'' for NADH, ``e'' for electrons. Numbers in square brackets below the transition represent transition durations with default of one time unit, if the number is missing.}
\label{fig:q7}
\end{center}
\end{figure}

We encode these cases in ASP with maximal firing set semantics and simulate them for 10 time steps. For the normal fluidity we get:
{\footnotesize
\begin{verbatim}
holds(is,27,h,10)
\end{verbatim}
}

for low fluidity we get:
{\footnotesize
\begin{verbatim}
holds(is,24,h,10)
\end{verbatim}
}

for lower fluidity we get:
{\footnotesize
\begin{verbatim}
holds(is,18,h,10)
\end{verbatim}
}

Note that as the fluidity decreases, so does the amount of H+ transported to intermembrane space, pointing to lower efficiency of electron transport chain. Trend of various runs is shown in Figure \ref{fig:q7:runs}.

\begin{figure}[H]
\begin{center}
\includegraphics[width=12cm]{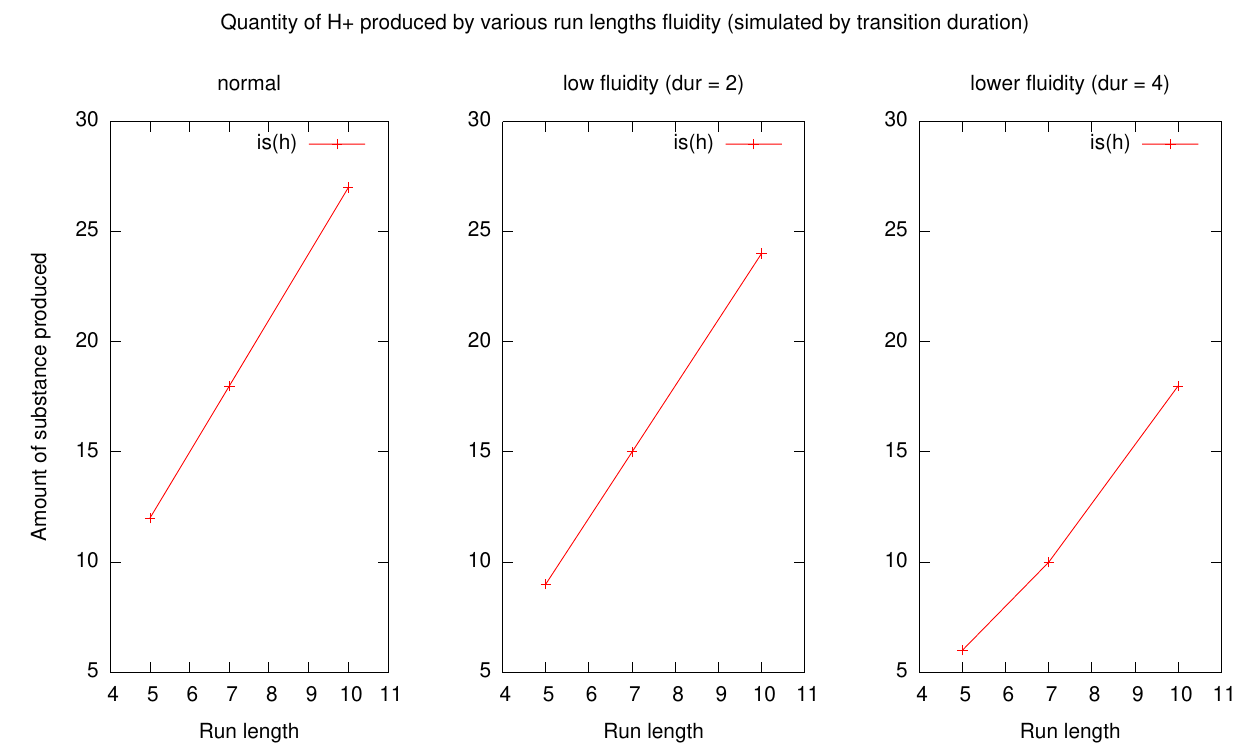}
\caption{Quantities of H+ produced in the intermembrane space at various run lengths and fluidities for the Petri Net model in Figure \ref{fig:q7}.}
\label{fig:q7:runs}
\end{center}
\end{figure}

\end{solution}

\section{Comparing Altered Trajectories due to Priority and Read Interventions}
\begin{question}\label{q8}
Phosphofructokinase (PFK) is allosterically regulated by ATP. Considering the result of glycolysis, is the allosteric regulation of PFK likely to increase or decrease the rate of activity for this enzyme?
\end{question}

Provided Answer: 
\begin{quote}
``Considering that one of the end products of glycolysis is ATP, PFK is inhibited when ATP is abundant and bound to the enzyme. The inhibition decreases ATP production along this pathway.''
\end{quote}

\begin{solution}
Regulation of Phosphofructokinase (PFK) is presented in Figure 9.20 of Campbell's book. We ignore substances upstream of Fructose 6-phosphate (F6P) by assuming they are available in abundance. We also ignore AMP by assuming normal supply of it. We also ignore any output of glycolysis other than ATP production since the downstream processes ultimately produce additional ATP. Citric acid is also ignored since it is not relevant to the question at hand. We monitor the rate of activity of PFK by the number of times it gets used for glycolysis.

We model this problem as a Petri Net shown in Figure \ref{fig:q8}. Allosteric regulation of PFK is modeled by a compound ``pfkatp'' which represents PFK's binding with ATP to form a compound. Details of allosteric regulation are not provided in the same chapter, they are background knowledge from external sources. Higher than normal quantity of ATP is modeled by a threshold arc (shown with arrow-heads at both ends) with an arbitrary threshold value of 4. This number can be increased as necessary. The output of glycolysis and down stream processes ``t3'' has been set to 2 to run the simulation in a reasonable amount of time. It can be made larger as necessary. The allosteric regulation transition ``t4'' has also been given a higher priority than glycolysis transition ``t3''. This way, ATP in excess will cause PFK to be converted to PFK+ATP compound, reducing action of PFK.

We assume that F6P is available in sufficient quantity and so is PFK. This requirement is fulfilled by having more quantity than can be consumed in the simulation duration. We model both the normal situation including transition $t4$ shown in dotted style and the abnormal situation where $t4$ is removed.

\begin{figure}[H]
\begin{center}
\includegraphics[width=6cm]{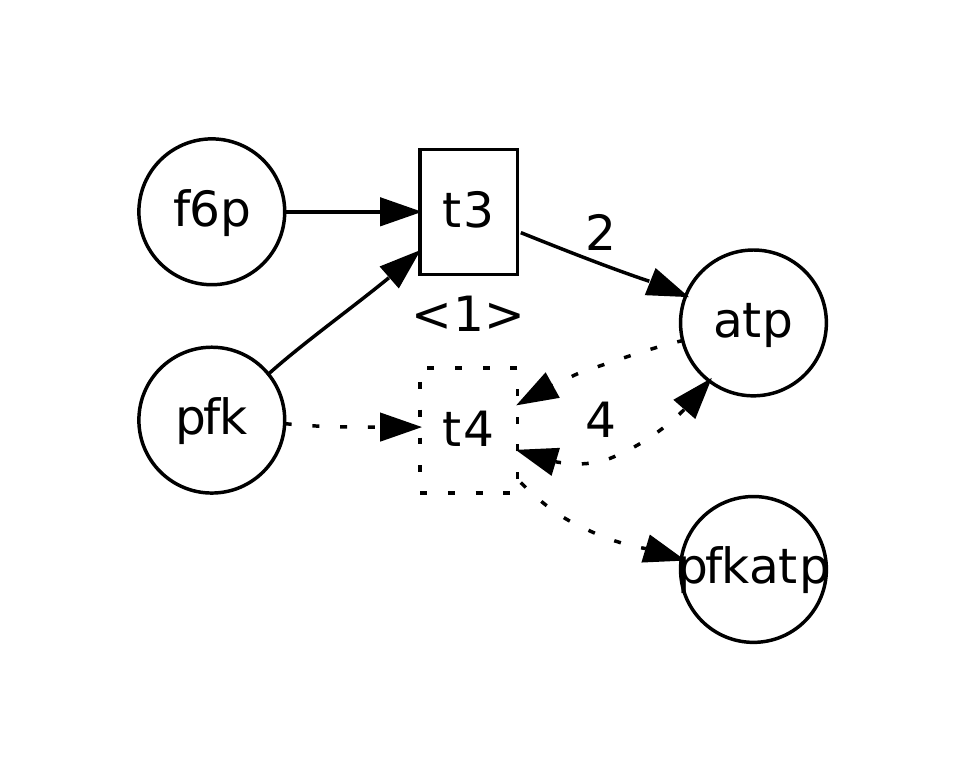}
\caption{Petri Net graph relevant to question \ref{q8}. ``pfk'' is phosphofructokinase, ``f6p'' is fructose 6-phosphate, ``atp'' is ATP and ``pfkatp'' is the pfk bound with atp for allosteric regulation. Transition ``t3'' represents enzymic action of pfk, ``t4'' represents the binding of pfk with atp. The double arrowed arc represents a threshold arc, which enables ``t4'' when there are at least 4 tokens available at ``atp''. Numbers above transitions in angular brackets represent arc priorities.}
\label{fig:q8}
\end{center}
\end{figure}

We encode both situations in ASP with maximal firing set policy and run them for 10 time steps. At the end of the run we compare the firing count of transition $t3$ for both cases. For the normal case (with $t4$), we get the following results:
{\footnotesize
\begin{verbatim}
holds(atp,0,c,0) holds(f6p,20,c,0) holds(pfk,20,c,0) holds(pfkatp,0,c,0)
fires(t3,0)
holds(atp,2,c,1) holds(f6p,19,c,1) holds(pfk,19,c,1) holds(pfkatp,0,c,1)
fires(t3,1)
holds(atp,4,c,2) holds(f6p,18,c,2) holds(pfk,18,c,2) holds(pfkatp,0,c,2)
fires(t4,2)
holds(atp,3,c,3) holds(f6p,18,c,3) holds(pfk,17,c,3) holds(pfkatp,1,c,3)
fires(t3,3)
holds(atp,5,c,4) holds(f6p,17,c,4) holds(pfk,16,c,4) holds(pfkatp,1,c,4)
fires(t4,4)
holds(atp,4,c,5) holds(f6p,17,c,5) holds(pfk,15,c,5) holds(pfkatp,2,c,5)
fires(t4,5)
holds(atp,3,c,6) holds(f6p,17,c,6) holds(pfk,14,c,6) holds(pfkatp,3,c,6)
fires(t3,6)
holds(atp,5,c,7) holds(f6p,16,c,7) holds(pfk,13,c,7) holds(pfkatp,3,c,7)
fires(t4,7)
holds(atp,4,c,8) holds(f6p,16,c,8) holds(pfk,12,c,8) holds(pfkatp,4,c,8)
fires(t4,8)
holds(atp,3,c,9) holds(f6p,16,c,9) holds(pfk,11,c,9) holds(pfkatp,5,c,9)
fires(t3,9)
holds(atp,5,c,10) holds(f6p,15,c,10) holds(pfk,10,c,10) holds(pfkatp,5,c,10)
fires(t4,10)
\end{verbatim}
}

Note that $t3$ fires only when the ATP falls below our set threshold, above it PFK is converted to PFK+ATP compound via $t4$. For the abnormal case (without $t4$) we get the following results:
{\footnotesize
\begin{verbatim}
holds(atp,0,c,0) holds(f6p,20,c,0) holds(pfk,20,c,0) holds(pfkatp,0,c,0)
fires(t3,0)
holds(atp,2,c,1) holds(f6p,19,c,1) holds(pfk,19,c,1) holds(pfkatp,0,c,1)
fires(t3,1)
holds(atp,4,c,2) holds(f6p,18,c,2) holds(pfk,18,c,2) holds(pfkatp,0,c,2)
fires(t3,2)
holds(atp,6,c,3) holds(f6p,17,c,3) holds(pfk,17,c,3) holds(pfkatp,0,c,3)
fires(t3,3)
holds(atp,8,c,4) holds(f6p,16,c,4) holds(pfk,16,c,4) holds(pfkatp,0,c,4)
fires(t3,4)
holds(atp,10,c,5) holds(f6p,15,c,5) holds(pfk,15,c,5) holds(pfkatp,0,c,5)
fires(t3,5)
holds(atp,12,c,6) holds(f6p,14,c,6) holds(pfk,14,c,6) holds(pfkatp,0,c,6)
fires(t3,6)
holds(atp,14,c,7) holds(f6p,13,c,7) holds(pfk,13,c,7) holds(pfkatp,0,c,7)
fires(t3,7)
holds(atp,16,c,8) holds(f6p,12,c,8) holds(pfk,12,c,8) holds(pfkatp,0,c,8)
fires(t3,8)
holds(atp,18,c,9) holds(f6p,11,c,9) holds(pfk,11,c,9) holds(pfkatp,0,c,9)
fires(t3,9)
holds(atp,20,c,10) holds(f6p,10,c,10) holds(pfk,10,c,10) holds(pfkatp,0,c,10)
fires(t3,10)
\end{verbatim}
}

Note that when ATP is not abundant, transition $t3$ fires continuously, which represents the enzymic activity that converts F6P to downstream substances. Trend of various runs is shown in Figure \ref{fig:q8:runs}.

\begin{figure}[H]
\begin{center}
\includegraphics[width=12cm]{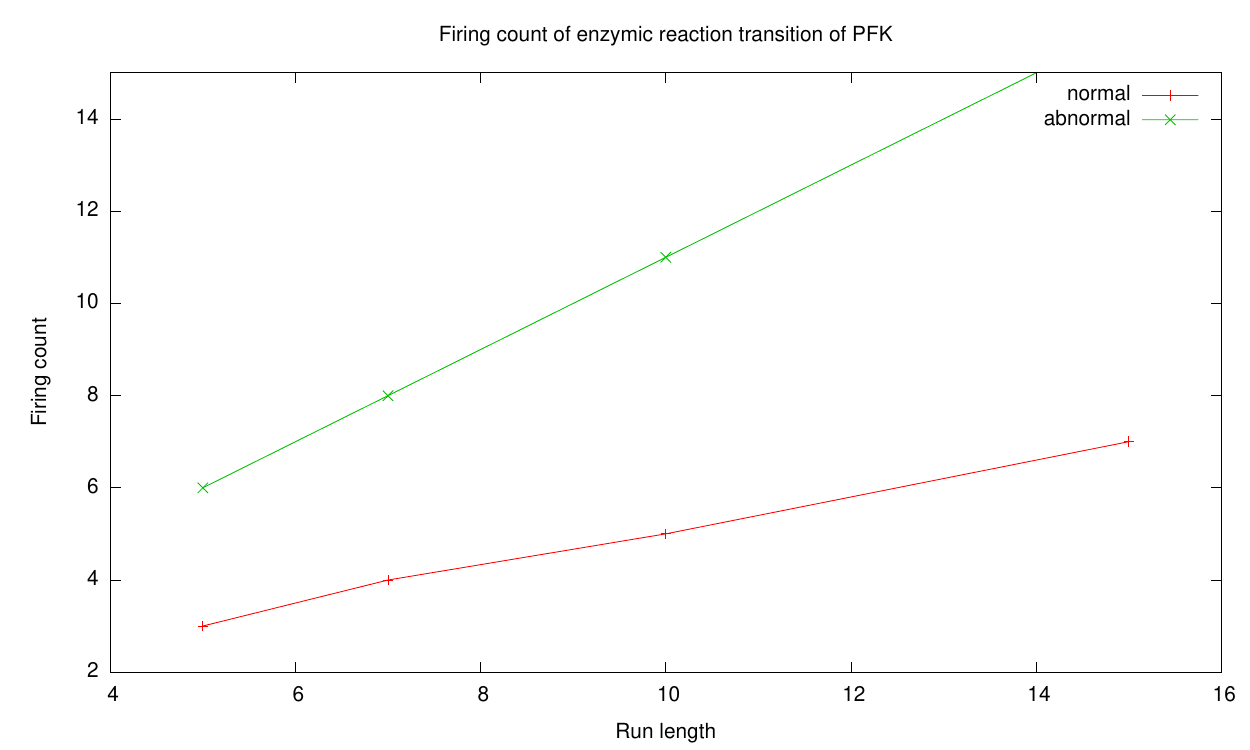}
\caption{Petri Net model in Figure \ref{fig:q8}.}
\label{fig:q8:runs}
\end{center}
\end{figure}

\end{solution}

\section{Comparing Altered Trajectories due to Automatic Conversion Intervention}
\begin{question}\label{q9}
How does the oxidation of NADH affect the rate of glycolysis?
\end{question}

Provided Answer:
\begin{quote}
``NADH must be oxidized back to NAD+ in order to be used in glycolysis. Without this molecule, glycolysis cannot occur.''
\end{quote}

\begin{solution}
Cellular respiration is summarized in Fig 9.6 of Campbell's book. NAD+ is reduced to NADH during glycolysis (see Campbell's Fig 9.9) during the process of converting Glyceraldehyde 3-phosphate (G3P) to 1,3-Bisphosphoglycerate (BPG13). NADH is oxidized back to NAD+ during oxidative phosphorylation by the electron transport chain (see Campbell's Fig 9.15). We can gauge the rate of glycolysis by the amount of Pyruvate produced, which is the end product of glycolysis. We simplify our model by abstracting glycolysis as a black-box that takes Glucose and NAD+ as input and produces NADH and Pyruvate as output, since there is a linear chain from Glucose to Pyruvate that depends upon the availability of NAD+. We also abstract oxidative phosphorylation as a black-box which takes NADH as input and produces NAD+ as output. None of the other inner workings of oxidative phosphorylation play a role in answering the question assuming they are functioning normally. We also ignore the pyruvate oxidation and citric acid cycle stages of cellular respiration since their end products only provide additional raw material for oxidative phosphorylation and do not add value to answering the question.

We assume a steady supply of Glucose and all other substances used in glycolysis but a limited supply of NAD+, since it can be recycled from NADH and we want to model its impact. We fulfill the steady supply requirement of Glucose with sufficient initial quantity in excess of what will be consumed during our simulation interval. We also ensure that we have sufficient initial quantity of NAD+ to maintain glycolysis as long as it can be recycled.

\begin{figure}[H]
\begin{center}
\includegraphics[width=9cm]{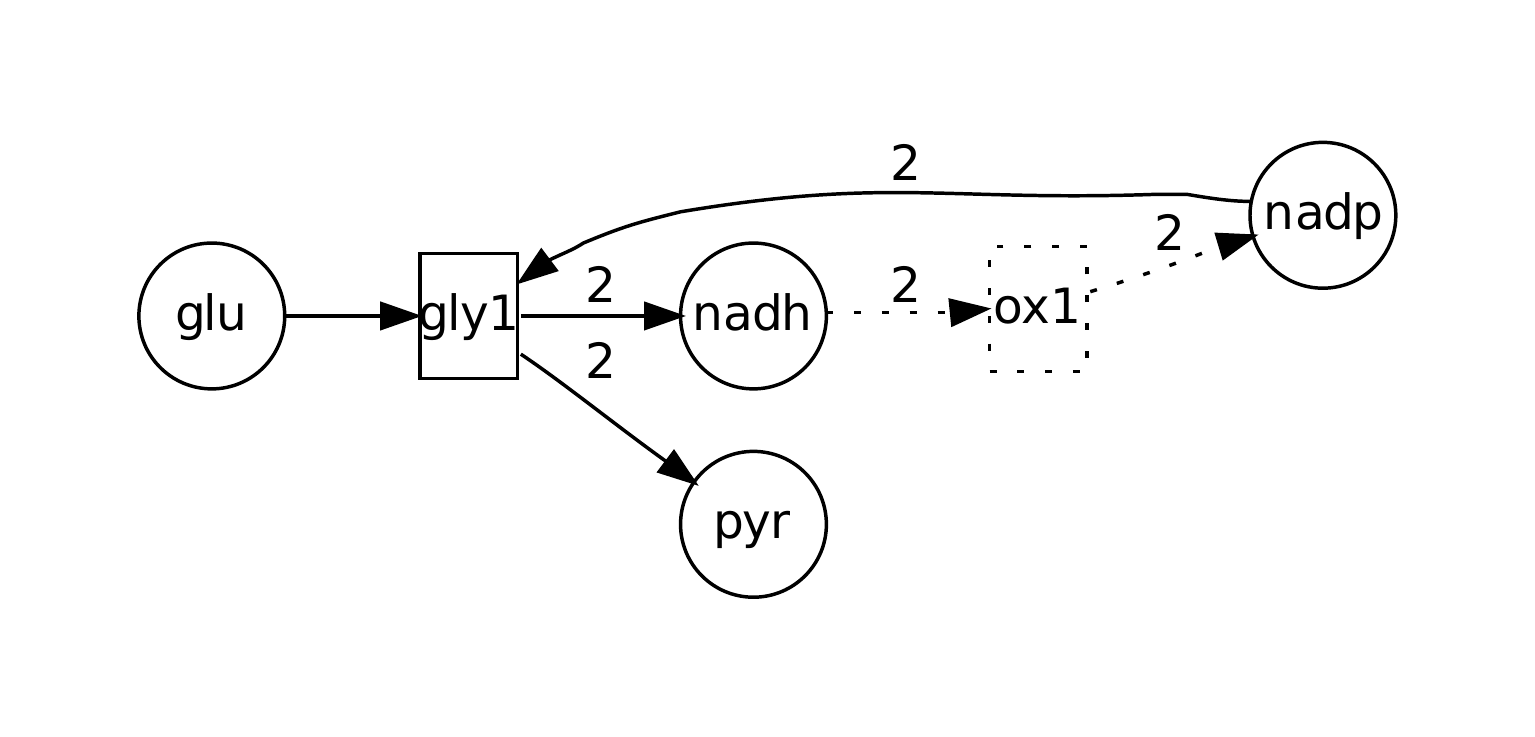}
\caption{Petri Net graph relevant to question \ref{q9}. ``glu'' represents glucose, ``gly1'' represents glycolysis, ``pyr'' represents pyruvate, ``ox1'' represents oxidative phosphorylation, ``nadh'' represents NADH and ``nadp'' represents NAD+. ``ox1'' is removed to model stoppage of oxidation of NADH to NAD+.}
\label{fig:q9}
\end{center}
\end{figure}

Figure \ref{fig:q9} is a Petri Net representation of our simplified model. Normal situation is modeled by the entire graph, where NADH is recycled back to NAD+, while the abnormal situation is modeled by the graph with the transition $ox1$ (shown in dotted style) removed. We encode both situations in ASP with the maximal firing set policy. Both situations are run for 5 steps and the amount of pyruvate is compared to determine the difference in the rate of glycolysis.

In normal situation (with $ox1$ transition), unique quantities of pyruvate ($pyr$) are as follows:
{\footnotesize
\begin{verbatim}
fires(gly1,0)
fires(gly1,1)
fires(gly1,2)
fires(gly1,3)
fires(gly1,4)

holds(pyr,10,5)
\end{verbatim}
}

while in abnormal situation (without $ox1$ transition), unique quantities of pyruvate are as follows:
{\footnotesize
\begin{verbatim}
fires(gly1,0)
fires(gly1,1)
fires(gly1,2)

holds(pyr,6,5)
\end{verbatim}
}

Note that the rate of glycolysis is lower when NADH is not recycled back to NAD+, as the glycolysis stops after the initial quantity of 6 NAD+ is consumed. Also, the $gly1$ transition does not fire after time-step 2, indicating glycolysis has stopped. Trend of various runs is shown in Figure \ref{fig:q9:runs}.

\begin{figure}[H]
\begin{center}
\includegraphics[width=10cm]{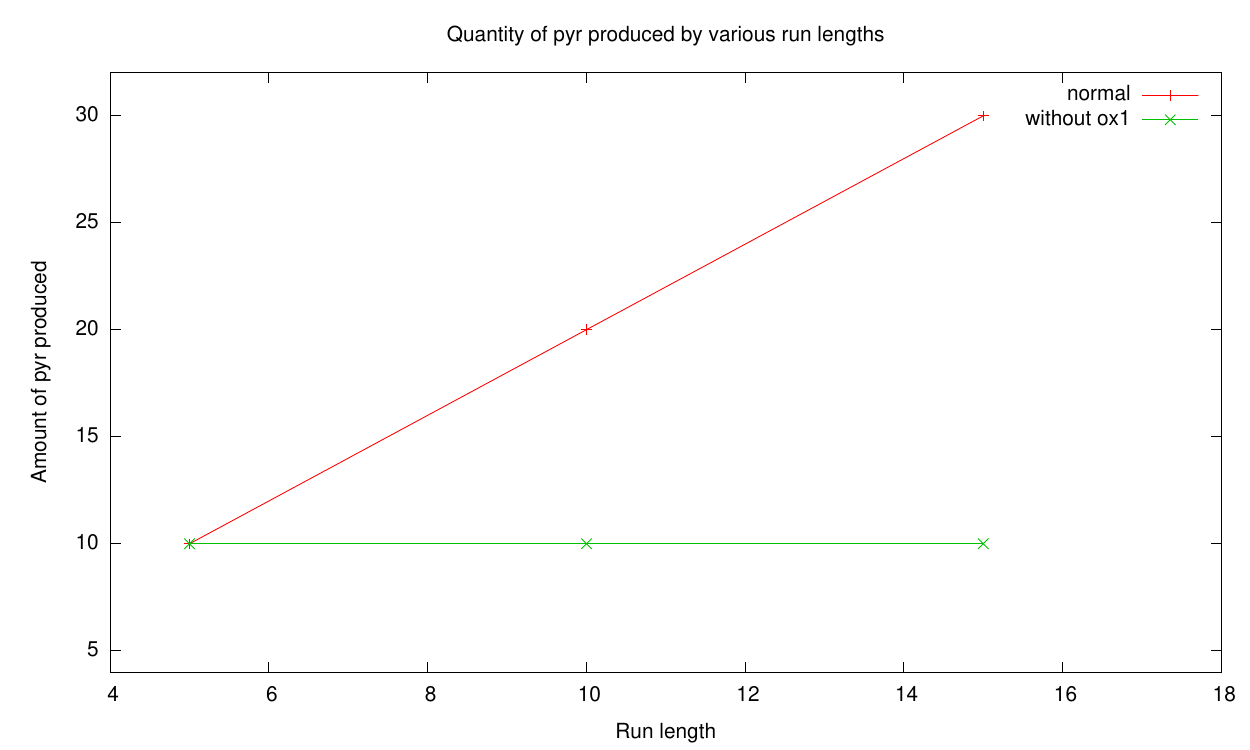}
\caption{Amount of ``pyr'' produced by runs of various lengths of Petri Net in Figure \ref{fig:q9}. It shows results for both normal situation where ``nadh'' is recycled to ``nadp'' as well as the abnormal situation where this recycling is stopped.}
\label{fig:q9:runs}
\end{center}
\end{figure}

\end{solution}

\section{Comparing Altered Trajectories due to Initial Value Intervention}
\begin{question}\label{q10}
During intense exercise, can a muscle cell use fat as a concentrated source of chemical energy? Explain.
\end{question}

Provided Answer: 
\begin{quote}
``When oxygen is present, the fatty acid chains containing most of the energy of a fat are oxidized and fed into the citric acid cycle and the electron transport chain. During intense exercise, however, oxygen is scarce in muscle cells, so ATP must be generated by glycolysis alone. A very small part of the fat molecule, the glycerol backbone, can be oxidized via glycolysis, but the amount of energy released by this portion is insignificant compared to that released by the fatty acid chains. (This is why moderate exercise, staying below 70\% maximum heart rate, is better for burning fat because enough oxygen remains available to the muscles.)''
\end{quote}

\begin{solution}
The process of fat consumption in glycolysis and citric acid cycle is summarized in Fig 9.19 of Campbell's book. Fats are digested into glycerol and fatty acids. Glycerol gets fed into glycolysis after being converted into Gyceraldehyde 3-phosphate (G3P), while fatty acids get fed into citric acid cycle after being broken down through beta oxidation and converted into Acetyl CoA. Campbell's Fig 9.18 identify a junction in catabolism where aerobic respiration or fermentation take place depending upon whether oxygen is present or not. Energy produced at various steps is in terms of ATP produced. In order to explain whether fat can be used as a concentrated source of chemical energy or not, we have to show the different ways of ATP production and when they kick in.

We combine the various pieces of information collected from Fig 9.19, second paragraph on second column of p/180, Fig 9.15, Fig 9.16 and Fig 9.18 of Campbell's book into Figure \ref{fig:q10}. We model two situations when oxygen is not available in the muscle cells (at the start of a intense exercise) and when oxygen is available in the muscle cells (after the exercise intensity is plateaued). We then compare and contrast them on the amount of ATP produced and the reasons for the firing sequences.

\begin{figure}[H]
\begin{center}
\includegraphics[width=15cm]{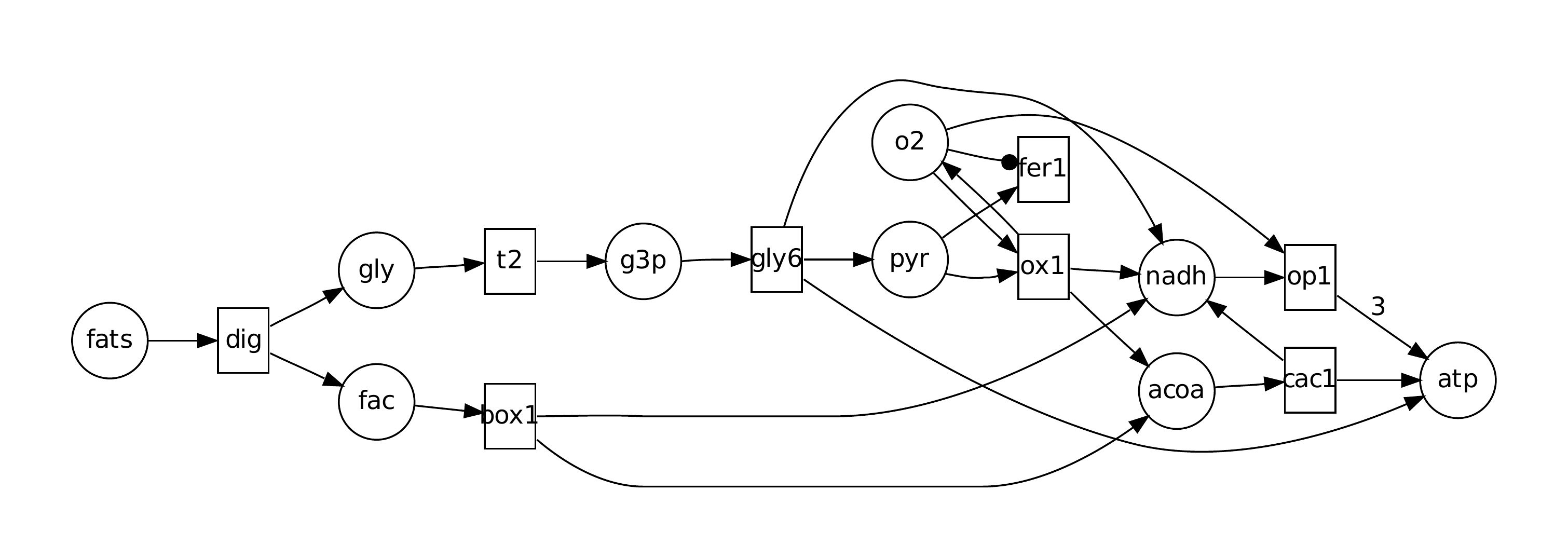}
\caption{Petri Net graph relevant to question \ref{q10}. ``fats'' are fats, ``dig'' is digestion of fats, ``gly'' is glycerol, ``fac'' is fatty acid, ``g3p'' is Glyceraldehyde 3-phosphate, ``pyr'' is pyruvate, ``o2'' is oxygen, ``nadh'' is NADH, ``acoa'' is Acyl CoA, ``atp'' is ATP, ``op1'' is oxidative phosphorylation, ``cac1'' is citric acid cycle, ``fer1'' is fermentation, ``ox1'' is oxidation of pyruvate to Acyl CoA and ``box1'' is beta oxidation.}
\label{fig:q10}
\end{center}
\end{figure}

Figure \ref{fig:q10} is a petri net representation of our simplified model. Our edge labels have lower numbers on them than the yield in Fig 9.16 of Campbell's book but they still capture the difference in volume that would be produced due to oxidative phosphorylation vs. glycolysis. Using exact amounts will only increase the difference of ATP production due to the two mechanisms. We encode both situations (when oxygen is present and when it is not) in ASP with maximal firing set policy. We run them for 10 steps. The firing sequence and the resulting yield of ATP explain what the possible use of fat as a source of chemical energy.

At he start of intense exercise, when oxygen is in short supply:
{\footnotesize
\begin{verbatim}
holds(acoa,0,0) holds(atp,0,0) holds(fac,0,0) holds(fats,5,0) 
holds(g3p,0,0) holds(gly,0,0) holds(nadh,0,0) holds(o2,0,0) holds(pyr,0,0)
fires(dig,0)
holds(acoa,0,1) holds(atp,0,1) holds(fac,1,1) holds(fats,4,1) 
holds(g3p,0,1) holds(gly,1,1) holds(nadh,0,1) holds(o2,0,1) holds(pyr,0,1)
fires(box1,1) fires(dig,1) fires(t2,1)
holds(acoa,1,2) holds(atp,0,2) holds(fac,1,2) holds(fats,3,2) 
holds(g3p,1,2) holds(gly,1,2) holds(nadh,1,2) holds(o2,0,2) holds(pyr,0,2)
fires(box1,2) fires(cac1,2) fires(dig,2) fires(gly6,2) fires(t2,2)
holds(acoa,1,3) holds(atp,2,3) holds(fac,1,3) holds(fats,2,3) 
holds(g3p,1,3) holds(gly,1,3) holds(nadh,3,3) holds(o2,0,3) holds(pyr,1,3)
fires(box1,3) fires(cac1,3) fires(dig,3) fires(fer1,3) fires(gly6,3) fires(t2,3)
holds(acoa,1,4) holds(atp,4,4) holds(fac,1,4) holds(fats,1,4) 
holds(g3p,1,4) holds(gly,1,4) holds(nadh,5,4) holds(o2,0,4) holds(pyr,1,4)
fires(box1,4) fires(cac1,4) fires(dig,4) fires(fer1,4) fires(gly6,4) fires(t2,4)
holds(acoa,1,5) holds(atp,6,5) holds(fac,1,5) holds(fats,0,5) 
holds(g3p,1,5) holds(gly,1,5) holds(nadh,7,5) holds(o2,0,5) holds(pyr,1,5)
fires(box1,5) fires(cac1,5) fires(fer1,5) fires(gly6,5) fires(t2,5)
holds(acoa,1,6) holds(atp,8,6) holds(fac,0,6) holds(fats,0,6) 
holds(g3p,1,6) holds(gly,0,6) holds(nadh,9,6) holds(o2,0,6) holds(pyr,1,6)
fires(cac1,6) fires(fer1,6) fires(gly6,6)
holds(acoa,0,7) holds(atp,10,7) holds(fac,0,7) holds(fats,0,7) 
holds(g3p,0,7) holds(gly,0,7) holds(nadh,10,7) holds(o2,0,7) holds(pyr,1,7)
fires(fer1,7)
holds(acoa,0,8) holds(atp,10,8) holds(fac,0,8) holds(fats,0,8) 
holds(g3p,0,8) holds(gly,0,8) holds(nadh,10,8) holds(o2,0,8) holds(pyr,0,8)
holds(acoa,0,9) holds(atp,10,9) holds(fac,0,9) holds(fats,0,9) 
holds(g3p,0,9) holds(gly,0,9) holds(nadh,10,9) holds(o2,0,9) holds(pyr,0,9)
holds(acoa,0,10) holds(atp,10,10) holds(fac,0,10) holds(fats,0,10) 
holds(g3p,0,10) holds(gly,0,10) holds(nadh,10,10) holds(o2,0,10) holds(pyr,0,10)
\end{verbatim}
}

when the exercise intensity has plateaued and oxygen is no longer in short supply:
{\footnotesize 
\begin{verbatim}
holds(acoa,0,0) holds(atp,0,0) holds(fac,0,0) holds(fats,5,0) 
holds(g3p,0,0) holds(gly,0,0) holds(nadh,0,0) holds(o2,10,0) holds(pyr,0,0)
fires(dig,0)
holds(acoa,0,1) holds(atp,0,1) holds(fac,1,1) holds(fats,4,1) 
holds(g3p,0,1) holds(gly,1,1) holds(nadh,0,1) holds(o2,10,1) holds(pyr,0,1)
fires(box1,1) fires(dig,1) fires(t2,1)
holds(acoa,1,2) holds(atp,0,2) holds(fac,1,2) holds(fats,3,2) 
holds(g3p,1,2) holds(gly,1,2) holds(nadh,1,2) holds(o2,10,2) holds(pyr,0,2)
fires(box1,2) fires(cac1,2) fires(dig,2) fires(gly6,2) fires(op1,2) fires(t2,2)
holds(acoa,1,3) holds(atp,5,3) holds(fac,1,3) holds(fats,2,3) 
holds(g3p,1,3) holds(gly,1,3) holds(nadh,2,3) holds(o2,9,3) holds(pyr,1,3)
fires(box1,3) fires(cac1,3) fires(dig,3) fires(gly6,3) fires(op1,3) 
        fires(ox1,3) fires(t2,3)
holds(acoa,2,4) holds(atp,10,4) holds(fac,1,4) holds(fats,1,4) 
holds(g3p,1,4) holds(gly,1,4) holds(nadh,4,4) holds(o2,8,4) holds(pyr,1,4)
fires(box1,4) fires(cac1,4) fires(dig,4) fires(gly6,4) fires(op1,4) 
        fires(ox1,4) fires(t2,4)
holds(acoa,3,5) holds(atp,15,5) holds(fac,1,5) holds(fats,0,5) 
holds(g3p,1,5) holds(gly,1,5) holds(nadh,6,5) holds(o2,7,5) holds(pyr,1,5)
fires(box1,5) fires(cac1,5) fires(gly6,5) fires(op1,5) fires(ox1,5) fires(t2,5)
holds(acoa,4,6) holds(atp,20,6) holds(fac,0,6) holds(fats,0,6) 
holds(g3p,1,6) holds(gly,0,6) holds(nadh,8,6) holds(o2,6,6) holds(pyr,1,6)
fires(cac1,6) fires(gly6,6) fires(op1,6) fires(ox1,6)
holds(acoa,4,7) holds(atp,25,7) holds(fac,0,7) holds(fats,0,7) 
holds(g3p,0,7) holds(gly,0,7) holds(nadh,9,7) holds(o2,5,7) holds(pyr,1,7)
fires(cac1,7) fires(op1,7) fires(ox1,7)
holds(acoa,4,8) holds(atp,29,8) holds(fac,0,8) holds(fats,0,8) 
holds(g3p,0,8) holds(gly,0,8) holds(nadh,10,8) holds(o2,4,8) holds(pyr,0,8)
fires(cac1,8) fires(op1,8)
holds(acoa,3,9) holds(atp,33,9) holds(fac,0,9) holds(fats,0,9) 
holds(g3p,0,9) holds(gly,0,9) holds(nadh,10,9) holds(o2,3,9) holds(pyr,0,9)
fires(cac1,9) fires(op1,9)
holds(acoa,2,10) holds(atp,37,10) holds(fac,0,10) holds(fats,0,10) 
holds(g3p,0,10) holds(gly,0,10) holds(nadh,10,10) holds(o2,2,10) holds(pyr,0,10)
fires(cac1,10) fires(op1,10)
\end{verbatim}
}

We see that more ATP is produced when oxygen is available. Most ATP (energy) is produced by the oxidative phosphorylation which requires oxygen. When oxygen is not available, small amount of energy is produced due to glycolysis of glycerol ($gly$). With oxygen a lot more energy is produced, most of it due to fatty acids ($fac$). Trend of various runs is shown in Figure \ref{fig:q10:runs}.

\begin{figure}[H]
\begin{center}
\includegraphics[width=10cm]{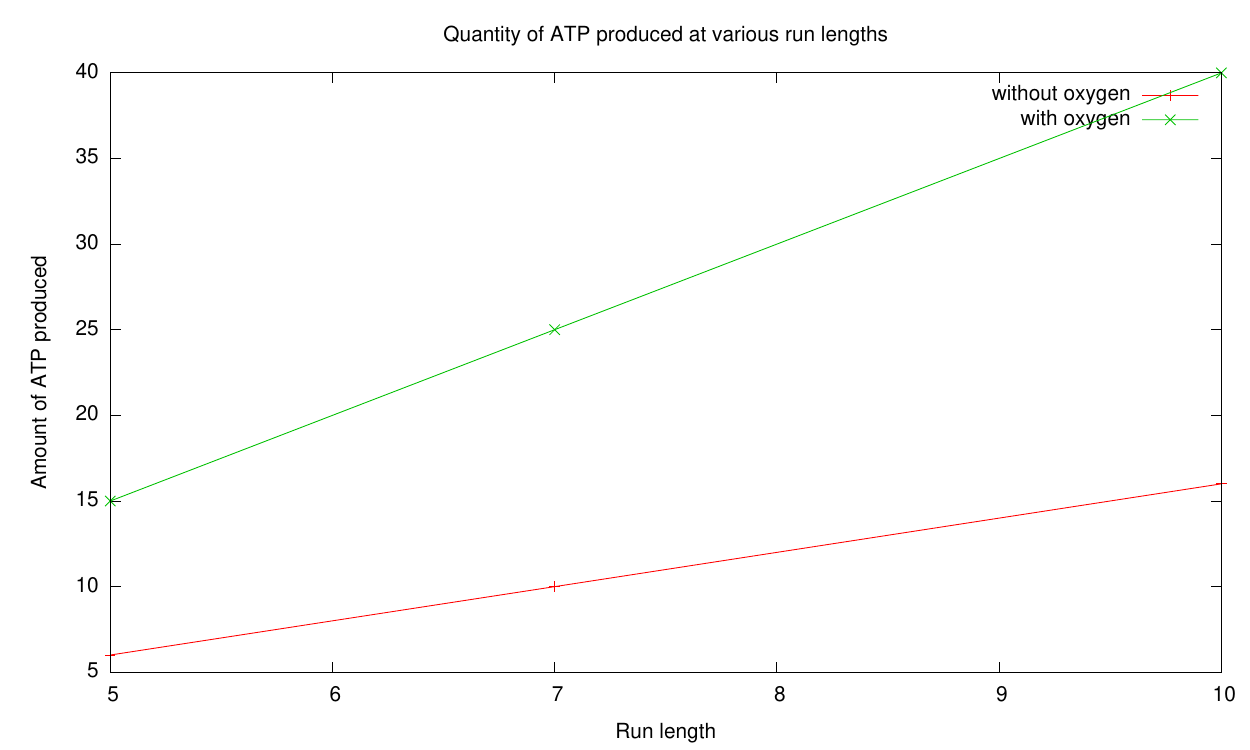}
\caption{Amount of ``atp'' produced by runs of various lengths of Petri Net in Figure \ref{fig:q10}. Two situations are shown: when oxygen is in short supply and when it is abundant.}
\label{fig:q10:runs}
\end{center}
\end{figure}

\end{solution}

\section{Conclusion}
In this chapter we presented how to model biological systems as Petri Nets, translated them into ASP, reasoned with them and answered questions about them. We used diagrams from Campbell's book, background knowledge and assumptions to facilitate our modeling work. However, source knowledge for real world applications comes from published papers, magazines and books. This means that we have to do text extraction. In one of the following chapters we look at some of the real applications that we have worked on in the past in collaboration with other researchers to develop models using text extraction. But first, we look at how we use the concept of answering questions using Petri Nets to build a question answering system. We will extend the Petri Nets even more for this.

\chapter[The BioPathQA System]{BioPathQA - A System for Modeling, Simulating, and Querying Biological Pathways}\label{ch:deepqa}

\section{Introduction}
In this chapter we combine the methods from Chapter~\ref{ch:modeling_qa}, notions from action languages, and ASP  to build a system {\em BioPathQA} and a language to specify pathways and query them. We show how various biological pathways are encoded in BioPathQA and how it computes answers of queries against them.

\section{Description of BioPathQA}
Our system has the following components: 
\begin{inparaenum}[(i)]
\item a pathway specification language 
\item a query language to specify the deep reasoning question,
\item an ASP program that encodes the pathway model and its extensions for simulation.
\end{inparaenum}

Knowledge about biological pathways comes in many different forms, such as cartoon diagrams, maps with well defined syntax and semantics (e.g. Kohn's maps~\cite{kohn2006molecular}), and biological publications. Similar to other technical domains, some amount of domain knowledge is also required. Users want to collect information from disparate sources and encode it in a pathway specification. We have developed a language to allow users to describe their pathway. This description includes describing the substances and actions that make up the pathway, the initial state of the substances, and how the state of the pathway changes due to the actions. An evolution of a pathway's state from the initial state, through a set of actions is called a trajectory. Being a specification language targeted at biological systems, multiple actions autonomously execute in parallel as soon as their preconditions are satisfied. The amount of parallelism is dictated by any resource conflicts between the actions. When that occurs, only one sub-set of the possible actions can execute, leading to multiple outcomes from that point on.

Questions are usually provided in natural language, which is vague. To avoid the vagaries of natural language, we developed a language with syntax close to natural language but with a well defined formal semantics. The query language allows a user to make changes to the pathway through interventions, and restrict its trajectories through observations and query on aggregate values in a trajectory, across a set of trajectories and even over two sets of trajectories. This allows the user to compare a base case of a pathway specification with an alternate case modified due to interventions and observations. This new feature is a major contribution of our research. 

Inspiration for our high level language comes from action languages and query languages such as~\cite{gelfond1993representing}. While action languages generally describe transition systems~\cite{gelfond1998action}, our language describes trajectories. In addition, our language is geared towards modeling natural systems, in which actions occur autonomously~\cite{reiter1996natural} when their pre-conditions are satisfied; and we do not allow the quantities to become negative (as the quantities represent amounts of physical entities).

Next we describe the syntax of our pathway specification language and the query language. Following that we will describe the syntax of our language and how we encode it in ASP.

\section{Syntax of Pathway Specification Language (BioPathQA-PL)}\label{plang:syntax}
The alphabet of pathway specification language $\mathcal{P}$ consists of disjoint nonempty domain-dependent sets $A$, $F$, $L$ representing actions, fluents, and locations, respectively; a fixed set $S$ of firing styles; a fixed set $K$ of keywords providing syntactic sugar (shown in bold face in pathway specification language below); a fixed set of punctuations $\{ `,' \}$; and a fixed set of special constants $\{`1',`*',`max'\}$; and integers. 

Each fluent $f \in F$ has a domain $dom(f)$ which is either integer or binary and specifies the values $f$ can take.  A fluent is either simple, such as $f$ or locational, such as $f[l]$, where $l \in L$. A {\em state} $s$ is an interpretation of $F$ that maps fluents to their values. We write $s(f)=v$ to represent ``$f$ has the value $v$ in state $s$''. States are indexed, such that consecutive states $s_i$ and $s_{i+1}$ represent an evolution over one time step from $i$ to $i+1$ due to firing of an action set $T_i$ in $s_i$.

We illustrate the role of various symbols in the alphabet with examples from the biological domain. Consider the following example of a hypothetical pathway specification:

{\footnotesize
\begin{align}
\label{pspec1:domain}&\mathbf{domain~of~}  sug \mathbf{~is~} integer, fac \mathbf{~is~} integer, acoa \mathbf{~is~} integer, h2o \mathbf{~is~} integer\\
\label{pspec1:gly}&gly \mathbf{~may~execute~causing~}  sug \mathbf{~change~value~by~} -1, acoa \mathbf{~change~value~by~} +1\\
\label{pspec1:box}&box \mathbf{~may~execute~causing~}  fac \mathbf{~change~value~by~} -1, acoa \mathbf{~change~value~by~} +1 \\
\label{pspec1:box:guard}&\mathbf{~~~~~~~~~~~if~}  h2o \mathbf{~has~value~} 1 \mathbf{~or~higher~}\\
\label{pspec1:box:inhibit}&\mathbf{inhibit~} box \mathbf{~if~}  sug \mathbf{~has~value~} 1 \mathbf{~or~higher~}\\
\label{pspec1:init}&\mathbf{initially~}  sug \mathbf{~has~value~} 3,  fac \mathbf{~has~value~} 4, acoa \mathbf{~has~value~} 0 & 
\end{align}
}

It describes two processes glycolysis and beta-oxidation represented by actions `$gly$' and `$box$' in lines \eqref{pspec1:gly} and \eqref{pspec1:box}-\eqref{pspec1:box:guard} respectively. Substances used by the pathway, i.e. sugar, fatty-acids, acetyl-CoA, and water are represented by numeric fluents `$sug$',`$fac$',`$acoa$', and `$h2o$' respectively in line~\eqref{pspec1:domain}. When glycolysis occurs, it consumes $1$ unit of sugar and produces $1$ unit of acetyl-CoA (line~\eqref{pspec1:gly}. When  beta-oxidation occurs, it consumes $1$ unit of fatty-acids and produces $1$ unit of acetyl-CoA (line~\eqref{pspec1:box}). The inputs of glycolysis implicitly impose a requirement that glycolysis can only occur when at least $1$ unit of sugar is available. Similarly, the input of beta-oxidation implicitly a requirement that beta-oxidation can only occur when at least $1$ unit of fatty-acids is available. Beta oxidation has an additional condition imposed on it in line~\eqref{pspec1:box:guard} that there must be at least $1$ unit of water available. We call this a {\em guard} condition on beta-oxidation. Line~\eqref{pspec1:box:inhibit} explictly inhibits beta-oxidation when there is any sugar available; and line~\eqref{pspec1:init} sets up the initial conditions of the pathway, i.e. Initially $3$ units of each sugar, $4$ units of fatty-acids are available and no acetyl-CoA is available. The words `$\{domain,$ $is,$ $may,$ $execute,$ $causing,$ $change,$ $value,$ $by,$ $has,$ $or,$ $higher,$ $inhibit,$ $if,$ $initially\}$' are keywords.

When locations are involved, locational fluents take place of simple fluents and our representation changes to include locations. For example:
{\footnotesize
\begin{align*}
&gly \mathbf{~may~execute~causing~}  \nonumber\\
&~~~~sug \mathbf{~atloc~} mm \mathbf{~change~value~by~} -1, \nonumber\\
&~~~~acoa \mathbf{~atloc~} mm \mathbf{~change~value~by~} +1
\end{align*}
}
represents glycolysis taking $1$ unit of sugar from mitochondrial matrix (represented by `$mm$') and produces acetyl-CoA in the mitochondrial matrix. Here `$atloc$' is an additional keyword.

A pathway is composed of a collection of different types of statements and clauses. We first introduce their syntax, following that we will give intuitive definitions, and following that we will show how they are combined together to construct a pathway specification.

\begin{definition}[Fluent domain declaration statement]
A fluent domain declaration statement declares the values a fluent can take. It has the form:
{\footnotesize
\begin{align}
\mathbf{domain~of~} f \mathbf{~is~} `integer'|`binary' \\ 
\mathbf{domain~of~} f \mathbf{~atloc~} l \mathbf{~is~} `integer'|`binary' 
\end{align}
}
for simple fluent ``$f$'', and locational fluent ``$f[l]$''. Multiple domain statements are compactly written as:
{\footnotesize
\begin{align}
\mathbf{domain~of~} f_1 \mathbf{~is~} `integer'|`binary' , \dots,  f_n \mathbf{~is~} `integer'|`binary' \\
\mathbf{domain~of~} f_1 \mathbf{~atloc~} l_1 \mathbf{~is~} `integer'|`binary' , \dots,  f_n \mathbf{~atloc~} l_n \mathbf{~is~} `integer'|`binary'
\end{align}
}
\end{definition}
Binary domain is usually used for representing substances in a signaling pathway, while a metabolic pathways take positive numeric values. Since the domain is for a physical entity, we do not allow negative values for fluents.

\begin{definition}[Guard condition]
A {\em guard condition} takes one of the following forms:
{\footnotesize
\begin{align}
\label{eqn:mayfire:cond:ge}&f \mathbf{~has~value~} w \mathbf{~or~higher~} \\
\label{eqn:mayfire:cond:ge:atloc}&f \mathbf{~atloc~} l \mathbf{~has~value~} w \mathbf{~or~higher~}\\
\label{eqn:mayfire:cond:lt}&f \mathbf{~has~value~lower~than~} w \\
\label{eqn:mayfire:cond:lt:atloc}&f \mathbf{~atloc~} l \mathbf{~has~value~lower~than~} w \\
\label{eqn:mayfire:cond:eq}&f \mathbf{~has~value~equal~to~} w \\
\label{eqn:mayfire:cond:eq:atloc}&f \mathbf{~atloc~} l \mathbf{~has~value~equal~to~} w \\
\label{eqn:mayfire:cond:grad}&f_1 \mathbf{~has~value~higher~than~} f_2\\
\label{eqn:mayfire:cond:grad:atloc}&f_1 \mathbf{~atloc~} l_1 \mathbf{~has~value~higher~than~} f_2 \mathbf{~atloc~} l_2
\end{align}
}
where, each $f$ in \eqref{eqn:mayfire:cond:ge}, \eqref{eqn:mayfire:cond:lt}, \eqref{eqn:mayfire:cond:eq}, \eqref{eqn:mayfire:cond:grad} is a simple fluent, each $f[l]$ in \eqref{eqn:mayfire:cond:ge:atloc}, \eqref{eqn:mayfire:cond:lt:atloc}, \eqref{eqn:mayfire:cond:eq:atloc}, \eqref{eqn:mayfire:cond:grad:atloc} is a locational fluent with location $l$, and each $w \in \mathbb{N}^+ \cup \{ 0 \}$.
\end{definition}

\begin{definition}[Effect clause]
An {\em effect} clause can take one of the following forms: 
{\footnotesize
\begin{align}
\label{eqn:effect:noloc}&f \mathbf{~change~value~by~} e\\
\label{eqn:effect:atloc}&f \mathbf{~atloc~} l \mathbf{~change~value~by~} e
\end{align}
}
where $a$ is an action, $f$ in \eqref{eqn:effect:noloc} is a simple fluent, $f[l]$ in \eqref{eqn:effect:atloc} is a locational fluent with location $l$, $e \in \mathbb{N}^+ \cup \{ * \}$ for integer fluents or $e \in \{1,-1,*\}$ for binary fluents.
\end{definition}

\begin{definition}[May-execute statement]
A {\em may-execute} statement captures the conditions for firing an action $a$ and its impact. It is of the form: 
{\footnotesize
\begin{align}\label{eqn:may:exec}
a \mathbf{~may~execute~causing~} & \mathit{effect}_1, \dots, \mathit{effect}_m \nonumber\\
\mathbf{~if~} &guard\_cond_1, \dots, guard\_cond_n
\end{align}
}
where $\mathit{effect}_i$ is an {\em effect} clause; and $\mathit{guard\_cond}_j$ is a guard condition clause, $m > 0$, and $n \geq 0$. If $n = 0$, the effect statement is unconditional (guarded by $\top$) and the $\mathbf{if}$ is dropped. A single {\em may-execute} statement must not have $\mathit{effect}_i, \mathit{effect}_j$ with $e_i < 0, e_j < 0$ for the same fluent; or $e_i > 0, e_j > 0$ for the same fluent.
\end{definition}

\begin{definition}[Must-execute statement]
An {\em must-execute} statement captures the impact of firing of an action $a$ that must fire when enabled (as long as it is not inhibited). It is an expression of the form:
{\footnotesize
\begin{align}\label{eqn:must:exec}
a \mathbf{~normally~must~execute~causing~} & \mathit{effect}_1, \dots, \mathit{effect}_m \nonumber\\ 
\mathbf{~if~} &guard\_cond_{1}, \dots, guard\_cond_n
\end{align}
}
where $\mathit{effect}_i$, and $\mathit{guard\_cond}_j$ are as in \eqref{eqn:may:exec} above.
\end{definition}

\begin{definition}[Inhibit statement]
An {\em inhibit} statement captures the conditions that inhibit an action from occurring. It is an expression of the form:
{\footnotesize
\begin{align}
\label{eqn:inhibit}\mathbf{inhibit~} a \mathbf{~if~} guard\_cond_1, \dots, guard\_cond_n 
\end{align}
}
where $a$ is an action, $guard\_cond_i$ is a guard condition clause, and $n \geq 0$. if $n = 0$, the inhibition of action `$a$' is unconditional `$\mathbf{if}$' is dropped.
\end{definition}

\begin{definition}[Initial condition statement]
An {\em initial condition} statement captures the initial state of pathway. It is of the form:
{\footnotesize
\begin{align}
\label{eqn:initial}\mathbf{initially~} & f \mathbf{~has~value~} w\\
\label{eqn:initial:atloc}\mathbf{initially~} & f \mathbf{~atloc~} l \mathbf{~has~value~} w
\end{align}
}
where each $f$ in \eqref{eqn:initial} is a simple fluent, $f[l]$ in \eqref{eqn:initial:atloc} is a locational fluent with location $l$, and each $w$ is a non-negative integer. Multiple {\em initial condition} statements  are written compactly as:
{\footnotesize
\begin{align*}
\mathbf{initially~} & f_1\mathbf{~has~value~} w_1, \dots, f_n \mathbf{~has~value~} w_n\\
\mathbf{initially~} & f_1 \mathbf{~atloc~} l_1 \mathbf{~has~value~} w_1, \dots, f_n \mathbf{~atloc~} l_n \mathbf{~has~value~} w_n
\end{align*}
}
\end{definition}

\begin{definition}[Duration Statement]
A {\em duration} statement represents the duration of an action that takes longer than a single time unit to execute. It is of the form:
{\footnotesize
\begin{align}\label{eqn:dur}
a \mathbf{~executes~in~} d \mathbf{~time~units} 
\end{align}
}
where $d$ is a positive integer representing the action duration. 
\end{definition}

\begin{definition}[Stimulate Statement]
A {\em stimulate} statement changes the rate of an action. It is an expression of the form:
{\footnotesize
\begin{align}\label{eqn:stimulate}
\mathbf{normally~stimulate~} a \mathbf{~by~factor~} n \mathbf{~if~} guard\_cond_1,\dots,guard\_cond_n
\end{align}
}
where $guard\_cond_i$ is a {\em condition}, $n > 0$. When $n=0$, the stimulation is unconditional and $\mathbf{~if~}$ is dropped. A stimulation causes the $\mathit{effect}$ in {\em may-cause} and {\em must-fire} multiplied by $n$. 
\end{definition}

Actions execute automatically when fireable, subject to the available fluent quantities. 
\begin{definition}[Firing Style Statement]
A {\em firing style} statement specifies how many actions execute simultaneously (or action parallelism). It is of the form:
{\footnotesize
\begin{equation}\label{eqn:firing:style}
\mathbf{firing~style~}  S
\end{equation}
}
where, $S$ is either ``$1$'', ``$*$'', or ``$max$'' for serial execution, interleaved execution, and maximum parallelism.
\end{definition}

We will now give the intuitive meaning of these statements and put them into context w.r.t. the biological domain. Though our description below uses simple fluents only, it applies to locational fluents in a obvious manner. The reason for having locational fluents at all is that they allow a more natural pathway specification when substance locations are involved instead of devising one's own encoding scheme. For example, in a mitochondria, hydrogen ions (H+) appear in multiple locations (intermembrane space and mitochondrial matrix), with each location carrying its distinct quantity separate from other locations.

Intuitively, a {\em may-execute statement}~\eqref{eqn:may:exec} represents an action $a$ that may fire if all conditions `$\mathit{guard\_cond}_{1},\dots,\mathit{guard\_cond}_n$' hold in the current state. When it executes, it impacts the state as specified in $\mathit{effect}$s. In biological context, action $a$ represents a process, such as a reaction, $\mathit{effect}$s represent the inputs / ingredients of the reaction, and $guard\_cond$ represent additional preconditions necessary for the reaction to proceed. Condition~\eqref{eqn:mayfire:cond:ge} holds in a state $s$ if $s(f) \geq w$. It could represent an initiation concentration $w$ of a substance $f$ which is higher than the quantity consumed by the reaction $a$. Condition~\eqref{eqn:mayfire:cond:lt} holds in a state $s$ if $s(f) < w$. Condition~\eqref{eqn:mayfire:cond:eq} holds in a state $s$ if $s(f) = w$. Condition~\eqref{eqn:mayfire:cond:grad} holds in a state $s$ if  $s(f_1) > s(f_2)$ capturing a situation where a substance gradient is required for a biological process to occur. An example of one such process is the synthesis of ATP by ATP Synthase, which requires a H+ (Hydrogen ion) gradient across the inner mitochondrial matrix~\cite[Figure 9.15]{CampbellBook}.

Intuitively, the {\em effect clause}~\eqref{eqn:effect:noloc} of an action describes the impact of an action on a fluent. When an action $a$ fires in a state $s$, the value of $f$ changes according to the effect clause for $f$. The value of $f$ increases by $e$ if $e > 0$, decreases by $e$ if $e < 0$, or decreases by $s(f)$ if $e = `*'$ (where $`*'$ can be interpreted as $-s(f)$). For a reaction $a$, a fluent with $e < 0$ represents an ingredient consumed in quantity $|e|$ by the reaction; a fluent with $e > 0$ represents a product of the reaction in quantity $e$; a fluent with $e = `*'$ represents consuming all quantity of  the substance due to the reaction. Since the fluents represent physical substances, their quantities cannot become negative. As a result, any action that may cause a fluent quantity to go below zero is disallowed. 

Intuitively, a {\em must-execute statement}~\eqref{eqn:must:exec} is similar to a {\em may-exec}, except that when enabled, it preferentially fires over other actions as long as there isn't an {\em inhibit} proposition that will cause the action to become inactive. It captures the effect of an action that must happen whenever enabled.

Intuitively, an {\em inhibit statement}~\eqref{eqn:inhibit} specifies the conditions that inhibits an action. In a biological context, it defines inhibition of reactions, e.g., through biological feedback control. Though we could have added these conditions to {\em may-exec}, it is more intuitive to keep them separate as inhibition conditions are usually discovered separately in a biological domain. Including them as part of {\em may-fire} would constitute a surgery of existing knowledge bases.

Intuitively, an {\em initial condition statement}~\eqref{eqn:initial} specifies the initial values of fluents. The collection of such propositions defines the initial state $s_0$ of the pathway. In a biological context, this defines the initial distribution of substances in the biological system.

Intuitively, an action {\em duration statement}~\eqref{eqn:dur} represents action durations, especially when an action takes longer to execute. When an action $a$ with duration $d$ fires in state $s_k$, it immediately decreases the values of fluents with $e < 0$ and $e = *$ upon execution, however, it does not increase the value of fluents with $e > 0$ until time the end of its execution in state $s_{k+d}$. In a biological context the action duration captures a reaction's duration. A reaction consumes its ingredients immediately on firing, processes them for duration $d$ and generates its products at the end of this duration. 

Intuitively, a {\em stimulate statement}~\eqref{eqn:stimulate} represents a change in the rate of an action $a$. The stimulation causes the action to change its rate of consumption of its ingredients and production of its products by a factor $n$. In biological context, this stimulation can be a result of an enzyme or a stimulant's availability, causing a reaction that normally proceeds slowly to occur faster.

Intuitively, a {\em firing style statement}~\eqref{eqn:firing:style} specifies the parallelism of actions. When it is ``$1$'', at most one action may fire, when it is ``$max$'', the maximum allowable actions must fire, and when it is ``$*$'', any subset of fireable actions may fire simultaneously. In a biological domain the firing style allows one to model serial operations, parallel operations and maximally parallel operations. The maximum parallelism is also useful in quickly discovering changes that occur in a biological system.

\begin{definition}[Pathway Specification]
A {\em pathway specification} is composed of one or more {\em may-execute}, {\em must-execute}, {\em effect}, {\em inhibit}, {\em stimulate}, {\em initially}, {\em priority}, {\em duration} statements, and one {\em firing style} statement. When a {\em duration} statement is not specified for an action, it is assumed to be $1$. Any fluents for which an initial quantity is not specified are assumed to have a value of zero.

A pathway specification is {\em consistent} if 
\begin{inparaenum}[(i)]
\item there is at most one firing style, priority, duration statement for each action $a$;
\item the $guard\_cond_1,\dots,guard\_cond_n$ from a {\em may-execute} or {\em must-execute} are disjoint from any other {\em may-execute} or {\em must-execute}~\footnote{Note that `$f1 \text{ has value } 5 \text{ or higher }$' overlaps with `$f1 \text{ has value } 7 \text{ or higher}$' and the two conditions are not considered disjoint.}; 
\item locational and non-locational fluents may not be intermixed; 
\item domain of fluents, effects, conditions and numeric values are consistent, i.e., effects and conditions on binary fluents must be binary; and
\item the pathway specification does not cause it to violate fluent domains by producing non-binary values for binary fluents.
\end{inparaenum}
\end{definition}

Each pathway specification $\mathbf{D}$ represents a collection of trajectories of the form: $\sigma = s_0, T_0, s_1, \dots, s_{k-1}, T_{k-1}, s_k$. Each trajectory encodes an evolution of the pathway starting from an initial state $s_0$, where $s_i$'s are states, and $T_i$'s are sets of actions that fired in state $s_i$. 

Intuitively, a trajectory starts from the {\em initial state} $s_0$. Each $s_i,s_{i+1}$ pair represents the state evolution in one time step due to the action set $T_i$. An action set $T_i$ is only executable in state $s_i$, if the sum of changes to fluents due to $e_i < 0$ and $e_i = *$ will not result in any of the fluents going negative. Changes to fluents due to $e_i > 0$ for the action set $T_i$ occur over subsequent time-steps depending upon the durations of actions involved. Thus, the state $s_i(f_i)$ is the sum of $e_i > 0$ for actions of duration $d$ that occurred $d$ time steps before (current time step) $i$, i.e. $a \in T_{i-d}$, where the default duration $d$ of an action is $1$ if none specified. 

Next we describe the semantics of the pathway specification language, which describes how these trajectories are generated.

\section{Semantics of Pathway Specification Language (BioPathQA-PL)}\label{sec:plang:sem}
The semantics of the pathway specification language are defined in terms of the trajectories of the domain description $\mathbf{D}$. Since our pathway specification language is inspired by Petri Nets, we use Petri Nets execution semantics to define its trajectories. However, some constructs in our pathway language specification are not directly representable in standard Petri Nets, as a result, we will have to extend them. 

Let an {\em arc-guard} be a conjunction of {\em guard conditions} of the form~\eqref{eqn:mayfire:cond:ge}-\eqref{eqn:mayfire:cond:grad:atloc}, such that it is wholly constructed of either locational or non-locational fluents, but not both.

We introduce a new type of Guarded-arc Petri Net in which each arc has an {\em arc-guard} expression associated with it. Arcs with the same arc-guard are traversed when a transition connected to them fires and the arc-guard is found to hold in the current state. The arc-guards of arcs connected to the same transition form an exclusive set, such that only arcs corresponding to one guard expression may fire (for one transition). This setup can lead to different outcomes of an action\footnote{Arcs for different guard expressions emanating / terminating at a place can further be combined into a single conditional arc with conditional arc-weights. If none of the condition applies then the arc is assumed to be missing.}. 

The transitions in this new type of Petri Net can have the following inscriptions on them:
\begin{enumerate}
\item Propositional formula, specifying the executability conditions of the transition.
\item Arc-durations, represented as ``$dur(n)$'', where $n \in \mathbb{N}^+$
\item A {\em must-execute} inscription, ``$must\text{-}execute(guard)$'', requires that when the $guard$ holds in a state where this transition is enabled, it must fire, unless explicitly inhibited. The $guard$ has the same form as an $arc\text{-}guard$
\item A {\em stimulation} inscription, ``$stimulate(n,guard)$'', applies a multiplication factor $n \in \mathbb{N}^+$ to the input and output quantities consumed and produced by the transition, when $guard$ hold in the current state, where $guard$ has the same form as an $arc\text{-}guard$.
\end{enumerate}

Certain aspects of our nets are similar to CPNs~\cite{jensen2007coloured}. However, the CPNs do not allow our semantics of the reset arcs, or must-fire guards.

\subsection{Guarded-Arc Petri Net}\label{sec:gpn}
\begin{figure}[htbp]
   \centering
   \includegraphics[width=0.7\linewidth]{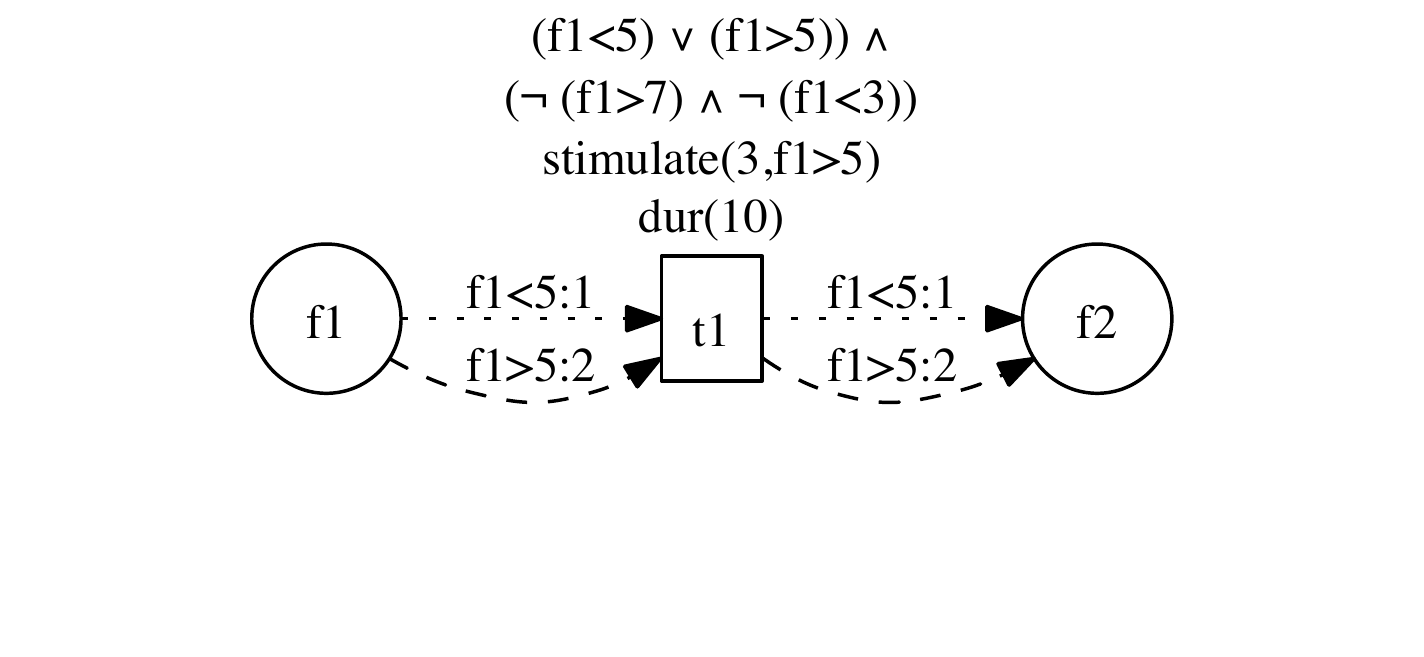} %
   \vspace{-40pt}
   \caption{Example of a guarded-arc Petri Net. }
   \label{fig:guarded:arc:pn}
\end{figure}
Figure~\ref{fig:guarded:arc:pn} shows an example of a guarded-arc Petri Net. There are two arc-guard expressions $f1<5$ and $f1>5$. When $f1<5$, $t1$ consumes one token from place $f1$ and produces one token in place $f2$. When $f1>5$, $t1$'s consumption and production of the same tokens doubles. The transition $t1$ implicitly gets the guards for each arc represented the {\em or}-ed conditions $(f1<5) \vee (f1>5))$. The arc also has two conditions inhibiting it, they are represented by the {\em and}-ed conditions $\lnot(f1>7) \wedge \lnot(f1<3)$, where `$\lnot$' represents logical not. Transition $t1$ is stimulated by factor $3$ when $f1>5$ and it has a duration of $10$ time units. A transition cannot fire even though one of its arc-guards is enabled, unless the token requirements on the arc itself are also fulfilled, e.g. if $f1$ has value $0$ in the current state, even though $f1 < 5$ guard is satisfied, the transition cannot execute, because the input arc $(f1,t1)$ for this guard needs $1$ token.

\begin{definition}[Guard]\label{def:guard}
A {\em condition} is of the form: $(f < v), (f \leq v), (f > v), (f \geq v), (f = v)$, where $f$ is a fluent and $v$ either a fluent or a numeric constant. Then, a guard is a propositional formula of conditions, with each condition treated as a proposition, subject to the restriction that all fluents in all conditions in a guard are either locational or simple, but not both.
\end{definition}

\begin{definition}[Interpretation of a Guard]\label{def:guard:interp}
An interpretation of a guard $G$ is a possible assignment of a value to each fleuent $f \in G$ from the domain of $f$.
\end{definition}

\begin{definition}[Guard Satisfaction]\label{def:guard:sat}
A guard $G$ with simple fluents is satisfied w.r.t. a state $s$, written $s \models G$ iff $G$ has an interpretation in which each of its fluents $f$ has the value $s(f)$ and $G$ is true. A guard $G$ with locational fluents is satisfied w.r.r. a state $s$, written $s \models G$ iff $G$ has an interpretation in which each of its fluents $f[l]$ has the value $m_{s(l)}(f)$ and $G$ is true, where $m_X(f)$ is the multiplicity of $f$ in $X$.
\end{definition}

\begin{definition}[Guarded-Arc Petri Net]\label{def:gpn}
A Guarded-Arc Petri Net is  a tuple \\$PN^G=(P,T,G,E,R,W,D,B,TG,MF,L)$, where:
\begin{align*}
P & \text{ is a finite set of places}\\
T & \text{ is a finite set of transitions}\\
G & \text{ is a set of guards as defined in definition~\eqref{def:guard}}\\
TG &: T \rightarrow G \text{ are the transition guards }\\
E &\subseteq (T \times P \times G) \cup (P \times T \times G) \text{ are the guarded arcs }\\
R &\subseteq P \times T \times G \text{ are the guarded reset arcs }\\
W &: E \rightarrow \mathbb{N}^+ \text{ are arc weights }\\
D &: T \rightarrow \mathbb{N}^+ \text{ are the transition durations}\\
B &: T \rightarrow G \times \mathbb{N}^+ \text{ transition stimulation or boost }\\
MF &: T \rightarrow 2^{G} \text{ must-fire guards for a transition}\\
L &: P \rightarrow \mathbb{N}^+ \text{ specifies maximum number to tokens for each place }
\end{align*}
subject to constraints:
\begin{enumerate}
\item $P \cap T = \emptyset$
\item $R \cap E = \emptyset$
\item Let $t_1 \in T$ and $t_2 \in T$ be any two distinct transitions, then $g1 \in MF(t_1)$ and $g2 \in MF(t_2)$ must not have an interpretation that make both $g1$ and $g2$ true.
\item Let $t \in T$ be a transition, and $gg_t = \{ g: (t,p,g) \in E \} \cup \{ g: (p,t,g) \in E \} \cup \{ g: (p,t,g) \in R \}$ be the set of arc-guards for normal and reset arcs connected to it, then $g_1 \in gg_t, g_2 \in gg_t$ must not have an interpretation that makes both $g_1$ and $g_2$ true. 
\item Let $t \in T$ be a transition, and $gg_t = \{ g: (g,n) \in B(t) \}$ be its stimulation arc-guards, then $g_1 \in gg_t, g_2 \in gg_t$ must not have an interpretation that makes both $g_1$ and $g_2$ true. 
\item Let $t \in T$ be a transition, and $(g,n) \in B(t) : n>1$, then there must not exist a place $p \in P : (p,t,g) \in E, L(p) = 1$. Intuitively, stimulation of binary inputs is not supported.
\end{enumerate}
\end{definition} 

We will make a simplifying assumption that all places are readable by using their place names. Execution of the $PN^G$ occurs in discrete time steps.
\begin{definition}[Marking (or State)]\label{def:gpn:marking}
Marking (or State) of a Guarded-Arc Petri Net $PN^G$ is the token assignment of each place $p_i \in P$. Initial marking $M_0 : P \rightarrow \mathbb{N}^0$, while the token assignment at step $k$ is written as $M_k$.
\end{definition}

Next we define the execution semantics of $PN^G$. First we introduce some terminology that will be used below. Let 
\begin{enumerate}
\item $s_0 = M_0$ represent the the initial marking (or state), $s_k = M_k$ represent the marking (or state) at time step $k$,
\item $s_k(p)$ represent the marking of place $p$ at time step $k$, such that $s_k = [s_k(p_0), \dots, $ $s_k(p_n)]$, where $P=\{p_0, \dots, p_n\}$ \item  $T_k$ be the firing-set that fired in step $k$, 
\item  $b_k(t)$ be the stimulation value applied to a transition $t$ w.r.t. step $k$
\item $en_k$ be the set of enabled transitions in state $s_k$, 
\item $mf_k$ be the set of must-execute transitions in state $s_k$,
\item $consume_k(p,\{t_1,\dots,t_n\})$ be the sum of tokens that will be consumed from place $p$ if transitions $t_1,\dots,t_n$ fired in state $s_k$,
\item $overc_k(\{t_1,\dots,t_n\})$ be the set of places that will have over-consumption of tokens if transitions $t_1,\dots,t_n$ were to fire simultaneously in state $s_k$,
\item $sel_k(fs)$ be the set of possible firing-set choices in state $s_k$ using $\mathit{fs}$ firing style
\item $produce_k(p)$ be the total production of tokens in place $p$ (in state $s_k$) due to actions terminating in state $s_k$, 
\item $s_{k+1}$ be the next state computed from current state $s_k$ due to firing of transition-set $T_k$
\item $min(a,b)$ gives the minimum of numbers $a,b$, such that $min(a,b) = a \text{ if } a < b \text{ or } b \text{ otherwise }$.
\end{enumerate}
Then, the execution semantics of the guarded-arc Petri Net starting from state $s_0$ using firing-style $\mathit{fs}$ is given as follows:
{\small
\begin{align}\label{gpn:exec:semantics}
b_k(t) &= 
\begin{cases}
n &\text{ if } (g,n) \in B(t), s_k \models g\\ 
1 &\text{ otherwise }
\end{cases}\nonumber\\
en_k &= \{t : t \in T, s_k \models TG(t), \forall (p,t,g) \in E, \nonumber\\
          &~~~~~~~~(s_k \models g, s_k(p) \geq W(p,t,g) * b_k(t)) \} \nonumber\\
mf_k &= \{t : t \in en_k, \exists g \in MF(t), s_k \models g \}\nonumber\\
consume_k(p,\{t_1,\dots,t_n\}) &= \sum_{i=1,\dots,n}{W(p,t_i,g) * b_k(t) : (p,t_i,g) \in E, s_k \models g} \nonumber\\
&+ \sum_{i=1,\dots,n}{s_k(p) : (p,t_i,g) \in R, s_k \models g} \nonumber\\
overc_k(\{t_1,\dots,t_n\}) &= \{p : p \in P, s_k(p) < consume_k(p,\{t_1,\dots,t_n\})  \}\nonumber\\
sel_k(1) &=
\begin{cases}
mf_k & \text{ if } |mf_k| = 1\\
\{ \{ t \} : t \in en_k \} & \text{ if } |mf_k| < 1
\end{cases}\nonumber\\
sel_k(*) &= \{ ss : ss \in 2^{en_k}, mf_k \subseteq ss, overc_k(ss) = \emptyset \} \nonumber\\
sel_k(max) &= \{ ss : ss \in 2^{en_k}, mf_k \subseteq ss, overc_k(p,ss) = \emptyset, \nonumber\\
&~~~~~~(\nexists ss' \in 2^{en_k} : ss \subset ss', mf_k \subseteq ss', overc_k(ss') = \emptyset) \} \nonumber\\
T_k &= T'_k : T'_k \in sel_k(\mathit{fs}), (\nexists t \in en_k \setminus T'_k, t \text{ is a reset transition } ) \nonumber\\
produce_k(p) &= \sum_{j=0,\dots,k}{W(t_i,p,g) * b_j(t_i) : (t_i,p,g) \in E, t_i \in T_j, D(t_i)+j = k+1}\nonumber\\
s_{k+1}(p) &= min(s_k(p) - consume_k(p,T_k) + produce_k(p), L(p))
\end{align} 
}

\begin{definition}[Trajectory]\label{def:gpn:traj}
$\sigma = s_0, T_0, s_1, \dots, s_{k}, T_{k}, s_{k+1}$ is a trajectory of $PN^G$ iff given $s_0 = M_0$, each $T_i$ is a possible firing-set in $s_i$ whose firing produces $s_{i+1}$ per $PN^G$'s execution semantics in~\eqref{gpn:exec:semantics}.
\end{definition}

\subsection{Construction of Guarded-Arc Petri Net from a Pathway Specification}
Now we describe how to construct such a Petri Net from a given pathway specification $\mathbf{D}$ with locational fluents. We proceed as follows:
\begin{enumerate}
\item The set of transitions $T = \{ a : a \text{ is an action in } \mathbf{D}\}$.
\item The set of places $P = \{ f : f \text{ is a fluent in } \mathbf{D}\}$.
\item The limit relation for places $L(f) = 
\begin{cases}
1 & \text{ if } `\mathbf{domain~of~} f \mathbf{~is~} binary' \in \mathbf{D}\\
\infty & \text{ otherwise }\\
\end{cases}$.
\item An arc-guard expression $guard\_cond_1,\dots,guard\_cond_n$ is translated into the conjunction $(guard_1,\dots,guard_n)$, where $guard_i$ is obtained from $guard\_cond_i$ as follows:
\begin{enumerate}
\item A {\em guard condition}~\eqref{eqn:mayfire:cond:ge} is translated to $f \geq w$
\item A {\em guard condition}~\eqref{eqn:mayfire:cond:lt} is translated to $f < w$
\item A {\em guard condition}~\eqref{eqn:mayfire:cond:eq} is translated to $f = w$
\item A {\em guard condition}~\eqref{eqn:mayfire:cond:grad} is translated to $f_1 > f_2$
\end{enumerate}
\item A {\em may-execute statement}~\eqref{eqn:may:exec} is translated into guarded arcs as follows:
\begin{enumerate}
\item Let guard $G$ be the translation of arc-guard conditions $guard\_cond_1,\dots,$ $guard\_cond_n$ specified in the {\em may-execute} proposition.
\item The {\em effect clause}~\eqref{eqn:effect:noloc} are translated into arcs as follows:
\begin{enumerate}
\item An effect clause with $e < 0$ is translated into an input arc $(f,a,G)$, with arc-weight $W(f,a,G) = |e|$.
\item An effect clause with $e = `*'$ is translated into a reset set $(f,a,G)$ with arc-weight $W(f,a,G) = *$.
\item An effect clause with $e > 0$ is translated into an output arc $(a,f,G)$, with arc-weight $W(a,f,G) = e$.
\end{enumerate}
\end{enumerate}
\item A {\em must-execute statement}~\eqref{eqn:must:exec} is translated into guarded arcs in the same way as {\em may execute}. 
In addition, it adds an arc-inscription $must\text{-}exec(G)$, where $G$ is the translation of the arc-guard.
\item An {\em inhibit statement}~\eqref{eqn:inhibit} is translated into $IG = (guard_1,\dots,$ $guard_n)$, where $(guard_1,\dots,$ $guard_n)$ is the translation of $(guard\_cond\_1,\dots,$ $guard\_cond_n)$
\item An {\em initial condition statement}~\eqref{eqn:initial:atloc} sets the initial marking of a specific place $p$ to $w$, i.e. $M_0(p) = w$.
\item An {\em duration statement}~\eqref{eqn:dur} adds a $dur(d)$ inscription to transition $a$.
\item A {\em stimulate statement}~\eqref{eqn:stimulate} adds a $stimulate(n,G)$ inscription to transition $a$, where $G$ is the translation of the stimulate guard, a conjunction of $guard\_cond_1,$ $\dots,guard\_cond_n$.
\item A guard $(G_1 \vee \dots \vee G_n) \wedge (\lnot IG_1 \wedge \dots \wedge \lnot IG_m)$ is added to each transition $a$, where $G_i, 1 \leq i \leq n$ is a guard for a {\em may-execute} or a {\em must-execute} statement and $IG_i, 1 \leq i \leq m$ is a guard for an {\em inhibit} statement.
\item A {\em firing style statement}~\eqref{eqn:firing:style} does not visibly appear on a Petri Net diagram, but it specifies the transition firing regime the Petri Net follows.
\end{enumerate}

{\bf Example:} Consider the following pathway specification:
{\small
\begin{equation}
\begin{array}{llll}
\mathbf{domain~of~} &f_1 \mathbf{~is~} integer, &f_2 \mathbf{~is~} integer\\ 
t_1 \mathbf{~may~execute~}\\
\;\;\;\;\;\;\;\;\;\mathbf{causing~} & f_1 \mathbf{~change~value~by~} -1, & f_2 \mathbf{~change~value~by~} +1\\
\;\;\;\;\;\;\;\;\;\mathbf{if~} & f_1 \mathbf{~has~value~lower~than~} 5\\
t_1 \mathbf{~may~execute~}\\
\;\;\;\;\;\;\;\;\;\mathbf{causing~} & f_1 \mathbf{~change~value~by~} -2, & f_2 \mathbf{~change~value~by~} +2\\
\;\;\;\;\;\;\;\;\;\mathbf{if~} & f_1 \mathbf{~has~value~higher~than~} 5\\
\mathbf{duration~of~} & t1 \mathbf{~is~} 10\\
\mathbf{inhibit~} t1 \mathbf{~if~} & f_1 \mathbf{~has~value~higher~than~} 7\\
\mathbf{inhibit~} t1 \mathbf{~if~} & f_1 \mathbf{~has~value~lower~than~} 3\\
\mathbf{normally~stimulate~} t1 & \mathbf{~by~factor~} 3 \\
\;\;\;\;\;\;\;\;\;\mathbf{~if~} & f_2 \mathbf{~has~value~higher~than~} 5\\
\mathbf{initially~} 
& f_1 \mathbf{~has~value~} 0, & f_2 \mathbf{~has~value~} 0,\\
\mathbf{firing~style~} & max 
\end{array}
\end{equation}
}

This pathway specification is encoded as the Petri Net in figure~\ref{fig:guarded:arc:pn}.

\subsection{Guarded-Arc Petri Net with Colored Tokens}\label{sec:gcpn}
Next we extend the Guarded-arc Petri Nets to add Colored tokens. We will use this extension to model pathways with locational fluents.
\begin{figure}[htbp]
   \centering
   \includegraphics[width=0.7\linewidth]{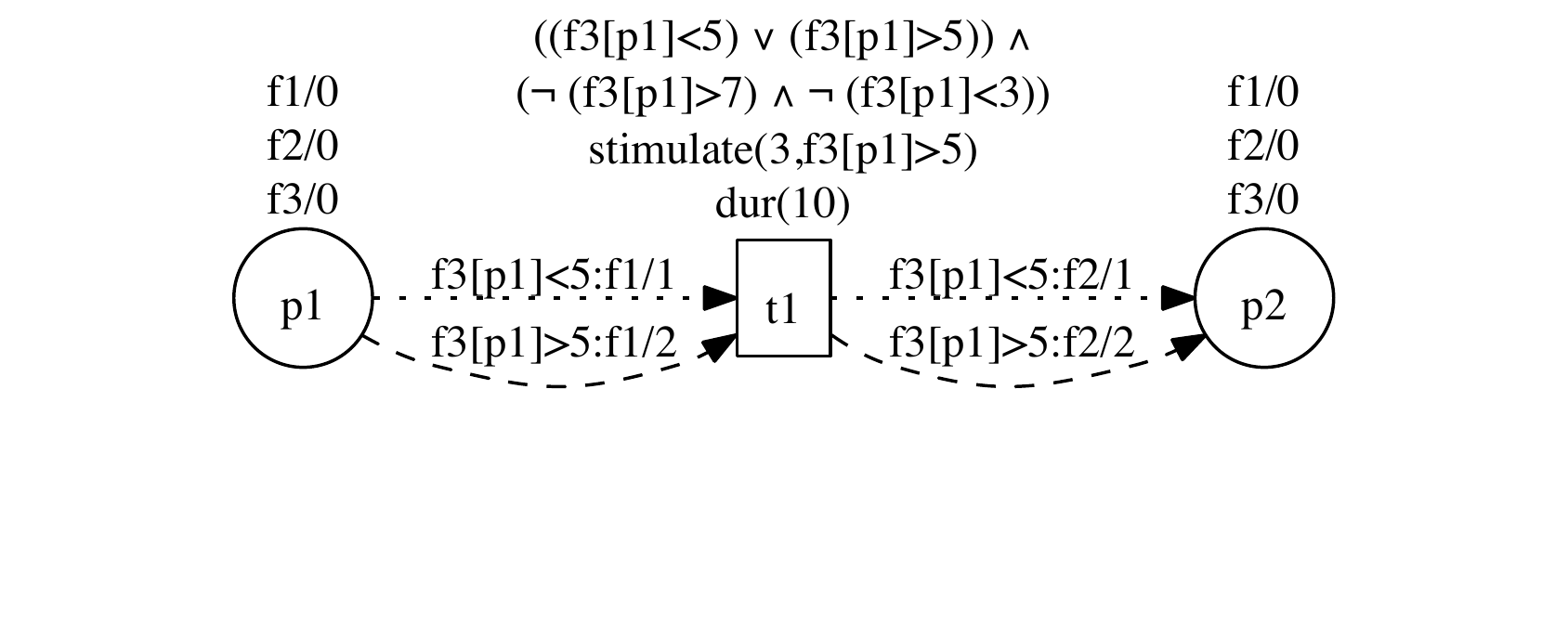} %
   \vspace{-40pt}
   \caption{Example of a guarded-arc Petri Net with colored tokens. }
   \label{fig:guarded:arc:cpn}
\end{figure}

Figure~\ref{fig:guarded:arc:cpn} shows an example of a guarded-arc Petri Net with colored tokens. There are two arc-guard expressions $f3[p1]<5$ and $f3[p1]>5$. When $f3[p1]<5$, $t1$ consumes one token of color $f1$ from place $p1$ and produces one token of color $f2$ in place $p2$. When $f3[p1]>5$, $t1$'s consumption and production of the same colored tokens doubles. The transition $t1$ implicitly gets the guards for each arc represented the {\em or}-ed conditions $((f3[p1]<5) \vee (f3[p1]>5))$. The arc also has two conditions inhibiting it, they are represented by the {\em and}-ed conditions $\lnot(f3[p1]>7) \wedge \lnot(f3[p1]<3)$, where `$\lnot$' represents logical not. Transition $t1$ is stimulated by factor $3$ when $f3[p1]>5$ and it has a duration of $10$ time units.

\begin{definition}[Guarded-Arc Petri Net with Colored Tokens]\label{def:gcpn}
A Guarded-Arc Petri Net with Colored Tokens is a tuple $PN^{GC}=(P,T,C,G,E,R,W,D,B,TG,$ $MF,L)$, such that:
\begin{align*}
P &: \text{finite set of places}\\
T &: \text{finite set of transitions}\\
C &: \text{finite set of colors}\\
G &: \text{set of guards as defined in definition~\eqref{def:guard} with locational fluents}\\
TG &: T \rightarrow G \text{ are the transition guards }\\
E &\subseteq (T \times P \times G) \cup (P \times T \times G) \text{ are the guarded arcs }\\
R &\subseteq P \times T \times G \text{ are the guarded reset arcs }\\
W &: E \rightarrow \langle C,m \rangle \text{ are arc weights; each arc weight is a multiset over } C\\
D &: T \rightarrow \mathbb{N}^+ \text{ are the transition durations}\\
B &: T \rightarrow G \times \mathbb{N}^+ \text{ transition stimulation or boost }\\
MF &: T \rightarrow 2^{G} \text{ must-fire guards for a transition}\\
L &: P \times C \rightarrow \mathbb{N}^+  \text{ specifies maximum number of tokens for each color in each place}\\
\end{align*}
subject to constraints:
\begin{enumerate}
\item $P \cap T = \emptyset$
\item $R \cap E = \emptyset$
\item Let $t_1 \in T$ and $t_2 \in T$ be any two distinct transitions, then $g1 \in MF(t_1)$ and $g2 \in MF(t_2)$ must not have an interpretation that make both $g1$ and $g2$ true.
\item Let $t \in T$ be a transition, and $gg_t = \{ g: (t,p,g) \in E \} \cup \{ g: (p,t,g) \in E \} \cup \{ g: (p,t,g) \in R \}$ be the set of arc-guards for normal and reset arcs connected to it, then $g_1 \in gg_t, g_2 \in gg_t$ must not have an interpretation that makes both $g_1$ and $g_2$ true. 
\item Let $t \in T$ be a transition, and $gg_t = \{ g: (g,n) \in B(t) \}$ be its stimulation guards, then $g_1 \in gg_t, g_2 \in gg_t$ must not have an interpretation that makes both $g_1$ and $g_2$ true. 
\item Let $t \in T$ be a transition, and $(g,n) \in B(t) : n>1$, then there must not exist a place $p \in P$ and a color $c \in C$ such that $(p,t,g) \in E, L(p,c) = 1$. Intuitively, stimulation of binary inputs is not supported.
\end{enumerate}
\end{definition}

\begin{definition}[Marking (or State)]\label{def:gcpn:marking}
Marking (or State) of a Guarded-Arc Petri Net with Colored Tokens $PN^{GC}$ is the colored token assignment of each place $p_i \in P$. Initial marking is written as $M_0 : P \rightarrow \langle C,m \rangle$, while the token assignment at step $k$ is written as $M_k$.
\end{definition}

We make a simplifying assumption that all places are readable by using their place name. Next we define the execution semantics of the guarded-arc Petri Net. First we introduce some terminology that will be used below. Let 
\begin{enumerate}
\item $s_0 = M_0$ represent the the initial marking (or state), $s_k = M_k$ represent the marking (or state) at time step $k$, 
\item $s_k(p)$ represent the marking of place $p$ at time step $k$, such that $s_k = [s_k(p_0),$ $\dots,s_k(p_n)]$, where $P=\{p_0,\dots,p_n\}$. 
\item  $T_k$ be the firing-set that fired in state $s_k$, 
\item  $b_k(t)$ be the stimulation value applied to transition $t$ w.r.t. state $s_k$, 
\item $en_k$ be the set of enabled transitions in state $s_k$, 
\item $mf_k$ be the set of must-fire transitions in state $s_k$,
\item $consume_k(p,\{t_1,\dots,t_n\})$ be the sum of colored tokens that will be consumed from place $p$ if transitions $t_1,\dots,t_n$ fired in state $s_k$,
\item $overc_k(\{t_1,\dots,t_n\})$ be the set of places that will have over-consumption of tokens if transitions $t_1,\dots,t_n$ were to fire simultaneously in state $s_k$,
\item $sel_k(fs)$ be the set of possible firing-sets in state $s_k$ using $\mathit{fs}$ firing style,
\item $produce_k(p)$ be the total production of tokens in place $p$ in state $s_k$ due to actions terminating in state $s_k$, 
\item $s_{k+1}$ be the next state computed from current state $s_k$ and $T_k$, 
\item $m_X(c)$ represents the multiplicity of $c \in C$ in multiset $X=\langle C,m \rangle$, 
\item $c/n$ represents repetition of an element $c$ of a multi-set $n$-times,
\item multiplication of multiset $X= \langle C,m \rangle$ with a number $n$ be defined in terms of multiplication of element multiplicities by $n$, i.e. $\forall c \in C, m_X(c)*n$, and
\item $min(a,b)$ gives the minimum of numbers $a,b$, such that $min(a,b) = a \text{ if } a < b \text{ or } b \text{ otherwise }$.
\end{enumerate}
Then, the execution semantics of the guarded-arc Petri Net starting from state $s_0$ using firing-style $\mathit{fs}$ is given as follows:

{\small
\begin{align}\label{gcpn:exec:semantics}
b_k(t) &= 
\begin{cases}
n & \text{ if } (g,n) \in B(t), s_k \models g\\ 
1 &\text{ otherwise }
\end{cases}\nonumber\\
en_k &= \{ t \in T, s_k \models TG(t), \forall (p,t,g) \in E, \nonumber\\
          &~~~~~~(s_k \models g, s_k(p) \geq W(p,t,g) * b_k(t)) \} \nonumber\\
mf_k &= \{ t \in en_k, \exists g \in MF(t), s_k \models g \}\nonumber\\
consume_k(p,\{t_1,\dots,t_n\}) &= \sum_{i=1,\dots,n}{W(p,t_i,g) * b_k(t) : (p,t_i,g) \in E, s_k \models g} \nonumber\\
&+ \sum_{i=1,\dots,n}{s_k(p) : (p,t_i,g) \in R, s_k \models g} \nonumber\\
overc_k(\{t_1,\dots,t_n\}) &= \{ p \in P : \exists c \in C, m_{s_k(p)}(c) < m_{consume_k(p,\{t_1,\dots,t_n\})}(c)  \}\nonumber\\
sel_k(1) &=
\begin{cases}
mf_k & \text{ if } |mf_k| = 1\\
\{ \{ t \} : t \in en_k \} & \text{ if } |mf_k| < 1
\end{cases}\nonumber\\
sel_k(*) &=
\begin{cases}
ss \in 2^{en_k} : mf_k \subseteq ss, overc_k(ss) = \emptyset %
\end{cases}\nonumber\\
sel_k(max) &=
\begin{cases}
ss \in 2^{en_k} : mf_k \subseteq ss, overc_k(p,ss) = \emptyset, \\
(\nexists ss' \in 2^{en_k} : ss \subset ss', mf_k \subseteq ss', overc_k(ss') = \emptyset)
\end{cases}\nonumber\\
T_k &= T'_k : T'_k \in sel_k(\mathit{fs}), (\nexists t \in en_k \setminus T'_k, t \text{ is a reset transition } ) \nonumber\\
produce_k(p) &= \sum_{j=0,\dots,k}{W(t_i,p,g) * b_j(t_i) : (t_i,p,g) \in E, t_i \in T_j), D(t_i)+j = k+1}\nonumber\\
s_{k+1} &= [ c/n : c \in C, \nonumber\\
             &~~~~n=min(m_{s_k(p)}(c) \nonumber\\
             &~~~~~~~~~~~~ - m_{consume_k(p,T_k)}(c) \nonumber\\
             &~~~~~~~~~~~~ + m_{produce_k(p)}(c), L(p,c)) ]
\end{align}
}

\begin{definition}[Trajectory]\label{def:gcpn:traj}
$\sigma = s_0, T_0, s_1, \dots, s_{k}, T_{k}, s_{k+1}$ is a trajectory of $PN^{GC}$ iff given $s_0=M_0$, each $T_i$ is a possible firing-set in $s_i$ whose firing produces $s_{i+1}$ per $PN^{GC}$'s execution semantics in~\eqref{gcpn:exec:semantics}.
\end{definition}

\subsection{Construction of Guarded-Arc Petri Net with Colored Tokens from a Pathway Specification with Locational Fluents}
Now we describe how to construct such a Petri Net from a given pathway specification $\mathbf{D}$ with locational fluents. We proceed as follows:
\begin{enumerate}
\item The set of transitions $T = \{ a : a \text{ is an action in } \mathbf{D}\}$.
\item The set of colors $C = \{ f : f[l] \text{ is a fluent in } \mathbf{D}\}$.
\item The set of places $P = \{ l : f[l] \text{ is a fluent in } \mathbf{D}\}$.
\item The limit relation for each colored token in a place \\ $L(f,c) = 
\begin{cases}
1 & \text{ if } `\mathbf{domain~of~} f \mathbf{~atloc~} l \mathbf{~is~} binary' \in \mathbf{D}\\
\infty & \text{ otherwise }\\
\end{cases}$.
\item A guard expression $guard\_cond_1,\dots,guard\_cond_n$ is translated into the conjunction $(guard_1,\dots,$ $guard_n)$, where $guard_i$ is obtained from $guard\_cond_i$ as follows:
\begin{enumerate}
\item A {\em guard condition}~\eqref{eqn:mayfire:cond:lt:atloc} is translated to $f[l] < w$
\item A {\em guard condition}~\eqref{eqn:mayfire:cond:eq:atloc} is translated to $f[l] = w$
\item A {\em guard condition}~\eqref{eqn:mayfire:cond:ge:atloc} is translated to $f[l] \geq w$
\item A {\em guard condition}~\eqref{eqn:mayfire:cond:grad:atloc} is translated to $f_1[l_1] > f_2[l_2]$
\end{enumerate}
\item A {\em may-execute statement}~\eqref{eqn:may:exec} is translated into guarded arcs as follows:
\begin{enumerate}
\item Let guard $G$ be the translation of guard conditions $guard\_cond_1,$ $\dots,guard\_cond_n$ specified in the {\em may-execute} proposition.
\item The {\em effect clause}s of the form~\eqref{eqn:effect:atloc} are grouped into input, reset and output effect sets for an action as follows:
\begin{enumerate}
\item The clauses with $e < 0$ for the same place $l$ are grouped together into an input set of $a$ requiring input from place $l$.
\item The clauses with $e = `*'$ for the same place $l$ are grouped together into a reset set of $a$ requiring input from place $l$.
\item The clauses with $e > 0$ for the same place $l$ are grouped together into an output set of $a$ to place $l$.
\end{enumerate}
\item A group of input {\em effect} clauses $\mathit{effect}_1,\dots,\mathit{effect}_m, m > 0$ of the form~\eqref{eqn:effect:atloc} is translated into an input arc $(l,a,G)$, with arc-weight $W(l,a,G) = w^+$, where $w^+$ is the multi-set union of  $f_i/|e_i|$ in $\mathit{effect}_i, 1 \leq i \leq m$.
\item A group of output {\em effect} clauses $\mathit{effect}_1,\dots,\mathit{effect}_m, m > 0$ of the form~\eqref{eqn:effect:atloc} is translated into an output arc $(a,l,G)$, with arc-weight $W(a,l,G) = w^-$, where $w^-$ is the multi-set union of $f_i/e_i$ in $\mathit{effect}_i, 1 \leq i \leq m$.
\item A group of reset {\em effect} clauses $\mathit{effect}_1,\dots,\mathit{effect}_m, m > 0$ of the form~\eqref{eqn:effect:atloc} is translated into a reset arc $(l,a,G)$ with arc-weight $W(l,a,G) = *$.
\end{enumerate}
\item A {\em must-execute statement}~\eqref{eqn:must:exec} is translated into guarded arcs in the same way as {\em may execute}. 
In addition, it adds an arc-inscription $must\text{-}exec(G)$, where $G$ is the guard, which is the translation of $guard\_cond_1,\dots,guard\_cond_n$.
\item An {\em inhibit statement}~\eqref{eqn:inhibit} is translated into $IG = (guard_1,\dots,guard_n)$, where $(guard_1,\dots,guard_n)$ is the translation of $(guard\_cond\_1,\dots,guard\_cond_n)$
\item An {\em initial condition statement}~\eqref{eqn:initial:atloc} sets the initial marking of a specific place $l$ for a specific color $f$ to $w$, i.e. $m_{(M_0(l))}(f) = w$.
\item An {\em duration statement}~\eqref{eqn:dur} adds a $dur(d)$ inscription to transition $a$.
\item A {\em stimulate statement}~\eqref{eqn:stimulate} adds a $stimulate(n,G)$ inscription to transition $a$, where guard $G$ is the translation of its guard expression $guard\_cond_1,$ $\dots,guard\_cond_n$.
\item A guard $(G_1 \vee \dots \vee G_n) \wedge (\lnot IG_1 \wedge \dots \wedge \lnot IG_m)$ is added to each transition $a$, where $G_i, 1 \leq i \leq n$ is the guard for a {\em may-execute} or a {\em must-execute} proposition and $IG_i, 1 \leq i \leq m$ is a guard for an {\em inhibit} proposition.
\item A {\em firing style statement}~\eqref{eqn:firing:style} does not visibly show on a Petri Net, but it specifies the transition firing regime the Petri Net follows.
\end{enumerate}

\subsubsection{Example}
Consider the following pathway specification:
{\small
\begin{align}
&\mathbf{domain~of~} \nonumber\\
&~~~~f_1 \mathbf{~atloc~} l_1 \mathbf{~is~} integer,  f_2 \mathbf{~atloc~} l_1 \mathbf{~is~} integer, \nonumber\\
&~~~~f_3 \mathbf{~atloc~} l_1 \mathbf{~is~} integer, f_1 \mathbf{~atloc~} l_2 \mathbf{~is~} integer, \nonumber\\
&~~~~f_2 \mathbf{~atloc~} l_2 \mathbf{~is~} integer, f_3 \mathbf{~atloc~} l_2 \mathbf{~is~} integer \nonumber\\
&t_1 \mathbf{~may~execute~causing} \nonumber\\
&~~~~~~~~f_1 \mathbf{~atloc~} l_1 \mathbf{~change~value~by~} -1, \nonumber\\
&~~~~~~~~f_2 \mathbf{~atloc~} l_2 \mathbf{~change~value~by~} +1 \nonumber\\
&~~~~\mathbf{if~} f_3 \mathbf{~atloc~} l_1 \mathbf{~has~value~lower~than~} 5 \nonumber\\
&t_1 \mathbf{~may~execute~causing} \nonumber\\
&~~~~~~~~f_1 \mathbf{~atloc~} l_1 \mathbf{~change~value~by~} -2, \nonumber\\
&~~~~~~~~f_2 \mathbf{~atloc~} l_2 \mathbf{~change~value~by~} +2 \nonumber\\
&~~~~\mathbf{if~} f_3 \mathbf{~atloc~} l_1 \mathbf{~has~value~higher~than~} 5 \nonumber\\
&\mathbf{duration~of~} t1 \mathbf{~is~} 10 \nonumber\\
&\mathbf{inhibit~} t1 \mathbf{~if~} f_3 \mathbf{~atloc~} l_1 \mathbf{~has~value~higher~than~} 7 \nonumber\\
&\mathbf{inhibit~} t1 \mathbf{~if~} f_3 \mathbf{~atloc~} l_1 \mathbf{~has~value~lower~than~} 3 \nonumber\\
&\mathbf{normally~stimulate~} t1 \mathbf{~by~factor~} 3 \nonumber\\
&~~~~\mathbf{~if~} f_2 \mathbf{~atloc~} l_2 \mathbf{~has~value~higher~than~} 5 \nonumber\\
&\mathbf{initially~} \nonumber\\
&~~~~f_1 \mathbf{~atloc~} l_1 \mathbf{~has~value~} 0, f_1 \mathbf{~atloc~} l_2 \mathbf{~has~value~} 0, \nonumber\\
&~~~~f_2 \mathbf{~atloc~} l_1 \mathbf{~has~value~} 0, f_2 \mathbf{~atloc~} l_2 \mathbf{~has~value~} 0, \nonumber\\
&~~~~f_3 \mathbf{~atloc~} l_1 \mathbf{~has~value~} 0, f_3 \mathbf{~atloc~} l_2 \mathbf{~has~value~} 0 \nonumber\\
&\mathbf{firing~style~} max 
\end{align}
}

\section{Syntax of Query Language (BioPathQA-QL)}\label{qlang:syntax}
The alphabet of query language $\mathcal{Q}$ consists of the same sets $A,F,L$ from $\mathcal{P}$ representing actions, fluents, and locations, respectively; a fixed set of reserved keywords $K$ shown in bold in syntax below; a fixed set $\{ `:', `;', `,' , `''\}$ of punctuations; a fixed set of $\{`<',`>',`='\}$ of directions; and constants. Our query language asks questions about biological entities and processes in a biological pathway described through the pathway specification language. This is our domain description. A query statement is composed of a query description (the quantity, observation, or condition being sought by the question), interventions (changes to the pathway), observations (about states and actions of the pathway), and initial setup conditions.

The query statement is evaluated against the trajectories of the pathway, generated by simulating the pathway. These trajectories are modified by the initial setup and interventions. The resulting trajectories are then filtered to retain only those which satisfy the observations specified in the query statement. 

A query statement can take various forms: The simplest queries do not modify the pathway and check if a specific observation is true on a trajectory or not. An observation can be a point observation or an interval observation depending upon whether they can be evaluated w.r.t. a point or an interval on a trajectory. More complex queries modify the pathway in various ways and ask for comparison of an observation before and after such modification.

Following query statements about the rate of production of  `$bpg13$' illustrate the kind of queries that can be asked from our system about the specified glycolysis pathway as given in~\cite[Figure 9.9]{CampbellBook}.

Determine if `$n$' is a possible rate of production of substance `$bpg13$':
{\footnotesize
\begin{align}
\mathbf{rate~} & \mathbf{of~production~of~} 'bpg13' \mathbf{~is~} n \nonumber\\
&\mathbf{when~observed~between~time~step~} 0 \mathbf{~and~} \mathbf{~time~step~} k ; &
\end{align}
}

Determine if `$n$' is a possible rate of production of substance `$bpg13$' in a pathway when it is being supplied with a limited supply of an upstream substance `$f16bp$':
{\footnotesize
\begin{align}
\mathbf{rate}&\mathbf{~of~production~of~} 'bpg13' \mathbf{~is~} n \nonumber\\
&\mathbf{when~observed~between~time~step~} 0 \mathbf{~and~} \mathbf{~time~step~} k ; \nonumber\\
&\mathbf{using~initial~setup:~} \mathbf{set~value~of~} `f16bp' \mathbf{~to~} 5 ; &
\end{align}
}

Determine if `$n$' is a possible rate of production of substance `$bpg13$' in a pathway when it is being supplied with a steady state supply of an upstream substance `$f16bp$' at the rate of $5$ units per time-step:
{\footnotesize
\begin{align}
\mathbf{rate}&\mathbf{~of~production~of~} 'bpg13' \mathbf{~is~} n \nonumber\\
&\mathbf{when~observed~between~time~step~} 0 \mathbf{~and~} \mathbf{~time~step~} k ; \nonumber\\
&\mathbf{using~initial~setup:~} \mathbf{continuously~supply~} `f16bp' \mathbf{~in~quantity~} 5 ; &
\end{align}
}

Determine if `$n$' is a possible rate of production of substance `$bpg13$' in a pathway when it is being supplied with a steady state supply of an upstream substance `$f16bp$' at the rate of $5$ units per time-step and the pathway is modified to remove all quantity of the substance '$dhap$' as soon as it is produced:
{\footnotesize
\begin{align}
\mathbf{rate}&\mathbf{~of~production~of~} 'bpg13' \mathbf{~is~} n \nonumber\\
&\mathbf{when~observed~between~time~step~} 0 \mathbf{~and~} \mathbf{~time~step~} k ; \nonumber\\
&\mathbf{due~to~interventions:~} \mathbf{remove~} `dhap' \mathbf{~as~soon~as~produced} ;\nonumber\\
&\mathbf{using~initial~setup:~} \mathbf{continuously~supply~} `f16bp' \mathbf{~in~quantity~} 5 ; &
\end{align}
}

Determine if `$n$' is a possible rate of production of substance `$bpg13$' in a pathway when it is being supplied with a steady state supply of an upstream substance `$f16bp$' at the rate of $5$ units per time-step and the pathway is modified to remove all quantity of the substance '$dhap$' as soon as it is produced and a non-functional pathway process / reaction named `$t5b$':
{\footnotesize
\begin{align}
\mathbf{rate}&\mathbf{~of~production~of~} 'bpg13' \mathbf{~is~} n \nonumber\\
&\mathbf{when~observed~between~time~step~} 0 \mathbf{~and~} \mathbf{~time~step~} k ; \nonumber\\
&\mathbf{due~to~interventions:~} \mathbf{remove~} `dhap' \mathbf{~as~soon~as~produced}  ;\nonumber\\
&\mathbf{due~to~observations:~} `t5b' \mathbf{~does~not~occur} ; \nonumber\\
&\mathbf{using~initial~setup:~} \mathbf{continuously~supply~} `f16bp' \mathbf{~in~quantity~} 5 ; &
\end{align}
}

Determine if `$n$' is the average rate of production of substance `$bpg13$' in a pathway when it is being supplied with a steady state supply of an upstream substance `$f16bp$' at the rate of $5$ units per time-step and the pathway is modified to remove all quantity of the substance '$dhap$' as soon as it is produced and a non-functional pathway process / reaction named `$t5b$':
{\footnotesize
\begin{align}
\mathbf{average}&\mathbf{~rate~of~production~of~} 'bpg13' \mathbf{~is~} n \nonumber\\
&\mathbf{when~observed~between~time~step~} 0 \mathbf{~and~} \mathbf{~time~step~} k ; \nonumber\\
&\mathbf{due~to~interventions:~} \mathbf{remove~} `dhap' \mathbf{~as~soon~as~produced} ;\nonumber\\
&\mathbf{due~to~observations:~} `t5b' \mathbf{~does~not~occur} ; \nonumber\\
&\mathbf{using~initial~setup:~} \mathbf{continuously~supply~} `f16bp' \mathbf{~in~quantity~} 5 ; &
\end{align}
}

Determine if `$d$'  is the direction of change in the average rate of production of substance `$bpg13$' with a steady state supply of an upstream pathway input when compared with a pathway with the same steady state supply of an upstream pathway input, but in which the substance `$dhap$' is removed from the pathway as soon as it is produced and pathway process / reaction called `$t5b$' is non-functional:
{\footnotesize
\begin{align}
\mathbf{dir}&\mathbf{ection~of~change~in~average~rate~of~production~of~} 'bpg13' \mathbf{~is~} d \nonumber\\
&\mathbf{when~observed~between~time~step~} 0 \mathbf{~and~} \mathbf{~time~step~} k ; \nonumber\\
&\mathbf{comparing~nominal~pathway~with~modified~pathway~obtained~} \nonumber\\
&\;\;\;\;\;\;\mathbf{due~to~interventions:~} \mathbf{remove~} `dhap' \mathbf{~as~soon~as~produced~} ;\nonumber\\
&\;\;\;\;\;\;\mathbf{due~to~observations:~} `t5b' \mathbf{~does~not~occur} ; \nonumber\\
&\mathbf{using~initial~setup:~} \mathbf{continuously~supply~} `f16bp' \mathbf{~in~quantity~} 5 ; &
\end{align}
}

Queries can also be about actions, as illustrated in the following examples.
Determine if action `$t5b$' ever occurs when there is a continuous supply of `$f16bp$' is available and `$t5a$' is disabled:
{\footnotesize
\begin{align}
`t5b' &\mathbf{~occurs~} ;\nonumber\\
&\mathbf{due~to~interventions:~} \mathbf{disable~} `t5a' ; \nonumber\\
&\mathbf{using~initial~setup:~} \mathbf{continuously~produce~} `f16bp' \mathbf{~in~quantity~} 5 ; &
\end{align} %
}

Determine if glycolysis ($`gly'$) gets replaced with beta-oxidation ($`box'$) when sugar ($`sug'$) is exhausted but fatty acids ($`fac'$) are available, when starting with a fixed initial supply of sugar and fatty acids in quantity $5$: 
{\footnotesize
\begin{align}
&`gly' \mathbf{~switches~to~} `box' \mathbf{~when~} \nonumber\\
&~~~~\mathbf{value~of~} 'sug' \mathbf{~is~} 0, \nonumber\\
&~~~~\mathbf{value~of~} 'fac' \mathbf{~is~higher~than~} 0 \nonumber\\
&~~~~\mathbf{in~all~trajectories} ; \nonumber\\
&\mathbf{due~to~observations:} \nonumber\\
&~~~~`gly' \mathbf{~switches~to~} `box' \nonumber\\
&\mathbf{using~initial~setup:~} \nonumber\\
&~~~~\mathbf{set~value~of~} `sug' \mathbf{~to~} 5, \nonumber\\
&~~~~\mathbf{set~value~of~} `fac' \mathbf{~to~} 5 ; &
\end{align}
}

Next we define various syntactic elements of a query statement, give their intuitive meaning, and how these components fit together to form a query statement. We will define the formal semantics in a later section. Note that some of the single-trajectory queries can be represented as LTL formulas. However, we have chosen to keep the current representation as it is more intuitive for our biological domain.

In the following description, $f_i$'s are fluents, $l_i$'s are locations, $n$'s are numbers, $q$'s are positive integer numbers, $d$ is one of the directions from $\{<,>,=\}$.

\begin{definition}[Point]
A point is a time-step on the trajectory. It has the form:
{\footnotesize
\begin{align}
&\label{dqa:syn:point}\mathbf{time~step~} ts 
\end{align}
}
\end{definition}

\begin{definition}[Interval]
An interval is a sub-sequence of time-steps on a trajectory. It has the form:
{\footnotesize
\begin{align}
&\label{dqa:syn:interval}\langle point \rangle \mathbf{~and~} \langle point \rangle 
\end{align}
}
\end{definition}

\begin{definition}[Aggregate Operator (aggop)]
An aggregate operator computes an aggregate quantity over a sequence of values. It can be one of the following:
{\footnotesize
\begin{align}
&\label{dqa:syn:aggop:min} \mathbf{minimum}\\
&\label{dqa:syn:aggop:max} \mathbf{maximum}\\
&\label{dqa:syn:aggop:avg} \mathbf{average}
\end{align}
}
\end{definition}

\begin{definition}[Quantitative Interval Formula]
A quantitative interval formula is a formula that is evaluated w.r.t. an interval on a trajectory for some quantity $n$.
{\footnotesize
\begin{align}
&\label{dqa:syn:quant:interval:formula:rate:prod}\mathbf{rate~of~production~of~} f \mathbf{~is~} n\\
&\label{dqa:syn:quant:interval:formula:rate:prod:atloc}\mathbf{rate~of~production~of~} f \mathbf{~atloc~} l \mathbf{~is~} n\\
&\label{dqa:syn:quant:interval:formula:rate:fire} \mathbf{rate~of~firing~of~} a \mathbf{~is~} n\\
&\label{dqa:syn:quant:interval:formula:total:prod}\mathbf{total~production~of~} f \mathbf{~is~} n\\
&\label{dqa:syn:quant:interval:formula:total:prod:atloc}\mathbf{total~production~of~} f \mathbf{~atloc~} l \mathbf{~is~} n
\end{align}
}
\end{definition}
Intuitively, the rate of production of a fluent $f$ in interval $s_i,\dots,s_j$ on a trajectory $s_0,T_0,\dots,T_{k-1},s_k$ is $n=(s_j(f)-s_i(f))/(j-i)$; rate of firing of an action $a$ in interval $s_i,\dots,s_j$ is $n=|\{T_l : a \in T_l, i \leq l \leq j-1\}|/(j-i)$; and total production of a fluent $f$ in interval $s_i,\dots,s_j$ is $n=s_j(f)-s_i(f)$. If the given $n$ equals the computed $n$, then the formula holds. The same intuition extends to locational fluents, except that fluent $f$ is replaced by $f[l]$, e.g. rate of production of fluent $f$ at location $l$ in interval $s_i,\dots,s_j$ on a trajectory is $n=(s_j(f[l])-s_i(f[l]))/(j-i)$. In biological context, the actions represent reactions and fluents substances used in these reactions. The rate and total production formulas are used in aggregate observations to determine if reactions are slowing down or speeding up during various portions of a simulation. 

\begin{definition}[Quantitative Point Formula]
A quantitative point formula is a formula that is evaluated w.r.t. a point on a trajectory for some quantity $n$.
{\footnotesize
\begin{align}
&\label{dqa:syn:quant:point:formula:gt}\mathbf{value~of~} f \mathbf{~is~higher~than~} n\\
&\label{dqa:syn:quant:point:formula:gt:atloc}\mathbf{value~of~} f \mathbf{~atloc~} l \mathbf{~is~higher~than~} n\\
&\label{dqa:syn:quant:point:formula:lt}\mathbf{value~of~} f \mathbf{~is~lower~than~} n\\
&\label{dqa:syn:quant:point:formula:lt:atloc}\mathbf{value~of~} f \mathbf{~atloc~} l \mathbf{~is~lower~than~} n\\
&\label{dqa:syn:quant:point:formula:eq}\mathbf{value~of~} f \mathbf{~is~} n\\
&\label{dqa:syn:quant:point:formula:eq:atloc}\mathbf{value~of~} f \mathbf{~atloc~} l \mathbf{~is~} n
\end{align}
}
\end{definition}

\begin{definition}[Qualitative Interval Formula]
A qualitative interval formula is a formula that is evaluated w.r.t. an interval on a trajectory.
{\footnotesize
\begin{align}
&\label{dqa:syn:qual:interval:formula:accum} f \mathbf{~is~accumulating~}\\
&\label{dqa:syn:qual:interval:formula:accum:atloc} f \mathbf{~is~accumulating~} \mathbf{~atloc~} l \\
&\label{dqa:syn:qual:interval:formula:decr} f \mathbf{~is~decreasing~}\\
&\label{dqa:syn:qual:interval:formula:decr:atloc} f \mathbf{~is~decreasing~} \mathbf{~atloc~} l 
\end{align}
}
\end{definition}
Intuitively, a fluent $f$ is accumulating in interval $s_i,\dots,s_j$ on a trajectory if $f$'s value monotonically increases during the interval. A fluent $f$ is decreasing in interval $s_i,\dots,s_j$ on a trajectory if $f$'s value monotonically decreases during the interval. The same intuition extends to locational fluents by replacing $f$ with $f[l]$.

\begin{definition}[Qualitative Point Formula]
A qualitative point formula is a formula that is evaluated w.r.t. a point on a trajectory.
{\footnotesize
\begin{align}
&\label{dqa:syn:qual:point:formula:occurs} a \mathbf{~occurs}\\
&\label{dqa:syn:qual:point:formula:notoccurs} a \mathbf{~does~not~occur}\\
&\label{dqa:syn:qual:point:formula:switches} a1 \mathbf{~switches~to~} a2\\
\end{align}
}
\end{definition}
Intuitively, an action occurs at a point $i$ on the trajectory if $a \in T_i$; an action does not occur at point $i$ if $a \notin T_i$; an action $a1$ switches to $a2$ at point $i$ if $a1 \in T_{i-1}$, $a2 \notin T_{i-1}$, $a1 \notin T_i$, $a2 \in T_i$.

\begin{definition}[Quantitative All Interval Formula]\label{dqa:syn:quant:all:interval:formula}
A quantitative all interval formula is a formula that is evaluated w.r.t. an interval on a set of trajectories $\sigma_1,\dots,\sigma_m$ and corresponding quantities $r_1,\dots,r_m$. 
{\footnotesize
\begin{align}
&\label{dqa:syn:quant:all:interval:formula:rate:prod}\mathbf{~rates~of~production~of~} f \mathbf{~are~} [r_1,\dots,r_m]\\
&\label{dqa:syn:quant:all:interval:formula:rate:prod:atloc}\mathbf{~rates~of~production~of~} f \mathbf{~altoc~} l \mathbf{~are~} [r_1,\dots,r_m]\\
&\label{dqa:syn:quant:all:interval:formula:rate:fire}\mathbf{~rates~of~firing~of~} f \mathbf{~are~} [r_1,\dots,r_m]\\
&\label{dqa:syn:quant:all:interval:formula:total:prod}\mathbf{~totals~of~production~of~} f \mathbf{~are~} [r_1,\dots,r_m]\\
&\label{dqa:syn:quant:all:interval:formula:total:prod:atloc}\mathbf{~totals~of~production~of~} f \mathbf{~altoc~} l \mathbf{~are~} [r_1,\dots,r_m]
\end{align}
}
\end{definition}
Intuitively, a quantitative all interval formula holds on some interval $[i,j]$ over a set of trajectories $\sigma_1,\dots,\sigma_m$ for values $[r_1,\dots,r_m]$ if for each $r_x$ the corresponding quantitative interval formula holds in interval $[i,j]$ in trajectory $\sigma_x$ . For example, $\mathbf{rates~of~production~of~} f \mathbf{~are~} [r_1,\dots,r_m]$ in interval $[i,j]$ over a set of trajectories $\sigma_1,\dots,\sigma_m$ if for each $x \in \{1 \dots m\}$, $\textbf{rate~of~production~of~} f \mathbf{~is~} r_x$ in interval $[i,j]$ in trajectory $\sigma_x$. 

\begin{definition}[Quantitative All Point Formula]\label{dqa:syn:quant:all:point:formula}
A quantitative all point formula is a formula that is evaluated w.r.t. a point on a set of trajectories $\sigma_1,\dots,\sigma_m$ and corresponding quantities $r_1,\dots,r_m$. 
{\footnotesize
\begin{align}
&\label{dqa:syn:quant:all:point:formula:eq}\mathbf{values~of~} f \mathbf{~are~} [r_1,\dots,r_m]\\
&\label{dqa:syn:quant:all:point:formula:eq:atloc}\mathbf{values~of~} f \mathbf{~atloc~} l \mathbf{~are~} [r_1,\dots,r_m]
\end{align}
}
\end{definition}
Intuitively, a quantitative all point formula holds at some point $i$ over a set of trajectories $\sigma_1,\dots,\sigma_m$ for values $[r_1,\dots,r_m]$ if for each $r_x$ the corresponding quantitative point formula holds at point $i$ in trajectory $\sigma_x$ . For example, $\mathbf{values~of~} f \mathbf{~are~} [r_1,\dots,r_m]$ at point $i$ over a set of trajectories $\sigma_1,\dots,\sigma_m$ if for each $x \in \{1\dots m\}$, $\textbf{value~of~} f \mathbf{~is~} r_x$ at point $i$ in trajectory $\sigma_x$.

\begin{definition}[Quantitative Aggregate Interval Formula]\label{dqa:syn:quant:agg:interval:formula}
A quantitative aggregate interval formula is a formula that is evaluated w.r.t. an interval on a set of trajectories $\sigma_1,\dots,\sigma_m$ and an aggregate value $r$, where $r$ is the aggregate of $[r_1,\dots,r_m]$ using $aggop$. 
{\footnotesize
\begin{align}
&\label{dqa:syn:quant:agg:interval:formula:rate:prod}\langle aggop \rangle \mathbf{~rate~of~production~of~} f \mathbf{~is~} n\\
&\label{dqa:syn:quant:agg:interval:formula:rate:prod:atloc}\langle aggop \rangle \mathbf{~rate~of~production~of~} f \mathbf{~atloc~} l \mathbf{~is~} n\\
&\label{dqa:syn:quant:agg:interval:formula:rate:fire}\langle aggop \rangle  \mathbf{~rate~of~firing~of~} a \mathbf{~is~} n\\
&\label{dqa:syn:quant:agg:interval:formula:total:prod}\langle aggop \rangle \mathbf{~total~production~of~} f \mathbf{~is~} n\\
&\label{dqa:syn:quant:agg:interval:formula:total:prod:atloc}\langle aggop \rangle \mathbf{~total~production~of~} f \mathbf{~atloc~} l \mathbf{~is~} n
\end{align}
}
\end{definition}
Intuitively, a quantitative aggregate interval formula holds on some interval $[i,j]$ over a set of trajectories $\sigma_1,\dots,\sigma_m$ for a value $r$ if there exist $[r_1,\dots,r_m]$ whose aggregate value per $aggop$ is $r$, such that for each $r_x$ the quantitative interval formula (corresponding to the quantitative aggregate interval formual) holds in interval $[i,j]$ in trajectory $\sigma_x$. For example, $average \mathbf{~rate~of~}$ $\mathbf{production~of~} f \mathbf{~is~} r$ in interval $[i,j]$ over a set of trajectories $\sigma_1,\dots,\sigma_m$ if  $r=(r_1+\dots+r_m)/m$ and  for each $x \in \{1\dots m\}$, $\textbf{rate~of~production~of~} f \mathbf{~is~} r_x$ in interval $[i,j]$ in trajectory $\sigma_x$.

\begin{definition}[Quantitative Aggregate Point Formula]\label{dqa:syn:quant:agg:point:formula}
A quantitative aggregate point formula is a formula that is evaluated w.r.t. a point on a set of trajectories $\sigma_1,\dots,\sigma_m$ and an aggregate value $r$, where $r$ is the aggregate of $[r_1,\dots,r_m]$ using $aggop$.
{\footnotesize
\begin{align}
&\label{dqa:syn:quant:agg:point:formula:eq}\langle aggop \rangle \mathbf{~value~of~} f \mathbf{~is~} r\\
&\label{dqa:syn:quant:agg:point:formula:eq:atloc}\langle aggop \rangle \mathbf{~value~of~} f \mathbf{~atloc~} l \mathbf{~is~} r
\end{align}
}
\end{definition}
Intuitively, a quantitative aggregate point formula holds at some point $i$ over a set of trajectories $\sigma_1,\dots,\sigma_m$ for a value $r$ if there exist $[r_1,\dots,r_m]$ whose aggregate value per $aggop$ is $r$, such that for each $r_x$ the quantitative point formula (corresponding to the quantitative aggregate point formual) holds at point $i$ in trajectory $\sigma_x$. For example, $average \mathbf{~value~of~} f \mathbf{~is~} r$ at point $i$ over a set of trajectories $\sigma_1,\dots,\sigma_m$ if  $r=(r_1+\dots+r_m)/m$ and  for each $x \in \{1\dots m\}$, $\textbf{value~of~} f \mathbf{~is~} r_x$ at point $i$ in trajectory $\sigma_x$.

\begin{definition}[Quantitative Comparative Aggregate Interval Formula]\label{dqa:syn:quant:cmpr:agg:interval:formula}
A quantitative comparative aggregate interval formula is a formula that is evaluated w.r.t. an interval over two sets of trajectories and a direction `$d$' such that `$d$' relates the two sets of trajectories.
{\footnotesize
\begin{align}
&\label{dqa:syn:quant:cmpr:agg:interval:formula:rate:prod}\mathbf{direction~of~change~in~} \langle aggop \rangle \mathbf{~rate~of~production~of~} f \mathbf{~is~} d\\
&\label{dqa:syn:quant:cmpr:agg:interval:formula:rate:prod:atloc}\mathbf{direction~of~change~in~} \langle aggop \rangle \mathbf{~rate~of~production~of~} f \mathbf{~atloc~} l \mathbf{~is~} d\\
&\label{dqa:syn:quant:cmpr:agg:interval:formula:rate:fire}\mathbf{direction~of~change~in~} \langle aggop \rangle  \mathbf{~rate~of~firing~of~} a \mathbf{~is~} d\\
&\label{dqa:syn:quant:cmpr:agg:interval:formula:total:prod}\mathbf{direction~of~change~in~} \langle aggop \rangle \mathbf{~total~production~of~} f \mathbf{~is~} d\\
&\label{dqa:syn:quant:cmpr:agg:interval:formula:total:prod:atloc}\mathbf{direction~of~change~in~} \langle aggop \rangle \mathbf{~total~production~of~} f \mathbf{~atloc~} l \mathbf{~is~} d
\end{align}
}
\end{definition}
Intuitively, a comparative quantitative aggregate interval formula compares two quantitative interval formulas over using the direction $d$ over a given interval.

\begin{definition}[Quantitative Comparative Aggregate Point Formula]\label{dqa:syn:quant:cmpr:agg:point:formula}
A quantitative comparative aggregate point formula is a formula that is evaluated w.r.t. a point over two sets of trajectories and a direction `$d$' such that `$d$' relates the two sets of trajectories.
{\footnotesize
\begin{align}
&\label{dqa:syn:quant:cmpr:agg:point:formula:eq}\mathbf{direction~of~change~in~} \langle aggop \rangle \mathbf{~value~of~} f \mathbf{~is~} d\\
&\label{dqa:syn:quant:cmpr:agg:point:formula:eq:atloc}\mathbf{direction~of~change~in~} \langle aggop \rangle \mathbf{~value~of~} f \mathbf{~atloc~} l \mathbf{~is~} d
\end{align}
}
\end{definition}
Intuitively, a comparative quantitative aggregate point formula compares two quantitative point formulas over using the direction $d$ at a given point.

\begin{definition}[Simple Interval Formula]\label{dqa:syn:simple:interval:formula}
A simple interval formula takes the following forms:
{\footnotesize
\begin{align}
&\langle \text{quantitative interval formula} \rangle \\
&\langle \text{qualitative interval formula} \rangle
\end{align}
}
\end{definition}

\begin{definition}[Simple Point Formula]\label{dqa:syn:simple:point:formula}
A simple interval formula takes the following forms:
{\footnotesize
\begin{align}
&\langle \text{quantitative point formula} \rangle \\
&\langle \text{qualitative point formula} \rangle 
\end{align}
}
\end{definition}

\begin{definition}[Internal Observation Description]\label{dqa:syn:internal:obs:desc}
An internal observation description takes the following form:
{\footnotesize
\begin{align}
&\langle \text{simple point formula} \rangle \\
&\langle \text{simple point formula} \rangle \mathbf{~at~} \langle point \rangle \\
&\langle \text{simple interval formula} \rangle \\
&\langle \text{simple interval formula} \rangle \mathbf{~when~observed~between~} \langle interval \rangle
\end{align}
}
\end{definition}

\begin{definition}[Simple Point Formula Cascade]\label{dqa:syn:simple:point:formula:cascade}
A simple point formula cascade takes the following form:
{\footnotesize
\begin{align}
\label{dqa:syn:point:formula:cascade:after}
&\langle \text{simple point formula} \rangle_0 \nonumber\\
&~~~~~~~~\mathbf{~after~} \langle \text{simple point formula} \rangle_{1,1}, \dots, \langle \text{simple point formula} \rangle_{1,n_1} \nonumber\\
&~~~~~~~~\vdots \nonumber\\
&~~~~~~~~\mathbf{~after~} \langle \text{simple point formula} \rangle_{u,1}, \dots, \langle \text{simple point formula} \rangle_{u,n_u} \\
\label{dqa:syn:point:formula:cascade:when}
&\langle \text{simple point formula} \rangle_0  \nonumber\\
&~~~~~~~~\mathbf{~when~} \langle \text{simple point formula} \rangle_{1,1}, \dots, \langle \text{simple point formula} \rangle_{1,n_1}\\
\label{dqa:syn:point:formula:cascade:when:cond}
&\langle \text{simple point formula} \rangle_0  \nonumber\\
&~~~~~~~~\mathbf{~when~} \langle \text{cond} \rangle
\end{align}
}
where $u \geq 1$ and `$\text{cond}$' is a conjunction of $\text{simple point formula}$s that is true in the same point as the $\text{simple point formula}$.
\end{definition}
Intuitively, the simple point formula cascade~\eqref{dqa:syn:point:formula:cascade:after} holds if  a given sequence of point formulas hold in order in a trajectory. 
Intuitively, simple point formula cascade~\eqref{dqa:syn:point:formula:cascade:when} holds if a given point formula occurs at the same point as a set of simple point formulas in a trajectory. 
Note that these formulas and many other of our single trajectory formulas can be replaced by an LTL~\cite{manna1992temporal} formula, but we have kept this syntax as it is more relevant to the question answering needs in the biological domain.

\begin{definition}[Query Description]\label{dqa:syn:qdesc}
A query description specifies a non-comparative observation that can be made either on a trajectory or a set of trajectories.
{\footnotesize
\begin{align}
&\label{dqa:syn:qdesc:agg:interval} \langle \text{quantitative aggregate interval formula} \rangle \mathbf{~when~observed~between~} \langle \text{interval} \rangle \\
&\label{dqa:syn:qdesc:agg:point} \langle \text{quantitative aggregate point formula} \rangle \mathbf{~when~observed~at~} \langle \text{point} \rangle \\
&\label{dqa:syn:qdesc:all:interval} \langle \text{quantitative all interval formula} \rangle \mathbf{~when~observed~between~} \langle \text{interval} \rangle \\
&\label{dqa:syn:qdesc:all:point} \langle \text{quantitative all point formula} \rangle \mathbf{~when~observed~at~} \langle \text{point} \rangle \\
&\label{dqa:syn:qdesc:internal:obs:desc} \langle \text{internal observation description} \rangle \\
&\label{dqa:syn:qdesc:internal:obs:desc:all} \langle \text{internal observation description} \rangle \mathbf{~in~all~trajectories}\\
&\label{dqa:syn:qdesc:simple:point:cascade} \langle \text{simple point formula cascade} \rangle\\
&\label{dqa:syn:qdesc:simple:point:cascade:all} \langle \text{simple point formula cascade} \rangle \mathbf{~in~all~trajectories}
\end{align}
}
\end{definition}

The single trajectory observations are can be represented using LTL formulas, but we have chosen to keep them in this form for ease of use by users from the biological domain.

\begin{definition}[Comparative Query Description]\label{dqa:syn:cmpr:qdesc}
A comparative query description specifies a comparative observation that can be made w.r.t. two sets of trajectories.
{\footnotesize
\begin{align}
&\label{dqa:syn:cmpr:qdesc:agg:interval} \langle \text{quantitative comparative aggregate interval formula} \rangle \nonumber\\
&\;\;\;\;\;\;\;\mathbf{~when~observed~between~} \langle \text{interval} \rangle \\
&\label{dqa:syn:cmpr:qdesc:agg:point} \langle \text{quantitative comparative aggregate point formula} \rangle \nonumber\\
&\;\;\;\;\;\;\;\mathbf{~when~observed~at~} \langle \text{point} \rangle 
\end{align}
}
\end{definition}

\begin{definition}[Intervention]\label{dqa:syn:interv}
Interventions define modifications to domain descriptions.
{\footnotesize
\begin{align}
&\label{idesc:remove:immed}\mathbf{remove~} f_1 \mathbf{~as~soon~as~produced} \\
&\label{idesc:remove:immed:atloc}\mathbf{remove~} f_1 \mathbf{~atloc~} l_1 \mathbf{~as~soon~as~produced}  \\
&\label{idesc:disable}\mathbf{disable~} a_2  \\
&\label{idesc:cont:xform:x2y}\mathbf{continuously~transform~} f_1\mathbf{~in~quantity~} q_1 \mathbf{~to~} f_2   \\
&\label{idesc:cont:xform:x2y:loc}\mathbf{continuously~transform~} f_1\mathbf{~atloc~} l_1 \mathbf{~in~quantity~} q_1 \mathbf{~to~} f_2 \mathbf{~atloc~} l_2   \\
&\label{idesc:inhibits}\mathbf{make~} f_3 \mathbf{~inhibit~} a_3  \\
&\label{idesc:inhibits:atloc}\mathbf{make~} f_3 \mathbf{~atloc~} l_3 \mathbf{~inhibit~} a_3  \\
&\label{idesc:contsupply}\mathbf{continuously~supply~} f_4 \mathbf{~in~quantity~} q_4   \\
&\label{idesc:contsupply:atloc}\mathbf{contiunously~supply~} f_4 \mathbf{~atloc~} l_4 \mathbf{~in~quantity~} q_4   \\
&\label{idesc:xfer}\mathbf{continuously~transfer~} f_1 \mathbf{~in~quantity~} q_1 \mathbf{~across~}  l_1,l_2 \mathbf{~to~lower~gradient~}  \\
&\label{idesc:delay}\mathbf{add~delay~of~} q_1 \mathbf{~time~units~in~availability~of~} f_1 \\
&\label{idesc:delay:atloc}\mathbf{add~delay~of~} q_1 \mathbf{~time~units~in~availability~of~} f_1 \mathbf{~atloc~} l_1   \\
&\label{idesc:fixedsupply}\mathbf{set~value~of~} f_4 \mathbf{~to~} q_4 \\
&\label{idesc:fixedsupply:atloc}\mathbf{set~value~of~} f_4 \mathbf{~atloc~} l_4 \mathbf{~to~} q_4 
\end{align}
}
\end{definition}
Intuitively, intervention~\eqref{idesc:remove:immed} modifies the pathway such that all quantity of $f_1$ is removed as soon as it is produced; intervention~\eqref{idesc:remove:immed:atloc} modifies the pathway such that all quantity of $f_1[l_1]$ is removed as soon as it is produced; intervention~\eqref{idesc:disable} disables the action $a_2$; intervention~\eqref{idesc:cont:xform:x2y} modifies the pathway such that $f_1$ gets converted to $f_2$ at the rate of $q_1$ units per time-unit; intervention~\eqref{idesc:cont:xform:x2y:loc} modifies the pathway such that $f_1[l_1]$ gets converted to $f_2[l_2]$ at the rate of $q_1$ units per time-unit; intervention~\eqref{idesc:inhibits} modifies the pathway such that action $a_3$ is now inhibited each time there is $1$ or more units of $f_3$ and sets value of $f_3$ to $1$ to initially inhibit $a_3$; intervention~\eqref{idesc:inhibits:atloc} modifies the pathway such that action $a_3$ is now inhibited each time there is $1$ or more units of $f_3[l_3]$ and sets value of $f_3[l_3]$ to $1$ to initially inhibit $a_3$; intervention~\eqref{idesc:contsupply} modifies the pathway to continuously supply $f_4$ at the rate of $q_4$ units per time-unit; intervention~\eqref{idesc:contsupply:atloc} modifies the pathway to continuously supply $f_4[l_4]$ at the rate of $q_4$ units per time-unit; intervention~\eqref{idesc:xfer} modifies the pathway to transfer $f_1[l_1]$ to $f_1[l_2]$ in quantity $q_1$ or back depending upon whether $f_1[l_1]$ is higher than $f_1[l_2]$ or lower; intervention~\eqref{idesc:delay} modifies the pathway to add delay of $q_1$ time units between when $f_1$ is produced to when it is made available to next action; intervention~\eqref{idesc:delay:atloc} modifies the pathway to add delay of $q_1$ time units between when $f_1[l_1]$ is produced to when it is made available to next action; intervention~\eqref{idesc:fixedsupply} modifies the pathway to set the initial value of $f_4$ to $q_4$; and intervention~\eqref{idesc:fixedsupply:atloc} modifies the pathway to set the initial value of $f_4[l_4]$ to $q_4$.

\begin{definition}[Initial Condition]\label{dqa:syn:icond}
An initial condition is one of the intervention~\eqref{idesc:contsupply}, \eqref{idesc:contsupply:atloc}, \eqref{idesc:fixedsupply}, \eqref{idesc:fixedsupply:atloc} as given in definition~\ref{dqa:syn:interv}.
\end{definition}
Intuitively, initial conditions are interventions that setup fixed or continuous supply of substances participating in a pathway.

\begin{definition}[Query Statement]\label{def:dqa:syn:qstmt}
A query statement can be of the following forms:
{\footnotesize
\begin{align}
\label{dqa:syn:qstmt}& \langle \text{query description} \rangle ; \nonumber\\
&\;\;\;\;\;\;\mathbf{due~to~interventions:} \; \langle intervention \rangle_1, \dots, \langle intervention \rangle_{N1}; \nonumber\\
&\;\;\;\;\;\;\mathbf{due~to~observations:} \; \langle \text{internal observation} \rangle_1, \dots, \langle \text{internal observation} \rangle_{N2}; \nonumber\\
&\;\;\;\;\;\;\mathbf{using~initial~setup:} \; \langle \text{initial condition} \rangle_1, \dots, \langle \text{initial condition} \rangle_{N3};\\
\nonumber\\
\label{dqa:syn:cmpr:qstmt}& \langle \text{comparative query description} \rangle ;  \nonumber\\
&\;\;\;\;\;\;\;\mathbf{comparing~nominal~pathway~with~modified~pathway~obtained~} \nonumber\\
&\;\;\;\;\;\;\;\;\;\;\;\;\;\mathbf{due~to~interventions:} \; \langle intervention \rangle_1, \dots, \langle intervention \rangle_{N1} ; \nonumber\\
&\;\;\;\;\;\;\;\;\;\;\;\;\;\mathbf{due~to~observations:} \; \langle \text{internal observation} \rangle_1, \dots, \langle \text{internal observation} \rangle_{N2}; \nonumber\\
&\;\;\;\;\;\;\mathbf{using~initial~setup:} \; \langle \text{initial condition} \rangle_1, \dots, \langle \text{initial condition} \rangle_{N3}; &
\end{align}
}
where interventions, observations, and initial setup are optional.
\end{definition}
Intuitively, a {\em query statement} asks whether a {\em query description} holds in a pathway, perhaps after modifying it with initial setup, interventions and observations. Intuitively, a {\em comparative query statement} asks whether a {\em comparative query description} holds with a nominal pathway is compared against a modified pathway, where both pathways have the same initial setup, but only the modified pathway has been modified with interventions and observations.

\section{Semantics of the Query Language (BioPathQA-QL)}\label{sec:dqa:sem}

In this section we give the semantics of our pathway specification language and the query language. The semantics of the query language is in terms of the trajectories of a domain description $\mathbf{D}$ that satisfy a query $\mathbf{Q}$. We will present the semantics using LTL-style formulas. First, we informally define the semantics of the query language as follows.

Let $\mathbf{Q}$ be a {\em query statement} of the form~\eqref{dqa:syn:qstmt} with a query description $U$, interventions $V_1,\dots,V_{|V|}$, internal observations $O_1,\dots,O_{|O|}$, and initial setup conditions $I_1,\dots,I_{|I|}$. We construct a modified domain description $\mathbf{D_1}$ by applying $I_1,\dots,I_{|I|}$ and$V_1,\dots,V_{|V|}$ to $\mathbf{D}$. We filter the trajectories of $\mathbf{D_1}$ to retain only those trajectories that satisfy the observations $O_1,\dots,O_{|O|}$. Then we determine if $U$ holds on any of the retained trajectories. If it does, then we say that $\mathbf{D}$ satisfies $\mathbf{Q}$.

Let $\mathbf{Q}$ be a {\em comparative query statement} of the form~\eqref{dqa:syn:cmpr:qstmt} with {\em quantitative comparative aggregate query description} $U$, interventions $V_1,\dots,V_{|V|}$, internal observations $O_1,\dots,O_{|O|}$, and initial conditions $I_1,\dots,I_{|I|}$. Then we evaluate $\mathbf{Q}$ by deriving two sub-query statements. $\mathbf{Q_0}$ is constructed by removing the interventions $V_1,\dots,V_{|V|}$ and observations $O_1,\dots,O_{|O|}$ from $\mathbf{Q}$ and replacing the {\em quantitative comparative aggregate query description} $U$ with the corresponding {\em quantitative aggregate query description} $U'$,  $\mathbf{Q_1}$ is constructed by replacing the quantitative comparative aggregate query description $U$ with the corresponding quantitative aggregate query description $U'$. Then $\mathbf{D}$ satisfies $\mathbf{Q}$ iff we can find $d \in \{ <,>,= \}$ s.t. $n \; d \; n'$, where $\mathbf{D}$ satisfies $\mathbf{Q_0}$ for some value $n$ and $\mathbf{D}$ satisfies $\mathbf{Q_1}$ for some value $n'$.

\subsection{An Illustrative Example}\label{sec:dqa:illustrative:example}
In this section, we illustrate with an example how we intuitively evaluate a comparative query statement. In the later sections, we will give the formal semantics of query satisfaction.

Consider the following simple pathway specification:
\begin{equation}\label{eqn:pathway:simple}
\begin{array}{llll}
&\mathbf{domain~of~} & f_1 \mathbf{~is~} integer, & f_2 \mathbf{~is~} integer \\
&t_1 \mathbf{~may~fire~causing~} & f_1 \mathbf{~change~value~by~} -1, & f_2 \mathbf{~change~value~by~} +1\\
& \mathbf{initially~} & f_1 \mathbf{~has~value~} 0, & f_2 \mathbf{~has~value~} 0,\\
&\mathbf{firing~style~} & max
\end{array}
\end{equation}

Let the following specify a query statement $\mathbf{Q}$:
{\footnotesize
\begin{align*}
\mathbf{dir}&\mathbf{ection~of~change~in~} average \mathbf{~rate~of~production~of~} f_2 \mathbf{~is~} d\\
&\mathbf{when~observed~between~time~step~} 0 \mathbf{~and~time~step~} k;\\
&\mathbf{comparing~nominal~pathway~with~modified~pathway~obtained~}\\
&\;\;\;\;\;\;\mathbf{due~to~interventions:~} \mathbf{remove~} f_2 \mathbf{~as~soon~as~produced} ;\\
&\mathbf{using~initial~setup:~} \mathbf{continuously~supply~} f_1 \mathbf{~in~quantity~} 1;
\end{align*}
}
that we want to evaluate against $\mathbf{D}$ using a simulation length $k$ with maximum $ntok$ tokens at any place to determine `$d$' that satisfies it.

We construct the baseline query $\mathbf{Q_0}$ by removing interventions and observations, and replacing the comparative aggregate quantitative query description with the corresponding aggregate quantitative query description as follows:
{\footnotesize
\begin{align*}
average &\mathbf{~rate~of~production~of~} f_2 \mathbf{~is~} n\\
&\mathbf{when~observed~between~time~step~} 0 \mathbf{~and~time~step~} k ;\\
&\mathbf{using~initial~setup:~} \mathbf{continuously~supply~} f_1 \mathbf{~in~quantity~} 1;\\
\end{align*}
}

We construct the alternate query $\mathbf{Q_1}$ by replacing the comparative aggregate quantitative query description with the corresponding aggregate quantitative query description as follows:
{\footnotesize
\begin{align*}
average&\mathbf{~rate~of~production~of~} f_2 \mathbf{~is~} n'\\
&\mathbf{when~observed~between~time~step~} 0 \mathbf{~and~time~step~} k ;\\
&\mathbf{due~to~interventions:~} \mathbf{remove~} f_2 \mathbf{~as~soon~as~produced} ;\\
&\mathbf{using~initial~setup:~} \mathbf{continuously~supply~} f_1 \mathbf{~in~quantity~} 1;\\
\end{align*}
}

We build a modified domain description $\mathbf{D_0}$ as $\mathbf{D} \diamond (\mathbf{continuously~supply~} f_1 $ $\mathbf{~in~quantity~} 1)$ based on initial conditions in $\mathbf{Q_0}$. We get: 
{\footnotesize
\begin{equation*}
\begin{array}{llll}
&\mathbf{domain~of~} & f_1 \mathbf{~is~} integer, & f_2 \mathbf{~is~} integer \\
&t_1 \mathbf{~may~fire~causing~} & f_1 \mathbf{~change~value~by~} -1, & f_2 \mathbf{~change~value~by~} +1\\
&t_{f_1} \mathbf{~may~fire~causing~} & f_1 \mathbf{~change~value~by~} +1\\
& \mathbf{initially~} & f_1 \mathbf{~has~value~} 0, & f_2 \mathbf{~has~value~} 0,\\
&\mathbf{firing~style~} & max
\end{array}
\end{equation*}
}

We evaluate $\mathbf{Q_0}$ against $\mathbf{D_0}$. It results in $m_0$ trajectories with rate of productions $n_j=(s_k(f_2)-s_0(f_2))/k$ on trajectory $\tau_j=s_0,\dots,s_k, 1 \leq j \leq m_0$ using time interval $[0,k]$. The rate of productions are averaged to produce $n=(n_1+\dots+n_{m_0})/m_0$.

Next, we construct the alternate domain description $\mathbf{D_1}$ as $\mathbf{D_0} \diamond (\mathbf{remove~} f_2 \mathbf{~as~soon~} $ $\mathbf{as~produced~})$ based on initial conditions and interventions in $\mathbf{Q_1}$. We get:
{\footnotesize
\begin{equation*}
\begin{array}{llll}
&\mathbf{domain~of~} & f_1 \mathbf{~is~} integer, & f_2 \mathbf{~is~} integer \\
&t_1 \mathbf{~may~fire~causing~} & f_1 \mathbf{~change~value~by~} -1, & f_2 \mathbf{~change~value~by~} +1\\
&t_{f_1} \mathbf{~may~fire~causing~} & f_1 \mathbf{~change~value~by~} +1\\
&t_{f_2} \mathbf{~may~fire~causing~} & f_2 \mathbf{~change~value~by~} *\\
& \mathbf{initially~} & f_1 \mathbf{~has~value~} 0, & f_2 \mathbf{~has~value~} 0\\
&\mathbf{firing~style~} & max
\end{array}
\end{equation*}
}

We evaluate $\mathbf{Q_1}$ against $\mathbf{D_1}$. Since there are no observations, no filtering is required. This results in $m_1$ trajectories, each with rate of production $n'_j=(s_k(f_2)-s_0(f_2))/k$ on trajectory $\tau'_j = s_0,\dots,s_k, 1 \leq j \leq m_1$ using time interval $[0,k]$. The rate of productions are averaged to produce $n'=(n'_1+\dots+n'_{m_1})/m_1$.

Due to the simple nature of our domain description, it has only one trajectory for each of the two domains. As a result, for any $k > 1$, $n' < n$. Thus, $\mathbf{D}$ satisfies $\mathbf{Q}$ iff $d = ``<''$.

We will now define the semantics of how a domain description $\mathbf{D}$ is modified according to the interventions and initial conditions, the semantics of conditions imposed by the internal observations. We will then formally define how $\mathbf{Q}$ is entailed in $\mathbf{D}$.

\subsection{Domain Transformation due to Interventions and Initial Conditions}\label{sec:semantics:idesc}\label{sec:semantics:icond}
An intervention $I$ modifies a given domain description $\mathbf{D}$, potentially resulting in a different set of trajectories than $\mathbf{D}$. We define a binary operator $\diamond$ that transforms $\mathbf{D}$ by applying an intervention $I$ as a set of edits to $\mathbf{D}$ using the pathway specification language. The trajectories of the modified domain description $\mathbf{D'} = \mathbf{D} \diamond I$ are given by the semantics of the pathway specification language. 
Below, we give the intuitive impact and edits required by each of the interventions. 

Domain modification by intervention \eqref{idesc:remove:immed} $\mathbf{D'} = \mathbf{D} \diamond (\mathbf{remove~} f_1 \mathbf{~as~soon~}$ $\mathbf{as~produced})$ modifies the pathway by removing all existing quantity of $f_1$ at each time step:
{\footnotesize
\begin{align*}
\mathbf{D'} &= \mathbf{D} 
+ \left\{
\begin{array}{l}
tr \mathbf{~may~execute~causing~} f_1 \mathbf{~change~value~by~} *
\end{array}
\right\} 
\end{align*}
}

Domain modification by intervention \eqref{idesc:remove:immed:atloc} $\mathbf{D'} = \mathbf{D} \diamond (\mathbf{remove~} f_1 \mathbf{~atloc~} l_1)$ modifies the pathway by removing all existing quantity of $f_1$ at each time step:
{\footnotesize
\begin{align*}
\mathbf{D'} &= \mathbf{D} 
+ \left\{
\begin{array}{l}
tr \mathbf{~may~execute~causing~} f_1 \mathbf{~atloc~} l_1 \mathbf{~change~value~by~} *
\end{array}
\right\} 
\end{align*}
}

Domain modification by intervention \eqref{idesc:disable} $\mathbf{D'} = \mathbf{D} \diamond (\mathbf{disable~} a_2)$ modifies the pathway such that its trajectories have $a_2 \notin T_i$, where $i \geq 0$.
{\footnotesize
\begin{align*}
\mathbf{D'} &= \mathbf{D} 
+ \left\{
\begin{array}{ll}
\mathbf{inhibit~} a_2
\end{array}
\right\} 
\end{align*}
}

Domain modification by intervention \eqref{idesc:cont:xform:x2y} $\mathbf{D'} = \mathbf{D} \diamond (\mathbf{continuously~transform~} f_1 $ $\mathbf{~in~quantity~} q_1 $ $\mathbf{~to~} f_2)$ where $s_{i+1}(f_1)$ decreases, and $s_{i+1}(f_2)$ increases by $q_1$ at each time step $i \geq 0$, when $s_i(f_1) \geq q_1$.
{\footnotesize
\begin{align*}
\mathbf{D'} = \mathbf{D} 
&+ \left\{
\begin{array}{ll}
a_{f_{1,2}} \mathbf{~may~execute~causing~} & f_1 \mathbf{~change~value~by~} -q_1, \\ & f_2 \mathbf{~change~value~by~} +q_1
\end{array}
\right\} 
\end{align*}
}

Domain modification by intervention \eqref{idesc:cont:xform:x2y:loc} $\mathbf{D'} = \mathbf{D} \diamond (\mathbf{continuously~transform~} $ $f_1 \mathbf{~atloc~} l_1$ $ \mathbf{~in~quantity~} q_1 $ $\mathbf{~to~} f_2 \mathbf{~atloc~} l_2)$ where $s_{i+1}(f_1[l_1])$ decreases, and $s_{i+1}(f_2[l_2])$ increases by $q_1$ at each time step $i \geq 0$, when $s_i(f_1[l_1]) \geq q_1$.
{\footnotesize
\begin{align*}
\mathbf{D'} = \mathbf{D} 
&+ \left\{
\begin{array}{ll}
a_{f_{1,2}} \mathbf{~may~execute~causing~} & f_1 \mathbf{~atloc~} l_1  \mathbf{~change~value~by~} -q_1, \\ & f_2  \mathbf{~atloc~} l_2 \mathbf{~change~value~by~} +q_1
\end{array}
\right\} 
\end{align*}
}

Domain modification by intervention \eqref{idesc:inhibits} $\mathbf{D'} = \mathbf{D} \diamond (\mathbf{make~} f_3 \mathbf{~inhibit~} a_3 )$ modifies the pathway such that it has $a_3$ inhibited due to $f_3$.
{\footnotesize
\begin{align*}
\mathbf{D'} = \mathbf{D} 
&- \left\{
\begin{array}{l}
\mathbf{initially~} f_3 \mathbf{~has~value~} q \in \mathbf{D}
\end{array}
\right\}
\\&+ \left\{
\begin{array}{ll}
\mathbf{inhibit~} a_3 \mathbf{~if~} & f_3 \mathbf{~has~value~} 1 \mathbf{~or~higher}\\
\mathbf{initially~} & f_3 \mathbf{~has~value~} 1
\end{array}
\right\} 
\end{align*}
}

Domain modification by intervention \eqref{idesc:inhibits:atloc} $\mathbf{D'} = \mathbf{D} \diamond (\mathbf{make~} f_3 \mathbf{~atloc~} l_3 \mathbf{~inhibit~} a_3)$ modifies the pathway such that it has $a_3$ inhibited due to $f_3[l_3]$.
{\footnotesize
\begin{align*}
\mathbf{D'} = \mathbf{D} 
&- \left\{
\begin{array}{l}
\mathbf{initially~} f_3 \mathbf{~atloc~} l_3 \mathbf{~has~value~} q \in \mathbf{D}
\end{array}
\right\}
\\&+ \left\{
\begin{array}{ll}
\mathbf{inhibit~} a_3 \mathbf{~if~} & f_3 \mathbf{~atloc~} l_3 \mathbf{~has~value~} 1 \mathbf{~or~higher}, \\
\mathbf{initially~} & f_3 \mathbf{~atloc~} l_3 \mathbf{~has~value~} 1
\end{array}
\right\} 
\end{align*}
}

Domain modification by intervention \eqref{idesc:contsupply} $\mathbf{D'} = \mathbf{D} \diamond (\mathbf{~continuously~supply~} f_4 $ $\mathbf{~in~quantity~} q_4)$ modifies the pathway such that a quantity $q_4$ of substance $f_4$ is supplied at each time step.
{\footnotesize
\begin{align*}
\mathbf{D'} = \mathbf{D} 
&+ \left\{
\begin{array}{ll}
t_{f_4} \mathbf{~may~execute~causing~} & f_4 \mathbf{~change~value~by~} +q_4
\end{array}
\right\} 
\end{align*}
}

Domain modification by intervention \eqref{idesc:contsupply:atloc} $\mathbf{D'} = \mathbf{D} \diamond (\mathbf{~continuously~supply~} f_4 $ $\mathbf{~atloc~} l_4 $ $\mathbf{~in~quantity~} q_4 )$ modifies the pathway such that a quantity $q_4$ of substance $f_4$ at location $l_4$ is supplied at each time step.
{\footnotesize
\begin{align*}
\mathbf{D'} = \mathbf{D} 
&+ \left\{
\begin{array}{ll}
t_{f_4} \mathbf{~may~execute~causing~} & f_4 \mathbf{~atloc~} l_4 \mathbf{~change~value~by~} +q_4
\end{array}
\right\} 
\end{align*}
}

Domain modification by intervention \eqref{idesc:xfer} $\mathbf{D'} = \mathbf{D} \diamond (\mathbf{continuously~transfer~} f_1 $ $\mathbf{~in~quantity~} q_1 $ $\mathbf{~across~} l_1, l_2 $ $\mathbf{~to~lower~gradient})$ modifies the pathway such that substance represented by $f_1$ is transferred from location $l_1$ to $l_2$ or $l_2$ to $l_1$ depending upon whether it is at a higher quantity at $l_1$ or $l_2$.
{\footnotesize
\begin{align*}
\mathbf{D'} = \mathbf{D} 
&+ \left\{
\begin{array}{ll}
t_{f_1} \mathbf{~may~execute~causing~} & f_1 \mathbf{~atloc~} l_1 \mathbf{~change~value~by~} -q_1, \\ & f_1 \mathbf{~atloc~} l_2 \mathbf{~change~value~by~} +q_1 \\
\mathbf{~~~~~~~if}& f_1 \mathbf{~atloc~} l_1 \mathbf{~has~higher~value~than~} f_1 \mathbf{~atloc~} l_2, \\
t'_{f_1} \mathbf{~may~execute~causing~} & f_1 \mathbf{~atloc~} l_2 \mathbf{~change~value~by~} -q_1, \\ & f_1 \mathbf{~atloc~} l_1 \mathbf{~change~value~by~} +q_1\\
 \mathbf{~~~~~~~if}& f_1 \mathbf{~atloc~} l_2 \mathbf{~has~higher~value~than~} f_1 \mathbf{~atloc~} l_1, \\
\end{array}
\right\} 
\end{align*}
}

Domain modification by intervention \eqref{idesc:delay} $\mathbf{D'} = \mathbf{D} \diamond (\mathbf{add~delay~of~} q_1 $ $\mathbf{~time~units~} $ $\mathbf{in~availability~of~}  f_1)$ modifies the pathway such that $f_1$'s arrival is delayed by $q_1$ time units. We create additional cases for all actions that produce $f_1$, such that it goes through an additional delay action.
{\footnotesize
\begin{align*}
\mathbf{D'} = \mathbf{D}
& - \left\{
\begin{array}{ll}
a \mathbf{~may~execute~causing~} & f_1 \mathbf{~change~value~by~} +w_1, \\ & \mathit{effect}_1,\dots,\mathit{effect}_n \\ & \mathbf{~if~} cond_1,\dots,cond_m \in \mathbf{D}\\
\end{array}
\right\} \\
& + \left\{
\begin{array}{ll}
a \mathbf{~may~execute~causing~} & f'_1 \mathbf{~change~value~by~} +w_1, \\ & \mathit{effect}_1,\dots,\mathit{effect}_n, \\ & \mathbf{~if~}  cond_1,\dots,cond_m\\
a_{f_1} \mathbf{~may~execute~causing~} & f'_1 \mathbf{~change~value~by~} -w_1, \\ & f_1 \mathbf{~change~value~by~} +w_1\\
a_{f_1} \mathbf{~executes~} & \mathbf{~in~} q_1 \mathbf{~time~units}\\
\end{array}
\right\}
\end{align*}
}

Domain modification by intervention \eqref{idesc:delay:atloc} $\mathbf{D'} = \mathbf{D} \diamond (\mathbf{add~delay~of} q_1 $ $\mathbf{~time~units~} $ $\mathbf{in~availability~of~}  f_1 \mathbf{~atloc~} l_1)$ modifies trajectories such that $f_1[l_1]$'s arrival is delayed by $q_1$ time units. We create additional cases for all actions that produce $f_1 \mathbf{~atloc~} l_1$, such that it goes through an additional delay action.
{\footnotesize
\begin{align*}
\mathbf{D'} = \mathbf{D} 
&- \left\{
\begin{array}{ll}
a \mathbf{~may~execute~causing~} & f_1 \mathbf{~atloc~} l_1 \mathbf{~change~value~by~} +w_1, \\ & \mathit{effect}_1,\dots,\mathit{effect}_n \\ & \mathbf{~if~} cond_1,\dots,cond_m \in \mathbf{D} 
\end{array}
\right\} \\
&+ \left\{
\begin{array}{ll}
a \mathbf{~may~execute~causing~} & f_1 \mathbf{~atloc~} l'_1 \mathbf{~change~value~by~} +w_1, \\ & \mathit{effect}_1,\dots,\mathit{effect}_n \\ & \mathbf{~if~} cond_1,\dots,cond_n\\
a_{f_1} \mathbf{~may~execute~causing~} & f_1 \mathbf{~atloc~} l'_1 \mathbf{~change~value~by~} -w_1,\\ & f_1 \mathbf{~atloc~} l_1 \mathbf{~change~value~by~} +w_1\\
a_{f_1} \mathbf{~executes~} & \mathbf{~in~} q_1 \mathbf{~time~units}\\
\end{array}
\right\}
\end{align*}
}

Domain modification by intervention \eqref{idesc:fixedsupply} $\mathbf{D'} = \mathbf{D} \diamond (\mathbf{~set~value~of~} f_4 $ $\mathbf{~to~} q_4)$ modifies the pathway such that its trajectories have $s_0(f_4) = q_4$.
{\footnotesize
\begin{align*}
\mathbf{D'} = \mathbf{D}
& - \left\{
\begin{array}{l}
\mathbf{initially~} f_4 \mathbf{~has~value~} n \in \mathbf{D}
\end{array}
\right\}\\ 
& + \left\{
\begin{array}{l}
\mathbf{initially~} f_4 \mathbf{~has~value~} q_4
\end{array}
\right\} 
\end{align*}
}

Domain modification by intervention \eqref{idesc:fixedsupply:atloc} $\mathbf{D'} = \mathbf{D} \diamond (\mathbf{~set~value~of~} f_4 $ $\mathbf{~atloc~} l_4 $ $\mathbf{~to~} q_4 )$ modifies the pathway such that its trajectories have $s_0({f_4}_{l_4}) = q_4$.
{\footnotesize
\begin{align*}
\mathbf{D'} = \mathbf{D} 
& - \left\{
\begin{array}{l}
\mathbf{initially~} f_4 \mathbf{~atloc~} l_4 \mathbf{~has~value~} n \in \mathbf{D}
\end{array}
\right\} \\
& + \left\{
\begin{array}{l}
\mathbf{initially~} f_4 \mathbf{~atloc~} l_4 \mathbf{~has~value~} q_4
\end{array}
\right\} 
\end{align*}
}

\subsection{Formula Semantics}\label{dqa:sem:formula}

We will now define the semantics of some common formulas that we will use in the following sections. First we introduce the LTL-style formulas that we will be using to define the syntax.

A formula $\langle s_i,\sigma \rangle \models F$ represents that $F$ holds at point $i$. 

A formula $\{ \langle s^1_i,\sigma_1 \rangle ,\dots, \langle s^m_i,\sigma_m \rangle \} \models F$ represents that $F$ holds at point $i$ on a set of trajectories  $\sigma_1,\dots,\sigma_m$.

A formula $\big\{ \{ \langle s^1_i,\sigma_1 \rangle ,\dots, \langle s^m_i,\sigma_m \rangle \}, \{ \langle \bar{s}^1_i,\bar{\sigma}_1 \rangle ,\dots, \langle \bar{s}^{\bar{m}}_i,\bar{\sigma}_{\bar{m}} \rangle \} \big\} \models F$ represents that $F$ holds at point $i$ on two sets of trajectories  $\sigma_1,\dots,\sigma_m$ and $\{ \bar{\sigma}_1,\dots,\bar{\sigma}_{\bar{m}} \}$.

A formula $(\langle s_i,\sigma \rangle,j) \models F$ represents that $F$ holds in the interval $[i,j]$ on trajectory $\sigma$.

A formula $(\{ \langle s^1_i,\sigma_1 \rangle ,\dots, \langle s^m_i,\sigma_m \rangle \}, j) \models F$ represents that $F$ holds in the interval $[i,j]$ on a set of trajectories $\sigma_1,\dots,\sigma_m$.

A formula $(\big\{ \{ \langle s^1_i,\sigma_1 \rangle ,\dots, \langle s^m_i,\sigma_m \rangle \}, \{ \langle \bar{s}^1_i,\bar{\sigma}_1 \rangle ,\dots, \langle \bar{s}^{\bar{m}}_i,\bar{\sigma}_{\bar{m}} \rangle \} \big\}, j) \models F$ represents that $F$ holds in the interval $[i,j]$ over two sets of trajectories $\{ \sigma_1,\dots, \sigma_m \}$ and $\{ \bar{\sigma}_1,\dots,\bar{\sigma}_{\bar{m}} \}$.

Given a domain description $\mathbf{D}$ with simple fluents represented by a Guarded-Arc Petri Net as defined in definition~\ref{def:gpn}.  Let $\sigma = s_0,T_0,s_1,\dots,T_{k-1},s_{k}$ be its trajectory as defined in~\eqref{def:gpn:traj}, and $s_i$ be a state on that trajectory. Let actions $T_i$ firing in state $s_i$ be observable in $s_i$ such that $T_i \subseteq s_i$. 

First we define how interval formulas are satisfied on a trajectory $\sigma$, starting state $s_i$ and an ending point $j$:
{\footnotesize
\begin{align}
(\langle s_i,\sigma \rangle, j) \models& \mathbf{~rate~of~production~of~} f \mathbf{~is~} n  \nonumber\\
&\text{ if } n=(s_j(f)-s_i(f))/(j-i)\\
(\langle s_i,\sigma \rangle, j) \models& \mathbf{~rate~of~firing~of~} a \mathbf{~is~} n  \nonumber\\
&\text{ if } n=\sum_{i\leq k \leq j, \langle s_k,\sigma \rangle \models a \mathbf{~occurs~}}{1}/(j-i)\\
(\langle s_i,\sigma \rangle, j) \models& \mathbf{~total~production~of~} f \mathbf{~is~} n  \nonumber\\
&\text{ if } n=(s_j(f)-s_i(f))\\
(\langle s_i,\sigma \rangle, j) \models& f \mathbf{~is~accumulating~}  \nonumber\\
&\text{ if } (\nexists k, i \leq k \leq j : s_{k+1}(f) < s_k(f)) \text{ and } s_j(f) > s_i(f)\\
(\langle s_i,\sigma \rangle, j) \models& f \mathbf{~is~decreasing~}  \nonumber\\
&\text{ if } (\nexists k, i \leq k \leq j : s_{k+1}(f) > s_k(f)) \text{ and } s_j(f) < s_i(f)
\end{align}
}

Next we define how a point formula is satisfied on a trajectory $\sigma$, in a state $s_i$:
{\footnotesize
\begin{align}
\langle s_i,\sigma \rangle \models& \mathbf{~value~of~} f \mathbf{~is~higher~than~} n \nonumber\\
&\text{ if } s_i(f) > n\\
\langle s_i,\sigma \rangle \models& \mathbf{~value~of~} f \mathbf{~is~lower~than~} n \nonumber\\
&\text{ if } s_i(f) < n\\
\langle s_i,\sigma \rangle \models& \mathbf{~value~of~} f \mathbf{~is~} n \nonumber\\
&\text{ if } s_i(f) = n\\
\langle s_i,\sigma \rangle \models& a \mathbf{~occurs~} \nonumber\\
&\text{ if } a \in s_i\\
\langle s_i,\sigma \rangle \models& a \mathbf{~does~not~occur~} \nonumber\\
&\text{ if } a \notin s_i\\
\langle s_i,\sigma \rangle \models& a_1 \mathbf{~switches~to~} a_2 \nonumber\\
&\text{ if } a_1 \in s_{i-1} \text{ and } a_2 \notin s_{i-1} \text{ and } a_1 \notin s_i \text{ and } a_2 \in s_i 
\end{align} 
}

Next we define how a quantitative all interval formula is satisfied on a set of trajectories $\sigma_1,\dots,\sigma_m$ with starting states $s^1_i,\dots,s^m_i$ and end point $j$:
{\footnotesize
\begin{align}
(\{ \langle s^1_i,\sigma_1 \rangle,\dots, \langle s^m_i,\sigma_m \rangle \}, j) \models& \mathbf{~rates~of~production~of~} f \mathbf{~are~} [r_1,\dots,r_m]  \nonumber\\
&\text{ if } (\langle s^1_i,\sigma_1 \rangle, j) \models \mathbf{~rate~of~production~of~} f \mathbf{~is~} r_1\nonumber\\
&\vdots \nonumber\\
&\text{ and } (\langle s^m_i,\sigma_m \rangle, j) \models \mathbf{~rate~of~production~of~} f \mathbf{~is~} r_m \\
(\{ \langle s^1_i,\sigma_1 \rangle,\dots, \langle s^m_i,\sigma_m \rangle \}, j) \models& \mathbf{~rates~of~firing~of~} a \mathbf{~are~} [r_1,\dots,r_m]  \nonumber\\
&\text{ if } (\langle s^1_i,\sigma_1 \rangle, j) \models \mathbf{~rate~of~firing~of~} a \mathbf{~is~} r_1\nonumber\\
&\vdots \nonumber\\
&\text{ and } (\langle s^m_i,\sigma_m \rangle, j) \models \mathbf{~rate~of~firing~of~} a \mathbf{~is~} r_m \\
(\{ \langle s^1_i,\sigma_1 \rangle,\dots, \langle s^m_i,\sigma_m \rangle \}, j) \models& \mathbf{~totals~of~production~of~} f \mathbf{~are~} [r_1,\dots,r_m]  \nonumber\\
&\text{ if } (\langle s^1_i,\sigma_1 \rangle, j) \models \mathbf{~total~production~of~} f \mathbf{~is~} r_1\nonumber\\
&\vdots \nonumber\\
&\text{ and } (\langle s^m_i,\sigma_m \rangle, j) \models \mathbf{~total~production~of~} f \mathbf{~is~} r_m 
\end{align}
}

Next we define how a quantitative all point formula is satisfied on a set of trajectories $\sigma_1,\dots,\sigma_m$ in states $s^1_i,\dots,s^m_i$:
{\footnotesize
\begin{align}
\{ \langle s^1_i,\sigma_1 \rangle,\dots, \langle s^m_i,\sigma_m \rangle \} \models& \mathbf{~values~of~} f \mathbf{~are~} [r_1,\dots,r_m]  \nonumber\\
&\text{ if } \langle s^1_i,\sigma_1 \rangle \models \mathbf{~value~of~} f \mathbf{~is~} r_1\nonumber\\
&\vdots \nonumber\\
&\text{ and } \langle s^m_i,\sigma_m \rangle \models \mathbf{~value~of~} f \mathbf{~is~} r_m
\end{align}
}

Next we define how a quantitative aggregate interval formula is satisfied on a set of trajectories $\sigma_1,\dots,\sigma_m$ with starting states $s^1_i,\dots,s^m_i$ and end point $j$:
{\footnotesize
\begin{align}
(\{ \langle s^1_i,\sigma_1 \rangle,\dots, \langle s^m_i,\sigma_m \rangle \}, j) \models& \; average \mathbf{~rate~of~production~of~} f \mathbf{~is~} r  \nonumber\\
&\text{ if } \exists [r_1,\dots,r_m] : \nonumber\\
&~~~~~~(\langle s^1_i,\sigma_1 \rangle, j) \models \mathbf{~rate~of~production~of~} f \mathbf{~is~} r_1\nonumber\\
&~~~~~~\vdots \nonumber\\
&~~~~~~(\langle s^m_i,\sigma_m \rangle, j) \models \mathbf{~rate~of~production~of~} f \mathbf{~is~} r_m  \nonumber\\
&\text{ and } r=(r_1+\dots+r_m)/m\\
(\{ \langle s^1_i,\sigma_1 \rangle,\dots, \langle s^m_i,\sigma_m \rangle \}, j) \models& \; minimum \mathbf{~rate~of~production~of~} f \mathbf{~is~} r  \nonumber\\
&\text{ if } \exists [r_1,\dots,r_m] : \nonumber\\
&~~~~~~(\langle s^1_i,\sigma_1 \rangle, j) \models \mathbf{~rate~of~production~of~} f \mathbf{~is~} r_1\nonumber\\
&~~~~~~\vdots \nonumber\\
&~~~~~~(\langle s^m_i,\sigma_m \rangle, j) \models \mathbf{~rate~of~production~of~} f \mathbf{~is~} r_m  \nonumber\\
&\text{ and } \exists k, 1 \leq k \leq m : r=r_k \text{ and } \forall x, 1 \leq x \leq m, r_k \leq r_x\\
(\{ \langle s^1_i,\sigma_1 \rangle,\dots, \langle s^m_i,\sigma_m \rangle \}, j) \models& \; maximum \mathbf{~rate~of~production~of~} f \mathbf{~is~} r  \nonumber\\
&\text{ if } \exists [r_1,\dots,r_m] : \nonumber\\
&~~~~~~(\langle s^1_i,\sigma_1 \rangle, j) \models \mathbf{~rate~of~production~of~} f \mathbf{~is~} r_1\nonumber\\
&~~~~~~\vdots \nonumber\\
&~~~~~~(\langle s^m_i,\sigma_m \rangle, j) \models \mathbf{~rate~of~production~of~} f \mathbf{~is~} r_m  \nonumber\\
&\text{ and } \exists k, 1 \leq k \leq m : r=r_k \text{ and } \forall x, 1 \leq x \leq m, r_k \geq r_x\\
\vspace{20pt}\nonumber\\
(\{ \langle s^1_i,\sigma_1 \rangle,\dots, \langle s^m_i,\sigma_m \rangle \}, j) \models& \; average \mathbf{~rate~of~firing~of~} f \mathbf{~is~} r  \nonumber\\
&\text{ iff } \exists [r_1,\dots,r_m] : \nonumber\\
&~~~~~~(\langle s^1_i,\sigma_1 \rangle, j) \models \mathbf{~rate~of~firing~of~} f \mathbf{~is~} r_1\nonumber\\
&~~~~~~\vdots \nonumber\\
&~~~~~~(\langle s^m_i,\sigma_m \rangle, j) \models \mathbf{~rate~of~firing~of~} f \mathbf{~is~} r_m  \nonumber\\
&\text{ and } r=(r_1+\dots+r_m)/m\\
(\{ \langle s^1_i,\sigma_1 \rangle,\dots, \langle s^m_i,\sigma_m \rangle \}, j) \models& \; minimum \mathbf{~rate~of~firing~of~} f \mathbf{~is~} r  \nonumber\\
&\text{ iff } \exists [r_1,\dots,r_m] : \nonumber\\
&~~~~~~(\langle s^1_i,\sigma_1 \rangle, j) \models \mathbf{~rate~of~firing~of~} f \mathbf{~is~} r_1\nonumber\\
&~~~~~~\vdots \nonumber\\
&~~~~~~(\langle s^m_i,\sigma_m \rangle, j) \models \mathbf{~rate~of~firing~of~} f \mathbf{~is~} r_m  \nonumber\\
&\text{ and } \exists k, 1 \leq k \leq m : r=r_k \text{ and } \forall x, 1 \leq x \leq m, r_k \leq r_x\\
(\{ \langle s^1_i,\sigma_1 \rangle,\dots, \langle s^m_i,\sigma_m \rangle \}, j) \models& \; maximum \mathbf{~rate~of~firing~of~} f \mathbf{~is~} r  \nonumber\\
&\text{ iff } \exists [r_1,\dots,r_m] : \nonumber\\
&~~~~~~(\langle s^1_i,\sigma_1 \rangle, j) \models \mathbf{~rate~of~firing~of~} f \mathbf{~is~} r_1\nonumber\\
&~~~~~~\vdots \nonumber\\
&~~~~~~(\langle s^m_i,\sigma_m \rangle, j) \models \mathbf{~rate~of~firing~of~} f \mathbf{~is~} r_m  \nonumber\\
&\text{ and } \exists k, 1 \leq k \leq m : r=r_k \text{ and } \forall x, 1 \leq x \leq m, r_k \geq r_x\\
\vspace{20pt}\nonumber\\
(\{ \langle s^1_i,\sigma_1 \rangle,\dots, \langle s^m_i,\sigma_m \rangle \}, j) \models& \; average \mathbf{~total~production~of~} f \mathbf{~is~} r  \nonumber\\
&\text{ if } \exists [r_1,\dots,r_m] : \nonumber\\
&~~~~~~(\langle s^1_i,\sigma_1 \rangle, j) \models \mathbf{~total~production~of~} f \mathbf{~is~} r_1\nonumber\\
&~~~~~~\vdots \nonumber\\
&~~~~~~(\langle s^m_i,\sigma_m \rangle, j) \models \mathbf{~total~production~of~} f \mathbf{~is~} r_m  \nonumber\\
&\text{ and } r=(r_1+\dots+r_m)/m\\
(\{ \langle s^1_i,\sigma_1 \rangle,\dots, \langle s^m_i,\sigma_m \rangle \}, j) \models& \; minimum \mathbf{~total~production~of~} f \mathbf{~is~} r  \nonumber\\
&\text{ if } \exists [r_1,\dots,r_m] : \nonumber\\
&~~~~~~(\langle s^1_i,\sigma_1 \rangle, j) \models \mathbf{~total~production~of~} f \mathbf{~is~} r_1\nonumber\\
&~~~~~~\vdots \nonumber\\
&~~~~~~(\langle s^m_i,\sigma_m \rangle, j) \models \mathbf{~total~production~of~} f \mathbf{~is~} r_m  \nonumber\\
&\text{ and } \exists k, 1 \leq k \leq m : r=r_k \text{ and } \forall x, 1 \leq x \leq m, r_k \leq r_x\\
(\{ \langle s^1_i,\sigma_1 \rangle,\dots, \langle s^m_i,\sigma_m \rangle \}, j) \models& \; maximum \mathbf{~total~production~of~} f \mathbf{~is~} r  \nonumber\\
&\text{ if } \exists [r_1,\dots,r_m] : \nonumber\\
&~~~~~~(\langle s^1_i,\sigma_1 \rangle, j) \models \mathbf{~total~production~of~} f \mathbf{~is~} r_1\nonumber\\
&~~~~~~\vdots \nonumber\\
&~~~~~~(\langle s^m_i,\sigma_m \rangle, j) \models \mathbf{~total~production~of~} f \mathbf{~is~} r_m  \nonumber\\
&\text{ and } \exists k, 1 \leq k \leq m : r=r_k \text{ and } \forall x, 1 \leq x \leq m, r_k \geq r_x
\end{align}
}

Next we define how a quantitative aggregate point formula is satisfied on a set of trajectories $\sigma_1,\dots,\sigma_m$ in states $s^1_i,\dots,s^m_i$:
{\footnotesize
\begin{align}
\{ \langle s^1_i,\sigma_1 \rangle,\dots, \langle s^m_i,\sigma_m \rangle \}\models& \; average \mathbf{~value~of~} f \mathbf{~is~} r \nonumber\\
&\text{ if } \exists [r_1,\dots,r_m] : \nonumber\\
&~~~~~~\langle s^1_i,\sigma_1 \rangle \models \mathbf{~value~of~} f \mathbf{~is~} r_1\nonumber\\
&~~~~~~\vdots \nonumber\\
&~~~~~~\langle s^m_i,\sigma_m \rangle \models \mathbf{~value~of~} f \mathbf{~is~} r_m \nonumber\\
&\text{ and } r=(r_1+\dots+r_m)/m\\
\{ \langle s^1_i,\sigma_1 \rangle,\dots, \langle s^m_i,\sigma_m \rangle \}\models& \; minimum \mathbf{~value~of~} f \mathbf{~is~} r \nonumber\\
&\text{ if } \exists [r_1,\dots,r_m] : \nonumber\\
&~~~~~~\langle s^1_i,\sigma_1 \rangle \models \mathbf{~value~of~} f \mathbf{~is~} r_1\nonumber\\
&~~~~~~\vdots \nonumber\\
&~~~~~~\langle s^m_i,\sigma_m \rangle \models \mathbf{~value~of~} f \mathbf{~is~} r_m \nonumber\\
&\text{ and } \exists k, 1 \leq k \leq m : r=r_k \text{ and } \forall x, 1 \leq x \leq m, r_k \leq r_x\\
\{ \langle s^1_i,\sigma_1 \rangle,\dots, \langle s^m_i,\sigma_m \rangle \}\models& \; maximum \mathbf{~value~of~} f \mathbf{~is~} r \nonumber\\
&\text{ if } \exists [r_1,\dots,r_m] : \nonumber\\
&~~~~~~\langle s^1_i,\sigma_1 \rangle \models \mathbf{~value~of~} f \mathbf{~is~} r_1\nonumber\\
&~~~~~~\vdots \nonumber\\
&~~~~~~\langle s^m_i,\sigma_m \rangle \models \mathbf{~value~of~} f \mathbf{~is~} r_m \nonumber\\
&\text{ and } \exists k, 1 \leq k \leq m : r=r_k \text{ and } \forall x, 1 \leq x \leq m, r_k \geq r_x
\end{align}
}

Next we define how a comparative quantitative aggregate interval formula is satisfied on two sets of trajectories $\sigma_1,\dots,\sigma_m$, and $\bar{\sigma}_1,\dots,\bar{\sigma}_{\bar{m}}$ a starting point $i$ and an ending point $j$:
{\footnotesize
\begin{align}
&\Big(\Big\{ \{ \langle s^1_i,\sigma_1 \rangle,\dots, \langle s^m_i,\sigma_m \rangle \}, \{ \langle \bar{s}^1_i,\bar{\sigma}_1 \rangle,\dots, \langle \bar{s}^{\bar{m}}_i,\bar{\sigma}_{\bar{m}} \rangle \} \Big\}, j \Big) \nonumber\\
&\;\;\;\;\;\;\models \; \mathbf{~direction~of~change~in~} average \mathbf{~rate~of~production~of~} f \mathbf{~is~} d  \nonumber\\
&\;\;\;\;\;\;\text{ if } \exists \; n_1 : (\{ \langle s^1_i,\sigma_1 \rangle,\dots, \langle s^m_i,\sigma_m \rangle \}, j) \models \; average \mathbf{~rate~of~production~of~} f \text{~is~} n_1   \nonumber\\
&\;\;\;\;\;\;\text{ and } \exists \; n_2 : (\{ \langle \bar{s}^1_i,\bar{\sigma}_1 \rangle,\dots, \langle \bar{s}^{\bar{m}}_i,\bar{\sigma}_{\bar{m}} \rangle \}, j) \models \; average \mathbf{~rate~of~production~of~} f \text{~is~} n_2   \nonumber\\
&\;\;\;\;\;\;\text{ and } n_2 \; d \; n_1\\ 
&\Big(\Big\{ \{ \langle s^1_i,\sigma_1 \rangle,\dots, \langle s^m_i,\sigma_m \rangle \}, \{ \langle \bar{s}^1_i,\bar{\sigma}_1 \rangle,\dots, \langle \bar{s}^{\bar{m}}_i,\bar{\sigma}_{\bar{m}} \rangle \} \Big\}, j \Big) \nonumber\\
&\;\;\;\;\;\;\models \; \mathbf{~direction~of~change~in~} minimum \mathbf{~rate~of~production~of~} f \mathbf{~is~} d  \nonumber\\
&\;\;\;\;\;\;\text{ if } \exists \; n_1 : (\{ \langle s^1_i,\sigma_1 \rangle,\dots, \langle s^m_i,\sigma_m \rangle \}, j) \models \; minimum \mathbf{~rate~of~production~of~} f \text{~is~} n_1   \nonumber\\
&\;\;\;\;\;\;\text{ and } \exists \; n_2 : (\{ \langle \bar{s}^1_i,\bar{\sigma}_1 \rangle,\dots, \langle \bar{s}^{\bar{m}}_i,\bar{\sigma}_{\bar{m}} \rangle \}, j) \models \; minimum \mathbf{~rate~of~production~of~} f \text{~is~} n_2   \nonumber\\
&\;\;\;\;\;\;\text{ and } n_2 \; d \; n_1\\ 
&\Big(\Big\{ \{ \langle s^1_i,\sigma_1 \rangle,\dots, \langle s^m_i,\sigma_m \rangle \}, \{ \langle \bar{s}^1_i,\bar{\sigma}_1 \rangle,\dots, \langle \bar{s}^{\bar{m}}_i,\bar{\sigma}_{\bar{m}} \rangle \} \Big\}, j \Big) \nonumber\\
&\;\;\;\;\;\;\models \; \mathbf{~direction~of~change~in~} maximum \mathbf{~rate~of~production~of~} f \mathbf{~is~} d  \nonumber\\
&\;\;\;\;\;\;\text{ if } \exists \; n_1 : (\{ \langle s^1_i,\sigma_1 \rangle,\dots, \langle s^m_i,\sigma_m \rangle \}, j) \models \; maximum \mathbf{~rate~of~production~of~} f \text{~is~} n_1   \nonumber\\
&\;\;\;\;\;\;\text{ and } \exists \; n_2 : (\{ \langle \bar{s}^1_i,\bar{\sigma}_1 \rangle,\dots, \langle \bar{s}^{\bar{m}}_i,\bar{\sigma}_{\bar{m}} \rangle \}, j) \models \; maximum \mathbf{~rate~of~production~of~} f \text{~is~} n_2   \nonumber\\
&\;\;\;\;\;\;\text{ and } n_2 \; d \; n_1\\ 
\vspace{20pt}\nonumber\\
&\Big(\Big\{ \{ \langle s^1_i,\sigma_1 \rangle,\dots, \langle s^m_i,\sigma_m \rangle \}, \{ \langle \bar{s}^1_i,\bar{\sigma}_1 \rangle,\dots, \langle \bar{s}^{\bar{m}}_i,\bar{\sigma}_{\bar{m}} \rangle \} \Big\}, j \Big) \nonumber\\
&\;\;\;\;\;\;\models \; \mathbf{~direction~of~change~in~} average \mathbf{~rate~of~firing~of~} f \mathbf{~is~} d  \nonumber\\
&\;\;\;\;\;\;\text{ if } \exists \; n_1 : (\{ \langle s^1_i,\sigma_1 \rangle,\dots, \langle s^m_i,\sigma_m \rangle \}, j) \models \; average \mathbf{~rate~of~firing~of~} f \text{~is~} n_1   \nonumber\\
&\;\;\;\;\;\;\text{ and } \exists \; n_2 : (\{ \langle \bar{s}^1_i,\bar{\sigma}_1 \rangle,\dots, \langle \bar{s}^{\bar{m}}_i,\bar{\sigma}_{\bar{m}} \rangle \}, j) \models \; average \mathbf{~rate~of~firing~of~} f \text{~is~} n_2   \nonumber\\
&\;\;\;\;\;\;\text{ and } n_2 \; d \; n_1\\ 
&\Big(\Big\{ \{ \langle s^1_i,\sigma_1 \rangle,\dots, \langle s^m_i,\sigma_m \rangle \}, \{ \langle \bar{s}^1_i,\bar{\sigma}_1 \rangle,\dots, \langle \bar{s}^{\bar{m}}_i,\bar{\sigma}_{\bar{m}} \rangle \} \Big\}, j \Big) \nonumber\\
&\;\;\;\;\;\;\models \; \mathbf{~direction~of~change~in~} minimum \mathbf{~rate~of~firing~of~} f \mathbf{~is~} d  \nonumber\\
&\;\;\;\;\;\;\text{ if } \exists \; n_1 : (\{ \langle s^1_i,\sigma_1 \rangle,\dots, \langle s^m_i,\sigma_m \rangle \}, j) \models \; minimum \mathbf{~rate~of~firing~of~} f \text{~is~} n_1   \nonumber\\
&\;\;\;\;\;\;\text{ and } \exists \; n_2 : (\{ \langle \bar{s}^1_i,\bar{\sigma}_1 \rangle,\dots, \langle \bar{s}^{\bar{m}}_i,\bar{\sigma}_{\bar{m}} \rangle \}, j) \models \; minimum \mathbf{~rate~of~firing~of~} f \text{~is~} n_2   \nonumber\\
&\;\;\;\;\;\;\text{ and } n_2 \; d \; n_1\\ 
&\Big(\Big\{ \{ \langle s^1_i,\sigma_1 \rangle,\dots, \langle s^m_i,\sigma_m \rangle \}, \{ \langle \bar{s}^1_i,\bar{\sigma}_1 \rangle,\dots, \langle \bar{s}^{\bar{m}}_i,\bar{\sigma}_{\bar{m}} \rangle \} \Big\}, j \Big) \nonumber\\
&\;\;\;\;\;\;\models \; \mathbf{~direction~of~change~in~} maximum \mathbf{~rate~of~firing~of~} f \mathbf{~is~} d  \nonumber\\
&\;\;\;\;\;\;\text{ if } \exists \; n_1 : (\{ \langle s^1_i,\sigma_1 \rangle,\dots, \langle s^m_i,\sigma_m \rangle \}, j) \models \; maximum \mathbf{~rate~of~firing~of~} f \text{~is~} n_1   \nonumber\\
&\;\;\;\;\;\;\text{ and } \exists \; n_2 : (\{ \langle \bar{s}^1_i,\bar{\sigma}_1 \rangle,\dots, \langle \bar{s}^{\bar{m}}_i,\bar{\sigma}_{\bar{m}} \rangle \}, j) \models \; maximum \mathbf{~rate~of~firing~of~} f \text{~is~} n_2   \nonumber\\
&\;\;\;\;\;\;\text{ and } n_2 \; d \; n_1\\
\vspace{20pt}\nonumber\\
&\Big(\Big\{ \{ \langle s^1_i,\sigma_1 \rangle,\dots, \langle s^m_i,\sigma_m \rangle \}, \{ \langle \bar{s}^1_i,\bar{\sigma}_1 \rangle,\dots, \langle \bar{s}^{\bar{m}}_i,\bar{\sigma}_{\bar{m}} \rangle \} \Big\}, j \Big) \nonumber\\
&\;\;\;\;\;\;\models \; \mathbf{~direction~of~change~in~} average \mathbf{~total~production~of~} f \mathbf{~is~} d  \nonumber\\
&\;\;\;\;\;\;\text{ if } \exists \; n_1 : (\{ \langle s^1_i,\sigma_1 \rangle,\dots, \langle s^m_i,\sigma_m \rangle \}, j) \models \; average \mathbf{~total~production~of~} f \text{~is~} n_1   \nonumber\\
&\;\;\;\;\;\;\text{ and } \exists \; n_2 : (\{ \langle \bar{s}^1_i,\bar{\sigma}_1 \rangle,\dots, \langle \bar{s}^{\bar{m}}_i,\bar{\sigma}_{\bar{m}} \rangle \}, j) \models \; average \mathbf{~total~production~of~} f \text{~is~} n_2   \nonumber\\
&\;\;\;\;\;\;\text{ and } n_2 \; d \; n_1\\ 
&\Big(\Big\{ \{ \langle s^1_i,\sigma_1 \rangle,\dots, \langle s^m_i,\sigma_m \rangle \}, \{ \langle \bar{s}^1_i,\bar{\sigma}_1 \rangle,\dots, \langle \bar{s}^{\bar{m}}_i,\bar{\sigma}_{\bar{m}} \rangle \} \Big\}, j \Big) \nonumber\\
&\;\;\;\;\;\;\models \; \mathbf{~direction~of~change~in~} minimum \mathbf{~total~production~of~} f \mathbf{~is~} d  \nonumber\\
&\;\;\;\;\;\;\text{ if } \exists \; n_1 : (\{ \langle s^1_i,\sigma_1 \rangle,\dots, \langle s^m_i,\sigma_m \rangle \}, j) \models \; minimum \mathbf{~total~production~of~} f \text{~is~} n_1   \nonumber\\
&\;\;\;\;\;\;\text{ and } \exists \; n_2 : (\{ \langle \bar{s}^1_i,\bar{\sigma}_1 \rangle,\dots, \langle \bar{s}^{\bar{m}}_i,\bar{\sigma}_{\bar{m}} \rangle \}, j) \models \; minimum \mathbf{~total~production~of~} f \text{~is~} n_2   \nonumber\\
&\;\;\;\;\;\;\text{ and } n_2 \; d \; n_1\\ 
&\Big(\Big\{ \{ \langle s^1_i,\sigma_1 \rangle,\dots, \langle s^m_i,\sigma_m \rangle \}, \{ \langle \bar{s}^1_i,\bar{\sigma}_1 \rangle,\dots, \langle \bar{s}^{\bar{m}}_i,\bar{\sigma}_{\bar{m}} \rangle \} \Big\}, j \Big) \nonumber\\
&\;\;\;\;\;\;\models \; \mathbf{~direction~of~change~in~} maximum \mathbf{~total~production~of~} f \mathbf{~is~} d  \nonumber\\
&\;\;\;\;\;\;\text{ if } \exists \; n_1 : (\{ \langle s^1_i,\sigma_1 \rangle,\dots, \langle s^m_i,\sigma_m \rangle \}, j) \models \; maximum \mathbf{~total~production~of~} f \text{~is~} n_1   \nonumber\\
&\;\;\;\;\;\;\text{ and } \exists \; n_2 : (\{ \langle \bar{s}^1_i,\bar{\sigma}_1 \rangle,\dots, \langle \bar{s}^{\bar{m}}_i,\bar{\sigma}_{\bar{m}} \rangle \}, j) \models \; maximum \mathbf{~total~production~of~} f \text{~is~} n_2   \nonumber\\
&\;\;\;\;\;\;\text{ and } n_2 \; d \; n_1
\end{align}
}

Next we define how a comparative quantitative aggregate point formula is satisfied on two sets of trajectories $\sigma_1,\dots,\sigma_m$, and $\bar{\sigma}_1,\dots,\bar{\sigma}_{\bar{m}}$ at point $i$:
{\footnotesize
\begin{align}
&\Big\{ \{ \langle s^1_i,\sigma_1 \rangle,\dots, \langle s^m_i,\sigma_m \rangle \}, \{ \langle \bar{s}^1_i,\bar{\sigma}_1 \rangle,\dots, \langle \bar{s}^{\bar{m}}_i,\bar{\sigma}_{\bar{m}} \rangle \} \Big\} \nonumber\\
&\;\;\;\;\;\;\models \; \mathbf{~direction~of~change~in~} average \mathbf{~value~of~} f \mathbf{~is~} d  \nonumber\\
&\;\;\;\;\;\;\text{ if } \exists \; n_1 : \{ \langle s^1_i,\sigma_1 \rangle,\dots, \langle s^m_i,\sigma_m \rangle \} \models \; average \mathbf{~value~of~} f \text{~is~} n_1   \nonumber\\
&\;\;\;\;\;\;\text{ and } \exists \; n_2 : \{ \langle \bar{s}^1_i,\bar{\sigma}_1 \rangle,\dots, \langle \bar{s}^{\bar{m}}_i,\bar{\sigma}_{\bar{m}} \rangle \} \models \; average \mathbf{~value~of~} f \text{~is~} n_2   \nonumber\\
&\;\;\;\;\;\;\text{ and } n_2 \; d \; n_1\\ 
&\Big\{ \{ \langle s^1_i,\sigma_1 \rangle,\dots, \langle s^m_i,\sigma_m \rangle \}, \{ \langle \bar{s}^1_i,\bar{\sigma}_1 \rangle,\dots, \langle \bar{s}^{\bar{m}}_i,\bar{\sigma}_{\bar{m}} \rangle \} \Big\} \nonumber\\
&\;\;\;\;\;\;\models \; \mathbf{~direction~of~change~in~} minimum \mathbf{~value~of~} f \mathbf{~is~} d  \nonumber\\
&\;\;\;\;\;\;\text{ if } \exists \; n_1 : \{ \langle s^1_i,\sigma_1 \rangle,\dots, \langle s^m_i,\sigma_m \rangle \} \models \; minimum \mathbf{~value~of~} f \text{~is~} n_1   \nonumber\\
&\;\;\;\;\;\;\text{ and } \exists \; n_2 : \{ \langle \bar{s}^1_i,\bar{\sigma}_1 \rangle,\dots, \langle \bar{s}^{\bar{m}}_i,\bar{\sigma}_{\bar{m}} \rangle \} \models \; minimum \mathbf{~value~of~} f \text{~is~} n_2   \nonumber\\
&\;\;\;\;\;\;\text{ and } n_2 \; d \; n_1\\ 
&\Big\{ \{ \langle s^1_i,\sigma_1 \rangle,\dots, \langle s^m_i,\sigma_m \rangle \}, \{ \langle \bar{s}^1_i,\bar{\sigma}_1 \rangle,\dots, \langle \bar{s}^{\bar{m}}_i,\bar{\sigma}_{\bar{m}} \rangle \} \Big\} \nonumber\\
&\;\;\;\;\;\;\models \; \mathbf{~direction~of~change~in~} maximum \mathbf{~value~of~} f \mathbf{~is~} d  \nonumber\\
&\;\;\;\;\;\;\text{ if } \exists \; n_1 : \{ \langle s^1_i,\sigma_1 \rangle,\dots, \langle s^m_i,\sigma_m \rangle \} \models \; maximum \mathbf{~value~of~} f \text{~is~} n_1   \nonumber\\
&\;\;\;\;\;\;\text{ and } \exists \; n_2 : \{ \langle \bar{s}^1_i,\bar{\sigma}_1 \rangle,\dots, \langle \bar{s}^{\bar{m}}_i,\bar{\sigma}_{\bar{m}} \rangle \} \models \; maximum \mathbf{~value~of~} f \text{~is~} n_2   \nonumber\\
&\;\;\;\;\;\;\text{ and } n_2 \; d \; n_1
\end{align}
}

\vspace{30pt}
Given a domain description $\mathbf{D}$ with simple fluents represented by a Guarded-Arc Petri Net with Colored tokens as defined in definition~\ref{def:gcpn}.  Let $\sigma = s_0,T_0,s_1,\dots,T_{k-1},s_k$ be its trajectory as defined in definition~\ref{def:gcpn:traj}, and $s_i$ be a state on that trajectory. Let actions $T_i$ firing in state $s_i$ be observable in $s_i$ such that $T_i \subseteq s_i$. We define observation semantics using LTL below. We will use $s_i(f[l])$ to represent $m_{s_i(l)}(f)$ (multiplicity / value of $f$ in location $l$) in state $s_i$.

First we define how interval formulas are satisfied on a trajectory $\sigma$, starting state $s_i$ and an ending point $j$:
{\footnotesize
\begin{align}
(\langle s_i,\sigma \rangle, j) \models& \mathbf{~rate~of~production~of~} f \mathbf{~atloc~} l \mathbf{~is~} n  \nonumber\\
&\text{ if } n=(s_j(f[l])-s_i(f[l]))/(j-i)\\
(\langle s_i,\sigma \rangle, j) \models& \mathbf{~rate~of~firing~of~} a \mathbf{~is~} n  \nonumber\\
&\text{ if } n=\sum_{i\leq k \leq j, \langle s_k,\sigma \rangle \models a \mathbf{~occurs~}}{1}/(j-i)\\
(\langle s_i,\sigma \rangle, j) \models& \mathbf{~total~production~of~} f \mathbf{~atloc~} l \mathbf{~is~} n  \nonumber\\
&\text{ if } n=(s_j(f[l])-s_i(f[l]))\\
(\langle s_i,\sigma \rangle, j) \models& f \mathbf{~is~accumulating~} \mathbf{~atloc~} l  \nonumber\\
&\text{ if } (\nexists k, i \leq k \leq j : s_{k+1}(f[l]) < s_k(f[l])) \text{ and } s_j(f[l]) > s_i(f[l])\\
(\langle s_i,\sigma \rangle, j) \models& f \mathbf{~is~decreasing~} \mathbf{~atloc~} l  \nonumber\\
&\text{ if } (\nexists k, i \leq k \leq j : s_{k+1}(f[l]) > s_k(f[l])) \text{ and } s_j(f[l]) < s_i(f[l])
\end{align}
}

Next we define how a point formula is satisfied on a trajectory $\sigma$, in a state $s_i$:
{\footnotesize
\begin{align}
\langle s_i,\sigma \rangle \models& \mathbf{~value~of~} f \mathbf{~atloc~} l \mathbf{~is~higher~than~} n \nonumber\\
&\text{ if } s_i(f[l]) > n\\
\langle s_i,\sigma \rangle \models& \mathbf{~value~of~} f \mathbf{~atloc~} l \mathbf{~is~lower~than~} n \nonumber\\
&\text{ if } s_i(f[l]) < n\\
\langle s_i,\sigma \rangle \models& \mathbf{~value~of~} f \mathbf{~atloc~} l \mathbf{~is~} n \nonumber\\
&\text{ if } s_i(f[l]) = n\\
\langle s_i,\sigma \rangle \models& a \mathbf{~occurs~} \nonumber\\
&\text{ if } a \in s_i\\
\langle s_i,\sigma \rangle \models& a \mathbf{~does~not~occur~} \nonumber\\
&\text{ if } a \notin s_i\\
\langle s_i,\sigma \rangle \models& a_1 \mathbf{~switches~to~} a_2 \nonumber\\
&\text{ if } a_1 \in s_{i-1} \text{ and } a_2 \notin s_{i-1} \text{ and } a_1 \notin s_i \text{ and } a_2 \in s_i 
\end{align} 
}

Next we define how a quantitative all interval formula is satisfied on a set of trajectories $\sigma_1,\dots,\sigma_m$ with starting states $s^1_i,\dots,s^m_i$ and end point $j$:
{\footnotesize
\begin{align}
(\{ \langle s^1_i,\sigma_1 \rangle,\dots, \langle s^m_i,\sigma_m \rangle \}, j) \models& \mathbf{~rates~of~production~of~} f \mathbf{~atloc~} l \mathbf{~are~} [r_1,\dots,r_m]  \nonumber\\
&\text{ if } (\langle s^1_i,\sigma_1 \rangle, j) \models \mathbf{~rate~of~production~of~} f \mathbf{~atloc~} l \mathbf{~is~} r_1\nonumber\\
&\vdots \nonumber\\
&\text{ and } (\langle s^m_i,\sigma_m \rangle, j) \models \mathbf{~rate~of~production~of~} f \mathbf{~atloc~} l \mathbf{~is~} r_m \\
(\{ \langle s^1_i,\sigma_1 \rangle,\dots, \langle s^m_i,\sigma_m \rangle \}, j) \models& \mathbf{~rates~of~firing~of~} a \mathbf{~are~} [r_1,\dots,r_m]  \nonumber\\
&\text{ if } (\langle s^1_i,\sigma_1 \rangle, j) \models \mathbf{~rate~of~firing~of~} a \mathbf{~is~} r_1\nonumber\\
&\vdots \nonumber\\
&\text{ and } (\langle s^m_i,\sigma_m \rangle, j) \models \mathbf{~rate~of~firing~of~} a \mathbf{~is~} r_m \\
(\{ \langle s^1_i,\sigma_1 \rangle,\dots, \langle s^m_i,\sigma_m \rangle \}, j) \models& \mathbf{~totals~of~production~of~} f \mathbf{~atloc~} l \mathbf{~are~} [r_1,\dots,r_m]  \nonumber\\
&\text{ if } (\langle s^1_i,\sigma_1 \rangle, j) \models \mathbf{~total~production~of~} f \mathbf{~atloc~} l \mathbf{~is~} r_1\nonumber\\
&\vdots \nonumber\\
&\text{ and } (\langle s^m_i,\sigma_m \rangle, j) \models \mathbf{~total~production~of~} f \mathbf{~atloc~} l \mathbf{~is~} r_m
\end{align}
}

Next we define how a quantitative all point formula is satisfied w.r.t. a set of trajectories $\sigma_1,\dots,\sigma_m$ in states $s^1_i,\dots,s^m_i$:
{\footnotesize
\begin{align}
\{ \langle s^1_i,\sigma_1 \rangle,\dots, \langle s^m_i,\sigma_m \rangle \} \models& \mathbf{~values~of~} f \mathbf{~atloc~} l\mathbf{~are~} [r_1,\dots,r_m]  \nonumber\\
&\text{ if } \langle s^1_i,\sigma_1 \rangle \models \mathbf{~value~of~} f \mathbf{~atloc~} l \mathbf{~is~} r_1\nonumber\\
&\vdots \nonumber\\
&\text{ and } \langle s^m_i,\sigma_m \rangle \models \mathbf{~value~of~} f \mathbf{~atloc~} l \mathbf{~is~} r_m
\end{align}
}

Next we define how a quantitative aggregate interval formula is satisfied w.r.t. a set of trajectories $\sigma_1,\dots,\sigma_m$ with starting states $s^1_i,\dots,s^m_i$ and end point $j$:
{\footnotesize
\begin{align}
(\{ \langle s^1_i,\sigma_1 \rangle,\dots, \langle s^m_i,\sigma_m \rangle \}, j) \models& \; average \mathbf{~rate~of~production~of~} f \mathbf{~atloc~} l \mathbf{~is~} r  \nonumber\\
&\text{ if } \exists [r_1,\dots,r_m] : \nonumber\\
&~~~~~~(\langle s^1_i,\sigma_1 \rangle, j) \models \mathbf{~rate~of~production~of~} f \mathbf{~atloc~} l \mathbf{~is~} r_1\nonumber\\
&~~~~~~\vdots \nonumber\\
&~~~~~~(\langle s^m_i,\sigma_m \rangle, j) \models \mathbf{~rate~of~production~of~} f \mathbf{~atloc~} l \mathbf{~is~} r_m  \nonumber\\
&\text{ and } r=(r_1+\dots+r_m)/m\\
(\{ \langle s^1_i,\sigma_1 \rangle,\dots, \langle s^m_i,\sigma_m \rangle \}, j) \models& \; minimum \mathbf{~rate~of~production~of~} f \mathbf{~atloc~} l \mathbf{~is~} r  \nonumber\\
&\text{ if } \exists [r_1,\dots,r_m] : \nonumber\\
&~~~~~~(\langle s^1_i,\sigma_1 \rangle, j) \models \mathbf{~rate~of~production~of~} f \mathbf{~atloc~} l \mathbf{~is~} r_1\nonumber\\
&~~~~~~\vdots \nonumber\\
&~~~~~~(\langle s^m_i,\sigma_m \rangle, j) \models \mathbf{~rate~of~production~of~} f \mathbf{~atloc~} l \mathbf{~is~} r_m  \nonumber\\
&\text{ and } \exists k, 1 \leq k \leq m : r=r_k \text{ and } \forall x, 1 \leq x \leq m, r_k \leq r_x\\
(\{ \langle s^1_i,\sigma_1 \rangle,\dots, \langle s^m_i,\sigma_m \rangle \}, j) \models& \; maximum \mathbf{~rate~of~production~of~} f \mathbf{~atloc~} l \mathbf{~is~} r  \nonumber\\
&\text{ if } \exists [r_1,\dots,r_m] : \nonumber\\
&~~~~~~(\langle s^1_i,\sigma_1 \rangle, j) \models \mathbf{~rate~of~production~of~} f \mathbf{~atloc~} l \mathbf{~is~} r_1\nonumber\\
&~~~~~~\vdots \nonumber\\
&~~~~~~(\langle s^m_i,\sigma_m \rangle, j) \models \mathbf{~rate~of~production~of~} f \mathbf{~atloc~} l \mathbf{~is~} r_m  \nonumber\\
&\text{ and } \exists k, 1 \leq k \leq m : r=r_k \text{ and } \forall x, 1 \leq x \leq m, r_k \geq r_x\\
\vspace{20pt}\nonumber\\
(\{ \langle s^1_i,\sigma_1 \rangle,\dots, \langle s^m_i,\sigma_m \rangle \}, j) \models& \; average \mathbf{~rate~of~firing~of~} f \mathbf{~is~} r  \nonumber\\
&\text{ iff } \exists [r_1,\dots,r_m] : \nonumber\\
&~~~~~~(\langle s^1_i,\sigma_1 \rangle, j) \models \mathbf{~rate~of~firing~of~} f \mathbf{~is~} r_1\nonumber\\
&~~~~~~\vdots \nonumber\\
&~~~~~~(\langle s^m_i,\sigma_m \rangle, j) \models \mathbf{~rate~of~firing~of~} f \mathbf{~is~} r_m  \nonumber\\
&\text{ and } r=(r_1+\dots+r_m)/m\\
(\{ \langle s^1_i,\sigma_1 \rangle,\dots, \langle s^m_i,\sigma_m \rangle \}, j) \models& \; minimum \mathbf{~rate~of~firing~of~} f \mathbf{~is~} r  \nonumber\\
&\text{ iff } \exists [r_1,\dots,r_m] : \nonumber\\
&~~~~~~(\langle s^1_i,\sigma_1 \rangle, j) \models \mathbf{~rate~of~firing~of~} f \mathbf{~is~} r_1\nonumber\\
&~~~~~~\vdots \nonumber\\
&~~~~~~(\langle s^m_i,\sigma_m \rangle, j) \models \mathbf{~rate~of~firing~of~} f \mathbf{~is~} r_m  \nonumber\\
&\text{ and } \exists k, 1 \leq k \leq m : r=r_k \text{ and } \forall x, 1 \leq x \leq m, r_k \leq r_x\\
(\{ \langle s^1_i,\sigma_1 \rangle,\dots, \langle s^m_i,\sigma_m \rangle \}, j) \models& \; maximum \mathbf{~rate~of~firing~of~} f \mathbf{~is~} r  \nonumber\\
&\text{ iff } \exists [r_1,\dots,r_m] : \nonumber\\
&~~~~~~(\langle s^1_i,\sigma_1 \rangle, j) \models \mathbf{~rate~of~firing~of~} f \mathbf{~is~} r_1\nonumber\\
&~~~~~~\vdots \nonumber\\
&~~~~~~(\langle s^m_i,\sigma_m \rangle, j) \models \mathbf{~rate~of~firing~of~} f \mathbf{~is~} r_m  \nonumber\\
&\text{ and } \exists k, 1 \leq k \leq m : r=r_k \text{ and } \forall x, 1 \leq x \leq m, r_k \geq r_x\\
\vspace{20pt}\nonumber\\
(\{ \langle s^1_i,\sigma_1 \rangle,\dots, \langle s^m_i,\sigma_m \rangle \}, j) \models& \; average \mathbf{~total~production~of~} f \mathbf{~atloc~} l \mathbf{~is~} r  \nonumber\\
&\text{ if } \exists [r_1,\dots,r_m] : \nonumber\\
&~~~~~~(\langle s^1_i,\sigma_1 \rangle, j) \models \mathbf{~total~production~of~} f \mathbf{~atloc~} l \mathbf{~is~} r_1\nonumber\\
&~~~~~~\vdots \nonumber\\
&~~~~~~(\langle s^m_i,\sigma_m \rangle, j) \models \mathbf{~total~production~of~} f \mathbf{~atloc~} l \mathbf{~is~} r_m  \nonumber\\
&\text{ and } r=(r_1+\dots+r_m)/m\\
(\{ \langle s^1_i,\sigma_1 \rangle,\dots, \langle s^m_i,\sigma_m \rangle \}, j) \models& \; minimum \mathbf{~total~production~of~} f \mathbf{~atloc~} l \mathbf{~is~} r  \nonumber\\
&\text{ if } \exists [r_1,\dots,r_m] : \nonumber\\
&~~~~~~(\langle s^1_i,\sigma_1 \rangle, j) \models \mathbf{~total~production~of~} f \mathbf{~atloc~} l \mathbf{~is~} r_1\nonumber\\
&~~~~~~\vdots \nonumber\\
&~~~~~~(\langle s^m_i,\sigma_m \rangle, j) \models \mathbf{~total~production~of~} f \mathbf{~atloc~} l \mathbf{~is~} r_m  \nonumber\\
&\text{ and } \exists k, 1 \leq k \leq m : r=r_k \text{ and } \forall x, 1 \leq x \leq m, r_k \leq r_x\\
(\{ \langle s^1_i,\sigma_1 \rangle,\dots, \langle s^m_i,\sigma_m \rangle \}, j) \models& \; maximum \mathbf{~total~production~of~} f \mathbf{~atloc~} l \mathbf{~is~} r  \nonumber\\
&\text{ if } \exists [r_1,\dots,r_m] : \nonumber\\
&~~~~~~(\langle s^1_i,\sigma_1 \rangle, j) \models \mathbf{~total~production~of~} f \mathbf{~atloc~} l \mathbf{~is~} r_1\nonumber\\
&~~~~~~\vdots \nonumber\\
&~~~~~~(\langle s^m_i,\sigma_m \rangle, j) \models \mathbf{~total~production~of~} f \mathbf{~atloc~} l \mathbf{~is~} r_m  \nonumber\\
&\text{ and } \exists k, 1 \leq k \leq m : r=r_k \text{ and } \forall x, 1 \leq x \leq m, r_k \geq r_x
\end{align}
}

Next we define how a quantitative aggregate point formula is satisfied w.r.t. a set of trajectories $\sigma_1,\dots,\sigma_m$ in states $s^1_i,\dots,s^m_i$:
{\footnotesize
\begin{align}
\{ \langle s^1_i,\sigma_1 \rangle,\dots, \langle s^m_i,\sigma_m \rangle \}\models& \; average \mathbf{~value~of~} f \mathbf{~atloc~} l \mathbf{~is~} r \nonumber\\
&\text{ if } \exists [r_1,\dots,r_m] : \nonumber\\
&~~~~~~\langle s^1_i,\sigma_1 \rangle \models \mathbf{~value~of~} f \mathbf{~atloc~} l \mathbf{~is~} r_1\nonumber\\
&~~~~~~\vdots \nonumber\\
&~~~~~~\langle s^m_i,\sigma_m \rangle \models \mathbf{~value~of~} f \mathbf{~atloc~} l \mathbf{~is~} r_m  \nonumber\\
&\text{ and } r=(r_1+\dots+r_m)/m\\
\{ \langle s^1_i,\sigma_1 \rangle,\dots, \langle s^m_i,\sigma_m \rangle \}\models& \; minimum \mathbf{~value~of~} f \mathbf{~atloc~} l \mathbf{~is~} r \nonumber\\
&\text{ if } \exists [r_1,\dots,r_m] : \nonumber\\
&~~~~~~\langle s^1_i,\sigma_1 \rangle \models \mathbf{~value~of~} f \mathbf{~atloc~} l \mathbf{~is~} r_1\nonumber\\
&~~~~~~\vdots \nonumber\\
&~~~~~~\langle s^m_i,\sigma_m \rangle \models \mathbf{~value~of~} f \mathbf{~atloc~} l \mathbf{~is~} r_m  \nonumber\\
&\text{ and } \exists k, 1 \leq k \leq m : r=r_k \text{ and } \forall x, 1 \leq x \leq m, r_k \leq r_x\\
\{ \langle s^1_i,\sigma_1 \rangle,\dots, \langle s^m_i,\sigma_m \rangle \}\models& \; maximum \mathbf{~value~of~} f \mathbf{~atloc~} l \mathbf{~is~} r  \nonumber\\
&\text{ if } \exists [r_1,\dots,r_m] : \nonumber\\
&~~~~~~\langle s^1_i,\sigma_1 \rangle \models \mathbf{~value~of~} f \mathbf{~atloc~} l \mathbf{~is~} r_1\nonumber\\
&~~~~~~\vdots \nonumber\\
&~~~~~~\langle s^m_i,\sigma_m \rangle \models \mathbf{~value~of~} f \mathbf{~atloc~} l \mathbf{~is~} r_m  \nonumber\\
&\text{ and } \exists k, 1 \leq k \leq m : r=r_k \text{ and } \forall x, 1 \leq x \leq m, r_k \geq r_x
\end{align}
}

Next we define how a comparative quantitative aggregate interval formula is satisfied w.r.t. two sets of trajectories $\sigma_1,\dots,\sigma_m$, and $\bar{\sigma}_1,\dots,\bar{\sigma}_{\bar{m}}$ a starting point $i$ and an ending point $j$:
{\footnotesize
\begin{align}
&\Big(\Big\{ \{ \langle s^1_i,\sigma_1 \rangle,\dots, \langle s^m_i,\sigma_m \rangle \}, \{ \langle \bar{s}^1_i,\bar{\sigma}_1 \rangle,\dots, \langle \bar{s}^{\bar{m}}_i,\bar{\sigma}_{\bar{m}} \rangle \} \Big\}, j \Big) \nonumber\\
&\;\;\;\;\;\;\models \; \mathbf{~direction~of~change~in~} average \mathbf{~rate~of~production~of~} f \mathbf{~atloc~} l \mathbf{~is~} d  \nonumber\\
&\;\;\;\;\;\;\text{ if } \exists \; n_1 : (\{ \langle s^1_i,\sigma_1 \rangle,\dots, \langle s^m_i,\sigma_m \rangle \}, j) \models \; average \mathbf{~rate~of~production~of~} f \mathbf{~atloc~} l \text{~is~} n_1   \nonumber\\
&\;\;\;\;\;\;\text{ and } \exists \; n_2 : (\{ \langle \bar{s}^1_i,\bar{\sigma}_1 \rangle,\dots, \langle \bar{s}^{\bar{m}}_i,\bar{\sigma}_{\bar{m}} \rangle \}, j) \models \; average \mathbf{~rate~of~production~of~} f \mathbf{~atloc~} l \text{~is~} n_2   \nonumber\\
&\;\;\;\;\;\;\text{ and } n_2 \; d \; n_1\\ 
&\Big(\Big\{ \{ \langle s^1_i,\sigma_1 \rangle,\dots, \langle s^m_i,\sigma_m \rangle \}, \{ \langle \bar{s}^1_i,\bar{\sigma}_1 \rangle,\dots, \langle \bar{s}^{\bar{m}}_i,\bar{\sigma}_{\bar{m}} \rangle \} \Big\}, j \Big) \nonumber\\
&\;\;\;\;\;\;\models \; \mathbf{~direction~of~change~in~} minimum \mathbf{~rate~of~production~of~} f \mathbf{~atloc~} l \mathbf{~is~} d  \nonumber\\
&\;\;\;\;\;\;\text{ if } \exists \; n_1 : (\{ \langle s^1_i,\sigma_1 \rangle,\dots, \langle s^m_i,\sigma_m \rangle \}, j) \models \; minimum \mathbf{~rate~of~production~of~} f \mathbf{~atloc~} l \text{~is~} n_1   \nonumber\\
&\;\;\;\;\;\;\text{ and } \exists \; n_2 : (\{ \langle \bar{s}^1_i,\bar{\sigma}_1 \rangle,\dots, \langle \bar{s}^{\bar{m}}_i,\bar{\sigma}_{\bar{m}} \rangle \}, j) \models \; minimum \mathbf{~rate~of~production~of~} f \mathbf{~atloc~} l \text{~is~} n_2   \nonumber\\
&\;\;\;\;\;\;\text{ and } n_2 \; d \; n_1\\ 
&\Big(\Big\{ \{ \langle s^1_i,\sigma_1 \rangle,\dots, \langle s^m_i,\sigma_m \rangle \}, \{ \langle \bar{s}^1_i,\bar{\sigma}_1 \rangle,\dots, \langle \bar{s}^{\bar{m}}_i,\bar{\sigma}_{\bar{m}} \rangle \} \Big\}, j \Big) \nonumber\\
&\;\;\;\;\;\;\models \; \mathbf{~direction~of~change~in~} maximum \mathbf{~rate~of~production~of~} f \mathbf{~atloc~} l \mathbf{~is~} d  \nonumber\\
&\;\;\;\;\;\;\text{ if } \exists \; n_1 : (\{ \langle s^1_i,\sigma_1 \rangle,\dots, \langle s^m_i,\sigma_m \rangle \}, j) \models \; maximum \mathbf{~rate~of~production~of~} f \mathbf{~atloc~} l \text{~is~} n_1   \nonumber\\
&\;\;\;\;\;\;\text{ and } \exists \; n_2 : (\{ \langle \bar{s}^1_i,\bar{\sigma}_1 \rangle,\dots, \langle \bar{s}^{\bar{m}}_i,\bar{\sigma}_{\bar{m}} \rangle \}, j) \models \; maximum \mathbf{~rate~of~production~of~} f \mathbf{~atloc~} l \text{~is~} n_2   \nonumber\\
&\;\;\;\;\;\;\text{ and } n_2 \; d \; n_1\\
\vspace{20pt}\nonumber\\
&\Big(\Big\{ \{ \langle s^1_i,\sigma_1 \rangle,\dots, \langle s^m_i,\sigma_m \rangle \}, \{ \langle \bar{s}^1_i,\bar{\sigma}_1 \rangle,\dots, \langle \bar{s}^{\bar{m}}_i,\bar{\sigma}_{\bar{m}} \rangle \} \Big\}, j \Big) \nonumber\\
&\;\;\;\;\;\;\models \; \mathbf{~direction~of~change~in~} average \mathbf{~rate~of~firing~of~} f \mathbf{~is~} d  \nonumber\\
&\;\;\;\;\;\;\text{ if } \exists \; n_1 : (\{ \langle s^1_i,\sigma_1 \rangle,\dots, \langle s^m_i,\sigma_m \rangle \}, j) \models \; average \mathbf{~rate~of~firing~of~} f \text{~is~} n_1   \nonumber\\
&\;\;\;\;\;\;\text{ and } \exists \; n_2 : (\{ \langle \bar{s}^1_i,\bar{\sigma}_1 \rangle,\dots, \langle \bar{s}^{\bar{m}}_i,\bar{\sigma}_{\bar{m}} \rangle \}, j) \models \; average \mathbf{~rate~of~firing~of~} f \text{~is~} n_2   \nonumber\\
&\;\;\;\;\;\;\text{ and } n_2 \; d \; n_1\\ 
&\Big(\Big\{ \{ \langle s^1_i,\sigma_1 \rangle,\dots, \langle s^m_i,\sigma_m \rangle \}, \{ \langle \bar{s}^1_i,\bar{\sigma}_1 \rangle,\dots, \langle \bar{s}^{\bar{m}}_i,\bar{\sigma}_{\bar{m}} \rangle \} \Big\}, j \Big) \nonumber\\
&\;\;\;\;\;\;\models \; \mathbf{~direction~of~change~in~} minimum \mathbf{~rate~of~firing~of~} f \mathbf{~is~} d  \nonumber\\
&\;\;\;\;\;\;\text{ if } \exists \; n_1 : (\{ \langle s^1_i,\sigma_1 \rangle,\dots, \langle s^m_i,\sigma_m \rangle \}, j) \models \; minimum \mathbf{~rate~of~firing~of~} f \text{~is~} n_1   \nonumber\\
&\;\;\;\;\;\;\text{ and } \exists \; n_2 : (\{ \langle \bar{s}^1_i,\bar{\sigma}_1 \rangle,\dots, \langle \bar{s}^{\bar{m}}_i,\bar{\sigma}_{\bar{m}} \rangle \}, j) \models \; minimum \mathbf{~rate~of~firing~of~} f \text{~is~} n_2   \nonumber\\
&\;\;\;\;\;\;\text{ and } n_2 \; d \; n_1\\ 
&\Big(\Big\{ \{ \langle s^1_i,\sigma_1 \rangle,\dots, \langle s^m_i,\sigma_m \rangle \}, \{ \langle \bar{s}^1_i,\bar{\sigma}_1 \rangle,\dots, \langle \bar{s}^{\bar{m}}_i,\bar{\sigma}_{\bar{m}} \rangle \} \Big\}, j \Big) \nonumber\\
&\;\;\;\;\;\;\models \; \mathbf{~direction~of~change~in~} maximum \mathbf{~rate~of~firing~of~} f \mathbf{~is~} d  \nonumber\\
&\;\;\;\;\;\;\text{ if } \exists \; n_1 : (\{ \langle s^1_i,\sigma_1 \rangle,\dots, \langle s^m_i,\sigma_m \rangle \}, j) \models \; maximum \mathbf{~rate~of~firing~of~} f \text{~is~} n_1   \nonumber\\
&\;\;\;\;\;\;\text{ and } \exists \; n_2 : (\{ \langle \bar{s}^1_i,\bar{\sigma}_1 \rangle,\dots, \langle \bar{s}^{\bar{m}}_i,\bar{\sigma}_{\bar{m}} \rangle \}, j) \models \; maximum \mathbf{~rate~of~firing~of~} f \text{~is~} n_2   \nonumber\\
&\;\;\;\;\;\;\text{ and } n_2 \; d \; n_1\\
\vspace{20pt}\nonumber\\
&\Big(\Big\{ \{ \langle s^1_i,\sigma_1 \rangle,\dots, \langle s^m_i,\sigma_m \rangle \}, \{ \langle \bar{s}^1_i,\bar{\sigma}_1 \rangle,\dots, \langle \bar{s}^{\bar{m}}_i,\bar{\sigma}_{\bar{m}} \rangle \} \Big\}, j \Big) \nonumber\\
&\;\;\;\;\;\;\models \; \mathbf{~direction~of~change~in~} average \mathbf{~total~production~of~} f \mathbf{~atloc~} l \mathbf{~is~} d  \nonumber\\
&\;\;\;\;\;\;\text{ if } \exists \; n_1 : (\{ \langle s^1_i,\sigma_1 \rangle,\dots, \langle s^m_i,\sigma_m \rangle \}, j) \models \; average \mathbf{~total~production~of~} f \mathbf{~atloc~} l \text{~is~} n_1   \nonumber\\
&\;\;\;\;\;\;\text{ and } \exists \; n_2 : (\{ \langle \bar{s}^1_i,\bar{\sigma}_1 \rangle,\dots, \langle \bar{s}^{\bar{m}}_i,\bar{\sigma}_{\bar{m}} \rangle \}, j) \models \; average \mathbf{~total~production~of~} f \mathbf{~atloc~} l \text{~is~} n_2   \nonumber\\
&\;\;\;\;\;\;\text{ and } n_2 \; d \; n_1\\ 
&\Big(\Big\{ \{ \langle s^1_i,\sigma_1 \rangle,\dots, \langle s^m_i,\sigma_m \rangle \}, \{ \langle \bar{s}^1_i,\bar{\sigma}_1 \rangle,\dots, \langle \bar{s}^{\bar{m}}_i,\bar{\sigma}_{\bar{m}} \rangle \} \Big\}, j \Big) \nonumber\\
&\;\;\;\;\;\;\models \; \mathbf{~direction~of~change~in~} minimum \mathbf{~total~production~of~} f \mathbf{~atloc~} l \mathbf{~is~} d  \nonumber\\
&\;\;\;\;\;\;\text{ if } \exists \; n_1 : (\{ \langle s^1_i,\sigma_1 \rangle,\dots, \langle s^m_i,\sigma_m \rangle \}, j) \models \; minimum \mathbf{~total~production~of~} f \mathbf{~atloc~} l \text{~is~} n_1   \nonumber\\
&\;\;\;\;\;\;\text{ and } \exists \; n_2 : (\{ \langle \bar{s}^1_i,\bar{\sigma}_1 \rangle,\dots, \langle \bar{s}^{\bar{m}}_i,\bar{\sigma}_{\bar{m}} \rangle \}, j) \models \; minimum \mathbf{~total~production~of~} f \mathbf{~atloc~} l \text{~is~} n_2   \nonumber\\
&\;\;\;\;\;\;\text{ and } n_2 \; d \; n_1\\ 
&\Big(\Big\{ \{ \langle s^1_i,\sigma_1 \rangle,\dots, \langle s^m_i,\sigma_m \rangle \}, \{ \langle \bar{s}^1_i,\bar{\sigma}_1 \rangle,\dots, \langle \bar{s}^{\bar{m}}_i,\bar{\sigma}_{\bar{m}} \rangle \} \Big\}, j \Big) \nonumber\\
&\;\;\;\;\;\;\models \; \mathbf{~direction~of~change~in~} maximum \mathbf{~total~production~of~} f \mathbf{~atloc~} l \mathbf{~is~} d  \nonumber\\
&\;\;\;\;\;\;\text{ if } \exists \; n_1 : (\{ \langle s^1_i,\sigma_1 \rangle,\dots, \langle s^m_i,\sigma_m \rangle \}, j) \models \; maximum \mathbf{~total~production~of~} f \mathbf{~atloc~} l \text{~is~} n_1   \nonumber\\
&\;\;\;\;\;\;\text{ and } \exists \; n_2 : (\{ \langle \bar{s}^1_i,\bar{\sigma}_1 \rangle,\dots, \langle \bar{s}^{\bar{m}}_i,\bar{\sigma}_{\bar{m}} \rangle \}, j) \models \; maximum \mathbf{~total~production~of~} f \mathbf{~atloc~} l \text{~is~} n_2   \nonumber\\
&\;\;\;\;\;\;\text{ and } n_2 \; d \; n_1
\end{align}
}

Next we define how a comparative quantitative aggregate point formula is satisfied w.r.t. two sets of trajectories $\sigma_1,\dots,\sigma_m$, and $\bar{\sigma}_1,\dots,\bar{\sigma}_{\bar{m}}$ at point $i$:
{\footnotesize
\begin{align}
&\Big\{ \{ \langle s^1_i,\sigma_1 \rangle,\dots, \langle s^m_i,\sigma_m \rangle \}, \{ \langle \bar{s}^1_i,\bar{\sigma}_1 \rangle,\dots, \langle \bar{s}^{\bar{m}}_i,\bar{\sigma}_{\bar{m}} \rangle \} \Big\} \nonumber\\
&\;\;\;\;\;\;\models \; \mathbf{~direction~of~change~in~} average \mathbf{~value~of~} f \mathbf{~atloc~} l \mathbf{~is~} d  \nonumber\\
&\;\;\;\;\;\;\text{ if } \exists \; n_1 : \{ \langle s^1_i,\sigma_1 \rangle,\dots, \langle s^m_i,\sigma_m \rangle \} \models \; average \mathbf{~value~of~} f \mathbf{~atloc~} l \text{~is~} n_1   \nonumber\\
&\;\;\;\;\;\;\text{ and } \exists \; n_2 : \{ \langle \bar{s}^1_i,\bar{\sigma}_1 \rangle,\dots, \langle \bar{s}^{\bar{m}}_i,\bar{\sigma}_{\bar{m}} \rangle \} \models \; average \mathbf{~value~of~} f \mathbf{~atloc~} l \text{~is~} n_2   \nonumber\\
&\;\;\;\;\;\;\text{ and } n_2 \; d \; n_1\\ 
&\Big\{ \{ \langle s^1_i,\sigma_1 \rangle,\dots, \langle s^m_i,\sigma_m \rangle \}, \{ \langle \bar{s}^1_i,\bar{\sigma}_1 \rangle,\dots, \langle \bar{s}^{\bar{m}}_i,\bar{\sigma}_{\bar{m}} \rangle \} \Big\} \nonumber\\
&\;\;\;\;\;\;\models \; \mathbf{~direction~of~change~in~} minimum \mathbf{~value~of~} f \mathbf{~atloc~} l \mathbf{~is~} d  \nonumber\\
&\;\;\;\;\;\;\text{ if } \exists \; n_1 : \{ \langle s^1_i,\sigma_1 \rangle,\dots, \langle s^m_i,\sigma_m \rangle \} \models \; minimum \mathbf{~value~of~} f \mathbf{~atloc~} l \text{~is~} n_1   \nonumber\\
&\;\;\;\;\;\;\text{ and } \exists \; n_2 : \{ \langle \bar{s}^1_i,\bar{\sigma}_1 \rangle,\dots, \langle \bar{s}^{\bar{m}}_i,\bar{\sigma}_{\bar{m}} \rangle \} \models \; minimum \mathbf{~value~of~} f \mathbf{~atloc~} l \text{~is~} n_2   \nonumber\\
&\;\;\;\;\;\;\text{ and } n_2 \; d \; n_1\\ 
&\Big\{ \{ \langle s^1_i,\sigma_1 \rangle,\dots, \langle s^m_i,\sigma_m \rangle \}, \{ \langle \bar{s}^1_i,\bar{\sigma}_1 \rangle,\dots, \langle \bar{s}^{\bar{m}}_i,\bar{\sigma}_{\bar{m}} \rangle \} \Big\} \nonumber\\
&\;\;\;\;\;\;\models \; \mathbf{~direction~of~change~in~} maximum \mathbf{~value~of~} f \mathbf{~atloc~} l \mathbf{~is~} d  \nonumber\\
&\;\;\;\;\;\;\text{ if } \exists \; n_1 : \{ \langle s^1_i,\sigma_1 \rangle,\dots, \langle s^m_i,\sigma_m \rangle \} \models \; maximum \mathbf{~value~of~} f \mathbf{~atloc~} l \text{~is~} n_1   \nonumber\\
&\;\;\;\;\;\;\text{ and } \exists \; n_2 : \{ \langle \bar{s}^1_i,\bar{\sigma}_1 \rangle,\dots, \langle \bar{s}^{\bar{m}}_i,\bar{\sigma}_{\bar{m}} \rangle \} \models \; maximum \mathbf{~value~of~} f \mathbf{~atloc~} l \text{~is~} n_2   \nonumber\\
&\;\;\;\;\;\;\text{ and } n_2 \; d \; n_1
\end{align}
}

\subsection{Trajectory Filtering due to Internal Observations}\label{dqa:sem:obs:filter}
The trajectories produced by the Guarded-Arc Petri Net execution are filtered to retain only the trajectories that satisfy all internal observations in a query statement. Let $\sigma = s_0,\dots,s_k$ be a trajectory as given in definition~\ref{def:gpn:traj}. Then $\sigma$ satisfies an observation:
{\footnotesize
\begin{align}
&\mathbf{~rate~of~production~of~} f \mathbf{~is~} n \nonumber\\
&~~~~~~~~\text{ if } (\langle s_0,\sigma \rangle, k) \models \mathbf{~rate~of~production~of~} f \mathbf{~is~} n \\
&\mathbf{~rate~of~production~of~} f \mathbf{~is~} n \nonumber\\
&~~~~~~~~\text{ if } (\langle s_0,\sigma \rangle, k) \models \mathbf{~rate~of~firing~of~} a \mathbf{~is~} n \\
&\mathbf{~total~production~of~} f \mathbf{~is~} n \nonumber\\
&~~~~~~~~\text{ if } (\langle s_0,\sigma \rangle, k) \models \mathbf{~total~production~of~} f \mathbf{~is~} n \\
&f \mathbf{~is~accumulating~} \nonumber\\
&~~~~~~~~\text{ if } (\langle s_0,\sigma \rangle, k) \models f \mathbf{~is~accumulating~}  \\
&f \mathbf{~is~decreasing~} \nonumber\\
&~~~~~~~~\text{ if } (\langle s_0,\sigma \rangle, k) \models f \mathbf{~is~decreasing~}  \\
\vspace{20pt}\nonumber\\
&\mathbf{~rate~of~production~of~} f \mathbf{~is~} n \nonumber\\
&~~~~~~~~\mathbf{when~observed~between~time~step~} i \mathbf{~and~time~step~} j \nonumber\\
&~~~~~~~~\text{ if } (\langle s_i,\sigma \rangle, j) \models \mathbf{~rate~of~production~of~} f \mathbf{~is~} n \\
&\mathbf{~rate~of~production~of~} f \mathbf{~is~} n \nonumber\\
&~~~~~~~~\mathbf{when~observed~between~time~step~} i \mathbf{~and~time~step~} j \nonumber\\
&~~~~~~~~\text{ if } (\langle s_i,\sigma \rangle, j) \models \mathbf{~rate~of~firing~of~} a \mathbf{~is~} n \\
&\mathbf{~total~production~of~} f \mathbf{~is~} n \nonumber\\
&~~~~~~~~\mathbf{when~observed~between~time~step~} i \mathbf{~and~time~step~} j \nonumber\\
&~~~~~~~~\text{ if } (\langle s_i,\sigma \rangle, j) \models \mathbf{~total~production~of~} f \mathbf{~is~} n \\
&f \mathbf{~is~accumulating~} \nonumber\\
&~~~~~~~~\mathbf{when~observed~between~time~step~} i \mathbf{~and~time~step~} j \nonumber\\
&~~~~~~~~\text{ if } (\langle s_i,\sigma \rangle, j) \models f \mathbf{~is~accumulating~}  \\
&f \mathbf{~is~decreasing~} \nonumber\\
&~~~~~~~~\mathbf{when~observed~between~time~step~} i \mathbf{~and~time~step~} j \nonumber\\
&~~~~~~~~\text{ if } (\langle s_i,\sigma \rangle, j) \models f \mathbf{~is~decreasing~}  \\
\vspace{20pt}\nonumber\\
&f \mathbf{~is~higher~than~} n \nonumber\\
&~~~~~~~~\text{ if } \exists i, 0 \leq i \leq k: \langle s_i,\sigma \rangle \models \mathbf{~value~of~} f \mathbf{~is~higher~than~} n \\
&f \mathbf{~is~lower~than~} n \nonumber\\
&~~~~~~~~\text{ if } \exists i, 0 \leq i \leq k: \langle s_i,\sigma \rangle \models \mathbf{~value~of~} f \mathbf{~is~lower~than~} n \\
&f \mathbf{~is~} n \nonumber\\
&~~~~~~~~\text{ if } \exists i, 0 \leq i \leq k: \langle s_i,\sigma \rangle \models \mathbf{~value~of~} f \mathbf{~is~} n \\
&a \mathbf{~occurs~} \nonumber\\
&~~~~~~~~\text{ if } \exists i, 0 \leq i \leq k: \langle s_i,\sigma \rangle \models a \mathbf{~occurs~} \\
&a \mathbf{~does~not~occur~} \nonumber\\
&~~~~~~~~\text{ if } \exists i, 0 \leq i \leq k: \langle s_i,\sigma \rangle \models a \mathbf{~does~not~occur~} \\
&a_1 \mathbf{~switches~to~} a_2 \nonumber\\
&~~~~~~~~\text{ if } \exists i, 0 \leq i \leq k: \langle s_i,\sigma \rangle \models a_1 \mathbf{~switches~to~} a_2 
\vspace{20pt}\nonumber\\
&f \mathbf{~is~higher~than~} n \nonumber\\
&~~~~~~~~\mathbf{at~time~step~} i \nonumber\\
&~~~~~~~~\text{ if } \langle s_i,\sigma \rangle \models \mathbf{~value~of~} f \mathbf{~is~higher~than~} n \\
&f \mathbf{~is~lower~than~} n \nonumber\\
&~~~~~~~~\mathbf{at~time~step~} i \nonumber\\
&~~~~~~~~\text{ if } \langle s_i,\sigma \rangle \models \mathbf{~value~of~} f \mathbf{~is~lower~than~} n \\
&f \mathbf{~is~} n \nonumber\\
&~~~~~~~~\mathbf{at~time~step~} i \nonumber\\
&~~~~~~~~\text{ if } \langle s_i,\sigma \rangle \models \mathbf{~value~of~} f \mathbf{~is~} n \\
&a \mathbf{~occurs~} \nonumber\\
&~~~~~~~~\mathbf{at~time~step~} i \nonumber\\
&~~~~~~~~\text{ if } \langle s_i,\sigma \rangle \models a \mathbf{~occurs~} \\
&a \mathbf{~does~not~occur~} \nonumber\\
&~~~~~~~~\mathbf{at~time~step~} i \nonumber\\
&~~~~~~~~\text{ if } \langle s_i,\sigma \rangle \models a \mathbf{~does~not~occur~} \\
&a_1 \mathbf{~switches~to~} a_2 \nonumber\\
&~~~~~~~~\mathbf{at~time~step~} i \nonumber\\
&~~~~~~~~\text{ if } \langle s_i,\sigma \rangle \models a_1 \mathbf{~switches~to~} a_2 
\end{align}
}

Let $\sigma = s_0,\dots,s_k$ be a trajectory of the form \eqref{def:gcpn:traj}. Then $\sigma$ satisfies an observation:
{\footnotesize
\begin{align}
&\mathbf{~rate~of~production~of~} f \mathbf{~atloc~} l \mathbf{~is~} n \nonumber\\
&~~~~~~~~\text{ if } (\langle s_0,\sigma \rangle, k) \models \mathbf{~rate~of~production~of~} f \mathbf{~atloc~} l \mathbf{~is~} n \\
&\mathbf{~rate~of~production~of~} f \mathbf{~atloc~} l \mathbf{~is~} n \nonumber\\
&~~~~~~~~\text{ if } (\langle s_0,\sigma \rangle, k) \models \mathbf{~rate~of~firing~of~} a \mathbf{~is~} n \\
&\mathbf{~total~production~of~} f \mathbf{~atloc~} l \mathbf{~is~} n \nonumber\\
&~~~~~~~~\text{ if } (\langle s_0,\sigma \rangle, k) \models \mathbf{~total~production~of~} f \mathbf{~atloc~} l \mathbf{~is~} n \\
&f \mathbf{~is~accumulating~} \mathbf{~atloc~} l \nonumber\\
&~~~~~~~~\text{ if } (\langle s_0,\sigma \rangle, k) \models f \mathbf{~is~accumulating~} \mathbf{~atloc~} l  \\
&f \mathbf{~is~decreasing~} \mathbf{~atloc~} l \nonumber\\
&~~~~~~~~\text{ if } (\langle s_0,\sigma \rangle, k) \models f \mathbf{~is~decreasing~} \mathbf{~atloc~} l  \\
\vspace{20pt}\nonumber\\
&\mathbf{~rate~of~production~of~} f \mathbf{~atloc~} l \mathbf{~is~} n \nonumber\\
&~~~~~~~~\mathbf{when~observed~between~time~step~} i \mathbf{~and~time~step~} j \nonumber\\
&~~~~~~~~\text{ if } (\langle s_i,\sigma \rangle, j) \models \mathbf{~rate~of~production~of~} f \mathbf{~atloc~} l \mathbf{~is~} n \\
&\mathbf{~rate~of~production~of~} f \mathbf{~atloc~} l \mathbf{~is~} n \nonumber\\
&~~~~~~~~\mathbf{when~observed~between~time~step~} i \mathbf{~and~time~step~} j \nonumber\\
&~~~~~~~~\text{ if } (\langle s_i,\sigma \rangle, j) \models \mathbf{~rate~of~firing~of~} a \mathbf{~is~} n \\
&\mathbf{~total~production~of~} f \mathbf{~atloc~} l \mathbf{~is~} n \nonumber\\
&~~~~~~~~\mathbf{when~observed~between~time~step~} i \mathbf{~and~time~step~} j \nonumber\\
&~~~~~~~~\text{ if } (\langle s_i,\sigma \rangle, j) \models \mathbf{~total~production~of~} f \mathbf{~atloc~} l \mathbf{~is~} n \\
&f \mathbf{~is~accumulating~} \mathbf{~atloc~} l \nonumber\\
&~~~~~~~~\mathbf{when~observed~between~time~step~} i \mathbf{~and~time~step~} j \nonumber\\
&~~~~~~~~\text{ if } (\langle s_i,\sigma \rangle, j) \models f \mathbf{~is~accumulating~} \mathbf{~atloc~} l  \\
&f \mathbf{~is~decreasing~} \mathbf{~atloc~} l \nonumber\\
&~~~~~~~~\mathbf{when~observed~between~time~step~} i \mathbf{~and~time~step~} j \nonumber\\
&~~~~~~~~\text{ if } (\langle s_i,\sigma \rangle, j) \models f \mathbf{~is~decreasing~} \mathbf{~atloc~} l  \\
\vspace{20pt}\nonumber\\
&f \mathbf{~atloc~} l \mathbf{~is~higher~than~} n \nonumber\\
&~~~~~~~~\text{ if } \exists i, 0 \leq i \leq k: \langle s_i,\sigma \rangle \models \mathbf{~value~of~} f \mathbf{~atloc~} l \mathbf{~is~higher~than~} n \\
&f \mathbf{~atloc~} l \mathbf{~is~lower~than~} n \nonumber\\
&~~~~~~~~\text{ if } \exists i, 0 \leq i \leq k: \langle s_i,\sigma \rangle \models \mathbf{~value~of~} f \mathbf{~atloc~} l \mathbf{~is~lower~than~} n \\
&f \mathbf{~atloc~} l \mathbf{~is~} n \nonumber\\
&~~~~~~~~\text{ if } \exists i, 0 \leq i \leq k: \langle s_i,\sigma \rangle \models \mathbf{~value~of~} f \mathbf{~atloc~} l \mathbf{~is~} n \\
&a \mathbf{~occurs~} \nonumber\\
&~~~~~~~~\text{ if } \exists i, 0 \leq i \leq k: \langle s_i,\sigma \rangle \models a \mathbf{~occurs~} \\
&a \mathbf{~does~not~occur~} \nonumber\\
&~~~~~~~~\text{ if } \exists i, 0 \leq i \leq k: \langle s_i,\sigma \rangle \models a \mathbf{~does~not~occur~} \\
&a_1 \mathbf{~switches~to~} a_2 \nonumber\\
&~~~~~~~~\text{ if } \exists i, 0 \leq i \leq k: \langle s_i,\sigma \rangle \models a_1 \mathbf{~switches~to~} a_2 
\vspace{20pt}\nonumber\\
&f \mathbf{~atloc~} l \mathbf{~is~higher~than~} n \nonumber\\
&~~~~~~~~\mathbf{at~time~step~} i \nonumber\\
&~~~~~~~~\text{ if } \langle s_i,\sigma \rangle \models \mathbf{~value~of~} f \mathbf{~atloc~} l \mathbf{~is~higher~than~} n \\
&f \mathbf{~atloc~} l \mathbf{~is~lower~than~} n \nonumber\\
&~~~~~~~~\mathbf{at~time~step~} i \nonumber\\
&~~~~~~~~\text{ if } \langle s_i,\sigma \rangle \models \mathbf{~value~of~} f \mathbf{~atloc~} l \mathbf{~is~lower~than~} n \\
&f \mathbf{~atloc~} l \mathbf{~is~} n \nonumber\\
&~~~~~~~~\mathbf{at~time~step~} i \nonumber\\
&~~~~~~~~\text{ if } \langle s_i,\sigma \rangle \models \mathbf{~value~of~} f \mathbf{~atloc~} l \mathbf{~is~} n \\
&a \mathbf{~occurs~} \nonumber\\
&~~~~~~~~\mathbf{at~time~step~} i \nonumber\\
&~~~~~~~~\text{ if } \langle s_i,\sigma \rangle \models a \mathbf{~occurs~} \\
&a \mathbf{~does~not~occur~} \nonumber\\
&~~~~~~~~\mathbf{at~time~step~} i \nonumber\\
&~~~~~~~~\text{ if } \langle s_i,\sigma \rangle \models a \mathbf{~does~not~occur~} \\
&a_1 \mathbf{~switches~to~} a_2 \nonumber\\
&~~~~~~~~\mathbf{at~time~step~} i \nonumber\\
&~~~~~~~~\text{ if } \langle s_i,\sigma \rangle \models a_1 \mathbf{~switches~to~} a_2 
\end{align}
}

A trajectory $\sigma$ is kept for further processing w.r.t. a set of internal observations\\
 $\langle \text{internal observation} \rangle_1,\dots,\langle \text{internal observation} \rangle_n$ if $\sigma \models \langle \text{internal observation} \rangle_i$, $1 \leq i \leq n$.

\subsection{Query Description Satisfaction}\label{sec:dqa:qdesc:satisfaction}
Now, we define query statement semantics using LTL syntax. Let $\mathbf{D}$ be a domain description with simple fluents and $\sigma = s_0,\dots,s_k$ be its trajectory of length $k$ as defined in~\eqref{def:gpn:traj}. Let $\sigma_1,\dots,\sigma_m$ represent the set of trajectories of $\mathbf{D}$ filtered by observations as necessary, with each trajectory has the form $\sigma_i = s^i_0,\dots,s^i_k$ $1\leq i \leq m$. Let $\mathbf{\bar{D}}$ be a modified domain description and $\bar{\sigma}_1,\dots,\bar{\sigma}_{\bar{m}}$ be its trajectories of the form $\bar{\sigma}_i = \bar{s}^i_0,\dots,\bar{s}^i_k$. Then we define query satisfaction using the formula satisfaction in section~\ref{dqa:sem:formula} as follows.

Two sets of trajectories $\sigma_1,\dots,\sigma_m$ and $\bar{\sigma}_1,\dots,\bar{\sigma}_{\bar{m}}$ satisfy a comparative query description based on formula satisfaction of section~\ref{dqa:sem:formula} as follows:
{\footnotesize
\begin{align}
&\Big\{ \{ \langle s^1_0,\sigma_1 \rangle,\dots, \langle s^m_0,\sigma_m \rangle \}, \{ \langle \bar{s}^1_0,\bar{\sigma}_1 \rangle,\dots, \langle \bar{s}^{\bar{m}}_0,\bar{\sigma}_{\bar{m}} \rangle \} \Big\} \nonumber\\
&\;\;\;\;\;\;\models \; \mathbf{~direction~of~change~in~} aggop \mathbf{~rate~of~production~of~} f \mathbf{~is~} d \nonumber\\
&\;\;\;\;\;\;\;\;\;\;\;\;\;\;\;\;\;\mathbf{~when~observed~between~time~step~} i \mathbf{~and~time~step~} j \nonumber\\
&\;\;\;\;\;\;\text{ if } \Big(\Big\{ \{ \langle s^1_i,\sigma_1 \rangle,\dots, \langle s^m_i,\sigma_m \rangle \}, \{ \langle \bar{s}^1_i,\bar{\sigma}_1 \rangle,\dots, \langle \bar{s}^{\bar{m}}_i,\bar{\sigma}_{\bar{m}} \rangle \} \Big\}, j \Big) \nonumber\\
&\;\;\;\;\;\;\;\;\;\;\;\;\models \mathbf{~direction~of~change~in~} aggop \mathbf{~rate~of~production~of~} f \mathbf{~is~} d \\
&\Big\{ \{ \langle s^1_0,\sigma_1 \rangle,\dots, \langle s^m_0,\sigma_m \rangle \}, \{ \langle \bar{s}^1_0,\bar{\sigma}_1 \rangle,\dots, \langle \bar{s}^{\bar{m}}_0,\bar{\sigma}_{\bar{m}} \rangle \} \Big\} \nonumber\\
&\;\;\;\;\;\;\models \; \mathbf{~direction~of~change~in~} aggop \mathbf{~rate~of~firing~of~} f \mathbf{~is~} d \nonumber\\
&\;\;\;\;\;\;\;\;\;\;\;\;\;\;\;\;\;\mathbf{~when~observed~between~time~step~} i \mathbf{~and~time~step~} j \nonumber\\
&\;\;\;\;\;\;\text{ if } \Big(\Big\{ \{ \langle s^1_i,\sigma_1 \rangle,\dots, \langle s^m_i,\sigma_m \rangle \}, \{ \langle \bar{s}^1_i,\bar{\sigma}_1 \rangle,\dots, \langle \bar{s}^{\bar{m}}_i,\bar{\sigma}_{\bar{m}} \rangle \} \Big\}, j\Big) \nonumber\\
&\;\;\;\;\;\;\;\;\;\;\;\;\models \mathbf{~direction~of~change~in~} aggop \mathbf{~rate~of~firing~of~} f \mathbf{~is~} d \\
&\Big\{ \{ \langle s^1_0,\sigma_1 \rangle,\dots, \langle s^m_0,\sigma_m \rangle \}, \{ \langle \bar{s}^1_0,\bar{\sigma}_1 \rangle,\dots, \langle \bar{s}^{\bar{m}}_0,\bar{\sigma}_{\bar{m}} \rangle \} \Big\} \nonumber\\
&\;\;\;\;\;\;\models \; \mathbf{~direction~of~change~in~} aggop \mathbf{~total~production~of~} f \mathbf{~is~} d \nonumber\\
&\;\;\;\;\;\;\;\;\;\;\;\;\;\;\;\;\;\mathbf{~when~observed~between~time~step~} i \mathbf{~and~time~step~} j \nonumber\\
&\;\;\;\;\;\;\text{ if } \Big(\Big\{ \{ \langle s^1_i,\sigma_1 \rangle,\dots, \langle s^m_i,\sigma_m \rangle \}, \{ \langle \bar{s}^1_i,\bar{\sigma}_1 \rangle,\dots, \langle \bar{s}^{\bar{m}}_i,\bar{\sigma}_{\bar{m}} \rangle \} \Big\}, j\Big) \nonumber\\
&\;\;\;\;\;\;\;\;\;\;\;\;\models \mathbf{~direction~of~change~in~} aggop \mathbf{~total~production~of~} f \mathbf{~is~} d \\
&\Big\{ \{ \langle s^1_0,\sigma_1 \rangle,\dots, \langle s^m_0,\sigma_m \rangle \}, \{ \langle \bar{s}^1_0,\bar{\sigma}_1 \rangle,\dots, \langle \bar{s}^{\bar{m}}_0,\bar{\sigma}_{\bar{m}} \rangle \} \Big\} \nonumber\\
&\;\;\;\;\;\;\models \; \mathbf{~direction~of~change~in~} aggop \mathbf{~value~of~} f \mathbf{~is~} d \nonumber\\
&\;\;\;\;\;\;\;\;\;\;\;\;\;\;\;\;\;\mathbf{~when~observed~at~time~step~} i \nonumber\\
&\;\;\;\;\;\;\text{ if } \Big\{ \{ \langle s^1_i,\sigma_1 \rangle,\dots, \langle s^m_i,\sigma_m \rangle \}, \{ \langle \bar{s}^1_i,\bar{\sigma}_1 \rangle,\dots, \langle \bar{s}^{\bar{m}}_i,\bar{\sigma}_{\bar{m}} \rangle \} \Big\} \nonumber\\
&\;\;\;\;\;\;\;\;\;\;\;\;\models \mathbf{~direction~of~change~in~} aggop \mathbf{~value~of~} f \mathbf{~is~} d 
\end{align}
}

A set of trajectories $\sigma_1,\dots,\sigma_m$, each of length $k$ satisfies a quantitative or a qualitative interval query description based on formula satisfaction of section~\ref{dqa:sem:formula} as follows:
{\footnotesize
\begin{align}
&\{ \langle s^1_0,\sigma_1 \rangle,\dots, \langle s^m_0,\sigma_m \rangle \} \models \mathbf{~rate~of~production~of~} f \mathbf{~is~} n \nonumber\\
&~~~~\text{ if } \exists x, 1 \leq x \leq m : (\langle s^x_0,\sigma_x \rangle, k) \models \mathbf{~rate~of~production~of~} f \mathbf{~is~} n  \\
&\{ \langle s^1_0,\sigma_1 \rangle,\dots, \langle s^m_0,\sigma_m \rangle \} \models \mathbf{~rate~of~firing~of~} a \mathbf{~is~} n \nonumber\\
&~~~~\text{ if } \exists x, 1 \leq x \leq m : (\langle s^x_0,\sigma_x \rangle, k) \models \mathbf{~rate~of~firing~of~} a \mathbf{~is~} n  \\
&\{ \langle s^1_0,\sigma_1 \rangle,\dots, \langle s^m_0,\sigma_m \rangle \} \models \mathbf{~total~production~of~} f \mathbf{~is~} n \nonumber\\
&~~~~\text{ if } \exists x, 1 \leq x \leq m : (\langle s^x_0,\sigma_x \rangle, k) \models \mathbf{~total~production~of~} f \mathbf{~is~} n  \\
\vspace{20pt}\nonumber\\
&\{ \langle s^1_0,\sigma_1 \rangle,\dots, \langle s^m_0,\sigma_m \rangle \} \models \mathbf{~rate~of~production~of~} f \mathbf{~is~} n \nonumber\\
&\hspace{30pt}\mathbf{~when~observed~between~time~step~} i \mathbf{~and~time~step~} j \nonumber\\
&~~~~\text{ if } \exists x, 1 \leq x \leq m : (\langle s^x_i,\sigma_x \rangle, j) \models \mathbf{~rate~of~production~of~} f \mathbf{~is~} n  \\
&\{ \langle s^1_0,\sigma_1 \rangle,\dots, \langle s^m_0,\sigma_m \rangle \} \models \mathbf{~rate~of~firing~of~} a \mathbf{~is~} n \nonumber\\
&\hspace{30pt}\mathbf{~when~observed~between~time~step~} i \mathbf{~and~time~step~} j \nonumber\\
&~~~~\text{ if } \exists x, 1 \leq x \leq m : (\langle s^x_i,\sigma_x \rangle, j) \models \mathbf{~rate~of~firing~of~} a \mathbf{~is~} n  \\
&\{ \langle s^1_0,\sigma_1 \rangle,\dots, \langle s^m_0,\sigma_m \rangle \} \models \mathbf{~total~production~of~} f \mathbf{~is~} n \nonumber\\
&\hspace{30pt}\mathbf{~when~observed~between~time~step~} i \mathbf{~and~time~step~} j \nonumber\\
&~~~~\text{ if } \exists x, 1 \leq x \leq m : (\langle s^x_i,\sigma_x \rangle, j) \models \mathbf{~total~production~of~} f \mathbf{~is~} n \\
\vspace{20pt}\nonumber\\
&\{ \langle s^1_0,\sigma_1 \rangle,\dots, \langle s^m_0,\sigma_m \rangle \} \models f \mathbf{~is~accumulating~} \nonumber\\
&~~~~\text{ if } \exists x, 1 \leq x \leq m : (\langle s^x_0,\sigma_x \rangle, k) \models f \mathbf{~is~accumulating~}  \\
&\{ \langle s^1_0,\sigma_1 \rangle,\dots, \langle s^m_0,\sigma_m \rangle \} \models f \mathbf{~is~decreasing~} \nonumber\\
&~~~~\text{ if } \exists x, 1 \leq x \leq m : (\langle s^x_0,\sigma_x \rangle, k) \models f \mathbf{~is~decreasing~}  \\
\vspace{20pt}\nonumber\\
&\{ \langle s^1_0,\sigma_1 \rangle,\dots, \langle s^m_0,\sigma_m \rangle \} \models f \mathbf{~is~accumulating~} \nonumber\\
&\hspace{30pt}\mathbf{~when~observed~between~time~step~} i \mathbf{~and~time~step~} j \nonumber\\
&~~~~\text{ if } \exists x, 1 \leq x \leq m : (\langle s^x_i,\sigma_x \rangle, j) \models f \mathbf{~is~accumulating~}  \\
&\{ \langle s^1_0,\sigma_1 \rangle,\dots, \langle s^m_0,\sigma_m \rangle \} \models f \mathbf{~is~decreasing~} \nonumber\\
&\hspace{30pt}\mathbf{~when~observed~between~time~step~} i \mathbf{~and~time~step~} j \nonumber\\
&~~~~\text{ if } \exists x, 1 \leq x \leq m : (\langle s^x_i,\sigma_x \rangle, j) \models f \mathbf{~is~decreasing~}  \\
\vspace{20pt}\nonumber\\
&\{ \langle s^1_0,\sigma_1 \rangle,\dots, \langle s^m_0,\sigma_m \rangle \} \models \mathbf{~rate~of~production~of~} f \mathbf{~is~} n \nonumber\\
&\hspace{30pt}\mathbf{~in~all~trajectories~} \nonumber\\
&~~~~\text{ if } \forall x, 1 \leq x \leq m : (\langle s^x_0,\sigma_x \rangle, k) \models \mathbf{~rate~of~production~of~} f \mathbf{~is~} n  \\
&\{ \langle s^1_0,\sigma_1 \rangle,\dots, \langle s^m_0,\sigma_m \rangle \} \models \mathbf{~rate~of~firing~of~} a \mathbf{~is~} n \nonumber\\
&\hspace{30pt}\mathbf{~in~all~trajectories~} \nonumber\\
&~~~~\text{ if } \forall x, 1 \leq x \leq m : (\langle s^x_0,\sigma_x \rangle, k) \models \mathbf{~rate~of~firing~of~} a \mathbf{~is~} n  \\
&\{ \langle s^1_0,\sigma_1 \rangle,\dots, \langle s^m_0,\sigma_m \rangle \} \models \mathbf{~total~production~of~} f \mathbf{~is~} n \nonumber\\
&\hspace{30pt}\mathbf{~in~all~trajectories~} \nonumber\\
&~~~~\text{ if } \forall x, 1 \leq x \leq m : (\langle s^x_0,\sigma_x \rangle, k) \models \mathbf{~total~production~of~} f \mathbf{~is~} n  \\
\vspace{20pt}\nonumber\\
&\{ \langle s^1_0,\sigma_1 \rangle,\dots, \langle s^m_0,\sigma_m \rangle \} \models \mathbf{~rate~of~production~of~} f \mathbf{~is~} n \nonumber\\
&\hspace{30pt}\mathbf{~when~observed~between~time~step~} i \mathbf{~and~time~step~} j \nonumber\\
&\hspace{30pt}\mathbf{~in~all~trajectories~} \nonumber\\
&~~~~\text{ if } \forall x, 1 \leq x \leq m : (\langle s^x_i,\sigma_x \rangle, j) \models \mathbf{~rate~of~production~of~} f \mathbf{~is~} n  \\
&\{ \langle s^1_0,\sigma_1 \rangle,\dots, \langle s^m_0,\sigma_m \rangle \} \models \mathbf{~rate~of~firing~of~} a \mathbf{~is~} n \nonumber\\
&\hspace{30pt}\mathbf{~when~observed~between~time~step~} i \mathbf{~and~time~step~} j \nonumber\\
&\hspace{30pt}\mathbf{~in~all~trajectories~} \nonumber\\
&~~~~\text{ if } \forall x, 1 \leq x \leq m : (\langle s^x_i,\sigma_x \rangle, j) \models \mathbf{~rate~of~firing~of~} a \mathbf{~is~} n  \\
&\{ \langle s^1_0,\sigma_1 \rangle,\dots, \langle s^m_0,\sigma_m \rangle \} \models \mathbf{~total~production~of~} f \mathbf{~is~} n \nonumber\\
&\hspace{30pt}\mathbf{~when~observed~between~time~step~} i \mathbf{~and~time~step~} j \nonumber\\
&\hspace{30pt}\mathbf{~in~all~trajectories~} \nonumber\\
&~~~~\text{ if } \forall x, 1 \leq x \leq m : (\langle s^x_i,\sigma_x \rangle, j) \models \mathbf{~total~production~of~} f \mathbf{~is~} n \\
\vspace{20pt}\nonumber\\
&\{ \langle s^1_0,\sigma_1 \rangle,\dots, \langle s^m_0,\sigma_m \rangle \} \models f \mathbf{~is~accumulating~} \nonumber\\
&\hspace{30pt}\mathbf{~in~all~trajectories~} \nonumber\\
&~~~~\text{ if } \forall x, 1 \leq x \leq m : (\langle s^x_0,\sigma_x \rangle, k) \models f \mathbf{~is~accumulating~}  \\
&\{ \langle s^1_0,\sigma_1 \rangle,\dots, \langle s^m_0,\sigma_m \rangle \} \models f \mathbf{~is~decreasing~} \nonumber\\
&\hspace{30pt}\mathbf{~in~all~trajectories~} \nonumber\\
&~~~~\text{ if } \forall x, 1 \leq x \leq m : (\langle s^x_0,\sigma_x \rangle, k) \models f \mathbf{~is~decreasing~}  \\
\vspace{20pt}\nonumber\\
&\{ \langle s^1_0,\sigma_1 \rangle,\dots, \langle s^m_0,\sigma_m \rangle \} \models f \mathbf{~is~accumulating~} \nonumber\\
&\hspace{30pt}\mathbf{~when~observed~between~time~step~} i \mathbf{~and~time~step~} j \nonumber\\
&\hspace{30pt}\mathbf{~in~all~trajectories~} \nonumber\\
&~~~~\text{ if } \forall x, 1 \leq x \leq m : (\langle s^x_i,\sigma_x \rangle, j) \models f \mathbf{~is~accumulating~}  \\
&\{ \langle s^1_0,\sigma_1 \rangle,\dots, \langle s^m_0,\sigma_m \rangle \} \models f \mathbf{~is~decreasing~} \nonumber\\
&\hspace{30pt}\mathbf{~when~observed~between~time~step~} i \mathbf{~and~time~step~} j \nonumber\\
&\hspace{30pt}\mathbf{~in~all~trajectories~} \nonumber\\
&~~~~\text{ if } \forall x, 1 \leq x \leq m : (\langle s^x_i,\sigma_x \rangle, j) \models f \mathbf{~is~decreasing~} 
\end{align}
}

A set of trajectories $\sigma_1,\dots,\sigma_m$, each of length $k$ satisfies a quantitative or a qualitative point query description based on formula satisfaction of section~\ref{dqa:sem:formula} as follows:
{\footnotesize
\begin{align}
&\{ \langle s^1_0,\sigma_1 \rangle,\dots, \langle s^m_0,\sigma_m \rangle \} \models \mathbf{~value~of~} f \mathbf{~is~higher~than~} n \nonumber\\
&~~~~\text{ if } \exists x \exists i, 1 \leq x \leq m, 0 \leq i \leq k : \langle s^x_i,\sigma_x \rangle \models \mathbf{~value~of~} f \mathbf{~is~higher~than~} n \\
&\{ \langle s^1_0,\sigma_1 \rangle,\dots, \langle s^m_0,\sigma_m \rangle \} \models \mathbf{~value~of~} f \mathbf{~is~lower~than~} n \nonumber\\
&~~~~\text{ if } \exists x \exists i, 1 \leq x \leq m, 0 \leq i \leq k : \langle s^x_i,\sigma_x \rangle \models \mathbf{~value~of~} f \mathbf{~is~lower~than~} n \\
&\{ \langle s^1_0,\sigma_1 \rangle,\dots, \langle s^m_0,\sigma_m \rangle \} \models \mathbf{~value~of~} f \mathbf{~is~} n \nonumber\\
&~~~~\text{ if } \exists x \exists i, 1 \leq x \leq m, 0 \leq i \leq k : \langle s^x_i,\sigma_x \rangle \models \mathbf{~value~of~} f \mathbf{~is~} n \\
\vspace{20pt}\nonumber\\
&\{ \langle s^1_0,\sigma_1 \rangle,\dots, \langle s^m_0,\sigma_m \rangle \} \models \mathbf{~value~of~} f \mathbf{~is~higher~than~} n \nonumber\\
&\hspace{30pt}\mathbf{~at~time~step~} i \nonumber\\
&~~~~\text{ if } \exists x, 1 \leq x \leq m : \langle s^x_i,\sigma_x \rangle \models \mathbf{~value~of~} f \mathbf{~is~higher~than~} n \\
&\{ \langle s^1_0,\sigma_1 \rangle,\dots, \langle s^m_0,\sigma_m \rangle \} \models \mathbf{~value~of~} f \mathbf{~is~lower~than~} n \nonumber\\
&\hspace{30pt}\mathbf{~at~time~step~} i \nonumber\\
&~~~~\text{ if } \exists x, 1 \leq x \leq m : \langle s^x_i,\sigma_x \rangle \models \mathbf{~value~of~} f \mathbf{~is~lower~than~} n \\
&\{ \langle s^1_0,\sigma_1 \rangle,\dots, \langle s^m_0,\sigma_m \rangle \} \models \mathbf{~value~of~} f \mathbf{~is~} n \nonumber\\
&\hspace{30pt}\mathbf{~at~time~step~} i \nonumber\\
&~~~~\text{ if } \exists x, 1 \leq x \leq m : \langle s^x_i,\sigma_x \rangle \models \mathbf{~value~of~} f \mathbf{~is~} n \\
\vspace{20pt}\nonumber\\
&\{ \langle s^1_0,\sigma_1 \rangle,\dots, \langle s^m_0,\sigma_m \rangle \} \models \mathbf{~value~of~} f \mathbf{~is~higher~than~} n \nonumber\\
&\hspace{30pt}\mathbf{~in~all~trajectories~} \nonumber\\
&~~~~\text{ if } \forall x \exists i, 1 \leq x \leq m, 0 \leq i \leq k : \langle s^x_i,\sigma_x \rangle \models \mathbf{~value~of~} f \mathbf{~is~higher~than~} n \\
&\{ \langle s^1_0,\sigma_1 \rangle,\dots, \langle s^m_0,\sigma_m \rangle \} \models \mathbf{~value~of~} f \mathbf{~is~lower~than~} n \nonumber\\
&\hspace{30pt}\mathbf{~in~all~trajectories~} \nonumber\\
&~~~~\text{ if } \forall x \exists i, 1 \leq x \leq m, 0 \leq i \leq k : \langle s^x_i,\sigma_x \rangle \models \mathbf{~value~of~} f \mathbf{~is~lower~than~} n \\
&\{ \langle s^1_0,\sigma_1 \rangle,\dots, \langle s^m_0,\sigma_m \rangle \} \models \mathbf{~value~of~} f \mathbf{~is~} n \nonumber\\
&\hspace{30pt}\mathbf{~in~all~trajectories~} \nonumber\\
&~~~~\text{ if } \forall x \exists i, 1 \leq x \leq m, 0 \leq i \leq k : \langle s^x_i,\sigma_x \rangle \models \mathbf{~value~of~} f \mathbf{~is~} n \\
\vspace{20pt}\nonumber\\
&\{ \langle s^1_0,\sigma_1 \rangle,\dots, \langle s^m_0,\sigma_m \rangle \} \models \mathbf{~value~of~} f \mathbf{~is~higher~than~} n \nonumber\\
&\hspace{30pt}\mathbf{~at~time~step~} i \nonumber\\
&\hspace{30pt}\mathbf{~in~all~trajectories~} \nonumber\\
&~~~~\text{ if } \forall x, 1 \leq x \leq m : \langle s^x_i,\sigma_x \rangle \models \mathbf{~value~of~} f \mathbf{~is~higher~than~} n \\
&\{ \langle s^1_0,\sigma_1 \rangle,\dots, \langle s^m_0,\sigma_m \rangle \} \models \mathbf{~value~of~} f \mathbf{~is~lower~than~} n \nonumber\\
&\hspace{30pt}\mathbf{~at~time~step~} i \nonumber\\
&\hspace{30pt}\mathbf{~in~all~trajectories~} \nonumber\\
&~~~~\text{ if } \forall x, 1 \leq x \leq m : \langle s^x_i,\sigma_x \rangle \models \mathbf{~value~of~} f \mathbf{~is~lower~than~} n \\
&\{ \langle s^1_0,\sigma_1 \rangle,\dots, \langle s^m_0,\sigma_m \rangle \} \models \mathbf{~value~of~} f \mathbf{~is~} n \nonumber\\
&\hspace{30pt}\mathbf{~at~time~step~} i \nonumber\\
&\hspace{30pt}\mathbf{~in~all~trajectories~} \nonumber\\
&~~~~\text{ if } \forall x, 1 \leq x \leq m : \langle s^x_i,\sigma_x \rangle \models \mathbf{~value~of~} f \mathbf{~is~} n
\end{align}
}

\vspace{30pt}
Now, we turn our attention to domain descriptions with locational fluents. Let $\mathbf{D}$ be a domain description and $\sigma = s_0,\dots,s_k$ be its trajectory of length $k$ as defined in \eqref{def:gcpn:traj}. Let $\sigma_1,\dots,\sigma_m$ represent the set of trajectories of $\mathbf{D}$ filtered by observations as necessary, with each trajectory has the form $\sigma_i = s^i_0,\dots,s^i_k$ $1\leq i \leq m$. Let $\mathbf{\bar{D}}$ be a modified domain description and $\bar{\sigma}_1,\dots,\bar{\sigma}_{\bar{m}}$ be its trajectories of the form $\bar{\sigma}_i = \bar{s}^i_0,\dots,\bar{s}^i_k$. Then we define query satisfaction using the formula satisfaction in section~\ref{dqa:sem:formula} as follows.

Two sets of trajectories $\sigma_1,\dots,\sigma_m$ and $\bar{\sigma}_1,\dots,\bar{\sigma}_{\bar{m}}$ satisfy a comparative query description based on formula satisfaction of section~\ref{dqa:sem:formula} as follows:
{\footnotesize
\begin{align}
&\Big\{ \{ \langle s^1_0,\sigma_1 \rangle,\dots, \langle s^m_0,\sigma_m \rangle \}, \{ \langle \bar{s}^1_0,\bar{\sigma}_1 \rangle,\dots, \langle \bar{s}^{\bar{m}}_0,\bar{\sigma}_{\bar{m}} \rangle \} \Big\} \nonumber\\
&\;\;\;\;\;\;\models \; \mathbf{~direction~of~change~in~} \langle \text{aggop} \rangle \mathbf{~rate~of~production~of~} f \mathbf{~atloc~} l \mathbf{~is~} d \nonumber\\
&\;\;\;\;\;\;\;\;\;\;\;\;\;\;\;\;\;\mathbf{~when~observed~between~time~step~} i \mathbf{~and~time~step~} j \nonumber\\
&\;\;\;\;\;\;\text{ if } \Big(\Big\{ \{ \langle s^1_i,\sigma_1 \rangle,\dots, \langle s^m_i,\sigma_m \rangle \}, \{ \langle \bar{s}^1_i,\bar{\sigma}_1 \rangle,\dots, \langle \bar{s}^{\bar{m}}_i,\bar{\sigma}_{\bar{m}} \rangle \} \Big\}, j \Big) \nonumber\\
&\;\;\;\;\;\;\;\;\;\;\;\;\models \mathbf{~direction~of~change~in~} \langle \text{aggop} \rangle \mathbf{~rate~of~production~of~} f \mathbf{~atloc~} l \mathbf{~is~} d \\
&\Big\{ \{ \langle s^1_0,\sigma_1 \rangle,\dots, \langle s^m_0,\sigma_m \rangle \}, \{ \langle \bar{s}^1_0,\bar{\sigma}_1 \rangle,\dots, \langle \bar{s}^{\bar{m}}_0,\bar{\sigma}_{\bar{m}} \rangle \} \Big\} \nonumber\\
&\;\;\;\;\;\;\models \; \mathbf{~direction~of~change~in~} \langle \text{aggop} \rangle \mathbf{~rate~of~firing~of~} f \mathbf{~is~} d \nonumber\\
&\;\;\;\;\;\;\;\;\;\;\;\;\;\;\;\;\;\mathbf{~when~observed~between~time~step~} i \mathbf{~and~time~step~} j \nonumber\\
&\;\;\;\;\;\;\text{ if } \Big(\Big\{ \{ \langle s^1_i,\sigma_1 \rangle,\dots, \langle s^m_i,\sigma_m \rangle \}, \{ \langle \bar{s}^1_i,\bar{\sigma}_1 \rangle,\dots, \langle \bar{s}^{\bar{m}}_i,\bar{\sigma}_{\bar{m}} \rangle \} \Big\}, j\Big) \nonumber\\
&\;\;\;\;\;\;\;\;\;\;\;\;\models \mathbf{~direction~of~change~in~} \langle \text{aggop} \rangle \mathbf{~rate~of~firing~of~} f \mathbf{~is~} d \\
&\Big\{ \{ \langle s^1_0,\sigma_1 \rangle,\dots, \langle s^m_0,\sigma_m \rangle \}, \{ \langle \bar{s}^1_0,\bar{\sigma}_1 \rangle,\dots, \langle \bar{s}^{\bar{m}}_0,\bar{\sigma}_{\bar{m}} \rangle \} \Big\} \nonumber\\
&\;\;\;\;\;\;\models \; \mathbf{~direction~of~change~in~} \langle \text{aggop} \rangle \mathbf{~total~production~of~} f \mathbf{~atloc~} l \mathbf{~is~} d \nonumber\\
&\;\;\;\;\;\;\;\;\;\;\;\;\;\;\;\;\;\mathbf{~when~observed~between~time~step~} i \mathbf{~and~time~step~} j \nonumber\\
&\;\;\;\;\;\;\text{ if } \Big(\Big\{ \{ \langle s^1_i,\sigma_1 \rangle,\dots, \langle s^m_i,\sigma_m \rangle \}, \{ \langle \bar{s}^1_i,\bar{\sigma}_1 \rangle,\dots, \langle \bar{s}^{\bar{m}}_i,\bar{\sigma}_{\bar{m}} \rangle \} \Big\}, j\Big) \nonumber\\
&\;\;\;\;\;\;\;\;\;\;\;\;\models \mathbf{~direction~of~change~in~} \langle \text{aggop} \rangle \mathbf{~total~production~of~} f \mathbf{~atloc~} l \mathbf{~is~} d \\
&\Big\{ \{ \langle s^1_0,\sigma_1 \rangle,\dots, \langle s^m_0,\sigma_m \rangle \}, \{ \langle \bar{s}^1_0,\bar{\sigma}_1 \rangle,\dots, \langle \bar{s}^{\bar{m}}_0,\bar{\sigma}_{\bar{m}} \rangle \} \Big\} \nonumber\\
&\;\;\;\;\;\;\models \; \mathbf{~direction~of~change~in~} \langle \text{aggop} \rangle \mathbf{~value~of~} f \mathbf{~atloc~} l \mathbf{~is~} d \nonumber\\
&\;\;\;\;\;\;\;\;\;\;\;\;\;\;\;\;\;\mathbf{~when~observed~at~time~step~} i \nonumber\\
&\;\;\;\;\;\;\text{ if } \Big\{ \{ \langle s^1_i,\sigma_1 \rangle,\dots, \langle s^m_i,\sigma_m \rangle \}, \{ \langle \bar{s}^1_i,\bar{\sigma}_1 \rangle,\dots, \langle \bar{s}^{\bar{m}}_i,\bar{\sigma}_{\bar{m}} \rangle \} \Big\} \nonumber\\
&\;\;\;\;\;\;\;\;\;\;\;\;\models \mathbf{~direction~of~change~in~} \langle \text{aggop} \rangle \mathbf{~value~of~} f \mathbf{~atloc~} l \mathbf{~is~} d 
\end{align}
}

A set of trajectories $\sigma_1,\dots,\sigma_m$, each of length $k$ satisfies a quantitative or a qualitative interval query description based on formula satisfaction of section~\ref{dqa:sem:formula} as follows:
{\footnotesize
\begin{align}
&\{ \langle s^1_0,\sigma_1 \rangle,\dots, \langle s^m_0,\sigma_m \rangle \} \models \mathbf{~rate~of~production~of~} f \mathbf{~atloc~} l \mathbf{~is~} n \nonumber\\
&~~~~\text{ if } \exists x, 1 \leq x \leq m : (\langle s^x_0,\sigma_x \rangle, k) \models \mathbf{~rate~of~production~of~} f \mathbf{~atloc~} l \mathbf{~is~} n  \\
&\{ \langle s^1_0,\sigma_1 \rangle,\dots, \langle s^m_0,\sigma_m \rangle \} \models \mathbf{~rate~of~firing~of~} a \mathbf{~is~} n \nonumber\\
&~~~~\text{ if } \exists x, 1 \leq x \leq m : (\langle s^x_0,\sigma_x \rangle, k) \models \mathbf{~rate~of~firing~of~} a \mathbf{~is~} n  \\
&\{ \langle s^1_0,\sigma_1 \rangle,\dots, \langle s^m_0,\sigma_m \rangle \} \models \mathbf{~total~production~of~} f \mathbf{~atloc~} l \mathbf{~is~} n \nonumber\\
&~~~~\text{ if } \exists x, 1 \leq x \leq m : (\langle s^x_0,\sigma_x \rangle, k) \models \mathbf{~total~production~of~} f \mathbf{~atloc~} l \mathbf{~is~} n  \\
\vspace{20pt}\nonumber\\
&\{ \langle s^1_0,\sigma_1 \rangle,\dots, \langle s^m_0,\sigma_m \rangle \} \models \mathbf{~rate~of~production~of~} f \mathbf{~atloc~} l \mathbf{~is~} n \nonumber\\
&\hspace{30pt}\mathbf{~when~observed~between~time~step~} i \mathbf{~and~time~step~} j \nonumber\\
&~~~~\text{ if } \exists x, 1 \leq x \leq m : (\langle s^x_i,\sigma_x \rangle, j) \models \mathbf{~rate~of~production~of~} f \mathbf{~atloc~} l \mathbf{~is~} n  \\
&\{ \langle s^1_0,\sigma_1 \rangle,\dots, \langle s^m_0,\sigma_m \rangle \} \models \mathbf{~rate~of~firing~of~} a \mathbf{~is~} n \nonumber\\
&\hspace{30pt}\mathbf{~when~observed~between~time~step~} i \mathbf{~and~time~step~} j \nonumber\\
&~~~~\text{ if } \exists x, 1 \leq x \leq m : (\langle s^x_i,\sigma_x \rangle, j) \models \mathbf{~rate~of~firing~of~} a \mathbf{~is~} n  \\
&\{ \langle s^1_0,\sigma_1 \rangle,\dots, \langle s^m_0,\sigma_m \rangle \} \models \mathbf{~total~production~of~} f \mathbf{~atloc~} l \mathbf{~is~} n \nonumber\\
&\hspace{30pt}\mathbf{~when~observed~between~time~step~} i \mathbf{~and~time~step~} j \nonumber\\
&~~~~\text{ if } \exists x, 1 \leq x \leq m : (\langle s^x_i,\sigma_x \rangle, j) \models \mathbf{~total~production~of~} f \mathbf{~atloc~} l \mathbf{~is~} n \\
\vspace{20pt}\nonumber\\
&\{ \langle s^1_0,\sigma_1 \rangle,\dots, \langle s^m_0,\sigma_m \rangle \} \models f \mathbf{~is~accumulating~} \mathbf{~atloc~} l \nonumber\\
&~~~~\text{ if } \exists x, 1 \leq x \leq m : (\langle s^x_0,\sigma_x \rangle, k) \models f \mathbf{~is~accumulating~} \mathbf{~atloc~} l  \\
&\{ \langle s^1_0,\sigma_1 \rangle,\dots, \langle s^m_0,\sigma_m \rangle \} \models f \mathbf{~is~decreasing~} \mathbf{~atloc~} l \nonumber\\
&~~~~\text{ if } \exists x, 1 \leq x \leq m : (\langle s^x_0,\sigma_x \rangle, k) \models f \mathbf{~is~decreasing~} \mathbf{~atloc~} l  \\
\vspace{20pt}\nonumber\\
&\{ \langle s^1_0,\sigma_1 \rangle,\dots, \langle s^m_0,\sigma_m \rangle \} \models f \mathbf{~is~accumulating~} \mathbf{~atloc~} l  \nonumber\\
&\hspace{30pt}\mathbf{~when~observed~between~time~step~} i \mathbf{~and~time~step~} j \nonumber\\
&~~~~\text{ if } \exists x, 1 \leq x \leq m : (\langle s^x_i,\sigma_x \rangle, j) \models f \mathbf{~is~accumulating~} \mathbf{~atloc~} l  \\
&\{ \langle s^1_0,\sigma_1 \rangle,\dots, \langle s^m_0,\sigma_m \rangle \} \models f \mathbf{~is~decreasing~} \mathbf{~atloc~} l \nonumber\\
&\hspace{30pt}\mathbf{~when~observed~between~time~step~} i \mathbf{~and~time~step~} j \nonumber\\
&~~~~\text{ if } \exists x, 1 \leq x \leq m : (\langle s^x_i,\sigma_x \rangle, j) \models f \mathbf{~is~decreasing~} \mathbf{~atloc~} l \\
\vspace{30pt} \nonumber\\
&\{ \langle s^1_0,\sigma_1 \rangle,\dots, \langle s^m_0,\sigma_m \rangle \} \models \mathbf{~rate~of~production~of~} f \mathbf{~atloc~} l \mathbf{~is~} n \nonumber\\
&\hspace{30pt}\mathbf{~in~all~trajectories~} \nonumber\\
&~~~~\text{ if } \forall x, 1 \leq x \leq m : (\langle s^x_0,\sigma_x \rangle, k) \models \mathbf{~rate~of~production~of~} f \mathbf{~atloc~} l \mathbf{~is~} n  \\
&\{ \langle s^1_0,\sigma_1 \rangle,\dots, \langle s^m_0,\sigma_m \rangle \} \models \mathbf{~rate~of~firing~of~} a \mathbf{~is~} n \nonumber\\
&\hspace{30pt}\mathbf{~in~all~trajectories~} \nonumber\\
&~~~~\text{ if } \forall x, 1 \leq x \leq m : (\langle s^x_0,\sigma_x \rangle, k) \models \mathbf{~rate~of~firing~of~} a \mathbf{~is~} n  \\
&\{ \langle s^1_0,\sigma_1 \rangle,\dots, \langle s^m_0,\sigma_m \rangle \} \models \mathbf{~total~production~of~} f \mathbf{~atloc~} l \mathbf{~is~} n \nonumber\\
&\hspace{30pt}\mathbf{~in~all~trajectories~} \nonumber\\
&~~~~\text{ if } \forall x, 1 \leq x \leq m : (\langle s^x_0,\sigma_x \rangle, k) \models \mathbf{~total~production~of~} f \mathbf{~atloc~} l \mathbf{~is~} n \\
\vspace{20pt}\nonumber\\
&\{ \langle s^1_0,\sigma_1 \rangle,\dots, \langle s^m_0,\sigma_m \rangle \} \models \mathbf{~rate~of~production~of~} f \mathbf{~atloc~} l \mathbf{~is~} n \nonumber\\
&\hspace{30pt}\mathbf{~when~observed~between~time~step~} i \mathbf{~and~time~step~} j \nonumber\\
&\hspace{30pt}\mathbf{~in~all~trajectories~} \nonumber\\
&~~~~\text{ if } \forall x, 1 \leq x \leq m : (\langle s^x_i,\sigma_x \rangle, j) \models \mathbf{~rate~of~production~of~} f \mathbf{~atloc~} l \mathbf{~is~} n  \\
&\{ \langle s^1_0,\sigma_1 \rangle,\dots, \langle s^m_0,\sigma_m \rangle \} \models \mathbf{~rate~of~firing~of~} a \mathbf{~is~} n \nonumber\\
&\hspace{30pt}\mathbf{~when~observed~between~time~step~} i \mathbf{~and~time~step~} j \nonumber\\
&\hspace{30pt}\mathbf{~in~all~trajectories~} \nonumber\\
&~~~~\text{ if } \forall x, 1 \leq x \leq m : (\langle s^x_i,\sigma_x \rangle, j) \models \mathbf{~rate~of~firing~of~} a \mathbf{~is~} n  \\
&\{ \langle s^1_0,\sigma_1 \rangle,\dots, \langle s^m_0,\sigma_m \rangle \} \models \mathbf{~total~production~of~} f \mathbf{~atloc~} l \mathbf{~is~} n \nonumber\\
&\hspace{30pt}\mathbf{~when~observed~between~time~step~} i \mathbf{~and~time~step~} j \nonumber\\
&\hspace{30pt}\mathbf{~in~all~trajectories~} \nonumber\\
&~~~~\text{ if } \forall x, 1 \leq x \leq m : (\langle s^x_i,\sigma_x \rangle, j) \models \mathbf{~total~production~of~} f \mathbf{~atloc~} l \mathbf{~is~} n \text{ till } j\\
\vspace{20pt}\nonumber\\
&\{ \langle s^1_0,\sigma_1 \rangle,\dots, \langle s^m_0,\sigma_m \rangle \} \models f \mathbf{~is~accumulating~} \mathbf{~atloc~} l \nonumber\\
&\hspace{30pt}\mathbf{~in~all~trajectories~} \nonumber\\
&~~~~\text{ if } \forall x, 1 \leq x \leq m : (\langle s^x_0,\sigma_x \rangle, k) \models f \mathbf{~is~accumulating~} \mathbf{~atloc~} l  \\
&\{ \langle s^1_0,\sigma_1 \rangle,\dots, \langle s^m_0,\sigma_m \rangle \} \models f \mathbf{~is~decreasing~} \mathbf{~atloc~} l \nonumber\\
&\hspace{30pt}\mathbf{~in~all~trajectories~} \nonumber\\
&~~~~\text{ if } \forall x, 1 \leq x \leq m : (\langle s^x_0,\sigma_x \rangle, k) \models f \mathbf{~is~decreasing~} \mathbf{~atloc~} l  \\
\vspace{20pt}\nonumber\\
&\{ \langle s^1_0,\sigma_1 \rangle,\dots, \langle s^m_0,\sigma_m \rangle \} \models f \mathbf{~is~accumulating~} \mathbf{~atloc~} l \nonumber\\
&\hspace{30pt}\mathbf{~when~observed~between~time~step~} i \mathbf{~and~time~step~} j \nonumber\\
&\hspace{30pt}\mathbf{~in~all~trajectories~} \nonumber\\
&~~~~\text{ if } \forall x, 1 \leq x \leq m : (\langle s^x_i,\sigma_x \rangle, j) \models f \mathbf{~is~accumulating~} \mathbf{~atloc~} l  \\
&\{ \langle s^1_0,\sigma_1 \rangle,\dots, \langle s^m_0,\sigma_m \rangle \} \models f \mathbf{~is~decreasing~} \mathbf{~atloc~} l \nonumber\\
&\hspace{30pt}\mathbf{~when~observed~between~time~step~} i \mathbf{~and~time~step~} j \nonumber\\
&\hspace{30pt}\mathbf{~in~all~trajectories~} \nonumber\\
&~~~~\text{ if } \forall x, 1 \leq x \leq m : (\langle s^x_i,\sigma_x \rangle, j) \models f \mathbf{~is~decreasing~} \mathbf{~atloc~} l  
\end{align}
}

A set of trajectories $\sigma_1,\dots,\sigma_m$, each of length $k$ satisfies a quantitative or a qualitative point query description based on formula satisfaction of section~\ref{dqa:sem:formula} as follows:
{\footnotesize
\begin{align}
&\{ \langle s^1_0,\sigma_1 \rangle,\dots, \langle s^m_0,\sigma_m \rangle \} \models \mathbf{~value~of~} f \mathbf{~atloc~} l\mathbf{~is~higher~than~} n \nonumber\\
&~~~~\text{ if } \exists x \exists i, 1 \leq x \leq m, 0 \leq i \leq k : \langle s^x_i,\sigma_x \rangle \models \mathbf{~value~of~} f \mathbf{~atloc~} l \mathbf{~is~higher~than~} n \\
&\{ \langle s^1_0,\sigma_1 \rangle,\dots, \langle s^m_0,\sigma_m \rangle \} \models \mathbf{~value~of~} f \mathbf{~atloc~} l\mathbf{~is~lower~than~} n \nonumber\\
&~~~~\text{ if } \exists x \exists i, 1 \leq x \leq m, 0 \leq i \leq k : \langle s^x_i,\sigma_x \rangle \models \mathbf{~value~of~} f \mathbf{~atloc~} l \mathbf{~is~lower~than~} n \\
&\{ \langle s^1_0,\sigma_1 \rangle,\dots, \langle s^m_0,\sigma_m \rangle \} \models \mathbf{~value~of~} f \mathbf{~atloc~} l\mathbf{~is~} n \nonumber\\
&~~~~\text{ if } \exists x \exists i, 1 \leq x \leq m, 0 \leq i \leq k : \langle s^x_i,\sigma_x \rangle \models \mathbf{~value~of~} f \mathbf{~atloc~} l \mathbf{~is~} n\\
\vspace{20pt}\nonumber\\
&\{ \langle s^1_0,\sigma_1 \rangle,\dots, \langle s^m_0,\sigma_m \rangle \} \models \mathbf{~value~of~} f \mathbf{~atloc~} l\mathbf{~is~higher~than~} n \nonumber\\
&\hspace{30pt}\mathbf{~at~time~step~} i \nonumber\\
&~~~~\text{ if } \exists x, 1 \leq x \leq m : \langle s^x_i,\sigma_x \rangle \models \mathbf{~value~of~} f \mathbf{~atloc~} l \mathbf{~is~higher~than~} n \\
&\{ \langle s^1_0,\sigma_1 \rangle,\dots, \langle s^m_0,\sigma_m \rangle \} \models \mathbf{~value~of~} f \mathbf{~atloc~} l\mathbf{~is~lower~than~} n \nonumber\\
&\hspace{30pt}\mathbf{~at~time~step~} i \nonumber\\
&~~~~\text{ if } \exists x, 1 \leq x \leq m : \langle s^x_i,\sigma_x \rangle \models \mathbf{~value~of~} f \mathbf{~atloc~} l \mathbf{~is~lower~than~} n \\
&\{ \langle s^1_0,\sigma_1 \rangle,\dots, \langle s^m_0,\sigma_m \rangle \} \models \mathbf{~value~of~} f \mathbf{~atloc~} l\mathbf{~is~} n \nonumber\\
&\hspace{30pt}\mathbf{~at~time~step~} i \nonumber\\
&~~~~\text{ if } \exists x, 1 \leq x \leq m : \langle s^x_i,\sigma_x \rangle \models \mathbf{~value~of~} f \mathbf{~atloc~} l \mathbf{~is~} n\\
\vspace{20pt}\nonumber\\
&\{ \langle s^1_0,\sigma_1 \rangle,\dots, \langle s^m_0,\sigma_m \rangle \} \models \mathbf{~value~of~} f \mathbf{~atloc~} l\mathbf{~is~higher~than~} n \nonumber\\
&\hspace{30pt}\mathbf{~in~all~trajectories~} \nonumber\\
&~~~~\text{ if } \forall x \exists i, 1 \leq x \leq m, 0 \leq i \leq k : \langle s^x_i,\sigma_x \rangle \models \mathbf{~value~of~} f \mathbf{~atloc~} l \mathbf{~is~higher~than~} n \\
&\{ \langle s^1_0,\sigma_1 \rangle,\dots, \langle s^m_0,\sigma_m \rangle \} \models \mathbf{~value~of~} f \mathbf{~atloc~} l \mathbf{~is~lower~than~} n \nonumber\\
&\hspace{30pt}\mathbf{~in~all~trajectories~} \nonumber\\
&~~~~\text{ if } \forall x \exists i, 1 \leq x \leq m, 0 \leq i \leq k : \langle s^x_i,\sigma_x \rangle \models \mathbf{~value~of~} f \mathbf{~atloc~} l \mathbf{~is~lower~than~} n \\
&\{ \langle s^1_0,\sigma_1 \rangle,\dots, \langle s^m_0,\sigma_m \rangle \} \models \mathbf{~value~of~} f \mathbf{~atloc~} l\mathbf{~is~} n \nonumber\\
&\hspace{30pt}\mathbf{~in~all~trajectories~} \nonumber\\
&~~~~\text{ if } \forall x \exists i, 1 \leq x \leq m, 0 \leq i \leq k : \langle s^x_i,\sigma_x \rangle \models \mathbf{~value~of~} f \mathbf{~atloc~} l \mathbf{~is~} n\\
\vspace{20pt}\nonumber\\
&\{ \langle s^1_0,\sigma_1 \rangle,\dots, \langle s^m_0,\sigma_m \rangle \} \models \mathbf{~value~of~} f \mathbf{~atloc~} l \mathbf{~is~higher~than~} n \nonumber\\
&\hspace{30pt}\mathbf{~at~time~step~} i \nonumber\\
&\hspace{30pt}\mathbf{~in~all~trajectories~} \nonumber\\
&~~~~\text{ if } \forall x, 1 \leq x \leq m : \langle s^x_i,\sigma_x \rangle \models \mathbf{~value~of~} f \mathbf{~atloc~} l \mathbf{~is~higher~than~} n \\
&\{ \langle s^1_0,\sigma_1 \rangle,\dots, \langle s^m_0,\sigma_m \rangle \} \models \mathbf{~value~of~} f \mathbf{~atloc~} l\mathbf{~is~lower~than~} n \nonumber\\
&\hspace{30pt}\mathbf{~at~time~step~} i \nonumber\\
&\hspace{30pt}\mathbf{~in~all~trajectories~} \nonumber\\
&~~~~\text{ if } \forall x, 1 \leq x \leq m : \langle s^x_i,\sigma_x \rangle \models \mathbf{~value~of~} f \mathbf{~atloc~} l \mathbf{~is~lower~than~} n \\
&\{ \langle s^1_0,\sigma_1 \rangle,\dots, \langle s^m_0,\sigma_m \rangle \} \models \mathbf{~value~of~} f \mathbf{~atloc~} l\mathbf{~is~} n \nonumber\\
&\hspace{30pt}\mathbf{~at~time~step~} i \nonumber\\
&\hspace{30pt}\mathbf{~in~all~trajectories~} \nonumber\\
&~~~~\text{ if } \forall x, 1 \leq x \leq m : \langle s^x_i,\sigma_x \rangle \models \mathbf{~value~of~} f \mathbf{~atloc~} l \mathbf{~is~} n
\end{align}
}

\vspace{30pt}
Next, we generically define the satisfaction of a {\em simple point formula cascade} query w.r.t. a set of trajectories $\sigma_1,\dots,\sigma_m$. The trajectories will either be as defined in definitions~\eqref{def:gpn:traj} or \eqref{def:gcpn:traj} for {\em simple point formula cascade} query statement made up of simple fluents or locational fluents, respectively.
A set of trajectories $\sigma_1,\dots,\sigma_m$, each of length $k$ satisfies a {\em simple point formula cascade} based on formula satisfaction of section~\ref{dqa:sem:formula} as follows:
{\footnotesize
\begin{align}
&\{ \langle s^1_0,\sigma_1 \rangle,\dots, \langle s^m_0,\sigma_m \rangle \} \models \langle \text{simple point formula} \rangle_0 \nonumber\\
&~~~~~~~~\mathbf{~after~} \langle \text{simple point formula} \rangle_{1,1}, \dots, \langle \text{simple point formula} \rangle_{1,n_1} \nonumber\\
&~~~~~~~~\vdots \nonumber\\
&~~~~~~~~\mathbf{~after~} \langle \text{simple point formula} \rangle_{u,1}, \dots, \langle \text{simple point formula} \rangle_{u,n_u} \nonumber\\
&~~~~\text{ if } \exists x \exists i_0 \exists i_1 \dots \exists i_u, 1 \leq x \leq m, i_1 < i_0 \leq k, \dots, i_u < i_{u-1} \leq k, 0 \leq i_u \leq k: \nonumber\\
&~~~~~~~~\langle s^x_{i_0},\sigma_x \rangle \models \langle \text{simple point formula} \rangle_0 \text{ and } \nonumber\\
&~~~~~~~~\langle s^x_{i_1},\sigma_x \rangle \models \langle \text{simple point formula} \rangle_{1,1} \text{ and } \nonumber \hdots
\text{ and } \langle s^x_{i_1},\sigma_x \rangle \models \langle \text{simple point formula} \rangle_{1,n_1} \text{ and } \nonumber\\
&~~~~~~~~~~~~\vdots \nonumber\\
&~~~~~~~~\langle s^x_{i_u},\sigma_x \rangle \models \langle \text{simple point formula} \rangle_{u,1} \text{ and } \nonumber \hdots
\text{ and } \langle s^x_{i_u},\sigma_x \rangle \models \langle \text{simple point formula} \rangle_{u,n_u} \\
\\
\vspace{10pt}\nonumber\\
&\{ \langle s^1_0,\sigma_1 \rangle,\dots, \langle s^m_0,\sigma_m \rangle \} \models \langle \text{simple point formula} \rangle_0 \nonumber\\
&~~~~~~~~\mathbf{~after~} \langle \text{simple point formula} \rangle_{1,1}, \dots, \langle \text{simple point formula} \rangle_{1,n_1} \nonumber\\
&~~~~~~~~\vdots \nonumber\\
&~~~~~~~~\mathbf{~after~} \langle \text{simple point formula} \rangle_{u,1}, \dots, \langle \text{simple point formula} \rangle_{u,n_u} \nonumber\\
&~~~~~~~~\mathbf{~in~all~trajectories} \nonumber\\
&~~~~\text{ if } \forall x, 1 \leq x \leq m, \exists i_0 \exists i_1 \dots \exists i_u, i_1 < i_0 \leq k, \dots, i_u < i_{u-1} \leq k, 0 \leq i_u \leq k: \nonumber\\
&~~~~~~~~\langle s^x_{i_0},\sigma_x \rangle \models \langle \text{simple point formula} \rangle_0 \text{ and } \nonumber\\
&~~~~~~~~\langle s^x_{i_1},\sigma_x \rangle \models \langle \text{simple point formula} \rangle_{1,1} \text{ and } \nonumber \hdots
\text{ and } \langle s^x_{i_1},\sigma_x \rangle \models \langle \text{simple point formula} \rangle_{1,n_1} \text{ and } \nonumber\\
&~~~~~~~~~~~~\vdots \nonumber\\
&~~~~~~~~\langle s^x_{i_u},\sigma_x \rangle \models \langle \text{simple point formula} \rangle_{u,1} \text{ and } \nonumber \hdots
\text{ and } \langle s^x_{i_u},\sigma_x \rangle \models \langle \text{simple point formula} \rangle_{u,n_u} \\
\\
\vspace{10pt}\nonumber\\
&\{ \langle s^1_0,\sigma_1 \rangle,\dots, \langle s^m_0,\sigma_m \rangle \} \models \langle \text{simple point formula} \rangle_0 \nonumber\\
&~~~~~~~~\mathbf{~when~} \langle \text{simple point formula} \rangle_{1,1}, \dots, \langle \text{simple point formula} \rangle_{1,n_1} \nonumber\\
&~~~~\text{ if } \exists x \exists i, 1 \leq x \leq m, 0 \leq i \leq k: \nonumber\\
&~~~~~~~~\langle s^x_{i_0},\sigma_x \rangle \models \langle \text{simple point formula} \rangle_0 \text{ and } \nonumber\\
&~~~~~~~~\langle s^x_{i_1},\sigma_x \rangle \models \langle \text{simple point formula} \rangle_{1,1} \text{ and } \nonumber \hdots
\text{ and } \langle s^x_{i_1},\sigma_x \rangle \models \langle \text{simple point formula} \rangle_{1,n_1}  \\
\\
\vspace{10pt}\nonumber\\
&\{ \langle s^1_0,\sigma_1 \rangle,\dots, \langle s^m_0,\sigma_m \rangle \} \models \langle \text{simple point formula} \rangle_0 \nonumber\\
&~~~~~~~~\mathbf{~when~} \langle \text{simple point formula} \rangle_{1,1}, \dots, \langle \text{simple point formula} \rangle_{1,n_1} \nonumber\\
&~~~~~~~~\mathbf{~in~all~trajectories} \nonumber\\
&~~~~\text{ if } \forall x, 1 \leq x \leq m, \exists i, 0 \leq i \leq k: \nonumber\\
&~~~~~~~~\langle s^x_{i_0},\sigma_x \rangle \models \langle \text{simple point formula} \rangle_0 \text{ and } \nonumber\\
&~~~~~~~~\langle s^x_{i_1},\sigma_x \rangle \models \langle \text{simple point formula} \rangle_{1,1} \text{ and } \nonumber \hdots
\text{ and } \langle s^x_{i_1},\sigma_x \rangle \models \langle \text{simple point formula} \rangle_{1,n_1} \\
\end{align}
}

\vspace{30pt}
\subsection{Query Statement Satisfaction}
Let $\mathbf{D}$ as defined in section~\ref{plang:syntax} be a domain description and $\mathbf{Q}$ be a query statement~\eqref{dqa:syn:qstmt} as defined in section~\ref{qlang:syntax} with query description $U$, interventions $V_1,\dots,V_{|V|}$, internal observations $O_1,\dots,O_{|O|}$, and initial conditions $I_1,\dots,I_{|I|}$. Let $\mathbf{D_1} \equiv \mathbf{D} \diamond I_1 \diamond \dots \diamond I_{|I|} \diamond V_1 \diamond \dots \diamond V_{|V|}$ be the modified domain description constructed by applying the initial conditions and interventions from $\mathbf{Q}$ as defined in section~\ref{sec:semantics:idesc}. Let $\sigma_1,\dots,\sigma_m$ be the trajectories of $\mathbf{D_1}$ that satisfy $O_1,\dots,O_{|O|}$ as given in section~\ref{dqa:sem:obs:filter}. Then, $\mathbf{D}$ satisfies $\mathbf{Q}$ if $\{\sigma_1,\dots,\sigma_m\} \models U$ as defined in section~\ref{sec:dqa:qdesc:satisfaction}.

Let $\mathbf{D}$ as defined in section~\ref{plang:syntax} be a domain description and $\mathbf{Q}$ be a query statement~\eqref{dqa:syn:cmpr:qstmt} as defined in section~\ref{qlang:syntax} with query description $U$, interventions $V_1,\dots,V_{|V|}$, internal observations $O_1,\dots,O_{|O|}$, and initial conditions $I_1,\dots,I_{|I|}$. Let $\mathbf{D_)} \equiv \mathbf{D} \diamond I_1 \diamond \dots \diamond I_{|I|} $ be the nominal domain description constructed by applying the initial conditions from $\mathbf{Q}$ as defined in section~\ref{sec:semantics:idesc}. Let $\sigma_1,\dots,\sigma_m$ be the trajectories of $\mathbf{D_0}$. Let $\mathbf{D_1} \equiv \mathbf{D} \diamond I_1 \diamond \dots \diamond I_{|I|} \diamond V_1 \diamond \dots \diamond V_{|V|}$ be the modified domain description constructed by applying the initial conditions and interventions from $\mathbf{Q}$ as defined in section~\ref{sec:semantics:idesc}. Let $\bar{\sigma}_1,\dots,\bar{\sigma}_{\bar{m}}$ be the trajectories of $\mathbf{D_1}$ that satisfy $O_1,\dots,O_{|O|}$ as given in section~\ref{dqa:sem:obs:filter}. Then, $\mathbf{D}$ satisfies $\mathbf{Q}$ if $\Big\{\{\sigma_1,\dots,\sigma_m\}, \{\bar{\sigma}_1,\dots,\bar{\sigma}_{\bar{m}}\} \Big\} \models U$ as defined in section~\ref{sec:dqa:qdesc:satisfaction}.

\subsection{Example Encodings}
In this section we give some examples of how we will encode queries and pathways related to these queries. We will also show how the pathway is modified to answer questions~\footnote{Some of the same pathways appear in previous chapters, they have been updated here with additional background knowledge.}.

\begin{question}\label{q:q1}
At one point in the process of glycolysis, both DHAP and G3P are produced. Isomerase catalyzes the reversible conversion between these two isomers. The conversion of DHAP to G3P never reaches equilibrium and G3P is used in the next step of glycolysis. What would happen to the rate of glycolysis if DHAP were removed from the process of glycolysis as quickly as it was produced?
\end{question}

\begin{figure}[htbp]
   \centering
   \includegraphics[width=0.7\linewidth]{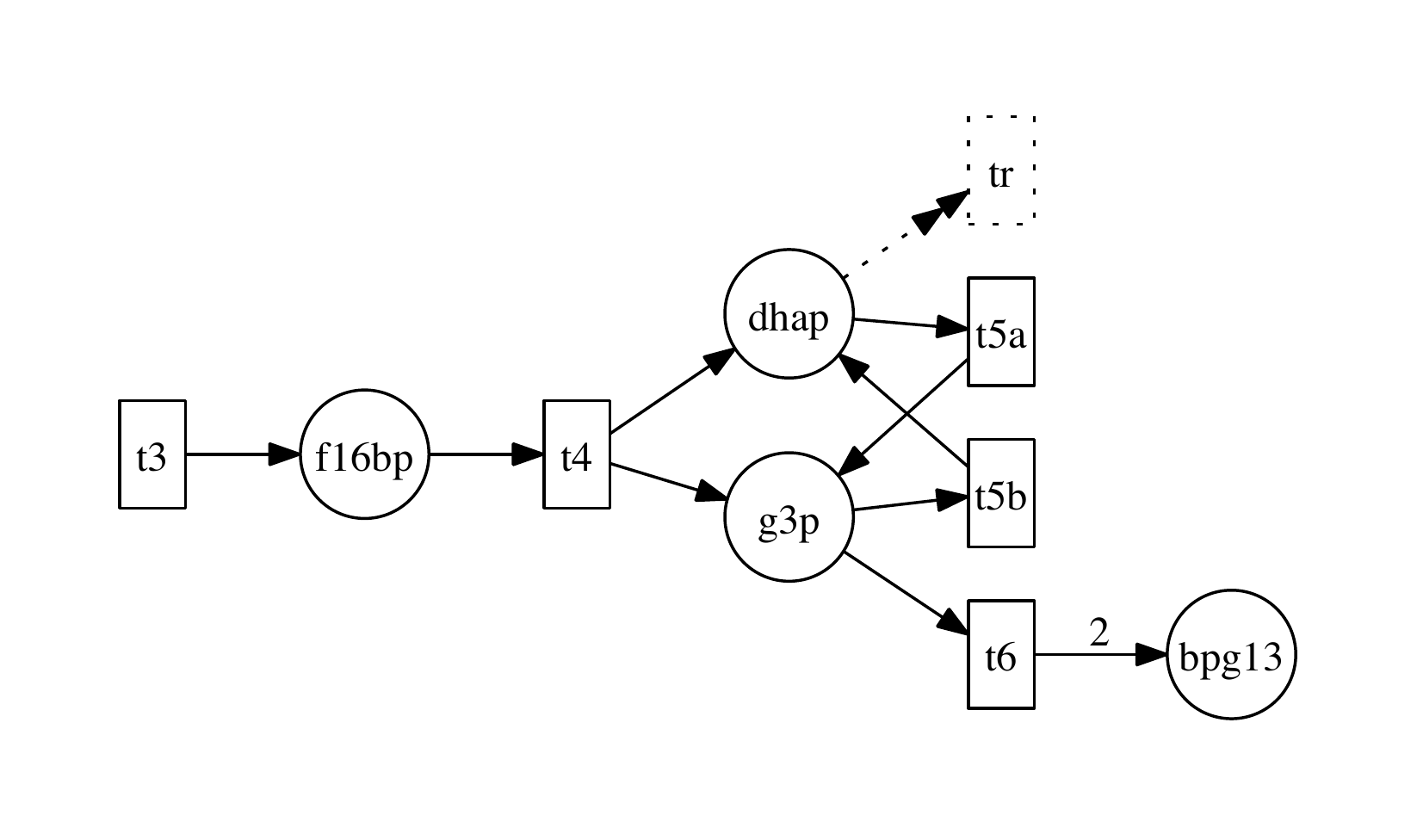}
   \caption{Petri Net for question~\ref{q:q1}}
   \label{fig:q:q1}
\end{figure}

The question is asking for the direction of change in the {\em rate of glycolysis} when the nominal glycolysis pathway is compared against a modified pathway in which {\em dhap} is removed as soon as it is produced. Since this rate can vary with the trajectory followed by the world evolution, we consider the average change in rate.
From the domain knowledge~\cite[Figure 9.9]{CampbellBook} we know that the rate of $glycolysis$ can be measured by the rate of $pyruvate$ (the end product of glycolysis) and that the rate of $pyruvate$ is equal to the rate of $bpg13$ (due to linear chain from $bpg13$ to $pyruvate$). Thus, we can monitor the rate of $bpg13$ instead to determine the rate of glycolysis. To ensure that our pathway is not starved due to source ingredients, we add a continuous supply of $f16bp$ in quantity $1$ to the pathway. 

Then, the following pathway specification encodes the domain description $\mathbf{D}$ for question~\ref{q:q1} and produces the PN in Fig.~\ref{fig:q:q1} minus the $tr,t3$ transitions:
{\footnotesize
\begin{eqnarray}\label{q:q1:pspec}
&\begin{array}{llll}
&\mathbf{domain~of~} & f16bp \mathbf{~is~} integer, & dhap \mathbf{~is~} integer, \\&& g3p \mathbf{~is~} integer, & bpg13 \mathbf{~is~} integer\nonumber\\
&t4 \mathbf{~may~execute~causing~} & f16bp \mathbf{~change~value~by~} -1, & dhap \mathbf{~change~value~by~} +1,\\ && g3p \mathbf{~change~value~by~} +1\nonumber\\
&t5a \mathbf{~may~execute~causing~} & dhap \mathbf{~change~value~by~} -1, & g3p \mathbf{~change~value~by~} +1\nonumber\\
&t5b \mathbf{~may~execute~causing~} & g3p \mathbf{~change~value~by~} -1, & dhap \mathbf{~change~value~by~} +1\nonumber\\
&t6 \mathbf{~may~execute~causing~} & g3p \mathbf{~change~value~by~} -1, & bpg13 \mathbf{~change~value~by~} +2\nonumber\\
& \mathbf{initially~} & f16bp \mathbf{~has~value~} 0, & dhap \mathbf{~has~value~} 0,\\ && g3p \mathbf{~has~value~} 0, & bpg13 \mathbf{~has~value~} 0\nonumber\\ 
&\mathbf{firing~style~} & max
\end{array}\\
\end{eqnarray}
}

And the following query $\mathbf{Q}$ for a simulation of length $k$ encodes the question:
{\footnotesize
\begin{align}\label{q:q1:qspec}
&\mathbf{direction~of~change~in~} average \mathbf{~rate~of~production~of~} bpg13 \mathbf{~is~} d\nonumber\\
&~~~~~~\mathbf{when~observed~between~time~step~} 0 \mathbf{~and~time~step~} k;\nonumber\\
&~~~~~~\mathbf{comparing~nominal~pathway~with~modified~pathway~obtained~}\nonumber\\
&~~~~~~~~~~~~\mathbf{due~to~interventions:} \mathbf{~remove~} dhap \mathbf{~as~soon~as~produced};\nonumber\\
&~~~~~~\mathbf{using~initial~setup:~} \mathbf{continuously~supply~} f16bp \mathbf{~in~quantity~} 1;
\end{align}
}

Since this is a comparative quantitative query statement, it is decomposed into two sub-queries, $\mathbf{Q_0}$ capturing the nominal case of average rate of production w.r.t. given initial conditions:
{\footnotesize
\begin{align*}
\begin{array}{lll}
average \mathbf{~rate~of~production~of~} bpg13 \mathbf{~is~} n_{avg}\\
~~~~~~~~~~~~\mathbf{when~observed~between~time~step~} 0 \mathbf{~and~time~step~} k; \\
~~~~~~\mathbf{using~initial~setup:~} \mathbf{continuously~supply~} f16bp \mathbf{~in~quantity~} 1;
\end{array}
\end{align*}
and $\mathbf{Q_1}$ capturing the modified case in which the pathway is subject to interventions and observations w.r.t. initial conditions:
\begin{align*}
\begin{array}{lll}
average \mathbf{~rate~of~production~of~} bpg13 \mathbf{~is~} n'_{avg}\\
~~~~~~~~~~~~\mathbf{when~observed~between~time~step~} 0 \mathbf{~and~time~step~} k; \\
~~~~~~\mathbf{due~to~interventions:} \mathbf{~remove~} dhap \mathbf{~as~soon~as~produced} ;\\
~~~~~~\mathbf{using~initial~setup:~} \mathbf{continuously~supply~} f16bp \mathbf{~in~quantity~} 1;
\end{array}
\end{align*} 
}

Then the task is to determine $d$, such that $\mathbf{D} \models \mathbf{Q_0}$ for some value of $n_{avg}$, $\mathbf{D} \models \mathbf{Q_1}$ for some value of $n'_{avg}$, and $n'_{avg} \; d \; n_{avg}$. To answer the sub-queries $\mathbf{Q_0}$ and $\mathbf{Q_1}$ we build modified domain descriptions $\mathbf{D_0}$ and $\mathbf{D_1}$, where, $\mathbf{D_0} \equiv \mathbf{D} \diamond (\mathbf{continuously~supply~} f16bp $ $\mathbf{~in~quantity~} 1)$ is the nominal domain description $\mathbf{D}$ modified according to $\mathbf{Q_0}$ to include the initial conditions; and $\mathbf{D_1} \equiv \mathbf{D_0} \diamond (\mathbf{remove~} dhap $ $\mathbf{~as~soon~as~}$ $\mathbf{produced})$ is the modified domain description $\mathbf{D}$ modified according to $\mathbf{Q_1}$ to include the initial conditions as well as the interventions. The $\diamond$ operator modifies the domain description to its left by adding, removing or modifying pathway specification language statements to add the {\em interventions} and {\em initial conditions} description to its right.  Thus,
{\footnotesize
\begin{align*}
\mathbf{D_0} &= \mathbf{D} + \left\{
\begin{array}{llll}
tf_{f16bp} \mathbf{~may~execute~causing~} &f16bp \mathbf{~change~value~by~} +1\\
\end{array}
\right.\\
\mathbf{D_1} &= \mathbf{D_0} + \left\{
\begin{array}{llll}
tr \mathbf{~may~execute~causing~} &dhap \mathbf{~change~value~by~} *\\
\end{array}
\right.
\end{align*}
}

Performing a simulation of $k=5$ steps with $ntok=20$ max tokens, we find that the average rate of $bpg13$ production decreases from $n_{avg}=0.83$ to $n'_{avg}=0.5$. Thus, $\mathbf{D} \models \mathbf{Q}$ iff $d = '<'$. Alternatively, we say that the rate of glycolysis decreases when DHAP is removed as quickly as it is produced.

\begin{question}\label{q:q2}
When and how does the body switch to B-oxidation versus glycolysis as the major way of burning fuel?
\end{question}

\begin{figure}[htbp]
\centering
\includegraphics[width=0.4\linewidth]{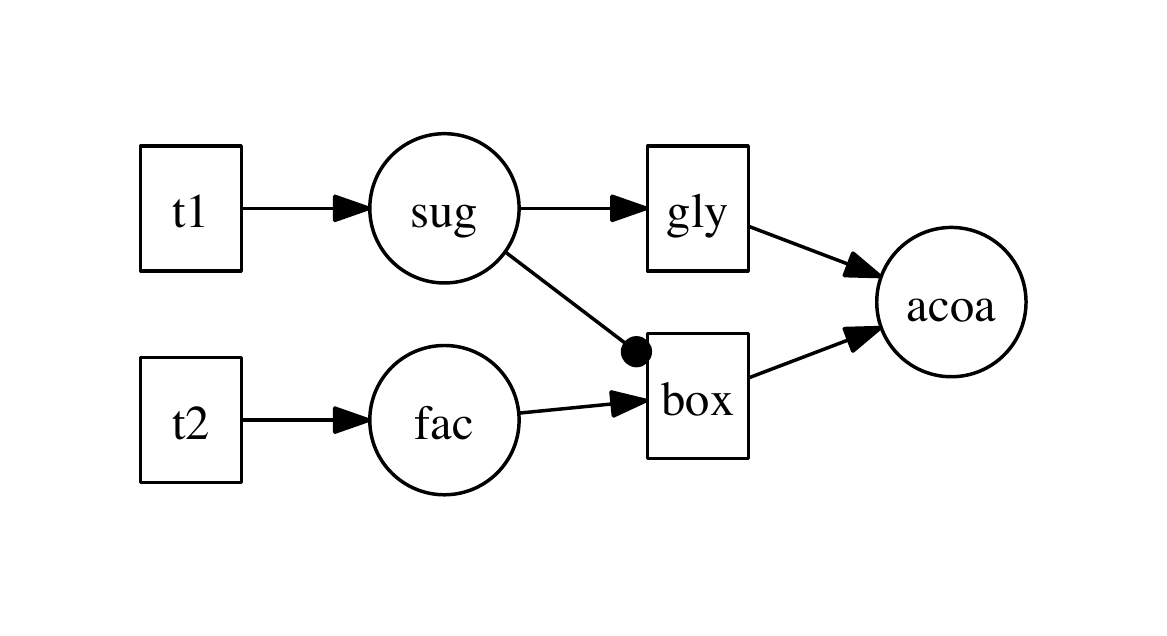}
\caption{Petri Net for question~\ref{q:q2}}
\label{fig:q:q2}
\end{figure}

\noindent
The following pathway specification encodes the domain description $\mathbf{D}$ for question~\ref{q:q2} and produces the PN in Fig.~\ref{fig:q:q2}:
{\footnotesize
\begin{align*}
\begin{array}{llll}
&\mathbf{domain~of~} & gly \mathbf{~is~} integer, & sug \mathbf{~is~} integer, \\ && fac \mathbf{~is~} integer, & acoa \mathbf{~is~} integer \\
&gly \mathbf{~may~execute~causing~} & sug \mathbf{~change~value~by~} -1, & acoa \mathbf{~change~value~by~} +1\\
&box \mathbf{~may~execute~causing~} & fac \mathbf{~change~value~by~} -1, & acoa \mathbf{~change~value~by~} +1\\
&\mathbf{inhibit~} box \mathbf{~if~} & sug \mathbf{~has~value~} 1 \mathbf{~or~higher}\\
&\mathbf{initially~} & sug \mathbf{~has~value~} 3, & fac \mathbf{~has~value~} 3\\ && acoa \mathbf{~has~value~} 0 \\
&t1 \mathbf{~may~execute~causing~} & sug \mathbf{~change~value~by~} +1\\
&t2 \mathbf{~may~execute~causing~} & fac \mathbf{~change~value~by~} +1\\
& \mathbf{firing~style~} & *
\end{array}
\end{align*}
}
where, $fac$ represents fatty acids, $sug$ represents sugar, $acoa$ represents acetyl coenzyme-A, $gly$ represents the process of glycolysis, and $box$ represents the process of beta oxidation.

The question is asking for the general conditions when glycolysis switches to beta-oxidation, which is some property ``$p$'' that holds after which the switch occurs. The query $\mathbf{Q}$ is encoded as:
{\footnotesize
\begin{align*}
&gly \mathbf{~switches~to~} box \mathbf{~when~} p; \nonumber\\
&~~~~\mathbf{due~to~observations:} \nonumber\\
&~~~~~~~~gly \mathbf{~switches~to~} box \nonumber\\
&~~~~\mathbf{~in~all~trajectories}
\end{align*}
}
where condition `$p$' is a conjunction of {\em simple point formulas}. Then the task is to determine a minimal such conjunction of formulas that is satisfied in the state where $`gly' \mathbf{~switches~to~} `box'$ holds over all trajectories.\footnote{Note that this could be an LTL formula that must hold in all trajectories, but we did not add it here to keep the language simple.}

Since there is no change in initial conditions of the pathway and there are no interventions, the modified domain description $\mathbf{D_1} \equiv \mathbf{D}$.

Intuitively, $p$ is the property that holds over fluents of the transitional state $s_j$ in which the switch takes palce, such that $gly \in T_{j-1}, box \notin T_{j-1}, gly \notin T_j, box \in T_j$ and the minimal set of firings leading up to it. The only trajectories to consider are the ones in which the observation is true. Thus the condition $p$ is determined as the intersection of sets of fluent based conditions that were true at the time of the switch, such as:
{\footnotesize
\begin{align*}
\{
&sug \mathbf{~has~value~} s_j(sug), sug \mathbf{~has~value~higher~than~} 0, \dots \\
&sug \mathbf{~has~value~higher~than~} s_j(sug)-1,sug \mathbf{~has~value~lower~than~} s_j(sug)+1,\\
&fac \mathbf{~has~value~} s_j(fac), fac \mathbf{~has~value~higher~than~} 0, \dots \\
&fac \mathbf{~has~value~higher~than~} s_j(fac)-1, fac \mathbf{~has~value~lower~than~} s_j(fac)+1, \\
&acoa \mathbf{~has~value~} s_j(acoa), acoa \mathbf{~has~value~higher~than~} 0, \dots \\
&acoa \mathbf{~has~value~higher~than~} s_j(acoa)-1, acoa \mathbf{~has~value~lower~than~} s_j(acoa)+1 
\}
\end{align*}
}

Simulating it for $k=5$ steps with $ntok=20$ max tokens, we find the condition $p = acoa \text{ has value greater than } 0, $ $sug \text{ has value } 0,  $ $sug \text{ has value lower than } 1, $ $fac \text{ has value higher than } 0 \}$. Thus, the state when this switch occurs must sugar ($sug$) depleted and available supply of fatty acids ($fac$).

\begin{question}\label{q:q5}
The final protein complex in the electron transport chain of the mitochondria is non-functional. Explain the effect of this on pH of the intermembrane space of the mitochondria.
\end{question}

\begin{figure}[htbp]
\centering
\includegraphics[width=1.0\linewidth]{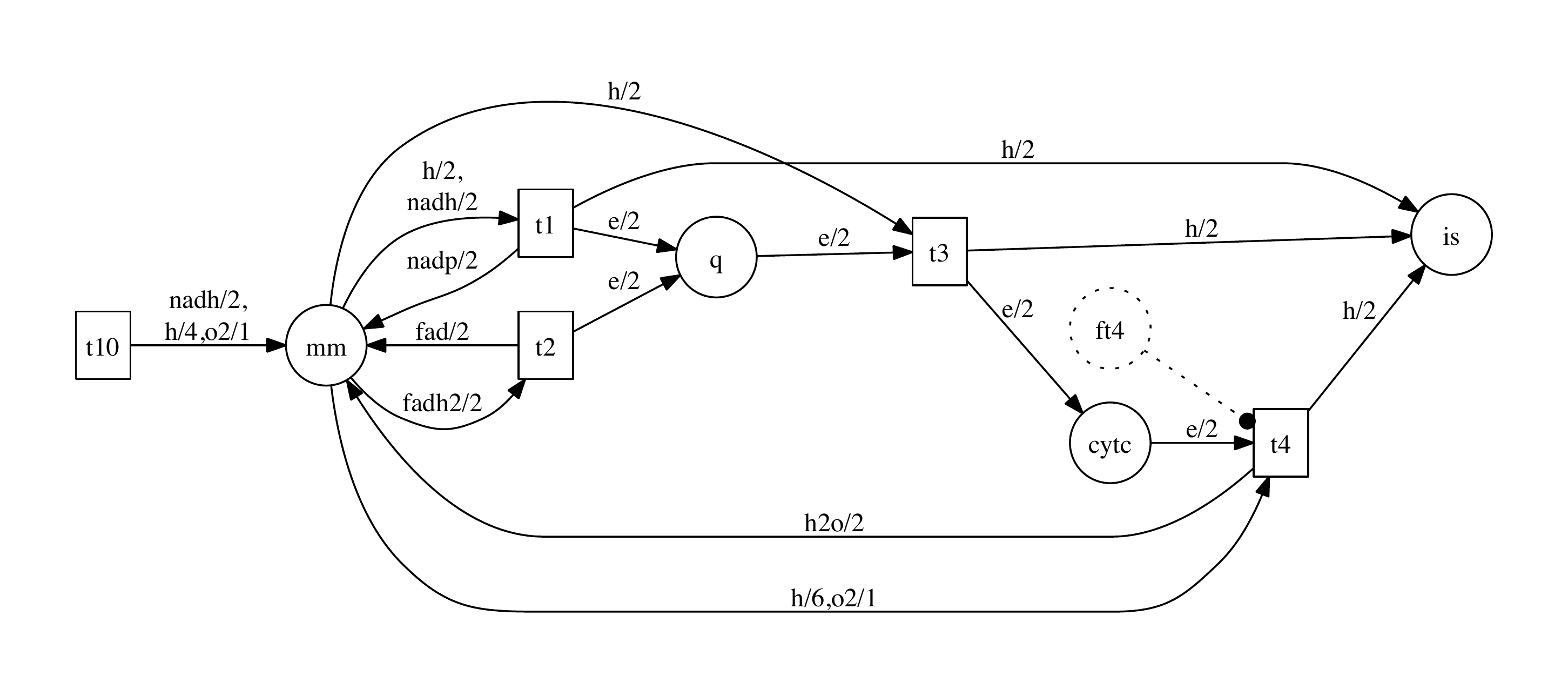}
\caption{Petri Net for question~\ref{q:q5}}
\label{fig:q:q5}
\end{figure}

The question is asking for the direction of change in the pH of the intermembrane space when the nominal case is compared against a modified pathway in which the complex 4 ($t4$) is defective. Since pH is defined as $-log_{10}(H+)$, we monitor the total production of $H+$ ions to determine the change in pH value. However, since different world evolutions can follow different trajectories, we consider the average production of H+. Furthermore, we model the defective $t4$ as being unable to carry out its reaction, by disabling/inhibiting it. 

Then the following pathway specification encodes the domain description $\mathbf{D}$ for question~\ref{q:q5} and produces the PN in Fig.~\ref{fig:q:q5} without the $ft4$ place node. In the pathway description below, we have included only one {\em domain} declaration for all fluents as an example and left the rest out to save space. Assume integer {\em domain} declaration for all fluents. 
{\footnotesize
\begin{align*}
&\mathbf{domain~of~} \\
&~~~~ nadh \mathbf{~atloc~} mm \mathbf{~is~} integer \\
&t1 \mathbf{~may~execute~causing~} \\
&~~~~ nadh \mathbf{~atloc~} mm \mathbf{~change~value~by~} -2, h \mathbf{~atloc~} mm \mathbf{~change~value~by~} -2,\\ 
&~~~~ e \mathbf{~atloc~} q \mathbf{~change~value~by~} +2, h \mathbf{~atloc~} is \mathbf{~change~value~by~} +2, \\
&~~~~ nadp \mathbf{~atloc~} mm \mathbf{~change~value~by~} +2\\
&t2 \mathbf{~may~execute~causing~} \\
&~~~~ fadh2 \mathbf{~atloc~} mm \mathbf{~change~value~by~} -2, e \mathbf{~atloc~} q \mathbf{~change~value~by~} +2,\\
&~~~~ fad \mathbf{~atloc~} mm \mathbf{~change~value~by~} +2\\
&t3 \mathbf{~may~execute~causing~} \\
&~~~~ e \mathbf{~atloc~} q \mathbf{~change~value~by~} -2, h \mathbf{~atloc~} mm \mathbf{~change~value~by~} -2,\\ 
&~~~~ e \mathbf{~atloc~} cytc \mathbf{~change~value~by~} +2, h \mathbf{~atloc~} is \mathbf{~change~value~by~} +2\\
&t4 \mathbf{~may~execute~causing~} \\
&~~~~ o2 \mathbf{~atloc~} mm \mathbf{~change~value~by~} -1, e \mathbf{~atloc~} cytc \mathbf{~change~value~by~} -2,\\ 
&~~~~ h \mathbf{~atloc~} mm \mathbf{~change~value~by~} -6, h2o \mathbf{~atloc~} mm \mathbf{~change~value~by~} +2,\\ 
&~~~~ h \mathbf{~atloc~} is \mathbf{~change~value~by~} +2\\
&t10 \mathbf{~may~execute~causing~} \\
&~~~~ nadh \mathbf{~atloc~} mm \mathbf{~change~value~by~} +2, \\ 
&~~~~ h \mathbf{~atloc~} \mathbf{~change~value~by~} +4, o2 \mathbf{~atloc~} mm \mathbf{~change~value~by~} +1\\
&\mathbf{initially~} \\
&~~~~ fadh2 \mathbf{~atloc~} mm \mathbf{~has~value~} 0, e \mathbf{~atloc~} mm \mathbf{~has~value~} 0,\\ 
&~~~~ fad \mathbf{~atloc~} mm \mathbf{~has~value~} 0, o2 \mathbf{~atloc~} mm \mathbf{~has~value~} 0,\\ 
&~~~~ h2o \mathbf{~atloc~} mm \mathbf{~has~value~} 0, atp \mathbf{~atloc~} mm \mathbf{~has~value~} 0\\
&\mathbf{initially~} \\
&~~~~ fadh2 \mathbf{~atloc~} is \mathbf{~has~value~} 0, e \mathbf{~atloc~} is \mathbf{~has~value~} 0,\\ 
&~~~~ fad \mathbf{~atloc~} is \mathbf{~has~value~} 0, o2 \mathbf{~atloc~} is \mathbf{~has~value~} 0,\\ 
&~~~~ h2o \mathbf{~atloc~} is \mathbf{~has~value~} 0, atp \mathbf{~atloc~} is \mathbf{~has~value~} 0\\
&\mathbf{initially~} \\
&~~~~ fadh2 \mathbf{~atloc~} q \mathbf{~has~value~} 0, e \mathbf{~atloc~} q \mathbf{~has~value~} 0,\\ 
&~~~~ fad \mathbf{~atloc~} q \mathbf{~has~value~} 0, o2 \mathbf{~atloc~} q \mathbf{~has~value~} 0,\\ 
&~~~~ h2o \mathbf{~atloc~} q \mathbf{~has~value~} 0, atp \mathbf{~atloc~} q \mathbf{~has~value~} 0\\
&\mathbf{initially~} \\
&~~~~ fadh2 \mathbf{~atloc~} cytc \mathbf{~has~value~} 0, e \mathbf{~atloc~} cytc \mathbf{~has~value~} 0,\\ 
&~~~~ fad \mathbf{~atloc~} cytc \mathbf{~has~value~} 0, o2 \mathbf{~atloc~} cytc \mathbf{~has~value~} 0,\\ 
&~~~~ h2o \mathbf{~atloc~} cytc \mathbf{~has~value~} 0, atp \mathbf{~atloc~} cytc \mathbf{~has~value~} 0\\
&~~~~ \mathbf{firing~style~}  max\\
\end{align*}
}
where $mm$ represents the mitochondrial matrix, $is$ represents the intermembrane space, $t1-t4$ represent the reaction of the four complexes making up the electron transport chain, $h$ is the $H+$ ion, $nadh$ is $NADH$, $fadh2$ is $FADH_2$, $fad$ is $FAD$, $e$ is electrons, $o2$ is oxygen $O_2$, $atp$ is $ATP$, $h2o$ is water $H_2O$, and $t10$ is a source transition that supply a continuous supply of source ingredients for the chain to function, such as $nadh$, $h$, $o2$.

As a result, the query $\mathbf{Q}$ asked by the question is encoded as follows:
{\footnotesize
\begin{align*}
\begin{array}{l}
\mathbf{direction~of~change~in~} average \mathbf{~total~production~of~} h \mathbf{~atloc~} is \mathbf{~is~} d\\
~~~~~~~~\mathbf{when~observed~between~time~step~} 0 \mathbf{~and~time~step~} k ;\\
~~~~~~~~\mathbf{comparing~nominal~pathway~with~modified~pathway~obtained}\\
~~~~~~~~\mathbf{due~to~intervention~} t4 \mathbf{~disabled} ;
\end{array}
\end{align*}
}

Since this is a comparative quantitative query statement, it is decomposed into two queries, $\mathbf{Q_0}$ capturing the nominal case of average production w.r.t. given initial conditions:
{\footnotesize
\begin{align*}
\begin{array}{l}
average \mathbf{~total~production~of~} h \mathbf{~atloc~} is \mathbf{~is~} n_{avg}\\
~~~~~~~~\mathbf{when~observed~between~time~step~} 0 \mathbf{~and~time~step~} k ;
\end{array}
\end{align*}
}
and $\mathbf{Q_1}$ the modified case w.r.t. initial conditions, modified to include interventions and subject to observations:
{\footnotesize
\begin{align*}
\begin{array}{l}
average \mathbf{~total~production~of~} h \mathbf{~atloc~} is \mathbf{~is~} n'_{avg}\\
~~~~~~~~\mathbf{when~observed~between~time~step~} 0 \mathbf{~and~time~step~} k ;\\
~~~~~~~~\mathbf{due~to~intervention~} t4 \mathbf{~disabled} ;
\end{array}
\end{align*}
}

The the task is to determine $d$, such that $\mathbf{D} \models \mathbf{Q_0}$ for some value of $n_{avg}$, $\mathbf{D} \models \mathbf{Q_1}$ for some value of $n'_{avg}$, and $n'_{avg} \; d \; n_{avg}$.  To answer the sub-queries $\mathbf{Q_0}$ and $\mathbf{Q_1}$  we build nominal description $\mathbf{D_0}$ and modified pathway $\mathbf{D_1}$, where $\mathbf{D_0} \equiv \mathbf{D}$ since there are no initial conditions; and $\mathbf{D_1} \equiv \mathbf{D_0} \diamond (t4 \mathbf{~disabled})$ is the domain description $\mathbf{D}$ modified according to $\mathbf{Q_1}$ to modified to include the initial conditions as well as interventions. The $\diamond$ operator modifies the domain description to its left by adding, removing or modifying pathway specification language statements to add the {\em intervention} and {\em initial conditions} to its right.  Thus,
{\footnotesize
\begin{align*}
\mathbf{D_1} = \mathbf{D_0} + \left\{
\begin{array}{llll}
\mathbf{inhibit~} t4\\
\end{array}
\right.
\end{align*}
}

Performing a simulation of $k=5$ steps with $ntok=20$ max tokens, we find that the average total production of H+ in the intermembrane space ($h$ at location $is$) reduces from $16$ to $14$. Lower quantity of H+ translates to a higer numeric value of $-log_{10}$, as a result the pH increases.

\begin{question}\label{q:q7}
Membranes must be fluid to function properly. How would decreased fluidity of the membrane affect the efficiency of the electron transport chain?
\end{question}

\begin{figure}[htbp]
   \centering
   \includegraphics[width=1.0\linewidth]{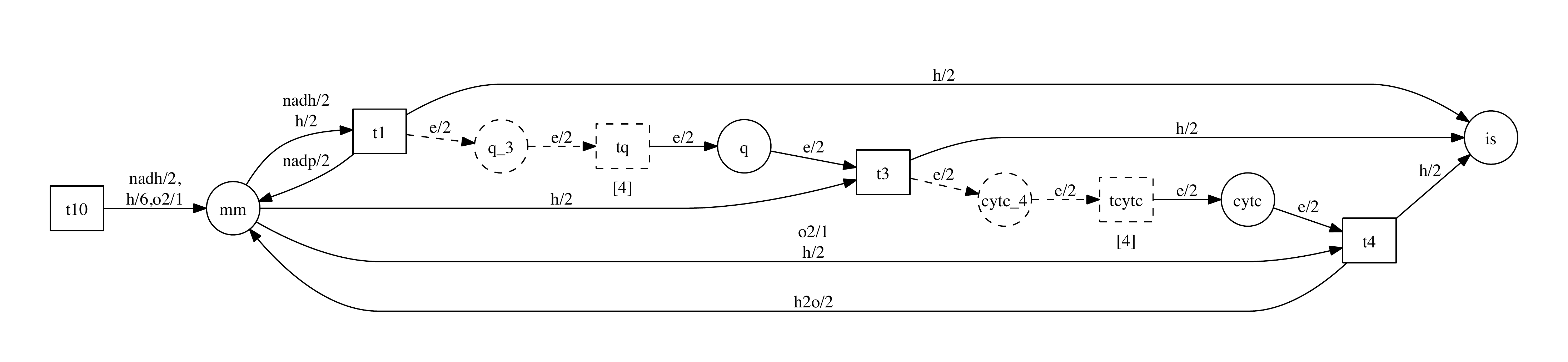}
   \caption{Petri Net with colored tokens alternate for question~\ref{q:q7}}
   \label{fig:q:q7}
\end{figure}

From background knowledge, we know that, ``Establishing the H+ gradient is a major function of the electron transport chain''~\cite[Chapter 9]{CampbellBook}, we measure the efficiency in terms of H+ ions moved to the intermembrane space ($is$) over time. Thus, we interpret the question is asking for the direction of change in the production of H+ moved to the intermembrane space when the nominal case is compared against a modified pathway with decreased fluidity of membrane. Additional background knowledge from~\cite{CampbellBook} tells us that the decreased fluidity reduces the speed of mobile carriers such as $q$ and $cytc$. Fluidity can span a range of values, but we will consider one such value $v$ per query. 

The following pathway specification encodes the domain description $\mathbf{D}$ for question~\ref{q:q7} and produces the PN in Fig.~\ref{fig:q:q7} minus the places $q\_3,cytc\_4$ and transitions $tq,tcytc$. In the pathway description below, we have included only one {\em domain} declaration for all fluents as an example and left the rest out to save space. Assume integer {\em domain} declaration for all fluents.
{\footnotesize 
\begin{align*}
&\mathbf{domain~of~} \\
&~~~~nadh \mathbf{~atloc~} mm \mathbf{~is~} integer \\
&t1 \mathbf{~may~execute~causing~} \\
&~~~~ nadh \mathbf{~atloc~} mm \mathbf{~change~value~by~} -2,  h \mathbf{~atloc~} mm \mathbf{~change~value~by~} -2,\\ 
&~~~~ e \mathbf{~atloc~} q \mathbf{~change~value~by~} +2,  h \mathbf{~atloc~} is \mathbf{~change~value~by~} +2,\\ 
&~~~~ nadp \mathbf{~atloc~} mm \mathbf{~change~value~by~} +2\\
&t3 \mathbf{~may~execute~causing~} \\
&~~~~ e \mathbf{~atloc~} q \mathbf{~change~value~by~} -2,  h \mathbf{~atloc~} mm \mathbf{~change~value~by~} -2,\\ 
&~~~~ e \mathbf{~atloc~} cytc \mathbf{~change~value~by~} +2,  h \mathbf{~atloc~} is \mathbf{~change~value~by~} +2\\
&t4 \mathbf{~may~execute~causing~} \\
&~~~~ o2 \mathbf{~atloc~} mm \mathbf{~change~value~by~} -1,  e \mathbf{~atloc~} cytc \mathbf{~change~value~by~} -2, \\
&~~~~ h \mathbf{~atloc~} mm \mathbf{~change~value~by~} -2,  h2o \mathbf{~atloc~} mm \mathbf{~change~value~by~} +2,\\ 
&~~~~ h \mathbf{~atloc~} is \mathbf{~change~value~by~} +2\\
&t10 \mathbf{~may~execute~causing~} \\
&~~~~ nadh \mathbf{~atloc~} mm \mathbf{~change~value~by~} +2,  h \mathbf{~atloc~} mm \mathbf{~change~value~by~} +4,\\ 
&~~~~ o2 \mathbf{~atloc~} mm \mathbf{~change~value~by~} +1\\
& \mathbf{initially~} \\
&~~~~ nadh \mathbf{~atloc~} mm \mathbf{~has~value~} 0,  h \mathbf{~atloc~} mm \mathbf{~has~value~} 0,\\ 
&~~~~ nadp \mathbf{~atloc~} mm \mathbf{~has~value~} 0,  o2 \mathbf{~atloc~} mm \mathbf{~has~value~} 0\\ 
&~~~~ h2o \mathbf{~atloc~} mm \mathbf{~has~value~} 0 \\
& \mathbf{initially~} \\
&~~~~ nadh \mathbf{~atloc~} is \mathbf{~has~value~} 0, h \mathbf{~atloc~} is \mathbf{~has~value~} 0,\\ 
&~~~~ nadp \mathbf{~atloc~} is \mathbf{~has~value~} 0,  o2 \mathbf{~atloc~} is \mathbf{~has~value~} 0\\ 
&~~~~ h2o \mathbf{~atloc~} is \mathbf{~has~value~} 0 \\
& \mathbf{initially~} \\
&~~~~ nadh \mathbf{~atloc~} q \mathbf{~has~value~} 0,  h \mathbf{~atloc~} q \mathbf{~has~value~} 0,\\ 
&~~~~ nadp \mathbf{~atloc~} q \mathbf{~has~value~} 0,  o2 \mathbf{~atloc~} q \mathbf{~has~value~} 0\\ 
&~~~~ h2o \mathbf{~atloc~} q \mathbf{~has~value~} 0 \\
& \mathbf{initially~} \\
&~~~~ nadh \mathbf{~atloc~} cytc \mathbf{~has~value~} 0,  h \mathbf{~atloc~} cytc \mathbf{~has~value~} 0,\\ 
&~~~~ nadp \mathbf{~atloc~} cytc \mathbf{~has~value~} 0,  o2 \mathbf{~atloc~} cytc \mathbf{~has~value~} 0\\ 
&~~~~ h2o \mathbf{~atloc~} cytc \mathbf{~has~value~} 0 \\
&\mathbf{firing~style~}  max\\
\end{align*}
}

The query $\mathbf{Q}$ asked by the question is encoded as follows:

{\footnotesize
\begin{align*}
\begin{array}{l}
\mathbf{direction~of~change~in~} average \mathbf{~total~production~of~} h \mathbf{~atloc~} is \mathbf{~is~} d\\
~~~~~~\mathbf{when~observed~between~time~step~} 0 \mathbf{~and~time~step~} k;\\
~~~~~~\mathbf{comparing~nominal~pathway~with~modified~pathway~obtained~}\\
~~~~~~~~~~~~\mathbf{due~to~interventions:} \\
~~~~~~~~~~~~~~~~~~\mathbf{add~delay~of~} v \mathbf{~time~units~in~availability~of~} e \mathbf{~atloc~} q,\\
~~~~~~~~~~~~~~~~~~\mathbf{add~delay~of~}  v \mathbf{~time~units~in~availability~of~} e \mathbf{~atloc~} cytc;
\end{array}
\end{align*}
}

Since this is a comparative quantitative query statement, we decompose it into two queries, $\mathbf{Q_0}$ capturing the nominal case of average rate of production w.r.t. given initial conditions:
{\footnotesize
\begin{align*}
\begin{array}{l}
average \mathbf{~total~production~of~} h \mathbf{~atloc~} is \mathbf{~is~} d\\
~~~~~~~~~~~~\mathbf{when~observed~between~time~step~} 0 \mathbf{~and~time~step~} k;\\
\end{array}
\end{align*}
}
and $\mathbf{Q_1}$ the modified case w.r.t. initial conditions, modified to include interventions and subject to observations:
{\footnotesize
\begin{align*}
\begin{array}{l}
average \mathbf{~total~production~of~} h \mathbf{~atloc~} is \mathbf{~is~} d\\
~~~~~~~~~~~~\mathbf{when~observed~between~time~step~} 0 \mathbf{~and~time~step~} k;\\
~~~~~~\mathbf{due~to~interventions:} \\
~~~~~~~~~~~~\mathbf{add~delay~of~} v \mathbf{~time~units~in~availability~of~} e \mathbf{~atloc~} q,\\
~~~~~~~~~~~~\mathbf{add~delay~of~}  v \mathbf{~time~units~in~availability~of~} e \mathbf{~atloc~} cytc;
\end{array}
\end{align*}
}

Then the task is to determine $d$, such that $\mathbf{D} \models \mathbf{Q_0}$ for some value of $n_{avg}$, $\mathbf{D} \models \mathbf{Q_1}$ for some value of $n'_{avg}$, and $n'_{avg} \; d \; n_{avg}$. To answer the sub-queries $\mathbf{Q_0}$ and $\mathbf{Q_1}$ we build modified domain descriptions $\mathbf{D_0}$ and $\mathbf{D_1}$, where, $\mathbf{D_0} \equiv \mathbf{D}$ is the nominal domain description $\mathbf{D}$ modified according to $\mathbf{Q_0}$ to include the initial conditions; and $\mathbf{D_1} \equiv \mathbf{D_0} \diamond (\mathbf{add~delay~of~} v \mathbf{~time~units~in~availability~of~} $ $e \mathbf{~atloc~} q) \diamond (\mathbf{add~delay~of~}  v \mathbf{~time~units~in~availability~of~} $ $e \mathbf{~atloc~} cytc)$ is the modified domain description $\mathbf{D}$ modified according to $\mathbf{Q_1}$ to include the initial conditions as well as the interventions. The $\diamond$ operator modifies the domain description to its left by adding, removing or modifying pathway specification language statements to add the {\em interventions} and {\em initial conditions} description to its right.  Thus,
{\footnotesize
\begin{align*}
\mathbf{D_1} = \mathbf{D} &- 
\left\{
\begin{array}{llll}
t1 \mathbf{~may~execute~causing~} & nadh \mathbf{~atloc~} mm \mathbf{~change~value~by~} -2, \\
	& h \mathbf{~atloc~} mm \mathbf{~change~value~by~} -2,\\
	& e \mathbf{~atloc~} q \mathbf{~change~value~by~} +2, \\
	& h \mathbf{~atloc~} is \mathbf{~change~value~by~} +2,\\ 
	& nadp \mathbf{~atloc~} mm \mathbf{~change~value~by~} +2\\
t3 \mathbf{~may~execute~causing~} & e \mathbf{~atloc~} q \mathbf{~change~value~by~} -2, \\
	& h \mathbf{~atloc~} mm \mathbf{~change~value~by~} -2,\\ 
	& e \mathbf{~atloc~} cytc \mathbf{~change~value~by~} +2, \\
	& h \mathbf{~atloc~} is \mathbf{~change~value~by~} +2\\
\end{array}
\right\}\\
& + 
\left\{
\begin{array}{llll}
t1 \mathbf{~may~execute~causing~} & nadh \mathbf{~atloc~} mm \mathbf{~change~value~by~} -2, \\
	& h \mathbf{~atloc~} mm \mathbf{~change~value~by~} -2,\\
	& e \mathbf{~atloc~} q\_3 \mathbf{~change~value~by~} +2, \\
	& h \mathbf{~atloc~} is \mathbf{~change~value~by~} +2,\\ 
	& nadp \mathbf{~atloc~} mm \mathbf{~change~value~by~} +2\\
t3 \mathbf{~may~execute~causing~} & e \mathbf{~atloc~} q \mathbf{~change~value~by~} -2, \\
	& h \mathbf{~atloc~} mm \mathbf{~change~value~by~} -2,\\ 
	& e \mathbf{~atloc~} cytc\_4 \mathbf{~change~value~by~} +2, \\
	& h \mathbf{~atloc~} is \mathbf{~change~value~by~} +2\\
tq \mathbf{~may~fire~causing~} & e \mathbf{~atloc~} q\_3 \mathbf{~change~value~by~} -2, \\ 
	& e \mathbf{~atloc~} q \mathbf{~change~value~by~} +2\\ 
tcytc \mathbf{~may~fire~causing~} & e \mathbf{~atloc~} cytc\_4 \mathbf{~change~value~by~} -2, \\ 
	& e \mathbf{~atloc~} cytc \mathbf{~change~value~by~} +2\\
tq \mathbf{~executes~} & \mathbf{~in~} 4 \mathbf{~time~units}\\
tcytc \mathbf{~executes~} & \mathbf{~in~} 4 \mathbf{~time~units}\\
\end{array}
\right\}\\
\end{align*}
}

Performing a simulation of k = 5 steps with ntok = 20 max tokens with a fluidity based delay of $2$, we find that the average total production of H+ in the intermembrane space (h at location is) reduces from 16 to 10. Lower quantity of H+ going into the intermembrane space means lower efficiency, where we define the efficiency as the total amount of $H+$ ions transferred to the intermembrane space over the simulation run.

\subsection{Example Encoding with Conditional Actions}
Next, we illustrate how conditional actions would be encoded in our high-level language with an example. Consider the pathway from question~\ref{q:q1}. Say, the reaction step $t4$ has developed a fault, in which it has two modes of operation, in the first mode, when $f16bp$ has less than $3$ units available, the reaction proceeds normally, but when $f16bp$ is available in $3$ units or higher, the reaction continues to produce $g3p$ but not $dhap$ directly. $dhap$ can still be produced by subsequent step from $g3p$. The modified pathway is given in our pathway specification language below:
{\footnotesize
\begin{equation*}
\begin{array}{llll}
&t3 \mathbf{~may~execute~causing~} & f16bp \mathbf{~change~value~by~} +1\\
&t4 \mathbf{~may~execute~causing~} & f16bp \mathbf{~change~value~by~} -1, & dhap \mathbf{~change~value~by~} +1,\\ && g3p \mathbf{~change~value~by~} +1\\
&~~~~~~~~~~~~~~~~\mathbf{if~}& g3p \mathbf{~has~value~lower~than~} 3\\
&t4 \mathbf{~may~execute~causing~} & f16bp \mathbf{~change~value~by~} -1, & dhap \mathbf{~change~value~by~} +1\\
&~~~~~~~~~~~~~~~~\mathbf{~if}& g3p \mathbf{~has~value~} 3 \mathbf{~or~higher}\\
&t5a \mathbf{~may~execute~causing~} & dhap \mathbf{~change~value~by~} -1, & g3p \mathbf{~change~value~by~} +1\\
&t5b \mathbf{~may~execute~causing~} & g3p \mathbf{~change~value~by~} -1, & dhap \mathbf{~change~value~by~} +1\\
&t6 \mathbf{~may~execute~causing~} & g3p \mathbf{~change~value~by~} -1, & bpg13 \mathbf{~change~value~by~} +2\\
& \mathbf{initially~} & f16bp \mathbf{~has~value~} 0, & dhap \mathbf{~has~value~} 0,\\ && g3p \mathbf{~has~value~} 0, & bpg13 \mathbf{~has~value~} 0\\ 
&\mathbf{firing~style~} & max
\end{array}
\end{equation*}
}

We ask the same question $\mathbf{Q}$:
{\footnotesize
\begin{align*}
&\mathbf{direction~of~change~in~} average \mathbf{~rate~of~production~of~} bpg13 \mathbf{~is~} d\\
&~~~~~~\mathbf{when~observed~between~time~step~} 0 \mathbf{~and~time~step~} k; \\
&~~~~~~\mathbf{comparing~nominal~pathway~with~modified~pathway~obtained~}\\
&~~~~~~~~~~~~\mathbf{due~to~interventions:~} \mathbf{remove~} dhap \mathbf{~as~soon~as~produced};\\
&~~~~~~~~~~~~\mathbf{using~initial~setup:~} \mathbf{continuously~supply~} f16bp \mathbf{~in~quantity~} 1; 
\end{align*}
}

Since this is a comparative quantitative query statement, we decompose it into two queries, $\mathbf{Q_0}$ capturing the nominal case of average rate of production w.r.t. given initial conditions:
{\footnotesize
\begin{align*}
\begin{array}{l}
average  \mathbf{~rate~of~production~of~} bpg13 \mathbf{~is~} n\\
~~~~~~~~~~~~\mathbf{when~observed~between~time~step~} 0 \mathbf{~and~time~step~} k;\\
~~~~~~\mathbf{using~initial~setup:~} \mathbf{continuously~supply~} f16bp \mathbf{~in~quantity~} 1; 
\end{array}
\end{align*}
}
and $\mathbf{Q_1}$ the modified case w.r.t. initial conditions, modified to include interventions and subject to observations:
{\footnotesize
\begin{align*}
\begin{array}{l}
average  \mathbf{~rate~of~production~of~} bpg13 \mathbf{~is~} n\\
~~~~~~~~~~~~\mathbf{when~observed~between~time~step~} 0 \mathbf{~and~time~step~} k;\\
~~~~~~\mathbf{due~to~interventions:~} \mathbf{remove~} dhap \mathbf{~as~soon~as~produced};\\
~~~~~~\mathbf{using~initial~setup:~} \mathbf{continuously~supply~} f16bp \mathbf{~in~quantity~} 1; 
\end{array}
\end{align*}
}

Then the task is to determine $d$, such that $\mathbf{D} \models \mathbf{Q_0}$ for some value of $n_{avg}$, $\mathbf{D} \models \mathbf{Q_1}$ for some value of $n'_{avg}$, and $n'_{avg} \; d \; n_{avg}$. To answer the sub-queries $\mathbf{Q_0}$ and $\mathbf{Q_1}$ we build modified domain descriptions $\mathbf{D_0}$ and $\mathbf{D_1}$, where, $\mathbf{D_0} \equiv \mathbf{D} \diamond (\mathbf{continuously~supply~} f16bp $ $\mathbf{~in~quantity~} 1)$ is the nominal domain description $\mathbf{D}$ modified according to $\mathbf{Q_0}$ to include the initial conditions; and $\mathbf{D_1} \equiv \mathbf{D_0} \diamond (\mathbf{remove~} dhap \mathbf{~as~soon~as~produced})$ is the modified domain description $\mathbf{D}$ modified according to $\mathbf{Q_1}$ to include the initial conditions as well as the interventions. The $\diamond$ operator modifies the domain description to its left by adding, removing or modifying pathway specification language statements to add the {\em interventions} and {\em initial conditions} description to its right.  Thus,
{\footnotesize
\begin{align*}
\mathbf{D_0} &= \mathbf{D} + \left\{
\begin{array}{llll}
tf_{f16bp} \mathbf{~may~execute~causing~} &f16bp \mathbf{~change~value~by~} +1\\
\end{array}
\right.\\
\mathbf{D_1} &= \mathbf{D_0} + \left\{
\begin{array}{llll}
tr \mathbf{~may~execute~causing~} &dhap \mathbf{~change~value~by~} *\\
\end{array}
\right.
\end{align*}
}

\subsection{ASP Program}
Next we briefly outline how the pathway specification and the query statement components are encoded in ASP (using Clingo syntax). In the following section, we will illustrate the process using an example.

As evident from the previous sections, we need to simulate non-comparative queries only. Any comparative queries are translated into non-comparative sub-queries, each of which is simulated and their results compared to evaluate the comparative query.

The ASP program is a concatenation of the translation of a pathway specification (domain description which includes the firing style) and internal observations. Any initial setup conditions and interventions are pre-applied to the pathway specification using intervention semantics in section~\ref{sec:semantics:idesc} before it is translated to ASP using the translation in chapter~\ref{ch:asp_enc} as our basis. The encoded pathway specification has the semantics defined in section~\ref{sec:plang:sem}. Internal observations in the `{\em due to observations:}' portion of query statement are translated into ASP constraints using the internal observation semantics defined in section~\ref{dqa:sem:obs:filter} and added to the encoding of the pathway specification.

The program if simulated for a specified simulation length $k$ produces all trajectories of the pathway for the specified firing style, filtered by the internal observations. The query description specified in the query statement is then evaluated w.r.t. these trajectories. Although this part can be done in ASP, we have currently implemented it outside ASP in our implementation for ease of using floating point math.

Next, we describe an implementation of our high level language and illustrate the construction of an ASP program, its simulation, and query statement evaluation.

\subsection{Implementation}
\begin{figure}[htbp]
   \centering
   \includegraphics[width=0.7\linewidth]{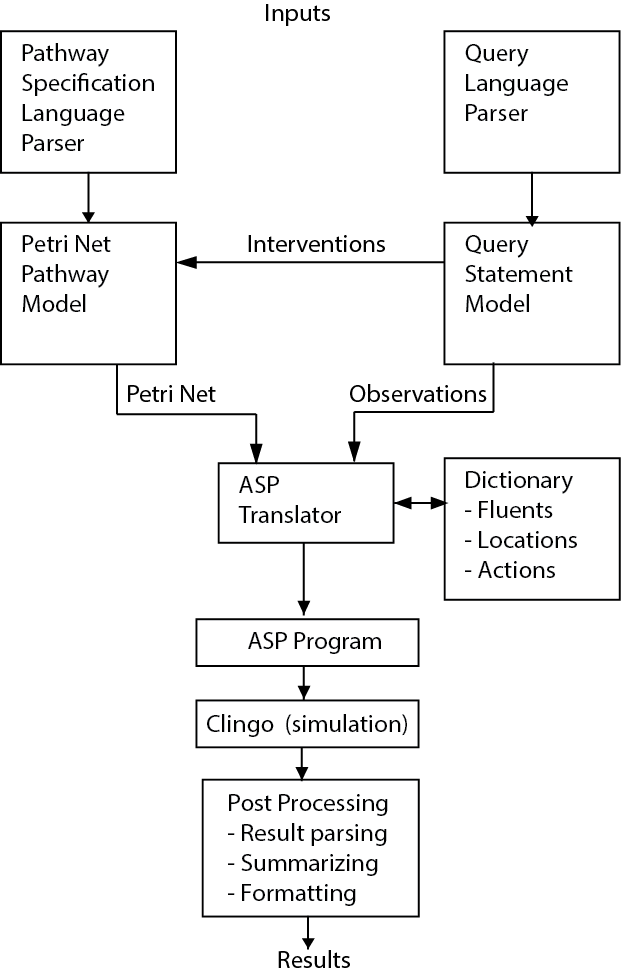} %
   \caption{BioPathQA Implementation System Architecture}
   \label{fig:dqa:sys:arch}
\end{figure}

We have developed an implementation~\footnote{Implementation available at: \texttt{https://sites.google.com/site/deepqa2014/}} of a subset of our high level (Pathway and Query specification) language in Python. We use the Clingo ASP implementation for our simulation. In this section we describe various components of this implementation. An architectural overview of our implementation is shown in figure~\ref{fig:dqa:sys:arch}.

The {\em Pathway Specification Language (BioPathQA-PL) Parser} component is responsible for parsing the Pathway Specification Language (BioPathQA-PL). It use PLY (Python Lex-Yacc)\footnote{http://www.dabeaz.com/ply} to parse a given pathway specification using grammar based on section~\ref{plang:syntax}. On a successful parse, a Guarded-Arc Petri Net pathway model based on section~\ref{sec:plang:sem} is constructed for the pathway specification.

The {\em Query Language Parser} component is responsible for parsing the Query Specification Language (BioPathQA-QL). It uses PLY to parse a given query statement using grammar based on section~\ref{qlang:syntax}. On a successful parse, an internal representation of the query statement is constructed. Elements of this internal representation include objects representing the query description, the list of interventions, the list of internal observations, and the list of initial setup conditions. Each intervention and initial setup condition object has logic in it to modify a given pathway per the intervention semantics described in section~\ref{sec:semantics:idesc}. The Query Statement Model component is also responsible for generating basic queries for aggregate queries and implementing interventions in the Petri Net Pathway Model.

The {\em Dictionary} of fluents, locations, and actions is consulted by the ASP code generator to standardize symbol names in the ASP code produced for the pathway specification and the internal observations.

The {\em ASP Translator} component is responsible for translating the Guarded-Arc Petri Net model into ASP facts and rules; and the driver needed to simulate the model using the firing semantics specified in the pathway model. The code generated is based on the ASP translation of Petri Nets and its various extensions given in chapter~\ref{ch:asp_enc}. To reduce the ASP code and its complexity, the translator limits the output model to the extensions used in the Petri Net model to be translated. Thus, the colored tokens extension code is not produced unless colored tokens have been used. Similarly, guarded-arcs code is not produced if no arc-guards are used in the model. 

The ASP Translator component is also responsible for translating internal observations from the Query Statement into ASP constraints to filter Petri Net trajectories based on the observation semantics in section~\ref{dqa:sem:obs:filter}. Following examples illustrate our encoding scheme.
The observation `$a_1 \mathbf{~switches~to~} a_2$' is encoded as a constraint using the following rules:
{\footnotesize
\begin{quote}
\begin{verbatim}
obs_1_occurred(TS+1) :- time(TS;TS+1), trans(a1;a2),
  fires(a1,TS), not fires(a2,TS),
  not fires(a1,TS+1), fires(a2,TS+1).
obs_1_occurred :- obs_1_occurred(TS), time(TS).
obs_1_had_occurred(TSS) :- obs_1_occurred(TS), TS<=TSS, time(TSS;TS).
:- not obs_1_occurred.
\end{verbatim}
\end{quote}
}
The observation `$a_1 \mathbf{~occurs~at~time~step~} 5$' is encoded as a constraint using the following rules:
{\footnotesize
\begin{quote}
\begin{verbatim}
obs_2_occurred(TS) :- fires(a1,TS), trans(a1), time(TS), TS=5.
obs_2_occurred :- obs_2_occurred(TS), time(TS).
obs_2_had_occurred(TSS) :- obs_2_occurred(TS), TS<=TSS, time(TSS;TS).
:- not obs_2_occurred.
\end{verbatim}
\end{quote}
}
The observation `$s_1 \mathbf{~is~decreasing~atloc~} l_1 \mathbf{~when~observed~between~time~step~} 0 $ $\mathbf{~and~time~step~} 5$' is encoded as a constraint using the following rules:
{\footnotesize
\begin{quote}
\begin{verbatim}
%
obs_3_violated(TS) :- place(l1), col(s1),
  holds(l1,Q1,s1,TS), holds(l1,Q2,s1,TS+1),
  num(Q1;Q2), Q2 > Q1, time(TS;TS+1), TS=0, TS+1=5.
obs_3_violated :- obs_3_violated(TS), time(TS).

%
obs_3_occurred(TS+1) :- not obs_3_violated,
  holds(l1,Q1,s1,TS), holds(l1,Q2,s1,TS+1), time(TS;TS+1), num(Q1;Q2),
  Q2<Q1, TS=0, TS+1=5.
obs_3_occurred :- obs_3_occurred(TS), time(TS).

obs_3_had_occurred(TSS) :- obs_3_occurred(TS), TS<=TSS, time(TSS;TS).
:- obs_3_occurred.
\end{verbatim}
\end{quote}
}

In addition, the translator is also responsible for any rules needed to ease post-processing of the query description. For example, for qualitative queries, a generic predicate \texttt{tgt\_obs\_occurred(TS)} is generated that is {\em true} when the given qualitative description holds in an answer-set at time step $TS$. The output of the translator is an ASP program, which when simulated using Clingo produces the (possibly) filtered trajectories of the pathway.

The {\em Post Processor} component is responsible for parsing the ASP answer sets, identifying the correct atoms from it, extracting quantities from atom-bodies as necessary, organizing them into a matrix form, and aggregating them as needed.

\begin{figure}[htbp]
   \centering
   \includegraphics[width=0.5\linewidth]{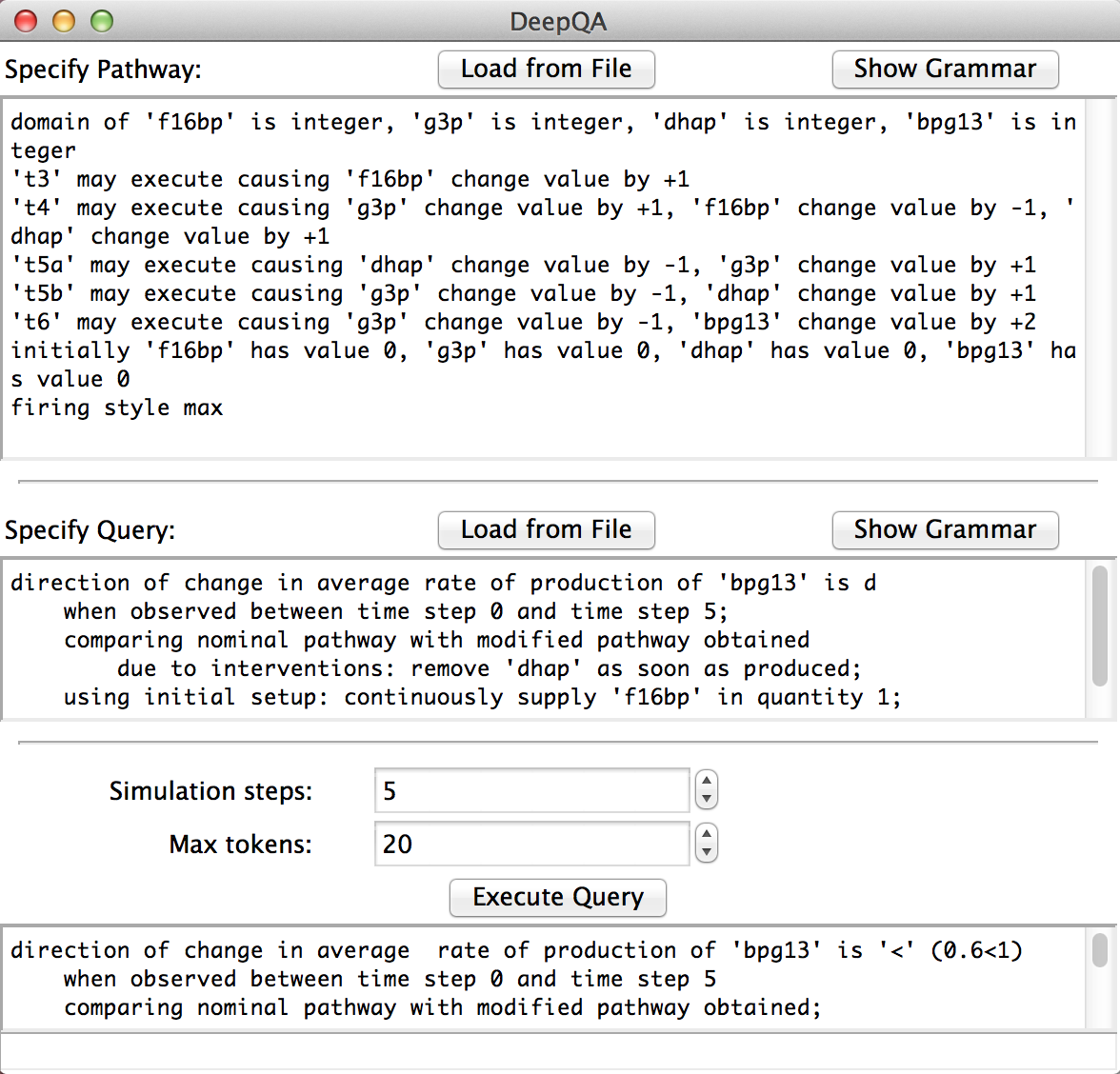} %
   \caption{BioPathQA Graphical User Interface}
   \label{fig:dqa:ui}
\end{figure}

The {\em User Interface} component is responsible for coordinating the processing of query statement. It presents the user with a graphical user interface shown in figure~\ref{fig:dqa:ui}. The user types a Pathway Specification (in BioPathQA-PL syntax), a Query Specification (in BioPathQA-QL syntax), and simulation parameters. On pressing ``Execute Query'', the user interface component processes the query as prints results in the bottom box. Query evaluation differs by the type of query. We describe the query evaluation methodology used below.

For non-comparative quantitative queries:
\begin{enumerate}
\item\label{dqa:qeval:1} Pathway specification is parsed into a Guarded-Arc Petri Net model.
\item Query statement is parsed into an internal form.
\item Initial conditions from the query are applied to the pathway model.
\item Interventions are applied to the pathway model.
\item Modified pathway model is translated to ASP.
\item \label{dqa:qeval:6}Internal observations are added to the ASP code as ASP constraints.
\item Answer sets of the ASP code are computed using Clingo.
\item Relevant atoms are extracted: \texttt{fires/2} predicate for firing rate, \texttt{holds/3} (or \texttt{holds/4} -- colored tokens) predicate for fluent quantity or rate formulas.
\item Fluent value or firing-count values are extracted and organized as matrices with rows representing answer-sets and columns representing time-steps.
\item Within answer-set interval or point value sub-select is done and the values converted to rates or totals as needed.
\item \label{dqa:qeval:11}If aggregation, such as average, minimum, or maximum is desired, it is performed over rows of values from the last step.
\item If a value was specified in the query, it is compared against the computed aggregate for boolean result.
\item If a value was not specified, the computed value is returned as the value satisfying the query statement.
\item For queries over all trajectories, the same value must hold over all trajectories, otherwise, only one match is required to satisfy the query.
\end{enumerate}

For non-comparative qualitative queries:
\begin{enumerate}
\item Follow steps \eqref{dqa:qeval:1}-\eqref{dqa:qeval:6} of non-comparative quantitative queries.
\item Add rules for the query description for post-processing.
\item Answer sets of the ASP code are computed using Clingo.
\item Relevant atoms are extracted: \texttt{tgt\_obs\_occurred/1} identifying the time step when the observation within the query description is satisfied.
\item Truth value of the query observation is determined, including determining the truth value over all trajectories.
\end{enumerate}

For comparative quantitative queries:
\begin{enumerate}
\item Query statement is decomposed into two non-comparative quantitative sub-query statements as illustrated in section~\ref{sec:dqa:illustrative:example}:
\begin{enumerate}
\item A nominal sub-query which has the same initial conditions as the comparative query, but none of its interventions or observations
\item A modified sub-query which has the same initial conditions, interventions, and observations as the comparative query
\end{enumerate}
both sub-query statements have the same query description, which is the non-aggregate form of the comparative query description. Thus, a comparative average rate query is translated to non-comparative average rate sub-queries.
\item Each sub-query statements is evaluated using steps \eqref{dqa:qeval:1}-\eqref{dqa:qeval:11} from the non-comparative quantitative query processing.
\item A direction of change is computed by comparing the computed aggregate value for the modified query statement to the nominal query statement.
\item If the comparative quantitative query has a direction specified, it is the compared against the computed value for a boolean result.
\item If the comparative quantitative query did not have a direction specified, the computed value is returned as the value satisfying the query statement.
\end{enumerate}

For explanation queries with query description with formula of the form~\eqref{dqa:syn:point:formula:cascade:when:cond}, it is expected that the number of answer-sets will be quite large. So, we avoid generating all answer-sets before processing them, instead we process them in-line as they are generated. It is a bit slower, but leads to a smaller memory foot print.
\begin{enumerate}
\item Follow steps \eqref{dqa:qeval:1}-\eqref{dqa:qeval:6} of non-comparative quantitative queries.
\item Add rules for the query description for post-processing.
\item Compute answer sets of the ASP code using Clingo.
\item Extract relevant atoms:
\begin{enumerate}
\item extract \texttt{tgt\_obs\_occurred/1} identifying the time step when the query description is satisfied
\item extract \texttt{holds/3} (or \texttt{holds/4} -- for colored tokens) at the same time-step as \texttt{tgt\_obs\_occurred/1} to construct fluent-based conditions
\end{enumerate} 
\item Construct fluent-based conditions as explanation of the query observation.
\item If the query is over all trajectories, fluent-based conditions for each trajectory are intersected across trajectories to determine the minimum set of conditions explaining the query observation.
\end{enumerate}

Next we illustrate query processing through an execution trace of question~\eqref{q:q1}. The following shows the encoding of the base case domain, which includes the pathway specification from~\eqref{q:q1:pspec} with initial setup conditions from the query statement~\eqref{q:q1:qspec} applied:
{\footnotesize
\begin{verbatim}
%
#const nts=1.
time(0..nts).

%
#const ntok=1.
num(0..ntok).

%
place(bpg13).
place(dhap).
place(f16bp).
place(g3p).

%
trans(src_f16bp_1).
trans(t3).
trans(t4).
trans(t5a).
trans(t5b).
trans(t6).

%
tparc(src_f16bp_1,f16bp,1,TS) :- time(TS).

tparc(t3,f16bp,1,TS) :- time(TS).

ptarc(f16bp,t4,1,TS) :- time(TS).
tparc(t4,g3p,1,TS) :- time(TS).
tparc(t4,dhap,1,TS) :- time(TS).

ptarc(dhap,t5a,1,TS) :- time(TS).
tparc(t5a,g3p,1,TS) :- time(TS).

ptarc(g3p,t5b,1,TS) :- time(TS).
tparc(t5b,dhap,1,TS) :- time(TS).

ptarc(g3p,t6,1,TS) :- time(TS).
tparc(t6,bpg13,2,TS) :- time(TS).

%
holds(bpg13,0,0).
holds(dhap,0,0).
holds(f16bp,0,0).
holds(g3p,0,0).

%

%
could_not_have(T,TS) :- enabled(T,TS), not fires(T,TS), ptarc(P,T,Q,TS),
    holds(P,QQ,TS), tot_decr(P,QQQ,TS), Q > QQ - QQQ.
:- not could_not_have(T,TS), time(TS), enabled(T,TS), not fires(T,TS), trans(T).


%

%
min(A,B,A) :- A<=B, num(A;B).
min(A,B,B) :- B<=A, num(A;B).
#hide min/3.

%
holdspos(P):- holds(P,N,0), place(P), num(N), N > 0.
holds(P,0,0) :- place(P), not holdspos(P).

%
notenabled(T,TS) :- ptarc(P,T,N,TS), holds(P,Q,TS), Q < N, place(P), trans(T), 
    time(TS), num(N), num(Q).

%
enabled(T,TS) :- trans(T), time(TS), not notenabled(T, TS).

%
{ fires(T,TS) } :- enabled(T,TS), trans(T), time(TS).

%
add(P,Q,T,TS) :- fires(T,TS), tparc(T,P,Q,TS), time(TS).
del(P,Q,T,TS) :- fires(T,TS), ptarc(P,T,Q,TS), time(TS).

%
tot_incr(P,QQ,TS) :- QQ = #sum[add(P,Q,T,TS) = Q : num(Q) : trans(T)], 
     time(TS), num(QQ), place(P).
tot_decr(P,QQ,TS) :- QQ = #sum[del(P,Q,T,TS) = Q : num(Q) : trans(T)], 
    time(TS), num(QQ), place(P).

%
holds(P,Q,TS+1) :- holds(P,Q1,TS), tot_incr(P,Q2,TS), tot_decr(P,Q3,TS), 
    Q=Q1+Q2-Q3, place(P), num(Q;Q1;Q2;Q3), time(TS), time(TS+1).

%
%
consumesmore(P,TS) :- holds(P,Q,TS), tot_decr(P,Q1,TS), Q1 > Q.
consumesmore :- consumesmore(P,TS).
:- consumesmore.
\end{verbatim}
}

The following shows encoding of the alternate case domain, which consists of the pathway specification from~\eqref{q:q1:pspec} with initial setup conditions and interventions applied; and any internal observations from the query statement~\eqref{q:q1:qspec} added:
{\footnotesize
\begin{verbatim}
%
#const nts=1.
time(0..nts).

%
#const ntok=1.
num(0..ntok).

%
place(bpg13).
place(dhap).
place(f16bp).
place(g3p).

%
trans(reset_dhap_1).
trans(src_f16bp_1).
trans(t3).
trans(t4).
trans(t5a).
trans(t5b).
trans(t6).

%
ptarc(dhap,reset_dhap_1,Q,TS) :- holds(dhap,Q,TS), Q>0, time(TS).
:- enabled(reset_dhap_1,TS), not fires(reset_dhap_1,TS), time(TS).

tparc(src_f16bp_1,f16bp,1,TS) :- time(TS).

tparc(t3,f16bp,1,TS) :- time(TS).

ptarc(f16bp,t4,1,TS) :- time(TS).
tparc(t4,g3p,1,TS) :- time(TS).
tparc(t4,dhap,1,TS) :- time(TS).

ptarc(dhap,t5a,1,TS) :- time(TS).
tparc(t5a,g3p,1,TS) :- time(TS).

ptarc(g3p,t5b,1,TS) :- time(TS).
tparc(t5b,dhap,1,TS) :- time(TS).

ptarc(g3p,t6,1,TS) :- time(TS).
tparc(t6,bpg13,2,TS) :- time(TS).

%
holds(bpg13,0,0).
holds(dhap,0,0).
holds(f16bp,0,0).
holds(g3p,0,0).

%

%
could_not_have(T,TS) :- enabled(T,TS), not fires(T,TS), ptarc(P,T,Q,TS),
    holds(P,QQ,TS), tot_decr(P,QQQ,TS), Q > QQ - QQQ.
:- not could_not_have(T,TS), time(TS), enabled(T,TS), not fires(T,TS), trans(T).


%

%
min(A,B,A) :- A<=B, num(A;B).
min(A,B,B) :- B<=A, num(A;B).
#hide min/3.

%
holdspos(P):- holds(P,N,0), place(P), num(N), N > 0.
holds(P,0,0) :- place(P), not holdspos(P).

%
notenabled(T,TS) :- ptarc(P,T,N,TS), holds(P,Q,TS), Q < N, place(P), trans(T), 
    time(TS), num(N), num(Q).

%
enabled(T,TS) :- trans(T), time(TS), not notenabled(T, TS).

%
{ fires(T,TS) } :- enabled(T,TS), trans(T), time(TS).

%
add(P,Q,T,TS) :- fires(T,TS), tparc(T,P,Q,TS), time(TS).
del(P,Q,T,TS) :- fires(T,TS), ptarc(P,T,Q,TS), time(TS).

%
tot_incr(P,QQ,TS) :- QQ = #sum[add(P,Q,T,TS) = Q : num(Q) : trans(T)], 
     time(TS), num(QQ), place(P).
tot_decr(P,QQ,TS) :- QQ = #sum[del(P,Q,T,TS) = Q : num(Q) : trans(T)], 
    time(TS), num(QQ), place(P).

%
holds(P,Q,TS+1) :- holds(P,Q1,TS), tot_incr(P,Q2,TS), tot_decr(P,Q3,TS), 
    Q=Q1+Q2-Q3, place(P), num(Q;Q1;Q2;Q3), time(TS), time(TS+1).

%
%
consumesmore(P,TS) :- holds(P,Q,TS), tot_decr(P,Q1,TS), Q1 > Q.
consumesmore :- consumesmore(P,TS).
:- consumesmore.
\end{verbatim}
}

Both programs are simulated for $5$ time-steps and $20$ max tokens using the following Clingo command:
{\footnotesize
\begin{quote}
\begin{verbatim}
clingo 0 -cntok=20 -cnts=5 program.lp
\end{verbatim}
\end{quote}
}

Answer sets of the base case are as follows:
{\footnotesize
\begin{verbatim}
Answer: 1
holds(bpg13,0,0) holds(dhap,0,0) holds(f16bp,0,0) holds(g3p,0,0)
fires(src_f16bp_1,0) fires(t3,0)
holds(bpg13,0,1) holds(dhap,0,1) holds(f16bp,2,1) holds(g3p,0,1)
fires(src_f16bp_1,1) fires(t3,1) fires(t4,1)
holds(bpg13,0,2) holds(dhap,1,2) holds(f16bp,3,2) holds(g3p,1,2)
fires(src_f16bp_1,2) fires(t3,2) fires(t4,2) fires(t5a,2) fires(t5b,2)
holds(bpg13,0,3) holds(dhap,2,3) holds(f16bp,4,3) holds(g3p,2,3)
fires(src_f16bp_1,3) fires(t3,3) fires(t4,3) fires(t5a,3) fires(t5b,3) fires(t6,3)
holds(bpg13,2,4) holds(dhap,3,4) holds(f16bp,5,4) holds(g3p,2,4)
fires(src_f16bp_1,4) fires(t3,4) fires(t4,4) fires(t5a,4) fires(t5b,4) fires(t6,4)
holds(bpg13,4,5) holds(dhap,4,5) holds(f16bp,6,5) holds(g3p,2,5)
fires(src_f16bp_1,5) fires(t3,5) fires(t4,5) fires(t5a,5) fires(t5b,5) fires(t6,5)
Answer: 2
holds(bpg13,0,0) holds(dhap,0,0) holds(f16bp,0,0) holds(g3p,0,0)
fires(src_f16bp_1,0) fires(t3,0)
holds(bpg13,0,1) holds(dhap,0,1) holds(f16bp,2,1) holds(g3p,0,1)
fires(src_f16bp_1,1) fires(t3,1) fires(t4,1)
holds(bpg13,0,2) holds(dhap,1,2) holds(f16bp,3,2) holds(g3p,1,2)
fires(src_f16bp_1,2) fires(t3,2) fires(t4,2) fires(t5a,2) fires(t6,2)
holds(bpg13,2,3) holds(dhap,1,3) holds(f16bp,4,3) holds(g3p,2,3)
fires(src_f16bp_1,3) fires(t3,3) fires(t4,3) fires(t5a,3) fires(t5b,3) fires(t6,3)
holds(bpg13,4,4) holds(dhap,2,4) holds(f16bp,5,4) holds(g3p,2,4)
fires(src_f16bp_1,4) fires(t3,4) fires(t4,4) fires(t5a,4) fires(t5b,4) fires(t6,4)
holds(bpg13,6,5) holds(dhap,3,5) holds(f16bp,6,5) holds(g3p,2,5)
fires(src_f16bp_1,5) fires(t3,5) fires(t4,5) fires(t5a,5) fires(t5b,5) fires(t6,5)
\end{verbatim}
}

Answer sets of the alternate case are as follows:
{\footnotesize
\begin{verbatim}
Answer: 1
holds(bpg13,0,0) holds(dhap,0,0) holds(f16bp,0,0) holds(g3p,0,0)
fires(reset_dhap_1,0) fires(src_f16bp_1,0) fires(t3,0)
holds(bpg13,0,1) holds(dhap,0,1) holds(f16bp,2,1) holds(g3p,0,1)
fires(reset_dhap_1,1) fires(src_f16bp_1,1) fires(t3,1) fires(t4,1)
holds(bpg13,0,2) holds(dhap,1,2) holds(f16bp,3,2) holds(g3p,1,2)
fires(reset_dhap_1,2) fires(src_f16bp_1,2) fires(t3,2) fires(t4,2) fires(t6,2)
holds(bpg13,2,3) holds(dhap,1,3) holds(f16bp,4,3) holds(g3p,1,3)
fires(reset_dhap_1,3) fires(src_f16bp_1,3) fires(t3,3) fires(t4,3) fires(t5b,3)
holds(bpg13,2,4) holds(dhap,2,4) holds(f16bp,5,4) holds(g3p,1,4)
fires(reset_dhap_1,4) fires(src_f16bp_1,4) fires(t3,4) fires(t4,4) fires(t5b,4)
holds(bpg13,2,5) holds(dhap,2,5) holds(f16bp,6,5) holds(g3p,1,5)
fires(reset_dhap_1,5) fires(src_f16bp_1,5) fires(t3,5) fires(t4,5) fires(t5b,5)
Answer: 2
holds(bpg13,0,0) holds(dhap,0,0) holds(f16bp,0,0) holds(g3p,0,0)
fires(reset_dhap_1,0) fires(src_f16bp_1,0) fires(t3,0)
holds(bpg13,0,1) holds(dhap,0,1) holds(f16bp,2,1) holds(g3p,0,1)
fires(reset_dhap_1,1) fires(src_f16bp_1,1) fires(t3,1) fires(t4,1)
holds(bpg13,0,2) holds(dhap,1,2) holds(f16bp,3,2) holds(g3p,1,2)
fires(reset_dhap_1,2) fires(src_f16bp_1,2) fires(t3,2) fires(t4,2) fires(t6,2)
holds(bpg13,2,3) holds(dhap,1,3) holds(f16bp,4,3) holds(g3p,1,3)
fires(reset_dhap_1,3) fires(src_f16bp_1,3) fires(t3,3) fires(t4,3) fires(t5b,3)
holds(bpg13,2,4) holds(dhap,2,4) holds(f16bp,5,4) holds(g3p,1,4)
fires(reset_dhap_1,4) fires(src_f16bp_1,4) fires(t3,4) fires(t4,4) fires(t5b,4)
holds(bpg13,2,5) holds(dhap,2,5) holds(f16bp,6,5) holds(g3p,1,5)
fires(reset_dhap_1,5) fires(src_f16bp_1,5) fires(t3,5) fires(t4,5) fires(t6,5)
Answer: 3
holds(bpg13,0,0) holds(dhap,0,0) holds(f16bp,0,0) holds(g3p,0,0)
fires(reset_dhap_1,0) fires(src_f16bp_1,0) fires(t3,0)
holds(bpg13,0,1) holds(dhap,0,1) holds(f16bp,2,1) holds(g3p,0,1)
fires(reset_dhap_1,1) fires(src_f16bp_1,1) fires(t3,1) fires(t4,1)
holds(bpg13,0,2) holds(dhap,1,2) holds(f16bp,3,2) holds(g3p,1,2)
fires(reset_dhap_1,2) fires(src_f16bp_1,2) fires(t3,2) fires(t4,2) fires(t6,2)
holds(bpg13,2,3) holds(dhap,1,3) holds(f16bp,4,3) holds(g3p,1,3)
fires(reset_dhap_1,3) fires(src_f16bp_1,3) fires(t3,3) fires(t4,3) fires(t5b,3)
holds(bpg13,2,4) holds(dhap,2,4) holds(f16bp,5,4) holds(g3p,1,4)
fires(reset_dhap_1,4) fires(src_f16bp_1,4) fires(t3,4) fires(t4,4) fires(t6,4)
holds(bpg13,4,5) holds(dhap,1,5) holds(f16bp,6,5) holds(g3p,1,5)
fires(reset_dhap_1,5) fires(src_f16bp_1,5) fires(t3,5) fires(t4,5) fires(t5b,5)
Answer: 4
holds(bpg13,0,0) holds(dhap,0,0) holds(f16bp,0,0) holds(g3p,0,0)
fires(reset_dhap_1,0) fires(src_f16bp_1,0) fires(t3,0)
holds(bpg13,0,1) holds(dhap,0,1) holds(f16bp,2,1) holds(g3p,0,1)
fires(reset_dhap_1,1) fires(src_f16bp_1,1) fires(t3,1) fires(t4,1)
holds(bpg13,0,2) holds(dhap,1,2) holds(f16bp,3,2) holds(g3p,1,2)
fires(reset_dhap_1,2) fires(src_f16bp_1,2) fires(t3,2) fires(t4,2) fires(t6,2)
holds(bpg13,2,3) holds(dhap,1,3) holds(f16bp,4,3) holds(g3p,1,3)
fires(reset_dhap_1,3) fires(src_f16bp_1,3) fires(t3,3) fires(t4,3) fires(t5b,3)
holds(bpg13,2,4) holds(dhap,2,4) holds(f16bp,5,4) holds(g3p,1,4)
fires(reset_dhap_1,4) fires(src_f16bp_1,4) fires(t3,4) fires(t4,4) fires(t6,4)
holds(bpg13,4,5) holds(dhap,1,5) holds(f16bp,6,5) holds(g3p,1,5)
fires(reset_dhap_1,5) fires(src_f16bp_1,5) fires(t3,5) fires(t4,5) fires(t6,5)
Answer: 5
holds(bpg13,0,0) holds(dhap,0,0) holds(f16bp,0,0) holds(g3p,0,0)
fires(reset_dhap_1,0) fires(src_f16bp_1,0) fires(t3,0)
holds(bpg13,0,1) holds(dhap,0,1) holds(f16bp,2,1) holds(g3p,0,1)
fires(reset_dhap_1,1) fires(src_f16bp_1,1) fires(t3,1) fires(t4,1)
holds(bpg13,0,2) holds(dhap,1,2) holds(f16bp,3,2) holds(g3p,1,2)
fires(reset_dhap_1,2) fires(src_f16bp_1,2) fires(t3,2) fires(t4,2) fires(t5b,2)
holds(bpg13,0,3) holds(dhap,2,3) holds(f16bp,4,3) holds(g3p,1,3)
fires(reset_dhap_1,3) fires(src_f16bp_1,3) fires(t3,3) fires(t4,3) fires(t5b,3)
holds(bpg13,0,4) holds(dhap,2,4) holds(f16bp,5,4) holds(g3p,1,4)
fires(reset_dhap_1,4) fires(src_f16bp_1,4) fires(t3,4) fires(t4,4) fires(t5b,4)
holds(bpg13,0,5) holds(dhap,2,5) holds(f16bp,6,5) holds(g3p,1,5)
fires(reset_dhap_1,5) fires(src_f16bp_1,5) fires(t3,5) fires(t4,5) fires(t5b,5)
Answer: 6
holds(bpg13,0,0) holds(dhap,0,0) holds(f16bp,0,0) holds(g3p,0,0)
fires(reset_dhap_1,0) fires(src_f16bp_1,0) fires(t3,0)
holds(bpg13,0,1) holds(dhap,0,1) holds(f16bp,2,1) holds(g3p,0,1)
fires(reset_dhap_1,1) fires(src_f16bp_1,1) fires(t3,1) fires(t4,1)
holds(bpg13,0,2) holds(dhap,1,2) holds(f16bp,3,2) holds(g3p,1,2)
fires(reset_dhap_1,2) fires(src_f16bp_1,2) fires(t3,2) fires(t4,2) fires(t5b,2)
holds(bpg13,0,3) holds(dhap,2,3) holds(f16bp,4,3) holds(g3p,1,3)
fires(reset_dhap_1,3) fires(src_f16bp_1,3) fires(t3,3) fires(t4,3) fires(t5b,3)
holds(bpg13,0,4) holds(dhap,2,4) holds(f16bp,5,4) holds(g3p,1,4)
fires(reset_dhap_1,4) fires(src_f16bp_1,4) fires(t3,4) fires(t4,4) fires(t5b,4)
holds(bpg13,0,5) holds(dhap,2,5) holds(f16bp,6,5) holds(g3p,1,5)
fires(reset_dhap_1,5) fires(src_f16bp_1,5) fires(t3,5) fires(t4,5) fires(t6,5)
Answer: 7
holds(bpg13,0,0) holds(dhap,0,0) holds(f16bp,0,0) holds(g3p,0,0)
fires(reset_dhap_1,0) fires(src_f16bp_1,0) fires(t3,0)
holds(bpg13,0,1) holds(dhap,0,1) holds(f16bp,2,1) holds(g3p,0,1)
fires(reset_dhap_1,1) fires(src_f16bp_1,1) fires(t3,1) fires(t4,1)
holds(bpg13,0,2) holds(dhap,1,2) holds(f16bp,3,2) holds(g3p,1,2)
fires(reset_dhap_1,2) fires(src_f16bp_1,2) fires(t3,2) fires(t4,2) fires(t5b,2)
holds(bpg13,0,3) holds(dhap,2,3) holds(f16bp,4,3) holds(g3p,1,3)
fires(reset_dhap_1,3) fires(src_f16bp_1,3) fires(t3,3) fires(t4,3) fires(t5b,3)
holds(bpg13,0,4) holds(dhap,2,4) holds(f16bp,5,4) holds(g3p,1,4)
fires(reset_dhap_1,4) fires(src_f16bp_1,4) fires(t3,4) fires(t4,4) fires(t6,4)
holds(bpg13,2,5) holds(dhap,1,5) holds(f16bp,6,5) holds(g3p,1,5)
fires(reset_dhap_1,5) fires(src_f16bp_1,5) fires(t3,5) fires(t4,5) fires(t5b,5)
Answer: 8
holds(bpg13,0,0) holds(dhap,0,0) holds(f16bp,0,0) holds(g3p,0,0)
fires(reset_dhap_1,0) fires(src_f16bp_1,0) fires(t3,0)
holds(bpg13,0,1) holds(dhap,0,1) holds(f16bp,2,1) holds(g3p,0,1)
fires(reset_dhap_1,1) fires(src_f16bp_1,1) fires(t3,1) fires(t4,1)
holds(bpg13,0,2) holds(dhap,1,2) holds(f16bp,3,2) holds(g3p,1,2)
fires(reset_dhap_1,2) fires(src_f16bp_1,2) fires(t3,2) fires(t4,2) fires(t5b,2)
holds(bpg13,0,3) holds(dhap,2,3) holds(f16bp,4,3) holds(g3p,1,3)
fires(reset_dhap_1,3) fires(src_f16bp_1,3) fires(t3,3) fires(t4,3) fires(t5b,3)
holds(bpg13,0,4) holds(dhap,2,4) holds(f16bp,5,4) holds(g3p,1,4)
fires(reset_dhap_1,4) fires(src_f16bp_1,4) fires(t3,4) fires(t4,4) fires(t6,4)
holds(bpg13,2,5) holds(dhap,1,5) holds(f16bp,6,5) holds(g3p,1,5)
fires(reset_dhap_1,5) fires(src_f16bp_1,5) fires(t3,5) fires(t4,5) fires(t6,5)
Answer: 9
holds(bpg13,0,0) holds(dhap,0,0) holds(f16bp,0,0) holds(g3p,0,0)
fires(reset_dhap_1,0) fires(src_f16bp_1,0) fires(t3,0)
holds(bpg13,0,1) holds(dhap,0,1) holds(f16bp,2,1) holds(g3p,0,1)
fires(reset_dhap_1,1) fires(src_f16bp_1,1) fires(t3,1) fires(t4,1)
holds(bpg13,0,2) holds(dhap,1,2) holds(f16bp,3,2) holds(g3p,1,2)
fires(reset_dhap_1,2) fires(src_f16bp_1,2) fires(t3,2) fires(t4,2) fires(t5b,2)
holds(bpg13,0,3) holds(dhap,2,3) holds(f16bp,4,3) holds(g3p,1,3)
fires(reset_dhap_1,3) fires(src_f16bp_1,3) fires(t3,3) fires(t4,3) fires(t6,3)
holds(bpg13,2,4) holds(dhap,1,4) holds(f16bp,5,4) holds(g3p,1,4)
fires(reset_dhap_1,4) fires(src_f16bp_1,4) fires(t3,4) fires(t4,4) fires(t5b,4)
holds(bpg13,2,5) holds(dhap,2,5) holds(f16bp,6,5) holds(g3p,1,5)
fires(reset_dhap_1,5) fires(src_f16bp_1,5) fires(t3,5) fires(t4,5) fires(t5b,5)
Answer: 10
holds(bpg13,0,0) holds(dhap,0,0) holds(f16bp,0,0) holds(g3p,0,0)
fires(reset_dhap_1,0) fires(src_f16bp_1,0) fires(t3,0)
holds(bpg13,0,1) holds(dhap,0,1) holds(f16bp,2,1) holds(g3p,0,1)
fires(reset_dhap_1,1) fires(src_f16bp_1,1) fires(t3,1) fires(t4,1)
holds(bpg13,0,2) holds(dhap,1,2) holds(f16bp,3,2) holds(g3p,1,2)
fires(reset_dhap_1,2) fires(src_f16bp_1,2) fires(t3,2) fires(t4,2) fires(t5b,2)
holds(bpg13,0,3) holds(dhap,2,3) holds(f16bp,4,3) holds(g3p,1,3)
fires(reset_dhap_1,3) fires(src_f16bp_1,3) fires(t3,3) fires(t4,3) fires(t6,3)
holds(bpg13,2,4) holds(dhap,1,4) holds(f16bp,5,4) holds(g3p,1,4)
fires(reset_dhap_1,4) fires(src_f16bp_1,4) fires(t3,4) fires(t4,4) fires(t5b,4)
holds(bpg13,2,5) holds(dhap,2,5) holds(f16bp,6,5) holds(g3p,1,5)
fires(reset_dhap_1,5) fires(src_f16bp_1,5) fires(t3,5) fires(t4,5) fires(t6,5)
Answer: 11
holds(bpg13,0,0) holds(dhap,0,0) holds(f16bp,0,0) holds(g3p,0,0)
fires(reset_dhap_1,0) fires(src_f16bp_1,0) fires(t3,0)
holds(bpg13,0,1) holds(dhap,0,1) holds(f16bp,2,1) holds(g3p,0,1)
fires(reset_dhap_1,1) fires(src_f16bp_1,1) fires(t3,1) fires(t4,1)
holds(bpg13,0,2) holds(dhap,1,2) holds(f16bp,3,2) holds(g3p,1,2)
fires(reset_dhap_1,2) fires(src_f16bp_1,2) fires(t3,2) fires(t4,2) fires(t5b,2)
holds(bpg13,0,3) holds(dhap,2,3) holds(f16bp,4,3) holds(g3p,1,3)
fires(reset_dhap_1,3) fires(src_f16bp_1,3) fires(t3,3) fires(t4,3) fires(t6,3)
holds(bpg13,2,4) holds(dhap,1,4) holds(f16bp,5,4) holds(g3p,1,4)
fires(reset_dhap_1,4) fires(src_f16bp_1,4) fires(t3,4) fires(t4,4) fires(t6,4)
holds(bpg13,4,5) holds(dhap,1,5) holds(f16bp,6,5) holds(g3p,1,5)
fires(reset_dhap_1,5) fires(src_f16bp_1,5) fires(t3,5) fires(t4,5) fires(t5b,5)
Answer: 12
holds(bpg13,0,0) holds(dhap,0,0) holds(f16bp,0,0) holds(g3p,0,0)
fires(reset_dhap_1,0) fires(src_f16bp_1,0) fires(t3,0)
holds(bpg13,0,1) holds(dhap,0,1) holds(f16bp,2,1) holds(g3p,0,1)
fires(reset_dhap_1,1) fires(src_f16bp_1,1) fires(t3,1) fires(t4,1)
holds(bpg13,0,2) holds(dhap,1,2) holds(f16bp,3,2) holds(g3p,1,2)
fires(reset_dhap_1,2) fires(src_f16bp_1,2) fires(t3,2) fires(t4,2) fires(t5b,2)
holds(bpg13,0,3) holds(dhap,2,3) holds(f16bp,4,3) holds(g3p,1,3)
fires(reset_dhap_1,3) fires(src_f16bp_1,3) fires(t3,3) fires(t4,3) fires(t6,3)
holds(bpg13,2,4) holds(dhap,1,4) holds(f16bp,5,4) holds(g3p,1,4)
fires(reset_dhap_1,4) fires(src_f16bp_1,4) fires(t3,4) fires(t4,4) fires(t6,4)
holds(bpg13,4,5) holds(dhap,1,5) holds(f16bp,6,5) holds(g3p,1,5)
fires(reset_dhap_1,5) fires(src_f16bp_1,5) fires(t3,5) fires(t4,5) fires(t6,5)
Answer: 13
holds(bpg13,0,0) holds(dhap,0,0) holds(f16bp,0,0) holds(g3p,0,0)
fires(reset_dhap_1,0) fires(src_f16bp_1,0) fires(t3,0)
holds(bpg13,0,1) holds(dhap,0,1) holds(f16bp,2,1) holds(g3p,0,1)
fires(reset_dhap_1,1) fires(src_f16bp_1,1) fires(t3,1) fires(t4,1)
holds(bpg13,0,2) holds(dhap,1,2) holds(f16bp,3,2) holds(g3p,1,2)
fires(reset_dhap_1,2) fires(src_f16bp_1,2) fires(t3,2) fires(t4,2) fires(t6,2)
holds(bpg13,2,3) holds(dhap,1,3) holds(f16bp,4,3) holds(g3p,1,3)
fires(reset_dhap_1,3) fires(src_f16bp_1,3) fires(t3,3) fires(t4,3) fires(t6,3)
holds(bpg13,4,4) holds(dhap,1,4) holds(f16bp,5,4) holds(g3p,1,4)
fires(reset_dhap_1,4) fires(src_f16bp_1,4) fires(t3,4) fires(t4,4) fires(t5b,4)
holds(bpg13,4,5) holds(dhap,2,5) holds(f16bp,6,5) holds(g3p,1,5)
fires(reset_dhap_1,5) fires(src_f16bp_1,5) fires(t3,5) fires(t4,5) fires(t6,5)
Answer: 14
holds(bpg13,0,0) holds(dhap,0,0) holds(f16bp,0,0) holds(g3p,0,0)
fires(reset_dhap_1,0) fires(src_f16bp_1,0) fires(t3,0)
holds(bpg13,0,1) holds(dhap,0,1) holds(f16bp,2,1) holds(g3p,0,1)
fires(reset_dhap_1,1) fires(src_f16bp_1,1) fires(t3,1) fires(t4,1)
holds(bpg13,0,2) holds(dhap,1,2) holds(f16bp,3,2) holds(g3p,1,2)
fires(reset_dhap_1,2) fires(src_f16bp_1,2) fires(t3,2) fires(t4,2) fires(t6,2)
holds(bpg13,2,3) holds(dhap,1,3) holds(f16bp,4,3) holds(g3p,1,3)
fires(reset_dhap_1,3) fires(src_f16bp_1,3) fires(t3,3) fires(t4,3) fires(t6,3)
holds(bpg13,4,4) holds(dhap,1,4) holds(f16bp,5,4) holds(g3p,1,4)
fires(reset_dhap_1,4) fires(src_f16bp_1,4) fires(t3,4) fires(t4,4) fires(t5b,4)
holds(bpg13,4,5) holds(dhap,2,5) holds(f16bp,6,5) holds(g3p,1,5)
fires(reset_dhap_1,5) fires(src_f16bp_1,5) fires(t3,5) fires(t4,5) fires(t5b,5)
Answer: 15
holds(bpg13,0,0) holds(dhap,0,0) holds(f16bp,0,0) holds(g3p,0,0)
fires(reset_dhap_1,0) fires(src_f16bp_1,0) fires(t3,0)
holds(bpg13,0,1) holds(dhap,0,1) holds(f16bp,2,1) holds(g3p,0,1)
fires(reset_dhap_1,1) fires(src_f16bp_1,1) fires(t3,1) fires(t4,1)
holds(bpg13,0,2) holds(dhap,1,2) holds(f16bp,3,2) holds(g3p,1,2)
fires(reset_dhap_1,2) fires(src_f16bp_1,2) fires(t3,2) fires(t4,2) fires(t6,2)
holds(bpg13,2,3) holds(dhap,1,3) holds(f16bp,4,3) holds(g3p,1,3)
fires(reset_dhap_1,3) fires(src_f16bp_1,3) fires(t3,3) fires(t4,3) fires(t6,3)
holds(bpg13,4,4) holds(dhap,1,4) holds(f16bp,5,4) holds(g3p,1,4)
fires(reset_dhap_1,4) fires(src_f16bp_1,4) fires(t3,4) fires(t4,4) fires(t6,4)
holds(bpg13,6,5) holds(dhap,1,5) holds(f16bp,6,5) holds(g3p,1,5)
fires(reset_dhap_1,5) fires(src_f16bp_1,5) fires(t3,5) fires(t4,5) fires(t5b,5)
Answer: 16
holds(bpg13,0,0) holds(dhap,0,0) holds(f16bp,0,0) holds(g3p,0,0)
fires(reset_dhap_1,0) fires(src_f16bp_1,0) fires(t3,0)
holds(bpg13,0,1) holds(dhap,0,1) holds(f16bp,2,1) holds(g3p,0,1)
fires(reset_dhap_1,1) fires(src_f16bp_1,1) fires(t3,1) fires(t4,1)
holds(bpg13,0,2) holds(dhap,1,2) holds(f16bp,3,2) holds(g3p,1,2)
fires(reset_dhap_1,2) fires(src_f16bp_1,2) fires(t3,2) fires(t4,2) fires(t6,2)
holds(bpg13,2,3) holds(dhap,1,3) holds(f16bp,4,3) holds(g3p,1,3)
fires(reset_dhap_1,3) fires(src_f16bp_1,3) fires(t3,3) fires(t4,3) fires(t6,3)
holds(bpg13,4,4) holds(dhap,1,4) holds(f16bp,5,4) holds(g3p,1,4)
fires(reset_dhap_1,4) fires(src_f16bp_1,4) fires(t3,4) fires(t4,4) fires(t6,4)
holds(bpg13,6,5) holds(dhap,1,5) holds(f16bp,6,5) holds(g3p,1,5)
fires(reset_dhap_1,5) fires(src_f16bp_1,5) fires(t3,5) fires(t4,5) fires(t6,5)
\end{verbatim}
}

Atoms selected for $bpg13$ quantity extraction for the nominal case:
{\footnotesize
\begin{verbatim}
[holds(bpg13,0,0),holds(bpg13,0,1),holds(bpg13,0,2),
	holds(bpg13,0,3),holds(bpg13,2,4),holds(bpg13,4,5)], 
[holds(bpg13,0,0),holds(bpg13,0,1),holds(bpg13,0,2),
	holds(bpg13,2,3),holds(bpg13,4,4),holds(bpg13,6,5)]
\end{verbatim}
}

Atoms selected for $bpg13$ quantity extraction for the modified case:
{\footnotesize
\begin{verbatim}
[holds(bpg13,0,0),holds(bpg13,0,1),holds(bpg13,0,2),
	holds(bpg13,2,3),holds(bpg13,2,4),holds(bpg13,2,5)], 
[holds(bpg13,0,0),holds(bpg13,0,1),holds(bpg13,0,2),
	holds(bpg13,2,3),holds(bpg13,2,4),holds(bpg13,2,5)], 
[holds(bpg13,0,0),holds(bpg13,0,1),holds(bpg13,0,2),
	holds(bpg13,2,3),holds(bpg13,2,4),holds(bpg13,4,5)], 
[holds(bpg13,0,0),holds(bpg13,0,1),holds(bpg13,0,2),
	holds(bpg13,2,3),holds(bpg13,2,4),holds(bpg13,4,5)], 
[holds(bpg13,0,0),holds(bpg13,0,1),holds(bpg13,0,2),
	holds(bpg13,0,3),holds(bpg13,0,4),holds(bpg13,0,5)], 
[holds(bpg13,0,0),holds(bpg13,0,1),holds(bpg13,0,2),
	holds(bpg13,0,3),holds(bpg13,0,4),holds(bpg13,0,5)], 
[holds(bpg13,0,0),holds(bpg13,0,1),holds(bpg13,0,2),
	holds(bpg13,0,3),holds(bpg13,0,4),holds(bpg13,2,5)], 
[holds(bpg13,0,0),holds(bpg13,0,1),holds(bpg13,0,2),
	holds(bpg13,0,3),holds(bpg13,0,4),holds(bpg13,2,5)], 
[holds(bpg13,0,0),holds(bpg13,0,1),holds(bpg13,0,2),
	holds(bpg13,0,3),holds(bpg13,2,4),holds(bpg13,2,5)], 
[holds(bpg13,0,0),holds(bpg13,0,1),holds(bpg13,0,2),
	holds(bpg13,0,3),holds(bpg13,2,4),holds(bpg13,2,5)], 
[holds(bpg13,0,0),holds(bpg13,0,1),holds(bpg13,0,2),
	holds(bpg13,0,3),holds(bpg13,2,4),holds(bpg13,4,5)], 
[holds(bpg13,0,0),holds(bpg13,0,1),holds(bpg13,0,2),
	holds(bpg13,0,3),holds(bpg13,2,4),holds(bpg13,4,5)], 
[holds(bpg13,0,0),holds(bpg13,0,1),holds(bpg13,0,2),
	holds(bpg13,2,3),holds(bpg13,4,4),holds(bpg13,4,5)], 
[holds(bpg13,0,0),holds(bpg13,0,1),holds(bpg13,0,2),
	holds(bpg13,2,3),holds(bpg13,4,4),holds(bpg13,4,5)], 
[holds(bpg13,0,0),holds(bpg13,0,1),holds(bpg13,0,2),
	holds(bpg13,2,3),holds(bpg13,4,4),holds(bpg13,6,5)], 
[holds(bpg13,0,0),holds(bpg13,0,1),holds(bpg13,0,2),
	holds(bpg13,2,3),holds(bpg13,4,4),holds(bpg13,6,5)]
\end{verbatim}
}

Progression from raw matrix of $bpg13$ quantities in various answer-sets (rows) at various simulation steps (columns) to rates and finally to the average aggregate for the nominal case:

$
\begin{bmatrix}
0 & 0 & 0 & 0 & 2 & 4\\
0 & 0 & 0 & 2 & 4 & 6\\
\end{bmatrix}
=rate=>
\begin{bmatrix}
0.8\\
1.2\\
\end{bmatrix}
=average=>
1.0
$

Progression from raw matrix of $bpg13$ quantities in various answer-sets (rows) at various simulation steps (columns) to rates and finally to the average aggregate for the modified case:

$
\begin{bmatrix}
0 & 0 & 0 & 2 & 2 & 2\\
0 & 0 & 0 & 2 & 2 & 2\\
0 & 0 & 0 & 2 & 2 & 4\\
0 & 0 & 0 & 2 & 2 & 4\\
0 & 0 & 0 & 0 & 0 & 0\\
0 & 0 & 0 & 0 & 0 & 0\\
0 & 0 & 0 & 0 & 0 & 2\\
0 & 0 & 0 & 0 & 0 & 2\\
0 & 0 & 0 & 0 & 2 & 2\\
0 & 0 & 0 & 0 & 2 & 2\\
0 & 0 & 0 & 0 & 2 & 4\\
0 & 0 & 0 & 0 & 2 & 4\\
0 & 0 & 0 & 2 & 4 & 4\\
0 & 0 & 0 & 2 & 4 & 4\\
0 & 0 & 0 & 2 & 4 & 6\\
0 & 0 & 0 & 2 & 4 & 6\\
\end{bmatrix}
=rate=>
\begin{bmatrix}
0.4\\
0.4\\  
0.8  \\
0.8  \\
0.   \\
0.   \\
0.4  \\
0.4  \\
0.4  \\
0.4  \\
0.8  \\
0.8  \\
0.8  \\
0.8  \\
1.2 \\
1.2\\
\end{bmatrix}
=average=>
0.6
$

We find that $d='<'$ comparing the modified case rate of $0.6$ to the nominal case rate of $1.0$. Since the direction $d$ was an unknown in the query statement, our system generates produces the full query specification with $d$ replaced by $'<'$ as follows:
{\footnotesize
\begin{verbatim}
direction of change in average  rate of production of 'bpg13' is '<' (0.6<1) 
    when observed between time step 0 and time step 5
    comparing nominal pathway with modified pathway obtained;
    due to interventions:
        remove 'dhap' as soon as produced;
    using initial setup:
        continuously supply 'f16bp' in quantity 1;
\end{verbatim}
}

\subsection{Evaluation Methodolgy}
A direct comparison against other tools is not possible, since most other programs explore one state evolution, while we explore all possible state evolutions.  In addition ASP has to ground the program completely, irrespective of whether we are computing one answer or all. So, to evaluate our system, we compare our results for the questions from the 2nd Deep KR Challenge against the answers they have provided. Our results in essence match the responses given for the questions.

\section{Related Work}

In this section, we relate our high level language with other high level action languages.

\subsection{Comparison with $\pi$-Calculus}
$\pi$-calculus is a formalism that is used to model biological systems and pathways by modeling biological systems as mobile communication systems. We use the biological model described by~\cite{regev2001representation} for comparison against our system. In their model they represent molecules and their domains as computational processes, interacting elements of molecules as communication channels (two molecules interact if they fit together as in a lock-and-key mechanism), and reactions as communication through channel transmission.
\begin{itemize}
\item $\pi$-calculus models have the ability of changing their structure during simulation. Our system on the other hand only allows modification of the pathway at the start of simulation.
\item Regular $\pi$-calculus models appear qualitative in nature. However, stochastic extensions allow representation of quantitative data~\cite{priami2001application}. In contrast, the focus of our system is on the quantitative+qualitative representation using numeric fluents.
\item It is unclear how one can easily implement maximal-parallelism of our system in $\pi$-calculus, where a maximum number of simultaneous actions occur such that they do not cause a conflict. Where, a set of actions is said to be in conflict if their simultaneous execution will cause a fluent to become negative.
\end{itemize}

\subsection{Comparison with Action Language $\mathcal{A}$}

Action language $\mathcal{A}$~\cite{gelfond1993representing} is a formalism that has been used to model biological systems and pathways. First we give a brief overview of $\mathcal{A}$ in an intuitive manner. Assume two sets of disjoint symbols containing fluent names and action names, then a {\em fluent expression} is either a fluent name $F$ or $\lnot F$. A domain description is composed of propositions of the following form:
\begin{description}
\item {\em value proposition}: $F \mathbf{~after~} A_1;\dots;A_m$, where $(m \geq 0)$, $F$ is a fluent and $A_1,\dots,A_m$ are fluents.
\item {\em effect propostion}: $A \mathbf{~causes~} F \mathbf{~if~} P_1,\dots,P_n$, where $(n \geq 0)$, $A$ is an action, $F,P_1,\dots,P_n$ are fluent expressions. $P_1,\dots,P_n$ are called preconditions of $A$ and the effect proposition describes the effect on $F$.
\end{description}

We relate it to our work:
\begin{itemize}
\item Fluents are boolean. We support numeric valued fluents, with binary fluents.
\item Fluents are non-inertial, but inertia can be added. Our fluents are always intertial.
\item Action description specifies the effect of an action. Our domain description specifies `natural'-actions, which execute automatically when their pre-conditions are satisfied, subject to certain conditions. As a result our domain description represents trajectories.
\item No built in support for aggregates exists. We support a selected set of aggregates, on a single trajectory and over multiple trajectories. 
\item Value propositions in $\mathcal{A}$ are representable as observations in our query language.
\end{itemize}

\subsection{Comparison with Action Language $\mathcal{B}$}

Action language $\mathcal{B}$ extends $\mathcal{A}$ by adding static causal laws, which allows one to specify indirect effects or ramifications of an action~\cite{gelfond2012common}. We relate it  to our work below:
\begin{itemize}
\item Inertia is built into the semantics of $\mathcal{B}$~\cite{gelfond2012common}. Our language also has intertia built in.
\item $\mathcal{B}$ supports static causal laws that allow defining a fluent in terms of other fluents. We do not support static causal laws.
\end{itemize}

\subsection{Comparison with Action Language $\mathcal{C}$}

Action language $\mathcal{C}$ is based on the theory of causal explanation, i.e. a formula is true if there is a cause for it to be true~\cite{giunchiglia1998action}. It has been previously used to represent and reason about biological pathways~\cite{dworschak2008tools}. We relate it to our work below:
\begin{itemize}
\item $\mathcal{C}$ supports boolean fluents only. We support numeric valued fluents, and binary fluents.
\item $\mathcal{C}$ allows both inertial and non-inertial fluents. While our fluents are always inertial.
\item $\mathcal{C}$ support static causal laws (or ramifications), that allow defining a fluent in terms of other fluents. We do not support them.
\item $\mathcal{C}$ describes causal relationships between fluents and actions. Our language on the other hand describes trajectories.
\end{itemize}

\subsection{Comparison with Action Language $\mathcal{C+}$}

First, we give a brief overview of $\mathcal{C+}$~\cite{giunchiglia2004nonmonotonic}.
Intuitively, atoms $\mathcal{C+}$ are of the form $c=v$, where $c$ is either a fluent or an action constant, $v$ belongs to the domain of $c$, and fluents and actions form a disjoint set. A formula is a propositional combination of atoms. A fluent formula is a formula in which all constants are fluent constants; and an action formula is a formula with one action constant but no fluent constants.
An action description in $\mathcal{C+}$ is composed of causal laws of the following forms:
\begin{description}
\item {\em static law}: $\mathbf{caused~} F \mathbf{~if~} G$, where $F$ and $G$ are fluent formulas
\item {\em action dynamic law}: $\mathbf{caused~} F \mathbf{~if~} G$, where $F$ is an action formula and $G$ is a formula
\item {\em fluent dynamic law}: $\mathbf{caused~} F \mathbf{~if~} G \mathbf{~after~} H$, where $F$ and $G$ are fluent formulas and $H$ is a formula
\end{description}
Concise forms of these laws exist, e.g. `$\mathbf{intertial~} f \equiv \mathbf{caused~} f=v \mathbf{~if~} f=v \mathbf{~after~} f=v, \forall v \in \text{ domain of } f$' that allow a more intuitive program declaration.

We now relate it to our work.

\begin{itemize}
\item Some of the main differences include:
\begin{itemize}
\item Fluents are multi-valued, fluent values can be integer, boolean or other types. We support integer and binary valued fluents only.
\item Actions are multi-valued. We do not support multi-valued actions.
\item Both inertial and non-inertial fluents are supported. In comparison we allow inertial fluents only.
\item Static causal laws are supported that allow changing the value of a fluent based on other fluents (ramifications). We do not allow static causal laws.
\item Effect of parallel actions on numeric fluents is not additive. However, the additive fluents extension~\cite{lee2003describing} adds the capability of additive fluents through new rules. The extended language, however, imposes certain restrictions on additive fluents and also restricts the domain of additive actions to boolean actions only. Our fluents are always additive.
\item Supports defaults. We do not have the same notion as defaults, but allow initial values for fluents in our domain description.
\item Action's occurrence and its effect are defined in separate statements. In our case, the action's occurrence and effect are generally combined in one statement.
\end{itemize}

\item Although parallel actions are supported, it is unclear how one can concisely describe the condition implicit in our system that simultaneously occurring actions may not conflict. Two actions conflict if their simultaneous execution will cause a fluent to become negative.

\item Exogenous actions seem the closest match to our {\em may~execute~} actions. However, our actions are `natural', in that they execute automatically when their pre-conditions are satisfied, they are not explicitly inhibited, and they do not conflict. Actions conflict when their simultaneous execution will cause one of the fluents to become negative. The exogenous-style character of our actions holds when the {\em firing style} is `$*$'. When the firing style changes, the execution semantics change as well. Consider the following two {\em may execute} statements in our language:
{\footnotesize
\begin{flalign*}
a_1 \mathbf{~may~execute~causing~} f_1 \mathbf{~change~value~by~} -5 \mathbf{~if~} f_2 \mathbf{~has~value~} 3 \mathbf{~or~higher} \\
a_2 \mathbf{~may~execute~causing~} f_1 \mathbf{~change~value~by~} -3 \mathbf{~if~} f_2 \mathbf{~has~value~} 2 \mathbf{~or~higher} 
\end{flalign*}
}
and two states: (i) $f_1=10, f_2=5$, (ii) $f_1=6,f_2=5$. In state (i) both $a_1,a_2$ can occur simultaneously (at one point) resulting in firing-choices $\big\{\{a_1,a_2 \},\{a_1\},\{a_2\},\{\},\big\}$; whereas, in state (ii) only one of $a_1$ or $a_2$ can occur at one point resulting in the firing-choices: $\big\{\{a_1\},\{a_2\},\{\},\big\}$ because of a conflict due to the limited amount of $f_1$. These firing choices apply for {\em firing style} `*', which allows any combination of fireable actions to occur.  If the {\em firing style} is set to `max', the maximum set of non-conflicting actions may fire, and the firing choices for state (i) change to $\big\{\{a_1,a_2 \}\big\}$ and the firing choices for state (ii) change to $\big\{\{a_1\},\{a_2\}\big\}$. If the {\em firing style} is set to `1', at most one action may fire at one point, and the firing choices for both state (i) and state (ii) reduce to $\big\{\{a_1\},\{a_2\},\{\}\big\}$. So, the case with `*' firing style can be represented in $\mathcal{C+}$ with exogenous actions $a_1,a_2$; the case with `1' firing style can be represented in $\mathcal{C+}$ with exogenous actions $a_1,a_2$ and a constraint requiring that both $a_1,a_2$ do not occur simultaneously; while the case with `max' firing style can be represented by exogenous actions $a_1,a_2$ with additional action dynamic laws. They will still be subject to the conflict checking.
Action dynamic laws can be used to force actions similar to our {\em must~execute~} actions. 

\item Specification of initial values of fluents seem possible through the query language. The {\em default} statement comes close, but it does not have the same notion as setting a fluent to a value once. We support specifying initial values both in the domain description as well as the query. 

\item There does not appear built-in support for aggregation of fluent values within the same answer set, such as sum, count, rate, minimum, maximum etc. Although some of it could be implemented using the additive fluents extension. We support a set of aggregates, such as total, and rate. Additional aggregates can be easily added.

\item We support queries over aggregates (such as minimum, maximum, average) of single-trajectory aggregates (such as total, and rate etc.) over a set of trajectories. We also support comparative queries over two sets of trajectories. Our queries allow modification of the domain description as part of query evaluation.

\end{itemize}

\subsection{Comparison with $\mathcal{BC}$}
Action language $\mathcal{BC}$ combines features of $\mathcal{B}$ and $\mathcal{C+}$~\cite{lee2013action}. First we give a brief overview of $\mathcal{BC}$.

Intuitively, $\mathcal{BC}$ has actions and valued fluents. A valued fluent, called an atom, is of the form `$f=v$', where $f$ is a fluent constant and $v \in domain(f)$. A fluent can be regular or statically determined. An action description in $\mathcal{BC}$ is composed of static and dynamic laws of the following form:
\begin{description}
\item {\em static law}: $A_0 \mathbf{~if~} A_1,\dots,A_m \mathbf{~ifcons~} A_{m+1},\dots,A_n$, where $(n \geq m \geq 0)$, and each $A_i$ is an atom.
\item {\em dynamic law}: $A_0 \mathbf{~after~} A_1,\dots,A_m \mathbf{~ifcons~} A_{m+1},\dots,A_n$, where $(n \geq m \geq 0)$, $A_0$ is a regular fluent atom, each of $A_1,\dots,A_m$ is an atom or an action constant, and $A_{m+1},\dots,A_n$ are atoms.
\end{description}
Concise forms of these laws exist that allow a more intuitive program declaration.

Now we relate it to our work.
\begin{itemize}
\item Some of the main differences include:
\begin{itemize}
\item Fluents are multi-valued, fluent values can which can be integer, boolean, or other types. We only support integer and binary fluents.
\item Static causal laws are allowed. We do not support static causal laws.
\item Similar to $\mathcal{C+}$ numeric fluent accumulation is not supported. It is supported in our system.
\item It is unclear how aggregate queries within a trajectory can be concisely represented. Aggregate queries such as rate are supported in our system.
\item It does not seem that queries over multiple trajectories or sets of trajectories are supported. Such queries are supported in our system.
\end{itemize}
\end{itemize}

\subsection{Comparison with ASPMT}
ASPMT combines ASP with Satisfiability Modulo Theories. We relate the work in~\cite{lee2013answer} where $\mathcal{C+}$ is extended using ASPMT with our work. 
\begin{itemize}
\item It adds support for real valued fluents to $\mathcal{C+}$ including additive fluents. Thus, it allows reasoning with continuous and discrete processes simultaneously. Our language does not support real numbers directly.
\end{itemize}

Several systems also exist to model and reason with biological pathway. For example:

\subsection{Comparison with BioSigNet-RR}
BioSigNet-RR~\cite{baral2004knowledge} is a system for representing and reasoning with signaling networks. We relate it to our work in an intuitive manner.
\begin{itemize}
\item Fluents are boolean, so qualitative queries are possible. We support both integer and binary fluents, so quantiative queries are also possible.
\item Indirect effects (or ramifications) are supported. We do not support these.
\item Action effects are captured separately in `$\mathbf{~causes~}$ statement' from action triggering statements `$\mathbf{~triggers~}$' and `$\mathbf{~n\_triggers~}$'. We capture both components in a `$\mathbf{~may~execute~causing~}$' or `$\mathbf{~normally~must~execute~causing~}$' statement.
\item Their action triggering semantics have some similarity to our actions. Just like their actions get triggered when the action's pre-conditions are satisfied, our actions are also triggered when their pre-conditions are satisfied. However, the triggering semantics are different, e.g. their {\bf triggers} statement causes an action to occur even if it is disabled, we do not have an equivalent for it; and their {\bf n\_triggers} is similar in semantics to {\em normally~must~execute~causing} statement. 
\item It is not clear how loops in biological systems can be modeled in their system. Loops are possible in our by virtue of the Petri Net semantics. 
\item Their queries can have time-points and their precedence relations as part of the query. Though our queries allow the specification of some time points for interval queries, time-points are not supported in a similar way. However, we do support certain types of observation relative queries.
\item The intervention in their planning queries has similarities to interventions in our system. However, it appears that our intervention descriptions are higher level. 
\end{itemize}

\section{Conclusion}
In this chapter we presented the BioPathQA system and the languages to represent and query biological pathways. We also presented a new type of Petri Net, the so called Guarded-Arc Petri Net that is used as a model behind our pathway specification language, which shares certain aspects with CPNs~\cite{jensen2007coloured}, but our semantics for reset arcs is different, and we allow must-fire actions that prioritize actions for firing over other actions. We also showed how the system can be applied to questions from college level text books that require deeper reasoning and cannot be answered by using surface level knowledge. Although our system is developed with respect to the biological domain, it can be applied to non-biological domain as well.%

Some of the features of our language include: natural-actions that automatically fire when their prerequisite conditions are met (subject to certain restrictions); an automatic default constraint that ensures fluents do not go negative, since they model natural systems substances;  a more natural representation of locations; and control of the level of parallelism to a certain degree. %
Our query language also allows interventions similar to Pearl's surgeries~\cite{pearl1995action}, which are more general than actions.

Next we want to apply BioPathQA to a real world application by doing text extraction. Knowledge for real world is extracted from research papers.  In the next chapter we show how such text extraction is done for pathway construction and drug development. We will then show how we can apply BioPathQA to the extracted knowledge to answer questions about the extracted knowledge. %
\chapter{Text Extraction for Real World Applications}\label{ch:prev_work}

In the previous chapter we looked at the BioPathQA system and how it answers simulation based reasoning questions about biological pathways, including questions that require comparison of alternate scenarios through simulation. These so called `what-if' questions arise in biological activities such as drug development, drug interaction, and personalized medicine. We will now put our system and language in context of such activities.

Cutting-edge knowledge about pathways for activities such as drug development, drug interaction, and personalized medicine comes in the form of natural language research papers, thousands of which are published each year. To use this knowledge with our system, we need to perform extraction. In this chapter we describe techniques we use for such knowledge extraction for discovering drug interactions. We illustrate with an example extraction how we organize the extracted knowledge into a pathway specification and give examples of relevant what-if questions that a researcher performing may ask in the drug development domain.

\section{Introduction}

Thousands of research papers are published each year about biological systems and pathways over a broad spectrum of activities, including interactions between dugs and diseases, the associated pathways, and genetic variation.  Thus, one has to perform text extraction to extract relationships between the biochemical processes, their involvement in diseases, and their interaction with drugs. For personalized medicine, one is also interested in how these interrelationships change in presence of genetic variation. In short, we are looking for relationships between various components of the biochemical processes and their internal and external stimuli. 

Many approaches exist for extracting relationships from a document. Most rely on some form of co-occurrence, relative distance, or order of  words in a single document. Some use shallow parsing as well. Although these techniques tend to have a higher recall, they focus on extracting explicit relationships, which are relationships that are fully captured in a sentence or a document. 
These techniques also do not capture implicit relationships that may be spread across multiple documents. are spread across multiple documents relating to different species.  
Additional issues arise from the level of detail from in older vs. newer texts and seemingly contradictory information due to various levels of confidence in the techniques used. Many do not handle negative statements.

We primarily use a system called PTQL~\cite{PTQL} to extract these relationships, which allows combining the syntactic structure (parse tree), semantic elements, and word order in a relationship query. The sentences are pre-processed by using named-entity recognition, and entity normalization to allow querying on classes of entity types, such as drugs, and diseases; and also to allow cross-linking relationships across documents when they refer to the same entity with a different name. Queries that use such semantic association between words/phrases are likely to produce higher precision results.

Source knowledge for extraction primarily comes from thousands of biological abstracts published each year in PubMed~\footnote{\texttt{http://www.ncbi.nlm.nih.gov/pubmed}}. 

Next we briefly describe how we extract relationships about drug interactions. Following that we briefly describe how we extract association of drugs, and diseases with genetic variation. We conclude this chapter with an illustrative example of how the drug interaction relationships are used with our system to answer questions about drug interactions and how genetic variation could be utilized in our system.

\section{Extracting Relationships about Drug Interactions}
\label{sec:ddi}
We summarize the extraction of relationships for our work on drug-drug interactions from~\cite{TariDDI2010}.

Studying drug-drug interactions are a major activity in drug development. Drug interactions occur due to the interactions between the biological processes / pathways that are responsible metabolizing and transporting drugs. Metabolic processes remove a drug from the system within a certain time period. For a drug to remain effective, it must be maintained within its therapeutic window for the period of treatment, requiring taking the drug periodically. Outside the therapeutic window, a drug can become toxic if a quantity greater than the therapeutic window is retained; or it can become ineffective if a quantity less than the therapeutic window is retained.

Since liver enzymes metabolize most drugs, it is the location where most metabolic-interaction takes place. Induction or inhibition of these enzymes can affect the bioavailability of a drug through transcriptional regulation, either directly or indirectly. For example, if drug $A$ inhibits enzyme $E$, which metabolizes drug $B$, then the bioavailability of drug $B$ will be higher than normal, rendering it toxic. On the other hand, if drug $A$ induces enzyme $E$, which metabolizes drug $B$, then drug $B$'s bioavailability will be lesser than normal, rendering it ineffective.

Inhibition of enzymes is a common form of drug-drug interactions~\cite{Boobis2009}. In direct inhibition, a drug $A$ inhibit enzyme $E$, which is responsible for metabolism of drug $B$. Drug $A$, leads to a decrease in the level of enzyme $E$, which in turn can increase bioavailability of drug $B$ potentially leading to toxicity. Alternatively, insufficient metabolism of drug $B$ can lead to smaller amount of drug $B$'s metabolites being produced, leading to therapeutic failure. An example of one such direct inhibition is the interaction between CYP2D6 inhibitor quinidine and CYP2D6 substrates (i.e. substances metabolized by CYP2D6), such as Codeine. The inhibition of CYP2D6 increases the bioavailability of drugs metabolized by CYP2D6 leading to adverse side effects.

Another form of drug interactions is through induction of enzymes~\cite{Boobis2009}. In direct induction, a drug $A$ induces enzyme $E$, which is responsible for metabolism of drug $B$. An example of such direct induction is between phenobarbital, a CYP2C9 inducer and warfarin (a CYP2C9 substrate). Phenobarbital leads to increased metabolism of warfarin, decreasing warfarinÕs bioavailability. Direct interaction due to induction though possible is not as common as indirect interaction through transcription factors, which regulate the drug metabolizing enzymes. In such an interaction, drug $A$ activates a transcription factor $TF$, which regulates and induces enzyme $E$, where enzyme $E$ metabolizes drug $B$. Transcription factors are referred to as regulators of xenobiotic-metabolizing enzymes. Examples of such regulators include aryl hydrocarbon receptor AhR, pregnane X receptor PXR and constitutive androstane receptor CAR.

Drug interactions can also occur due to the induction or inhibition of transporters. Transporters are mainly responsible for cellular uptake or efflux (elimination) of drugs. They play an important part in drug disposition, by transporting drugs into the liver cells, for example. Transporter-based drug interactions, however, are not as well studies as metabolism-based interactions~\cite{Boobis2009}.

\subsection{Method}
\begin{figure}[htbp]
\begin{center}
\includegraphics[width=6cm]{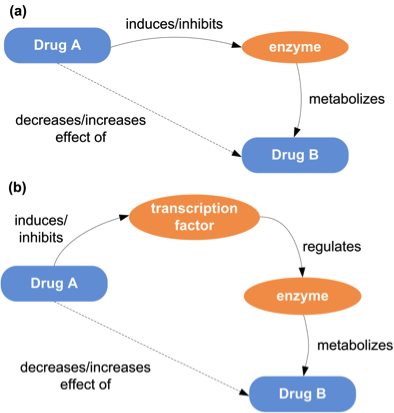}
\caption{This figure from \cite{TariDDI2010} outlines the effects of drug A on drug B through (a) direct induction/inhibition of enzymes; (b) indirect induction/inhibition of transportation factors that regulate the drug-metabolizing enzymes.}
\label{fig:ddi}
\end{center}
\end{figure}

Extraction of drug-drug interactions from the text can either be explicit or implicit. Explicit extraction refers to extraction of drug-drug interaction mentioned within a single sentence, while implicit extraction requires extraction of bio-properties of drug transport, metabolism and elimination that can lead to drug-drug interaction. This type of indirect extraction combines background information about biological processes, identification of protein families and the interactions that are involved in drug metabolism. Our approach is to extract both explicit and implicit drug interactions as summarized in Fig~\ref{fig:ddi} and it builds upon the work done in~\cite{tari2009querying}.

\subsubsection{Explicit Drug Interaction Extraction}
Explicit extraction mainly extracts drug-drug interactions directly mentioned in PubMed (or Medline) abstracts. For example, the following sentences each have a metabolic interaction mentioned within the same sentence:
\begin{enumerate}
\item[S1:] \textit{Ciprofloxacin} strongly \underline{inhibits} \textit{clozapine} \underline{metabolism}. (PMID: 19067475)
\item[S2:] Enantioselective \underline{induction} of \textit{cyclophosphamide} \underline{metabolism} by \textit{phenytoin}.
\end{enumerate}
which can be extracted by using the following PTQL query using the underlined keywords from above sentences:
\begin{quote}
{\footnotesize
\begin{verbatim}
//S{//?[Tag=`Drug'](d1) =>
   //?[Value IN {`induce',`induces',`inhibit',`inhibits'}](v) =>
   //?[Tag=`Drug'](d2) => //?[Value=`metabolism'](w)} :::
   [d1 v d2 w] 5 : d1.value, v.value, d2.value.
\end{verbatim}
}
\end{quote}
This PTQL query specifies that a drug (denoted by d1) must be followed by one of the keywords from $\{`induce',`inhibit', `inhibits'\}$ (denoted by v), which in turn must be followed by another drug (denoted by d2) followed the keyword $`metabolism'$ (denoted by w); all found within a proximity of 5 words of each other. The query produces tripes of $\langle d1, v, d2 \rangle$ values as output. Thus the results will produce triples $\langle d1, induces, d2 \rangle$ and $\langle d1, inhibits, d2 \rangle$ which mean that the drug d1 increases the effect of d2 (i.e. $\langle d1, increases, d2 \rangle$) and decreases the effect of d2 (i.e. $\langle d1, decreases, d2 \rangle$) respectively. For example, the sentence S1 above matches this PTQL query and the query will produce the triplet $\langle \text{ciprofloxacin}, \text{increases}, \text{clozapine} \rangle$.

\subsubsection{Implicit Drug Interaction Extraction}
Implicit extraction mainly extracts drug-drug interactions not yet published, but which can be inferred from published articles and properties of drug metabolism. The metabolic properties themselves have their origin in various publications. The metabolic interactions extracted from published articles and the background knowledge of properties of drug metabolism are reasoned with in an automated fashion to infer drug interactions. The following table outlines the kinds of interactions extracted from the text: 

\begin{table}\scriptsize
\begin{centering}
\begin{tabular}{ l p{6cm} }
\hline
$\langle actor1, relation, actor2 \rangle$ & Description \\
\hline
$\langle d,metabolizes,p \rangle$ & Drug \textit{d} is metabolized by protein \textit{p} \\
$\langle d,inhibits,p \rangle$ & Drug \textit{d} inhibits the activity of protein \textit{p} \\
$\langle d,induces,p \rangle$ & Drug \textit{d} induces the activity of protein \textit{p} \\
$\langle p1,regulates,p2 \rangle$ & Protein \textit{p1} regulates the activity of protein \textit{p2} \\
\hline
\end{tabular}
\caption{This table from \cite{TariDDI2010} shows the triplets representing various properties relevant to the extraction of implicit drug interactions and their description.}
\label{tab:ddi:triplets}
\end{centering}
\end{table}
which require multiple PTQL queries for extraction. As an example, the following PTQL query is used to extract  $\langle protein, metabolizes, drug \rangle$ triplets:
\begin{quote}
{\footnotesize
\begin{verbatim}
//S{/?[Tag=`Drug'](d1) =>
   //VP{//?[Value IN {`metabolized',`metabolised'}](rel) =>
   //?[Tag=`GENE'](g1)}} ::: g1.value, rel.value, d1.value
\end{verbatim}
}
\end{quote}
which specifies that the extracted triplets must have a drug (denoted by d1) followed by a verb phrase (denoted by VP) with the verb in $\{ `metabolized',`metabolised' \}$, followed by a gene (denoted by g1). Table~\ref{tab:pathsynth:2} shows examples of extracted triplets.
\begin{table}\scriptsize
\begin{centering}
\begin{tabular}{ l l p{7cm} }
\hline
PMID & Extracted Interaction & Evidence \\
\hline
8689812 & $\langle cyp3a4,metabolizes,lovastatin \rangle$ & Lovastatin is metabolized by CYP3A4 \\
8477556 & $\langle fluoxetine,inhibits,cyp2d6 \rangle$ & Inhibition by fluoxetine of cytochrome P450 2D6 activity \\
10678302 & $\langle phenytoin,induces,cyp2c \rangle$ & Phenytoin induces CYP2C and CYP3A4 isoforms, but not CYP2E1 \\
11502872 & $\langle pxr,regulates,cyp2b6 \rangle$ & The CYP2B6 gene is directly regulated by PXR \\
\hline
\end{tabular}
\caption{This table from \cite{TariDDI2010} shows the triplets representing various properties relevant to the extraction of implicit drug interactions and their description.}
\label{tab:pathsynth:2}
\end{centering}
\end{table}

\subsubsection{Data Cleaning}
The protein-protein and protein-drug relationships extracted from the parse tree database need an extra step of refinement to ensure that they correspond to the known properties of drug metabolism. For instance, for a protein to metabolize a drug, the protein must be an enzyme. Similarly, for a protein to regulate an enzyme, the protein must be a transcription factor. Thus, the $\langle protein, metabolizes, drug \rangle$ facts get refined to $\langle enzyme, metabolizes, drug \rangle$ and $\langle protein, regulates, protein \rangle$ gets refined to $\langle transcription factor, regulates, enzyme \rangle$ respectively. Classification of proteins is done using UniProt, the Gene Ontology (GO) and Entrez Gene summary by applying rules such as:
\begin{itemize}
\item A protein p is an enzyme if it belongs to a known enzyme family, such as CYP, UGT or SULT gene families; or is annotated under UniProt with the hydrolase, ligase, lyase or transferase keywords; or is listed under the \textit{``metabolic process''} GO-term; or its Entrez Gene summary mentions key phrases like \textit{``drug metabolism''} or roots for \textit{``enzyme''} or \textit{``catalyzes''}.
\item A protein p is considered as a transcription factor if it is annotated with keywords transcription, transcription-regulator or activator under UniProt; or it is listed under the \textit{``transcription factor activity''} category in GO; or its Entrez Gene summary contains the phrase \textit{``transcription factor''}.
\end{itemize}

Additional rules are applied to remove conflicting information, such as, favoring negative extractions (such as `$P$ does not metabolize $D$') over positive extractions (such as `$P$ metabolizes $D$'). For details, see~\cite{TariDDI2010}.

\subsection{Results}
The correctness of extracted interactions was determined by manually compiling a gold standard for each type of interaction using co-occurrence queries. For example, for $\langle protein, metabolizes, drug \rangle$ relations, we examined sentences that contain co-occurrence of protein, drug and one of the keywords ``metabolized'', ``metabolize'', ``metabolises'', ``metabolise'', ``substrate'' etc. Table~\ref{tab:tariddi2010:results:2} summarizes the performance of our extraction approach.

\begin{table}\scriptsize
\begin{centering}
\begin{tabular}{ p{4.5cm} p{2.5cm} p{2cm} p{1.5cm} }
\hline
Relations & Precision (\# TP) & Recall (\# FN) & F-measure \\
\hline
$<protein,metabolizes,drug>$ & 93.1\% (54) & 26.7\% (148) & 41.5\% \\
$<drug,induces,protein>$ & 61.8\% (42) & 30.7\% (95) & 41.0\% \\
$<drug,inhibits,protein>$ & 58.6\% (99) & 48.5\% (105) & 53.1\% \\
$<protein,regulates,protein>$ & 68.7\% (46) & 100.0\% (0) & 81.4\% \\
negation & 84.4\% (38) & -- & -- \\
\hline
\end{tabular}
\caption{Performance of interactions extracted from Medline abstracts. TP represents true-positives, while FN represents false-negatives~\cite{TariDDI2010}}
\label{tab:tariddi2010:results:2}
\end{centering}
\end{table}

\section{Extracting Knowledge About Genetic Variants}
\label{sec:snpshot}
We summarize the relevant portion of our work on associating genetic variants with drugs, diseases and adverse reactions as presented in ~\cite{SNPshot}.

Incorrect drug dosage is the leading cause of adverse drug reactions. Doctors prescribe the same amount of medicine to a patient for most drugs using the average drug response, even though a particular person's drug response may be higher or lower than the average case. A large part of the difference in drug response can be attributed to single nucleotide polymorphisms (SNPs). For example, the enzyme CYP2D6 has 70 known allelic variations, 50 of which are non-functional~\cite{Wuttke2002}. Patients with \textit{poor metabolizer} variations may retain higher concentration of drug for typical dosage, while patients with \textit{rapid metabolizers} may not even reach therapeutic level of drug or toxic level of drug metabolites~\cite{ZhouSF2009}. Thus, it is important to consider the individual's genetic composition for dosage determination, especially for narrow therapeutic index drugs. 

Scientists studying these variations have grouped metabolizers into categories of poor (PM), intermediate (IM), rapid (RM) and ultra-rapid metabolizers (UM) and found that for some drugs, only 20\% of usual dosage was required for PM and up to 140\% for UM~\cite{Kirchheiner2001}. Information about SNPs, their frequency in various population groups, their effect on genes (enzymic activity) and related data is stored in research papers, clinical studies and surveys. However, it is spread-out among them. Various databases collect this information in different forms. PharmGKB collects information such information and how it related to drug response~\cite{Altman2010}. However, it is a small percentage of the total number of articles on PharmGKB, due to time consuming manual curation of data.

Our work focuses on automatically extracting genetic variations and their impact on drug responses from PubMed abstracts to catch up with the current state of research in the biological domain, producing a repository of knowledge to assist in personalized medicine. Our approach leverages on as many existing tools as possible.

\subsection{Methods}
Next, we describe the methods used in our extraction, including: named entity recognition, entity normalization, and relation extraction.

\subsubsection{Named Entity Recognition}
We want to identify entities including genes (also proteins and enzymes), drugs, diseases, ADRs (adverse drug reactions), SNPs, RefSNPs (rs-numbers), names of alleles, populations and frequencies. 

For genes, we use BANNER~\cite{BANNER} trained on BioCreative II GM training set~\cite{Krallinger2008}. For genotypes (genetic variations including SNPs) we used a combination of MutationFinder~\cite{MutationFinder} and custom components. Custom components were targeted mostly on non-SNPs (``c.76\_78delACT'', 11MV324KF'') and insertions/deletions (``1707 del T'', ``c.76\_77insG''), RefSNPs (rs-numbers) and names of alleles/haplotypes (``CYP2D6*4'', ``T allele'', ``null allele'').

For diseases (and ADRs), we used BANNER trained on a corpus of 3000 sentences with disease annotations~\cite{Leaman2009}. An additional 200 random sentences containing no disease were added from BioCreative II GM data to offset the low percentage (10\%) of sentences without disease in the 3000 sentence corpus. In addition to BANNER, we used a dictionary extracted from UMLS. This dictionary consisted of 162k terms for 54k concepts from the six categories ``Disease or Syndrome'', ''Neoplastic Process'',``Congenital Abnormality'',``Mental or Behavioral Dysfunction'',``Experimental Model of Disease'' and ``Acquired Abnormality''. The list was filtered to remove unspecific as well as spurious disease names such as ``symptoms'', ``disorder'', \dots. A dictionary for adverse drug reactions originated from SIDER Side Effect Resource~\cite{Kuhn2010}, which provides a mapping of ADR terminology to UMLS CUIs. It consisted of 1.6k UMLS concepts with 6.5k terms.

For drugs, we used a dictionary based on DrugBank\cite{DrugBank} containing about 29k different drug names including both generic as well as brand names. We used the cross-linking information from DrugBank to collect additional synonyms and IDs from PharmGKB. We cross linked to Compound and Substance IDs from PubChem to provide hyperlinks to additional information.

For population, we collected a dictionary of terms referring to countries, regions, regions inhabitants and their ethnicities from WikiPedia, e.g. ``Caucasian'', ``Italian'', ``North African'', \dots. We filtered out irrelevant  phrases like ``Chinese [hamster]''.

For frequencies, we extract all numbers and percentages as well as ranges from sentences that contain the word ``allele'', ``variant'', ``mutation'', or ``population''. The output is filtered in this case as well to remove false positives referring to p-values, odd ratios, confidence intervals and common trigger words.

\subsubsection{Entity Normalization}
Genes, diseases and drugs can appear with many different names in the text. For example, ``CYP2D6'' can appear as ``Cytochrome p450 2D6'' or ``P450 IID6'' among others, but they all refer to the same enzyme (EntrezGene ID 1565). We use GNAT on recognized genes~\cite{Hakenberg2008}, but limit them to human, mouse and rat genes. The gene name recognized by BANNER is filtered by GNAT to remove non-useful modifiers and looked up against EntrezGene restricted to human, mouse and rat genes to find candidate IDs for each gene name. Ambiguity (multiple matches) is resolved by matching the text surrounding the gene mention with gene's annotation from a set of resources, like EntrezGene, UniProt.

Drugs and diseases/ADRs are resolved to their official IDs from DrugBank or UMLS. If none is found, we choose an ID for it based on its name. Genetic variants 

Genetic variations are converted to HGVScite~\cite{Dunnen2000} recommended format. Alleles were converted to the star notation (e.g. ``CYP2D6*1'') and the genotype (``TT allele'') or fixed terms such as ``null allele'' and ``variant allele''.

Populations mentions are mapped to controlled vocabulary to remove orthographic and lexical variations.

\subsubsection{Relation Extraction}
Twelve type of relations were extracted between the detected entities as given in Table \ref{tab:snpshot:tab2}. Different methods were applied to detect different relations depending upon relation type, sentence structure and whether another method was able to extraction a relation beforehand. Gene-drug, gene-disease, drug-disease were extracted using sentence based co-occurrence (possibly refined by using relation-specific keywords) due to its good precision yield of this method for these relations. For other relations additional extraction methods were implemented. These include:
\begin{description}
\item[High-confidence co-occurrence that includes keywords]
These co-occurences have the relation keyword in them. This method is applied to gene-drug, gene-disease, drug-ADR, drug-disease and mutation-disease associations. It uses keywords from PolySearch \cite{DrugBank2008} as well as our own.
\item[Co-occurrence without keywords]
Such co-occurrences do not require any relationship keyword. This method is used for allele-population and variant-population relationships. This method can misidentify negative relationships. High-confidence relationships, if not found with a keyword drop down to this method for a lower confidence result.
\item[1:n co-occurrence]
Relationships where one entity has one instance in a given sentence and the other occurs one or more times. Single instance entity may have more than one occurrence. This method is useful in identifying gene mutations, where a gene is mentioned in a sentence along with a number of its mutations. The gene itself may be repeated.
\item[Enumerations with matching counts]
Captures entities in sentences where a list of entities is followed by an equal number of counts. This method is useful in capturing alleles and their associated frequencies, e.g. ``The frequencies of CYP1B1*1, *2, *3, and *4 alleles were 0.087, 0.293, 0.444, and 0.175, respectively.''
\item[Least common ancestor (LCA) sub-tree]
Assigns associations based on distance in parse tree. We used Stanford parser \cite{Manning2003} to get grammatical structure of a sentence as a dependency tree. This allows relating verb to its subject and noun to its modifiers. This method picks the closest pair in the lowest common ancestors (dependency) sub-tree of the entities. Maximum distance in terms of edges connecting the entity nodes was set to 10, which was determined empirically to provide the best balance between precision and recall. This method associates frequencies with alleles in the sentence ``The allele frequencies were 18.3\% (-24T) and 21.2\% (1249A)''.
\item[m:n co-occurrence]
This method builds associations between all pairs of entities.
\item[Low confidence co-occurrence]
This acts as the catch-all case if none of the above methods work. 
\end{description}

\begin{table}\scriptsize
\centering
\begin{tabular}{ l r r l }
\hline
& Total & Unique & PharmGKB \\
\hline
Gene-drug & 191,054 & 31,593 & 6820/3014 \\
Gene-disease & 709,987 & 102,881 & 8147/4478 \\
Drug-disease & 117,834 & 26,268 & 4343/939 \\
Drug-adverse effect & 73,696 & 16,569 \\
Gene-variant & 101,477 & 21,704 & 645/516 \\
Gene-allele & 65,569 & 6802 & 146/99 \\
Gene-RefSNP & 12,881 & 5748 & 1820/1125 \\
Allele-population & 7181 & 1891 \\
Variant-population & 12,897 & 6765 \\
Allele frequency & 6,893 & 279 \\
Variant frequency & 6,646 & 1654 \\
Population frequency & 8404 & 144 \\
Drugs-populations & 12,849 & 4388 \\
Drug-alleles & 6,778 & 1858 \\
Drug-RefSNP & 1,161 & 721 \\
Drug-mutation & 10,809 & 5491 \\
Sum & 1,315,811 & 233,964 \\
\hline
\end{tabular}
\caption{Unique binary relations identified between detected entities from~\cite{SNPshot}. 
}
\label{tab:snpshot:tab2}
\end{table}

These methods were applied in order to each sentence, stoping at the first method that extracted the desired relationship. Order of these methods was determined empirically based of their precision. The order of the method used determines our confidence in the result. If none of the higher confidence methods are successful, a co-occurrence based method is used for extraction with low confidence.

Abstract-level co-occurrence are also extracted to provide hits on potential relations. They appear in the database only when they appear in more than a pre-set threshold number of  abstracts.

\subsection{Results}
Performance was evaluated by evaluating the precision and recall of individual components and coverage of existing results. Precision and recall were tested by processing 3500 PubMed abstracts found via PharmGKB relations and manually checking the 2500 predictions. Coverage was tested against DrugBank and PharmGKB.  Extracted relations went through manual evaluation for correctness. Each extraction was also assigned a confidence value based on the confidence in the method of extraction used. We got a coverage of 91\% of data in DrugBank and 94\% in PharmGKB. Taking into false positive rates for genes, drugs and gene-drug relations, SNPshot has more than 10,000 new relations.

\section{Applying BioPathQA to Drug-Drug Interaction Discovery}
Now we use our BioPathQA system from chapter~\ref{ch:deepqa} to answer questions about drug-drug interaction using knowledge extracted from research publications using the approach in sections~\ref{sec:ddi},\ref{sec:snpshot}. We supplement the extracted knowledge with domain knowledge as needed.

Let the extracted facts be as follows:
\begin{itemize}
\item The drug $gefitinib$ is metabolized by $CYP3A4$. 
\item The drug $phenytoin$ induces $CYP3A4$.
\end{itemize}

Following additional facts have been provided about a new drug currently in development:
\begin{itemize}
\item A new drug being developed $test\_drug$ is a CYP3A4 inhibitor
\end{itemize}

We show the pathway specification based on the above facts and background knowledge, then elaborate on each component:
{\small
\begin{flalign}
\label{ddi:pathway:domain}
&\mathbf{domain~of~} gefitinib \mathbf{~is~} integer, cyp3a4 \mathbf{~is~} integer,  \nonumber\\
&~~~~~~~~~~~~~~~~~~~~ phenytoin \mathbf{~is~} integer, test\_drug \mathbf{~is~} integer\\
\label{ddi:pathway:cyp3a4:activity}
&t1 \mathbf{~may~execute~causing~} \nonumber\\
&~~~~~~~~~~~~~~~~~~~~ gefitinib \mathbf{~change~value~by~} -1, \nonumber\\
&~~~~~~~~~~~~~~~~~~~~ cyp3a4 \mathbf{~change~value~by~} -1, \nonumber\\
&~~~~~~~~~~~~~~~~~~~~ cyp3a4 \mathbf{~change~value~by~} +1\\
\label{ddi:pathway:cyp3a4:stimulate}
&\mathbf{normally~stimulate~} t1 \mathbf{~by~factor~} 2   \nonumber\\
&~~~~~~~~~~~~~~~~~~~~ \mathbf{~if~} phenytoin \mathbf{~has~value~} 1 \mathbf{~or~higher~} \\
\label{ddi:pathway:cyp3a4:inhibit}
&\mathbf{inhibit~} t1  \mathbf{~if~} test\_drug \mathbf{~has~value~} 1 \mathbf{~or~higher~} \\
\label{ddi:pathway:init}
&\mathbf{initially~}  gefitinib \mathbf{~has~value~} 20, cyp3a4 \mathbf{~has~value~} 60, \nonumber\\
&~~~~~~~~~~~~~~~~~~~~ phenytoin \mathbf{~has~value~} 0, test\_drug \mathbf{~has~value~} 0 \\
\label{ddi:pathway:firing:style}
&\mathbf{firing~style~} max 
\end{flalign}
}

Line~\ref{ddi:pathway:domain} declares the domain of the fluents as integer numbers. Line~\ref{ddi:pathway:cyp3a4:activity} represents the activity of enzyme $cyp3a4$ as the action $t1$. Due to the enzymic action $t1$, one unit of $gefitinib$ is metabolized, and thus converted to various metabolites (not shown here). The enzymic action uses one unit of $cyp3a4$ as catalyst, which is used in the reaction and released afterwards. Line~\ref{ddi:pathway:cyp3a4:stimulate} represents the knowledge that $phenytoin$ induces the activity of $cyp3a4$. From background knowledge we find out that the stimulation in the activity can be as high as $2$-times~\cite{luo2002cyp3a4}. Line~\ref{ddi:pathway:cyp3a4:inhibit} represents the knowledge that there is a new drug $test\_drug$ being tested that is known to inhibit the activity of $cyp3a4$. Line~\ref{ddi:pathway:init} specifies the initial distribution of the drugs and enzymes in the pathway. Assuming the patient has been given some fixed dose, say $20$ units, of the medicine $gefitinib$. It also specifies there is a large $60$ units quantity of $cyp3a4$ available to ensure reactions do not slow down due to unavailability of enzyme availability. Additionaly, the drug $phenytoin$ is absent from the system and a new drug $test\_drug$ to be tested is not in the system either. This gives us our pathway specification. Now we consider two application scenarios for drug development.

\subsection{Drug Administration}
A patient is taking 20 units of $gefitinib$, and is being prescribed additional drugs to be co-administered. The drug administrator wants to know if there will be an interaction with $gefitinib$ if 5 units of $phenytoin$ are co-administered. If there is an interaction, what will be the bioavailability of $gefitinib$ so that its dosage could be appropriately adjusted.

The first question is asking whether giving the patient 5-units of $phenytoin$ in addition to the existing $gefitinib$ dose will cause a drug-interaction. It is encoded as the following query statement $\mathbf{Q}$ for a $k$-step simulation:
\begin{flalign*}
&\mathbf{direction~of~change~in~} average \mathbf{~value~of~} gefitinib \mathbf{~is~} d\\
&~~~~~~\mathbf{when~observed~at~time~step~} k; \\
&~~~~~~\mathbf{comparing~nominal~pathway~with~modified~pathway~obtained~}\\
&~~~~~~~~~~~~\mathbf{due~to~interventions:~} \mathbf{set~value~of~} phenytoin \mathbf{~to~} 5;
\end{flalign*}
If the direction of change is ``$=$'' then there was no drug-interaction. Otherwise, an interaction was noticed. For a simulation of length $k=5$, we find $15$ units of $gefitinib$ remained at the end of simulation in the nominal case when no $phenytoin$ is administered. The amount drops to $10$ units of $gefitinib$ when $phenytoin$ is co-administered. The change in direction is ``$<$''. Thus there is an interaction.

The second question is asking about the bioavailability of the drug $gefitinib$ after some after giving $phenytoin$ in 5 units. If this bioavailability falls below the efficacy level of the drug, then the drug would not treat the disease effectively. It is encoded as the following query statement $\mathbf{Q}$ for a $k$-step simulation:
\begin{flalign*}
&average \mathbf{~value~of~} gefitinib \mathbf{~is~} n\\
&~~~~\mathbf{when~observed~at~time~step~}  k; \\
&~~~~\mathbf{due~to~interventions:~} \\
&~~~~~~~~\mathbf{set~value~of~} phenytoin \mathbf{~to~} 5;
\end{flalign*}
For a simulation of length $k=5$, we find $10$ units of $gefitinib$ remain. A drug administrator (such as a pharmacist) can adjust the drug accordingly.

\subsection{Drug Development}
A drug manufacturer is developing a new drug $test\_drug$ that is known to inhibit CYP3A4 that will be co-administered with drugs $gefitinib$ and $phenytoin$. He wants to determine the bioavailability of $gefitinib$ over time to determine the risk of toxicity. 

The question is asking about the bioavailability of the drug $gefitinib$ after $10$ time units after giving $phenytoin$ in $5$ units and the new drug $test\_drug$ in $5$ units. If this bioavailability remains high, there is chance for toxicity due to the drug at the subsequent dosage intervals. It is encoded as the following query statement $\mathbf{Q}$ for a $k$-step simulation:
\begin{flalign*}
&average \mathbf{~value~of~} gefitinib \mathbf{~is~} n\\
&~~~~\mathbf{when~observed~at~time~step~}  k; \\
&~~~~\mathbf{due~to~interventions:~} \\
&~~~~~~~~\mathbf{set~value~of~} phenytoin \mathbf{~to~} 5,\\
&~~~~~~~~\mathbf{set~value~of~} test\_drug \mathbf{~to~} 5;
\end{flalign*}
For a simulation of length $k=5$, we find all $20$ units of $gefitinib$ remain. This could lead to toxicity by building high concentration of $gefitinib$ in the body.

\subsection{Drug Administration in Presence of Genetic Variation}
A drug administrator wants to establish the dosage of $morphine$ for a person based on its genetic profile using its bioavailability.

Consider the following facts extracted about a simplified morphine pathway:
\begin{itemize}
\item $codeine$ is metabolized by $CYP2D6$ producing $morphine$
\item $CYP2D6$ has three allelic variations
\begin{itemize}
\item ``*1'' -- (EM) effective metabolizer (normal case)
\item ``*2'' -- (UM) ultra rapid metabolizer
\item ``*9'' -- (PM) poor metabolizer
\end{itemize}
\end{itemize}

For simplicity, assume UM allele doubles the metabolic rate, while PM allele halves the metabolic rate of CYP2D6. Then, the resulting pathway is given by:
\begin{align*}
&\mathbf{domain~of~} cyp2d6\_allele \mathbf{~is~} integer, cyp2d6 \mathbf{~is~} integer\\
&\mathbf{domain~of~} codeine \mathbf{~is~} integer, morphine \mathbf{~is~} integer\\
&cyp2d6\_action \mathbf{~may~execute~causing}\\
&~~~~codeine \mathbf{~change~value~by~} -2, morphine \mathbf{~change~value~by~} +2\\
&~~~~\mathbf{if~} cyp2d6\_allele \mathbf{~has~value~equal~to~} 1\\
&cyp2d6\_action \mathbf{~may~execute~causing}\\
&~~~~codeine \mathbf{~change~value~by~} -4, morphine \mathbf{~change~value~by~} +4\\
&~~~~\mathbf{if~} cyp2d6\_allele \mathbf{~has~value~equal~to~} 2\\
&cyp2d6\_action \mathbf{~may~execute~causing~}\\
&~~~~codeine \mathbf{~change~value~by~} -1, morphine \mathbf{~change~value~by~} +1\\
&~~~~\mathbf{if~} cyp2d6\_allele \mathbf{~has~value~equal~to~} 9\\
&\mathbf{initially~}\\ 
&~~~~codeine \mathbf{~has~value~} 0, morphine \mathbf{~has~value~} 0,\\
&~~~~cyp2d6 \mathbf{~has~value~} 20, cyp2d6\_allele \mathbf{~has~value~} 1\\
&\mathbf{firing~style~} max\\
\end{align*}

Then, the bioavailability of $morphine$ can be determined by the following query:
\begin{flalign*}
&average \mathbf{~value~of~} morphine \mathbf{~is~} n\\
&~~~~\mathbf{when~observed~at~time~step~} k; \\
&~~~~\mathbf{due~to~interventions:~} \\
&~~~~~~~~\mathbf{set~value~of~} codeine \mathbf{~to~} 20,\\
&~~~~~~~~\mathbf{set~value~of~} cyp2d6 \mathbf{~to~} 20,\\
&~~~~~~~~\mathbf{set~value~of~} cyp2d6\_allele \mathbf{~to~} 9;
\end{flalign*}
Simulation for $5$ time steps reveal that the average bioavailability of $morphine$ after $5$ time-steps is $5$ for PM (down from $10$ for EM).

Although this is a toy example, it is easy to see the potential of capturing known genetic variations in the pathway and setting the complete genetic profile of a person in the intervention part of the query.

\section{Conclusion}
In this chapter we presented how we extract biological pathway knowledge from text, including knowledge about drug-gene interactions and their relationship to genetic variation. We showed how the information extracted is used to build pathway specification and illustrated how biologically relevant questions can be answered about drug-drug interaction using the BioPathQA system developed in chapter~\ref{ch:deepqa}. Next we look at the future directions in which the research work done in this thesis can be extended.

\chapter{Conclusion and Future Work}\label{ch:future_work}

The field of knowledge representation and reasoning (KR) is currently one of the most active research areas. It represents the next step in the evolution of systems that know how to organize knowledge, and have the ability to intelligently respond to questions about this knowledge. Such questions could be about static knowledge or the dynamic processes.

Biological systems are uniquely positioned as role models for this next evolutionary step due to their precise vocabulary and mechanical nature. As a result, a number of recent research challenges in the KR field are focused on it. The biological field itself needs systems that can intelligently answer questions about such biological processes and systems in an automated fashion, given the large number of research papers published each year. Curating these publications is time consuming and expensive, as a result the state of over all knowledge about biological systems lags behind the cutting edge research.

An important class of questions asked about biological systems are the so called ``what-if'' questions that compare alternate scenarios of a biological pathway. To answer such questions, one has to perform simulation on a nominal pathway against a pathway modified due to the interventions specified for the alternate scenario. Often, this means creating two pathways (for nominal and alternate cases) and simulate them separately. This opens up the possibility that the two pathways can become out of synchronization. A better approach is to allow the user to specify the needed interventions in the query statement itself. In addition, to understand the full spread of possible outcomes, given the parallel nature of biological pathways, one must consider all possible pathway evolutions, otherwise, some outcomes may remain hidden.

If a system is to be used by biologists, it must have a simple interface, lowering the barrier of entry. Since biological pathway knowledge can arrive from different sources, including books, published articles, and lab experiments, a common input format is desired. Such a format allows specification of pathways due to automatic extraction, as well as any changes / additions due to locally available information.

A comprehensive end-to-end system that accomplish all the goals would take a natural language query along with any additional specific knowledge about the pathway as input, extract the relevant portion of the relevant pathway from published material (and background knowledge),  simulate it based on the query, and generate the results in a visual format. Each of these tasks comes with its own challenges, some of which have been addressed in this thesis.  

In this thesis, we have developed a system and a high level language to specify a biological pathway and answer simulation based reasoning questions about it. The high level language uses controlled-English vocabulary to make it more natural for a researcher to use directly. The high level language has two components: a pathway specification language, and a query specification language. The pathway specification language allows the user to specify a pathway in a source independent form, thus locally obtained knowledge (e.g. from lab) can be combined with automatically extracted knowledge. We believe that our pathway specification language is easy for a person to understand and encode, lowering the bar to using our system. Our pathway specification language allows conditional actions, enabling the encoding of alternate action outcomes due to genetic variation. An important aspect of our pathway specification language is that it specifies trajectories, which includes specifying the initial configuration of substances, as well as state evolution style, such as maximal firing of actions, or serialized actions etc.

Our query specification language provides a bridge between natural language questions and their formal representation. It is English-like but with precise semantics. A main feature of our query language is its support for comparative queries over alternate scenarios, which is not currently supported by any of the query languages (associated with action languages) we have reviewed. Our specification of alternate scenarios uses interventions (a general form of actions), that allow the user to modify the pathway as part of the query processing. We believe our query language is easier for a biologist to understand without requiring formal training.

To model the pathways, we use Petri Nets, which have been used in the past to model and simulate biological pathways. Petri Nets have a simple visual representation, which closely matches biological pathways; and they inherently support parallelism. We extended the Petri Nets to add features that we needed to suit our domain, e.g., reset arcs that remove all quantity of a substance as soon as it is produced, and conditional arcs that specify the conditional outcome of an action. 

For simulation, we use ASP, which allowed us straight forward way to implement Petri Nets. It also gave us the ability to add extensions to the Petri Net by making local edits, implement different firing semantics, filter trajectories based on observations, and reason with the results. One of the major advantage of using Petri Net based simulation is the ability to generate all possible state evolutions, enabling us to process queries that determine the conditions when a certain observation becomes true.

Our post-processing step is done in Python, which allows strong text processing capabilities using regular expressions, as well as libraries to easy process large matrices of numbers for summarization of results.

Now we present additional challenges that need to be addressed.

\section{Pathway Extraction}
In Chapter~\ref{ch:prev_work} we described how we extract facts for drug-drug interaction and gene variation. This work needs to be extended to include some of the newer databases that have come online recently. This may provide us with enzyme reaction rates, and substance quantities used in various reactions. The relation extraction for pathways must also be cognizant of any genetic variation mentioned in the text. 

Since the knowledge about the pathway appears in relationships at varying degree of detail, a process needs to be devised to assemble the pathway from the same level to granularity together, while also maintaining pathways at different levels of granularities.

Since pathway extraction is a time consuming task, it would be best to create a catalog of the pathways. The cataloged pathways could be manually edited by the user as necessary. Storing pathways in this way means that would have to be updated periodically, requiring merging of new knowledge into existing pathways. Manual edits would have to be identified, such that the updated pathway does not overwrite them without the user's knowledge.

\section{Pathway Selection}
Questions presented in biological texts do not explicitly mention the relevant pathway to use for answering the question. One way to address this issue is to maintain a catalog of pre-defined pathways with keywords associated with them. Such keywords can include names of the substances, names of the processes, and other relevant meta-data about the pathway. The catalog can be searched to find the closest match to the query being asked.

An additional aspect in proper pathway selection is to use the proper abstraction level. If our catalog contains a pathway at different abstraction levels, the coarsest pathway that contains the processes and substances in the query should be selected. Any higher fidelity will increase the processing time and generate too much irrelevant data. Alternatively, the catalog could contain the pathway in a hierarchical form, allowing extraction of all elements of a pathway at the same depth. A common way to hierarchically organize the pathway related to our system is to have hierarchical actions, which is the approach taken by hierarchical Petri nets.

Lastly, the question may only ask about a small subsection of a much larger pathway. For better performance, it is beneficial to extract the smallest biological pathway network model that can answer the question.

\section{Pathway Modeling}
In Chapter~\ref{ch:modeling_qa}, we presented our modeling of biological questions using Petri Nets and their extensions encoded in ASP. We came across concepts like allosteric regulation, inhibition of inhibition, and inhibition of activation that we currently do not model. In allosteric regulation, an enzyme is not fully enabled or disabled, the enzyme's shape changes, making it more or less active. The level of its activity depends upon concentrations of activators and inhibitors. In inhibition of inhibition, the inhibition of a reaction is itself inhibited by another inhibition; while in inhibition of activation (or stimulation), a substance inhibits the stimulation produced by a substance. Both of these appear to be actions on actions, something that Petri Nets do not allow. An alternate coding for these would have to be devised.

As more detailed information about pathways becomes available, the reactions and processes that we have in current pathways may get replaced with more detailed sub-pathways themselves. However, such refinement may not come at the same time for separate legs of the pathway. Just replacing the coarse transition with a refined transition may not be sufficient due to relative timing constraints. Hence, a hierarchical Petri Net model may need to be implemented (see~\cite{fehling1993concept,huber1991hierarchies}).

\section{Pathway Simulation}
In Chapter~\ref{ch:asp_enc} we presented our approach to encode Petri Nets and their extensions. We used a discrete solver called {\em clingo} for our ASP encoding. As the number of simulation length increases in size or larger quantities are involved, the solver slows down significantly. This is due to an increased grounding cost of the program. Incremental solving (using {\em iclingo}) does not help, since the program size still increases, and the increments merely delays the slow down but does not stop it.

Systems such as constraint logic solvers (such as~\cite{ostrowski2012asp}) could be used for discrete cases. Alternatively, a system developed on the ASPMT~\cite{lee2013answer} approach could be used, since it can represent longer runs, larger quantities, and real number quantities.

\section{Extend High Level Language}
In Chapter~\ref{ch:deepqa} we described the BioPathQA system, the pathway specification and the query specification high level languages. As we enhance the modeling of the biological pathways, we will need to improve or extend the system as well as the high level language. We give a few examples of such extensions.

Our pathway specification language currently does not support continuous quantities (real numbers). Extending to real numbers will improve the coverage of the pathways that can be modeled. In addition, derived quantities (fluents) can be added, e.g. {\em pH} could be defined as a formula that is read-only in nature.

Certain observations and queries can be easily specified using a language such as LTL, especially for questions requiring conditions leading to an action or a state. As a result, it may be useful to add LTL formulas to the query language. We did not take this approach because it would have presented an additional non-English style syntax for the biologists.

Our substance production / consumption rates and amounts are currently tied to the fluents. In certain situations it is desirable to analyze the quantity of a substance produced / consumed by a specific action, e.g. one is interested in finding the amount of H+ ions produced by a multi-protein complex IV only.

Interventions (that are a part of the query statement) presented in this thesis are applied at the start of the simulation. Eliminating this restriction would allow applying surgeries to the pathway mid execution. Thus, instead of specifying the steady state conditions in the query statement, one could apply the intervention when a certain steady state is reached.

\section{Result Formatting and Visualization}
In Chapter~\ref{ch:deepqa} we described our system that answers questions specified in our high level language. At the end of its process, it outputs the final result. This output method can be enhanced by allowing to look at the progression of results in addition to the final result. This provides the biologist with the whole spread of possible outcomes. An example of such a spread is shown in Fig.~{fig:q1:result} for question~\ref{q1}. A graphical representation of the simulation progression is also beneficial in enhancing the confidence of the biologist. Indeed many existing tools do this. A similar effect can be achieved by parsing and showing the relevant portion of the answer set.

\section{Summary}

In Chapter~\ref{ch:intro} we introduced the thesis topic and summarized specific research contributions

In Chapter~\ref{ch:asp_enc} we introduced the foundational material of this thesis including Petri Nets and ASP. We showed how ASP could be used to encode basic Petri Nets. We also showed how ASP's elaboration tolerance and declarative syntax allows us to encode various Petri Net extensions with small localized changes. We also introduced a new firing semantics, the so called {\em maximal firing set semantics} to simulate a Petri Net with maximum parallel activity.

In Chapter~\ref{ch:modeling_qa} we showed how the Petri Net extensions and the ASP encoding can be used to answer simulation based deep reasoning questions. This and the work in Chapter~\ref{ch:asp_enc} was published in \cite{anwar2013encodinghl,anwar2013encoding}.

In Chapter~\ref{ch:deepqa} we developed a system called BioPathQA to allow users to specify a pathway and answer queries against it. We also developed a pathway specification language and a query language for this system in order to avoid the vagaries of natural language. We introduced a new type of Guarded-arc Petri Nets to model conditional actions as a model for pathway simulation. We also described our implementation developed around a subset of the pathway specification language. 

In Chapter~\ref{ch:prev_work} we briefly described how text extraction is done to extract real world knowledge about pathways and drug interactions. We then used the extracted knowledge to answer question using BioPathQA. The text extraction work was published in \cite{TariDDI2010,TariPathway2010,SNPshot}.

\appendix

\chapter{Proofs of Various Propositions}\label{ch:proofs}

\newpage
{\bf Assumption:} The definitions in this section assume the programs $\Pi$ do not have recursion through aggregate atoms. Our ASP translation ensures this due to the construction of programs $\Pi$.

First we extend some definitions and properties related to ASP, such that they apply to rules with aggregate atoms. We will refer to the non-aggregate atoms as basic atoms. Recall the definitions of an ASP program given in section~\ref{sec:asp}.

\begin{proposition}[Forced Atom Proposition]\label{prop:forced_atom}
Let $S$ be an answer set of a ground ASP program $\Pi$ as defined in definition~\ref{def:asp:program}. For any ground instance of a rule R in $\Pi$ of the form $A_0 \text{:-} A_1,\dots,$ $A_m,\mathbf{not~} B_{1},\dots,$ $\mathbf{not~} B_n, C_1,\dots,$ $C_k.$ if $\forall A_i, 1 \leq i \leq m, S \models A_i$, and $\forall B_j, 1 \leq j \leq n, S \not\models B_j$, $\forall C_l, 1 \leq l \leq k, S \models C_l$ then $S \models A_0$. 
\end{proposition}
\noindent
{\em Proof:} Let $S$ be an answer set of a ground ASP program $\Pi$, $R \in \Pi$ be a ground rule such that $\forall B_j, 1 \leq j \leq n, S \not\models B_j$; and $\forall C_l, 1 \leq l \leq k, S \models C_l$. Then, the reduct $R^S \equiv \{ p_1 \text{:-} A_1,\dots,A_m. ; \dots ; p_h \text{:-} A_1,\dots,A_m. \; | \; \{ p_1,\dots,p_h \} = S \cap lit(A_0) \}$ . Since $S$ is an answer set of $\Pi$, it is a model of $\Pi^S$. As a result, whenever, $\forall A_i, 1 \leq i \leq m, S \models A_i$, $\{ p_1,\dots,p_h \} \subseteq S$ and $S \models A_0$.

\begin{proposition}[Supporting Rule Proposition]\label{prop:supp_rule}
If $S$ is an answer set of a ground ASP program $\Pi$ as defined in definition~\ref{def:asp:program} then $S$ is supported by $\Pi$. That is, if $S \models A_0$, then there exists a ground instance of a rule R in $\Pi$ of the type $A_0 \text{:-} A_1,\dots,$ $A_m,$ $\mathbf{not~} B_{1},\dots,$ $\mathbf{not~} B_n, $ $C_1,\dots,C_k.$ such that $\forall A_i, 1 \leq i \leq m, S \models A_i$, $\forall B_j, 1 \leq j \leq n, S \not\models B_j$, and $\forall C_l, 1\leq l \leq k, S \models C_l$.
\end{proposition}

\noindent
{\em Proof:} For $S$ to be an answer set of $\Pi$, it must be the deductive closure of reduct $\Pi^S$. The deductive closure $S$ of $\Pi^S$ is iteratively built by starting from an empty set $S$, and adding head atoms of rules $R_h^S \equiv p_h \text{:-} A_1,\dots,A_m., R_h^S \in \Pi^S$, whenever, $S \models A_i, 1 \leq i \leq m$, where, $R_h^S$ is a rule in the reduct of ground rule $R \in \Pi$ with $p_h \in lit(A_0) \cap S$. Thus, there is a rule $R \equiv A_0 \text{:-} A_1,\dots,$ $A_m,$ $\mathbf{not~} B_{1},\dots,$ $\mathbf{not~}B_n, $ $C_1,\dots,$ $C_k.$, $R \in \Pi$, such that $\forall C_l, 1 \leq l \leq k$ and $\forall B_j, 1 \leq j \leq n, S \not\models B_j$. Nothing else belongs in $S$.

Next, we extend the splitting set theorem~\cite{lifschitz1994splitting} to include aggregate atoms.

\begin{definition}[Splitting Set]\label{def:splitting_set}
A {\em Splitting Set} for a program $\Pi$ is any set $U$ of literals such that, for every rule $R \in \Pi$, if $U \models head(R)$ then $lit(R) \subset U$. The set $U$ splits $\Pi$ into upper and lower parts. The set of rules $R \in \Pi$ s.t. $lit(R) \subset U$ is called the bottom of $\Pi$ w.r.t. $U$, denoted by $bot_U(\Pi)$. The rest of the rules, i.e. $\Pi \setminus bot_U(\Pi)$ is called the top of $\Pi$ w.r.t. $U$, denoted by $top_U(\Pi)$.
\end{definition}

\begin{proposition}\label{prop:split:reduct}
Let $U$ be a splitting set of $\Pi$ with answer set $S$ and let $X = S \cap U$ and $Y = S \setminus U$. Then, the reduct of $\Pi$ w.r.t. $S$, i.e. $\Pi^S$ is equal to $bot_U(\Pi)^X \cup (\Pi \setminus bot_U(\Pi))^{X \cup Y}$.
\end{proposition}
{\bf Proof:} We can rewrite $\Pi$ as $bot_U(\Pi) \cup (\Pi \setminus bot_U(\Pi))$ using the definition of splitting set. Then the reduct of $\Pi$ w.r.t. $S$ can be written in terms of $X$ and $Y$, since $S = X \cup Y$. $\Pi^S =$ $\Pi^{X \cup Y}  =$ $(bot_U(\Pi) \cup (\Pi \setminus bot_U(\Pi)))^{X \cup Y} =$ $bot_U(\Pi)^{X \cup Y} \cup $ $(\Pi \setminus bot_U(\Pi))^{X \cup Y}$. Since $lit(bot_U(\Pi)) \subseteq U$ and $Y \cap U = \emptyset$, the reduct of $bot_U(\Pi)^{X \cup Y} = bot_U(\Pi)^X$. Thus, $\Pi^{X \cup Y} = bot_U(\Pi)^X \cup (\Pi \setminus bot_U(\Pi))^{X \cup Y}$.

\begin{proposition}\label{prop:ans:in:lit:head}
Let $S$ be an answer set of a program $\Pi$, then $S \subseteq lit(head(\Pi))$.
\end{proposition}
\noindent
{\em Proof:} If $S$ is an answer set of a program $\Pi$ then $S$ is produced by the deductive closure of $\Pi^S$ (the reduct of $\Pi$ w.r.t $S$). By definition of the deductive closure, nothing can be in $S$ unless it is the head of some rule supported by $S$.

Splitting allow computing the answer set of a program $\Pi$ in layers. Answer sets of the bottom layer are first used to partially evaluate the top layer, and then answer sets of the top layer are computed. Next, we define how a program is partially evaluated.

Intuitively, the partial evaluation of an aggregate atom $c$ given splitting set $U$ w.r.t. a set of literals $X$ removes all literals that are part of the splitting set $U$ from $c$ and updates $c$'s lower and upper bounds based on the literals in $X$, which usually come from $bot_U$ of a program. The set $X$ represents our knowledge about the positive literals, while the set $U \setminus X$ represents our knowledge about naf-literals at this stage. We can remove all literals in $U$ from $c$, since the literals in $U$ will not appear in the head of any rule in $top_U$. 

\begin{definition}[Partial Evaluation of Aggregate Atom]\label{def:partial_eval:agg}
The partial evaluation of an aggregate atom $c = l \; [ B_0=w_0,\dots, B_m=w_m  ] \; u$, given splitting set $U$ w.r.t. a set of literals $X$, written $eval_U(c,X)$ is a new aggregate atom $c'$ constructed from $c$ as follows:
\begin{enumerate}
\item $pos(c') = pos(c) \setminus U$
\item $d=\sum_{B_i \in pos(c) \cap U \cap X}{w_i} $ 
\item $l' = l-d$, $u' = u-d$ are the lower and upper limits of $c'$
\end{enumerate}
\end{definition}

Next, we define how a program is partially evaluated given a splitting set $U$ w.r.t. a set of literals $X$ that form the answer-set of the lower layer. 
Intuitively, a partial evaluation deletes all rules from the partial evaluation for which the body of the rule is determined to be not supported by $U$ w.r.t. $X$. This includes rules which have an aggregate atom $c$ in their body s.t. $lit(c) \subseteq U$, but $X \not\models c$~\footnote{Note that we can fully evaluate an aggregate atom $c$ w.r.t. answer-set $X$ if $lit(c) \subseteq U$.}. In the remaining rules, the positive and negative literals that overlap with $U$ are deleted, and so are the aggregate atoms that have $lit(c) \subseteq U$ (since such a $c$ can be fully evaluated w.r.t. $X$). Each remaining aggregate atom is updated by removing atoms that belong to $U$~\footnote{Since the atoms in $U$ will not appear in the head of any atoms in $top_U$ and hence will not form a basis in future evaluations of $c$.}, and updating its limits based on the answer-set $X$
~\footnote{The limit update requires knowledge of the current answer-set to update limit values.}.
 The head atom is not modified, since $eval_U(...)$ is performed on $\Pi \setminus bot_U(\Pi)$, which already removes all rules with heads atoms that intersect $U$.

\begin{definition}[Partial Evaluation]\label{def:partial_eval}
The {\em partial evaluation} of $\Pi$, given splitting set $U$ w.r.t. a set of literals $X$ is the program $eval_U(\Pi,X)$ composed of rules $R'$ for each $R \in \Pi$ that satisfies all the following conditions:
\begin{enumerate}
\item $pos(R) \cap U \subseteq X,$
\item $((neg(R) \cap U) \cap X) = \emptyset, \text{ and }$ 
\item if there is a $c \in agg(R)$ s.t. $lit(c) \subseteq U$, then $X \models c$
\end{enumerate}
A new rule $R'$ is constructed from a rule $R$ as follows:
\begin{enumerate}
\item $head(R') = head(R)$,
\item $pos(R') = pos(R) \setminus U$, 
\item $neg(R') = neg(R) \setminus U$, 
\item $agg(R') = \{ eval_U(c,X) : c \in agg(R), lit(c) \not\subseteq U \}$
\end{enumerate}
\end{definition}

\begin{proposition}\label{prop:x:in:lit:bot:y:in:lit:eval}
Let $U$ be a splitting set for $\Pi$, $X$ be an answer set of $bot_U(\Pi)$, and $Y$ be an answer set of $eval_U(\Pi \setminus bot_U(\Pi),X)$. Then, $X \subseteq lit(\Pi) \cap U$ and $Y \subseteq lit(\Pi) \setminus U$.
\end{proposition}

\noindent
{\em Proof:} By proposition~\ref{prop:ans:in:lit:head}, $X \subseteq lit(head(bot_U(\Pi)))$, and $Y \subseteq lit(head(eval_U(\Pi \setminus bot_U(\Pi),X)))$. In addition, $lit(head(bot_U(\Pi))) \subseteq lit(bot_U(\Pi))$ and $lit(bot_U(\Pi)) \subseteq lit(\Pi) \cap U$ by definition of $bot_U(\Pi)$. Then $X \subseteq lit(\Pi) \cap U$, and $Y \subseteq lit(\Pi) \setminus U$.

\begin{definition}[Solution]\label{def:solution}
\cite{Baral2003} Let $U$ be a splitting set for a program $\Pi$. A solution to $\Pi$ w.r.t. $U$ is a pair $\langle X,Y \rangle$ of literals such that:
\begin{itemize}
\item $X$ is an answer set for $bot_U(\Pi)$
\item $Y$ is an answer set for $eval_U(top_U(\Pi),X)$; and
\item $X \cup Y$ is consistent.
\end{itemize}
\end{definition}

\begin{proposition}[Splitting Theorem]\label{def:split_theorem}
\cite{Baral2003} Let $U$ be a splitting set for a program $\Pi$. A set $S$ of literals is a consistent answer set for $\Pi$ iff $S = X \cup Y$ for some solution $\langle X,Y \rangle$ of $\Pi$ w.r.t. $U$.
\end{proposition}

\begin{lemma}\label{lemma:eval:agg}
Let $U$ be a splitting set of $\Pi$, $C$ be an aggregate atom in $\Pi$, and $X$ and $Y$ be subsets of $lit(\Pi)$ s.t. $X \subseteq U$, and $Y \cap U = \emptyset$. Then, $X \cup Y \models C$ iff $Y \models eval_U(C,X)$.
\end{lemma}

\noindent
{\em Proof:} 
\begin{enumerate}
\item Let $C' = eval_U(C,X)$, then by definition of partial evaluation of aggregate atom, $pos(C') = pos(C) \setminus U$, with lower limit $l' = l-d$, and upper limit $u' = u-d$, computed from $l,u$, the lower and upper limits of $C$, where
\[
d=\displaystyle\sum_{B_i \in pos(C) \cap U \cap X}{w_i} 
\]
\item $Y \models C'$ iff  
\[
l' \leq \left( \displaystyle\sum_{B'_i \in pos(C') \cap Y}{w'_i} \right) \leq u' 
\] 
-- by definition of aggregate atom satisfaction.
\item then $Y \models C$ iff 
\[
l \leq \left( \displaystyle\sum_{B_i \in pos(C) \cap U \cap X}{w_i} +\displaystyle\sum_{B'_i \in (pos(C) \setminus U) \cap Y}{w'_i}  \right) \leq u
\]
\item however, $(pos(C) \cap U) \cap X$ and $(pos(C) \setminus U) \cap Y$ combined represent $pos(C) \cap (X \cup Y)$ -- since 
\begin{align*} 
pos(C) \cap (X \cup Y) &= ((pos(C) \cap U) \cup (pos(C) \setminus U)) \cap (X \cup Y) \\
&= [((pos(C) \cap U) \cup (pos(C) \setminus U)) \cap X] \\
&~~~~~~\cup [((pos(C) \cap U) \cup (pos(C) \setminus U)) \cap Y]\\
&= [(pos(C) \cap U) \cap X) \cup ((pos(C) \setminus U) \cap X)]\\
&~~~~~~\cup [(pos(C) \cap U) \cap Y) \cup ((pos(C) \setminus U) \cap Y)]\\
&= [((pos(C) \cap U) \cap X) \cup \emptyset] \cup [\emptyset \cup ((pos(C) \setminus U) \cap Y)]\\
&= ((pos(C) \cap U) \cap X) \cup ((pos(C) \setminus U) \cap Y)
\end{align*}
where $X \subseteq U \text{ and } Y \cap U = \emptyset$

\item thus, $Y \models C$ iff 
\[
l \leq \left( \displaystyle\sum_{B_i \in pos(C) \cap (X \cup Y)}{w_i}  \right) \leq u
\]
\item which is the same as $X \cup Y \models C$
\end{enumerate}

\begin{lemma}\label{lemma:body:sat:eval}
Let $U$ be a splitting set for $\Pi$, and $X, Y$ be subsets of $lit(\Pi)$ s.t. $X \subseteq U$ and $Y \cap U = \emptyset$. Then the body of a rule $R' \in eval_U(\Pi,X)$ is satisfied by $Y$ iff the body of the rule $R \in \Pi$ it was constructed from is satisfied by $X \cup Y$.  
\end{lemma}

\noindent
{\em Proof:} 
$Y$ satisfies $body(R')$

iff $pos(R') \subseteq Y$, $neg(R') \cap Y = \emptyset$, $Y \models C'$ for each $C' \in agg(R')$ -- by definition of rule satisfaction

iff $(pos(R) \cap U) \subseteq X$, $(pos(R) \setminus U) \subseteq Y$, $(neg(R) \cap U) \cap X) = \emptyset$, $(neg(R) \setminus U) \cap Y) = \emptyset$, $X$ satisfies $C$ for all $C \in agg(C)$ in which $lit(C) \subseteq U$, and $Y$ satisfies $eval_U(C,X)$ for all $ C \in agg(C)$ in which $lit(C) \not\subseteq U$ -- using definition of partial evaluation

iff $pos(R) \subseteq X \cup Y$, $neg(R) \cap (X \cup Y) = \emptyset$, $X \cup Y \models C$ -- using 
\begin{itemize}
\item $(A \cap U) \cup (A \setminus U) = A$
\item $A \cap (X \cup Y) = ((A \cap U) \cup (A \setminus U)) \cap (X \cup Y) = ((A \cap U) \cap  (X \cup Y)) \cup ((A \setminus U) \cap (X \cup Y)) =  (A \cap U) \cap X) \cup ((A \setminus U) \cap Y)$ -- given $X \subseteq U$ and $Y \cap U = \emptyset$.
\item and lemma~\ref{lemma:eval:agg}
\end{itemize}

\noindent
{\em Proof of Splitting Theorem:} Let $U$ be a splitting set of $\Pi$, then a consistent set of literals $S$ is an answer set of $\Pi$ iff it can be written as $S = X \cup Y$, where $X$ is an answer set of $bot_U(\Pi)$; and $Y$ is an answer set of $eval_U(\Pi \setminus bot_U(\Pi),Y)$. %

\vspace{20pt}
\noindent
($\Leftarrow$) Let $X$ is an answer set of $bot_U(\Pi)$; and $Y$ is an answer set of $eval_U(\Pi \setminus bot_U(\Pi),X)$; we show that $X \cup Y$ is an answer set of $\Pi$. 

By definition of $bot_U(\Pi)$, $lit(bot_U(\Pi)) \subseteq U$. In addition, by proposition~\ref{prop:x:in:lit:bot:y:in:lit:eval}, $Y \cap U = \emptyset$. Then, $\Pi^{X \cup Y} = (bot_U(\Pi) \cup (\Pi \setminus bot_U(\Pi)))^{X \cup Y} = bot_U(\Pi)^{X \cup Y} \cup (\Pi \setminus bot_U(\Pi))^{X \cup Y} = bot_U(\Pi)^X \cup (\Pi \setminus bot_U(\Pi))^{X \cup Y}$.

Let $r$ be a rule in $\Pi^{X \cup Y}$, s.t. $X \cup Y \models body(r)$ then we show that $X \cup Y \models head(r)$. The rule $r$ either belongs to $bot_U(\Pi)^X$ or $(\Pi \setminus bot_U(\Pi))^{X \cup Y}$. %

\vspace{10pt}
\noindent
{\em Case 1:} say $r \in bot_U(\Pi)^X$ be a rule whose body is satisfied by $X \cup Y$
\begin{enumerate}
\item then there is a rule $R \in bot_U(\Pi)$ s.t. $r \in R^X$
\item then $X \models body(R)$ -- since $X \cup Y \models body(r)$; $lit(bot_U(\Pi)) \subseteq U$ and $Y \cap U = \emptyset$
\item we already have $X \models head(R)$ -- since $X$ is an answer set of $bot_U(\Pi)$; given
\item then $X \cup Y \models head(R)$ -- because $lit(R) \subseteq U$ and $Y \cap U = \emptyset$ 
\item consequently, $X \cup Y \models head(r)$
\end{enumerate}

\vspace{10pt}
\noindent
{\em Case 2:} say $r \in (\Pi \setminus bot_U(\Pi))^{X \cup Y}$ be a rule whose body is satisfied by $X \cup Y$ %
\begin{enumerate}
\item then there is a rule $R \in (\Pi \setminus bot_U(\Pi))$ s.t. $r \in R^{X \cup Y}$
\item \label{split:thm:left:case2:head:r} then $lit(head(R)) \cap U = \emptyset$ -- otherwise, $R$ would have belonged to $bot_U(\Pi)$, by definition of splitting set
\item then $head(r) \in Y$ -- since $X \subseteq U$
\item in addition, $pos(R) \subseteq X \cup Y$, $neg(R) \cap (X \cup Y) = \emptyset$, $X \cup Y \models C$ for each $C \in agg(R)$ -- using definition of reduct
\item then $pos(R) \cap U \subseteq X$ or $pos(R) \setminus U \subseteq Y$; $(neg(R) \cap U) \cap X = \emptyset$ and $(neg(R) \setminus U) \cap Y = \emptyset$; and for each $C \in agg(R)$, either $X \models C$ if $lit(C) \subseteq U$, or $Y \models eval_U(C,X)$ if $lit(C) \not\subseteq U$ -- by rearranging, lemma~\ref{lemma:eval:agg}, $X \subseteq U$, $Y \cap U = \emptyset$, and definition of partial evaluation of an aggregate atom
\item note that $pos(R) \cap U \subseteq X$, $(neg(R) \cap U) \cap X = \emptyset$, and for each $C \in agg(R)$, s.t. $lit(C) \subseteq U$, $X \models C$, represent conditions satisfied by each rule that become part of a partial evaluation -- using definition of partial evaluation
\item and $pos(R) \setminus U$, $neg(R) \setminus U$, and for each $C \in agg(R)$, $eval_U(C,X)$ are the modifications made to the rule during partial evaluation given splitting set $U$ w.r.t. $X$ -- using definition of partial evaluation
\item \label{split:thm:left:case2:reduct:y} and $pos(R) \setminus U \subseteq Y$, $(neg(R) \setminus U) \cap Y = \emptyset$, and for each $C \in agg(R)$, $Y \models eval_U(C,X)$ if $lit(C) \not\subseteq U$ represent conditions satisfied by rules that become part of the reduct w.r.t $Y$ -- using definition of partial evaluation and reduct
\item then $r$ is a rule in reduct $eval_U(\Pi \setminus bot_U(\Pi),X)^Y$ -- using \eqref{split:thm:left:case2:reduct:y}, \eqref{split:thm:left:case2:head:r} above
\item in addition, given that $Y$ satisfies $eval_U(\Pi \setminus bot_U(\Pi),X)$, and $head(r) \cap U = \emptyset$, we have $X \cup Y \models head(r)$

\end{enumerate}
Next we show that $X \cup Y$ satisfies all rules of $\Pi$. Say, $R$ is a rule in $\Pi$ not satisfied by $X \cup Y$. Then, either it belongs to $bot_U(\Pi)$ or $(\Pi \setminus bot_U(\Pi))$. If it belongs to $bot_U(\Pi)$, it must not be satisfied by $X$, since $lit(bot_U(\Pi)) \subseteq U$ and $Y \cap U = \emptyset$. However, the contrary is given to be true. On the other hand if it belongs to $(\Pi \setminus bot_U(\Pi))$, then $X \cup Y$ satisfies $body(R)$ but not $head(R)$. That would mean that its $head(R)$ is not satisfied by $Y$, since $head(R) \cap U = \emptyset$ by definition of splitting set. However, from lemma~\ref{lemma:body:sat:eval} we know that if $body(R)$ is satisfied by $X \cup Y$, $body(R')$ is satisfied by $Y$ for $R' \in eval_U(\Pi \setminus bot_U(\Pi),X)$. We also know that $Y$ satisfies all rules in $eval_U(\Pi \setminus bot_U(\Pi),X)$. So, $R'$ must be satisfied by $Y$ contradicting our assumption. Thus, all rules of $\Pi$ are satisfied by $X \cup Y$ and  $X \cup Y$ is an answer set of $\Pi$.

\vspace{20pt}
\noindent
($\Rightarrow$) Let $S$ be a consistent answer set of $\Pi$, we show that $S = X \cup Y$ for sets $X$ and $Y$ s.t. $X$ is an answer set of $bot_U(\Pi)$ and $Y$ is an answer set of $eval_U(\Pi \setminus bot_U(\Pi),X)$. We take $X=S \cap U$, $Y=S \setminus U$, then $S=X \cup Y$.

\vspace{10pt}
\noindent
{\em Case 1:} We show that $X$ is answer set of $bot_U(\Pi)$
\begin{enumerate}
\item $\Pi$ can be split into $bot_U(\Pi) \cup (\Pi \setminus bot_U(\Pi))$ -- by definition of splitting
\item then $X \cup Y$ satisfies $bot_U(\Pi)$ -- $X \cup Y$ is an answer set of $\Pi$; given
\item however $lit(bot_U(\Pi)) \subseteq U$, $Y \cap U = \emptyset$ -- by definition of splitting
\item then $X$ satisfies $bot_U(\Pi)$ -- since elements of $Y$ do not appear in the rules of $bot_U(\Pi)$
\item then $X$ is an answer set of $bot_U(\Pi)$
\end{enumerate}

\vspace{10pt}
\noindent
{\em Case 2:} We show that $Y$ is answer set of $eval_U(\Pi \setminus bot_U(\Pi),X)$

\begin{enumerate}
\item let $r$ be a rule in $eval_U(\Pi \setminus bot_U(\Pi),X)^Y$, s.t. its body is satisfied by $Y$
\item then $r \in R^Y$ for an $R \in eval_U(\Pi \setminus bot_U(\Pi),X)$ s.t.
\begin{inparaenum}[(i)]
\item $pos(R) \subseteq Y$
\item $neg(R) \cap Y = \emptyset$
\item $Y \models C$ for all  $C \in agg(R)$
\item $head(R) \cap Y \neq \emptyset$ -- using definition of reduct
\end{inparaenum}
\item each $R$ is constructed from $R' \in \Pi$ that satisfies all the following conditions
\begin{inparaenum}[(i)]
\item $pos(R') \subseteq U \cap X$
\item $(neg(R') \cap U) \cap X = \emptyset$
\item if there is a $C' \in agg(R')$ s.t. $lit(C') \subseteq U$, then $X \models C'$ 
\end{inparaenum}; 
and each $C \in agg(R)$ is a partial evaluation of $C' \in agg(R')$ s.t. $C = eval_U(C',X)$ -- using definition of partial evaluation
\item then the $body(R')$ satisfies all the following conditions: 
\begin{enumerate} 
\item $pos(R') \subseteq X \cup Y$  -- since $X \subseteq U$, $X \cap Y = \emptyset$ %
\item $neg(R') \cap (X \cup Y) = \emptyset$ -- since $X \subseteq U$, $X \cap Y = \emptyset$ %
\item $X \cup Y \models C'$ for each $C' \in agg(R')$ -- since
\begin{inparaenum}[(i)]
\item each $C' \in agg(R')$ with $lit(C') \subseteq U$ satisfied by $X$ is also satisfied by $X \cup Y$ as $lit(Y) \cap lit(C') = \emptyset$; and
\item each $C' \in agg(R')$ with $lit(C') \not\subseteq U$ is satisfied by $X \cup Y$ -- using partial evaluation, reduct construction, and $X \cap Y = \emptyset$
\end{inparaenum}
\end{enumerate}
\item then $X \cup Y$ satisfies $body(R')$ -- from previous line
\item in addition, $lit(head(R')) \cap U = \emptyset$, otherwise, $R'$ would have belonged to $bot_U(\Pi)$ by definition of splitting set
\item then $R'$ is a rule in $\Pi \setminus bot_U(\Pi)$ -- from the last two lines 
\item \label{split:thm:right:case2:head:y} we know that $X \cup Y$ satisfies every rule in $(\Pi \setminus bot_U(\Pi))$ -- given; and that elements of $U$ do not appear in the head of rules in $(\Pi \setminus bot_U(\Pi))$ -- from definition of splitting; then $Y$ must satisfy the head of these rules
\item then $Y$ satisfies $head(R')$ -- from \eqref{split:thm:right:case2:head:y}

\item Next we show that $Y$ satisfies all rules of $eval_U(\Pi \setminus bot_U(\Pi),X)$. Let $R'$ be a rule in $eval_U(\Pi \setminus bot_U(\Pi),X)$ such that $body(R')$ is satisfied by $Y$ but not $head(R')$. Since $head(R') \cap Y = \emptyset$, $head(R')$ is not satisfied by $X \cup Y$ either. Then, there is an $R \in (\Pi \setminus bot_U(\Pi))$ such that $X \cup Y$ satisfies $body(R)$ but not $head(R)$, which contradicts given. Thus, $Y$ satisfies all rules of $eval_U(\Pi \setminus bot_U(\Pi),X)$.

\item Then $Y$ is an answer set of $eval_U(\Pi \setminus bot_U(\Pi),X)$
\end{enumerate}

\begin{definition}[Splitting Sequence]\label{def:split_seq}
\cite{Baral2003} A {\em splitting sequence} for a program $\Pi$ is a monotone, continuous sequence ${\langle U_{\alpha} \rangle}_{\alpha < \mu}$ of splitting sets of $\Pi$ such that $\bigcup_{\alpha < \mu}{U_{\mu}} = lit(\Pi)$.
\end{definition}

\begin{definition}[Solution]\label{def:seq:solution}
\cite{Baral2003} Let $U={\langle U_{\alpha} \rangle}_{\alpha < \mu}$ be a splitting sequence for a program $\Pi$. A {\em solution} to $\Pi$ w.r.t $U$ is a sequence ${\langle X_{\alpha} \rangle}_{\alpha < \mu}$ of sets of literals such that:
\begin{enumerate}
\item $X_0$ is an answer set for $bot_{U_0}(\Pi)$
\item for any ordinal $\alpha + 1 < \mu$, $X_{\alpha+1}$ is an answer set of the program:
\[
eval_{U_{\alpha}}(bot_{U_{\alpha+1}}(\Pi) \setminus bot_{U_{\alpha}}(\Pi), \bigcup_{\nu \leq \alpha}{X_{\nu}})
\]
\item for any limit ordinal $\alpha < \mu, X_{\alpha} = \emptyset$, and
\item $\bigcup_{\alpha \leq \mu}(X_{\alpha})$ is consistent
\end{enumerate}
\end{definition}

\begin{proposition}[Splitting Sequence Theorem]\label{def:split_seq_thm}
\cite{Baral2003} Let $U={\langle U_{\alpha} \rangle}_{\alpha < \mu}$ be a splitting sequence for a program $\Pi$. A set $S$ of literals is a consistent answer set for $\Pi$ iff $S=\bigcup_{\alpha < \mu}{X_{\alpha}}$ for some solution ${\langle X_{\alpha} \rangle}_{\alpha < \mu}$ to $\Pi$ w.r.t $U$.
\end{proposition}

\noindent
{\bf Proof:} Let $U = \langle U_\alpha \rangle_{\alpha < \mu}$ be a splitting sequence of $\Pi$, then a consistent set of literals $S=\bigcup_{\alpha < \mu}{X_{\alpha}}$ is an answer set of $\Pi^S$ iff $X_0$ is an answer set of $bot_{U_0}(\Pi)$ and for any ordinal $\alpha + 1 < \mu$, $X_{\alpha+1}$ is an answer set of $eval_{U_{\alpha}}(bot_{U_{\alpha+1}}(\Pi) \setminus bot_{U_{\alpha}}(\Pi), \bigcup_{\nu \leq \alpha}X_{\nu})$.

Note that every literal in $bot_{U_0}(\Pi)$ belongs to $lit(\Pi) \cap U_0$, and every literal occurring in $eval_{U_{\alpha}}(bot_{U_{\alpha+1}}(\Pi) \setminus bot_{U_{\alpha}}(\Pi), \bigcup_{\nu \leq \alpha}X_{\nu})$, $(\alpha + 1 < \mu)$ belongs to $lit(\Pi) \cap (U_{\alpha +1} \setminus U_{\alpha})$. In addition, $X_0$, and all $X_{\alpha+1}$ are pairwise disjoint.

\vspace{20pt}
\noindent
We prove the theorem by induction over the splitting sequence. %

\noindent
{\bf Base case:} $\alpha = 1$. The splitting sequence is $U_0 \subseteq U_1$.

Then the sub-program $\Pi_1 = bot_{U_1}(\Pi)$ contains all literals in $U_1$; and $U_0$ splits $\Pi_1$ into $bot_{U_0}(\Pi_1)$ and $bot_{U_1}(\Pi_1) \setminus bot_{U_0}(\Pi_1)$. Then, $S_1 = X_0 \cup X_1$ is a consistent answer set of $\Pi_1$ iff $X_0 = S_1 \cap U_0$ is an answer set of $bot_{U_0}(\Pi_1)$ and $X_1 = S_1 \setminus U_0$ is an answer set of $eval_{U_0}(\Pi_1 \setminus bot_{U_0}(\Pi_1),X_1)$  -- by the splitting theorem

Since $bot_{U_0}(\Pi_1) = bot_{U_0}(\Pi)$ and $bot_{U_1}(\Pi_1) \setminus bot_{U_0}(\Pi_1) = bot_{U_1}(\Pi) \setminus bot_{U_0}(\Pi)$; $S_1 = X_0 \cup X_1$ is an answer set for $\Pi \setminus bot_{U_1}(\Pi)$.

\noindent
{\bf Induction:}  Assume theorem holds for $\alpha = k$, show theorem holds for $\alpha = k+1$.

The inductive assumption holds for the splitting sequence $U_0 \subseteq \dots \subseteq U_k$. Then the sub-program $\Pi_k = bot_{U_k}(\Pi)$ contains all literals in $U_k$ and $S_k = X_0 \cup \dots \cup X_k$ is a consistent  answer set of $\Pi^{S_k}$ iff $X_0$ is an answer set for $bot_{U_0}(\Pi_k)$ and for any $\alpha \leq k$, $X_{\alpha+1}$ is answer set of $eval_{U_{\alpha}}(bot_{U_{\alpha+1}}(\Pi) \setminus bot_{U_{\alpha}}(\Pi), X_0 \cup \dots \cup X_{\alpha})$

We show that the theorem holds for $\alpha = k+1$. The splitting sequence is $U_0 \subseteq U_{k+1}$.

Then the sub-program $\Pi_{k+1} = bot_{U_{k+1}}(\Pi)$ contains all literals $U_{k+1}$. We have $U_k$ split $\Pi_{k+1}$ into $bot_{U_k}(\Pi_{k+1})$ and $bot_{U_{k+1}}(\Pi_{k+1}) \setminus bot_{U_k}(\Pi_{k+1})$. Then, $S_{k+1} = X_{0:k} \cup X_{k+1}$ is a consistent answer set of $\Pi_{k+1}$ iff $X_{0:k} = S_{k+1} \cap U_k$ is an answer set of $bot_{U_k}(\Pi_{k+1})$ and $X_{k+1} = S_{k+1} \setminus U_k$ is an answer set of $eval_{U_k}(\Pi_{k+1} \setminus bot_{U_k}(\Pi_{k+1},X_{k+1})$ -- by the splitting theorem

Since $bot_{U_k}(\Pi_{k+1}) = bot_{U_k}(\Pi)$ and $bot_{U_{k+1}}(\Pi_{k+1}) \setminus bot_{U_k}(\Pi_{k+1}) = bot_{U_{k+1}}(\Pi) \setminus bot_{U_k}(\Pi)$; $S_{k+1} = X_{0:k} \cup X_1$ is an answer set for $\Pi \setminus bot_{U_{k+1}}(\Pi)$.

From the inductive assumption we know that $X_0 \cup \dots \cup X_k$ is a consistent answer set of $bot_{U_k}(\Pi)$, $X_0$ is the answer set of $bot_{U_0}(\Pi)$, and for each $0 \leq \alpha \leq k$, $X_{\alpha+1}$ is answer set of $eval_{U_{\alpha}}(bot_{U_{\alpha+1}}(\Pi) \setminus bot_{U_{\alpha}}(\Pi), X_0 \cup \dots \cup X_{\alpha})$. Thus, $X_{0:k} = X_0 \cup \dots \cup X_k$.

Combining above with the inductive assumption, we get $S_{k+1} = X_0 \cup \dots \cup X_{k+1}$ is a consistent answer set of $\Pi^{S_{k+1}}$ iff $X_0$ is an answer set for $bot_{U_0}(\Pi_{k+1})$ and for any $\alpha \leq k+1$, $X_{\alpha+1}$ is answer set of $eval_{U_{\alpha}}(bot_{U_{\alpha+1}}(\Pi) \setminus bot_{U_{\alpha}}(\Pi), X_0 \cup \dots \cup X_{\alpha})$.

In addition, for some $\alpha < \mu$, where $\mu$ is the length of the splitting sequence $U = \langle U_{\alpha} \rangle_{\alpha < \mu}$ of $\Pi$, $bot_{U_{\alpha}}(\Pi)$ will be the entire $\Pi$, i.e. $lit(\Pi) = U_{\alpha}$. Then the set $S$ of literals is a consistent answer set of $\Pi$ iff $S=\bigcup_{\alpha < \mu}(X_{\alpha})$ for some solution $\langle X_{\alpha} \rangle_{\alpha < \mu}$ to $\Pi$ w.r.t $U$.

\section{Proof of Proposition~\ref{prop:basic_enc}}

Let $PN=(P,T,E,W)$ be a Petri Net, $M_0$ be its initial marking and let $\Pi^0(PN,M_0,$ $k,ntok)$ be the ASP encoding of $PN$ and $M_0$ over a simulation length $k$, with maximum $ntok$ tokens on any place node, as defined in section~\ref{sec:enc_basic}. Then $X=M_0,T_0,M_1,\dots,M_k,T_k,M_{k+1}$ is an execution sequence of a $PN$ (w.r.t. $M_0$) iff there is an answer set $A$ of $\Pi^0(PN,M_0,k,ntok)$ such that:
\begin{equation}\label{eqn:fires}
\{ fires(t,ts) : t \in T_{ts}, 0 \leq ts \leq k \} = \{ fires(t,ts) : fires(t,ts) \in A  \}
\end{equation}
\begin{equation}\label{eqn:holds}
\begin{split}
\{ holds(p,q,ts) &: p \in P, q = M_{ts}(p), 0 \leq ts \leq k+1 \} \\
&= \{ holds(p,q,ts) : holds(p,q,ts) \in A \}
\end{split}
\end{equation}

We prove this by showing that:
\begin{enumerate}[(I)]
\item Given an execution sequence $X$, we create a set $A$ such that it satisfies \eqref{eqn:fires} and \eqref{eqn:holds} and show that $A$ is an answer set of $\Pi^0$ \label{prove:x2a}
\item Given an answer set $A$ of $\Pi^0$, we create an execution sequence $X$ such that (\ref{eqn:fires}) and (\ref{eqn:holds}) are satisfied \label{prove:a2x}
\end{enumerate}

\noindent
{\bf First we show (\ref{prove:x2a}):} Given $PN$ and its execution sequence $X$, we create a set $A$ as a union of the following sets:
\begin{enumerate}
\item $A_1=\{ num(n) : 0 \leq n \leq ntok \}$ %
\item $A_2=\{ time(ts) : 0 \leq ts \leq k\}$ %
\item $A_3=\{ place(p) : p \in P \}$ %
\item $A_4=\{ trans(t) : t \in T \}$ %
\item $A_5=\{ ptarc(p,t,n) : (p,t) \in E^-, n=W(p,t) \}$, where $E^- \subseteq E$ %
\item $A_6=\{ tparc(t,p,n) : (t,p) \in E^+, n=W(t,p) \}$, where $E^+ \subseteq E$ %
\item $A_7=\{ holds(p,q,0) : p \in P, q=M_{0}(p) \}$ %
\item $A_8=\{ notenabled(t,ts) : t \in T, 0 \leq ts \leq k, \exists p \in \bullet t, M_{ts}(p) < W(p,t) \}$ \newline per definition~\ref{def:pn:enable} (enabled transition) %
\item $A_9=\{ enabled(t,ts) : t \in T, 0 \leq ts \leq k, \forall p \in \bullet t, W(p,t) \leq M_{ts}(p) \}$ \newline per definition~\ref{def:pn:enable} (enabled transition) %
\item $A_{10}=\{ fires(t,ts) : t \in T_{ts}, 0 \leq ts \leq k \}$ \newline from definition~\ref{def:pn:firing_set} (firing set) 
, only an enabled transition may fire
\item $A_{11}=\{ add(p,q,t,ts) : t \in T_{ts}, p \in t \bullet, q=W(t,p), 0 \leq ts \leq k \}$ \newline per definition~\ref{def:pn:texec} (transition execution) %
\item $A_{12}=\{ del(p,q,t,ts) : t \in T_{ts}, p \in \bullet t, q=W(p,t), 0 \leq ts \leq k \}$ \newline per definition~\ref{def:pn:texec} (transition execution) %
\item $A_{13}=\{ tot\_incr(p,q,ts) : p \in P, q=\sum_{t \in T_{ts}, p \in t \bullet}{W(t,p)}, 0 \leq ts \leq k \}$ \newline per definition~\ref{def:pn:exec} (firing set execution) %
\item $A_{14}=\{ tot\_decr(p,q,ts): p \in P, q=\sum_{t \in T_{ts}, p \in \bullet t}{W(p,t)}, 0 \leq ts \leq k \}$ \newline per definition~\ref{def:pn:exec} (firing set execution) %
\item $A_{15}=\{ consumesmore(p,ts) : p \in P, q=M_{ts}(p), q1=\sum_{t \in T_{ts}, p \in \bullet t}{W(p,t)}, q1 > q, 0 \leq ts \leq k \}$ \newline per definition~\ref{def:pn:conflict} (conflicting transitions) for enabled transition set $T_{ts}$ %
\item $A_{16}=\{ consumesmore : \exists p \in P : q=M_{ts}(p), q1=\sum_{t \in T_{ts}, p \in \bullet t}{W(p,t)}, q1 > q, 0 \leq ts \leq k \}$ \newline per definition~\ref{def:pn:conflict} (conflicting transitions) %
\item $A_{17}=\{ holds(p,q,ts+1) : p \in P, q=M_{ts+1}(p), 0 \leq ts < k\}$, \newline where $M_{ts+1}(p) = M_{ts}(p) - \sum_{\substack{t \in T_{ts}, p \in \bullet t}}{W(p,t)} + \sum_{\substack{t \in T_{ts}, p \in t \bullet}}{W(t,p)}$ \newline according to definition~\ref{def:pn:firing_set} (firing set execution) %
\end{enumerate}

\noindent
{\bf We show that $A$ satisfies (\ref{eqn:fires}) and (\ref{eqn:holds}), and $A$ is an answer set of $\Pi^0$.}

$A$ satisfies (\ref{eqn:fires}) and (\ref{eqn:holds}) by its construction (given above). We show $A$ is an answer set of $\Pi^0$ by splitting. We split $lit(\Pi^0)$ (literals of $\Pi^0$) into a sequence of $6(k+1)+2$ sets:
\begin{itemize}\renewcommand{\labelitemi}{$\bullet$}
\item $U_0= head(f\ref{f:place}) \cup head(f\ref{f:trans}) \cup head(f\ref{f:ptarc}) \cup head(f\ref{f:tparc}) \cup head(f\ref{f:time}) \cup head(f\ref{f:num}) \cup head(i\ref{i:holds}) = \{place(p) : p \in P\} \cup \{ trans(t) : t \in T\} \cup \{ ptarc(p,t,n) : (p,t) \in E^-, n=W(p,t) \} \cup \{  tparc(t,p,n) : (t,p) \in E^+, n=W(t,p) \} \cup $ $\{ time(0), \dots, $ $time(k)\} \cup \{num(0), \dots, num(ntok)\} \cup \{ holds(p,q,0) : p \in P, q=M_0(p) \} $
\item $U_{6k+1}=U_{6k+0} \cup head(e\ref{e:ne:ptarc})^{ts=k} = U_{6k+0} \cup \{ notenabled(t,k) : t \in T \}$
\item $U_{6k+2}=U_{6k+1} \cup head(e\ref{e:enabled})^{ts=k} = U_{6k+1} \cup \{ enabled(t,k) : t \in T \}$
\item $U_{6k+3}=U_{6k+2} \cup head(a\ref{a:fires})^{ts=k} = U_{6k+2} \cup \{ fires(t,k) : t \in T \}$
\item $U_{6k+4}=U_{6k+3}  \cup head(r\ref{r:add})^{ts=k} \cup head(r\ref{r:del})^{ts=k} = U_{6k+3} \cup \{ add(p,q,t,k) : p \in P, t \in T, q=W(t,p) \} \cup \{ del(p,q,t,k) : p \in P, t \in T, q=W(p,t) \}$
\item $U_{6k+5}=U_{6k+4} \cup head(r\ref{r:totincr})^{ts=k} \cup head(r\ref{r:totdecr})^{ts=k} = U_{6k+4} \cup \{ tot\_incr(p,q,k) : p \in P, 0 \leq q \leq ntok \} \cup \{ tot\_decr(p,q,k) : p \in P, 0 \leq q \leq ntok \}$
\item $U_{6k+6}=U_{6k+5} \cup head(r\ref{r:nextstate})^{ts=k} \cup head(a\ref{a:overc:place})^{ts=k} = U_{6k+5} \cup \{ holds(p,q,k+1) : p \in P, 0 \leq q \leq ntok \} \cup \{ consumesmore(p,k) : p \in P\}$
\item $U_{6k+7}=U_{6k+6} \cup head(a\ref{a:overc:gen}) = U_{7k+6} \cup \{ consumesmore \}$
\end{itemize}
where $head(r_i)^{ts=k}$ are head atoms of ground rule $r_i$ in which $ts=k$. We write $A_i^{ts=k} = \{ a(\dots,ts) : a(\dots,ts) \in A_i, ts=k \}$ as short hand for all atoms in $A_i$ with $ts=k$. $U_{\alpha}, 0 \leq \alpha \leq 6k+7$ form a splitting sequence, since each $U_i$ is a splitting set of $\Pi^0$, and $\langle U_{\alpha}\rangle_{\alpha < \mu}$ is a monotone continuous sequence, where $U_0 \subseteq U_1 \dots \subseteq U_{6k+7}$ and $\bigcup_{\alpha < \mu}{U_{\alpha}} = lit(\Pi^0)$. 

We compute the answer set of $\Pi^0$ using the splitting sets as follows:

\begin{enumerate}
\item $bot_{U_0}(\Pi^0) = f\ref{f:place} \cup f\ref{f:trans} \cup f\ref{f:ptarc} \cup f\ref{f:tparc} \cup f\ref{f:time} \cup i\ref{i:holds} \cup f\ref{f:num}$ and $X_0 = A_1 \cup \dots \cup A_7$ ($= U_0$) is its answer set -- using forced atom proposition

\item $eval_{U_0}(bot_{U_1}(\Pi^0) \setminus bot_{U_0}(\Pi^0), X_0) = \{ notenabled(t,0) \text{:-} . | \{ trans(t), $ $ ptarc(p,t,n), $ $ holds(p,q,0) \} \subseteq X_0, \text{~where~}  q < n \}$. Its answer set $X_1=A_8^{ts=0}$ -- using  forced atom proposition and construction of $A_8$.
\begin{enumerate}
\item where, $q=M_0(p)$, $n=W(p,t)$, and for an arc $(p,t) \in E^-$ -- by construction of $i\ref{i:holds}$ and $f\ref{f:ptarc}$ in $\Pi^0$, and 
\item in an arc $(p,t) \in E^-$, $p \in \bullet t$ (by definition~\ref{def:pn:preset} of preset)
\item thus, $notenabled(t,0) \in X_1$ represents $\exists p \in \bullet t : M_0(p) < W(p,t)$.
\end{enumerate}

\item $eval_{U_1}(bot_{U_2}(\Pi^0) \setminus bot_{U_1}(\Pi^0), X_0 \cup X_1) = \{ enabled(t,0) \text{:-}. | trans(t) \in X_0 \cup X_1, $ $notenabled(t,0) \notin X_0 \cup X_1 \}$. Its answer set is $X_2 = A_9^{ts=0}$ -- using forced atom proposition and construction of $A_9$.
\begin{enumerate}
\item since an $enabled(t,0) \in X_2$ if $\nexists ~notenabled(t,0) \in X_0 \cup X_1$, which is equivalent to $\nexists p \in \bullet t : M_0(p) < W(p,t) \equiv \forall p \in \bullet t: M_0(p) \geq W(p,t)$.
\end{enumerate}

\item $eval_{U_2}(bot_{U_3}(\Pi^0) \setminus bot_{U_2}(\Pi^0), X_0 \cup X_1 \cup X_2) = \{\{fires(t,0)\} \text{:-}. | enabled(t,0) \\ \text{~holds in~} X_0 \cup X_1 \cup X_2 \}$. It has multiple answer sets $X_{3.1}, \dots, X_{3.n}$, corresponding to elements of power set of $fires(t,0)$ atoms in $eval_{U_2}(...)$ -- using supported rule proposition. Since we are showing that the union of answer sets of $\Pi^0$ determined using splitting is equal to $A$, we only consider the set that matches the $fires(t,0)$ elements in $A$ and call it $X_3$, ignoring the rest. Thus, $X_3 = A_{10}^{ts=0}$, representing $T_0$.

\item $eval_{U_3}(bot_{U_4}(\Pi^0) \setminus bot_{U_3}(\Pi^0), X_0 \cup \dots \cup X_3) = \{add(p,n,t,0) \text{:-}. | \{fires(t,0), \\ tparc(t,p,n) \} \subseteq X_0 \cup \dots \cup X_3 \} \cup \{ del(p,n,t,0) \text{:-}. | \{ fires(t,0), ptarc(p,t,n) \} \subseteq X_0 \cup \dots \cup X_3 \}$. It's answer set is $X_4 = A_{11}^{ts=0} \cup A_{12}^{ts=0}$ -- using forced atom proposition and definitions of $A_{11}$ and $A_{12}$. 
\begin{enumerate}
\item where each $add$ atom is equivalent to $n=W(t,p) : p \in t \bullet$, 
\item and each $del$ atom is equivalent to $n=W(p,t) : p \in \bullet t$,
\item representing the effect of transitions in $T_0$ -- by construction
\end{enumerate}

\item $eval_{U_4}(bot_{U_5}(\Pi^0) \setminus bot_{U_4}(\Pi^0), X_0 \cup \dots \cup X_4) = \{tot\_incr(p,qq,0) \text{:-}. | \\ qq=\sum_{add(p,q,t,0) \in X_0 \cup \dots \cup X_4}{q} \} \cup $ $\{ tot\_decr(p,qq,0) \text{:-}. | qq=\sum_{del(p,q,t,0) \in X_0 \cup \dots \cup X_4}{q} \}$. It's answer set is $X_5 = A_{13}^{ts=0} \cup A_{14}^{ts=0}$ --  using forced atom proposition and definitions of $A_{13}$ and $A_{14}$, ad definition~\ref{def:agg:atom} (semantics of aggregate assignment atom).
\begin{enumerate}
\item where each $tot\_incr(p,qq,0)$, $qq=\sum_{add(p,q,t,0) \in X_0 \cup \dots X_4}{q} \\ \equiv $ $qq=\sum_{t \in X_3, p \in t \bullet}{W(p,t)}$,
\item and each $tot\_decr(p,qq,0)$, $qq=\sum_{del(p,q,t,0) \in X_0 \cup \dots X_4}{q} \\ \equiv $ $qq=\sum_{t \in X_3, p \in \bullet t}{W(t,p)}$,
\item represent the net effect of transitions in $T_0$ -- by construction
\end{enumerate}
\item $eval_{U_5}(bot_{U_6}(\Pi^0) \setminus bot_{U_5}(\Pi^0), X_0 \cup \dots \cup X_5) = \{ consumesmore(p,0) \text{:-}. | \\ \{holds(p,q,0), $ $tot\_decr(p,q1,0) \} \subseteq X_0 \cup \dots \cup X_5, $ $q1 > q \} \cup $ $\{ holds(p,q,1) \text{:-}. | $ $ \{ holds(p,q1,0), $ $tot\_incr(p,q2,0), $ $tot\_decr(p,q3,0) \} \subseteq X_0 \cup \dots \cup X_5, $ $q=q1+q2-q3 \}$. It's answer set is $X_6 = A_{15}^{ts=0} \cup A_{17}^{ts=0}$ -- using forced atom proposition.
\begin{enumerate}
\item where, $consumesmore(p,0)$ represents $\exists p : q=M_0(p), \\ q1=\sum_{t \in T_0, p \in \bullet t}{W(p,t)}, q1 > q$ -- indicating place $p$ will be overconsumed if $T_0$ is fired as defined in definition~\ref{def:pn:conflict} (conflicting transitions)
\item and $holds(p,q,1)$ represents $q=M_1(p)$ -- by construction 
\end{enumerate}

\[ \vdots \]

\item $eval_{U_{6k+0}}(bot_{U_{6k+1}}(\Pi^0) \setminus bot_{U_{6k+0}}(\Pi^0), X_0 \cup \dots \cup X_{6k+0}) = \\ \{ notenabled(t,k) \text{:-} . | \{ trans(t), ptarc(p,t,n), holds(p,q,k) \} \subseteq X_0 \cup \dots \cup X_{6k+0}, \\ \text{~where~}  q < n \}$. Its answer set $X_{6k+1}=A_8^{ts=k}$ -- using forced atom proposition and construction of $A_8$.
\begin{enumerate}
\item where, $q=M_k(p)$, and $n=W(p,t)$ for an arc $(p,t) \in E^-$ -- by construction of $holds$ and $ptarc$ predicates in $\Pi^0$, and 
\item in an arc $(p,t) \in E^-$, $p \in \bullet t$ (by definition~\ref{def:pn:preset} of preset)
\item thus, $notenabled(t,k) \in X_{6k+1}$ represents $\exists p \in \bullet t : M_k(p) < W(p,t)$.
\end{enumerate}

\item $eval_{U_{6k+1}}(bot_{U_{6k+2}}(\Pi^0) \setminus bot_{U_{6k+1}}(\Pi^0), X_0 \cup \dots \cup X_{6k+1}) = \{ enabled(t,k) \text{:-}. | \\ trans(t) \in $ $X_0 \cup \dots \cup X_{6k+1}, notenabled(t,k) \notin X_0 \cup \dots \cup X_{6k+1} \}$. Its answer set is $X_{6k+2} = A_9^{ts=k}$ -- using forced atom proposition and construction of $A_9$.
\begin{enumerate}
\item since an $enabled(t,k) \in X_{6k+2}$ if $\nexists ~notenabled(t,k) \in X_0 \cup \dots \cup X_{6k+1}$, which is equivalent to $\nexists p \in \bullet t : M_k(p) < W(p,t) \equiv \forall p \in \bullet t: M_k(p) \geq W(p,t)$.
\end{enumerate}

\item $eval_{U_{6k+2}}(bot_{U_{6k+3}}(\Pi^0) \setminus bot_{U_{6k+2}}(\Pi^0), X_0 \cup \dots \cup X_{6k+2}) = \\ \{\{fires(t,k)\} \text{:-}. | $ $enabled(t,k) \text{~holds in~} $ $X_0 \cup \dots \cup X_{6k+2} \}$. It has multiple answer sets $X_{{6k+3}.1}, \dots, X_{{6k+3}.n}$, corresponding to elements of power set of $fires(t,k)$ atoms in $eval_{U_{6k+2}}(...)$ -- using supported rule proposition. Since we are showing that the union of answer sets of $\Pi^0$ determined using splitting is equal to $A$, we only consider the set that matches the $fires(t,k)$ elements in $A$ and call it $X_{6k+3}$, ignoring the reset. Thus, $X_{6k+3} = A_{10}^{ts=k}$, representing $T_k$.

\item $eval_{U_{6k+3}}(bot_{U_{6k+4}}(\Pi^0) \setminus bot_{U_{6k+3}}(\Pi^0), X_0 \cup \dots \cup X_{6k+3}) = $ $ \{add(p,n,t,k) \text{:-}. | \\ \{fires(t,k), tparc(t,p,n) \} \subseteq X_0 \cup \dots \cup X_{6k+3} \} \cup $ $ \{ del(p,n,t,k) \text{:-}. | $ $\{ fires(t,k), $ $ptarc(p,t,n) \} \subseteq X_0 \cup \dots \cup X_{6k+3} \}$. It's answer set is $X_{6k+4} = A_{11}^{ts=k} \cup A_{12}^{ts=k}$ -- using forced atom proposition and definitions of $A_{11}$ and $A_{12}$. 
\begin{enumerate}
\item where, each $add$ atom is equivalent to $n=W(t,p) : p \in t \bullet$, 
\item and each $del$ atom is equivalent to $n=W(p,t) : p \in \bullet t$
\item representing the effect of transitions in $T_k$
\end{enumerate}

\item $eval_{U_{6k+4}}(bot_{U_{6k+5}}(\Pi^0) \setminus bot_{U_{6k+4}}(\Pi^0), X_0 \cup \dots \cup X_{6k+4}) = $ $\{tot\_incr(p,qq,k) \text{:-}. | $ \\$ qq=\sum_{add(p,q,t,k) \in X_0 \cup \dots \cup X_{6k+4}}{q} \} \cup $ $\{ tot\_decr(p,qq,k) \text{:-}. | \\ qq=\sum_{del(p,q,t,k) \in X_0 \cup \dots \cup X_{6k+4}}{q} \}$. It's answer set is $X_5 = A_{13}^{ts=k} \cup A_{14}^{ts=k}$ --  using forced atom proposition and definitions of $A_{13}$ and $A_{14}$.
\begin{enumerate}
\item where, each $tot\_incr(p,qq,k)$, $qq=\sum_{add(p,q,t,k) \in X_0 \cup \dots X_{7k+4}}{q}$ \\$\equiv qq=\sum_{t \in X_{6k+3}, p \in t \bullet}{W(p,t)}$, 
\item and each $tot\_decr(p,qq,k)$, $qq=\sum_{del(p,q,t,k) \in X_0 \cup \dots X_{7k+4}}{q}$ \\$\equiv qq=\sum_{t \in X_{6k+3}, p \in \bullet t}{W(t,p)}$,
\item represent the net effect of transitions in $T_k$
\end{enumerate}
\item $eval_{U_{6k+5}}(bot_{U_{6k+6}}(\Pi^0) \setminus bot_{U_{6k+5}}(\Pi^0), X_0 \cup \dots \cup X_{6k+5}) = $ $ \{ consumesmore(p,k) \text{:-}. | $ $ \{holds(p,q,k), $ $ tot\_decr(p,q1,k) \} \subseteq X_0 \cup \dots \cup X_{6k+5} , q1 > q \} \cup \{ holds(p,q,k+1) \text{:-}., |  \{ holds(p,q1,k), tot\_incr(p,q2,k), \\ tot\_decr(p,q3,k) \} \subseteq X_0 \cup \dots \cup X_{6k+5}, q=q1+q2-q3 \}$. It's answer set is $X_{6k+6} = A_{15}^{ts=k} \cup A_{17}^{ts=k}$ -- using forced atom proposition.
\begin{enumerate}
\item where, $holds(p,q,k+1)$ represents the marking of place $p$ in the next time step due to firing $T_k$,
\item and, $consumesmore(p,k)$ represents $\exists p : q=M_{k}(p), q1=\sum_{t \in T_{k}, p \in \bullet t}{W(p,t)}, $ $q1 > q$ indicating place $p$ that will be overconsumed if $T_k$ is fired as defined in definition~\ref{def:pn:conflict} (conflicting transitions)
\end{enumerate}

\item $eval_{U_{6k+6}}(bot_{U_{6k+7}}(\Pi^0) \setminus bot_{U_{6k+6}}(\Pi^0), X_0 \cup \dots \cup X_{6k+6}) = \{ consumesmore \text{:-}. | $ $\{consumesmore(p,0),$ $ \dots, $ $consumesmore(p,k)\} \cap (X_0 \cup \dots \cup X_{6k+7}) \neq \emptyset \}$. It's answer set is $X_{6k+7} = A_{16}^{ts=k}$ -- using forced atom proposition
\begin{enumerate}
\item $X_{6k+7}$ will be empty since none of $consumesmore(p,0),\dots, \\ consumesmore(p,k)$ hold in $X_0 \cup \dots \cup X_{6k+6}$ due to the construction of $A$ and encoding of $a\ref{a:overc:place}$, and it is not eliminated by the constraint $a\ref{a:overc:elim}$.
\end{enumerate}

\end{enumerate}

The set $X=X_{0} \cup \dots \cup X_{6k+7}$ is the answer set of $\Pi^0$ by the splitting sequence theorem~\ref{def:split_seq_thm}. Each $X_i, 0 \leq i \leq 6k+7$ matches a distinct partition of $A$, and $X = A$, thus $A$ is an answer set of $\Pi^0$.

\vspace{30pt}
{\bf Next we show (\ref{prove:a2x}):} Given $\Pi^0$ be the encoding of a Petri Net $PN(P,T,E,W)$ with initial marking $M_0$, and $A$ be an answer set of $\Pi^0$ that satisfies (\ref{eqn:fires}) and (\ref{eqn:holds}), then we can construct $X=M_0,T_0,\dots,M_k,T_k,M_{k+1}$ from $A$, such that it is an execution sequence of $PN$.

We construct the $X$ as follows:
\begin{enumerate}
\item $M_i = (M_i(p_0), \dots, M_i(p_n))$, where $\{ holds(p_0,M_i(p_0),i), \dots holds(p_n,M_i(p_n),i) \} \\ \subseteq A$, for $0 \leq i \leq k+1$
\item $T_i = \{ t : fires(t,i) \in A\}$, for $0 \leq i \leq k$ 
\end{enumerate}
and show that $X$ is indeed an execution sequence of $PN$. We show this by induction over $k$ (i.e. given $M_k$, $T_k$ is a valid firing set and its firing produces marking $M_{k+1}$).

\vspace{20pt}

\noindent
{\bf Base case:} Let $k=0$, show
\begin{inparaenum}[(1)]
\item $T_0$ is a valid firing set for $M_0$, and 
\item $T_0$'s firing in $M_0$ producing marking $M_1$.
\end{inparaenum} 

\noindent
\begin{enumerate}
\item We show $T_0$ is a valid firing set w.r.t. $M_0$. Let $\{ fires(t_0,0), \dots, fires(t_x,0) \}$ be the set of all $fires(\dots,0)$ atoms in $A$, \label{prove:fires_t0}

\begin{enumerate}
\item Then for each $fires(t_i,0) \in A$

\begin{enumerate}
\item $enabled(t_i,0) \in A$ -- from rule $a\ref{a:fires}$ and supported rule proposition
\item Then $notenabled(t_i,0) \notin A$ -- from rule $e\ref{e:enabled}$ and supported rule proposition
\item Then $body(e\ref{e:ne:ptarc})$ must not hold in $A$ -- from rule $e\ref{e:ne:ptarc}$ and forced atom proposition
\item Then $q \not< n_i \equiv q \geq n_i$ in $e\ref{e:ne:ptarc}$ for all $\{holds(p,q,0), ptarc(p,t_i,n_i)\} \subseteq A$ -- from $e\ref{e:ne:ptarc}$, forced atom proposition, and the following:
\begin{enumerate}
\item $holds(p,q,0) \in A$ represents $q=M_0(p)$ -- rule $i\ref{i:holds}$ construction
\item $ptarc(p,t_i,n_i) \in A$ represents $n_i=W(p,t_i)$ -- rule $f\ref{f:ptarc} $ construction
\end{enumerate}
\item Then $\forall p \in \bullet t_i, M_0(p) > W(p,t_i)$ -- from definition~\ref{def:pn:preset} of preset $\bullet t_i$ in PN
\item Then $t_i$ is enabled and can fire in $PN$, as a result it can belong to $T_0$ -- from definition~\ref{def:pn:enable} of enabled transition

\end{enumerate}
\item And $consumesmore \notin A$, since $A$ is an answer set of $\Pi^0$ -- from rule $a\ref{a:overc:elim}$ and supported rule proposition
\begin{enumerate}
\item Then $\nexists consumesmore(p,0) \in A$ -- from rule $a\ref{a:overc:gen}$ and supported rule proposition
\item  Then $\nexists \{ holds(p,q,0), tot\_decr(p,q1,0) \} \subseteq A : q1>q$ in $body(a\ref{a:overc:place})$ -- from $a\ref{a:overc:place}$ and forced atom proposition
\item Then $\nexists p : \sum_{t_i \in \{t_0,\dots,t_x\}, p \in \bullet t_i}{W(p,t_i)} > M_0(p)$ -- from the following
\begin{enumerate}
\item $holds(p,q,0)$ represents $q=M_0(p)$ -- from rule $i\ref{i:holds}$ construction, given
\item $tot\_decr(p,q1,0) \in A$ if $\{ del(p,q1_0,t_0,0), \dots del(p,q1_x,t_x,0) \} \subseteq A$, where $q1=q1_0+\dots+q1_x$ -- from $r\ref{r:totdecr}$ and forced atom proposition
\item $del(p,q1_i,t_i,0) \in A$ if $\{ fires(t_i,0), ptarc(p,t_i,q1_i) \} \subseteq A$ -- from $r\ref{r:del}$ and supported rule proposition
\item $del(p,q1_i,t_i,0) \in A$ represents removal of $q1_i = W(p,t_i)$ tokens from $p \in \bullet t_i$ -- from $r\ref{r:del}$, supported rule proposition, and definition~\ref{def:pn:texec} of transition execution in PN
\end{enumerate}
\item Then the set of transitions in $T_0$ do not conflict -- by the definition~\ref{def:pn:conflict} of conflicting transitions
\end{enumerate}

\item Then $\{t_0, \dots, t_x\} = T_0$ -- using 1(a),1(b) above, and definition of firing set 
\end{enumerate}

\item We show $M_1$ is produced by firing $T_0$ in $M_0$. Let $holds(p,q,1) \in A$
\begin{enumerate}
\item Then $\{ holds(p,q1,0), tot\_incr(p,q2,0), tot\_decr(p,q3,0) \} \subseteq A : q=q1+q2-q3$ -- from rule $r\ref{r:nextstate}$ and supported rule proposition \label{x:1}
\item \label{x:2} Then $holds(p,q1,0) \in A$ represents $q1=M_0(p)$ -- given, rule $i\ref{i:holds}$ construction; and 
 $\{add(p,q_0,t_0,0), \dots, $ $add(p,q_j,t_j,0)\} \subseteq A : q_0 + \dots + q_j = q2$ ; %
 and $\{del(p,q_0,t_0,0), \dots, $ $del(p,q_l,t_l,0)\} \subseteq A : q_0 + \dots + q_l = q3$ %
   -- rules $r\ref{r:totincr},r\ref{r:totdecr}$ and supported rule proposition, respectively
\item Then $\{ fires(t_0,0), \dots, fires(t_j,0) \} \subseteq A$ and $\{ fires(t_0,0), \dots, fires(t_l,0) \} $ $\subseteq A$ -- rules $r\ref{r:add},r\ref{r:del}$ and supported rule proposition; and the following
\begin{enumerate}
\item $tparc(t_y,p,q_y) \in A, 0 \leq y \leq j$ represents $q_y=W(t_y,p)$ -- given
\item $ptarc(p,t_z,q_z) \in A, 0 \leq z \leq l$ represents $q_z=W(p,t_z)$ -- given
\end{enumerate}
\item Then $\{ fires(t_0,0), \dots, fires(t_j,0) \} \cup \{ fires(t_0,0), \dots, fires(t_l,0) \} \subseteq A $ \\$= \{ fires(t_0,0), \dots, fires(t_x,0) \} \subseteq A$ -- set union of subsets
\item Then for each $fires(t_x,0) \in A$ we have $t_x \in T_0$ -- already shown in item~\ref{prove:fires_t0} above
\item Then $q = M_0(p) + \sum_{t_x \in T_0 \wedge p \in t_x \bullet}{W(t_x,p)} - \sum_{t_x \in T_0 \wedge p \in \bullet t_x}{W(p,t_x)}$ -- from %
\eqref{x:2} above and the following
\begin{enumerate}
\item Each $add(p,q_j,t_j,0) \in A$ represents $q_j=W(t_j,p)$ for $p \in t_j \bullet$ -- rule $r\ref{r:add}$ encoding, and definition~\ref{def:pn:texec} of transition execution in $PN$ %
\item Each $del(p,t_y,q_y,0) \in A$ represents $q_y=W(p,t_y)$ for $p \in \bullet t_y$ -- from rule $r\ref{r:del}$ encoding, and definition~\ref{def:pn:texec} of transition execution in $PN$
\item Each $tot\_incr(p,q2,0) \in A$ represents $q2=\sum_{t_x \in T_0 \wedge p \in t_x  \bullet}{W(t_x,p)}$ -- aggregate assignment atom semantics in rule $r\ref{r:totincr}$
\item Each $tot\_decr(p,q3,0) \in A$ represents $q3=\sum_{t_x \in T_0 \wedge p \in \bullet t_x}{W(p,t_x)}$ -- aggregate assignment atom semantics in rule $r\ref{r:totdecr}$
\end{enumerate}
\item Then, $M_1(p) = q$ -- since $holds(p,q,1) \in A$ encodes $q=M_1(p)$ -- by construction 
\end{enumerate}
\end{enumerate}

\noindent
{\bf Inductive Step:} Assume $M_k$ is a valid marking in $X$ for $PN$, show 
\begin{inparaenum}[(1)]
\item $T_k$ is a valid firing set for $M_k$, and 
\item firing $T_k$ in $M_k$ produces marking $M_{k+1}$.
\end{inparaenum}

\begin{enumerate}
\item We show that $T_k$ is a valid firing set for $M_k$. Let $\{ fires(t_0,k),\dots,fires(t_x,k) \}$ be the set of all $fires(\dots,k)$ atoms in $A$,
\begin{enumerate}
\item Then for each $fires(t_i,k) \in A$\label{prove:fires_tk}
\begin{enumerate}
\item $enabled(t_i,k) \in A$ -- from rule $a\ref{a:fires}$ and supported rule proposition
\item Then $notenabled(t_i,k) \notin A$ -- from rule $e\ref{e:enabled}$ and supported rule proposition
\item Then body of $e\ref{e:ne:ptarc}$ must hold in $A$ -- from rule $e\ref{e:ne:ptarc}$ and forced proposition
\item Then $q \not< n_i \equiv q \geq n_i$ in $e\ref{e:ne:ptarc}$ for all $\{ holds(p,q,k), ptarc(p,t_i,n_i) \} \subseteq A$ -- from $e\ref{e:ne:ptarc}$ using forced atom proposition, and the following
\begin{enumerate}
\item $holds(p,q,k) \in A$ represents $q=M_k(p)$ -- construction, inductive assumption
\item $ptarc(p,t,n_i) \in A$ represents $n_i=W(p,t)$ -- rule $f\ref{f:tparc}$ construction
\end{enumerate}
\item Then $\forall p \in \bullet t_i, M_k(p) > W(p,t_i)$ -- from definition~\ref{def:pn:preset} of preset $\bullet t_i$ in PN
\item Then $t_i$ is enabled and can fire in $PN$, as a result it can belong to $T_k$ -- from definition~\ref{def:pn:enable} of enabled transition
\end{enumerate}

\item And $consumesmore \notin A$, since $A$ is an answer set of $\Pi^0$ -- from rule $a\ref{a:overc:elim}$ and supported rule proposition
\begin{enumerate}
\item Then $\nexists consumesmore(p,k) \in A$ -- from rule $a\ref{a:overc:gen}$ and forced atom proposition
\item Then $\nexists \{ holds(p,q,k), tot\_decr(p,q1,k) \} \subseteq A : q1 > q$ in $body(a\ref{a:overc:place})$ -- from $a\ref{a:overc:place}$ and forced atom proposition
\item Then $\nexists p : \sum_{t_i \in \{ t_0,\dots,t_x \}, p \in \bullet t_i}{W(p,t_i)} > M_k(p)$ -- from the following
\begin{enumerate}
\item $holds(p,q,k) \in A$ represents $q=M_k(p)$ -- by construction, and the inductive assumption 
\item $tot\_decr(p,q1,k) \in A$ if $\{ del(p,q1_0,t_0,k), \dots, del(p,q1_x,t_x,k) \} \subseteq A$, where $q1=q1_0+\dots+q1_x$ -- from $r\ref{r:totdecr}$ and forced atom proposition
\item $del(p,q1_i,t_i,k) \in A$ if $\{ fires(t_i,k), ptarc(p,t_i,q1_i) \} \subseteq A$ -- from $r\ref{r:del}$ and supported rule proposition
\item $del(p,q1_i,t_i,k)$ represents removal of $q1_i = W(p,t_i)$ tokens from $p \in \bullet t_i$ -- from construction rule $r\ref{r:del}$, supported rule proposition, and definition~\ref{def:pn:texec} of transition execution in $PN$
\end{enumerate}

\item Then the set of transitions $T_k$ does conflict -- by the definition~\ref{def:pn:conflict} of conflicting transitions

\end{enumerate}

\item Then $\{t_0,\dots,t_x\} = T_k$ -- using 1(a),1(b) above
\end{enumerate}

\item We show $M_{k+1}$ is produced by firing $T_k$ in $M_k$. Let $holds(p,q,k+1) \in A$
\begin{enumerate}
\item Then $\{ holds(p,q1,k), tot\_incr(p,q2,k), tot\_decr(p,q3,k) \} \in A : q=q2+q2-q3$ -- from rule $r\ref{r:nextstate}$ and supported rule proposition \label{x:3}
\item \label{x:4}Then $holds(p,q1,k) \in A$ represents $q1=M_k(p)$ -- inductive assumption and construction;
and $\{ add(p,q2_0,t_0,k), \dots, add(p,q2_j,t_j,k) \} \subseteq A : q2_0+\dots+q2_j=q2$ and $\{ del(p,q3_0,t_0,k), \dots, del(p,q3_l,t_l,k) \} \subseteq A : q3_0+\dots+q3_l=q3$ -- from rules $r\ref{r:totincr},r\ref{r:totdecr}$ using supported rule proposition, respectively
\item Then $\{ fires(t_0,k), \dots, fires(t_j,k) \} \subseteq A$ and $\{ fires(t_0,k), \dots, fires(t_l,k) \} $ $\subseteq A$ -- from rules $r\ref{r:add} ,r\ref{r:del} $ using supported rule proposition, respectively
\item Then $\{ fires(t_0,k), \dots, fires(t_j,k) \} \cup \{ fires(t_0,k), \dots, fires(t_l,k) \} $ \\$= \{ fires(t_0,k), \dots, fires(t_x,k) \} \subseteq A$ -- subset union property
\item Then for each $fires(t_x,k) \in A$ we have $t_x \in T_k$ - already shown in item~\ref{prove:max:fires_tk} above
\item Then $q = M_k(p) + \sum_{t_x \in T_k \wedge p \in t_x \bullet}{W(t_x,p)} - \sum_{t_x \in T_k \wedge p \in \bullet t_x}{W(p,t_x}$ -- from %
\eqref{x:4} above and the following
\begin{enumerate}
\item Each $add(p,q_j,t_j,0) \in A$ represents $q_j = W(t_j,p)$ for $p \in t_j \bullet$ -- encoding of $r\ref{r:add}$ and definition~\ref{def:pn:texec} of transition execution in PN
\item Each $del(p,t_y,q_y,0) \in A$ represents $q_y = W(p,t_y)$ for $p \in \bullet t_y$ -- encoding of $r\ref{r:del}$ and definition~\ref{def:pn:texec} of transition execution in PN
\item Each $tot\_incr(p,q2,0) \in A$ represents $q2 = \sum_{t_x \in T_k \wedge p \in t_x \bullet}{W(t_x,p)}$ -- aggregate assignment atom semantics in rule $r\ref{r:totincr}$
\item Each $tot\_decr(p,q3,0) \in A$ represents $q3 = \sum_{t_x \in T_k \wedge p \in \bullet t_x}{W(p,t_x)}$ -- aggregate assignment atom semantics in rule $r\ref{r:totdecr}$
\end{enumerate}
\item Then $M_{k+1}(p) = q$ -- since $holds(p,q,k+1) \in A$ encodes $q=M_{k+1}(p)$ by construction
\end{enumerate}
\end{enumerate}

\noindent
As a result, for any $n > k$, $T_n$ is a valid firing set w.r.t. $M_n$ and its firing produces marking $M_{n+1}$.

\vspace{20pt}
\noindent
{\bf Conclusion:} Since both \ref{prove:x2a} and \ref{prove:a2x} hold, $X=M_0,T_0,M_1,\dots,M_k,T_k,M_{k+1}$ is an execution sequence of $PN(P,T,E,W)$ (w.r.t. $M_0$) iff there is an answer set $A$ of $\Pi^0(PN,M_0,k,ntok)$ such that (\ref{eqn:fires}) and (\ref{eqn:holds}) hold.

\section{Proof of Proposition \ref{prop:max}}

Let $PN=(P,T,E,W)$ be a Petri Net, $M_0$ be its initial marking and let $\Pi^1(PN,M_0,$ $k,ntok)$ be the ASP encoding of $PN$ and $M_0$ over a simulation length $k$, with maximum $ntok$ tokens on any place node, as defined in section~\ref{sec:enc_max}. Then $X=M_0,T_0,M_1,\dots,M_k,T_k,M_{k+1}$ is an execution sequence of $PN$ (w.r.t. $M_0$) iff there is an answer set $A$ of $\Pi^1(PN,M_0,k,ntok)$ such that: 
\begin{equation}
\{ fires(t,ts) : t \in T_{ts}, 0 \leq ts \leq k\} = \{ fires(t,ts) : fires(t,ts) \in A \} \label{eqn:max:fires}
\end{equation}
\begin{equation}
\begin{split}
\{ holds(p,q,ts) &: p \in P, q = M_{ts}(p), 0 \leq ts \leq k+1 \} \\
&= \{ holds(p,q,ts) : holds(p,q,ts) \in A \} \label{eqn:max:holds}
\end{split}
\end{equation}

We prove this by showing that:
\begin{enumerate}[(I)]
\item Given an execution sequence $X$, we create a set $A$ such that it satisfies \eqref{eqn:max:fires} and \eqref{eqn:max:holds} and show that $A$ is an answer set of $\Pi^1$ \label{prove:x2a:max}
\item Given an answer set $A$ of $\Pi^1$, we create an execution sequence $X$ such that \eqref{eqn:max:fires} and \eqref{eqn:max:holds} are satisfied. \label{prove:a2x:max}
\end{enumerate}

\noindent
{\bf First we show (\ref{prove:x2a:max})}: Given a $PN$ and an execution sequence $X$ of $PN$, we create a set $A$ as a union of the following sets:
\begin{enumerate}
\item $A_1=\{ num(n) : 0 \leq n \leq ntok \}$ %
\item $A_2=\{ time(ts) : 0 \leq ts \leq k\}$ %
\item $A_3=\{ place(p) : p \in P \}$ %
\item $A_4=\{ trans(t) : t \in T \}$ %
\item $A_5=\{ ptarc(p,t,n) : (p,t) \in E^-, n=W(p,t) \}$, where $E^- \subseteq E$ %
\item $A_6=\{ tparc(t,p,n) : (t,p) \in E^+, n=W(t,p) \}$, where $E^+ \subseteq E$ %
\item $A_7=\{ holds(p,q,0) : p \in P, q=M_{0}(p) \}$ %
\item $A_8=\{ notenabled(t,ts) : t \in T, 0 \leq ts \leq k, \exists p \in \bullet t, M_{ts}(p) < W(p,t) \}$ \newline per definition~\ref{def:pn:enable} (enabled transition) %
\item $A_9=\{ enabled(t,ts) : t \in T, 0 \leq ts \leq k, \forall p \in \bullet t, W(p,t) \leq M_{ts}(p) \}$ \newline per definition~\ref{def:pn:enable} (enabled transition) %
\item $A_{10}=\{ fires(t,ts) : t \in T_{ts}, 0 \leq ts \leq k \}$ \newline per definition~\ref{def:pn:firing_set} (firing set), only an enabled transition may fire
\item $A_{11}=\{ add(p,q,t,ts) : t \in T_{ts}, p \in t \bullet, q=W(t,p), 0 \leq ts \leq k \}$ \newline per definition~\ref{def:pn:texec} (transition execution)
\item $A_{12}=\{ del(p,q,t,ts) : t \in T_{ts}, p \in \bullet t, q=W(p,t), 0 \leq ts \leq k \}$ \newline per definition~\ref{def:pn:texec} (transition execution)
\item $A_{13}=\{ tot\_incr(p,q,ts) : p \in P, q=\sum_{t \in T_{ts}, p \in t \bullet}{W(t,p)}, 0 \leq ts \leq k \}$ \newline per definition~\ref{def:pn:firing_set} (firing set execution) %
\item $A_{14}=\{ tot\_decr(p,q,ts): p \in P, q=\sum_{t \in T_{ts}, p \in \bullet t}{W(p,t)}, 0 \leq ts \leq k \}$ \newline per definition~\ref{def:pn:firing_set} (firing set execution) %
\item $A_{15}=\{ consumesmore(p,ts) : p \in P, q=M_{ts}(p), q1=\sum_{t \in T_{ts}, p \in \bullet t}{W(p,t)}, q1 > q, 0 \leq ts \leq k \}$ \newline per definition~\ref{def:pn:conflict} (conflicting transitions) for enabled transition set $T_{ts}$ %
\item $A_{16}=\{ consumesmore : \exists p \in P : q=M_{ts}(p), q1=\sum_{t \in T_{ts}, p \in \bullet t}{W(p,t)}, q1 > q, 0 \leq ts \leq k \}$ \newline per definition~\ref{def:pn:conflict} (conflicting transitions) %
\item $A_{17}=\{ could\_not\_have(t,ts) :  t \in T, (\forall p \in \bullet t, W(p,t) \leq M_{ts}(p)), t \not\in T_{ts}, (\exists p \in \bullet t : W(p,t) > M_{ts}(p) - \sum_{t' \in T_{ts}, p \in \bullet t'}{W(p,t')}), 0 \leq ts \leq k \}$ \newline
per the maximal firing set semantics
\item $A_{18}=\{ holds(p,q,ts+1) : p \in P, q=M_{ts+1}(p), 0 \leq ts < k\}$, \newline where $M_{ts+1}(p) = M_{ts}(p) - \sum_{\substack{t \in T_{ts}, p \in \bullet t}}{W(p,t)} + \sum_{\substack{t \in T_{ts}, p \in t \bullet}}{W(t,p)}$ \newline according to definition~\ref{def:pn:exec} (firing set execution) %
\end{enumerate}

\noindent
{\bf We show that $A$ satisfies \eqref{eqn:max:fires} and \eqref{eqn:max:holds}, and $A$ is an answer set of $\Pi^1$.}

$A$ satisfies \eqref{eqn:max:fires} and \eqref{eqn:max:holds} by its construction above. We show $A$ is an answer set of $\Pi^1$ by splitting. We split $lit(\Pi^1)$ into a sequence of $6k+8$ sets:

\begin{itemize}\renewcommand{\labelitemi}{$\bullet$}
\item $U_0= head(f\ref{f:place}) \cup head(f\ref{f:trans}) \cup head(f\ref{f:ptarc}) \cup head(f\ref{f:tparc}) \cup head(f\ref{f:time}) \cup head(f\ref{f:num}) $ $\cup head(i\ref{i:holds}) = \{place(p) : p \in P\} \cup \{ trans(t) : t \in T\} \cup \{ ptarc(p,t,n) : (p,t) \in E^-, n=W(p,t) \} \cup \{  tparc(t,p,n) : (t,p) \in E^+, n=W(t,p) \} \cup $ $ \{ time(0), \dots, time(k)\} \cup \{num(0), \dots, num(ntok)\} \cup \{ holds(p,q,0) : p \in P, q=M_0(p) \} $
\item $U_{6k+1}=U_{6k+0} \cup head(e\ref{e:ne:ptarc})^{ts=k} = U_{6k+0} \cup \{ notenabled(t,k) : t \in T \}$
\item $U_{6k+2}=U_{6k+1} \cup head(e\ref{e:enabled})^{ts=k} = U_{6k+1} \cup \{ enabled(t,k) : t \in T \}$
\item $U_{6k+3}=U_{6k+2} \cup head(a\ref{a:fires})^{ts=k} = U_{6k+2} \cup \{ fires(t,k) : t \in T \}$
\item $U_{6k+4}=U_{6k+3}  \cup head(r\ref{r:add})^{ts=k} \cup head(r\ref{r:del})^{ts=k} = U_{6k+3} \cup \{ add(p,q,t,k) : p \in P, t \in T, q=W(t,p) \} \cup \{ del(p,q,t,k) : p \in P, t \in T, q=W(p,t) \}$
\item $U_{6k+5}=U_{6k+4} \cup head(r\ref{r:totincr})^{ts=k} \cup head(r\ref{r:totdecr})^{ts=k} = U_{6k+4} \cup \{ tot\_incr(p,q,k) : p \in P, 0 \leq q \leq ntok \} \cup \{ tot\_decr(p,q,k) : p \in P, 0 \leq q \leq ntok \}$
\item $U_{6k+6}=U_{6k+5} \cup head(r\ref{r:nextstate})^{ts=k} \cup head(a\ref{a:overc:place})^{ts=k} \cup head(a\ref{a:maxfire:cnh})^{ts=k} $ $= U_{6k+5} \\ \cup \{ consumesmore(p,k) : p \in P\} \cup \{ holds(p,q,k+1) : p \in P, 0 \leq q \leq ntok \} \cup $ $\{ could\_not\_have(t,k) : t \in T \}$ 
\item $U_{6k+7}=U_{6k+6} \cup head(a\ref{a:overc:gen}) = U_{6k+6} \cup \{ consumesmore \}$
\end{itemize}
where $head(r_i)^{ts=k}$ are head atoms of ground rule $r_i$ in which $ts=k$. We write $A_i^{ts=k} = \{ a(\dots,ts) : a(\dots,ts) \in A_i, ts=k \}$ as short hand for all atoms in $A_i$ with $ts=k$. $U_{\alpha}, 0 \leq \alpha \leq 6k+7$ form a splitting sequence, since each $U_i$ is a splitting set of $\Pi^1$, and $\langle U_{\alpha}\rangle_{\alpha < \mu}$ is a monotone continuous sequence, where $U_0 \subseteq U_1 \dots \subseteq U_{6k+7}$ and $\bigcup_{\alpha < \mu}{U_{\alpha}} = lit(\Pi^1)$. 

We compute the answer set of $\Pi^1$ using the splitting sets as follows:
\begin{enumerate}
\item $bot_{U_0}(\Pi^1) = f\ref{f:place} \cup f\ref{f:trans} \cup f\ref{f:ptarc} \cup f\ref{f:tparc} \cup f\ref{f:time} \cup i\ref{i:holds} \cup f\ref{f:num}$ and $X_0 = A_1 \cup \dots \cup A_7$ ($= U_0$) is its answer set -- using forced atom proposition

\item $eval_{U_0}(bot_{U_1}(\Pi^1) \setminus bot_{U_0}(\Pi^1), X_0) = \{ notenabled(t,0) \text{:-} . | \{ trans(t), \\ ptarc(p,t,n), holds(p,q,0) \} \subseteq X_0, \text{~where~}  q < n \}$. Its answer set $X_1=A_8^{ts=0}$ -- using  forced atom proposition and construction of $A_8$.
\begin{enumerate}
\item where, $q=M_0(p)$, and $n=W(p,t)$ for an arc $(p,t) \in E^-$ -- by construction of $i\ref{i:holds}$ and $f\ref{f:ptarc}$ in $\Pi^1$, and 
\item in an arc $(p,t) \in E^-$, $p \in \bullet t$ (by definition~\ref{def:pn:preset} of preset)
\item thus, $notenabled(t,0) \in X_1$ represents $\exists p \in \bullet t : M_0(p) < W(p,t)$.
\end{enumerate}

\item $eval_{U_1}(bot_{U_2}(\Pi^1) \setminus bot_{U_1}(\Pi^1), X_0 \cup X_1) = \{ enabled(t,0) \text{:-}. | trans(t) \in X_0 \cup X_1, $ $notenabled(t,0) \notin X_0 \cup X_1 \}$. Its answer set is $X_2 = A_9^{ts=0}$ -- using forced atom proposition and construction of $A_9$.
\begin{enumerate}
\item since an $enabled(t,0) \in X_2$ if $\nexists ~notenabled(t,0) \in X_0 \cup X_1$, which is equivalent to $\nexists p \in \bullet t : M_0(p) < W(p,t) \equiv \forall p \in \bullet t: M_0(p) \geq W(p,t)$.
\end{enumerate}

\item $eval_{U_2}(bot_{U_3}(\Pi^1) \setminus bot_{U_2}(\Pi^1), X_0 \cup X_1 \cup X_2) = \{\{fires(t,0)\} \text{:-}. | enabled(t,0) \\ \text{~holds in~} X_0 \cup X_1 \cup X_2 \}$. It has multiple answer sets $X_{3.1}, \dots, X_{3.n}$, corresponding to elements of power set of $fires(t,0)$ atoms in $eval_{U_2}(...)$ -- using supported rule proposition. Since we are showing that the union of answer sets of $\Pi^1$ determined using splitting is equal to $A$, we only consider the set that matches the $fires(t,0)$ elements in $A$ and call it $X_3$, ignoring the rest. Thus, $X_3 = A_{10}^{ts=0}$, representing $T_0$.

\item $eval_{U_3}(bot_{U_4}(\Pi^1) \setminus bot_{U_3}(\Pi^1), X_0 \cup \dots \cup X_3) = \{add(p,n,t,0) \text{:-}. | \{fires(t,0), \\ tparc(t,p,n) \} \subseteq X_0 \cup \dots \cup X_3 \} \cup \{ del(p,n,t,0) \text{:-}. | \{ fires(t,0), ptarc(p,t,n) \} \subseteq X_0 \cup \dots \cup X_3 \}$. It's answer set is $X_4 = A_{11}^{ts=0} \cup A_{12}^{ts=0}$ -- using forced atom proposition and definitions of $A_{11}$ and $A_{12}$. 
\begin{enumerate}
\item where, each $add$ atom encodes $n=W(t,p) : p \in t \bullet$, 
\item and each $del$ atom encodes $n=W(p,t) : p \in \bullet t$
\item representing the effect of transitions in $T_0$ -- by construction
\end{enumerate}

\item $eval_{U_4}(bot_{U_5}(\Pi^1) \setminus bot_{U_4}(\Pi^1), X_0 \cup \dots \cup X_4) = \{tot\_incr(p,qq,0) \text{:-}. | $ \\$qq=\sum_{add(p,q,t,0) \in X_0 \cup \dots \cup X_4}{q} \} \cup \{ tot\_decr(p,qq,0) \text{:-}. | qq=\sum_{del(p,q,t,0) \in X_0 \cup \dots \cup X_4}{q} \}$. It's answer set is $X_5 = A_{13}^{ts=0} \cup A_{14}^{ts=0}$ --  using forced atom proposition and definitions of $A_{13}$, $A_{14}$, and 
semantics of aggregate assignment atom
\begin{enumerate}
\item where, each $tot\_incr(p,qq,0)$, $qq=\sum_{add(p,q,t,0) \in X_0 \cup \dots X_4}{q}$ \\$\equiv qq=\sum_{t \in X_3, p \in t \bullet}{W(p,t)}$, 
\item and, each $tot\_decr(p,qq,0)$, $qq=\sum_{del(p,q,t,0) \in X_0 \cup \dots X_4}{q}$ \\$\equiv qq=\sum_{t \in X_3, p \in \bullet t}{W(t,p)}$, 
\item represent the net effect of actions in $T_0$ -- by construction
\end{enumerate}
\item $eval_{U_5}(bot_{U_6}(\Pi^1) \setminus bot_{U_5}(\Pi^1), X_0 \cup \dots \cup X_5) = \{ consumesmore(p,0) \text{:-}. | \\ \{holds(p,q,0), tot\_decr(p,q1,0) \} \subseteq X_0 \cup \dots \cup X_5 : q1 > q \} \cup \{ holds(p,q,1) \text{:-}., | \\ \{ holds(p,q1,0), tot\_incr(p,q2,0), tot\_decr(p,q3,0) \} \subseteq X_0 \cup \dots \cup X_5, q=q1+q2-q3 \} \cup \{ could\_not\_have(t,0) \text{:-}. | \{ enabled(t,0), ptarc(s,t,q), holds(s,qq,0), \\ tot\_decr(s,qqq,0) \} \subseteq X_0 \cup \dots \cup X_5, fires(t,0) \notin (X_0 \cup \dots \cup X_5), q > qq-qqq \}$. It's answer set is $X_6 = A_{15}^{ts=0} \cup A_{17}^{ts=0} \cup A_{18}^{ts=0}$ -- using forced atom proposition and definitions of $A_{15}, A_{17}, A_{18}$.
\begin{enumerate}
\item where, $consumesmore(p,0)$ represents $\exists p : q=M_0(p), q1=\sum_{t \in T_0, p \in \bullet t}{W(p,t)}, $ $q1 > q$ indicating place $p$ will be overconsumed if $T_0$ is fired, as defined in definition~\ref{def:pn:conflict} (conflicting transitions) 
\item and, $holds(p,q,1)$ encodes $q=M_1(p)$ -- by construction
\item and $could\_not\_have(t,0)$ represents an enabled transition $t$ in $T_0$ that could not fire due to insufficient tokens
\item $X_6$ does not contain $could\_not\_have(t,0)$, when $enabled(t,0) \in X_0 \cup \dots \cup X_5$ and $fires(t,0) \notin X_0 \cup \dots \cup X_5$ due to construction of $A$, encoding of $a\ref{a:maxfire:cnh}$ and its body atoms. As a result it is not eliminated by the constraint $a\ref{a:maxfire:elim}$
\end{enumerate}

\[ \vdots \]

\item $eval_{U_{6k+0}}(bot_{U_{6k+1}}(\Pi^1) \setminus bot_{U_{6k+0}}(\Pi^1), X_0 \cup \dots \cup X_{6k+0}) = \\ \{ notenabled(t,k) \text{:-} . | \{ trans(t), ptarc(p,t,n), holds(p,q,k) \} \subseteq X_0 \cup \dots \cup X_{6k+0}, \\ \text{~where~}  q < n \}$. Its answer set $X_{6k+1}=A_8^{ts=k}$ -- using forced atom proposition and construction of $A_8$.
\begin{enumerate}
\item where, $q=M_k(p)$, and $n=W(p,t)$ for an arc $(p,t) \in E^-$ -- by construction of $holds$ and $ptarc$ predicates in $\Pi^1$, and 
\item in an arc $(p,t) \in E^-$, $p \in \bullet t$ (by definition~\ref{def:pn:preset} of preset)
\item thus, $notenabled(t,k) \in X_{6k+1}$ represents $\exists p \in \bullet t : M_k(p) < W(p,t)$.
\end{enumerate}

\item $eval_{U_{6k+1}}(bot_{U_{6k+2}}(\Pi^1) \setminus bot_{U_{6k+1}}(\Pi^1), X_0 \cup \dots \cup X_{6k+1}) = \{ enabled(t,k) \text{:-}. | \\ trans(t) \in X_0 \cup \dots \cup X_{6k+1}, notenabled(t,k) \notin X_0 \cup \dots \cup X_{6k+1} \}$. Its answer set is $X_{6k+2} = A_9^{ts=k}$ -- using forced atom proposition and construction of $A_9$.
\begin{enumerate}
\item since an $enabled(t,k) \in X_{6k+2}$ if $\nexists ~notenabled(t,k) \in X_0 \cup \dots \cup X_{6k+1}$, which is equivalent to $\nexists p \in \bullet t : M_k(p) < W(p,t) \equiv \forall p \in \bullet t: M_k(p) \geq W(p,t)$.
\end{enumerate}

\item $eval_{U_{6k+2}}(bot_{U_{6k+3}}(\Pi^1) \setminus bot_{U_{6k+2}}(\Pi^1), X_0 \cup \dots \cup X_{6k+2}) = \\ \{\{fires(t,k)\} \text{:-}. | enabled(t,k) \text{~holds in~} X_0 \cup \dots \cup X_{6k+2} \}$. It has multiple answer sets $X_{{6k+3}.1}, \dots, X_{{6k+3}.n}$, corresponding to elements of power set of $fires(t,k)$ atoms in $eval_{U_{6k+2}}(...)$ -- using supported rule proposition. Since we are showing that the union of answer sets of $\Pi^1$ determined using splitting is equal to $A$, we only consider the set that matches the $fires(t,k)$ elements in $A$ and call it $X_{6k+3}$, ignoring the reset. Thus, $X_{6k+3} = A_{10}^{ts=k}$, representing $T_k$.

\item $eval_{U_{6k+3}}(bot_{U_{6k+4}}(\Pi^1) \setminus bot_{U_{6k+3}}(\Pi^1), X_0 \cup \dots \cup X_{6k+3}) = \\ \{add(p,n,t,k) \text{:-}. | \{fires(t,k), tparc(t,p,n) \} \subseteq X_0 \cup \dots \cup X_{6k+3} \} \cup \\ \{ del(p,n,t,k) \text{:-}. | \{ fires(t,k), ptarc(p,t,n) \} \subseteq X_0 \cup \dots \cup X_{6k+3} \}$. It's answer set is $X_{6k+4} = A_{11}^{ts=k} \cup A_{12}^{ts=k}$ -- using forced atom proposition and definitions of $A_{11}$ and $A_{12}$. 
\begin{enumerate}
\item where, each $add$ atom is equivalent to $n=W(t,p) : p \in t \bullet$, 
\item and, each $del$ atom is equivalent to $n=W(p,t) : p \in \bullet t$,
\item representing the effect of transitions in $T_k$
\end{enumerate}

\item $eval_{U_{6k+4}}(bot_{U_{6k+5}}(\Pi^1) \setminus bot_{U_{6k+4}}(\Pi^1), X_0 \cup \dots \cup X_{6k+4}) = \{tot\_incr(p,qq,k) \text{:-}. | \\ qq=\sum_{add(p,q,t,k) \in X_0 \cup \dots \cup X_{6k+4}}{q} \} \cup \{ tot\_decr(p,qq,k) \text{:-}. | \\ qq=\sum_{del(p,q,t,k) \in X_0 \cup \dots \cup X_{6k+4}}{q} \}$. It's answer set is $X_{6k+5} = A_{13}^{ts=k} \cup A_{14}^{ts=k}$ --  using forced atom proposition and definitions of $A_{13}$ and $A_{14}$.
\begin{enumerate}
\item where, each $tot\_incr(p,qq,k)$, $qq=\sum_{add(p,q,t,k) \in X_0 \cup \dots X_{6k+4}}{q}$ $\equiv \\ qq=\sum_{t \in X_{6k+3}, p \in t \bullet}{W(p,t)}$, 
\item and, each $tot\_decr(p,qq,k)$, $qq=\sum_{del(p,q,t,k) \in X_0 \cup \dots X_{6k+4}}{q}$ $\equiv \\ qq=\sum_{t \in X_{6k+3}, p \in \bullet t}{W(t,p)}$, 
\item represent the net effect of transitions in $T_k$
\end{enumerate}
\item $eval_{U_{6k+5}}(bot_{U_{6k+6}}(\Pi^1) \setminus bot_{U_{6k+5}}(\Pi^1), X_0 \cup \dots \cup X_{6k+5}) = \\ \{ consumesmore(p,k) \text{:-}. |  \{holds(p,q,k), tot\_decr(p,q1,k) \} \subseteq X_0 \cup \dots \cup X_{6k+5} : q1 > q \} \cup \{ holds(p,q,k+1) \text{:-}., |  \{ holds(p,q1,k), tot\_incr(p,q2,k), \\ tot\_decr(p,q3,k) \} \subseteq X_0 \cup \dots \cup X_{6k+5} : q=q1+q2-q3 \} \cup \{ could\_not\_have(t,k) \text{:-} \\ \{ enabled(t,k), ptarc(s,t,q), holds(s,qq,k), tot\_decr(s,qqq,k) \} \subseteq X_0 \cup \dots \cup X_{6k+5}, $ $fires(t,k) \notin (X_0 \cup \dots \cup X_{6k+5}), q > qq-qqq \}$. It's answer set is $X_{6k+6} = A_{15}^{ts=k} \cup A_{17}^{ts=k} \cup A_{18}^{ts=k}$ -- using forced atom proposition.
\begin{enumerate}
\item where, $consumesmore(p,k)$ represents $\exists p : q=M_{k}(p), \\ q1=\sum_{t \in T_{k}, p \in \bullet t}{W(p,t)}, q1 > q$
\item $holds(p,q,k+1)$ represents $q=M_{k+1}(p)$ indicating place $p$ that will be over consumed if $T_k$ is fired, as defined in definition~\ref{def:pn:conflict} (conflicting transitions), 
\item $holds(p,q,k+1)$ represents $q=M_{k+1}(p)$ -- by construction
\item and $could\_not\_have(t,k)$ represents an enabled transition $t$ in $T_k$ that could not fire due to insufficient tokens
\item $X_{6k+6}$ does not contain $could\_not\_have(t,k)$, when $enabled(t,k) \in X_0 \cup \dots \cup X_{6k+5}$ and $fires(t,k) \notin X_0 \cup \dots \cup X_{6k+5}$ due to construction of $A$, encoding of $a\ref{a:maxfire:cnh}$ and its body atoms. As a result it is note eliminated by the constraint $a\ref{a:maxfire:elim}$
\end{enumerate}

\item $eval_{U_{6k+6}}(bot_{U_{6k+7}}(\Pi^1) \setminus bot_{U_{6k+6}}(\Pi^1), X_0 \cup \dots \cup X_{6k+6}) = \{ consumesmore \text{:-}. | $ \\$\{ consumesmore(p,0),$ $\dots, $ $consumesmore(p,k) \} \cap $ $(X_0 \cup \dots \cup X_{6k+6}) \neq \emptyset \}$. It's answer set is $X_{6k+7} = A_{16}$ -- using forced atom proposition
\begin{enumerate}
\item $X_{6k+7}$ will be empty since none of $consumesmore(p,0),\dots, \\ consumesmore(p,k)$ hold in $X_0 \cup \dots \cup X_{6k+6}$ due to the construction of $A$, encoding of $a\ref{a:overc:place}$ and its body atoms. As a result, it is not eliminated by the constraint $a\ref{a:overc:elim}$
\end{enumerate}

\end{enumerate}

The set $X = X_0 \cup \dots \cup X_{6k+7}$ is the answer set of $\Pi^0$ by the splitting sequence theorem~\ref{def:split_seq_thm}. Each $X_i, 0 \leq i \leq 6k+7$ matches a distinct portion of $A$, and $X = A$, thus $A$ is an answer set of $\Pi^1$.

\vspace{30pt}
\noindent
{\bf Next we show (\ref{prove:a2x:max}):} Given $\Pi^1$ be the encoding of a Petri Net $PN(P,T,E,W)$ with initial marking $M_0$, and $A$ be an answer set of $\Pi^1$ that satisfies (\ref{eqn:max:fires}) and (\ref{eqn:max:holds}), then we can construct $X=M_0,T_0,\dots,M_k,T_k,M_{k+1}$ from $A$, such that it is an execution sequence of $PN$.

We construct the $X$ as follows:
\begin{enumerate}
\item $M_i = (M_i(p_0), \dots, M_i(p_n))$, where $\{ holds(p_0,M_i(p_0),i), \dots holds(p_n,M_i(p_n),i) \} \\ \subseteq A$, for $0 \leq i \leq k+1$
\item $T_i = \{ t : fires(t,i) \in A\}$, for $0 \leq i \leq k$ 
\end{enumerate}
and show that $X$ is indeed an execution sequence of $PN$. We show this by induction over $k$ (i.e. given $M_k$, $T_k$ is a valid firing set and its firing produces marking $M_{k+1}$).

\vspace{20pt}

\noindent
{\bf Base case:} Let $k=0$, and $M_0$ is a valid marking in $X$ for $PN$, show
\begin{inparaenum}[(1)]
\item $T_0$ is a valid firing set for $M_0$, and 
\item $T_0$'s firing w.r.t. marking $M_0$ produces $M_1$. 
\end{inparaenum} 

\noindent
\begin{enumerate}
\item We show $T_0$ is a valid firing set for $M_0$. Let $\{ fires(t_0,0), \dots, fires(t_x,0) \}$ be the set of all $fires(\dots,0)$ atoms in $A$, \label{prove:max:fires_t0}

\begin{enumerate}
\item Then for each $fires(t_i,0) \in A$

\begin{enumerate}
\item $enabled(t_i,0) \in A$ -- from rule $a\ref{a:fires}$ and supported rule proposition
\item Then $notenabled(t_i,0) \notin A$ -- from rule $e\ref{e:enabled}$ and supported rule proposition
\item Then $body(e\ref{e:ne:ptarc})$ must not hold in $A$ -- from rule $e\ref{e:ne:ptarc}$ and forced atom proposition
\item Then $q \not< n_i \equiv q \geq n_i$ in $e\ref{e:ne:ptarc}$ for all $\{holds(p,q,0), ptarc(p,t_i,n_i)\} \subseteq A$ -- from $e\ref{e:ne:ptarc}$, forced atom proposition, and the following
\begin{enumerate}
\item $holds(p,q,0) \in A$ represents $q=M_0(p)$ -- rule $i\ref{i:holds}$ construction
\item $ptarc(p,t_i,n_i) \in A$ represents $n_i=W(p,t_i)$ -- rule $f\ref{f:ptarc} $ construction
\end{enumerate}
\item Then $\forall p \in \bullet t_i, M_0(p) > W(p,t_i)$ -- from definition~\ref{def:pn:preset} of preset $\bullet t_i$ in PN
\item Then $t_i$ is enabled and can fire in $PN$, as a result it can belong to $T_0$ -- from definition~\ref{def:pn:enable} of enabled transition

\end{enumerate}
\item And $consumesmore \notin A$, since $A$ is an answer set of $\Pi^1$ -- from rule $a\ref{a:overc:elim}$ and supported rule proposition
\begin{enumerate}
\item Then $\nexists consumesmore(p,0) \in A$ -- from rule $a\ref{a:overc:gen}$ and supported rule proposition
\item  Then $\nexists \{ holds(p,q,0), tot\_decr(p,q1,0) \} \subseteq A : q1>q$ in $body(a\ref{a:overc:place})$ -- from $a\ref{a:overc:place}$ and forced atom proposition
\item Then $\nexists p : \sum_{t_i \in \{t_0,\dots,t_x\}, p \in \bullet t_i}{W(p,t_i)} > M_0(p)$ -- from the following
\begin{enumerate}
\item $holds(p,q,0)$ represents $q=M_0(p)$ -- from rule $i\ref{i:holds}$ construction, given
\item $tot\_decr(p,q1,0) \in A$ if $\{ del(p,q1_0,t_0,0), \dots, del(p,q1_x,t_x,0) \} \subseteq A$, where $q1 = q1_0+\dots+q1_x$ -- from $r\ref{r:totdecr}$ and forced atom proposition
\item $del(p,q1_i,t_i,0) \in A$ if $\{ fires(t_i,0), ptarc(p,t_i,q1_i) \} \subseteq A$ -- from $r\ref{r:del}$ and supported rule proposition
\item $del(p,q1_i,t_i,0)$ represents removal of $q1_i = W(p,t_i)$ tokens from $p \in \bullet t_i$ -- from rule $r\ref{r:del}$, supported rule proposition, and definition~\ref{def:pn:texec} of transition execution in $PN$
\end{enumerate}
\item Then the set of transitions in $T_0$ do not conflict -- by the definition~\ref{def:pn:conflict} of conflicting transitions
\end{enumerate}

\item And for each $enabled(t_j,0) \in A$ and $fires(t_j,0) \notin A$, $could\_not\_have(t_j,0) \in A$, since $A$ is an answer set of $\Pi^1$ - from rule $a\ref{a:maxfire:elim}$ and supported rule proposition
\begin{enumerate}
\item Then $\{ enabled(t_j,0), holds(s,qq,0), ptarc(s,t_j,q,0), \\ tot\_decr(s,qqq,0) \} \subseteq A$, such that $q > qq - qqq$ and $fires(t_j,0) \notin A$ - from rule $a\ref{a:maxfire:cnh}$ and supported rule proposition
\item Then for an $s \in \bullet t_j$, $W(s,t_j) > M_0(s) - \sum_{t_i \in T_0, s \in \bullet t_i}{W(s,t_i)}$ - from the following:%
\begin{enumerate}
\item $ptarc(s,t_i,q)$ represents $q=W(s,t_i)$ -- from rule $f\ref{f:r:ptarc}$ construction
\item $holds(s,qq,0)$ represents $qq=M_0(s)$ -- from $i\ref{i:holds}$ construction
\item $tot\_decr(s,qqq,0) \in A$ if $\{ del(s,qqq_0,t_0,0), \dots, del(s,qqq_x,t_x,0) \} \subseteq A$ -- from rule $r\ref{r:totdecr}$ construction and supported rule proposition
\item $del(s,qqq_i,t_i,0) \in A$ if $\{ fires(t_i,0), ptarc(s,t_i,qqq_i) \} \subseteq A$ -- from rule $r\ref{r:r:del} $ and supported rule proposition
\item $del(s,qqq_i,t_i,0)$ represents $qqq_i = W(s,t_i) : t_i \in T_0, (s,t_i) \in E^-$  -- from rule $f\ref{f:r:ptarc}$ construction 
\item $tot\_decr(q,qqq,0)$ represents $\sum_{t_i \in T_0, s \in \bullet t_i}{W(s,t_i)}$ -- from (C,D,E) above 
\end{enumerate}

\item Then firing $T_0 \cup \{ t_j \}$ would have required more tokens than are present at its source place $s \in \bullet t_j$. Thus, $T_0$ is a maximal set of transitions that can simultaneously fire.
\end{enumerate}

\item Then $\{t_0, \dots, t_x\} = T_0$ -- using 1(a),1(b) above; and using 1(c) it is a maximal firing set  
\end{enumerate}

\item We show $M_1$ is produced by firing $T_0$ in $M_0$. Let $holds(p,q,1) \in A$
\begin{enumerate}
\item Then $\{ holds(p,q1,0), tot\_incr(p,q2,0), tot\_decr(p,q3,0) \} \subseteq A : q=q1+q2-q3$ -- from rule $r\ref{r:nextstate}$ and supported rule proposition \label{max:x:1}
\item \label{max:x:2} Then, $holds(p,q1,0) \in A$ represents $q1=M_0(p)$ -- given, rule $i\ref{i:holds}$ construction; and  
$\{add(p,q2_0,t_0,0), \dots, $ $add(p,q2_j,t_j,0)\} \subseteq A : q2_0 + \dots + q2_j = q2$  %
and $\{del(p,q3_0,t_0,0), \dots, $ $del(p,q3_l,t_l,0)\} \subseteq A : q3_0 + \dots + q3_l = q3$ %
  -- rules $r\ref{r:totincr},r\ref{r:totdecr}$ using supported rule proposition
\item Then $\{ fires(t_0,0), \dots, fires(t_j,0) \} \subseteq A$ and $\{ fires(t_0,0), \dots, fires(t_l,0) \} \\ \subseteq A$ -- rules $r\ref{r:add},r\ref{r:del}$ and supported rule proposition, respectively
\item Then $\{ fires(t_0,0), \dots, $ $fires(t_j,0) \} \cup \{ fires(t_0,0), \dots, $ $fires(t_l,0) \} \subseteq A = \{ fires(t_0,0), \dots, $ $fires(t_x,0) \} \subseteq A$ -- set union of subsets
\item Then for each $fires(t_x,0) \in A$ we have $t_x \in T_0$ -- already shown in item~\ref{prove:max:fires_t0} above
\item Then $q = M_0(p) + \sum_{t_x \in T_0 \wedge p \in t_x \bullet}{W(t_x,p)} - \sum_{t_x \in T_0 \wedge p \in \bullet t_x}{W(p,t_x)}$ -- from %
\eqref{max:x:2} above and the following
\begin{enumerate}
\item Each $add(p,q_j,t_j,0) \in A$ represents $q_j=W(t_j,p)$ for $p \in t_j \bullet$ -- rule $r\ref{r:add}$ encoding, and definition~\ref{def:pn:texec} of transition execution in $PN$ %
\item Each $del(p,t_y,q_y,0) \in A$ represents $q_y=W(p,t_y)$ for $p \in \bullet t_y$ -- from rule $r\ref{r:del}$ encoding, and definition~\ref{def:pn:texec} of transition execution in $PN$
\item Each $tot\_incr(p,q2,0) \in A$ represents $q2=\sum_{t_x \in T_0 \wedge p \in t_x  \bullet}{W(t_x,p)}$ -- aggregate assignment atom semantics in rule $r\ref{r:totincr}$
\item Each $tot\_decr(p,q3,0) \in A$ represents $q3=\sum_{t_x \in T_0 \wedge p \in \bullet t_x}{W(p,t_x)}$ -- aggregate assignment atom semantics in rule $r\ref{r:totdecr}$
\end{enumerate}
\item Then, $M_1(p) = q$ -- since $holds(p,q,1) \in A$ encodes $q=M_1(p)$  from construction
\end{enumerate}
\end{enumerate}

\noindent
{\bf Inductive Step:} Let $k > 0$, and $M_k$ is a valid marking in $X$ for $PN$, show 
\begin{inparaenum}[(1)]
\item $T_k$ is a valid firing set for $M_k$, and 
\item $T_k$'s firing in $M_k$ produces marking $M_{k+1}$.
\end{inparaenum}

\begin{enumerate}
\item We show $T_k$ is a valid firing set. Let $\{ fires(t_0,k),\dots,fires(t_x,k) \}$ be the set of all $fires(\dots,k)$ atoms in $A$,\label{prove:max:fires_tk}
\begin{enumerate}
\item Then for each $fires(t_i,k) \in A$%
\begin{enumerate}
\item $enabled(t_i,k) \in A$ -- from rule $a\ref{a:fires}$ and supported rule proposition
\item Then $notenabled(t_i,k) \notin A$ -- from rule $e\ref{e:enabled}$ and supported rule proposition
\item Then $body(e\ref{e:ne:ptarc})$ must hold in $A$ -- from rule $e\ref{e:ne:ptarc}$ and forced atom proposition
\item Then $q \not< n_i \equiv q \geq n_i$ in $e\ref{e:ne:ptarc}$ for all $\{ holds(p,q,k), ptarc(p,t_i,n) \} \subseteq A$ -- from $e\ref{e:ne:ptarc}$, forced atom proposition, and the following
\begin{enumerate}
\item $holds(p,q,k) \in A$ represents $q=M_k(p)$ -- construction, inductive assumption
\item $ptarc(p,t,n_i) \in A$ represents $n_i=W(p,t_i)$ -- rule $f\ref{f:ptarc}$ construction
\end{enumerate}
\item Then $\forall p \in \bullet t_i$, $M_k(p) \geq W(p,t_i)$ -- from definition~\ref{def:pn:preset} of preset $\bullet t_i$ in $PN$
\item Then $t_i$ is enabled and can fire in $PN$, as a result it can belong to $T_k$ -- from definition~\ref{def:pn:enable} of enabled transition
\end{enumerate}

\item And $consumesmore \notin A$, since $A$ is an answer set of $\Pi^1$ -- from rule $a\ref{a:overc:elim}$ and supported rule proposition
\begin{enumerate}
\item Then $\nexists consumesmore(p,k) \in A$ -- from rule $a\ref{a:overc:gen}$ and forced atom proposition
\item Then $\nexists \{ holds(p,q,k), tot\_decr(p,q1,k) \} \subseteq A : q1 > q$ in $body(a\ref{a:overc:place})$ -- from $a\ref{a:overc:place}$ and forced atom proposition
\item Then $\nexists p : \sum_{t_i \in \{ t_0,\dots,t_x \}, p \in \bullet t_i}{W(p,t_i)} > M_k(p)$ -- from the following
\begin{enumerate}
\item $holds(p,q,k) \in A$ represents $q=M_k(p)$ -- by construction of $\Pi^1$, and the inductive assumption about $M_k(p)$ 
\item $tot\_decr(p,q1,k) \in A$ if $\{ del(p,q1_0,t_0,k), \dots, del(p,q1_x,t_x,k) \} \subseteq A$, where $q1=q1_0+\dots+q1_x$ -- from $r\ref{r:totdecr}$ and forced atom proposition
\item $del(p,q1_i,t_i,k) \in A$ if $\{ fires(t_i,k), ptarc(p,t_i,q1_i) \} \subseteq A$ -- from $r\ref{r:del}$ and supported rule proposition
\item $del(p,q1_i,t_i,k)$ represents removal of $q1_i = W(p,t_i)$ tokens from $p \in \bullet t_i$ -- from construction rule $r\ref{r:del}$, supported rule proposition, and definition~\ref{def:pn:texec} of transition execution in $PN$
\end{enumerate}

\item Then the set of transitions $T_k$ do not conf -- by the definition~\ref{def:pn:conflict} of conflicting transitions
\end{enumerate}

\item And for each $enabled(t_j,k) \in A$ and $fires(t_j,k) \notin A$, $could\_not\_have(t_j,k) \in A$, since $A$ is an answer set of $\Pi^1$ - from rule $a\ref{a:maxfire:elim}$ and supported rule proposition
\begin{enumerate}
\item Then $\{ enabled(t_j,k), holds(s,qq,k), ptarc(s,t_j,q,k), \\ tot\_decr(s,qqq,k) \} \subseteq A$, such that $q > qq - qqq$ and $fires(t_j,0) \notin A$ - from rule $a\ref{a:maxfire:cnh}$ and supported rule proposition
\item Then for an $s \in \bullet t_j$, $W(s,t_j) > M_k(s) - \sum_{t_i \in T_k, s \in \bullet t_i}{W(s,t_i)}$ - from the following:%
\begin{enumerate}
\item $ptarc(s,t_i,q)$ represents $q=W(s,t_i)$ -- from rule $f\ref{f:r:ptarc}$ construction
\item $holds(s,qq,k)$ represents $qq=M_k(s)$ -- from $i\ref{i:holds}$ construction
\item $tot\_decr(s,qqq,k) \in A$ if $\{ del(s,qqq_0,t_0,k), \dots, del(s,qqq_x,t_x,k) \} \subseteq A$ -- from rule $r\ref{r:totdecr}$ construction and supported rule proposition
\item $del(s,qqq_i,t_i,k) \in A$ if $\{ fires(t_i,k), ptarc(s,t_i,qqq_i) \} \subseteq A$ -- from rule $r\ref{r:r:del} $ and supported rule proposition
\item $del(s,qqq_i,t_i,k)$ represents $qqq_i = W(s,t_i) : t_i \in T_k, (s,t_i) \in E^-$  -- from rule $f\ref{f:r:ptarc}$ construction 
\item $tot\_decr(q,qqq,k)$ represents $\sum_{t_i \in T_k, s \in \bullet t_i}{W(s,t_i)}$ -- from (C,D,E) above 
\end{enumerate}

\item Then firing $T_k \cup \{ t_j \}$ would have required more tokens than are present at its source place $s \in \bullet t_j$. Thus, $T_k$ is a maximal set of transitions that can simultaneously fire.
\end{enumerate}

\item Then $\{ t_0,\dots,t_x \} = T_k$ -- using 1(a),1(b) above; and using 1(c) it is a maximal firing set

\end{enumerate}

\item We show $M_{k+1}$ is produced by firing $T_k$ in $M_k$. Let $holds(p,q,k+1) \in A$
\begin{enumerate}
\item Then $\{ holds(p,q1,k), tot\_incr(p,q2,k), tot\_decr(p,q3,k) \} \in A : q=q2+q2-q3$ -- from rule $r\ref{r:nextstate}$ and supported rule proposition \label{max:x:3}
\item \label{max:x:4} $holds(p,q1,k) \in A$ represents $q1=M_k(p)$ -- construction, inductive assumption; 
and $\{ add(p,q2_0,t_0,k), \dots, $ $add(p,q2_j,t_j,k) \} \subseteq A : q2_0+\dots+q2_j=q2$ and $\{ del(p,q3_0,t_0,k), \dots, $ $del(p,q3_l,t_l,k) \} \subseteq A : q3_0+\dots+q3_l=q3$ -- from rules $r\ref{r:totincr},r\ref{r:totdecr}$ using supported rule proposition, respectively
\item Then $\{ fires(t_0,k), \dots, fires(t_j,k) \} \subseteq A$ and $\{ fires(t_0,k), \dots, fires(t_l,k) \} \\ \subseteq A$ -- from rules $r\ref{r:add} ,r\ref{r:del} $ using supported rule proposition, respectively
\item Then $\{ fires(t_0,k), \dots, $ $fires(t_j,k) \} \cup \{ fires(t_0,k), \dots, $ $fires(t_l,k) \} = \{ fires(t_0,k), \dots, $ $fires(t_x,k) \} \subseteq A$ -- subset union property
\item Then for each $fires(t_x,k) \in A$ we have $t_x \in T_k$ - already shown in item~\eqref{prove:fires_tk} above
\item Then $q = M_k(p) + \sum_{t_x \in T_k \wedge p \in t_x \bullet}{W(t_x,p)} - \sum_{t_x \in T_k \wedge p \in \bullet t_x}{W(p,t_x)}$ -- from %
\eqref{max:x:4} above and the following
\begin{enumerate}
\item Each $add(p,q_j,t_j,k) \in A$ represents $q_j = W(t_j,p)$ for $p \in t_j \bullet$ -- rule $r\ref{r:add}$ encoding and definition~\ref{def:pn:texec} of transition execution in PN
\item Each $del(p,t_y,q_y,k) \in A$ represents $q_y = W(p,t_y)$ for $p \in \bullet t_y$ -- rule $r\ref{r:del}$ encoding and definition~\ref{def:pn:texec} of transition execution in PN
\item Each $tot\_incr(p,q2,k) \in A$ represents $q2 = \sum_{t_x \in T_k \wedge p \in t_x \bullet}{W(t_x,p)}$ -- aggregate assignment atom semantics in rule $r\ref{r:totincr}$
\item Each $tot\_decr(p,q3,k) \in A$ represents $q3 = \sum_{t_x \in T_k \wedge p \in \bullet t_x}{W(p,t_x)}$ -- aggregate assignment atom semantics in rule $r\ref{r:totdecr}$
\end{enumerate}
\item Then $M_{k+1}(p) = q$ -- since $holds(p,q,k+1) \in A$ encodes $q=M_{k+1}(p)$ -- from construction
\end{enumerate}
\end{enumerate}

\noindent
As a result, for any $n > k$, $T_n$ will be a valid firing set w.r.t. $M_n$ and its firing produces marking $M_{n+1}$.

\vspace{20pt}
\noindent
{\bf Conclusion:} Since both \ref{prove:x2a:max} and \ref{prove:a2x:max} hold, $X=M_0,T_0,M_1,\dots,M_k,T_k,M_{k+1}$ is an execution sequence of $PN(P,T,E,W)$ (w.r.t. $M_0$) iff there is an answer set $A$ of $\Pi^1(PN,M_0,k,ntok)$ such that (\ref{eqn:max:fires}) and (\ref{eqn:max:holds}) hold.

\section{Proof of Proposition~\ref{prop:reset}}
Let $PN=(P,T,E,W,R)$ be a Petri Net, $M_0$ be its initial marking and let $\Pi^2(PN,M_0,k,ntok)$ by the ASP encoding of $PN$ and $M_0$ over a simulation length $k$, with maximum $ntok$ tokens on any place node, as defined in section~\ref{sec:enc_reset}. Then $X=M_0,T_0,M_1,\dots,M_k,T_k,M_{k+1}$ is an execution sequence of $PN$ (w.r.t $M_0$) iff there is an answer set $A$ of $\Pi^2(PN,M_0,k,ntok)$ such that:
\begin{equation}
\{ fires(t,ts) : t \in T_{ts}, 0 \leq ts \leq k\} = \{ fires(t,ts) : fires(t,ts) \in A \} \label{eqn:reset:fires}
\end{equation}
\begin{equation}
\begin{split}
\{ holds(p,q,ts) &: p \in P, q = M_{ts}(p), 0 \leq ts \leq k+1 \} \\
&= \{ holds(p,q,ts) : holds(p,q,ts) \in A \} \label{eqn:reset:holds}
\end{split}
\end{equation}

\noindent
We prove this by showing that:
\begin{enumerate}[(I)]
\item Given an execution sequence $X$, we create a set $A$ such that it satisfies \eqref{eqn:reset:fires} and \eqref{eqn:reset:holds} and show that $A$ is an answer set of $\Pi^2$ \label{prove:x2a:reset}
\item Given an answer set $A$ of $\Pi^2$, we create an execution sequence $X$ such that \eqref{eqn:reset:fires} and \eqref{eqn:reset:holds} are satisfied. \label{prove:a2x:reset}
\end{enumerate}

\noindent
{\bf First we show (\ref{prove:x2a:reset})}: Given $PN$ and an execution sequence $X$ of $PN$, we create a set $A$ as a union of the following sets:
\begin{enumerate}
\item $A_1=\{ num(n) : 0 \leq n \leq ntok \}$ %
\item $A_2=\{ time(ts) : 0 \leq ts \leq k\}$ %
\item $A_3=\{ place(p) : p \in P \}$ %
\item $A_4=\{ trans(t) : t \in T \}$ %
\item $A_5=\{ ptarc(p,t,n,ts) : (p,t) \in E^-, n=W(p,t), 0 \leq ts \leq k\}$, where $E^- \subseteq E$ %
\item $A_6=\{ tparc(t,p,n,ts) : (t,p) \in E^+, n=W(t,p), 0 \leq ts \leq k\}$, where $E^+ \subseteq E$ %
\item $A_7=\{ holds(p,q,0) : p \in P, q=M_{0}(p) \}$ %
\item $A_8=\{ notenabled(t,ts) : t \in T, 0 \leq ts \leq k, \exists p \in \bullet t, M_{ts}(p) < W(p,t) \}$ \newline per definition~\ref{def:pn:enable} (enabled transition) %
\item $A_9=\{ enabled(t,ts) : t \in T, 0 \leq ts \leq k, \forall p \in \bullet t, W(p,t) \leq M_{ts}(p) \}$ \newline per definition~\ref{def:pn:enable} (enabled transition) %
\item $A_{10}=\{ fires(t,ts) : t \in T_{ts}, 0 \leq ts \leq k \}$ \newline per definition~\ref{def:pnr:firing_set} (firing set), only an enabled transition may fire
\item $A_{11}=\{ add(p,q,t,ts) : t \in T_{ts}, p \in t \bullet, q=W(t,p), 0 \leq ts \leq k \}$ \newline per definition~\ref{def:pnr:texec} (transition execution) %
\item $A_{12}=\{ del(p,q,t,ts) : t \in T_{ts}, p \in \bullet t, q=W(p,t), 0 \leq ts \leq k \} \cup \{  del(p,q,t,ts) : t \in T_{ts}, p \in R(t), q=M_{ts}(p), 0 \leq ts \leq k \}$ \newline per definition~\ref{def:pnr:texec} (transition execution) %
\item $A_{13}=\{ tot\_incr(p,q,ts) : p \in P, q=\sum_{t \in T_{ts}, p \in t \bullet}{W(t,p)}, 0 \leq ts \leq k \}$ \newline per definition~\ref{def:pnr:exec} (execution) %
\item $A_{14}=\{ tot\_decr(p,q,ts): p \in P, q=\sum_{t \in T_{ts}, p \in \bullet t}{W(p,t)}+\sum_{t \in T_{ts}, p \in R(t)}{M_{ts}(p)}, \\ 0 \leq ts \leq k \}$ \newline per definition~\ref{def:pnr:exec} (execution) %
\item $A_{15}=\{ consumesmore(p,ts) : p \in P, q=M_{ts}(p), q1=\sum_{t \in T_{ts}, p \in \bullet t}{W(p,t)}+\sum_{t \in T_{ts}, p \in R(t)}{M_{ts}(p)}, q1 > q, 0 \leq ts \leq k \}$ \newline per definition~\ref{def:pnr:conflict} (conflicting transitions) %
\item $A_{16}=\{ consumesmore : \exists p \in P : q=M_{ts}(p), q1=\sum_{t \in T_{ts}, p \in \bullet t}{W(p,t)}+\sum_{t \in T_{ts}, p \in R(t)}{M_{ts}(p)}, q1 > q, 0 \leq ts \leq k \}$ \newline per definition~\ref{def:pnr:conflict} (conflicting transitions) %
\item $A_{17}=\{ could\_not\_have(t,ts) :  t \in T, (\forall p \in \bullet t, W(p,t) \leq M_{ts}(p)), t \not\in T_{ts}, (\exists p \in \bullet t \cup R(t) : q > M_{ts}(p) - (\sum_{t' \in T_{ts}, p \in \bullet t'}{W(p,t')} + \sum_{t' \in T_{ts}, p \in R(t')}{M_{ts}(p)}), q=W(p,t) \text{~if~} p \in \bullet t \text{~or~} M_{ts}(p) \text{~otherwise~}), 0 \leq ts \leq k \}$ \newline
per the maximal firing set semantics
\item $A_{18}=\{ holds(p,q,ts+1) : p \in P, q=M_{ts+1}(p), 0 \leq ts < k\}$, where $M_{ts+1}(p) = M_{ts}(p) - (\sum_{\substack{t \in T_{ts}, p \in \bullet t}}{W(p,t)} +  \sum_{\substack{t \in T_{ts}, p \in R(t)}}{M_{ts}(p)}) + \sum_{\substack{t \in T_{ts}, p \in t \bullet}}{W(t,p)}$ \newline according to definition~\ref{def:pnr:exec} (firing set execution) %
\item $A_{19}=\{ ptarc(p,t,n,ts) : p \in R(t), n=M_{ts}(p), n > 0, 0 \leq ts \leq k\}$ %
\end{enumerate}

{\bf We show that $A$ satisfies \eqref{eqn:reset:fires} and \eqref{eqn:reset:holds}, and $A$ is an answer set of $\Pi^1$.}

$A$ satisfies \eqref{eqn:reset:fires} and \eqref{eqn:reset:holds} by its construction above. We show $A$ is an answer set of $\Pi^1$ by splitting. We split $lit(\Pi^1)$ into a sequence of $7(k+1)+2$ sets:

\begin{itemize}\renewcommand{\labelitemi}{$\bullet$}
\item $U_0= head(f\ref{f:place}) \cup head(f\ref{f:trans}) \cup head(f\ref{f:time}) \cup head(f\ref{f:num}) \cup head(i\ref{i:holds}) = \{place(p) : p \in P\} \cup \{ trans(t) : t \in T\} \cup \{ time(0), \dots, time(k)\} \cup \{num(0), \dots, num(ntok)\} \cup \{ holds(p,q,0) : p \in P, q=M_0(p) \} $
\item $U_{7k+1}=U_{7k+0} \cup head(f\ref{f:r:ptarc})^{ts=k} \cup head(f\ref{f:r:tparc})^{ts=k} \cup head(f\ref{f:rptarc})^{ts=k} = U_{7k+0} \cup \\ \{ ptarc(p,t,n,k) : (p,t) \in E^-, n=W(p,t) \} \cup \{  tparc(t,p,n,k) : (t,p) \in E^+, n=W(t,p) \} \cup \{ ptarc(p,t,n,k) : p \in R(t), n=M_{k}(p), n > 0 \}$
\item $U_{7k+2}=U_{7k+1} \cup head(e\ref{e:r:ne:ptarc} )^{ts=k} = U_{7k+1} \cup \{ notenabled(t,k) : t \in T \}$
\item $U_{7k+3}=U_{7k+2} \cup head(e\ref{e:enabled})^{ts=k} = U_{7k+2} \cup \{ enabled(t,k) : t \in T \}$
\item $U_{7k+4}=U_{7k+3} \cup head(a\ref{a:fires})^{ts=k} = U_{7k+3} \cup \{ fires(t,k) : t \in T \}$
\item $U_{7k+5}=U_{7k+4}  \cup head(r\ref{r:r:add} )^{ts=k} \cup head(r\ref{r:r:del} )^{ts=k} = U_{7k+4} \cup \{ add(p,q,t,k) : p \in P, t \in T, q=W(t,p) \} \cup \{ del(p,q,t,k) : p \in P, t \in T, q=W(p,t) \} \cup \{ del(p,q,t,k) : p \in P, t \in T, q=M_{k}(p) \}$
\item $U_{7k+6}=U_{7k+5} \cup head(r\ref{r:totincr})^{ts=k} \cup head(r\ref{r:totdecr})^{ts=k} = U_{7k+5} \cup \{ tot\_incr(p,q,k) : p \in P, 0 \leq q \leq ntok \} \cup \{ tot\_decr(p,q,k) : p \in P, 0 \leq q \leq ntok \}$
\item $U_{7k+7}=U_{7k+6} \cup head(r\ref{r:nextstate})^{ts=k} \cup head(a\ref{a:overc:place})^{ts=k} \cup head(a\ref{a:r:maxfire:cnh})^{ts=k} $ $= U_{7k+6} \cup \\ \{ consumesmore(p,k) : p \in P\} \cup \{ holds(p,q,k+1) : p \in P, 0 \leq q \leq ntok \} \cup \{ could\_not\_have(t,k) : t \in T \}$
\item $U_{7k+8}=U_{7k+7} \cup head(a\ref{a:overc:gen}) = U_{7k+7} \cup \{ consumesmore \}$
\end{itemize}
where $head(r_i)^{ts=k}$ are head atoms of ground rule $r_i$ in which $ts=k$. We write $A_i^{ts=k} = \{ a(\dots,ts) : a(\dots,ts) \in A_i, ts=k \}$ as short hand for all atoms in $A_i$ with $ts=k$. $U_{\alpha}, 0 \leq \alpha \leq 7k+8$ form a splitting sequence, since each $U_i$ is a splitting set of $\Pi^1$, and $\langle U_{\alpha}\rangle_{\alpha < \mu}$ is a monotone continuous sequence, where $U_0 \subseteq U_1 \dots \subseteq U_{7k+8}$ and $\bigcup_{\alpha < \mu}{U_{\alpha}} = lit(\Pi^1)$. 

We compute the answer set of $\Pi^2$ using the splitting sets as follows:
\begin{enumerate}
\item $bot_{U_0}(\Pi^2) = f\ref{f:place} \cup f\ref{f:trans} \cup f\ref{f:time} \cup i\ref{i:holds} \cup f\ref{f:num}$ and $X_0 = A_1 \cup \dots \cup A_4 \cup A_7$ ($= U_0$) is its answer set -- using forced atom proposition

\item $eval_{U_0}(bot_{U_1}(\Pi^2) \setminus bot_{U_0}(\Pi^2), X_0) = \{ ptarc(p,t,q,0) \text{:-}. | q=W(p,t) \} \cup \\ \{ tparc(t,p,q,0) \text{:-}. | $ $q=W(t,p) \} \cup \{ ptarc(p,t,q,0) \text{:-}. | $ $q=M_0(p) \} $. Its answer set $X_1=A_5^{ts=0} \cup A_6^{ts=0} \cup A_{19}^{ts=0}$ -- using forced atom proposition and construction of $A_5, A_6, A_{19}$.

\item $eval_{U_1}(bot_{U_2}(\Pi^2) \setminus bot_{U_1}(\Pi^2), X_0 \cup X_1) = \{ notenabled(t,0) \text{:-} . | \{ trans(t), \\ ptarc(p,t,n,0), holds(p,q,0) \} \subseteq X_0 \cup X_1, \text{~where~}  q < n \}$. Its answer set $X_2=A_8^{ts=0}$ -- using  forced atom proposition and construction of $A_8$.
\begin{enumerate}
\item where, $q=M_0(p)$, and $n=W(p,t)$ for an arc $(p,t) \in E^-$ -- by construction of $i\ref{i:holds}$ and $f\ref{f:r:ptarc}$ in $\Pi^2$, and 
\item in an arc $(p,t) \in E^-$, $p \in \bullet t$ (by definition~\ref{def:pn:preset} of preset)
\item thus, $notenabled(t,0) \in X_1$ represents $\exists p \in \bullet t : M_0(p) < W(p,t)$.
\end{enumerate}

\item $eval_{U_2}(bot_{U_3}(\Pi^2) \setminus bot_{U_2}(\Pi^2), X_0 \cup \dots \cup X_2) = \{ enabled(t,0) \text{:-}. | trans(t) \in X_0 \cup \dots \cup X_2, notenabled(t,0) \notin X_0 \cup \dots \cup X_2 \}$. Its answer set is $X_3 = A_9^{ts=0}$ -- using forced atom proposition and construction of $A_9$.
\begin{enumerate}
\item since an $enabled(t,0) \in X_3$ if $\nexists ~notenabled(t,0) \in X_0 \cup \dots \cup X_2$; which is equivalent to $\nexists p \in \bullet t : M_0(p) < W(p,t) \equiv \forall p \in \bullet t: M_0(p) \geq W(p,t)$.
\end{enumerate}

\item $eval_{U_3}(bot_{U_4}(\Pi^2) \setminus bot_{U_3}(\Pi^2), X_0 \cup \dots \cup X_3) = \{\{fires(t,0)\} \text{:-}. | enabled(t,0) \\ \text{~holds in~} X_0 \cup \dots \cup X_3 \}$. It has multiple answer sets $X_{4.1}, \dots, X_{4.n}$, corresponding to elements of power set of $fires(t,0)$ atoms in $eval_{U_3}(...)$ -- using supported rule proposition. Since we are showing that the union of answer sets of $\Pi^2$ determined using splitting is equal to $A$, we only consider the set that matches the $fires(t,0)$ elements in $A$ and call it $X_4$, ignoring the rest. Thus, $X_4 = A_{10}^{ts=0}$, representing $T_0$.
\begin{enumerate}
\item in addition, for every $t$ such that $enabled(t,0) \in X_0 \cup \dots \cup X_3,  R(t) \neq \emptyset$; $fires(t,0) \in X_4$ -- per definition~\ref{def:pnr:firing_set} (firing set); requiring that a reset transition is fired when enabled
\item thus, the firing set $T_0$ will not be eliminated by the constraint $f\ref{f:c:rptarc:elim}$ 
\end{enumerate}

\item $eval_{U_4}(bot_{U_5}(\Pi^2) \setminus bot_{U_4}(\Pi^2), X_0 \cup \dots \cup X_4) = \{add(p,n,t,0) \text{:-}. | \{fires(t,0), \\ tparc(t,p,n,0) \} \subseteq X_0 \cup \dots \cup X_4 \} \cup \{ del(p,n,t,0) \text{:-}. | \{ fires(t,0), ptarc(p,t,n,0) \} \subseteq X_0 \cup \dots \cup X_4 \}$. It's answer set is $X_5 = A_{11}^{ts=0} \cup A_{12}^{ts=0}$ -- using forced atom proposition and definitions of $A_{11}$ and $A_{12}$. 
\begin{enumerate}
\item where, each $add$ atom is equivalent to $n=W(t,p) : p \in t \bullet$, 
\item and each $del$ atom is equivalent to $n=W(p,t) : p \in \bullet t$; or $n=M_k(p) : p \in R(t)$,
\item representing the effect of transitions in $T_0$ -- by construction
\end{enumerate}

\item $eval_{U_5}(bot_{U_6}(\Pi^2) \setminus bot_{U_5}(\Pi^2), X_0 \cup \dots \cup X_5) = \{tot\_incr(p,qq,0) \text{:-}. | $ \\$qq=\sum_{add(p,q,t,0) \in X_0 \cup \dots \cup X_5}{q} \} \cup \{ tot\_decr(p,qq,0) \text{:-}. | qq=\sum_{del(p,q,t,0) \in X_0 \cup \dots \cup X_5}{q} \}$. It's answer set is $X_6 = A_{13}^{ts=0} \cup A_{14}^{ts=0}$ --  using forced atom proposition, definitions of $A_{13}$, $A_{14}$, and definition~\ref{def:agg:sat} (semantics of aggregate assignment atom).
\begin{enumerate}
\item where, each for $tot\_incr(p,qq,0)$, $qq=\sum_{add(p,q,t,0) \in X_0 \cup \dots X_5}{q}$ \\$\equiv qq=\sum_{t \in X_4, p \in t \bullet}{W(p,t)}$, 
\item and each $tot\_decr(p,qq,0)$, $qq=\sum_{del(p,q,t,0) \in X_0 \cup \dots X_5}{q}$ \\$\equiv qq=\sum_{t \in X_4, p \in \bullet t}{W(t,p)} + \sum_{t \in X_4, p \in R(t)}{M_{k}(p)}$, 
\item represent the net effect of transitions in $T_0$ -- by construction 
\end{enumerate}
\item $eval_{U_6}(bot_{U_7}(\Pi^2) \setminus bot_{U_6}(\Pi^2), X_0 \cup \dots \cup X_6) = \{ consumesmore(p,0) \text{:-}. | \\ \{holds(p,q,0), tot\_decr(p,q1,0) \} \subseteq X_0 \cup \dots \cup X_6, q1 > q \} \cup \{ holds(p,q,1) \text{:-}., | \\ \{ holds(p,q1,0), tot\_incr(p,q2,0), tot\_decr(p,q3,0) \} \subseteq X_0 \cup \dots \cup X_6, q=q1+q2-q3 \} \cup \{ could\_not\_have(t,0) \text{:-}. | \{ enabled(t,0), ptarc(s,t,q), holds(s,qq,0), \\ tot\_decr(s,qqq,0) \} \subseteq X_0 \cup \dots \cup X_6, fires(t,0) \notin (X_0 \cup \dots \cup X_6), q > qq-qqq \}$. It's answer set is $X_7 = A_{15}^{ts=0} \cup A_{17}^{ts=0} \cup A_{18}^{ts=0}$ -- using forced atom proposition and definitions of $A_{15}, A_{17}, A_{18}$.
\begin{enumerate}
\item where, $consumesmore(p,0)$ represents $\exists p : q=M_0(p), q1= \\ \sum_{t \in T_0, p \in \bullet t}{W(p,t)}+$ $\sum_{t \in T_0, p \in R(t)}{M_k(p)}, $ $q1 > q$, indicating place $p$ will be overconsumed if $T_0$ is fired -- as defined in definition~\ref{def:pnr:conflict} (conflicting transitions)
\item and, $holds(p,q,1)$ represents $q=M_1(p)$ -- by construction of $\Pi^2$
\item and $could\_not\_have(t,0)$ represents an enabled transition $t$ in $T_0$ that could not fire due to insufficient tokens
\item $X_7$ does not contain $could\_not\_have(t,0)$, when $enabled(t,0) \in X_0 \cup \dots \cup X_6$ and $fires(t,0) \notin X_0 \cup \dots \cup X_6$ due to construction of $A$, encoding of $a\ref{a:r:maxfire:cnh}$ and its body atoms. As a result it is not eliminated by the constraint $a\ref{a:maxfire:elim}$
\end{enumerate}

\[ \vdots \]

\item $eval_{U_{7k+0}}(bot_{U_{7k+1}}(\Pi^2) \setminus bot_{U_{7k+0}}(\Pi^2), X_0 \cup \dots \cup X_{7k+0}) = \{ ptarc(p,t,q,k) \text{:-}. | q=W(p,t) \} \cup \{ tparc(t,p,q,k) \text{:-}. | q=W(t,p) \} \cup \{ ptarc(p,t,q,k) \text{:-}. | q=M_0(p) \} $. Its answer set $X_{7k+1}=A_5^{ts=k} \cup A_6^{ts=k} \cup A_{19}^{ts=k}$ -- using forced atom proposition and construction of $A_5, A_6, A_{19}$.

\item $eval_{U_{7k+1}}(bot_{U_{7k+2}}(\Pi^2) \setminus bot_{U_{7k+1}}(\Pi^2), X_0 \cup X_{7k+1}) = \{ notenabled(t,k) \text{:-} . | \{ trans(t), \\ ptarc(p,t,n,k), holds(p,q,k) \} \subseteq X_0 \cup X_{7k+1}, \text{~where~}  q < n \}$. Its answer set $X_{7k+2}=A_8^{ts=k}$ -- using  forced atom proposition and construction of $A_8$.
\begin{enumerate}
\item where, $q=M_0(p)$, and $n=W(p,t)$ for an arc $(p,t) \in E^-$ -- by construction of $holds$ and $ptarc$ predicates in $\Pi^2$, and 
\item in an arc $(p,t) \in E^-$, $p \in \bullet t$ (by definition~\ref{def:pn:preset} of preset)
\item thus, $notenabled(t,k) \in X_{7k+1}$ represents $\exists p \in \bullet t : M_0(p) < W(p,t)$.
\end{enumerate}

\item $eval_{U_{7k+2}}(bot_{U_{7k+3}}(\Pi^2) \setminus bot_{U_{7k+2}}(\Pi^2), X_0 \cup \dots \cup X_{7k+2}) = \{ enabled(t,k) \text{:-}. | trans(t) \in X_0 \cup \dots \cup X_{7k+2} , notenabled(t,k) \notin X_0 \cup \dots \cup X_{7k+2} \}$. Its answer set is $X_{7k+3} = A_9^{ts=k}$ -- using forced atom proposition and construction of $A_9$.
\begin{enumerate}
\item since an $enabled(t,k) \in X_{7k+3}$ if $\nexists ~notenabled(t,k) \in X_0 \cup \dots \cup X_{7k+2}$; which is equivalent to $\nexists p \in \bullet t : M_0(p) < W(p,t) \equiv \forall p \in \bullet t: M_0(p) \geq W(p,t)$.
\end{enumerate}

\item $eval_{U_{7k+3}}(bot_{U_{7k+4}}(\Pi^2) \setminus bot_{U_{7k+3}}(\Pi^2), X_0 \cup \dots \cup X_{7k+3}) = \{\{fires(t,k)\} \text{:-}. | \\ enabled(t,k) \\ \text{~holds in~} X_0 \cup \dots \cup X_{7k+3} \}$. It has multiple answer sets $X_{7k+4.1}, \dots, X_{1k+4.n}$, corresponding to elements of power set of $fires(t,k)$ atoms in $eval_{U_{7k+3}}(...)$ -- using supported rule proposition. Since we are showing that the union of answer sets of $\Pi^2$ determined using splitting is equal to $A$, we only consider the set that matches the $fires(t,k)$ elements in $A$ and call it $X_{7k+4}$, ignoring the rest. Thus, $X_{7k+4} = A_{10}^{ts=k}$, representing $T_k$.
\begin{enumerate}
\item in addition, for every $t$ such that $enabled(t,k) \in X_0 \cup \dots \cup X_{7k+3},  R(t) \neq \emptyset$; $fires(t,k) \in X_{7k+4}$ -- per definition~\ref{def:pnr:firing_set} (firing set); requiring that a reset transition is fired when enabled
\item thus, the firing set $T_k$ will not be eliminated by the constraint $f\ref{f:c:rptarc:elim}$ 
\end{enumerate}

\item $eval_{U_{7k+4}}(bot_{U_{7k+5}}(\Pi^2) \setminus bot_{U_{7k+4}}(\Pi^2), X_0 \cup \dots \cup X_{7k+4}) = \\ \{add(p,n,t,k) \text{:-}. | \{fires(t,k), tparc(t,p,n,k) \} \subseteq X_0 \cup \dots \cup X_{7k+4} \} \cup \\ \{ del(p,n,t,k) \text{:-}. | \{ fires(t,k), ptarc(p,t,n,k) \} \subseteq X_0 \cup \dots \cup X_{7k+4} \}$. It's answer set is $X_{7k+5} = A_{11}^{ts=k} \cup A_{12}^{ts=k}$ -- using forced atom proposition and definitions of $A_{11}$ and $A_{12}$. 
\begin{enumerate}
\item where, each $add$ atom is equivalent to $n=W(t,p) : p \in \bullet t$, 
\item and each $del$ atom is equivalent to $n=W(p,t) : p \in t \bullet$; or $n=M_k(p) : p \in R(t)$,
\item representing the effect of transitions in $T_k$
\end{enumerate}

\item $eval_{U_{7k+5}}(bot_{U_{7k+6}}(\Pi^2) \setminus bot_{U_{7k+5}}(\Pi^2), X_0 \cup \dots \cup X_{7k+5}) = \\ \{tot\_incr(p,qq,k) \text{:-}. | qq=\sum_{add(p,q,t,k) \in X_0 \cup \dots \cup X_{7k+5}}{q} \} \cup \\ \{ tot\_decr(p,qq,k) \text{:-}. | qq=\sum_{del(p,q,t,k) \in X_0 \cup \dots \cup X_{7k+5}}{q} \}$. It's answer set is $X_{7k+6} = A_{13}^{ts=k} \cup A_{14}^{ts=k}$ --  using forced atom proposition and definitions of $A_{13}$ and $A_{14}$.
\begin{enumerate}
\item where, each $tot\_incr(p,qq,k)$, $qq=\sum_{add(p,q,t,k) \in X_0 \cup \dots X_{7k+5}}{q}$ \\$\equiv qq=\sum_{t \in X_{7k+4}, p \in t \bullet}{W(p,t)}$, 
\item and each $tot\_decr(p,qq,k)$, $qq=\sum_{del(p,q,t,k) \in X_0 \cup \dots X_{7k+5}}{q}$ \\$\equiv qq=\sum_{t \in X_{7k+4}, p \in \bullet t}{W(t,p)} + \sum_{t \in X_{7k+4}, p \in R(t)}{M_{k}(p)}$, 
\item represent the net effect of transitions in $T_k$
\end{enumerate}
\item $eval_{U_{7k+6}}(bot_{U_{7k+7}}(\Pi^2) \setminus bot_{U_{7k+6}}(\Pi^2), X_0 \cup \dots \cup X_{7k+6}) = \{ consumesmore(p,k) \text{:-}. | \\ \{holds(p,q,k), tot\_decr(p,q1,k) \} \subseteq X_0 \cup \dots \cup X_{7k+6}, q1 > q \} \cup \{ holds(p,q,k+1) \text{:-}., | \\ \{ holds(p,q1,k), tot\_incr(p,q2,k), tot\_decr(p,q3,k) \} \subseteq X_0 \cup \dots \cup X_{7k+6}, q=q1+q2-q3 \} \cup \{ could\_not\_have(t,k) \text{:-}. | \{ enabled(t,k), ptarc(s,t,q), holds(s,qq,k), \\ tot\_decr(s,qqq,k) \} \subseteq X_0 \cup \dots \cup X_{7k+6}, fires(t,k) \notin (X_0 \cup \dots \cup X_{7k+6}), q > qq-qqq \}$. It's answer set is $X_{7k+7} = A_{15}^{ts=k} \cup A_{17}^{ts=k} \cup A_{18}^{ts=k}$ -- using forced atom proposition and definitions of $A_{15}, A_{17}, A_{18}$.
\begin{enumerate}
\item where, $consumesmore(p,k)$ represents $\exists p : q=M_0(p), q1= \\ \sum_{t \in T_0, p \in \bullet t}{W(p,t)}+\sum_{t \in T_0, p \in R(t)}{M_k(p)}, q1 > q$, indicating place $p$ that will be over consumed if $T_k$ is fired, as defined in definition~\ref{def:pnr:conflict} (conflicting transitions)
\item $holds(p,q,k+1)$ represents $q=M_{k+1}(p)$ -- by construction of $\Pi^2$,
\item and $could\_not\_have(t,k)$ represents enabled transition $t$ in $T_k$ that could not be fired due to insufficient tokens
\item $X_{7k+7}$ does not contain $could\_not\_have(t,k)$, when $enabled(t,k) \in X_0 \cup \dots \cup X_{7k+6}$ and $fires(t,k) \notin X_0 \cup \dots \cup X_{7k+6}$ due to the construction of $A$, encoding of $a\ref{a:r:maxfire:cnh}$ and its body atoms. As a result it is not eliminated by the constraint $a\ref{a:maxfire:elim}$

\end{enumerate}

\item $eval_{U_{7k+7}}(bot_{U_{7k+8}}(\Pi^2) \setminus bot_{U_{7k+7}}(\Pi^2), X_0 \cup \dots \cup X_{7k+7}) = \{ consumesmore \text{:-}. | $ \\$ \{ consumesmore(p,0),\dots,$ $consumesmore(p,k) \} \subseteq X_0 \cup \dots \cup X_{7k+7} \}$. It's answer set is $X_{7k+8} = A_{16}$ -- using forced atom proposition
\begin{enumerate}
\item $X_{7k+8}$ will be empty since none of $consumesmore(p,0),\dots, \\ consumesmore(p,k)$ hold in $X_0 \cup \dots \cup X_{7k+7}$ due to the construction of $A$, encoding of $a\ref{a:overc:place}$ and its body atoms. As a result, it is not eliminated by the constraint $a\ref{a:overc:elim}$
\end{enumerate}

\end{enumerate}

The set $X = X_0 \cup \dots \cup X_{7k+8}$ is the answer set of $\Pi^2$ by the splitting sequence theorem~\ref{def:split_seq_thm}. Each $X_i, 0 \leq i \leq 7k+8$ matches a distinct portion of $A$, and $X = A$, thus $A$ is an answer set of $\Pi^2$.

\vspace{30pt}
\noindent
{\bf Next we show (\ref{prove:a2x:reset}):} Given $\Pi^2$ be the encoding of a Petri Net $PN(P,T,E,W,R)$ with initial marking $M_0$, and $A$ be an answer set of $\Pi^2$ that satisfies (\ref{eqn:reset:fires}) and (\ref{eqn:reset:holds}), then we can construct $X=M_0,T_0,\dots,M_k,T_k,M_{k+1}$ from $A$, such that it is an execution sequence of $PN$.

We construct the $X$ as follows:
\begin{enumerate}
\item $M_i = (M_i(p_0), \dots, M_i(p_n))$, where $\{ holds(p_0,M_i(p_0),i), \dots holds(p_n,M_i(p_n),i) \} \\ \subseteq A$, for $0 \leq i \leq k+1$
\item $T_i = \{ t : fires(t,i) \in A\}$, for $0 \leq i \leq k$ 
\end{enumerate}
and show that $X$ is indeed an execution sequence of $PN$. We show this by induction over $k$ (i.e. given $M_k$, $T_k$ is a valid firing set and its firing produces marking $M_{k+1}$).

\vspace{20pt}

\noindent
{\bf Base case:} Let $k=0$, and $M_0$ is a valid marking in $X$ for $PN$, show
\begin{inparaenum}[(1)]
\item $T_0$ is a valid firing set for $M_0$, and 
\item $T_0$'s firing in $M_0$ produces marking $M_1$.
\end{inparaenum} 

\begin{enumerate}
\item We show $T_0$ is a valid firing set for $M_0$. Let $\{ fires(t_0,0), \dots, fires(t_x,0) \}$ be the set of all $fires(\dots,0)$ atoms in $A$, \label{prove:reset:fires_t0}

\begin{enumerate}
\item Then for each $fires(t_i,0) \in A$

\begin{enumerate}
\item $enabled(t_i,0) \in A$ -- from rule $a\ref{a:fires}$ and supported rule proposition
\item Then $notenabled(t_i,0) \notin A$ -- from rule $e\ref{e:enabled}$ and supported rule proposition
\item Then $body(e\ref{e:r:ne:ptarc})$ must not hold in $A$ -- from rule $e\ref{e:r:ne:ptarc}$ and forced atom proposition
\item Then $q \not< n_i \equiv q \geq n_i$ in $e\ref{e:r:ne:ptarc}$ for all $\{holds(p,q,0), ptarc(p,t_i,n_i,0)\} \subseteq A$ -- from $e\ref{e:r:ne:ptarc}$, forced atom proposition, and the following
\begin{enumerate}
\item $holds(p,q,0) \in A$ represents $q=M_0(p)$ -- rule $i\ref{i:holds}$ construction
\item $ptarc(p,t_i,n_i,0) \in A$ represents $n_i=W(p,t_i)$ -- rule $f\ref{f:r:ptarc}$ construction; or it represents $n_i = M_0(p)$ -- rule $f\ref{f:rptarc}$ construction; the construction of $f\ref{f:rptarc}$ ensures that $notenabled(t,0)$ is never true due to the reset arc
\end{enumerate}
\item Then $\forall p \in \bullet t_i, M_0(p) \geq W(p,t_i)$ -- from  definition~\ref{def:pn:preset} of preset $\bullet t_i$ in PN
\item Then $t_i$ is enabled and can fire in $PN$, as a result it can belong to $T_0$ -- from definition~\ref{def:pn:enable} of enabled transition

\end{enumerate}
\item And $consumesmore \notin A$, since $A$ is an answer set of $\Pi^2$ -- from rule $a\ref{a:overc:elim}$ and supported rule proposition
\begin{enumerate}
\item Then $\nexists consumesmore(p,0) \in A$ -- from rule $a\ref{a:overc:gen}$ and supported rule proposition
\item  Then $\nexists \{ holds(p,q,0), tot\_decr(p,q1,0) \} \subseteq A : q1>q$ in $body(a\ref{a:overc:place})$ -- from $a\ref{a:overc:place}$ and forced atom proposition
\item Then $\nexists p : (\sum_{t_i \in \{t_0,\dots,t_x\}, p \in \bullet t_i}{W(p,t_i)}+\sum_{t_i \in \{t_0,\dots,t_x\}, p \in R(t_i)}{M_0(p)}) > M_0(p)$ -- from the following
\begin{enumerate}
\item $holds(p,q,0)$ represents $q=M_0(p)$ -- from rule $i\ref{i:holds}$ construction, given
\item $tot\_decr(p,q1,0) \in A$ if $\{ del(p,q1_0,t_0,0), \dots, del(p,q1_x,t_x,0) \} \subseteq A$, where $q1 = q1_0+\dots+q1_x$ -- from $r\ref{r:totdecr}$ and forced atom proposition
\item $del(p,q1_i,t_i,0) \in A$ if $\{ fires(t_i,0), ptarc(p,t_i,q1_i,0) \} \subseteq A$ -- from $r\ref{r:r:del} $ and supported rule proposition
\item $del(p,q1_i,t_i,0)$ either represents removal of $q1_i = W(p,t_i)$ tokens from $p \in \bullet t_i$; or it represents removal of $q1_i = M_0(p)$ tokens from $p \in R(t_i)$-- from rule $r\ref{r:r:del} $, supported rule proposition, and definition~\ref{def:pnr:texec} of transition execution in $PN$
\end{enumerate}
\item Then the set of transitions in $T_0$ do not conflict -- by the definition~\ref{def:pnr:conflict} of conflicting transitions
\end{enumerate}

\item And for each $enabled(t_j,0) \in A$ and $fires(t_j,0) \notin A$, $could\_not\_have(t_j,0) \in A$, since $A$ is an answer set of $\Pi^2$ - from rule $a\ref{a:maxfire:elim}$ and supported rule proposition
\begin{enumerate}
\item Then $\{ enabled(t_j,0), holds(s,qq,0), ptarc(s,t_j,q,0), \\ tot\_decr(s,qqq,0) \} \subseteq A$, such that $q > qq - qqq$ and $fires(t_j,0) \notin A$ - from rule $a\ref{a:r:maxfire:cnh}$ and supported rule proposition
\item Then for an $s \in \bullet t_j \cup R(t_j)$, $q > M_0(s) - (\sum_{t_i \in T_0, s \in \bullet t_i}{W(s,t_i)} + \sum_{t_i \in T_0, s \in R(t_i)}{M_0(s))}$, where $q=W(s,t_j) \text{~if~} s \in \bullet t_j, \text{~or~} M_0(s) \text{~otherwise}$ - from the following: %
\begin{enumerate}
\item $ptarc(s,t_i,q,0)$ represents $q=W(s,t_i)$ if $(s,t_i) \in E^-$ or $q=M_0(s)$ if $s \in R(t_i)$ -- from rule $f\ref{f:r:ptarc},f\ref{f:rptarc}$ construction
\item $holds(s,qq,0)$ represents $qq=M_0(s)$ -- from $i\ref{i:holds}$ construction
\item $tot\_decr(s,qqq,0) \in A$ if $\{ del(s,qqq_0,t_0,0), \dots, del(s,qqq_x,t_x,0) \} \subseteq A$ -- from rule $r\ref{r:totdecr}$ construction and supported rule proposition
\item $del(s,qqq_i,t_i,0) \in A$ if $\{ fires(t_i,0), ptarc(s,t_i,qqq_i,0) \} \subseteq A$ -- from rule $r\ref{r:r:del} $ and supported rule proposition
\item $del(s,qqq_i,t_i,0)$ represents $qqq_i = W(s,t_i) : t_i \in T_0, (s,t_i) \in E^-$, or $qqq_i = M_0(t_i) : t_i \in T_0, s \in R(t_i)$ -- from rule $f\ref{f:r:ptarc},f\ref{f:rptarc}$ construction 
\item $tot\_decr(q,qqq,0)$ represents $\sum_{t_i \in T_0, s \in \bullet t_i}{W(s,t_i)} + \\ \sum_{t_i \in T_0, s \in R(t_i)}{M_0(s)}$ -- from (C,D,E) above 
\end{enumerate}

\item Then firing $T_0 \cup \{ t_j \}$ would have required more tokens than are present at its source place $s \in \bullet t_j \cup R(t_j)$. Thus, $T_0$ is a maximal set of transitions that can simultaneously fire.
\end{enumerate}

\item And for each reset transition $t_r$ with $enabled(t_r,0) \in A$, $fires(t_r,0) \in A$, since $A$ is an answer set of $\Pi^2$ - from rule $f\ref{f:c:rptarc:elim}$ and supported rule proposition
\begin{enumerate}
\item Then, the firing set $T_0$ satisfies the reset-transition requirement of definition~\ref{def:pnr:firing_set} (firing set)
\end{enumerate}

\item Then $\{t_0, \dots, t_x\} = T_0$ -- using 1(a),1(b),1(d) above; and using 1(c) it is a maximal firing set  
\end{enumerate}

\item We show $M_1$ is produced by firing $T_0$ in $M_0$. Let $holds(p,q,1) \in A$
\begin{enumerate}
\item Then $\{ holds(p,q1,0), tot\_incr(p,q2,0), tot\_decr(p,q3,0) \} \subseteq A : q=q1+q2-q3$ -- from rule $r\ref{r:nextstate}$ and supported rule proposition \label{reset:x:1}
\item \label{reset:x:2} $holds(p,q1,0) \in A$ represents $q1=M_0(p)$ -- given, rule $i\ref{i:holds}$ construction; 
and $\{add(p,q2_0,t_0,0), \dots, $ $add(p,q2_j,t_j,0)\} \subseteq A : q2_0 + \dots + q2_j = q2$  %
and $\{del(p,q3_0,t_0,0), \dots, $ $del(p,q3_l,t_l,0)\} \subseteq A : q3_0 + \dots + q3_l = q3$ %
 -- rules $r\ref{r:totincr},r\ref{r:totdecr}$ using supported rule proposition

\item Then $\{ fires(t_0,0), \dots, fires(t_j,0) \} \subseteq A$ and $\{ fires(t_0,0), \dots, fires(t_l,0) \} \\ \subseteq A$ -- rules $r\ref{r:r:add},r\ref{r:r:del}$ and supported rule proposition, respectively
\item Then $\{ fires(t_0,0), \dots, fires(t_j,0) \} \cup \{ fires(t_0,0), \dots, fires(t_l,0) \} \subseteq A = \{ fires(t_0,0), \dots, fires(t_x,0) \} \subseteq A$ -- set union of subsets
\item Then for each $fires(t_x,0) \in A$ we have $t_x \in T_0$ -- already shown in item~\ref{prove:reset:fires_t0} above
\item Then $q = M_0(p) + \sum_{t_x \in T_0 \wedge p \in t_x \bullet}{W(t_x,p)} - (\sum_{t_x \in T_0 \wedge p \in \bullet t_x}{W(p,t_x)} + \\ \sum_{t_x \in T_0 \wedge p \in R(t_x)}{M_0(p)})$ -- from %
(\ref{reset:x:1}),(\ref{reset:x:2}) 
\eqref{reset:x:2} above and the following
\begin{enumerate}
\item Each $add(p,q_j,t_j,0) \in A$ represents $q_j=W(t_j,p)$ for $p \in t_j \bullet$ -- rule $r\ref{r:r:add} $ encoding, and definition~\ref{def:pn:preset} of postset in $PN$ %
\item Each $del(p,t_y,q_y,0) \in A$ represents either $q_y=W(p,t_y)$ for $p \in \bullet t_y$, or $q_y=M_0(p)$ for $p \in R(t_y)$ -- from rule $r\ref{r:r:del} ,f\ref{f:r:ptarc} $ encoding and definition~\ref{def:pn:preset} of preset in $PN$; or from rule $r\ref{r:r:del} ,f\ref{f:rptarc}$ encoding and definition of reset arc in $PN$
\item Each $tot\_incr(p,q2,0) \in A$ represents $q2=\sum_{t_x \in T_0 \wedge p \in t_x  \bullet}{W(t_x,p)}$ -- aggregate assignment atom semantics in rule $r\ref{r:totincr}$
\item Each $tot\_decr(p,q3,0) \in A$ represents $q3=\sum_{t_x \in T_0 \wedge p \in \bullet t_x}{W(p,t_x)} + $ $\sum_{t_x \in T_0 \wedge p \in R(t_x)}{M_0(p)}$ -- aggregate assignment atom semantics in rule $r\ref{r:totdecr}$
\end{enumerate}
\item Then $M_1(p) = q$ -- since $holds(p,q,1) \in A$ encodes $q=M_1(p)$ -- from construction 
\end{enumerate}
\end{enumerate}

\noindent
{\bf Inductive Step:} Let $k > 0$, and $M_k$ is a valid marking in $X$ for $PN$, show 
\begin{inparaenum}[(1)]
\item $T_k$ is a valid firing set in $M_k$, and 
\item firing $T_k$ in $M_k$ produces marking $M_{k+1}$.
\end{inparaenum}

\begin{enumerate}
\item We show that $T_k$ is a valid firing set in $M_k$. Let $\{ fires(t_0,k), \dots, fires(t_x,k) \}$ be the set of all $fires(\dots,k)$ atoms in $A$, \label{prove:reset:fires_tk}

\begin{enumerate}
\item Then for each $fires(t_i,k) \in A$

\begin{enumerate}
\item $enabled(t_i,k) \in A$ -- from rule $a\ref{a:fires}$ and supported rule proposition
\item Then $notenabled(t_i,k) \notin A$ -- from rule $e\ref{e:enabled}$ and supported rule proposition
\item Then $body(e\ref{e:r:ne:ptarc})$ must not hold in $A$ -- from rule $e\ref{e:r:ne:ptarc}$ and forced atom proposition
\item Then $q \not< n_i \equiv q \geq n_i$ in $e\ref{e:r:ne:ptarc}$ for all $\{holds(p,q,k), ptarc(p,t_i,n_i,k)\} \subseteq A$ -- from $e\ref{e:r:ne:ptarc}$, forced atom proposition, and the following %
\begin{enumerate}
\item $holds(p,q,k) \in A$ represents $q=M_k(p)$ -- construction, inductive assumption
\item $ptarc(p,t_i,n_i,k) \in A$ represents $n_i=W(p,t_i)$ -- rule $f\ref{f:r:ptarc}$ construction; or it represents $n_i=M_k(p)$ -- rule $f\ref{f:rptarc}$ construction; the construction of $f\ref{f:rptarc}$ ensures that $notenabled(t,0)$ is never true due to the reset arc 
\end{enumerate}
\item Then $\forall p \in \bullet t_i, M_k(p) \geq W(p,t_i)$ -- from definition~\ref{def:pn:preset} of preset $\bullet t_i$ in PN
\item Then $t_i$ is enabled and can fire in $PN$, as a result it can belong to $T_k$ -- from definition~\ref{def:pn:enable} of enabled transition

\end{enumerate}
\item And $consumesmore \notin A$, since $A$ is an answer set of $\Pi^2$ -- from rule $a\ref{a:overc:elim}$ and supported rule proposition
\begin{enumerate}
\item Then $\nexists consumesmore(p,k) \in A$ -- from rule $a\ref{a:overc:gen}$ and supported rule proposition
\item  Then $\nexists \{ holds(p,q,k), tot\_decr(p,q1,k) \} \subseteq A : q1>q$ in $body(a\ref{a:overc:place})$ -- from $a\ref{a:overc:place}$ and forced atom proposition
\item Then $\nexists p : (\sum_{t_i \in \{t_0,\dots,t_x\}, p \in \bullet t_i}{W(p,t_i)}+\sum_{t_i \in \{t_0,\dots,t_x\}, p \in R(t_i)}{M_k(p)}) > M_k(p)$ -- from the following
\begin{enumerate}
\item $holds(p,q,k)$ represents $q=M_k(p)$ -- inductive assumption, given
\item $tot\_decr(p,q1,k) \in A$ if $\{ del(p,q1_0,t_0,k), \dots, del(p,q1_x,t_x,k) \} \subseteq A$, where $q1 = q1_0+\dots+q1_x$ -- from $r\ref{r:totdecr}$ and forced atom proposition
\item $del(p,q1_i,t_i,k) \in A$ if $\{ fires(t_i,k), ptarc(p,t_i,q1_i,k) \} \subseteq A$ -- from $r\ref{r:r:del} $ and supported rule proposition
\item $del(p,q1_i,t_i,k)$ either represents removal of $q1_i = W(p,t_i)$ tokens from $p \in \bullet t_i$; or it represents removal of $q1_i = M_k(p)$ tokens from $p \in R(t_i)$-- from rule $r\ref{r:r:del} $, supported rule proposition, and definition~\ref{def:pnr:texec} of transition execution in $PN$
\end{enumerate}
\item Then the set of transitions in $T_k$ do not conflict -- by the definition~\ref{def:pnr:conflict} of conflicting transitions
\end{enumerate}

\item And for each $enabled(t_j,k) \in A$ and $fires(t_j,k) \notin A$, $could\_not\_have(t_j,k) \in A$, since $A$ is an answer set of $\Pi^2$ - from rule $a\ref{a:maxfire:elim}$ and supported rule proposition
\begin{enumerate}
\item Then $\{ enabled(t_j,k), holds(s,qq,k), ptarc(s,t_j,q,k), \\ tot\_decr(s,qqq,k) \} \subseteq A$, such that $q > qq - qqq$ and $fires(t_j,k) \notin A$ - from rule $a\ref{a:r:maxfire:cnh}$ and supported rule proposition
\item Then for an $s \in \bullet t_j \cup R(t_j)$, $q > M_k(s) - (\sum_{t_i \in T_k, s \in \bullet t_i}{W(s,t_i)} + \sum_{t_i \in T_k, s \in R(t_i)}{M_k(s))}$, where $q=W(s,t_j) \text{~if~} s \in \bullet t_j, \text{~or~} M_k(s) \text{~otherwise}$ - from the following: %
\begin{enumerate}
\item $ptarc(s,t_i,q,k)$ represents $q=W(s,t_i)$ if $(s,t_i) \in E^-$ or $q=M_k(s)$ if $s \in R(t_i)$ -- from rule $f\ref{f:r:ptarc},f\ref{f:rptarc}$ construction
\item $holds(s,qq,k)$ represents $qq=M_k(s)$ -- construction
\item $tot\_decr(s,qqq,k) \in A$ if $\{ del(s,qqq_0,t_0,k), \dots, del(s,qqq_x,t_x,k) \} \subseteq A$ -- from rule $r\ref{r:totdecr}$ construction and supported rule proposition
\item $del(s,qqq_i,t_i,k) \in A$ if $\{ fires(t_i,k), ptarc(s,t_i,qqq_i,k) \} \subseteq A$ -- from rule $r\ref{r:r:del} $ and supported rule proposition
\item $del(s,qqq_i,t_i,k)$ represents $qqq_i = W(s,t_i) : t_i \in T_k, (s,t_i) \in E^-$, or $qqq_i = M_k(t_i) : t_i \in T_k, s \in R(t_i)$ -- from rule $f\ref{f:r:ptarc},f\ref{f:rptarc}$ construction 
\item $tot\_decr(q,qqq,k)$ represents $\sum_{t_i \in T_k, s \in \bullet t_i}{W(s,t_i)} + \\ \sum_{t_i \in T_k, s \in R(t_i)}{M_k(s)}$ -- from (C,D,E) above 
\end{enumerate}

\item Then firing $T_k \cup \{ t_j \}$ would have required more tokens than are present at its source place $s \in \bullet t_j \cup R(t_j)$. Thus, $T_k$ is a maximal set of transitions that can simultaneously fire.
\end{enumerate}

\item And for each reset transition $t_r$ with $enabled(t_r,k) \in A$, $fires(t_r,k) \in A$, since $A$ is an answer set of $\Pi^2$ - from rule $f\ref{f:c:rptarc:elim}$ and supported rule proposition
\begin{enumerate}
\item Then the firing set $T_k$ satisfies the reset transition requirement of definition~\ref{def:pnr:firing_set} (firing set)
\end{enumerate}

\item Then $\{t_0, \dots, t_x\} = T_k$ -- using 1(a),1(b), 1(d) above; and using 1(c) it is a maximal firing set  
\end{enumerate}

\item We show that $M_{k+1}$ is produced by firing $T_k$ in $M_k$. Let $holds(p,q,k+1) \in A$
\begin{enumerate}
\item Then $\{ holds(p,q1,k), tot\_incr(p,q2,k), tot\_decr(p,q3,k) \} \subseteq A : q=q1+q2-q3$ -- from rule $r\ref{r:nextstate}$ and supported rule proposition \label{reset:x:1:k}
\item \label{reset:x:2:k} $holds(p,q1,k) \in A$ represents $q1=M_k(p)$ -- construction, inductive assumption; 
and $\{add(p,q2_0,t_0,k), \dots, $ $add(p,q2_j,t_j,k)\} \subseteq A : q2_0 + \dots + q2_j = q2$  %
and $\{del(p,q3_0,t_0,k), \dots, $ $del(p,q3_l,t_l,k)\} \subseteq A : q3_0 + \dots + q3_l = q3$ %
 -- rules $r\ref{r:totincr},r\ref{r:totdecr}$ using supported rule proposition
\item Then $\{ fires(t_0,k), \dots, $ $fires(t_j,k) \} \subseteq A$ and $\{ fires(t_0,k), \dots, $ $fires(t_l,k) \} \\ \subseteq A$ -- rules $r\ref{r:r:add},r\ref{r:r:del}$ and supported rule proposition, respectively
\item Then $\{ fires(t_0,k), \dots, fires(t_j,k) \} \cup \{ fires(t_0,k), \dots, $ $fires(t_l,k) \} \subseteq $ $A = \{ fires(t_0,k), \dots, $ $ fires(t_x,k) \} \subseteq A$ -- set union of subsets
\item Then for each $fires(t_x,k) \in A$ we have $t_x \in T_k$ -- already shown in item~\ref{prove:reset:fires_tk} above
\item Then $q = M_k(p) + \sum_{t_x \in T_0 \wedge p \in t_x \bullet}{W(t_x,p)} - (\sum_{t_x \in T_k \wedge p \in \bullet t_x}{W(p,t_x)} + \\ \sum_{t_x \in T_k \wedge p \in R(t_x)}{M_k(p)})$ -- from 
\eqref{reset:x:2:k} above and the following
\begin{enumerate}
\item Each $add(p,q_j,t_j,k) \in A$ represents $q_j=W(t_j,p)$ for $p \in t_j \bullet$ -- rule $r\ref{r:r:add} $ encoding, and definition~\ref{def:pnr:texec} of transition execution in $PN$ %
\item Each $del(p,t_y,q_y,k) \in A$ represents either $q_y=W(p,t_y)$ for $p \in \bullet t_y$, or $q_y=M_k(p)$ for $p \in R(t_y)$ -- from rule $r\ref{r:r:del} ,f\ref{f:r:ptarc} $ encoding and definition~\ref{def:pnr:texec} of transition execution in $PN$; or from rule $r\ref{r:r:del} ,f\ref{f:rptarc}$ encoding and definition of reset arc in $PN$
\item Each $tot\_incr(p,q2,k) \in A$ represents $q2=\sum_{t_x \in T_k \wedge p \in t_x  \bullet}{W(t_x,p)}$ -- aggregate assignment atom semantics in rule $r\ref{r:totincr}$
\item Each $tot\_decr(p,q3,0) \in A$ represents $q3=\sum_{t_x \in T_k \wedge p \in \bullet t_x}{W(p,t_x)} + $ $\sum_{t_x \in T_k \wedge p \in R(t_x)}{M_0(p)}$ -- aggregate assignment atom semantics in rule $r\ref{r:totdecr}$
\end{enumerate}
\item Then, $M_{k+1}(p) = q$ -- since $holds(p,q,k+1) \in A$ encodes $q=M_{k+1}(p)$ -- from construction 
\end{enumerate}
\end{enumerate}

\noindent
As a result, for any $n > k$, $T_n$ is a valid firing set w.r.t. $M_n$ and its firing produces marking $M_{n+1}$. 

\noindent
{\bf Conclusion:} Since both \eqref{prove:x2a:reset} and \eqref{prove:a2x:reset} hold, $X=M_0,T_0,M_1,\dots,M_k,T_{k+1}$ is an execution sequence of $PN(P,T,E,W,R)$ (w.r.t $M_0$) iff there is an answer set $A$ of $\Pi^2(PN,M_0,k,ntok)$ such that \eqref{eqn:reset:fires} and \eqref{eqn:reset:holds} hold.

\section{Poof of Proposition~\ref{prop:inhibit}}

Let $PN=(P,T,E,W,R,I)$ be a Petri Net, $M_0$ be its initial marking and let $\Pi^3(PN,M_0,k,ntok)$ be the ASP encoding of $PN$ and $M_0$ over a simulation length $k$, with maximum $ntok$ tokens on any place node, as defined in section~\ref{sec:enc_inhibit}. Then $X=M_0,T_0,M_1,\dots,M_k,T_k,M_{k+1}$ is an execution sequence of $PN$ (w.r.t. $M_0$) iff there is an answer set $A$ of $\Pi^3(PN,M_0,k,ntok)$ such that: 
\begin{equation}
\{ fires(t,ts) : t \in T_{ts}, 0 \leq ts \leq k\} = \{ fires(t,ts) : fires(t,ts) \in A \} \label{eqn:inhibit:fires}
\end{equation}
\begin{equation}
\begin{split}
\{ holds(p,q,ts) &: p \in P, q = M_{ts}(p), 0 \leq ts \leq k+1 \} \\
&= \{ holds(p,q,ts) : holds(p,q,ts) \in A \} \label{eqn:inhibit:holds}
\end{split}
\end{equation}

We prove this by showing that:
\begin{enumerate}[(I)]
\item Given an execution sequence $X$, we create a set $A$ such that it satisfies \eqref{eqn:inhibit:fires} and \eqref{eqn:max:holds} and show that $A$ is an answer set of $\Pi^3$ \label{prove:x2a:inhibit}
\item Given an answer set $A$ of $\Pi^3$, we create an execution sequence $X$ such that \eqref{eqn:max:fires} and \eqref{eqn:inhibit:holds} are satisfied. \label{prove:a2x:inhibit}
\end{enumerate}

\noindent
{\bf First we show (\ref{prove:x2a:inhibit})}: Given $PN$ and an execution sequence $X$ of $PN$, we create a set $A$ as a union of the following sets:
\begin{enumerate}
\item $A_1=\{ num(n) : 0 \leq n \leq ntok \}$ %
\item $A_2=\{ time(ts) : 0 \leq ts \leq k\}$ %
\item $A_3=\{ place(p) : p \in P \}$ %
\item $A_4=\{ trans(t) : t \in T \}$ %
\item $A_5=\{ ptarc(p,t,n,ts) : (p,t) \in E^-, n=W(p,t), 0 \leq ts \leq k \}$, where $E^- \subseteq E$ %
\item $A_6=\{ tparc(t,p,n,ts) : (t,p) \in E^+, n=W(t,p), 0 \leq ts \leq k \}$, where $E^+ \subseteq E$ %
\item $A_7=\{ holds(p,q,0) : p \in P, q=M_{0}(p) \}$ %
\item $A_8=\{ notenabled(t,ts) : t \in T, 0 \leq ts \leq k, (\exists p \in \bullet t, M_{ts}(p) < W(p,t)) \vee (\exists p \in I(t), M_{ts}(p) \neq 0) \}$ \newline per definition~\ref{def:pnri:enable} (enabled transition) %
\item $A_9=\{ enabled(t,ts) : t \in T, 0 \leq ts \leq k, (\forall p \in \bullet t, W(p,t) \leq M_{ts}(p)) \wedge (\forall p \in I(t), M_{ts}(p) = 0) \}$ \newline per definition~\ref{def:pnri:enable} (enabled transition) %
\item $A_{10}=\{ fires(t,ts) : t \in T_{ts}, 0 \leq ts \leq k \}$ \newline per definition~\ref{def:pnri:firing_set} (firing set), only an enabled transition may fire
\item $A_{11}=\{ add(p,q,t,ts) : t \in T_{ts}, p \in t \bullet, q=W(t,p), 0 \leq ts \leq k \}$ \newline per definition~\ref{def:pnri:texec} (transition execution) %
\item $A_{12}=\{ del(p,q,t,ts) : t \in T_{ts}, p \in \bullet t, q=W(p,t), 0 \leq ts \leq k \} \cup \{ del(p,q,t,ts) : t \in T_{ts}, p \in R(t), q=M_{ts}(p), 0 \leq ts \leq k \}$ \newline per definition~\ref{def:pnri:exec} (transition execution) %
\item $A_{13}=\{ tot\_incr(p,q,ts) : p \in P, q=\sum_{t \in T_{ts}, p \in t \bullet}{W(t,p)}, 0 \leq ts \leq k \}$ \newline per definition~\ref{def:pnri:exec} (firing set execution) %
\item $A_{14}=\{ tot\_decr(p,q,ts): p \in P, q=\sum_{t \in T_{ts}, p \in \bullet t}{W(p,t)}+\sum_{t \in T_{ts}, p \in R(t) }{M_{ts}(p)}, 0 \leq ts \leq k \}$ \newline per definition~\ref{def:pnri:exec} (firing set execution) %
\item $A_{15}=\{ consumesmore(p,ts) : p \in P, q=M_{ts}(p), q1=\sum_{t \in T_{ts}, p \in \bullet t}{W(p,t)} + \sum_{t \in T_{ts}, p \in R(t)}{M_{ts}(p)}, q1 > q, 0 \leq ts \leq k \}$ \newline per definition~\ref{def:pnri:conflict} (conflicting transitions) %
\item $A_{16}=\{ consumesmore : \exists p \in P : q=M_{ts}(p), q1=\sum_{t \in T_{ts}, p \in \bullet t}{W(p,t)} + \sum_{t \in T_{ts}, p \in R(t)}(M_{ts}(p)), q1 > q, 0 \leq ts \leq k \}$ \newline per definition~\ref{def:pnri:conflict} (conflicting transitions) %
\item $A_{17}=\{ could\_not\_have(t,ts) :  t \in T, (\forall p \in \bullet t, W(p,t) \leq M_{ts}(p)), t \not\in T_{ts}, (\exists p \in \bullet t : W(p,t) > M_{ts}(p) - (\sum_{t' \in T_{ts}, p \in \bullet t'}{W(p,t')} + \sum_{t' \in T_{ts}, p \in R(t')} M_{ts}(p)), 0 \leq ts \leq k \}$ \newline
per the maximal firing set semantics
\item $A_{18}=\{ holds(p,q,ts+1) : p \in P, q=M_{ts+1}(p), 0 \leq ts < k\}$, \newline where $M_{ts+1}(p) = M_{ts}(p) - (\sum_{\substack{t \in T_{ts}, p \in \bullet t}}{W(p,t)} + $ $\sum_{t \in T_{ts}, p \in R(t)}M_{ts}(p)) + \\ \sum_{\substack{t \in T_{ts}, p \in t \bullet}}{W(t,p)}$ \newline according to definition~\ref{def:pnri:firing_set} (firing set execution) %
\item $A_{19}=\{ ptarc(p,t,n,ts) : p \in R(t), n = M_{ts}(p), n > 0, 0 \leq ts \leq k \}$
\item $A_{20}=\{ iptarc(p,t,1,ts) : p \in P, 0 \leq ts < k \}$
\end{enumerate}

\noindent
{\bf We show that $A$ satisfies \eqref{eqn:inhibit:fires} and \eqref{eqn:inhibit:holds}, and $A$ is an answer set of $\Pi^3$.}

$A$ satisfies \eqref{eqn:inhibit:fires} and \eqref{eqn:inhibit:holds} by its construction above. We show $A$ is an answer set of $\Pi^3$ by splitting. We split $lit(\Pi^3)$ into a sequence of $7k+9$ sets:

\begin{itemize}\renewcommand{\labelitemi}{$\bullet$}
\item $U_0= head(f\ref{f:place}) \cup head(f\ref{f:trans}) \cup head(f\ref{f:time}) \cup head(f\ref{f:num}) \cup head(i\ref{i:holds}) = \{place(p) : p \in P\} \cup \{ trans(t) : t \in T\} \cup \{ time(0), \dots, time(k)\} \cup \{num(0), \dots, num(ntok)\} \cup \{ holds(p,q,0) : p \in P, q=M_0(p) \} $
\item $U_{7k+1}=U_{7k+0} \cup head(f\ref{f:r:ptarc})^{ts=k} \cup head(f\ref{f:r:tparc})^{ts=k} \cup head(f\ref{f:rptarc})^{ts=k} \cup head(f\ref{f:iptarc})^{ts=k} = U_{7k+0} \cup \{ ptarc(p,t,n,k) : (p,t) \in E^-, n=W(p,t) \} \cup \{  tparc(t,p,n,k) : (t,p) \in E^+, n=W(t,p) \} \cup \{ ptarc(p,t,n,k) : p \in R(t), n=M_{k}(p), n > 0 \} \cup \\ \{ iptarc(p,t,1,k) : p \in I(t) \}$
\item $U_{7k+2}=U_{7k+1} \cup head(e\ref{e:r:ne:ptarc} )^{ts=k} \cup head(e\ref{e:ne:iptarc})^{ts=k} = U_{7k+1} \cup \{ notenabled(t,k) : t \in T \}$
\item $U_{7k+3}=U_{7k+2} \cup head(e\ref{e:enabled})^{ts=k} = U_{7k+2} \cup \{ enabled(t,k) : t \in T \}$
\item $U_{7k+4}=U_{7k+3} \cup head(a\ref{a:fires})^{ts=k} = U_{7k+3} \cup \{ fires(t,k) : t \in T \}$
\item $U_{7k+5}=U_{7k+4}  \cup head(r\ref{r:r:add} )^{ts=k} \cup head(r\ref{r:r:del} )^{ts=k} = U_{7k+4} \cup \{ add(p,q,t,k) : p \in P, t \in T, q=W(t,p) \} \cup \{ del(p,q,t,k) : p \in P, t \in T, q=W(p,t) \} \cup \{ del(p,q,t,k) : p \in P, t \in T, q=M_{k}(p) \}$
\item $U_{7k+6}=U_{7k+5} \cup head(r\ref{r:totincr})^{ts=k} \cup head(r\ref{r:totdecr})^{ts=k} = U_{7k+5} \cup \{ tot\_incr(p,q,k) : p \in P, 0 \leq q \leq ntok \} \cup \{ tot\_decr(p,q,k) : p \in P, 0 \leq q \leq ntok \}$
\item $U_{7k+7}=U_{7k+6} \cup head(r\ref{r:nextstate})^{ts=k} \cup head(a\ref{a:overc:place})^{ts=k} \cup head(a\ref{a:r:maxfire:cnh})^{ts=k} = $ $U_{7k+6} \cup \\ \{ consumesmore(p,k) : p \in P\} \cup \{ holds(p,q,k+1) : p \in P, 0 \leq q \leq ntok \} \cup \{ could\_not\_have(t,k) : t \in T \}$
\item $U_{7k+8}=U_{7k+7} \cup head(a\ref{a:overc:gen})^{ts=k} = U_{7k+7} \cup \{ consumesmore \}$
\end{itemize}
where $head(r_i)^{ts=k}$ are head atoms of ground rule $r_i$ in which $ts=k$. We write $A_i^{ts=k} = \{ a(\dots,ts) : a(\dots,ts) \in A_i, ts=k \}$ as short hand for all atoms in $A_i$ with $ts=k$. $U_{\alpha}, 0 \leq \alpha \leq 7k+8$ form a splitting sequence, since each $U_i$ is a splitting set of $\Pi^1$, and $\langle U_{\alpha}\rangle_{\alpha < \mu}$ is a monotone continuous sequence, where $U_0 \subseteq U_1 \dots \subseteq U_{7k+8}$ and $\bigcup_{\alpha < \mu}{U_{\alpha}} = lit(\Pi^1)$. 

We compute the answer set of $\Pi^3$ using the splitting sets as follows:
\begin{enumerate}
\item $bot_{U_0}(\Pi^3) = f\ref{f:place} \cup f\ref{f:trans} \cup f\ref{f:time} \cup i\ref{i:holds} \cup f\ref{f:num}$ and $X_0 = A_1 \cup \dots \cup A_4 \cup A_7$ ($= U_0$) is its answer set -- using forced atom proposition

\item $eval_{U_0}(bot_{U_1}(\Pi^3) \setminus bot_{U_0}(\Pi^3), X_0) = \{ ptarc(p,t,q,0) \text{:-}. | q=W(p,t) \} \cup \\ \{ tparc(t,p,q,0) \text{:-}. | q=W(t,p) \} \cup \{ ptarc(p,t,q,0) \text{:-}. | q=M_0(p) \} \cup \\ \{ iptarc(p,t,1,0) \text{:-}. \} $. Its answer set $X_1=A_5^{ts=0} \cup A_6^{ts=0} \cup A_{19}^{ts=0} \cup A_{20}^{ts=0}$ -- using forced atom proposition and construction of $A_5, A_6, A_{19}, A_{20}$.

\item $eval_{U_1}(bot_{U_2}(\Pi^3) \setminus bot_{U_1}(\Pi^3), X_0 \cup X_1) = \{ notenabled(t,0) \text{:-} . | (\{ trans(t), \\ ptarc(p,t,n,0), holds(p,q,0) \} \subseteq X_0 \cup X_1, \text{~where~}  q < n) \text{~or~} \{ notenabled(t,0) \text{:-} . | \\ (\{ trans(t), iptarc(p,t,n2,0), holds(p,q,0) \} \subseteq X_0 \cup X_1, \text{~where~}  q \geq n2 \} \}$. Its answer set $X_2=A_8^{ts=0}$ -- using  forced atom proposition and construction of $A_8$.
\begin{enumerate}
\item where, $q=M_0(p)$, and $n=W(p,t)$ for an arc $(p,t) \in E^-$ -- by construction $i\ref{i:holds}$ and $f\ref{f:r:ptarc}$ in $\Pi^3$, and 
\item in an arc $(p,t) \in E^-$, $p \in \bullet t$ (by definition~\ref{def:pn:preset} of preset)
\item $n2=1$ -- by construction of $iptarc$ predicates in $\Pi^3$, meaning $q \geq n2 \equiv q \geq 1 \equiv q > 0$,
\item thus, $notenabled(t,0) \in X_1$ represents $(\exists p \in \bullet t : M_0(p) < W(p,t)) \vee (\exists p \in I(t) : M_0(p) > 0)$.
\end{enumerate}

\item $eval_{U_2}(bot_{U_3}(\Pi^3) \setminus bot_{U_2}(\Pi^3), X_0 \cup \dots \cup X_2) = \{ enabled(t,0) \text{:-}. | trans(t) \in X_0 \cup \dots \cup X_2, notenabled(t,0) \notin X_0 \cup \dots \cup X_2 \}$. Its answer set is $X_3 = A_9^{ts=0}$ -- using forced atom proposition and construction of $A_9$.
\begin{enumerate}
\item since an $enabled(t,0) \in X_3$ if $\nexists ~notenabled(t,0) \in X_0 \cup \dots \cup X_2$; which is equivalent to $(\nexists p \in \bullet t : M_0(p) < W(p,t)) \wedge (\nexists p \in I(t) : M_0(p) > 0) \equiv (\forall p \in \bullet t: M_0(p) \geq W(p,t)) \wedge (\forall p \in I(t) : M_0(p) = 0)$.
\end{enumerate}

\item $eval_{U_3}(bot_{U_4}(\Pi^3) \setminus bot_{U_3}(\Pi^3), X_0 \cup \dots \cup X_3) = \{\{fires(t,0)\} \text{:-}. | enabled(t,0) \\ \text{~holds in~} X_0 \cup \dots \cup X_3 \}$. It has multiple answer sets $X_{4.1}, \dots, X_{4.n}$, corresponding to elements of power set of $fires(t,0)$ atoms in $eval_{U_3}(...)$ -- using supported rule proposition. Since we are showing that the union of answer sets of $\Pi^3$ determined using splitting is equal to $A$, we only consider the set that matches the $fires(t,0)$ elements in $A$ and call it $X_4$, ignoring the rest. Thus, $X_4 = A_{10}^{ts=0}$, representing $T_0$.
\begin{enumerate}
\item in addition, for every $t$ such that $enabled(t,0) \in X_0 \cup \dots \cup X_3,  R(t) \neq \emptyset$; $fires(t,0) \in X_4$ -- per definition~\ref{def:pnri:firing_set} (firing set); requiring that a reset transition is fired when enabled
\item thus, the firing set $T_0$ will not be eliminated by the constraint $f\ref{f:c:rptarc:elim}$ 
\end{enumerate}

\item $eval_{U_4}(bot_{U_5}(\Pi^3) \setminus bot_{U_4}(\Pi^3), X_0 \cup \dots \cup X_4) = \{add(p,n,t,0) \text{:-}. | \{fires(t,0), \\ tparc(t,p,n,0) \} \subseteq X_0 \cup \dots \cup X_4 \} \cup \{ del(p,n,t,0) \text{:-}. | \{ fires(t,0), ptarc(p,t,n,0) \} \\ \subseteq X_0 \cup \dots \cup X_4 \}$. It's answer set is $X_5 = A_{11}^{ts=0} \cup A_{12}^{ts=0}$ -- using forced atom proposition and definitions of $A_{11}$ and $A_{12}$. 
\begin{enumerate}
\item where, each $add$ atom is equivalent to $n=W(t,p) : p \in t \bullet$,
\item and each $del$ atom is equivalent to $n=W(p,t) : p \in \bullet t$; or $n=M_k(p) : p \in R(t)$,
\item representing the effect of transitions in $T_0$.
\end{enumerate}

\item $eval_{U_5}(bot_{U_6}(\Pi^3) \setminus bot_{U_5}(\Pi^3), X_0 \cup \dots \cup X_5) = \{tot\_incr(p,qq,0) \text{:-}. | $ \\$qq=\sum_{add(p,q,t,0) \in X_0 \cup \dots \cup X_5}{q} \} \cup \{ tot\_decr(p,qq,0) \text{:-}. | qq=\sum_{del(p,q,t,0) \in X_0 \cup \dots \cup X_5}{q} \}$. It's answer set is $X_6 = A_{13}^{ts=0} \cup A_{14}^{ts=0}$ --  using forced atom proposition and definitions of $A_{13}$ and $A_{14}$.
\begin{enumerate}
\item where, each $tot\_incr(p,qq,0)$, $qq=\sum_{add(p,q,t,0) \in X_0 \cup \dots X_5}{q}$ \\$\equiv qq=\sum_{t \in X_4, p \in t \bullet}{W(p,t)}$,
\item and each $tot\_decr(p,qq,0)$, $qq=\sum_{del(p,q,t,0) \in X_0 \cup \dots X_5}{q}$ \\$\equiv qq=\sum_{t \in X_4, p \in \bullet t}{W(t,p)} + \sum_{t \in X_4, p \in R(t)}{M_{k}(p)}$, 
\item represent the net effect of transitions in $T_0$.
\end{enumerate}
\item $eval_{U_6}(bot_{U_7}(\Pi^3) \setminus bot_{U_6}(\Pi^3), X_0 \cup \dots \cup X_6) = \{ consumesmore(p,0) \text{:-}. | \\ \{holds(p,q,0), tot\_decr(p,q1,0) \} \subseteq X_0 \cup \dots \cup X_6, q1 > q \} \cup \{ holds(p,q,1) \text{:-}., | \\ \{ holds(p,q1,0), tot\_incr(p,q2,0), tot\_decr(p,q3,0) \} \subseteq X_0 \cup \dots \cup X_6, q=q1+q2-q3 \} \cup \{ could\_not\_have(t,0) \text{:-}. | \{ enabled(t,0), ptarc(s,t,q), holds(s,qq,0), \\ tot\_decr(s,qqq,0) \} \subseteq X_0 \cup \dots \cup X_6, fires(t,0) \notin (X_0 \cup \dots \cup X_6), q > qq-qqq \}$. It's answer set is $X_7 = A_{15}^{ts=0} \cup A_{17}^{ts=0} \cup A_{18}^{ts=0}$ -- using forced atom proposition and definitions of $A_{15}, A_{17}, A_{18}, A_9$.
\begin{enumerate}
\item where, $consumesmore(p,0)$ represents $\exists p : q=M_0(p), q1= \\ \sum_{t \in T_0, p \in \bullet t}{W(p,t)}+\sum_{t \in T_0, p \in R(t)}{M_k(p)}, q1 > q$, indicating place $p$ will be over consumed if $T_0$ is fired, as defined in definition~\ref{def:pnri:conflict} (conflicting transitions)
\item $holds(p,q,1)$ represents $q=M_1(p)$ -- by construction of $\Pi^3$
\item and $could\_not\_have(t,0)$ represents enabled transition $t \in T_0$ that could not fire due to insufficient tokens
\item $X_7$ does not contain $could\_not\_have(t,0)$, when $enabled(t,0) \in X_0 \cup \dots \cup X_5$ and $fires(t,0) \notin X_0 \cup \dots \cup X_6$ due to construction of $A$, encoding of $a\ref{a:r:maxfire:cnh}$ and its body atoms. As a result it is not eliminated by the constraint $a\ref{a:maxfire:elim}$
\end{enumerate}

\[ \vdots \]

\item $eval_{U_{7k+0}}(bot_{U_{7k+1}}(\Pi^3) \setminus bot_{U_{7k+0}}(\Pi^3), X_0 \cup \dots \cup X_{7k+0}) = \{ ptarc(p,t,q,k) \text{:-}. | q=W(p,t) \} \cup \{ tparc(t,p,q,k) \text{:-}. | q=W(t,p) \} \cup \{ ptarc(p,t,q,k) \text{:-}. | q=M_k(p) \} \cup \{ iptarc(p,t,1,k) \text{:-}. \} $. Its answer set $X_{7k+1}=A_5^{ts=k} \cup A_6^{ts=k} \cup A_{19}^{ts=k} \cup A_{20}^{ts=k}$ -- using forced atom proposition and construction of $A_5, A_6, A_{19}, A_{20}$.

\item $eval_{U_{7k+1}}(bot_{U_{7k+2}}(\Pi^3) \setminus bot_{U_{7k+1}}(\Pi^3), X_0 \cup \dots \cup X_{7k+1}) = \{ notenabled(t,k) \text{:-} . | \\ (\{ trans(t), ptarc(p,t,n,k), holds(p,q,k) \} \subseteq X_0 \cup \dots \cup X_{7k+1}, \text{~where~}  q < n) \} \text{~or~} $ $\{ notenabled(t,k) \text{:-} . | (\{ trans(t), iptarc(p,t,n2,k), holds(p,q,k) \} \subseteq \\ X_0 \cup \dots \cup X_{7k+1}, \text{~where~}  q \geq n2 \} \}$. Its answer set $X_{7k+2}=A_8^{ts=k}$ -- using  forced atom proposition and construction of $A_8$.
\begin{enumerate}
\item where, $q=M_k(p)$, and $n=W(p,t)$ for an arc $(p,t) \in E^-$ -- by construction of $holds$ and $ptarc$ predicates in $\Pi^3$, and 
\item in an arc $(p,t) \in E^-$, $p \in \bullet t$ (by definition~\ref{def:pn:preset} of preset)
\item $n2=1$ -- by construction of $iptarc$ predicates in $\Pi^3$, meaning $q \geq n2 \equiv q \geq 1 \equiv q > 0$,
\item thus, $notenabled(t,k) \in X_{7k+1}$ represents $(\exists p \in \bullet t : M_k(p) < W(p,t)) \vee (\exists p \in I(t) : M_k(p) > k)$.
\end{enumerate}

\item $eval_{U_{7k+2}}(bot_{U_{7k+3}}(\Pi^3) \setminus bot_{U_{7k+2}}(\Pi^3), X_0 \cup \dots \cup X_{7k+2}) = \{ enabled(t,k) \text{:-}. | \\ trans(t) \in X_0 \cup \dots \cup X_{7k+2} \wedge notenabled(t,k) \notin X_0 \cup \dots \cup X_{7k+2} \}$. Its answer set is $X_{7k+3} = A_9^{ts=k}$ -- using forced atom proposition and construction of $A_9$.
\begin{enumerate}
\item since an $enabled(t,k) \in X_{7k+3}$ if $\nexists ~notenabled(t,k) \in X_0 \cup \dots \cup X_{7k+2}$; which is equivalent to $(\nexists p \in \bullet t : M_k(p) < W(p,t)) \wedge (\nexists p \in I(t) : M_k(p) > k) \equiv (\forall p \in \bullet t: M_k(p) \geq W(p,t)) \wedge (\forall p \in I(t) : M_k(p) = k)$.
\end{enumerate}

\item $eval_{U_{7k+3}}(bot_{U_{7k+4}}(\Pi^3) \setminus bot_{U_{7k+3}}(\Pi^3), X_0 \cup \dots \cup X_{7k+3}) = \{\{fires(t,k)\} \text{:-}. | \\ enabled(t,k) \\ \text{~holds in~} X_0 \cup \dots \cup X_{7k+3} \}$. It has multiple answer sets $X_{7k+4.1}, \dots, X_{1k+4.n}$, corresponding to elements of power set of $fires(t,k)$ atoms in $eval_{U_{7k+3}}(...)$ -- using supported rule proposition. Since we are showing that the union of answer sets of $\Pi^3$ determined using splitting is equal to $A$, we only consider the set that matches the $fires(t,k)$ elements in $A$ and call it $X_{7k+4}$, ignoring the rest. Thus, $X_{7k+4} = A_{10}^{ts=k}$, representing $T_k$.
\begin{enumerate}
\item in addition, for every $t$ such that $enabled(t,k) \in X_0 \cup \dots \cup X_{7k+3},  R(t) \neq \emptyset$; $fires(t,k) \in X_{7k+4}$ -- per definition~\ref{def:pnri:firing_set} (firing set); requiring that a reset transition is fired when enabled
\item thus, the firing set $T_k$ will not be eliminated by the constraint $f\ref{f:c:rptarc:elim}$ 
\end{enumerate}

\item $eval_{U_{7k+4}}(bot_{U_{7k+5}}(\Pi^3) \setminus bot_{U_{7k+4}}(\Pi^3), X_0 \cup \dots \cup X_{7k+4}) = \{add(p,n,t,k) \text{:-}. | \\ \{fires(t,k), tparc(t,p,n,k) \} \subseteq X_0 \cup \dots \cup X_{7k+4} \} \cup \{ del(p,n,t,k) \text{:-}. | \{ fires(t,k), \\ ptarc(p,t,n,k) \} \subseteq X_0 \cup \dots \cup X_{7k+4} \}$. It's answer set is $X_{7k+5} = A_{11}^{ts=k} \cup A_{12}^{ts=k}$ -- using forced atom proposition and definitions of $A_{11}$ and $A_{12}$. 
\begin{enumerate}
\item where, each $add$ atom is equivalent to $n=W(t,p) : p \in t \bullet$, 
\item and each $del$ atom is equivalent to $n=W(p,t) : p \in \bullet t$; or $n=M_k(p) : p \in R(t)$,
\item representing the effect of transitions in $T_k$
\end{enumerate}

\item $eval_{U_{7k+5}}(bot_{U_{7k+6}}(\Pi^3) \setminus bot_{U_{7k+5}}(\Pi^3), X_0 \cup \dots \cup X_{7k+5}) = \\  \{tot\_incr(p,qq,k) \text{:-}. | $ $qq=\sum_{add(p,q,t,k) \in X_0 \cup \dots \cup X_{7k+5}}{q} \} \cup $ $\{ tot\_decr(p,qq,k) \text{:-}. | $ $qq=\sum_{del(p,q,t,k) \in X_0 \cup \dots \cup X_{7k+5}}{q} \}$. It's answer set is $X_{7k+6} = A_{13}^{ts=k} \cup A_{14}^{ts=k}$ --  using forced atom proposition and definitions of $A_{13}$ and $A_{14}$.
\begin{enumerate}
\item where, each $tot\_incr(p,qq,k)$, $qq=\sum_{add(p,q,t,k) \in X_0 \cup \dots X_{7k+5}}{q}$ \\$\equiv qq=\sum_{t \in X_{7k+4}, p \in t \bullet}{W(p,t)}$, 
\item and each $tot\_decr(p,qq,k)$, $qq=\sum_{del(p,q,t,k) \in X_0 \cup \dots X_{7k+5}}{q}$ \\$\equiv qq=\sum_{t \in X_{7k+4}, p \in \bullet t}{W(t,p)} + \sum_{t \in X_{7k+4}, p \in R(t)}{M_{k}(p)}$,
\item representing the net effect of transitions in $T_k$
\end{enumerate}
\item $eval_{U_{7k+6}}(bot_{U_{7k+7}}(\Pi^3) \setminus bot_{U_{7k+6}}(\Pi^3), X_0 \cup \dots \cup X_{7k+6}) = \{ consumesmore(p,k) \text{:-}. | \\ \{holds(p,q,k), tot\_decr(p,q1,k) \} \subseteq X_0 \cup \dots \cup X_{7k+6}, q1 > q \} \cup \{ holds(p,q,1) \text{:-}., | \\ \{ holds(p,q1,k), tot\_incr(p,q2,k), tot\_decr(p,q3,k) \} \subseteq X_0 \cup \dots \cup X_{7k+6}, q=q1+q2-q3 \} \cup \{ could\_not\_have(t,k) \text{:-}. | \{ enabled(t,k), ptarc(s,t,q), holds(s,qq,k), \\ tot\_decr(s,qqq,k) \} \subseteq X_0 \cup \dots \cup X_{7k+6}, fires(t,k) \notin (X_0 \cup \dots \cup X_{7k+6}), q > qq-qqq \}$. It's answer set is $X_{7k+7} = A_{15}^{ts=k} \cup A_{17}^{ts=k} \cup A_{18}^{ts=k}$ -- using forced atom proposition and definitions of $A_{15}, A_{17}, A_{18}, A_9$.
\begin{enumerate}
\item where, $consumesmore(p,k)$ represents $\exists p : q=M_k(p), \\ q1=\sum_{t \in T_k, p \in \bullet t}{W(p,t)}+\sum_{t \in T_k, p \in R(t)}{M_k(p)}, q1 > q$, indicating place $p$ that will be over consumed if $T_k$ is fired, as defined in definition~\ref{def:pnri:conflict} (conflicting transitions),
\item $holds(p,q,k+1)$ represents $q=M_{k+1}(p)$ -- by construction of $\Pi^3$
\item and $could\_not\_have(t,k)$ represents enabled transition $t$ in $T_k$ that could not be fired due to insufficient tokens
\item $X_{7k+7}$ does not contain $could\_not\_have(t,k)$, when $enabled(t,k) \in X_0 \cup \dots \cup X_{7k+6}$ and $fires(t,k) \notin X_0 \cup \dots \cup X_{7k+6}$ due to construction of $A$, encoding of $a\ref{a:r:maxfire:cnh}$ and its body atoms. As a result it is not eliminated by the constraint $a\ref{a:maxfire:elim}$

\end{enumerate}

\item $eval_{U_{7k+7}}(bot_{U_{7k+8}}(\Pi^3) \setminus bot_{U_{7k+7}}(\Pi^3), X_0 \cup \dots \cup X_{7k+7}) = \{ consumesmore \text{:-}. | \\ \{ consumesmore(p,0),\dots,$ $consumesmore(p,k) \} \cap (X_0 \cup \dots \cup X_{7k+7}) \neq \emptyset \}$. It's answer set is $X_{7k+8} = A_{16}$ -- using forced atom proposition
\begin{enumerate}
\item $X_{7k+8}$ will be empty since none of $consumesmore(p,0),\dots, \\ consumesmore(p,k)$ hold in $X_0 \cup \dots \cup X_{7k+7}$ due to the construction of $A$, encoding of $a\ref{a:overc:place}$ and its body atoms. As a result, it is not eliminated by the constraint $a\ref{a:overc:elim}$
\end{enumerate}

\end{enumerate}

The set $X = X_0 \cup \dots \cup X_{7k+8}$ is the answer set of $\Pi^3$ by the splitting sequence theorem~\ref{def:split_seq_thm}. Each $X_i, 0 \leq i \leq 7k+8$ matches a distinct portion of $A$, and $X = A$, thus $A$ is an answer set of $\Pi^3$.

\vspace{30pt}
\noindent
{\bf Next we show (\ref{prove:a2x:inhibit}):} Given $\Pi^3$ be the encoding of a Petri Net $PN(P,T,E,W,R,I)$ with initial marking $M_0$, and $A$ be an answer set of $\Pi^3$ that satisfies (\ref{eqn:inhibit:fires}) and (\ref{eqn:inhibit:holds}), then we can construct $X=M_0,T_0,\dots,M_k,T_k,M_{k+1}$ from $A$, such that it is an execution sequence of $PN$.

We construct the $X$ as follows:
\begin{enumerate}
\item $M_i = (M_i(p_0), \dots, M_i(p_n))$, where $\{ holds(p_0,M_i(p_0),i), \dots holds(p_n,M_i(p_n),i) \} \\ \subseteq A$, for $0 \leq i \leq k+1$
\item $T_i = \{ t : fires(t,i) \in A\}$, for $0 \leq i \leq k$ 
\end{enumerate}
and show that $X$ is indeed an execution sequence of $PN$. We show this by induction over $k$ (i.e. given $M_k$, $T_k$ is a valid firing set and its firing produces marking $M_{k+1}$).

\vspace{20pt}

\noindent
{\bf Base case:} Let $k=0$, and $M_0$ is a valid marking in $X$ for $PN$, show
\begin{inparaenum}[(1)]
\item $T_0$ is a valid firing set for $M_0$, and 
\item $M_1$ is $T_0$'s target marking w.r.t. $M_0$.
\end{inparaenum} 

\begin{enumerate}
\item We show $T_0$ is a valid firing set for $M_0$. Let $\{ fires(t_0,0), \dots, fires(t_x,0) \}$ be the set of all $fires(\dots,0)$ atoms in $A$, \label{prove:inhibit:fires_t0}

\begin{enumerate}
\item Then for each $fires(t_i,0) \in A$

\begin{enumerate}
\item $enabled(t_i,0) \in A$ -- from rule $a\ref{a:fires}$ and supported rule proposition
\item Then $notenabled(t_i,0) \notin A$ -- from rule $e\ref{e:enabled}$ and supported rule proposition
\item Then $body(e\ref{e:r:ne:ptarc} )$ must not hold in $A$ and $body(e\ref{e:ne:iptarc} )$ must not hold in $A$ -- from rules $e\ref{e:r:ne:ptarc} ,e\ref{e:ne:iptarc} $ and forced atom proposition
\item Then $q \not< n_i \equiv q \geq n_i$ in $e\ref{e:r:ne:ptarc}$ for all $\{holds(p,q,0), ptarc(p,t_i,n_i,0)\} \subseteq A$ -- from $e\ref{e:r:ne:ptarc}$, forced atom proposition, and given facts ($holds(p,q,0) \in A, ptarc(p,t_i,n_i,0) \in A$)
\item And $q \not\geq n_i \equiv q < n_i$ in $e\ref{e:ne:iptarc} $ for all $\{ holds(p,q,0), iptarc(p,t_i,n_i,0) \} \subseteq A, n_i=1$; $q > n_i \equiv q = 0$ -- from $e\ref{e:ne:iptarc} $, forced atom proposition, given facts ($holds(p,q,0) \in A, iptarc(p,t_i,1,0) \in A$), and $q$ is a positive integer
\item Then $(\forall p \in \bullet t_i, M_0(p) \geq W(p,t_i)) \wedge (\forall p \in I(t_i), M_0(p) = 0)$ -- from 
\begin{enumerate}
\item $holds(p,q,0) \in A$ represents $q=M_0(p)$ -- rule $i\ref{i:holds}$ construction
\item $ptarc(p,t_i,n_i,0) \in A$ represents $n_i=W(p,t_i)$ -- rule $f\ref{f:r:ptarc}$ construction; or it represents $n_i = M_0(p)$ -- rule $f\ref{f:rptarc}$ construction; the construction of $f\ref{f:rptarc}$ ensures that $notenabled(t,0)$ is never true due to the reset arc
\item definition~\ref{def:pn:preset} of preset $\bullet t_i$ in PN
\end{enumerate}
\item Then $t_i$ is enabled and can fire in $PN$, as a result it can belong to $T_0$ -- from definition~\ref{def:pnri:enable} of enabled transition

\end{enumerate}
\item And $consumesmore \notin A$, since $A$ is an answer set of $\Pi^3$ -- from rule $a\ref{a:overc:elim}$ and supported rule proposition
\begin{enumerate}
\item Then $\nexists consumesmore(p,0) \in A$ -- from rule $a\ref{a:overc:gen}$ and supported rule proposition
\item  Then $\nexists \{ holds(p,q,0), tot\_decr(p,q1,0) \} \subseteq A : q1>q$ in $body(a\ref{a:overc:place})$ -- from $a\ref{a:overc:place}$ and forced atom proposition
\item Then $\nexists p : (\sum_{t_i \in \{t_0,\dots,t_x\}, p \in \bullet t_i}{W(p,t_i)}+\sum_{t_i \in \{t_0,\dots,t_x\}, p \in R(t_i)}{M_0(p)}) > M_0(p)$ -- from the following
\begin{enumerate}
\item $holds(p,q,0)$ represents $q=M_0(p)$ -- from rule $i\ref{i:holds}$ encoding, given
\item $tot\_decr(p,q1,0) \in A$ if $\{ del(p,q1_0,t_0,0), \dots, del(p,q1_x,t_x,0) \} \subseteq A$, where $q1 = q1_0+\dots+q1_x$ -- from $r\ref{r:totdecr}$ and forced atom proposition
\item $del(p,q1_i,t_i,0) \in A$ if $\{ fires(t_i,0), ptarc(p,t_i,q1_i,0) \} \subseteq A$ -- from $r\ref{r:r:del} $ and supported rule proposition
\item $del(p,q1_i,t_i,0)$ either represents removal of $q1_i = W(p,t_i)$ tokens from $p \in \bullet t_i$; or it represents removal of $q1_i = M_0(p)$ tokens from $p \in R(t_i)$-- from rule $r\ref{r:r:del} $, supported rule proposition, and definition~\ref{def:pnri:texec} of transition execution in $PN$
\end{enumerate}
\item Then the set of transitions in $T_0$ do not conflict -- by the definition~\ref{def:pnr:conflict} of conflicting transitions
\end{enumerate}

\item And for each $enabled(t_j,0) \in A$ and $fires(t_j,0) \notin A$, \\ $could\_not\_have(t_j,0) \in A$, since $A$ is an answer set of $\Pi^3$ - from rule $a\ref{a:maxfire:elim}$ and supported rule proposition
\begin{enumerate}
\item Then $\{ enabled(t_j,0), holds(s,qq,0), ptarc(s,t_j,q,0), \\ tot\_decr(s,qqq,0) \} \subseteq A$, such that $q > qq - qqq$ and $fires(t_j,0) \notin A$ - from rule $a\ref{a:r:maxfire:cnh}$ and supported rule proposition
\item Then for an $s \in \bullet t_j \cup R(t_j)$, $q > M_0(s) - (\sum_{t_i \in T_0, s \in \bullet t_i}{W(s,t_i)} + \sum_{t_i \in T_0, s \in R(t_i)}{M_0(s))}$, where $q=W(s,t_j) \text{~if~} s \in \bullet t_j, \text{~or~} M_0(s) \text{~otherwise}$ - from the following: %
\begin{enumerate}
\item $ptarc(s,t_i,q,0)$ represents $q=W(s,t_i)$ if $(s,t_i) \in E^-$ or $q=M_0(s)$ if $s \in R(t_i)$ -- from rule $f\ref{f:r:ptarc},f\ref{f:rptarc}$ construction
\item $holds(s,qq,0)$ represents $qq=M_0(s)$ -- from $i\ref{i:holds}$ construction
\item $tot\_decr(s,qqq,0) \in A$ if $\{ del(s,qqq_0,t_0,0), \dots, del(s,qqq_x,t_x,0) \} \subseteq A$ -- from rule $r\ref{r:totdecr}$ construction and supported rule proposition
\item $del(s,qqq_i,t_i,0) \in A$ if $\{ fires(t_i,0), ptarc(s,t_i,qqq_i,0) \} \subseteq A$ -- from rule $r\ref{r:r:del} $ and supported rule proposition
\item $del(s,qqq_i,t_i,0)$ represents $qqq_i = W(s,t_i) : t_i \in T_0, (s,t_i) \in E^-$, or $qqq_i = M_0(t_i) : t_i \in T_0, s \in R(t_i)$ -- from rule $f\ref{f:r:ptarc},f\ref{f:rptarc}$ construction 
\item $tot\_decr(q,qqq,0)$ represents $\sum_{t_i \in T_0, s \in \bullet t_i}{W(s,t_i)} + \\ \sum_{t_i \in T_0, s \in R(t_i)}{M_0(s)}$ -- from (C,D,E) above 
\end{enumerate}

\item Then firing $T_0 \cup \{ t_j \}$ would have required more tokens than are present at its source place $s \in \bullet t_j \cup R(t_j)$. Thus, $T_0$ is a maximal set of transitions that can simultaneously fire.
\end{enumerate}

\item And for each reset transition $t_r$ with $enabled(t_r,0) \in A$, $fires(t_r,0) \in A$, since $A$ is an answer set of $\Pi^2$ - from rule $f\ref{f:c:rptarc:elim}$ and supported rule proposition
\begin{enumerate}
\item Then, the firing set $T_0$ satisfies the reset-transition requirement of definition~\ref{def:pnri:firing_set} (firing set)
\end{enumerate}

\item Then $\{t_0, \dots, t_x\} = T_0$ -- using 1(a),1(b),1(d) above; and using 1(c) it is a maximal firing set  
\end{enumerate}

\item We show $M_1$ is produced by firing $T_0$ in $M_0$. Let $holds(p,q,1) \in A$
\begin{enumerate}
\item Then $\{ holds(p,q1,0), tot\_incr(p,q2,0), tot\_decr(p,q3,0) \} \subseteq A : q=q1+q2-q3$ -- from rule $r\ref{r:nextstate}$ and supported rule proposition \label{inhibit:x:1}
\item \label{inhibit:x:2} Then $holds(p,q1,0) \in A$ represents $q1=M_0(p)$ -- given, rule $i\ref{i:holds}$ construction;  
and $\{add(p,q2_0,t_0,0), \dots, $ $add(p,q2_j,t_j,0)\} \subseteq A : q2_0 + \dots + q2_j = q2$  %
 and $\{del(p,q3_0,t_0,0), \dots, $ $del(p,q3_l,t_l,0)\} \subseteq A : q3_0 + \dots + q3_l = q3$ %
   -- rules $r\ref{r:totincr},r\ref{r:totdecr}$ and supported rule proposition, respectively
\item Then $\{ fires(t_0,0), \dots, fires(t_j,0) \} \subseteq A$ and $\{ fires(t_0,0), \dots, fires(t_l,0) \} \\ \subseteq A$ -- rules $r\ref{r:r:add},r\ref{r:r:del}$ and supported rule proposition, respectively
\item Then $\{ fires(t_0,0), \dots, $ $fires(t_j,0) \} \cup \{ fires(t_0,0), \dots, $ $fires(t_l,0) \} \subseteq A = \{ fires(t_0,0), \dots, $ $fires(t_x,0) \} \subseteq A$ -- set union of subsets
\item Then for each $fires(t_x,0) \in A$ we have $t_x \in T_0$ -- already shown in item~\ref{prove:inhibit:fires_t0} above
\item Then $q = M_0(p) + \sum_{t_x \in T_0 \wedge p \in t_x \bullet}{W(t_x,p)} - (\sum_{t_x \in T_0 \wedge p \in \bullet t_x}{W(p,t_x)} + \\ \sum_{t_x \in T_0 \wedge p \in R(t_x)}{M_0(p)})$ -- from 
\eqref{inhibit:x:2} above and the following
\begin{enumerate}
\item Each $add(p,q_j,t_j,0) \in A$ represents $q_j=W(t_j,p)$ for $p \in t_j \bullet$ -- rule $r\ref{r:r:add} $ encoding, and definition~\ref{def:pnri:texec} of transition execution in $PN$ %
\item Each $del(p,t_y,q_y,0) \in A$ represents either $q_y=W(p,t_y)$ for $p \in \bullet t_y$, or $q_y=M_0(p)$ for $p \in R(t_y)$ -- from rule $r\ref{r:r:del} ,f\ref{f:r:ptarc} $ encoding and definition~\ref{def:pnri:texec} of transition execution in $PN$; or from rule $r\ref{r:r:del} ,f\ref{f:rptarc}$ encoding and definition of reset arc in $PN$
\item Each $tot\_incr(p,q2,0) \in A$ represents $q2=\sum_{t_x \in T_0 \wedge p \in t_x  \bullet}{W(t_x,p)}$ -- aggregate assignment atom semantics in rule $r\ref{r:totincr}$
\item Each $tot\_decr(p,q3,0) \in A$ represents $q3=\sum_{t_x \in T_0 \wedge p \in \bullet t_x}{W(p,t_x)} + $ $\sum_{t_x \in T_0 \wedge p \in R(t_x)}{M_0(p)}$ -- aggregate assignment atom semantics in rule $r\ref{r:totdecr}$
\end{enumerate}
\item Then, $M_1(p) = q$ -- since $holds(p,q,1) \in A$ encodes $q=M_1(p)$ -- from construction
\end{enumerate}
\end{enumerate}

\noindent
{\bf Inductive Step:} Let $k > 0$, and $M_k$ is a valid marking in $X$ for $PN$, show 
\begin{inparaenum}[(1)]
\item $T_k$ is a valid firing set for $M_k$, and 
\item firing $T_k$ in $M_k$ produces marking $M_{k+1}$.
\end{inparaenum}

\begin{enumerate}
\item We show that $T_k$ is a valid firing set in $M_k$. Let $\{ fires(t_0,k), \dots, fires(t_x,k) \}$ be the set of all $fires(\dots,k)$ atoms in $A$, \label{prove:inhibit:fires_tk}

\begin{enumerate}
\item Then for each $fires(t_i,k) \in A$

\begin{enumerate}
\item $enabled(t_i,k) \in A$ -- from rule $a\ref{a:fires}$ and supported rule proposition
\item Then $notenabled(t_i,k) \notin A$ -- from rule $e\ref{e:enabled}$ and supported rule proposition
\item Then $body(e\ref{e:r:ne:ptarc} )$ must not hold in $A$ and $body(e\ref{e:ne:iptarc} )$ must not hold in $A$ -- from rule $e\ref{e:r:ne:ptarc} ,e\ref{e:ne:iptarc} $ and forced atom proposition
\item Then $q \not< n_i \equiv q \geq n_i$ in $e\ref{e:r:ne:ptarc}$ for all $\{holds(p,q,k), ptarc(p,t_i,n_i,k)\} \subseteq A$ -- from $e\ref{e:r:ne:ptarc}$, forced atom proposition, and given facts ($holds(p,q,k) \in A, ptarc(p,t_i,n_i,k) \in A$)
\item And $q \not\geq n_i \equiv q < n_i$ in $e\ref{e:ne:iptarc} $ for all $\{ holds(p,q,k), iptarc(p,t_i,n_i,k) \} \subseteq A, n_i=1; q > n_i \equiv q = 0$ -- from $e\ref{e:ne:iptarc} $, forced atom proposition, given facts $(holds(p,q,k) \in A, iptarc(p,t_i,1,k) \in A)$, and $q$ is a positive integer
\item Then $(\forall p \in \bullet t_i, M_k(p) \geq W(p,t_i)) \wedge (\forall p \in I(t_i), M_k(p) = 0)$ -- from 
\begin{enumerate}
\item $holds(p,q,k) \in A$ represents $q=M_k(p)$ -- inductive assumption, given
\item $ptarc(p,t_i,n_i,k) \in A$ represents $n_i=W(p,t_i)$ -- rule $f\ref{f:r:ptarc}$ construction; or it represents $n_i=M_k(p)$ -- rule $f\ref{f:rptarc}$ construction; the construction of $f\ref{f:rptarc}$ ensures that $notenabled(t,0)$ is never true due to the reset arc 
\item definition~\ref{def:pn:preset} of preset $\bullet t_i$ in PN
\end{enumerate}
\item Then $t_i$ is enabled and can fire in $PN$, as a result it can belong to $T_k$ -- from definition~\ref{def:pnri:enable} of enabled transition

\end{enumerate}
\item And $consumesmore \notin A$, since $A$ is an answer set of $\Pi^3$ -- from rule $a\ref{a:overc:elim}$ and supported rule proposition
\begin{enumerate}
\item Then $\nexists consumesmore(p,k) \in A$ -- from rule $a\ref{a:overc:gen}$ and supported rule proposition
\item  Then $\nexists \{ holds(p,q,k), tot\_decr(p,q1,k) \} \subseteq A : q1>q$ in $body(a\ref{a:overc:place})$ -- from $a\ref{a:overc:place}$ and forced atom proposition
\item Then $\nexists p : (\sum_{t_i \in \{t_0,\dots,t_x\}, p \in \bullet t_i}{W(p,t_i)}+\sum_{t_i \in \{t_0,\dots,t_x\}, p \in R(t_i)}{M_k(p)}) > M_k(p)$ -- from the following
\begin{enumerate}
\item $holds(p,q,k)$ represents $q=M_k(p)$ -- inductive assumption, construction
\item $tot\_decr(p,q1,k) \in A$ if $\{ del(p,q1_0,t_0,k), \dots, del(p,q1_x,t_x,k) \} \subseteq A$, where $q1 = q1_0+\dots+q1_x$ -- from $r\ref{r:totdecr}$ and forced atom proposition
\item $del(p,q1_i,t_i,k) \in A$ if $\{ fires(t_i,k), ptarc(p,t_i,q1_i,k) \} \subseteq A$ -- from $r\ref{r:r:del} $ and supported rule proposition
\item $del(p,q1_i,t_i,k)$ either represents removal of $q1_i = W(p,t_i)$ tokens from $p \in \bullet t_i$; or it represents removal of $q1_i = M_k(p)$ tokens from $p \in R(t_i)$-- from rule $r\ref{r:r:del} $, supported rule proposition, and definition~\ref{def:pnri:texec} of transition execution in $PN$
\end{enumerate}
\item Then $T_k$ does not contain conflicting transitions -- by the definition~\ref{def:pnr:conflict} of conflicting transitions
\end{enumerate}

\item And for each $enabled(t_j,k) \in A$ and $fires(t_j,k) \notin A$, $could\_not\_have(t_j,k) \in A$, since $A$ is an answer set of $\Pi^3$ - from rule $a\ref{a:maxfire:elim}$ and supported rule proposition
\begin{enumerate}
\item Then $\{ enabled(t_j,k), holds(s,qq,k), ptarc(s,t_j,q,k), \\ tot\_decr(s,qqq,k) \} \subseteq A$, such that $q > qq - qqq$ and $fires(t_j,k) \notin A$ - from rule $a\ref{a:r:maxfire:cnh}$ and supported rule proposition
\item Then for an $s \in \bullet t_j \cup R(t_j)$, $q > M_k(s) - (\sum_{t_i \in T_k, s \in \bullet t_i}{W(s,t_i)} + \sum_{t_i \in T_k, s \in R(t_i)}{M_k(s))}$, where $q=W(s,t_j) \text{~if~} s \in \bullet t_j, \text{~or~} M_k(s) \text{~otherwise}$ - from the following: %
\begin{enumerate}
\item $ptarc(s,t_i,q,k)$ represents $q=W(s,t_i)$ if $(s,t_i) \in E^-$ or $q=M_k(s)$ if $s \in R(t_i)$ -- from rule $f\ref{f:r:ptarc},f\ref{f:rptarc}$ construction
\item $holds(s,qq,k)$ represents $qq=M_k(s)$ -- construction
\item $tot\_decr(s,qqq,k) \in A$ if $\{ del(s,qqq_0,t_0,k), \dots, del(s,qqq_x,t_x,k) \} \\ \subseteq A$ -- from rule $r\ref{r:totdecr}$ construction and supported rule proposition
\item $del(s,qqq_i,t_i,k) \in A$ if $\{ fires(t_i,k), ptarc(s,t_i,qqq_i,k) \} \subseteq A$ -- from rule $r\ref{r:r:del} $ and supported rule proposition
\item $del(s,qqq_i,t_i,k)$ represents $qqq_i = W(s,t_i) : t_i \in T_k, (s,t_i) \in E^-$, or $qqq_i = M_k(t_i) : t_i \in T_k, s \in R(t_i)$ -- from rule $f\ref{f:r:ptarc},f\ref{f:rptarc}$ construction 
\item $tot\_decr(q,qqq,k)$ represents $\sum_{t_i \in T_k, s \in \bullet t_i}{W(s,t_i)} + \\ \sum_{t_i \in T_k, s \in R(t_i)}{M_k(s)}$ -- from (C,D,E) above 
\end{enumerate}

\item Then firing $T_k \cup \{ t_j \}$ would have required more tokens than are present at its source place $s \in \bullet t_j \cup R(t_j)$. Thus, $T_k$ is a maximal set of transitions that can simultaneously fire.
\end{enumerate}

\item And for each reset transition $t_r$ with $enabled(t_r,k) \in A$, $fires(t_r,k) \in A$, since $A$ is an answer set of $\Pi^2$ - from rule $f\ref{f:c:rptarc:elim}$ and supported rule proposition
\begin{enumerate}
\item Then the firing set $T_k$ satisfies the reset transition requirement of definition~\ref{def:pnri:firing_set} (firing set)
\end{enumerate}

\item Then $\{t_0, \dots, t_x\} = T_k$ -- using 1(a),1(b), 1(d) above; and using 1(c) it is a maximal firing set  
\end{enumerate}

\item We show that $M_{k+1}$ is produced by firing $T_k$ in $M_k$. Let $holds(p,q,k+1) \in A$
\begin{enumerate}
\item Then $\{ holds(p,q1,k), tot\_incr(p,q2,k), tot\_decr(p,q3,k) \} \subseteq A : q=q1+q2-q3$ -- from rule $r\ref{r:nextstate}$ and supported rule proposition \label{inhibit:x:1:k}
\item Then, $holds(p,q1,k) \in A$ represents $q1=M_k(p)$ -- inductive assumption, construction \label{inhibit:x:2:k};
and $\{add(p,q2_0,t_0,k), \dots, $ $add(p,q2_j,t_j,k)\} \subseteq A : q2_0 + \dots + q2_j = q2$  %
and $\{del(p,q3_0,t_0,k), \dots, $ $del(p,q3_l,t_l,k)\} \subseteq A : q3_0 + \dots + q3_l = q3$ %
 -- rules $r\ref{r:totincr},r\ref{r:totdecr}$ and supported rule proposition, respectively
\item Then $\{ fires(t_0,k), \dots, fires(t_j,k) \} \subseteq A$ and $\{ fires(t_0,k), \dots, fires(t_l,k) \} \\ \subseteq A$ -- rules $r\ref{r:r:add},r\ref{r:r:del}$ and supported rule proposition, respectively
\item Then $\{ fires(t_0,k), \dots, $ $fires(t_j,k) \} \cup \{ fires(t_0,k), \dots, $ $fires(t_l,k) \} \subseteq A = $ $\{ fires(t_0,k), \dots, $ $fires(t_x,k) \} \subseteq A$ -- set union of subsets
\item Then for each $fires(t_x,k) \in A$ we have $t_x \in T_k$ -- already shown in item~\ref{prove:inhibit:fires_tk} above
\item Then $q = M_k(p) + \sum_{t_x \in T_0 \wedge p \in t_x \bullet}{W(t_x,p)} - (\sum_{t_x \in T_k \wedge p \in \bullet t_x}{W(p,t_x)} + \\ \sum_{t_x \in T_k \wedge p \in R(t_x)}{M_k(p)})$ -- from 
\eqref{inhibit:x:2:k} above and the following
\begin{enumerate}
\item Each $add(p,q_j,t_j,k) \in A$ represents $q_j=W(t_j,p)$ for $p \in t_j \bullet$ -- rule $r\ref{r:r:add} $ encoding, and definition~\ref{def:pnri:texec} of transition execution in $PN$ %
\item Each $del(p,t_y,q_y,k) \in A$ represents either $q_y=W(p,t_y)$ for $p \in \bullet t_y$, or $q_y=M_k(p)$ for $p \in R(t_y)$ -- from rule $r\ref{r:r:del} ,f\ref{f:r:ptarc} $ encoding and definition~\ref{def:pnr:texec} of transition execution in $PN$; or from rule $r\ref{r:r:del} ,f\ref{f:rptarc}$ encoding and definition of reset arc in $PN$
\item Each $tot\_incr(p,q2,k) \in A$ represents $q2=\sum_{t_x \in T_k \wedge p \in t_x  \bullet}{W(t_x,p)}$ -- aggregate assignment atom semantics in rule $r\ref{r:totincr}$
\item Each $tot\_decr(p,q3,0) \in A$ represents $q3=\sum_{t_x \in T_k \wedge p \in \bullet t_x}{W(p,t_x)} + $ $\sum_{t_x \in T_k \wedge p \in R(t_x)}{M_k(p)}$ -- aggregate assignment atom semantics in rule $r\ref{r:totdecr}$
\end{enumerate}
\item Then, $M_{k+1}(p) = q$ -- since $holds(p,q,k+1) \in A$ encodes $q=M_{k+1}(p)$ -- from construction
\end{enumerate}
\end{enumerate}

\noindent
As a result, for any $n > k$, $T_n$ will be a valid firing set for $M_n$ and $M_{n+1}$ will be its target marking. 

\noindent
{\bf Conclusion:} Since both \eqref{prove:x2a:reset} and \eqref{prove:a2x:reset} hold, $X=M_0,T_0,M_1,\dots,M_k,T_{k+1}$ is an execution sequence of $PN(P,T,E,W,R)$ (w.r.t $M_0$) iff there is an answer set $A$ of $\Pi^3(PN,M_0,k,ntok)$ such that \eqref{eqn:reset:fires} and \eqref{eqn:reset:holds} hold.

\section{Proof of Proposition~\ref{prop:query}}

Let $PN=(P,T,E,W,R,I,Q,QW)$ be a Petri Net, $M_0$ be its initial marking and let $\Pi^4(PN,M_0,k,ntok)$ be the ASP encoding of $PN$ and $M_0$ over a simulation length $k$, with maximum $ntok$ tokens on any place node, as defined in section~\ref{sec:enc_query}. Then $X=M_0,T_0,M_1,\dots,M_k,T_k,M_{k+1}$ is an execution sequence of $PN$ (w.r.t. $M_0$) iff there is an answer set $A$ of $\Pi^4(PN,M_0,k,ntok)$ such that: 
\begin{equation}
\{ fires(t,ts) : t \in T_{ts}, 0 \leq ts \leq k\} = \{ fires(t,ts) : fires(t,ts) \in A \} \label{eqn:query:fires}
\end{equation}
\begin{equation}
\begin{split}
\{ holds(p,q,ts) &: p \in P, q = M_{ts}(p), 0 \leq ts \leq k+1 \} \\
&= \{ holds(p,q,ts) : holds(p,q,ts) \in A \} \label{eqn:query:holds}
\end{split}
\end{equation}

We prove this by showing that:
\begin{enumerate}[(I)]
\item Given an execution sequence $X$, we create a set $A$ such that it satisfies \eqref{eqn:query:fires} and \eqref{eqn:query:holds} and show that $A$ is an answer set of $\Pi^4$ \label{prove:x2a:query}
\item Given an answer set $A$ of $\Pi^4$, we create an execution sequence $X$ such that \eqref{eqn:query:fires} and \eqref{eqn:query:holds} are satisfied. \label{prove:a2x:query}
\end{enumerate}

\noindent
{\bf First we show (\ref{prove:x2a:query})}: Given $PN$ and an execution sequence $X$ of $PN$, we create a set $A$ as a union of the following sets:
\begin{enumerate}
\item $A_1=\{ num(n) : 0 \leq n \leq ntok \}$ %
\item $A_2=\{ time(ts) : 0 \leq ts \leq k\}$ %
\item $A_3=\{ place(p) : p \in P \}$ %
\item $A_4=\{ trans(t) : t \in T \}$ %
\item $A_5=\{ ptarc(p,t,n,ts) : (p,t) \in E^-, n=W(p,t), 0 \leq ts \leq k \}$, where $E^- \subseteq E$ %
\item $A_6=\{ tparc(t,p,n,ts) : (t,p) \in E^+, n=W(t,p), 0 \leq ts \leq k \}$, where $E^+ \subseteq E$ %
\item $A_7=\{ holds(p,q,0) : p \in P, q=M_{0}(p) \}$ %
\item $A_8=\{ notenabled(t,ts) : t \in T, 0 \leq ts \leq k, (\exists p \in \bullet t, M_{ts}(p) < W(p,t)) \vee (\exists p \in I(t), M_{ts}(p) \neq 0) \vee (\exists (p,t) \in Q, M_{ts}(p) < QW(p,t)) \}$ \newline per definition~\ref{def:pnriq:enable} (enabled transition) %
\item $A_9=\{ enabled(t,ts) : t \in T, 0 \leq ts \leq k, (\forall p \in \bullet t, W(p,t) \leq M_{ts}(p)) \wedge (\forall p \in I(t), M_{ts}(p) = 0) \wedge (\forall (p,t) \in Q, M_{ts}(p) \geq QW(p,t)) \}$ \newline per definition~\ref{def:pnriq:enable} (enabled transition) %
\item $A_{10}=\{ fires(t,ts) : t \in T_{ts}, 0 \leq ts \leq k \}$ \newline per definition~\ref{def:pnriq:enable} (enabled transitions), only an enabled transition may fire
\item $A_{11}=\{ add(p,q,t,ts) : t \in T_{ts}, p \in t \bullet, q=W(t,p), 0 \leq ts \leq k \}$ \newline per definition~\ref{def:pnriq:texec} (transition execution) %
\item $A_{12}=\{ del(p,q,t,ts) : t \in T_{ts}, p \in \bullet t, q=W(p,t), 0 \leq ts \leq k \} \cup \{ del(p,q,t,ts) : t \in T_{ts}, p \in R(t), q=M_{ts}(p), 0 \leq ts \leq k \}$ \newline per definition~\ref{def:pnriq:texec} (transition execution) %
\item $A_{13}=\{ tot\_incr(p,q,ts) : p \in P, q=\sum_{t \in T_{ts}, p \in t \bullet}{W(t,p)}, 0 \leq ts \leq k \}$ \newline per definition~\ref{def:pnriq:exec} (firing set execution) %
\item $A_{14}=\{ tot\_decr(p,q,ts): p \in P, q=\sum_{t \in T_{ts}, p \in \bullet t}{W(p,t)}+\sum_{t \in T_{ts}, p \in R(t) }{M_{ts}(p)}, 0 \leq ts \leq k \}$ \newline per definition~\ref{def:pnriq:exec} (firing set execution) %
\item $A_{15}=\{ consumesmore(p,ts) : p \in P, q=M_{ts}(p), q1=\sum_{t \in T_{ts}, p \in \bullet t}{W(p,t)} + \sum_{t \in T_{ts}, p \in R(t)}{M_{ts}(p)}, q1 > q, 0 \leq ts \leq k \}$ \newline per definition~\ref{def:pnriq:conflict} (conflicting transitions) %
\item $A_{16}=\{ consumesmore : \exists p \in P : q=M_{ts}(p), q1=\sum_{t \in T_{ts}, p \in \bullet t}{W(p,t)} + \sum_{t \in T_{ts}, p \in R(t)}(M_{ts}(p)), q1 > q, 0 \leq ts \leq k \}$ \newline per definition~\ref{def:pnriq:conflict} (conflicting transitions) %
\item $A_{17}=\{ could\_not\_have(t,ts) :  t \in T, (\forall p \in \bullet t, W(p,t) \leq M_{ts}(p)), t \not\in T_{ts}, (\exists p \in \bullet t : W(p,t) > M_{ts}(p) - (\sum_{t' \in T_{ts}, p \in \bullet t'}{W(p,t')} + \sum_{t' \in T_{ts}, p \in R(t')} M_{ts}(p)), 0 \leq ts \leq k \}$ \newline
per the maximal firing set semantics
\item $A_{18}=\{ holds(p,q,ts+1) : p \in P, q=M_{ts+1}(p), 0 \leq ts < k\}$, \newline where $M_{ts+1}(p) = M_{ts}(p) - (\sum_{\substack{t \in T_{ts}, p \in \bullet t}}{W(p,t)} + \\ \sum_{t \in T_{ts}, p \in R(t)}M_{ts}(p)) + $ $\sum_{\substack{t \in T_{ts}, p \in t \bullet}}{W(t,p)}$ \newline according to definition~\ref{def:pnriq:firing_set} (firing set execution) %
\item $A_{19}=\{ ptarc(p,t,n,ts) : p \in R(t), n = M_{ts}(p), n > 0, 0 \leq ts \leq k \}$
\item $A_{20}=\{ iptarc(p,t,1,ts) : p \in P, 0 \leq ts < k \}$
\item $A_{21}=\{ tptarc(p,t,n,ts) : (p,t) \in Q, n=QW(p,t), 0 \leq ts \leq k \}$
\end{enumerate}

\noindent
{\bf We show that $A$ satisfies \eqref{eqn:inhibit:fires} and \eqref{eqn:inhibit:holds}, and $A$ is an answer set of $\Pi^4$.}

$A$ satisfies \eqref{eqn:inhibit:fires} and \eqref{eqn:inhibit:holds} by its construction above. We show $A$ is an answer set of $\Pi^4$ by splitting. We split $lit(\Pi^4)$ into a sequence of $7k+9$ sets:

\begin{itemize}\renewcommand{\labelitemi}{$\bullet$}
\item $U_0= head(f\ref{f:place}) \cup head(f\ref{f:trans}) \cup head(f\ref{f:time}) \cup head(f\ref{f:num}) \cup head(i\ref{i:holds}) = \{place(p) : p \in P\} \cup \{ trans(t) : t \in T\} \cup \{ time(0), \dots, time(k)\} \cup \{num(0), \dots, num(ntok)\} \cup \{ holds(p,q,0) : p \in P, q=M_0(p) \} $
\item $U_{7k+1}=U_{7k+0} \cup head(f\ref{f:r:ptarc})^{ts=k} \cup head(f\ref{f:r:tparc})^{ts=k} \cup head(f\ref{f:rptarc})^{ts=k} \cup head(f\ref{f:iptarc})^{ts=k} \cup head(f\ref{f:tptarc})^{ts=k} = U_{7k+0} \cup \{ ptarc(p,t,n,k) : (p,t) \in E^-, n=W(p,t) \} \cup \\ \{  tparc(t,p,n,k) : (t,p) \in E^+, n=W(t,p) \} \cup $ $\{ ptarc(p,t,n,k) : p \in R(t), n=M_{k}(p), n > 0 \} \cup $ $\{ iptarc(p,t,1,k) : p \in I(t) \} \cup $ $\{ tptarc(p,t,n,k) : (p,t) \in Q, n=QW(p,t) \}$
\item $U_{7k+2}=U_{7k+1} \cup head(e\ref{e:r:ne:ptarc} )^{ts=k} = U_{7k+1} \cup \{ notenabled(t,k) : t \in T \}$
\item $U_{7k+3}=U_{7k+2} \cup head(e\ref{e:enabled})^{ts=k} = U_{7k+2} \cup \{ enabled(t,k) : t \in T \}$
\item $U_{7k+4}=U_{7k+3} \cup head(a\ref{a:fires})^{ts=k} = U_{7k+3} \cup \{ fires(t,k) : t \in T \}$
\item $U_{7k+5}=U_{7k+4}  \cup head(r\ref{r:r:add} )^{ts=k} \cup head(r\ref{r:r:del} )^{ts=k} = U_{7k+4} \cup \{ add(p,q,t,k) : p \in P, t \in T, q=W(t,p) \} \cup \{ del(p,q,t,k) : p \in P, t \in T, q=W(p,t) \} \cup \{ del(p,q,t,k) : p \in P, t \in T, q=M_{k}(p) \}$
\item $U_{7k+6}=U_{7k+5} \cup head(r\ref{r:totincr})^{ts=k} \cup head(r\ref{r:totdecr})^{ts=k} = U_{7k+5} \cup \{ tot\_incr(p,q,k) : p \in P, 0 \leq q \leq ntok \} \cup \{ tot\_decr(p,q,k) : p \in P, 0 \leq q \leq ntok \}$
\item $U_{7k+7}=U_{7k+6} \cup head(r\ref{r:nextstate})^{ts=k} \cup head(a\ref{a:overc:place})^{ts=k} \cup head(a\ref{a:r:maxfire:cnh})^{ts=k} = U_{7k+6} \cup \\ \{ consumesmore(p,k) : p \in P\} \cup \{ holds(p,q,k+1) : p \in P, 0 \leq q \leq ntok \} \cup \{ could\_not\_have(t,k) : t \in T \}$
\item $U_{7k+8}=U_{7k+7} \cup head(a\ref{a:overc:gen}) = U_{7k+7} \cup \{ consumesmore \}$
\end{itemize}
where $head(r_i)^{ts=k}$ are head atoms of ground rule $r_i$ in which $ts=k$. We write $A_i^{ts=k} = \{ a(\dots,ts) : a(\dots,ts) \in A_i, ts=k \}$ as short hand for all atoms in $A_i$ with $ts=k$. $U_{\alpha}, 0 \leq \alpha \leq 7k+8$ form a splitting sequence, since each $U_i$ is a splitting set of $\Pi^4$, and $\langle U_{\alpha}\rangle_{\alpha < \mu}$ is a monotone continuous sequence, where $U_0 \subseteq U_1 \dots \subseteq U_{8(k+1)}$ and $\bigcup_{\alpha < \mu}{U_{\alpha}} = lit(\Pi^4)$. 

We compute the answer set of $\Pi^4$ using the splitting sets as follows:
\begin{enumerate}
\item $bot_{U_0}(\Pi^4) = f\ref{f:place} \cup f\ref{f:trans} \cup f\ref{f:time} \cup i\ref{i:holds} \cup f\ref{f:num}$ and $X_0 = A_1 \cup \dots \cup A_4 \cup A_7$ ($= U_0$) is its answer set -- using forced atom proposition

\item $eval_{U_0}(bot_{U_1}(\Pi^4) \setminus bot_{U_0}(\Pi^4), X_0) = \{ ptarc(p,t,q,0) \text{:-}. | q=W(p,t) \} \cup \\ \{ tparc(t,p,q,0) \text{:-}. | q=W(t,p) \} \cup \{ ptarc(p,t,q,0) \text{:-}. | q=M_0(p) \} \cup \\ \{ iptarc(p,t,1,0) \text{:-}. \} \cup \{ tptarc(p,t,q,0) \text{:-}. | q = QW(p,t) \} $. Its answer set $X_1=A_5^{ts=0} \cup A_6^{ts=0} \cup A_{19}^{ts=0} \cup A_{20}^{ts=0} \cup A_{21}^{ts=0}$ -- using forced atom proposition and construction of $A_5, A_6, A_{19}, A_{20}, A_{21}$.

\item $eval_{U_1}(bot_{U_2}(\Pi^4) \setminus bot_{U_1}(\Pi^4), X_0 \cup X_1) = \{ notenabled(t,0) \text{:-} . | (\{ trans(t), \\ ptarc(p,t,n,0), holds(p,q,0) \} \subseteq X_0 \cup X_1, \text{~where~}  q < n) \text{~or~} (\{ notenabled(t,0) \text{:-} . | \\ (\{ trans(t), iptarc(p,t,n2,0), holds(p,q,0) \} \subseteq X_0 \cup X_1, \text{~where~}  q \geq n2 \}) \text{~or~} \\ (\{ trans(t), tptarc(p,t,n3,0), holds(p,q,0) \} \subseteq X_0 \cup X_1, \text{~where~} q < n3) \}$. Its answer set $X_2=A_8^{ts=0}$ -- using  forced atom proposition and construction of $A_8$.
\begin{enumerate}
\item where, $q=M_0(p)$, and $n=W(p,t)$ for an arc $(p,t) \in E^-$ -- by construction of $i\ref{i:holds}$ and $f\ref{f:r:ptarc}$ in $\Pi^4$, and 
\item in an arc $(p,t) \in E^-$, $p \in \bullet t$ (by definition~\ref{def:pn:preset} of preset)
\item $n2=1$ -- by construction of $iptarc$ predicates in $\Pi^4$, meaning $q \geq n2 \equiv q \geq 1 \equiv q > 0$,
\item $tptarc(p,t,n3,0)$ represents $n3=QW(p,t)$, where $(p,t) \in Q$ 
\item thus, $notenabled(t,0) \in X_1$ represents $(\exists p \in \bullet t : M_0(p) < W(p,t)) \vee (\exists p \in I(t) : M_0(p) > 0) \vee (\exists (p,t) \in Q : M_{ts}(p) < QW(p,t))$.
\end{enumerate}

\item $eval_{U_2}(bot_{U_3}(\Pi^4) \setminus bot_{U_2}(\Pi^4), X_0 \cup \dots \cup X_2) = \{ enabled(t,0) \text{:-}. | trans(t) \in X_0 \cup \dots \cup X_2, notenabled(t,0) \notin X_0 \cup \dots \cup X_2 \}$. Its answer set is $X_3 = A_9^{ts=0}$ -- using forced atom proposition and construction of $A_9$.
\begin{enumerate}
\item since an $enabled(t,0) \in X_3$ if $\nexists ~notenabled(t,0) \in X_0 \cup \dots \cup X_2$; which is equivalent to $(\nexists p \in \bullet t : M_0(p) < W(p,t)) \wedge (\nexists p \in I(t) : M_0(p) > 0) \wedge (\nexists (p,t) \in Q : M_0(p) < QW(p,t) ) \equiv (\forall p \in \bullet t: M_0(p) \geq W(p,t)) \wedge (\forall p \in I(t) : M_0(p) = 0)$.
\end{enumerate}

\item $eval_{U_3}(bot_{U_4}(\Pi^4) \setminus bot_{U_3}(\Pi^4), X_0 \cup \dots \cup X_3) = \{\{fires(t,0)\} \text{:-}. | enabled(t,0) \\ \text{~holds in~} X_0 \cup \dots \cup X_3 \}$. It has multiple answer sets $X_{4.1}, \dots, X_{4.n}$, corresponding to elements of power set of $fires(t,0)$ atoms in $eval_{U_3}(...)$ -- using supported rule proposition. Since we are showing that the union of answer sets of $\Pi^4$ determined using splitting is equal to $A$, we only consider the set that matches the $fires(t,0)$ elements in $A$ and call it $X_4$, ignoring the rest. Thus, $X_4 = A_{10}^{ts=0}$, representing $T_0$.
\begin{enumerate}
\item in addition, for every $t$ such that $enabled(t,0) \in X_0 \cup \dots \cup X_3,  R(t) \neq \emptyset$; $fires(t,0) \in X_4$ -- per definition~\ref{def:pnriq:firing_set} (firing set); requiring that a reset transition is fired when enabled
\item thus, the firing set $T_0$ will not be eliminated by the constraint $f\ref{f:c:rptarc:elim}$ 
\end{enumerate}

\item $eval_{U_4}(bot_{U_5}(\Pi^4) \setminus bot_{U_4}(\Pi^4), X_0 \cup \dots \cup X_4) = \{add(p,n,t,0) \text{:-}. | \{fires(t,0), \\ tparc(t,p,n,0) \} \subseteq X_0 \cup \dots \cup X_4 \} \cup \{ del(p,n,t,0) \text{:-}. | \{ fires(t,0), ptarc(p,t,n,0) \} \\ \subseteq X_0 \cup \dots \cup X_4 \}$. It's answer set is $X_5 = A_{11}^{ts=0} \cup A_{12}^{ts=0}$ -- using forced atom proposition and definitions of $A_{11}$ and $A_{12}$. 
\begin{enumerate}
\item where, each $add$ atom is equivalent to $n=W(t,p),p \in t \bullet$,  
\item and each $del$ atom is equivalent to $n=W(p,t), p \in \bullet t$; or $n=M_k(p), p \in R(t)$,
\item representing the effect of transitions in $T_0$ -- by construction
\end{enumerate}

\item $eval_{U_5}(bot_{U_6}(\Pi^4) \setminus bot_{U_5}(\Pi^4), X_0 \cup \dots \cup X_5) = \{tot\_incr(p,qq,0) \text{:-}. | $ \\$qq=\sum_{add(p,q,t,0) \in X_0 \cup \dots \cup X_5}{q} \} \cup \{ tot\_decr(p,qq,0) \text{:-}. | qq=\sum_{del(p,q,t,0) \in X_0 \cup \dots \cup X_5}{q} \}$. It's answer set is $X_6 = A_{13}^{ts=0} \cup A_{14}^{ts=0}$ --  using forced atom proposition and definitions of $A_{13}$ and $A_{14}$.
\begin{enumerate}
\item where, each $tot\_incr(p,qq,0)$, $qq=\sum_{add(p,q,t,0) \in X_0 \cup \dots X_5}{q}$ \\$\equiv qq=\sum_{t \in X_4, p \in t \bullet}{W(p,t)}$, 
\item and each $tot\_decr(p,qq,0)$, $qq=\sum_{del(p,q,t,0) \in X_0 \cup \dots X_5}{q}$ \\$\equiv qq=\sum_{t \in X_4, p \in \bullet t}{W(t,p)} + \sum_{t \in X_4, p \in R(t)}{M_{k}(p)}$,
\item represent the net effect of transitions in $T_0$ -- by construction
\end{enumerate}
\item $eval_{U_6}(bot_{U_7}(\Pi^4) \setminus bot_{U_6}(\Pi^4), X_0 \cup \dots \cup X_6) = \{ consumesmore(p,0) \text{:-}. | \\ \{holds(p,q,0), tot\_decr(p,q1,0) \} \subseteq X_0 \cup \dots \cup X_6, q1 > q \} \cup \{ holds(p,q,1) \text{:-}., | \\ \{ holds(p,q1,0), tot\_incr(p,q2,0), tot\_decr(p,q3,0) \} \subseteq X_0 \cup \dots \cup X_6, q=q1+q2-q3 \} \cup \{ could\_not\_have(t,0) \text{:-}. | \{ enabled(t,0), ptarc(s,t,q), holds(s,qq,0), \\ tot\_decr(s,qqq,0) \} \subseteq X_0 \cup \dots \cup X_6, fires(t,0) \notin (X_0 \cup \dots \cup X_6), q > qq-qqq \}$. It's answer set is $X_7 = A_{15}^{ts=0} \cup A_{17}^{ts=0} \cup A_{18}^{ts=0}$ -- using forced atom proposition and definitions of $A_{15}, A_{17}, A_{18}, A_9$.
\begin{enumerate}
\item where, $consumesmore(p,0)$ represents $\exists p : q=M_0(p), q1= \\ \sum_{t \in T_0, p \in \bullet t}{W(p,t)}+\sum_{t \in T_0, p \in R(t)}{M_0(p)}, q1 > q$, indicating place $p$ will be over consumed if $T_0$ is fired, as defined in definition~\ref{def:pnriq:conflict} (conflicting transitions),
\item $holds(p,q,1)$ represents $q=M_1(p)$ -- by construction of $\Pi^4$,
\item and $could\_not\_have(t,0)$ represents enabled transition $t \in T_0$ that could not fire due to insufficient tokens
\item $X_7$ does not contain $could\_not\_have(t,0)$, when $enabled(t,0) \in X_0 \cup \dots \cup X_6$ and $fires(t,0) \notin X_0 \cup \dots \cup X_6$ due to construction of $A$, encoding of $a\ref{a:r:maxfire:cnh}$ and its body atoms. As a result it is not eliminated by the constraint $a\ref{a:maxfire:elim}$
\end{enumerate}

\[ \vdots \]

\item $eval_{U_{7k+0}}(bot_{U_{7k+1}}(\Pi^4) \setminus bot_{U_{7k+0}}(\Pi^4), X_0 \cup \dots \cup X_{7k+0}) = \{ ptarc(p,t,q,k) \text{:-}. | q=W(p,t) \} \cup \{ tparc(t,p,q,k) \text{:-}. | q=W(t,p) \} \cup \{ ptarc(p,t,q,k) \text{:-}. | q=M_k(p) \} \cup \{ iptarc(p,t,1,k) \text{:-}. \} $. Its answer set $X_{7k+1}=A_5^{ts=k} \cup A_6^{ts=k} \cup A_{19}^{ts=k} \cup A_{20}^{ts=k}$ -- using forced atom proposition and construction of $A_5, A_6, A_{19}, A_{20}$.

\item $eval_{U_{7k+1}}(bot_{U_{7k+2}}(\Pi^4) \setminus bot_{U_{7k+1}}(\Pi^4), X_0 \cup \dots \cup X_{7k+1}) = \{ notenabled(t,k) \text{:-} . | $ $(\{ trans(t), ptarc(p,t,n,k), holds(p,q,k) \} \subseteq X_0 \cup \dots \cup X_{7k+1}, \text{~where~}  q < n) \text{~or~} \\ \{ notenabled(t,k) \text{:-} . |  (\{ trans(t), iptarc(p,t,n2,k), holds(p,q,k) \} \subseteq X_0 \cup \dots \cup X_{7k+1}, \\ \text{~where~}  q \geq n2 \} \}$. Its answer set $X_{7k+2}=A_8^{ts=k}$ -- using  forced atom proposition and construction of $A_8$.
\begin{enumerate}
\item where, $q=M_k(p)$, and $n=W(p,t)$ for an arc $(p,t) \in E^-$ -- by construction of $holds$ and $ptarc$ predicates in $\Pi^4$, and 
\item in an arc $(p,t) \in E^-$, $p \in \bullet t$ (by definition~\ref{def:pn:preset} of preset)
\item $n2=1$ -- by construction of $iptarc$ predicates in $\Pi^4$, meaning $q \geq n2 \equiv q \geq 1 \equiv q > 0$,
\item thus, $notenabled(t,k) \in X_{7k+1}$ represents $(\exists p \in \bullet t : M_k(p) < W(p,t)) \vee (\exists p \in I(t) : M_k(p) > k)$.
\end{enumerate}

\item $eval_{U_{7k+2}}(bot_{U_{7k+3}}(\Pi^4) \setminus bot_{U_{7k+2}}(\Pi^4), X_0 \cup \dots \cup X_{7k+2}) = \{ enabled(t,k) \text{:-}. | \\ trans(t) \in X_0 \cup \dots \cup X_{7k+2} \wedge notenabled(t,k) \notin X_0 \cup \dots \cup X_{7k+2} \}$. Its answer set is $X_{7k+3} = A_9^{ts=k}$ -- using forced atom proposition and construction of $A_9$.
\begin{enumerate}
\item since an $enabled(t,k) \in X_{7k+3}$ if $\nexists ~notenabled(t,k) \in X_0 \cup \dots \cup X_{7k+2}$; which is equivalent to $(\nexists p \in \bullet t : M_k(p) < W(p,t)) \wedge (\nexists p \in I(t) : M_k(p) > k) \equiv (\forall p \in \bullet t: M_k(p) \geq W(p,t)) \wedge (\forall p \in I(t) : M_k(p) = k)$.
\end{enumerate}

\item $eval_{U_{7k+3}}(bot_{U_{7k+4}}(\Pi^4) \setminus bot_{U_{7k+3}}(\Pi^4), X_0 \cup \dots \cup X_{7k+3}) = \{\{fires(t,k)\} \text{:-}. | \\ enabled(t,k) \\ \text{~holds in~} X_0 \cup \dots \cup X_{7k+3} \}$. It has multiple answer sets $X_{7k+4.1}, \dots, X_{7k+4.n}$, corresponding to elements of power set of $fires(t,k)$ atoms in $eval_{U_{7k+3}}(...)$ -- using supported rule proposition. Since we are showing that the union of answer sets of $\Pi^4$ determined using splitting is equal to $A$, we only consider the set that matches the $fires(t,k)$ elements in $A$ and call it $X_{7k+4}$, ignoring the rest. Thus, $X_{7k+4} = A_{10}^{ts=k}$, representing $T_k$.
\begin{enumerate}
\item in addition, for every $t$ such that $enabled(t,k) \in X_0 \cup \dots \cup X_{7k+3},  R(t) \neq \emptyset$; $fires(t,k) \in X_{7k+4}$ -- per definition~\ref{def:pnriq:firing_set} (firing set); requiring that a reset transition is fired when enabled
\item thus, the firing set $T_k$ will not be eliminated by the constraint $f\ref{f:c:rptarc:elim}$ 
\end{enumerate}

\item $eval_{U_{7k+4}}(bot_{U_{7k+5}}(\Pi^4) \setminus bot_{U_{7k+4}}(\Pi^4), X_0 \cup \dots \cup X_{7k+4}) = \{add(p,n,t,k) \text{:-}. | \\ \{fires(t,k), tparc(t,p,n,k) \} \subseteq X_0 \cup \dots \cup X_{7k+4} \} \cup \{ del(p,n,t,k) \text{:-}. | \{ fires(t,k), \\ ptarc(p,t,n,k) \} \subseteq X_0 \cup \dots \cup X_{7k+4} \}$. It's answer set is $X_{7k+5} = A_{11}^{ts=k} \cup A_{12}^{ts=k}$ -- using forced atom proposition and definitions of $A_{11}$ and $A_{12}$. 
\begin{enumerate}
\item where, each $add$ atom is equivalent to $n=W(t,p) : p \in t \bullet$, 
\item and each $del$ atom is equivalent to $n=W(p,t) : p \in \bullet t$; or $n=M_k(p) : p \in R(t)$,
\item representing the effect of transitions in $T_k$
\end{enumerate}

\item $eval_{U_{7k+5}}(bot_{U_{7k+6}}(\Pi^4) \setminus bot_{U_{7k+5}}(\Pi^4), X_0 \cup \dots \cup X_{7k+5}) = \\ \{tot\_incr(p,qq,k) \text{:-}. | qq=\sum_{add(p,q,t,k) \in X_0 \cup \dots \cup X_{7k+5}}{q} \} \cup \\ \{ tot\_decr(p,qq,k) \text{:-}. | qq=\sum_{del(p,q,t,k) \in X_0 \cup \dots \cup X_{7k+5}}{q} \}$. It's answer set is $X_{7k+6} = A_{13}^{ts=k} \cup A_{14}^{ts=k}$ --  using forced atom proposition and definitions of $A_{13}$ and $A_{14}$.
\begin{enumerate}
\item where, each $tot\_incr(p,qq,k)$, $qq=\sum_{add(p,q,t,k) \in X_0 \cup \dots X_{7k+5}}{q}$ \\$\equiv qq=\sum_{t \in X_{7k+4}, p \in t \bullet}{W(p,t)}$,
\item and each $tot\_decr(p,qq,k)$, $qq=\sum_{del(p,q,t,k) \in X_0 \cup \dots X_{7k+5}}{q}$ \\$\equiv qq=\sum_{t \in X_{7k+4}, p \in \bullet t}{W(t,p)} + \sum_{t \in X_{7k+4}, p \in R(t)}{M_{k}(p)}$, 
\item represent the net effect of transition in $T_k$
\end{enumerate}
\item $eval_{U_{7k+6}}(bot_{U_{7k+7}}(\Pi^4) \setminus bot_{U_{7k+6}}(\Pi^4), X_0 \cup \dots \cup X_{7k+6}) = \{ consumesmore(p,k) \text{:-}. | \\ \{holds(p,q,k), tot\_decr(p,q1,k) \} \subseteq X_0 \cup \dots \cup X_{7k+6}, q1 > q \} \cup \{ holds(p,q,1) \text{:-}., | \\ \{ holds(p,q1,k), tot\_incr(p,q2,k), tot\_decr(p,q3,k) \} \subseteq X_0 \cup \dots \cup X_{7k+6}, q=q1+q2-q3 \} \cup \{ could\_not\_have(t,k) \text{:-}. | \{ enabled(t,k), ptarc(s,t,q), holds(s,qq,k), \\ tot\_decr(s,qqq,k) \} \subseteq X_0 \cup \dots \cup X_{7k+6}, fires(t,k) \notin (X_0 \cup \dots \cup X_{7k+6}), q > qq-qqq \}$. It's answer set is $X_{7k+7} = A_{15}^{ts=k} \cup A_{17}^{ts=k} \cup A_{18}^{ts=k}$ -- using forced atom proposition and definitions of $A_{15}, A_{17}, A_{18}, A_9$.
\begin{enumerate}
\item where, $consumesmore(p,k)$ represents $\exists p : q=M_k(p), q1= \\ \sum_{t \in T_k, p \in \bullet t}{W(p,t)}+\sum_{t \in T_k, p \in R(t)}{M_k(p)}, q1 > q$, indicating place $p$ that will be over consumed if $T_k$ is fired, as defined in definition~\ref{def:pnriq:conflict} (conflicting transitions),
\item $holds(p,q,k+1)$ represents $q=M_{k+1}(p)$ -- by construction of $\Pi^4$,
\item and $could\_not\_have(t,k)$ represents enabled transition $t$ in $T_k$ that could not be fired due to insufficient tokens
\item $X_{7k+7}$ does not contain $could\_not\_have(t,k)$, when $enabled(t,k) \in X_0 \cup \dots \cup X_{7k+6}$ and $fires(t,k) \notin X_0 \cup \dots \cup X_{7k+6}$ due to construction of $A$, encoding of $a\ref{a:r:maxfire:cnh}$ and its body atoms. As a result it is not eliminated by the constraint $a\ref{a:maxfire:elim}$
\end{enumerate}

\item $eval_{U_{7k+7}}(bot_{U_{7k+8}}(\Pi^4) \setminus bot_{U_{7k+7}}(\Pi^4), X_0 \cup \dots \cup X_{7k+7}) = \{ consumesmore \text{:-}. | \\ \{ consumesmore(p,0),\dots, $ $consumesmore(p,k) \} \cap (X_0 \cup \dots \cup X_{7k+7}) \neq \emptyset \}$. It's answer set is $X_{7k+8} = A_{16}$ -- using forced atom proposition
\begin{enumerate}
\item $X_{7k+8}$ will be empty since none of $consumesmore(p,0),\dots, \\ consumesmore(p,k)$ hold in $X_0 \cup \dots \cup X_{7k+8}$ due to the construction of $A$, encoding of $a\ref{a:overc:place}$ and its body atoms. As a result, it is not eliminated by the constraint $a\ref{a:overc:elim}$
\end{enumerate}

\end{enumerate}

The set $X = X_0 \cup \dots \cup X_{7k+8}$ is the answer set of $\Pi^4$ by the splitting sequence theorem~\ref{def:split_seq_thm}. Each $X_i, 0 \leq i \leq 7k+8$ matches a distinct portion of $A$, and $X = A$, thus $A$ is an answer set of $\Pi^4$.

\vspace{30pt}
\noindent
{\bf Next we show (\ref{prove:a2x:inhibit}):} Given $\Pi^4$ be the encoding of a Petri Net $PN(P,T,E,W,R,I)$ with initial marking $M_0$, and $A$ be an answer set of $\Pi^4$ that satisfies (\ref{eqn:inhibit:fires}) and (\ref{eqn:inhibit:holds}), then we can construct $X=M_0,T_0,\dots,M_k,T_k,M_{k+1}$ from $A$, such that it is an execution sequence of $PN$.

We construct the $X$ as follows:
\begin{enumerate}
\item $M_i = (M_i(p_0), \dots, M_i(p_n))$, where $\{ holds(p_0,M_i(p_0),i), \dots holds(p_n,M_i(p_n),i) \} \\ \subseteq A$, for $0 \leq i \leq k+1$
\item $T_i = \{ t : fires(t,i) \in A\}$, for $0 \leq i \leq k$ 
\end{enumerate}
and show that $X$ is indeed an execution sequence of $PN$. We show this by induction over $k$ (i.e. given $M_k$, $T_k$ is a valid firing set and its firing produces marking $M_{k+1}$).

\vspace{20pt}

\noindent
{\bf Base case:} Let $k=0$, and $M_0$ is a valid marking in $X$ for $PN$, show
\begin{inparaenum}[(1)]
\item $T_0$ is a valid firing set for $M_0$, and 
\item firing of $T_0$ in $M_0$ produces marking $M_1$.
\end{inparaenum} 

\begin{enumerate}
\item We show $T_0$ is a valid firing set for $M_0$. Let $\{ fires(t_0,0), \dots, fires(t_x,0) \}$ be the set of all $fires(\dots,0)$ atoms in $A$, \label{prove:query:fires_t0}

\begin{enumerate}
\item Then for each $fires(t_i,0) \in A$

\begin{enumerate}
\item $enabled(t_i,0) \in A$ -- from rule $a\ref{a:fires}$ and supported rule proposition
\item Then $notenabled(t_i,0) \notin A$ -- from rule $e\ref{e:enabled}$ and supported rule proposition
\item Then either of $body(e\ref{e:r:ne:ptarc} )$, $body(e\ref{e:ne:iptarc} )$, or $body(e\ref{e:ne:tptarc} )$ must not hold in $A$ -- from rules $e\ref{e:r:ne:ptarc} ,e\ref{e:ne:iptarc} ,e\ref{e:ne:tptarc} $ and forced atom proposition
\item Then $q \not< n_i \equiv q \geq n_i$ in $e\ref{e:r:ne:ptarc} $ for all $\{holds(p,q,0), ptarc(p,t_i,n_i,0)\} \subseteq A$ -- from $e\ref{e:r:ne:ptarc} $, forced atom proposition, and given facts ($holds(p,q,0) \in A, ptarc(p,t_i,n_i,0) \in A$)
\item And $q \not\geq n_i \equiv q < n_i$ in $e\ref{e:ne:iptarc} $ for all $\{ holds(p,q,0), iptarc(p,t_i,n_i,0) \} \subseteq A, n_i=1$; $q > n_i \equiv q = 0$ -- from $e\ref{e:ne:iptarc} $, forced atom proposition, given facts ($holds(p,q,0) \in A, iptarc(p,t_i,1,0) \in A$), and $q$ is a positive integer
\item And $q \not< n_i \equiv q \geq n_i$ in $e\ref{e:ne:tptarc} $ for all $\{ holds(p,q,0), tptarc(p,t_i,n_i,0) \} \subseteq A$ -- from $e\ref{e:ne:tptarc} $, forced atom proposition, and given facts
\item Then $(\forall p \in \bullet t_i, M_0(p) \geq W(p,t_i)) \wedge (\forall p \in I(t_i), M_0(p) = 0) \wedge (\forall (p,t_i) \in Q, M_0(p) \geq QW(p,t_i))$ -- from the following
\begin{enumerate}
\item $holds(p,q,0) \in A$ represents $q=M_0(p)$ -- rule $i\ref{i:holds}$ construction
\item $ptarc(p,t_i,n_i,0) \in A$ represents $n_i=W(p,t_i)$ -- rule $f\ref{f:r:ptarc}$ construction; or it represents $n_i = M_0(p)$ -- rule $f\ref{f:rptarc}$ construction; the construction of $f\ref{f:rptarc}$ ensures that $notenabled(t,0)$ is never true due to the reset arc
\item definition~\ref{def:pn:preset} of preset $\bullet t_i$ in $PN$
\item definition~\ref{def:pnriq:enable} of enabled transition in $PN$
\end{enumerate}
\item Then $t_i$ is enabled and can fire in $PN$, as a result it can belong to $T_0$ -- from definition~\ref{def:pnriq:enable} of enabled transition

\end{enumerate}
\item And $consumesmore \notin A$, since $A$ is an answer set of $\Pi^4$ -- from rule $a\ref{a:overc:elim}$ and supported rule proposition
\begin{enumerate}
\item Then $\nexists consumesmore(p,0) \in A$ -- from rule $a\ref{a:overc:gen}$ and supported rule proposition
\item  Then $\nexists \{ holds(p,q,0), tot\_decr(p,q1,0) \} \subseteq A : q1>q$ in $body(a\ref{a:overc:place})$ -- from $a\ref{a:overc:place}$ and forced atom proposition
\item Then $\nexists p : (\sum_{t_i \in \{t_0,\dots,t_x\}, p \in \bullet t_i}{W(p,t_i)}+\sum_{t_i \in \{t_0,\dots,t_x\}, p \in R(t_i)}{M_0(p)}) > M_0(p)$ -- from the following
\begin{enumerate}
\item $holds(p,q,0)$ represents $q=M_0(p)$ -- from rule $i\ref{i:holds}$ construction, given
\item $tot\_decr(p,q1,0) \in A$ if $\{ del(p,q1_0,t_0,0), \dots, del(p,q1_x,t_x,0) \} \subseteq A$, where $q1 = q1_0+\dots+q1_x$ -- from $r\ref{r:totdecr}$ and forced atom proposition
\item $del(p,q1_i,t_i,0) \in A$ if $\{ fires(t_i,0), ptarc(p,t_i,q1_i,0) \} \subseteq A$ -- from $r\ref{r:r:del} $ and supported rule proposition
\item $del(p,q1_i,t_i,0)$ represents removal of $q1_i = W(p,t_i)$ tokens from $p \in \bullet t_i$; or it represents removal of $q1_i = M_0(p)$ tokens from $p \in R(t_i)$-- from rule $r\ref{r:r:del} $, supported rule proposition, and definition~\ref{def:pnriq:texec} of transition execution in $PN$
\end{enumerate}
\item Then the set of transitions in $T_0$ do not conflict -- by the definition~\ref{def:pnr:conflict} of conflicting transitions
\end{enumerate}

\item And for each $enabled(t_j,0) \in A$ and $fires(t_j,0) \notin A$, \\ $could\_not\_have(t_j,0) \in A$, since $A$ is an answer set of $\Pi^4$ - from rule $a\ref{a:maxfire:elim}$ and supported rule proposition
\begin{enumerate}
\item Then $\{ enabled(t_j,0), holds(s,qq,0), ptarc(s,t_j,q,0), \\ tot\_decr(s,qqq,0) \} \subseteq A$, such that $q > qq - qqq$ and $fires(t_j,0) \notin A$ - from rule $a\ref{a:r:maxfire:cnh}$ and supported rule proposition
\item Then for an $s \in \bullet t_j \cup R(t_j)$, $q > M_0(s) - (\sum_{t_i \in T_0, s \in \bullet t_i}{W(s,t_i)} + \sum_{t_i \in T_0, s \in R(t_i)}{M_0(s))}$, where $q=W(s,t_j) \text{~if~} s \in \bullet t_j, \text{~or~} M_0(s) \text{~otherwise}$ - from the following: %
\begin{enumerate}
\item $ptarc(s,t_i,q,0)$ represents $q=W(s,t_i)$ if $(s,t_i) \in E^-$ or $q=M_0(s)$ if $s \in R(t_i)$ -- from rule $f\ref{f:r:ptarc},f\ref{f:rptarc}$ construction
\item $holds(s,qq,0)$ represents $qq=M_0(s)$ -- from $i\ref{i:holds}$ construction
\item $tot\_decr(s,qqq,0) \in A$ if $\{ del(s,qqq_0,t_0,0), \dots, del(s,qqq_x,t_x,0) \} \subseteq A$ -- from rule $r\ref{r:totdecr}$ construction and supported rule proposition
\item $del(s,qqq_i,t_i,0) \in A$ if $\{ fires(t_i,0), ptarc(s,t_i,qqq_i,0) \} \subseteq A$ -- from rule $r\ref{r:r:del} $ and supported rule proposition
\item $del(s,qqq_i,t_i,0)$ represents $qqq_i = W(s,t_i) : t_i \in T_0, (s,t_i) \in E^-$, or $qqq_i = M_0(t_i) : t_i \in T_0, s \in R(t_i)$ -- from rule $f\ref{f:r:ptarc},f\ref{f:rptarc}$ construction 
\item $tot\_decr(q,qqq,0)$ represents $\sum_{t_i \in T_0, s \in \bullet t_i}{W(s,t_i)} + \\ \sum_{t_i \in T_0, s \in R(t_i)}{M_0(s)}$ -- from (C,D,E) above 
\end{enumerate}

\item Then firing $T_0 \cup \{ t_j \}$ would have required more tokens than are present at its source place $s \in \bullet t_j \cup R(t_j)$. Thus, $T_0$ is a maximal set of transitions that can simultaneously fire.
\end{enumerate}

\item And for each reset transition $t_r$ with $enabled(t_r,0) \in A$, $fires(t_r,0) \in A$, since $A$ is an answer set of $\Pi^2$ - from rule $f\ref{f:c:rptarc:elim}$ and supported rule proposition
\begin{enumerate}
\item Then, the firing set $T_0$ satisfies the reset-transition requirement of definition~\ref{def:pnr:firing_set} (firing set)
\end{enumerate}

\item Then $\{t_0, \dots, t_x\} = T_0$ -- using 1(a),1(b),1(d) above; and using 1(c) it is a maximal firing set  
\end{enumerate}

\item We show $M_1$ is produced by firing $T_0$ in $M_0$. Let $holds(p,q,1) \in A$
\begin{enumerate}
\item Then $\{ holds(p,q1,0), tot\_incr(p,q2,0), tot\_decr(p,q3,0) \} \subseteq A : q=q1+q2-q3$ -- from rule $r\ref{r:nextstate}$ and supported rule proposition \label{query:x:1}
\item \label{query:x:2} Then $holds(p,q1,0) \in A$ represents $q1=M_0(p)$ -- given, rule $i\ref{i:holds}$ construction;  
\item Then $\{add(p,q2_0,t_0,0), \dots, add(p,q2_j,t_j,0)\} \subseteq A : q2_0 + \dots + q2_j = q2$  %
and $\{del(p,q3_0,t_0,0), \dots, del(p,q3_l,t_l,0)\} \subseteq A : q3_0 + \dots + q3_l = q3$ %
 -- rules $r\ref{r:totincr},r\ref{r:totdecr}$ and supported rule proposition, respectively
\item Then $\{ fires(t_0,0), \dots, fires(t_j,0) \} \subseteq A$ and $\{ fires(t_0,0), \dots, fires(t_l,0) \} \\ \subseteq A$ -- rules $r\ref{r:r:add},r\ref{r:r:del}$ and supported rule proposition, respectively
\item Then $\{ fires(t_0,0), \dots, fires(t_j,0) \} \cup \{ fires(t_0,0), \dots, fires(t_l,0) \} \subseteq A = \{ fires(t_0,0), \dots, $ $fires(t_x,0) \} \subseteq A$ -- set union of subsets
\item Then for each $fires(t_x,0) \in A$ we have $t_x \in T_0$ -- already shown in item~\ref{prove:query:fires_t0} above
\item Then $q = M_0(p) + \sum_{t_x \in T_0 \wedge p \in t_x \bullet}{W(t_x,p)} - (\sum_{t_x \in T_0 \wedge p \in \bullet t_x}{W(p,t_x)} + \\ \sum_{t_x \in T_0 \wedge p \in R(t_x)}{M_0(p)})$ -- from 
\eqref{query:x:2} above and the following
\begin{enumerate}
\item Each $add(p,q_j,t_j,0) \in A$ represents $q_j=W(t_j,p)$ for $p \in t_j \bullet$ -- rule $r\ref{r:r:add} $ encoding, and definition~\ref{def:pnriq:texec} of transition execution in $PN$ %
\item Each $del(p,t_y,q_y,0) \in A$ represents either $q_y=W(p,t_y)$ for $p \in \bullet t_y$, or $q_y=M_0(p)$ for $p \in R(t_y)$ -- from rule $r\ref{r:r:del} ,f\ref{f:r:ptarc} $ encoding and definition~\ref{def:pnriq:texec} of transition execution in $PN$; or from rule $r\ref{r:r:del} ,f\ref{f:rptarc}$ encoding and definition of reset arc in $PN$
\item Each $tot\_incr(p,q2,0) \in A$ represents $q2=\sum_{t_x \in T_0 \wedge p \in t_x  \bullet}{W(t_x,p)}$ -- aggregate assignment atom semantics in rule $r\ref{r:totincr}$
\item Each $tot\_decr(p,q3,0) \in A$ represents $q3=\sum_{t_x \in T_0 \wedge p \in \bullet t_x}{W(p,t_x)} + \sum_{t_x \in T_0 \wedge p \in R(t_x)}{M_0(p)}$ -- aggregate assignment atom semantics in rule $r\ref{r:totdecr}$
\end{enumerate}
\item Then, $M_1(p) = q$ -- since $holds(p,q,1) \in A$ encodes $q=M_1(p)$ -- from construction
\end{enumerate}
\end{enumerate}

\noindent
{\bf Inductive Step:} Let $k > 0$, and $M_k$ is a valid marking in $X$ for $PN$, show 
\begin{inparaenum}[(1)]
\item $T_k$ is a valid firing set for $M_k$, and 
\item firing $T_k$ in $M_k$ produces marking $M_{k+1}$.
\end{inparaenum}

\begin{enumerate}
\item We show that $T_k$ is a valid firing set in $M_k$. Let $\{ fires(t_0,k), \dots, fires(t_x,k) \}$ be the set of all $fires(\dots,k)$ atoms in $A$, \label{prove:query:fires_tk}

\begin{enumerate}
\item Then for each $fires(t_i,k) \in A$

\begin{enumerate}
\item $enabled(t_i,k) \in A$ -- from rule $a\ref{a:fires}$ and supported rule proposition
\item Then $notenabled(t_i,k) \notin A$ -- from rule $e\ref{e:enabled}$ and supported rule proposition
\item Then either of $body(e\ref{e:r:ne:ptarc} )$, $body(e\ref{e:ne:iptarc} )$, or $body(e\ref{e:ne:tptarc} )$ must not hold in $A$ -- from rule $e\ref{e:r:ne:ptarc} ,e\ref{e:ne:iptarc} ,e\ref{e:ne:tptarc} $ and forced atom proposition
\item Then $q \not< n_i \equiv q \geq n_i$ in $e\ref{e:r:ne:ptarc}$ for all $\{holds(p,q,k), ptarc(p,t_i,n_i,k)\} \subseteq A$ -- from $e\ref{e:r:ne:ptarc}$, forced atom proposition, and given facts ($holds(p,q,k) \in A, ptarc(p,t_i,n_i,k) \in A$)
\item And $q \not\geq n_i \equiv q < n_i$ in $e\ref{e:ne:iptarc} $ for all $\{ holds(p,q,k), iptarc(p,t_i,n_i,k) \} \subseteq A, n_i=1; q > n_i \equiv q = 0$ -- from $e\ref{e:ne:iptarc} $, forced atom proposition, given facts $(holds(p,q,k) \in A, iptarc(p,t_i,1,k) \in A)$, and $q$ is a positive integer
\item And $q \not< n_i \equiv q \geq n_i$ in $e\ref{e:ne:tptarc} $ for all $\{ holds(p,q,k), tptarc(p,t_i,n_i,k) \} \subseteq A$ -- from $e\ref{e:ne:tptarc} $, forced atom proposition, and given facts
\item Then $(\forall p \in \bullet t_i, M_k(p) \geq W(p,t_i)) \wedge (\forall p \in I(t_i), M_k(p) = 0) \wedge (\forall (p,t_i) \in Q, M_k(p) \geq QW(p,t_i))$ -- from 
\begin{enumerate}
\item $holds(p,q,k) \in A$ represents $q=M_k(p)$ -- inductive assumption, given
\item $ptarc(p,t_i,n_i,k) \in A$ represents $n_i=W(p,t_i)$ -- rule $f\ref{f:r:ptarc}$ construction; or it represents $n_i=M_k(p)$ -- rule $f\ref{f:rptarc}$ construction; the construction of $f\ref{f:rptarc}$ ensures that $notenabled(t,k)$ is never true due to the reset arc 
\item definition~\ref{def:pn:preset} of preset $\bullet t_i$ in $PN$
\item definition~\ref{def:pnriq:enable} of enabled transition in $PN$
\end{enumerate}
\item Then $t_i$ is enabled and can fire in $PN$, as a result it can belong to $T_k$ -- from definition~\ref{def:pnriq:enable} of enabled transition

\end{enumerate}
\item And $consumesmore \notin A$, since $A$ is an answer set of $\Pi^4$ -- from rule $a\ref{a:overc:elim}$ and supported rule proposition
\begin{enumerate}
\item Then $\nexists consumesmore(p,k) \in A$ -- from rule $a\ref{a:overc:gen}$ and supported rule proposition
\item  Then $\nexists \{ holds(p,q,k), tot\_decr(p,q1,k) \} \subseteq A : q1>q$ in $body(e\ref{e:r:ne:ptarc})$ -- from $a\ref{a:overc:place}$ and forced atom proposition
\item Then $\nexists p : (\sum_{t_i \in \{t_0,\dots,t_x\}, p \in \bullet t_i}{W(p,t_i)}+\sum_{t_i \in \{t_0,\dots,t_x\}, p \in R(t_i)}{M_k(p)}) > M_k(p)$ -- from the following
\begin{enumerate}
\item $holds(p,q,k)$ represents $q=M_k(p)$ -- inductive assumption, given
\item $tot\_decr(p,q1,k) \in A$ if $\{ del(p,q1_0,t_0,k), \dots, del(p,q1_x,t_x,k) \} \subseteq A$, where $q1 = q1_0+\dots+q1_x$ -- from $r\ref{r:totdecr}$ and forced atom proposition
\item $del(p,q1_i,t_i,k) \in A$ if $\{ fires(t_i,k), ptarc(p,t_i,q1_i,k) \} \subseteq A$ -- from $r\ref{r:r:del} $ and supported rule proposition
\item $del(p,q1_i,t_i,k)$ either represents removal of $q1_i = W(p,t_i)$ tokens from $p \in \bullet t_i$; or it represents removal of $q1_i = M_k(p)$ tokens from $p \in R(t_i)$-- from rule $r\ref{r:r:del} $, supported rule proposition, and definition~\ref{def:pnriq:texec} of transition execution in $PN$
\end{enumerate}
\item Then $T_k$ does not contain conflicting transitions -- by the definition~\ref{def:pnriq:conflict} of conflicting transitions
\end{enumerate}

\item And for each $enabled(t_j,k) \in A$ and $fires(t_j,k) \notin A$, \\ $could\_not\_have(t_j,k) \in A$, since $A$ is an answer set of $\Pi^4$ - from rule $a\ref{a:maxfire:elim}$ and supported rule proposition
\begin{enumerate}
\item Then $\{ enabled(t_j,k), holds(s,qq,k), ptarc(s,t_j,q,k), \\ tot\_decr(s,qqq,k) \} \subseteq A$, such that $q > qq - qqq$ and $fires(t_j,k) \notin A$ - from rule $a\ref{a:r:maxfire:cnh}$ and supported rule proposition
\item Then for an $s \in \bullet t_j \cup R(t_j)$, $q > M_k(s) - (\sum_{t_i \in T_k, s \in \bullet t_i}{W(s,t_i)} + \sum_{t_i \in T_k, s \in R(t_i)}{M_k(s))}$, where $q=W(s,t_j) \text{~if~} s \in \bullet t_j, \text{~or~} M_k(s) \text{~otherwise}$ - from the following: %
\begin{enumerate}
\item $ptarc(s,t_i,q,k)$ represents $q=W(s,t_i)$ if $(s,t_i) \in E^-$ or $q=M_k(s)$ if $s \in R(t_i)$ -- from rule $f\ref{f:r:ptarc},f\ref{f:rptarc}$ construction
\item $holds(s,qq,k)$ represents $qq=M_k(s)$ -- construction
\item $tot\_decr(s,qqq,k) \in A$ if $\{ del(s,qqq_0,t_0,k), \dots, del(s,qqq_x,t_x,k) \} \\ \subseteq A$ -- from rule $r\ref{r:totdecr}$ construction and supported rule proposition
\item $del(s,qqq_i,t_i,k) \in A$ if $\{ fires(t_i,k), ptarc(s,t_i,qqq_i,k) \} \subseteq A$ -- from rule $r\ref{r:r:del} $ and supported rule proposition
\item $del(s,qqq_i,t_i,k)$ represents $qqq_i = W(s,t_i) : t_i \in T_k, (s,t_i) \in E^-$, or $qqq_i = M_k(t_i) : t_i \in T_k, s \in R(t_i)$ -- from rule $f\ref{f:r:ptarc},f\ref{f:rptarc}$ construction 
\item $tot\_decr(q,qqq,k)$ represents $\sum_{t_i \in T_k, s \in \bullet t_i}{W(s,t_i)} + \\ \sum_{t_i \in T_k, s \in R(t_i)}{M_k(s)}$ -- from (C,D,E) above 
\end{enumerate}

\item Then firing $T_k \cup \{ t_j \}$ would have required more tokens than are present at its source place $s \in \bullet t_j \cup R(t_j)$. Thus, $T_k$ is a maximal set of transitions that can simultaneously fire.
\end{enumerate}

\item And for each reset transition $t_r$ with $enabled(t_r,k) \in A$, $fires(t_r,k) \in A$, since $A$ is an answer set of $\Pi^2$ - from rule $f\ref{f:c:rptarc:elim}$ and supported rule proposition
\begin{enumerate}
\item Then the firing set $T_k$ satisfies the reset transition requirement of definition~\ref{def:pnriq:firing_set} (firing set)
\end{enumerate}

\item Then $\{t_0, \dots, t_x\} = T_k$ -- using 1(a),1(b), 1(d) above; and using 1(c) it is a maximal firing set  
\end{enumerate}

\item We show that $M_{k+1}$ is produced by firing $T_k$ in $M_k$. Let $holds(p,q,k+1) \in A$
\begin{enumerate}
\item Then $\{ holds(p,q1,k), tot\_incr(p,q2,k), tot\_decr(p,q3,k) \} \subseteq A : q=q1+q2-q3$ -- from rule $r\ref{r:nextstate}$ and supported rule proposition \label{query:x:1:k}
\item \label{query:x:2:k} Then $holds(p,q1,k) \in A$ represents $q1=M_k(p)$ -- construction, inductive assumption; 
and $\{add(p,q2_0,t_0,k), \dots, add(p,q2_j,t_j,k)\} \subseteq A : q2_0 + \dots + q2_j = q2$  %
 and $\{del(p,q3_0,t_0,k), \dots, del(p,q3_l,t_l,k)\} \subseteq A : q3_0 + \dots + q3_l = q3$ %
  -- rules $r\ref{r:totincr},r\ref{r:totdecr}$ and supported rule proposition, respectively
\item Then $\{ fires(t_0,k), \dots, fires(t_j,k) \} \subseteq A$ and $\{ fires(t_0,k), \dots, fires(t_l,k) \} \\ \subseteq A$ -- rules $r\ref{r:r:add},r\ref{r:r:del}$ and supported rule proposition, respectively
\item Then $\{ fires(t_0,k), \dots, $ $fires(t_j,k) \} \cup \{ fires(t_0,k), \dots, $ $fires(t_l,k) \} \subseteq A = \{ fires(t_0,k), \dots, $ $fires(t_x,k) \} \subseteq A$ -- set union of subsets
\item Then for each $fires(t_x,k) \in A$ we have $t_x \in T_k$ -- already shown in item~\ref{prove:query:fires_tk} above
\item Then $q = M_k(p) + \sum_{t_x \in T_0 \wedge p \in t_x \bullet}{W(t_x,p)} - (\sum_{t_x \in T_k \wedge p \in \bullet t_x}{W(p,t_x)} + \\ \sum_{t_x \in T_k \wedge p \in R(t_x)}{M_k(p)})$ -- from 
\eqref{query:x:2:k} above and the following
\begin{enumerate}
\item Each $add(p,q_j,t_j,k) \in A$ represents $q_j=W(t_j,p)$ for $p \in t_j \bullet$ -- rule $r\ref{r:r:add} $ encoding, and definition~\ref{def:pnriq:texec} of transition execution in $PN$ %
\item Each $del(p,t_y,q_y,k) \in A$ represents either $q_y=W(p,t_y)$ for $p \in \bullet t_y$, or $q_y=M_k(p)$ for $p \in R(t_y)$ -- from rule $r\ref{r:r:del} ,f\ref{f:r:ptarc} $ encoding and definition~\ref{def:pnriq:texec} of transition execution in $PN$; or from rule $r\ref{r:r:del} ,f\ref{f:rptarc}$ encoding and definition of reset arc in $PN$
\item Each $tot\_incr(p,q2,k) \in A$ represents $q2=\sum_{t_x \in T_k \wedge p \in t_x  \bullet}{W(t_x,p)}$ -- aggregate assignment atom semantics in rule $r\ref{r:totincr}$
\item Each $tot\_decr(p,q3,0) \in A$ represents $q3=\sum_{t_x \in T_k \wedge p \in \bullet t_x}{W(p,t_x)} + \sum_{t_x \in T_k \wedge p \in R(t_x)}{M_0(p)}$ -- aggregate assignment atom semantics in rule $r\ref{r:totdecr}$
\end{enumerate}
\item Then, $M_{k+1}(p) = q$ -- since $holds(p,q,k+1) \in A$ encodes $q=M_{k+1}(p)$ -- from construction 
\end{enumerate}
\end{enumerate}

\noindent
As a result, for any $n > k$, $T_n$ will be a valid firing set for $M_n$ and $M_{n+1}$ will be its target marking. 

\noindent
{\bf Conclusion:} Since both \eqref{prove:x2a:reset} and \eqref{prove:a2x:reset} hold, $X=M_0,T_0,M_1,\dots,M_k,T_{k+1}$ is an execution sequence of $PN(P,T,E,W,R)$ (w.r.t $M_0$) iff there is an answer set $A$ of $\Pi^4(PN,M_0,k,ntok)$ such that \eqref{eqn:reset:fires} and \eqref{eqn:reset:holds} hold.

\section{Proof of Proposition~\ref{prop:pnc}}

Let $PN=(P,T,E,C,W,R,I,Q,QW)$ be a Petri Net, $M_0$ be its initial marking and let $\Pi^5(PN,M_0,k,ntok)$ be the ASP encoding of $PN$ and $M_0$ over a simulation length $k$, with maximum $ntok$ tokens on any place node, as defined in section~\ref{sec:pnc}. Then $X=M_0,T_k,M_1,\dots,M_k,T_k,M_{k+1}$ is an execution sequence of $PN$ (w.r.t. $M_0$) iff there is an answer set $A$ of $\Pi^5(PN,M_0,k,ntok)$ such that: 
\begin{equation}
\{ fires(t,ts) : t \in T_{ts}, 0 \leq ts \leq k\} = \{ fires(t,ts) : fires(t,ts) \in A \} \label{eqn:pnc:fires}
\end{equation}
\begin{equation}
\begin{split}
\{ holds(p,q,c,ts) &: p \in P, c/q = M_{ts}(p), 0 \leq ts \leq k+1 \} \\
&= \{ holds(p,q,c,ts) : holds(p,q,c,ts) \in A \} \label{eqn:pnc:holds}
\end{split}
\end{equation}

We prove this by showing that:
\begin{enumerate}[(I)]
\item Given an execution sequence $X$, we create a set $A$ such that it satisfies \eqref{eqn:pnc:fires} and \eqref{eqn:pnc:holds} and show that $A$ is an answer set of $\Pi^5$ \label{prove:x2a:pnc}
\item Given an answer set $A$ of $\Pi^5$, we create an execution sequence $X$ such that \eqref{eqn:pnc:fires} and \eqref{eqn:pnc:holds} are satisfied. \label{prove:a2x:pnc}
\end{enumerate}

\noindent
{\bf First we show (\ref{prove:x2a:pnc})}: Given $PN$ and an execution sequence $X$ of $PN$, we create a set $A$ as a union of the following sets:
\begin{enumerate}
\item $A_1=\{ num(n) : 0 \leq n \leq ntok \}$ %
\item $A_2=\{ time(ts) : 0 \leq ts \leq k\}$ %
\item $A_3=\{ place(p) : p \in P \}$ %
\item $A_4=\{ trans(t) : t \in T \}$ %
\item $A_5=\{ color(c) : c \in C \}$
\item $A_6=\{ ptarc(p,t,n_c,c,ts) : (p,t) \in E^-, c \in C, n_c=m_{W(p,t)}(c), n_c > 0, 0 \leq ts \leq k \}$, where $E^- \subseteq E$ %
\item $A_7=\{ tparc(t,p,n_c,c,ts) : (t,p) \in E^+, c \in C, n_c=m_{W(t,p)}(c), n_c > 0, 0 \leq ts \leq k \}$, where $E^+ \subseteq E$ %
\item $A_8=\{ holds(p,q_c,c,0) : p \in P, c \in C, q_c=m_{M_{0}(p)}(c) \}$ %
\item $A_9=\{ ptarc(p,t,n_c,c,ts) : p \in R(t), c \in C, n_c = m_{M_{ts}(p)}, n_c > 0, 0 \leq ts \leq k \}$
\item $A_{10}=\{ iptarc(p,t,1,c,ts) : p \in I(t), c \in C, 0 \leq ts < k \}$
\item $A_{11}=\{ tptarc(p,t,n_c,c,ts) : (p,t) \in Q, c \in C, n_c=m_{QW(p,t)}(c), n_c > 0, 0 \leq ts \leq k \}$
\item $A_{12}=\{ notenabled(t,ts) : t \in T, 0 \leq ts \leq k, \exists c \in C, (\exists p \in \bullet t, m_{M_{ts}(p)}(c) < m_{W(p,t)}(c)) \vee (\exists p \in I(t), m_{M_{ts}(p)}(c) > 0) \vee (\exists (p,t) \in Q, m_{M_{ts}(p)}(c) < m_{QW(p,t)}(c)) \}$ \newline per definition~\ref{def:pnc:enable} (enabled transition) %
\item $A_{13}=\{ enabled(t,ts) : t \in T, 0 \leq ts \leq k, \forall c \in C, (\forall p \in \bullet t, m_{W(p,t)}(c) \leq m_{M_{ts}(p)}(c)) \wedge (\forall p \in I(t), m_{M_{ts}(p)}(c) = 0) \wedge (\forall (p,t) \in Q, m_{M_{ts}(p)}(c) \geq m_{QW(p,t)}(c)) \}$ \newline per definition~\ref{def:pnc:enable} (enabled transition) %
\item $A_{14}=\{ fires(t,ts) : t \in T_{ts}, 0 \leq ts \leq k \}$ \newline per definition~\ref{def:pnc:firing_set} (firing set), only an enabled transition may fire\label{builda:fires}
\item $A_{15}=\{ add(p,q_c,t,c,ts) : t \in T_{ts}, p \in t \bullet, c \in C, q_c=m_{W(t,p)}(c), 0 \leq ts \leq k \}$ \newline per definition~\ref{def:pnc:texec} (transition execution) %
\item $A_{16}=\{ del(p,q_c,t,c,ts) : t \in T_{ts}, p \in \bullet t, c \in C, q_c=m_{W(p,t)}(c), 0 \leq ts \leq k \} \cup \{ del(p,q_c,t,c,ts) : t \in T_{ts}, p \in R(t), c \in C, q_c=m_{M_{ts}(p)}(c), 0 \leq ts \leq k \}$ \newline per definition~\ref{def:pnc:texec} (transition execution) %
\item $A_{17}=\{ tot\_incr(p,q_c,c,ts) : p \in P, c \in C, q_c=\sum_{t \in T_{ts}, p \in t \bullet}{m_{W(t,p)}(c)}, 0 \leq ts \leq k \}$ \newline per definition~\ref{def:pnc:exec} (firing set execution) %
\item $A_{18}=\{ tot\_decr(p,q_c,c,ts): p \in P, c \in C, q_c=\sum_{t \in T_{ts}, p \in \bullet t}{m_{W(p,t)}(c)}+ \\ \sum_{t \in T_{ts}, p \in R(t) }{m_{M_{ts}(p)}(c)}, 0 \leq ts \leq k \}$ \newline per definition~\ref{def:pnc:exec} (firing set execution) %
\item $A_{19}=\{ consumesmore(p,ts) : p \in P, c \in C, q_c=m_{M_{ts}(p)}(c), \\ q1_c=\sum_{t \in T_{ts}, p \in \bullet t}{m_{W(p,t)}(c)} + \sum_{t \in T_{ts}, p \in R(t)}{m_{M_{ts}(p)}(c)}, q1 > q, 0 \leq ts \leq k \}$ \newline per definition~\ref{def:pnc:conflict} (conflicting transitions) %
\item $A_{20}=\{ consumesmore : \exists p \in P, c \in C, q_c=m_{M_{ts}(p)}(c), \\q1_c=\sum_{t \in T_{ts}, p \in \bullet t}{m_{W(p,t)}(c)} + \sum_{t \in T_{ts}, p \in R(t)}(m_{M_{ts}(p)}(c)), q1 > q, 0 \leq ts \leq k \}$ \newline per definition~\ref{def:pnc:conflict} (conflicting transitions) %
per the maximal firing set semantics
\item $A_{21}=\{ holds(p,q_c,c,ts+1) : p \in P, c \in C, q_c=m_{M_{ts+1}(p)}(c), 0 \leq ts < k\}$, \newline where $M_{ts+1}(p) = M_{ts}(p) - (\sum_{\substack{t \in T_{ts}, p \in \bullet t}}{W(p,t)} + \\ \sum_{t \in T_{ts}, p \in R(t)}M_{ts}(p)) + \sum_{\substack{t \in T_{ts}, p \in t \bullet}}{W(t,p)}$ \newline according to definition~\ref{def:pnc:exec} (firing set execution) %
\end{enumerate}

\noindent
{\bf We show that $A$ satisfies \eqref{eqn:pnc:fires} and \eqref{eqn:pnc:holds}, and $A$ is an answer set of $\Pi^5$.}

$A$ satisfies \eqref{eqn:pnc:fires} and \eqref{eqn:pnc:holds} by its construction above. We show $A$ is an answer set of $\Pi^5$ by splitting. We split $lit(\Pi^5)$ into a sequence of $7k+9$ sets:

\begin{itemize}\renewcommand{\labelitemi}{$\bullet$}
\item $U_0= head(f\ref{f:c:place}) \cup head(f\ref{f:c:trans}) \cup head(f\ref{f:c:col}) \cup head(f\ref{f:c:time}) \cup head(f\ref{f:c:num}) \cup head(i\ref{i:c:init}) = \{place(p) : p \in P\} \cup \{ trans(t) : t \in T\} \cup \{ col(c) : c \in C \} \cup \{ time(0), \dots, time(k)\} \cup \{num(0), \dots, num(ntok)\} \cup \{ holds(p,q_c,c,0) : p \in P, c \in C, q_c=m_{M_0(p)}(c) \} $
\item $U_{7k+1}=U_{7k+0} \cup head(f\ref{f:c:ptarc})^{ts=k} \cup head(f\ref{f:c:tparc})^{ts=k} \cup head(f\ref{f:c:rptarc})^{ts=k} \cup head(f\ref{f:c:iptarc})^{ts=k} \cup head(f\ref{f:c:tptarc})^{ts=k} = U_{7k+0} \cup \{ ptarc(p,t,n_c,c,k) : (p,t) \in E^-, c \in C, n_c=m_{W(p,t)}(c) \} \cup \{  tparc(t,p,n_c,c,k) : (t,p) \in E^+, c \in C, n_c=m_{W(t,p)}(c) \} \cup \{ ptarc(p,t,n_c,c,k) : p \in R(t), c \in C, n_c=m_{M_{k}(p)}(c), n > 0 \} \cup \{ iptarc(p,t,1,c,k) : p \in I(t), c \in C \} \cup \{ tptarc(p,t,n_c,c,k) : (p,t) \in Q, c \in C, n_c=m_{QW(p,t)}(c) \}$
\item $U_{7k+2}=U_{7k+1} \cup head(e\ref{e:c:ne:ptarc})^{ts=k} \cup head(e\ref{e:c:ne:iptarc})^{ts=k} \cup head(e\ref{e:c:ne:tptarc})^{ts=k} = U_{7k+1} \cup \\ \{ notenabled(t,k) : t \in T \}$
\item $U_{7k+3}=U_{7k+2} \cup head(e\ref{e:c:enabled})^{ts=k} = U_{7k+2} \cup \{ enabled(t,k) : t \in T \}$
\item $U_{7k+4}=U_{7k+3} \cup head(a\ref{a:c:fires})^{ts=k} = U_{7k+3} \cup \{ fires(t,k) : t \in T \}$
\item $U_{7k+5}=U_{7k+4}  \cup head(r\ref{r:c:add})^{ts=k} \cup head(r\ref{r:c:del})^{ts=k} = U_{7k+4} \cup \{ add(p,q_c,t,c,k) : p \in P, t \in T, c \in C, q_c=m_{W(t,p)}(c) \} \cup \{ del(p,q_c,t,c,k) : p \in P, t \in T, c \in C, q_c=m_{W(p,t)}(c) \} \cup \{ del(p,q_c,t,c,k) : p \in P, t \in T, c \in C, q_c=m_{M_{k}(p)}(c) \}$
\item $U_{7k+6}=U_{7k+5} \cup head(r\ref{r:c:totincr})^{ts=k} \cup head(r\ref{r:c:totdecr})^{ts=k} = U_{7k+5} \cup \{ tot\_incr(p,q_c,c,k) : p \in P, c \in C, 0 \leq q_c \leq ntok \} \cup \{ tot\_decr(p,q_c,c,k) : p \in P, c \in C, 0 \leq q_c \leq ntok \}$
\item $U_{7k+7}=U_{7k+6} \cup head(a\ref{a:c:overc:place})^{ts=k} \cup head(r\ref{r:c:nextstate})^{ts=k} = U_{7k+6} \cup \{ consumesmore(p,k) : p \in P\} \cup \{ holds(p,q,k+1) : p \in P, 0 \leq q \leq ntok \} $
\item $U_{7k+8}=U_{7k+7} \cup head(a\ref{a:c:overc:gen}) = U_{7k+7} \cup \{ consumesmore \}$
\end{itemize}
where $head(r_i)^{ts=k}$ are head atoms of ground rule $r_i$ in which $ts=k$. We write $A_i^{ts=k} = \{ a(\dots,ts) : a(\dots,ts) \in A_i, ts=k \}$ as short hand for all atoms in $A_i$ with $ts=k$. $U_{\alpha}, 0 \leq \alpha \leq 7k+8$ form a splitting sequence, since each $U_i$ is a splitting set of $\Pi^5$, and $\langle U_{\alpha}\rangle_{\alpha < \mu}$ is a monotone continuous sequence, where $U_0 \subseteq U_1 \dots \subseteq U_{7k+8}$ and $\bigcup_{\alpha < \mu}{U_{\alpha}} = lit(\Pi^5)$. 

We compute the answer set of $\Pi^5$ using the splitting sets as follows:
\begin{enumerate}
\item $bot_{U_0}(\Pi^5) = f\ref{f:place} \cup f\ref{f:trans} \cup f\ref{f:c:ptarc} \cup f9 \cup f\ref{f:c:tptarc} \cup i\ref{i:c:init}$ and $X_0 = A_1 \cup \dots \cup A_5 \cup A_8$ ($= U_0$) is its answer set -- using forced atom proposition

\item $eval_{U_0}(bot_{U_1}(\Pi^5) \setminus bot_{U_0}(\Pi^5), X_0) = \{ ptarc(p,t,q_c,c,0) \text{:-}. |  c \in C, q_c=m_{W(p,t)}(c), q_c > 0 \} \cup \{ tparc(t,p,q_c,c,0) \text{:-}. | c \in C, q_c=m_{W(t,p)}(c), q_c > 0 \} \cup \{ ptarc(p,t,q_c,c,0) \text{:-}. | c \in C, q_c=m_{M_0(p)}(c), q_c > 0 \} \cup \{ iptarc(p,t,1,c,0) \text{:-}. | c \in C \} \cup \{ tptarc(p,t,q_c,c,0) \text{:-}. | c \in C, q_c = m_{QW(p,t)}(c), q_c > 0 \} $. Its answer set $X_1=A_6^{ts=0} \cup A_7^{ts=0} \cup A_9^{ts=0} \cup A_{10}^{ts=0} \cup A_{11}^{ts=0}$ -- using forced atom proposition and construction of $A_6, A_7, A_9, A_{10}, A_{11}$.

\item $eval_{U_1}(bot_{U_2}(\Pi^5) \setminus bot_{U_1}(\Pi^5), X_0 \cup X_1) = \{ notenabled(t,0) \text{:-} . | (\{ trans(t), \\ ptarc(p,t,n_c,c,0), holds(p,q_c,c,0) \} \subseteq X_0 \cup X_1, \text{~where~}  q_c < n_c) \text{~or~} \\ (\{ notenabled(t,0) \text{:-} . | (\{ trans(t), iptarc(p,t,n2_c,c,0), holds(p,q_c,c,0) \} \subseteq X_0 \cup X_1, \\ \text{~where~}  q_c \geq n2_c \}) \text{~or~} $ $(\{ trans(t), tptarc(p,t,n3_c,c,0), holds(p,q_c,c,0) \} \subseteq X_0 \cup X_1, \text{~where~} q_c < n3_c) \}$. Its answer set $X_2=A_{12}^{ts=0}$ -- using  forced atom proposition and construction of $A_{12}$.
\begin{enumerate}
\item where, $q_c=m_{M_0(p)}(c)$, and $n_c=m_{W(p,t)}(c)$ for an arc $(p,t) \in E^-$ -- by construction of $i\ref{i:c:init}$ and $f\ref{f:c:ptarc}$ in $\Pi^5$, and 
\item in an arc $(p,t) \in E^-$, $p \in \bullet t$ (by definition~\ref{def:pnc:preset} of preset)
\item $n2_c=1$ -- by construction of $iptarc$ predicates in $\Pi^5$, meaning $q_c \geq n2_c \equiv q_c \geq 1 \equiv q_c > 0$,
\item $tptarc(p,t,n3_c,c,0)$ represents $n3_c=m_{QW(p,t)}(c)$, where $(p,t) \in Q$ 
\item thus, $notenabled(t,0) \in X_1$ represents $\exists c \in C, (\exists p \in \bullet t : m_{M_0(p)}(c) < m_{W(p,t)}(c)) \vee (\exists p \in I(t) : m_{M_0(p)}(c) > 0) \vee (\exists (p,t) \in Q : m_{M_0(p)}(c) < m_{QW(p,t)}(c))$.
\end{enumerate}

\item $eval_{U_2}(bot_{U_3}(\Pi^5) \setminus bot_{U_2}(\Pi^5), X_0 \cup \dots \cup X_2) = \{ enabled(t,0) \text{:-}. | trans(t) \in X_0 \cup \dots \cup X_2, notenabled(t,0) \notin X_0 \cup \dots \cup X_2 \}$. Its answer set is $X_3 = A_{13}^{ts=0}$ -- using forced atom proposition and construction of $A_{13}$.
\begin{enumerate}
\item where, an $enabled(t,0) \in X_3$ if $\nexists ~notenabled(t,0) \in X_0 \cup \dots \cup X_2$; which is equivalent to $\forall c \in C, (\nexists p \in \bullet t : m_{M_0(p)}(c) < m_{W(p,t)}(c)) \wedge (\nexists p \in I(t) : m_{M_0(p)}(c) > 0) \wedge (\nexists (p,t) \in Q : m_{M_0(p)}(c) < m_{QW(p,t)}(c) ) \equiv \forall c \in C, (\forall p \in \bullet t: m_{M_0(p)}(c) \geq m_{W(p,t)}(c)) \wedge (\forall p \in I(t) : m_{M_0(p)}(c) = 0)$.
\end{enumerate}

\item $eval_{U_3}(bot_{U_4}(\Pi^5) \setminus bot_{U_3}(\Pi^5), X_0 \cup \dots \cup X_3) = \{\{fires(t,0)\} \text{:-}. | enabled(t,0) \\ \text{~holds in~} X_0 \cup \dots \cup X_3 \}$. It has multiple answer sets $X_{4.1}, \dots, X_{4.n}$, corresponding to elements of power set of $fires(t,0)$ atoms in $eval_{U_3}(...)$ -- using supported rule proposition. Since we are showing that the union of answer sets of $\Pi^5$ determined using splitting is equal to $A$, we only consider the set that matches the $fires(t,0)$ elements in $A$ and call it $X_4$, ignoring the rest. Thus, $X_4 = A_{14}^{ts=0}$, representing $T_0$.
\begin{enumerate}
\item in addition, for every $t$ such that $enabled(t,0) \in X_0 \cup \dots \cup X_3,  R(t) \neq \emptyset$; $fires(t,0) \in X_4$ -- per definition~\ref{def:pnc:firing_set} (firing set); requiring that a reset transition is fired when enabled
\item thus, the firing set $T_0$ will not be eliminated by the constraint $f\ref{f:c:rptarc:elim}$ 
\end{enumerate}

\item $eval_{U_4}(bot_{U_5}(\Pi^5) \setminus bot_{U_4}(\Pi^5), X_0 \cup \dots \cup X_4) = \{add(p,n_c,t,c,0) \text{:-}. | \{fires(t,0), \\ tparc(t,p,n_c,c,0) \} \subseteq X_0 \cup \dots \cup X_4 \} \cup \{ del(p,n_c,t,c,0) \text{:-}. | \{ fires(t,0), \\ ptarc(p,t,n_c,c,0) \} \subseteq X_0 \cup \dots \cup X_4 \}$. It's answer set is $X_5 = A_{15}^{ts=0} \cup A_{16}^{ts=0}$ -- using forced atom proposition and definitions of $A_{15}$ and $A_{16}$. 
\begin{enumerate}
\item where, each $add$ atom is equivalent to $n_c=m_{W(t,p)}(c),c \in C, p \in t \bullet$, 
\item and each $del$ atom is equivalent to $n_c=m_{W(p,t)}(c), c \in C, p \in \bullet t$; or $n_c=m_{M_0(p)}(c), c \in C, p \in R(t)$,
\item representing the effect of transitions in $T_0$
\end{enumerate}

\item $eval_{U_5}(bot_{U_6}(\Pi^5) \setminus bot_{U_5}(\Pi^5), X_0 \cup \dots \cup X_5) = \{tot\_incr(p,qq_c,c,0) \text{:-}. | qq_c=\sum_{add(p,q_c,t,c,0) \in X_0 \cup \dots \cup X_5}{q_c} \} \cup \{ tot\_decr(p,qq_c,c,0) \text{:-}. | qq_c= \\ \sum_{del(p,q_c,t,c,0) \in X_0 \cup \dots \cup X_5}{q_c} \}$. It's answer set is $X_6 = A_{17}^{ts=0} \cup A_{18}^{ts=0}$ --  using forced atom proposition and definitions of $A_{17}$ and $A_{18}$.
\begin{enumerate}
\item where, each $tot\_incr(p,qq_c,c,0)$, $qq_c=\sum_{add(p,q_c,t,c,0) \in X_0 \cup \dots X_5}{q_c}$ \\ $\equiv qq_c=\sum_{t \in X_4, p \in t \bullet}{m_{W(p,t)}(c)}$, 
\item and each $tot\_decr(p,qq_c,c,0)$, $qq_c=\sum_{del(p,q_c,t,c,0) \in X_0 \cup \dots X_5}{q_c}$ \\ $\equiv qq=\sum_{t \in X_4, p \in \bullet t}{m_{W(t,p)}(c)} + \sum_{t \in X_4, p \in R(t)}{m_{M_0(p)}(c)}$, 
\item represent the net effect of transitions in $T_0$
\end{enumerate}
\item $eval_{U_6}(bot_{U_7}(\Pi^5) \setminus bot_{U_6}(\Pi^5), X_0 \cup \dots \cup X_6) = \{ consumesmore(p,0) \text{:-}. | \\ \{holds(p,q_c,c,0), tot\_decr(p,q1_c,c,0) \} \subseteq X_0 \cup \dots \cup X_6, q1_c > q_c \} \cup \\ \{ holds(p,q_c,c,1) \text{:-}., |  \{ holds(p,q1_c,c,0), $ $tot\_incr(p,q2_c,c,0), $ $tot\_decr(p,q3_c,c,0) \} \\ \subseteq X_0 \cup \dots \cup X_6, q_c=q1_c+q2_c-q3_c \} $. It's answer set is $X_7 = A_{19}^{ts=0} \cup A_{21}^{ts=0}$  -- using forced atom proposition and definitions of $A_{19}, A_{21}$.
\begin{enumerate}
\item where, $consumesmore(p,0)$ represents $\exists p : q_c=m_{M_0(p)}(c), q1_c= \\ \sum_{t \in T_0, p \in \bullet t}{m_{W(p,t)}(c)}+\sum_{t \in T_0, p \in R(t)}{m_{M_0(p)}(c)}, q1_c > q_c, c \in C$, indicating place $p$ will be over consumed if $T_0$ is fired, as defined in definition~\ref{def:pnc:conflict} (conflicting transitions),
\item $holds(p,q_c,c,1)$ represents $q_c=m_{M_1(p)}(c)$ -- by construction of $\Pi^5$
\end{enumerate}

\[ \vdots \]

\item $eval_{U_{7k+0}}(bot_{U_{7k+1}}(\Pi^5) \setminus bot_{U_{7k+0}}(\Pi^5), X_0 \cup \dots \cup X_{7k+0}) = \{ ptarc(p,t,q_c,c,k) \text{:-}. | c \in C, q_c=m{W(p,t)}(c), q_c > 0 \} \cup \{ tparc(t,p,q_c,c,k) \text{:-}. | c \in C, q_c=m_{W(t,p)}(c), q_c > 0 \} \cup $ $\{ ptarc(p,t,q_c,c,k) \text{:-}. | $ $c \in C, q_c=m_{M_k(p)}(c), q_c > 0 \} \cup \{ iptarc(p,t,1,c,k) \text{:-}. | c \in C \} \cup \{ tptarc(p,t,q_c,c,k) \text{:-}. | c \in C, q_c = m_{QW(p,t)}(c), q_c > 0 \} $. Its answer set $X_{7k+1}=A_6^{ts=k} \cup A_7^{ts=k} \cup A_9^{ts=k} \cup A_{10}^{ts=k} \cup A_{11}^{ts=k}$ -- using forced atom proposition and construction of $A_6, A_7, A_9, A_{10}, A_{11}$.

\item $eval_{U_{7k+1}}(bot_{U_{7k+2}}(\Pi^5) \setminus bot_{U_{7k+1}}(\Pi^5), X_0 \cup \dots \cup X_{7k+1}) = \{ notenabled(t,k) \text{:-} . | \\ (\{ trans(t), ptarc(p,t,n_c,c,k), holds(p,q_c,c,k) \} \subseteq X_0 \cup \dots \cup X_{7k+1}, \text{~where~}  q_c < n_c) \text{~or~} $ $ (\{ notenabled(t,k) \text{:-} . |  (\{ trans(t), iptarc(p,t,n2_c,c,k),  holds(p,q_c,c,k) \} \subseteq X_0 \cup \dots \cup X_{7k+1}, \text{~where~}  q_c \geq n2_c \}) \text{~or~} (\{ trans(t), tptarc(p,t,n3_c,c,k), \\ holds(p,q_c,c,k) \} \subseteq X_0 \cup \dots \cup X_{7k+1}, \text{~where~} q_c < n3_c) \}$. Its answer set $X_{7k+2}=A_{12}^{ts=k}$ -- using  forced atom proposition and construction of $A_{12}$.
\begin{enumerate}
\item since $q_c=m_{M_k(p)}(c)$, and $n_c=m_{W(p,t)}(c)$ for an arc $(p,t) \in E^-$ -- by construction of $holds$ and $ptarc$ predicates in $\Pi^5$, and 
\item in an arc $(p,t) \in E^-$, $p \in \bullet t$ (by definition~\ref{def:pnc:preset} of preset)
\item $n2_c=1$ -- by construction of $iptarc$ predicates in $\Pi^5$, meaning $q_c \geq n2_c \equiv q_c \geq 1 \equiv q_c > 0$,
\item $tptarc(p,t,n3_c,c,k)$ represents $n3_c=m_{QW(p,t)}(c)$, where $(p,t) \in Q$ 
\item thus, $notenabled(t,k) \in X_{8k+1}$ represents $\exists c \in C (\exists p \in \bullet t : m_{M_k(p)}(c) < m_{W(p,t)}(c)) \vee (\exists p \in I(t) : m_{M_k(p)}(c) > 0) \vee (\exists (p,t) \in Q : m_{M_{ts}(p)}(c) < m_{QW(p,t)}(c))$.
\end{enumerate}

\item $eval_{U_{7k+2}}(bot_{U_{7k+3}}(\Pi^5) \setminus bot_{U_{7k+2}}(\Pi^5), X_0 \cup \dots \cup X_{7k+2}) = \{ enabled(t,k) \text{:-}. | trans(t) \in X_0 \cup \dots \cup X_{7k+2} \wedge notenabled(t,k) \notin X_0 \cup \dots \cup X_{7k+2} \}$. Its answer set is $X_{7k+3} = A_{13}^{ts=k}$ -- using forced atom proposition and construction of $A_{13}$.
\begin{enumerate}
\item since an $enabled(t,k) \in X_{7k+3}$ if $\nexists ~notenabled(t,k) \in X_0 \cup \dots \cup X_{7k+2}$; which is equivalent to $\forall c \in C, (\nexists p \in \bullet t : m_{M_k(p)}(c) < m_{W(p,t)}(c)) \wedge (\nexists p \in I(t) : m_{M_k(p)}(c) > 0) \wedge (\nexists (p,t) \in Q : m_{M_k(p)}(c) < m_{QW(p,t)}(c) ) \equiv \forall c \in C, (\forall p \in \bullet t: m_{M_k(p)}(c) \geq m_{W(p,t)}(c)) \wedge (\forall p \in I(t) : m_{M_k(p)}(c) = 0)$.
\end{enumerate}

\item $eval_{U_{7k+3}}(bot_{U_{7k+4}}(\Pi^5) \setminus bot_{U_{7k+3}}(\Pi^5), X_0 \cup \dots \cup X_{7k+3}) = \{\{fires(t,k)\} \text{:-}. | \\ enabled(t,k) \text{~holds in~} X_0 \cup \dots \cup X_{7k+3} \}$. It has multiple answer sets \\ $X_{7k+}{4.1}, \dots, X_{7k+}{4.n}$, corresponding to elements of power set of $fires(t,k)$ atoms in $eval_{U_3}(...)$ -- using supported rule proposition. Since we are showing that the union of answer sets of $\Pi^5$ determined using splitting is equal to $A$, we only consider the set that matches the $fires(t,k)$ elements in $A$ and call it $X_{7k+4}$, ignoring the rest. Thus, $X_{7k+4} = A_{14}^{ts=k}$, representing $T_k$.
\begin{enumerate}
\item in addition, for every $t$ such that $enabled(t,k) \in X_0 \cup \dots \cup X_{7k+3},  R(t) \neq \emptyset$; $fires(t,k) \in X_{7k+4}$ -- per definition~\ref{def:pnc:firing_set} (firing set); requiring that a reset transition is fired when enabled
\item thus, the firing set $T_k$ will not be eliminated by the constraint $f\ref{f:c:rptarc:elim}$ 
\end{enumerate}

\item $eval_{U_{7k+4}}(bot_{U_{7k+5}}(\Pi^5) \setminus bot_{U_{7k+4}}(\Pi^5), X_0 \cup \dots \cup X_{7k+4}) = \{add(p,n_c,t,c,k) \text{:-}. | $ $ \{fires(t,k), \\ tparc(t,p,n_c,c,k) \} \subseteq $ $X_0 \cup \dots \cup X_{7k+4} \} \cup $ $\{ del(p,n_c,t,c,k) \text{:-}. | $ $\{ fires(t,k), \\ ptarc(p,t,n_c,c,k) \} \subseteq $ $X_0 \cup \dots \cup X_{7k+4} \}$. It's answer set is $X_{7k+5} = A_{15}^{ts=k} \cup A_{16}^{ts=k}$ -- using forced atom proposition and definitions of $A_{15}$ and $A_{16}$. 
\begin{enumerate}
\item where each $add$ atom is equivalent to $n_c=m_{W(t,p)}(c),c \in C, p \in t \bullet$,  
\item and each $del$ atom is equivalent to $n_c=m_{W(p,t)}(c), c \in C, p \in \bullet t$; or $n_c=m_{M_k(p)}(c), c \in C, p \in R(t)$,
\item representing the effect of transitions in $T_k$
\end{enumerate}

\item $eval_{U_{7k+5}}(bot_{U_{7k+6}}(\Pi^5) \setminus bot_{U_{7k+5}}(\Pi^5), X_0 \cup \dots \cup X_{7k+5}) = \{tot\_incr(p,qq_c,c,k) \text{:-}. | $ $qq=\sum_{add(p,q_c,t,c,k) \in X_0 \cup \dots \cup X_{7k+5}}{q_c} \} \cup $ $ \{ tot\_decr(p,qq_c,c,k) \text{:-}. | \\ qq_c=\sum_{del(p,q_c,t,c,k) \in X_0 \cup \dots \cup X_{7k+5}}{q_c} \}$. It's answer set is $X_{7k+6} = A_{17}^{ts=k} \cup A_{18}^{ts=k}$ --  using forced atom proposition and definitions of $A_{17}$ and $A_{18}$.
\begin{enumerate}
\item where, each $tot\_incr(p,qq_c,c,k)$, $qq_c=\sum_{add(p,q_c,t,c,k) \in X_0 \cup \dots X_{8k+5}}{q_c}$ \\ $\equiv qq_c=\sum_{t \in X_{7k+4}, p \in t \bullet}{m_{W(p,t)}(c)}$, 
\item and each $tot\_decr(p,qq_c,c,k)$, $qq_c=\sum_{del(p,q_c,t,c,k) \in X_0 \cup \dots X_{8k+5}}{q_c}$ \\ $\equiv qq=\sum_{t \in X_{7k+4}, p \in \bullet t}{m_{W(t,p)}(c)} + \sum_{t \in X_{7k+4}, p \in R(t)}{m_{M_{k}(p)}(c)}$,
\item represent the net effect of transitions in $T_k$
\end{enumerate}
\item $eval_{U_{7k+6}}(bot_{U_{7k+7}}(\Pi^5) \setminus bot_{U_{7k+6}}(\Pi^5), X_0 \cup \dots \cup X_{7k+6}) = \{ consumesmore(p,k) \text{:-}. | \\ \{holds(p,q_c,c,k), tot\_decr(p,q1_c,c,k) \} \subseteq X_0 \cup \dots \cup X_{7k+6}, q1 > q \} \cup \\ \{ holds(p,q_c,c,k+1) \text{:-}. |  \{ holds(p,q1_c,c,k), tot\_incr(p,q2_c,c,k), \\ tot\_decr(p,q3_c,c,k) \} \subseteq $ $X_0 \cup \dots \cup X_{7k+6}, q_c=q1_c+q2_c-q3_c \} $. It's answer set is $X_{7k+7} = A_{19}^{ts=k} \cup A_{21}^{ts=k}$  -- using forced atom proposition and definitions of $A_{19}, A_{21}$.
\begin{enumerate}
\item where, $consumesmore(p,k)$ represents $\exists p : q_c=m_{M_k(p)}(c), q1_c= \\ \sum_{t \in T_k, p \in \bullet t}{m_{W(p,t)}(c)}+\sum_{t \in T_k, p \in R(t)}{m_{M_k(p)}(c)}, q1_c > q_c, c \in C$, indicating place $p$ that will be over consumed if $T_k$ is fired, as defined in definition~\ref{def:pnc:conflict} (conflicting transitions),
\item and $holds(p,q_c,c,k+1)$ represents $q_c=m_{M_{k+1}(p)}(c)$ -- by construction of $\Pi^5$
\end{enumerate}

\item $eval_{U_{7k+7}}(bot_{U_{7k+8}}(\Pi^5) \setminus bot_{U_{7k+7}}(\Pi^5), X_0 \cup \dots \cup X_{7k+7}) = \{ consumesmore \text{:-}. | \\ \{ consumesmore(p,0),\dots,$ $consumesmore(p,k) \} \cap (X_0 \cup \dots \cup X_{7k+7}) \neq \emptyset \}$. It's answer set is $X_{7k+8} = A_{20}^{ts=k}$ -- using forced atom proposition and the definition of $A_{20}$
\begin{enumerate}
\item $X_{7k+7}$ will be empty since none of $consumesmore(p,0),\dots, \\ consumesmore(p,k)$ hold in $X_0 \cup \dots \cup X_{7k+7}$ due to the construction of $A$, encoding of $a\ref{a:c:overc:place}$ and its body atoms. As a result, it is not eliminated by the constraint $a\ref{a:c:overc:elim}$
\end{enumerate}

\end{enumerate}

The set $X = X_0 \cup \dots \cup X_{7k+8}$ is the answer set of $\Pi^5$ by the splitting sequence theorem~\ref{def:split_seq_thm}. Each $X_i, 0 \leq i \leq 7k+8$ matches a distinct portion of $A$, and $X = A$, thus $A$ is an answer set of $\Pi^5$.

\vspace{30pt}
\noindent
{\bf Next we show (\ref{prove:a2x:pnc}):} Given $\Pi^5$ be the encoding of a Petri Net $PN(P,T,E,C,W,R,I,$ $Q,WQ)$ with initial marking $M_0$, and $A$ be an answer set of $\Pi^5$ that satisfies (\ref{eqn:pnc:fires}) and (\ref{eqn:pnc:holds}), then we can construct $X=M_0,T_k,\dots,M_k,T_k,M_{k+1}$ from $A$, such that it is an execution sequence of $PN$.

We construct the $X$ as follows:
\begin{enumerate}
\item $M_i = (M_i(p_0), \dots, M_i(p_n))$, where $\{ holds(p_0,m_{M_i(p_0)}(c),c,i), \dots \\ holds(p_n,m_{M_i(p_n)}(c),c,i) \} \subseteq A$, for $c \in C, 0 \leq i \leq k+1$
\item $T_i = \{ t : fires(t,i) \in A\}$, for $0 \leq i \leq k$ 
\end{enumerate}
and show that $X$ is indeed an execution sequence of $PN$. We show this by induction over $k$ (i.e. given $M_k$, $T_k$ is a valid firing set and its firing produces marking $M_{k+1}$).

\vspace{20pt}

\noindent
{\bf Base case:} Let $k=0$, and $M_0$ is a valid marking in $X$ for $PN$, show
\begin{inparaenum}[(1)]
\item $T_0$ is a valid firing set for $M_0$, and 
\item firing $T_0$ in $M_0$ results in marking $M_1$.
\end{inparaenum} 

\begin{enumerate}
\item We show $T_0$ is a valid firing set for $M_0$. Let $\{ fires(t_0,0), \dots, fires(t_x,0) \}$ be the set of all $fires(\dots,0)$ atoms in $A$,\label{prove:pri:fires_t0}

\begin{enumerate}
\item Then for each $fires(t_i,0) \in A$

\begin{enumerate}
\item $enabled(t_i,0) \in A$ -- from rule $a\ref{a:c:fires}$ and supported rule proposition
\item Then $notenabled(t_i,0) \notin A$ -- from rule $e\ref{e:c:enabled}$ and supported rule proposition
\item Then either of $body(e\ref{e:c:ne:ptarc})$, $body(e\ref{e:c:ne:iptarc})$, or $body(e\ref{e:c:ne:tptarc})$ must not hold in $A$ -- from rules $e\ref{e:c:ne:ptarc},e\ref{e:c:ne:iptarc},e\ref{e:c:ne:tptarc}$ and forced atom proposition
\item Then $q_c \not< {n_i}_c \equiv q_c \geq {n_i}_c$ in $e\ref{e:c:ne:ptarc}$ for all $\{holds(p,q_c,c,0), \\ ptarc(p,t_i,{n_i}_c,c,0)\} \subseteq A$ -- from $e\ref{e:c:ne:ptarc}$, forced atom proposition, and given facts ($holds(p,q_c,c,0) \in A, ptarc(p,t_i,{n_i}_c,0) \in A$)
\item And $q_c \not\geq {n_i}_c \equiv q_c < {n_i}_c$ in $e\ref{e:c:ne:iptarc}$ for all $\{ holds(p,q_c,c,0), \\ iptarc(p,t_i,{n_i}_c,c,0) \} \subseteq A, {n_i}_c=1$; $q_c > {n_i}_c \equiv q_c = 0$ -- from $e\ref{e:c:ne:iptarc}$, forced atom proposition, given facts ($holds(p,q_c,c,0) \in A, iptarc(p,t_i,1,c,0) \in A$), and $q_c$ is a positive integer
\item And $q_c \not< {n_i}_c \equiv q_c \geq {n_i}_c$ in $e\ref{e:c:ne:tptarc}$ for all $\{ holds(p,q_c,c,0),  \\ tptarc(p,t_i,{n_i}_c,c,0) \} \subseteq A$ -- from $e\ref{e:c:ne:tptarc}$, forced atom proposition, and given facts
\item Then $\forall c \in C, (\forall p \in \bullet t_i, m_{M_0(p)}(c) \geq m_{W(p,t_i)}(c)) \wedge (\forall p \in I(t_i), m_{M_0(p)}(c) = 0) \wedge (\forall (p,t_i) \in Q, m_{M_0(p)}(c) \geq m_{QW(p,t_i)}(c))$ -- from the following
\begin{enumerate}
\item $holds(p,q_c,c,0) \in A$ represents $q_c=m_{M_0(p)}(c)$ -- rule $i\ref{i:c:init}$ construction
\item $ptarc(p,t_i,{n_i}_c,0) \in A$ represents ${n_i}_c=m_{W(p,t_i)}(c)$ -- rule $f\ref{f:c:ptarc}$ construction; or it represents ${n_i}_c = m_{M_0(p)}(c)$ -- rule $f\ref{f:c:rptarc}$ construction; the construction of $f\ref{f:c:rptarc}$ ensures that $notenabled(t,0)$ is never true for a reset arc
\item definition~\ref{def:pnc:preset} of preset $\bullet t_i$ in $PN$
\item definition~\ref{def:pnc:enable} of enabled transition in $PN$
\end{enumerate}
\item Then $t_i$ is enabled and can fire in $PN$, as a result it can belong to $T_0$ -- from definition~\ref{def:pnc:enable} of enabled transition

\end{enumerate}
\item And $consumesmore \notin A$, since $A$ is an answer set of $\Pi^5$ -- from rule $a\ref{a:c:overc:gen}$ and supported rule proposition
\begin{enumerate}
\item Then $\nexists consumesmore(p,0) \in A$ -- from rule $a\ref{a:c:overc:place}$ and supported rule proposition
\item  Then $\nexists \{ holds(p,q_c,c,0), tot\_decr(p,q1_c,c,0) \} \subseteq A, q1_c>q_c$ in $body(a\ref{a:c:overc:place})$ -- from $a\ref{a:c:overc:place}$ and forced atom proposition
\item Then $\nexists c \in C \nexists p \in P, (\sum_{t_i \in \{t_0,\dots,t_x\}, p \in \bullet t_i}{m_{W(p,t_i)}(c)}+ \\ \sum_{t_i \in \{t_0,\dots,t_x\}, p \in R(t_i)}{m_{M_0(p)}(c)}) > m_{M_0(p)}(c)$ -- from the following
\begin{enumerate}
\item $holds(p,q_c,c,0)$ represents $q_c=m_{M_0(p)}(c)$ -- from rule $i\ref{i:c:init}$ construction, given
\item $tot\_decr(p,q1_c,c,0) \in A$ if $\{ del(p,{q1_0}_c,t_0,c,0), \dots, \\ del(p,{q1_x}_c,t_x,c,0) \} \subseteq A$, where $q1_c = {q1_0}_c+\dots+{q1_x}_c$ -- from $r\ref{r:c:totdecr}$ and forced atom proposition
\item $del(p,{q1_i}_c,t_i,c,0) \in A$ if $\{ fires(t_i,0), ptarc(p,t_i,{q1_i}_c,c,0) \} \subseteq A$ -- from $r\ref{r:c:del}$ and supported rule proposition
\item $del(p,{q1_i}_c,t_i,c,0)$ either represents removal of ${q1_i}_c = m_{W(p,t_i)}(c)$ tokens from $p \in \bullet t_i$; or it represents removal of ${q1_i}_c = m_{M_0(p)}(c)$ tokens from $p \in R(t_i)$-- from rules $r\ref{r:c:del},f\ref{f:c:ptarc},f\ref{f:c:rptarc}$, supported rule proposition, and definition~\ref{def:pnc:texec} of transition execution in $PN$
\end{enumerate}
\item Then the set of transitions in $T_0$ do not conflict -- by the definition~\ref{def:pnc:conflict} of conflicting transitions
\end{enumerate}

\item And for each reset transition $t_r$ with $enabled(t_r,0) \in A$, $fires(t_r,0) \in A$, since $A$ is an answer set of $\Pi^5$ - from rule $f\ref{f:c:rptarc:elim}$ and supported rule proposition
\begin{enumerate}
\item Then, the firing set $T_0$ satisfies the reset-transition requirement of definition~\ref{def:pnc:firing_set} (firing set)
\end{enumerate}

\item Then $\{t_0, \dots, t_x\} = T_0$ -- using 1(a),1(b),1(d) above; and using 1(c) it is a maximal firing set  

\end{enumerate}

\item We show $M_1$ is produced by firing $T_0$ in $M_0$. Let $holds(p,q_c,c,1) \in A$
\begin{enumerate}
\item Then $\{ holds(p,q1_c,c,0), tot\_incr(p,q2_c,c,0), tot\_decr(p,q3_c,c,0) \} \subseteq A : q_c=q1_c+q2_c-q3_c$ -- from rule $r\ref{r:c:nextstate}$ and supported rule proposition \label{x:1:base}
\item \label{x:2:base} Then, $holds(p,q1_c,c,0) \in A$ represents $q1_c=m_{M_0(p)}(c)$ -- given, rule $i\ref{i:c:init}$ construction;  
and $\{add(p,{q2_0}_c,t_0,c,0), \dots, $ $add(p,{q2_j}_c,t_j,c,0)\} \subseteq A : {q2_0}_c + \dots + {q2_j}_c = q2_c$  \label{stmt:add:base} and $\{del(p,{q3_0}_c,t_0,c,0), \dots, $ $del(p,{q3_l}_c,t_l,c,0)\} \subseteq A : {q3_0}_c + \dots + {q3_l}_c = q3_c$ \label{stmt:del:base}  -- rules $r\ref{r:c:totincr},r\ref{r:c:totdecr}$ and supported rule proposition, respectively
\item Then $\{ fires(t_0,0), \dots, fires(t_j,0) \} \subseteq A$ and $\{ fires(t_0,0), \dots, fires(t_l,0) \} $ $\subseteq A$ -- rules $r\ref{r:c:add},r\ref{r:c:del}$ and supported rule proposition, respectively
\item Then $\{ fires(t_0,0), \dots, $ $fires(t_j,0) \} \cup $ $\{ fires(t_0,0), \dots, $ $fires(t_l,0) \} \subseteq $ $A = \{ fires(t_0,0), \dots, $ $fires(t_x,0) \} \subseteq A$ -- set union of subsets
\item Then for each $fires(t_x,0) \in A$ we have $t_x \in T_0$ -- already shown in item~\ref{prove:pri:fires_t0} above
\item Then $q_c = m_{M_0(p)}(c) + \sum_{t_x \in T_0 \wedge p \in t_x \bullet}{m_{W(t_x,p)}(c)} - (\sum_{t_x \in T_0 \wedge p \in \bullet t_x}{m_{W(p,t_x)}(c)} + \sum_{t_x \in T_0 \wedge p \in R(t_x)}{m_{M_0(p)}(c)})$ -- from 
\eqref{x:2:base} above and the following
\begin{enumerate}
\item Each $add(p,{q_j}_c,t_j,c,0) \in A$ represents ${q_j}_c=m_{W(t_j,p)}(c)$ for $p \in t_j \bullet$ -- rule $r\ref{r:c:add},f\ref{f:c:tparc}$ encoding, and definition~\ref{def:pnc:texec} of transition execution in $PN$ %
\item Each $del(p,t_y,{q_y}_c,c,0) \in A$ represents either ${q_y}_c=m_{W(p,t_y)}(c)$ for $p \in \bullet t_y$, or ${q_y}_c=m_{M_0(p)}(c)$ for $p \in R(t_y)$ -- from rule $r\ref{r:c:del},f\ref{f:c:ptarc}$ encoding and definition~\ref{def:pnc:texec} of transition execution in $PN$; or from rule $r\ref{r:c:del},f\ref{f:c:rptarc}$ encoding and definition of reset arc in $PN$
\item Each $tot\_incr(p,q2_c,c,0) \in A$ represents $q2_c=\sum_{t_x \in T_0 \wedge p \in t_x  \bullet}{m_{W(t_x,p)}(c)}$ -- aggregate assignment atom semantics in rule $r\ref{r:c:totincr}$
\item Each $tot\_decr(p,q3_c,c,0) \in A$ represents $q3_c=\sum_{t_x \in T_0 \wedge p \in \bullet t_x}{m_{W(p,t_x)}(c)} + \sum_{t_x \in T_0 \wedge p \in R(t_x)}{m_{M_0(p)}(c)}$ -- aggregate assignment atom semantics in rule $r\ref{r:c:totdecr}$
\end{enumerate}
\item Then, $m_{M_1(p)}(c) = q_c$ -- since $holds(p,q_c,c,1) \in A$ encodes $q_c=m_{M_1(p)}(c)$ -- from construction
\end{enumerate}
\end{enumerate}

\noindent
{\bf Inductive Step:} Let $k > 0$, and $M_k$ is a valid marking in $X$ for $PN$, show 
\begin{inparaenum}[(1)]
\item $T_k$ is a valid firing set for $M_k$, and 
\item firing $T_k$ in $M_k$ produces marking $M_{k+1}$.
\end{inparaenum}

\begin{enumerate}
\item We show that $T_k$ is a valid firing set in $M_k$. Let $\{ fires(t_0,k), \dots, fires(t_x,k) \}$ be the set of all $fires(\dots,k)$ atoms in $A$,\label{prove:pri:fires_tk}

\begin{enumerate}
\item Then for each $fires(t_i,k) \in A$

\begin{enumerate}
\item $enabled(t_i,k) \in A$ -- from rule $a\ref{a:c:fires}$ and supported rule proposition
\item Then $notenabled(t_i,k) \notin A$ -- from rule $e\ref{e:c:enabled}$ and supported rule proposition
\item Then either of $body(e\ref{e:c:ne:ptarc})$, $body(e\ref{e:c:ne:iptarc})$, or $body(e\ref{e:c:ne:tptarc})$ must not hold in $A$ -- from rules $e\ref{e:c:ne:ptarc},e\ref{e:c:ne:iptarc},e\ref{e:c:ne:tptarc}$ and forced atom proposition
\item Then $q_c \not< {n_i}_c \equiv q_c \geq {n_i}_c$ in $e\ref{e:c:ne:ptarc}$ for all $\{holds(p,q_c,c,k), \\ ptarc(p,t_i,{n_i}_c,c,k)\} \subseteq A$ -- from $e\ref{e:c:ne:ptarc}$, forced atom proposition, given facts, and the inductive assumption 
\item And $q_c \not\geq {n_i}_c \equiv q_c < {n_i}_c$ in $e\ref{e:c:ne:iptarc}$ for all $\{ holds(p,q_c,c,k), \\ iptarc(p,t_i,{n_i}_c,c,k) \} \subseteq A, {n_i}_c=1$; $q_c > {n_i}_c \equiv q_c = 0$ -- from $e\ref{e:c:ne:iptarc}$, forced atom proposition, given facts ($holds(p,q_c,c,k) \in A, iptarc(p,t_i,1,c,k) \in A$), and $q_c$ is a positive integer
\item And $q_c \not< {n_i}_c \equiv q_c \geq {n_i}_c$ in $e\ref{e:c:ne:tptarc}$ for all $\{ holds(p,q_c,c,k), \\ tptarc(p,t_i,{n_i}_c,c,k) \} \subseteq A$ -- from $e\ref{e:c:ne:tptarc}$, forced atom proposition, and inductive assumption
\item Then $(\forall p \in \bullet t_i, m_{M_k(p)}(c) \geq m_{W(p,t_i)}(c)) \wedge (\forall p \in I(t_i), m_{M_k(p)}(c) = 0) \wedge (\forall (p,t_i) \in Q, m_{M_k(p)}(c) \geq m_{QW(p,t_i)}(c))$ -- from 
\begin{enumerate}
\item $holds(p,q_c,c,k) \in A$ represents $q_c=m_{M_k(p)}(c)$ -- construction of $\Pi^5$
\item $ptarc(p,t_i,{n_i}_c,k) \in A$ represents ${n_i}_c=m_{W(p,t_i)}(c)$ -- rule $f\ref{f:c:ptarc}$ construction; or it represents ${n_i}_c = m_{M_k(p)}(c)$ -- rule $f\ref{f:c:rptarc}$ construction; the construction of $f\ref{f:c:rptarc}$ ensures that $notenabled(t,k)$ is never true for a reset arc
\item definition~\ref{def:pnc:preset} of preset $\bullet t_i$ in $PN$
\item definition~\ref{def:pnc:enable} of enabled transition in $PN$
\end{enumerate}
\item Then $t_i$ is enabled and can fire in $PN$, as a result it can belong to $T_k$ -- from definition~\ref{def:pnc:enable} of enabled transition

\end{enumerate}
\item And $consumesmore \notin A$, since $A$ is an answer set of $\Pi^5$ -- from rule $a\ref{a:c:overc:gen}$ and supported rule proposition
\begin{enumerate}
\item Then $\nexists consumesmore(p,k) \in A$ -- from rule $a\ref{a:c:overc:place}$ and supported rule proposition
\item  Then $\nexists \{ holds(p,q_c,c,k), tot\_decr(p,q1_c,c,k) \} \subseteq A, q1_c>q_c$ in $body(a\ref{a:c:overc:place})$ -- from $a\ref{a:c:overc:place}$ and forced atom proposition
\item Then $\nexists c \in C, \nexists p \in P, (\sum_{t_i \in \{t_0,\dots,t_x\}, p \in \bullet t_i}{m_{W(p,t_i)}(c)}+ \\ \sum_{t_i \in \{t_0,\dots,t_x\}, p \in R(t_i)}{m_{M_k(p)}(c)}) > m_{M_k(p)}(c)$ -- from the following
\begin{enumerate}
\item $holds(p,q_c,c,k)$ represents $q_c=m_{M_k(p)}(c)$ -- construction of $\Pi^5$, inductive assumption
\item $tot\_decr(p,q1_c,c,k) \in A$ if $\{ del(p,{q1_0}_c,t_0,c,k), \dots, \\ del(p,{q1_x}_c,t_x,c,k) \} \subseteq A$, where $q1_c = {q1_0}_c+\dots+{q1_x}_c$ -- from $r\ref{r:c:totdecr}$ and forced atom proposition
\item $del(p,{q1_i}_c,t_i,c,k) \in A$ if $\{ fires(t_i,k), ptarc(p,t_i,{q1_i}_c,c,k) \} \subseteq A$ -- from $r\ref{r:c:del}$ and supported rule proposition
\item $del(p,{q1_i}_c,t_i,c,k)$ either represents removal of ${q1_i}_c = m_{W(p,t_i)}(c)$ tokens from $p \in \bullet t_i$; or it represents removal of ${q1_i}_c = m_{M_k(p)}(c)$ tokens from $p \in R(t_i)$-- from rules $r\ref{r:c:del},f\ref{f:c:ptarc},f\ref{f:c:rptarc}$, supported rule proposition, and definition~\ref{def:pnc:texec} of transition execution in $PN$
\end{enumerate}
\item Then the set of transitions in $T_k$ do not conflict -- by the definition~\ref{def:pnc:conflict} of conflicting transitions
\end{enumerate}

\item And for each reset transition $t_r$ with $enabled(t_r,k) \in A$, $fires(t_r,k) \in A$, since $A$ is an answer set of $\Pi^5$ - from rule $f\ref{f:c:rptarc:elim}$ and supported rule proposition
\begin{enumerate}
\item Then the firing set $T_k$ satisfies the reset transition requirement of definition~\ref{def:pnc:firing_set} (firing set)
\end{enumerate}

\item Then $\{t_0, \dots, t_x\} = T_k$ -- using 1(a),1(b), 1(d) above; and using 1(c) it is a maximal firing set  

\end{enumerate}

\item We show that $M_{k+1}$ is produced by firing $T_k$ in $M_k$. Let $holds(p,q_c,c,k+1) \in A$
\begin{enumerate}
\item Then $\{ holds(p,q1_c,c,k), tot\_incr(p,q2_c,c,k), tot\_decr(p,q3_c,c,k) \} \subseteq A : q_c=q1_c+q2_c-q3_c$ -- from rule $r\ref{r:c:nextstate}$ and supported rule proposition \label{x:1:induction}
\item \label{x:2:induction} Then, $holds(p,q1_c,c,k) \in A$ represents $q1_c=m_{M_k(p)}(c)$ -- construction, inductive assumption;  
and $\{add(p,{q2_0}_c,t_0,c,k), \dots, add(p,{q2_j}_c,t_j,c,k)\} \subseteq A : {q2_0}_c + \dots + {q2_j}_c = q2_c$  \label{stmt:add:induction} and $\{del(p,{q3_0}_c,t_0,c,k), \dots, del(p,{q3_l}_c,t_l,c,k)\} \subseteq A : {q3_0}_c + \dots + {q3_l}_c = q3_c$ \label{stmt:del:induction}  -- rules $r\ref{r:c:totincr},r\ref{r:c:totdecr}$ and supported rule proposition, respectively
\item Then $\{ fires(t_0,k), \dots, fires(t_j,k) \} \subseteq A$ and $\{ fires(t_0,k), \dots, fires(t_l,k) \} \\ \subseteq A$ -- rules $r\ref{r:c:add},r\ref{r:c:del}$ and supported rule proposition, respectively
\item Then $\{ fires(t_0,k), \dots, $ $fires(t_j,k) \} \cup \{ fires(t_0,k), \dots, $ $fires(t_l,k) \} \subseteq $ $A = \{ fires(t_0,k), \dots, $ $fires(t_x,k) \} \subseteq A$ -- set union of subsets
\item Then for each $fires(t_x,k) \in A$ we have $t_x \in T_k$ -- already shown in item~\ref{prove:pri:fires_tk} above
\item Then $q_c = m_{M_k(p)}(c) + \sum_{t_x \in T_k \wedge p \in t_x \bullet}{m_{W(t_x,p)}(c)} - (\sum_{t_x \in T_k \wedge p \in \bullet t_x}{m_{W(p,t_x)}(c)} + \sum_{t_x \in T_k \wedge p \in R(t_x)}{m_{M_k(p)}(c)})$ -- from 
\eqref{x:2:induction} above and the following
\begin{enumerate}
\item Each $add(p,{q_j}_c,t_j,c,k) \in A$ represents ${q_j}_c=m_{W(t_j,p)}(c)$ for $p \in t_j \bullet$ -- rule $r\ref{r:c:add},f\ref{f:c:tparc}$ encoding, and definition~\ref{def:pnc:texec} of transition execution in $PN$ %
\item Each $del(p,t_y,{q_y}_c,c,k) \in A$ represents either ${q_y}_c=m_{W(p,t_y)}(c)$ for $p \in \bullet t_y$, or ${q_y}_c=m_{M_k(p)}(c)$ for $p \in R(t_y)$ -- from rule $r\ref{r:c:del},f\ref{f:c:ptarc}$ encoding and definition~\ref{def:pnc:texec} of transition execution in $PN$; or from rule $r\ref{r:c:del},f\ref{f:c:rptarc}$ encoding and definition of reset arc in $PN$
\item Each $tot\_incr(p,q2_c,c,k) \in A$ represents $q2_c=\sum_{t_x \in T_k \wedge p \in t_x  \bullet}{m_{W(t_x,p)}(c)}$ -- aggregate assignment atom semantics in rule $r\ref{r:c:totincr}$
\item Each $tot\_decr(p,q3_c,c,k) \in A$ represents $q3_c=\sum_{t_x \in T_k \wedge p \in \bullet t_x}{m_{W(p,t_x)}(c)} + \sum_{t_x \in T_k \wedge p \in R(t_x)}{m_{M_k(p)}(c)}$ -- aggregate assignment atom semantics in rule $r\ref{r:c:totdecr}$
\end{enumerate}
\item Then, $m_{M_{k+1}(p)}(c) = q_c$ -- since $holds(p,q_c,c,k+1) \in A$ encodes $q_c=m_{M_{k+1}(p)}(c)$ -- from construction
\end{enumerate}
\end{enumerate}

\noindent
As a result, for any $n > k$, $T_n$ will be a valid firing set for $M_n$ and $M_{n+1}$ will be its target marking. 

\noindent
{\bf Conclusion:} Since both \eqref{prove:x2a:pnc} and \eqref{prove:a2x:pnc} hold, $X=M_0,T_k,M_1,\dots,M_k,T_{k+1}$ is an execution sequence of $PN(P,T,E,C,W,R,I,Q,QW)$ (w.r.t $M_0$) iff there is an answer set $A$ of $\Pi^5(PN,M_0,k,ntok)$ such that \eqref{eqn:pnc:fires} and \eqref{eqn:pnc:holds} hold.

\section{Proof of Proposition~\ref{prop:pri}}

Let $PN=(P,T,E,C,W,R,I,Q,QW,Z)$ be a Petri Net, $M_0$ be its initial marking and let $\Pi^6(PN,M_0,k,ntok)$ be the ASP encoding of $PN$ and $M_0$ over a simulation length $k$, with maximum $ntok$ tokens on any place node, as defined in section~\ref{sec:enc_priority}. Then $X=M_0,T_k,M_1,\dots,M_k,T_k,M_{k+1}$ is an execution sequence of $PN$ (w.r.t. $M_0$) iff there is an answer set $A$ of $\Pi^6(PN,M_0,k,ntok)$ such that: 
\begin{equation}
\{ fires(t,ts) : t \in T_{ts}, 0 \leq ts \leq k\} = \{ fires(t,ts) : fires(t,ts) \in A \} \label{eqn:pnpri:fires}
\end{equation}
\begin{equation}
\begin{split}
\{ holds(p,q,c,ts) &: p \in P, c/q = M_{ts}(p), 0 \leq ts \leq k+1 \} \\
&= \{ holds(p,q,c,ts) : holds(p,q,c,ts) \in A \} \label{eqn:pnpri:holds}
\end{split}
\end{equation}

We prove this by showing that:
\begin{enumerate}[(I)]
\item Given an execution sequence $X$, we create a set $A$ such that it satisfies \eqref{eqn:pnpri:fires} and \eqref{eqn:pnpri:holds} and show that $A$ is an answer set of $\Pi^6$ \label{prove:x2a:pnpri}
\item Given an answer set $A$ of $\Pi^6$, we create an execution sequence $X$ such that \eqref{eqn:pnpri:fires} and \eqref{eqn:pnpri:holds} are satisfied. \label{prove:a2x:pnpri}
\end{enumerate}

\noindent
{\bf First we show (\ref{prove:x2a:pnpri})}: We create a set $A$ as a union of the following sets:
\begin{enumerate}
\item $A_1=\{ num(n) : 0 \leq n \leq ntok \}$\label{pnpri:builda:num}
\item $A_2=\{ time(ts) : 0 \leq ts \leq k\}$\label{pnpri:builda:time}
\item $A_3=\{ place(p) : p \in P \}$\label{pnpri:builda:place}
\item $A_4=\{ trans(t) : t \in T \}$\label{pnpri:builda:trans}
\item $A_5=\{ color(c) : c \in C \}$
\item $A_6=\{ ptarc(p,t,n_c,c,ts) : (p,t) \in E^-, c \in C, n_c=m_{W(p,t)}(c), n_c > 0, 0 \leq ts \leq k \}$, where $E^- \subseteq E$\label{pnpri:builda:ptarc}
\item $A_7=\{ tparc(t,p,n_c,c,ts) : (t,p) \in E^+, c \in C, n_c=m_{W(t,p)}(c), n_c > 0, 0 \leq ts \leq k \}$, where $E^+ \subseteq E$\label{pnpri:builda:tparc}
\item $A_8=\{ holds(p,q_c,c,0) : p \in P, c \in C, q_c=m_{M_{0}(p)}(c) \}$\label{pnpri:builda:holds0}
\item $A_9=\{ ptarc(p,t,n_c,c,ts) : p \in R(t), c \in C, n_c = m_{M_{ts}(p)}, n_c > 0, 0 \leq ts \leq k \}$
\item $A_{10}=\{ iptarc(p,t,1,c,ts) : p \in I(t), c \in C, 0 \leq ts < k \}$
\item $A_{11}=\{ tptarc(p,t,n_c,c,ts) : (p,t) \in Q, c \in C, n_c=m_{QW(p,t)}(c), n_c > 0, 0 \leq ts \leq k \}$
\item $A_{12}=\{ notenabled(t,ts) : t \in T, 0 \leq ts \leq k, \exists c \in C, (\exists p \in \bullet t, m_{M_{ts}(p)}(c) < m_{W(p,t)}(c)) \vee (\exists p \in I(t), m_{M_{ts}(p)}(c) > 0) \vee (\exists (p,t) \in Q, m_{M_{ts}(p)}(c) < m_{QW(p,t)}(c)) \}$ \newline per definition~\ref{def:pnpri:enable} (enabled transition)\label{pnpri:builda:notenabled} %
\item $A_{13}=\{ enabled(t,ts) : t \in T, 0 \leq ts \leq k, \forall c \in C, (\forall p \in \bullet t, m_{W(p,t)}(c) \leq m_{M_{ts}(p)}(c)) \wedge (\forall p \in I(t), m_{M_{ts}(p)}(c) = 0) \wedge (\forall (p,t) \in Q, m_{M_{ts}(p)}(c) \geq m_{QW(p,t)}(c)) \}$ \newline per definition~\ref{def:pnpri:enable} (enabled transition)\label{pnpri:builda:enabled} %
\item $A_{14}=\{ fires(t,ts) : t \in T_{ts}, 0 \leq ts \leq k \}$ \newline per definition~\ref{def:pnpri:firing_set} (firing set), only an enabled transition may fire\label{pnpri:builda:fires}
\item $A_{15}=\{ add(p,q_c,t,c,ts) : t \in T_{ts}, p \in t \bullet, c \in C, q_c=m_{W(t,p)}(c), 0 \leq ts \leq k \}$ \newline per definition~\ref{def:pnpri:texec} (transition execution)\label{pnpri:builda:add}
\item $A_{16}=\{ del(p,q_c,t,c,ts) : t \in T_{ts}, p \in \bullet t, c \in C, q_c=m_{W(p,t)}(c), 0 \leq ts \leq k \} \cup \{ del(p,q_c,t,c,ts) : t \in T_{ts}, p \in R(t), c \in C, q_c=m_{M_{ts}(p)}(c), 0 \leq ts \leq k \}$ \newline per definition~\ref{def:pnpri:texec} (transition execution) %
\item $A_{17}=\{ tot\_incr(p,q_c,c,ts) : p \in P, c \in C, q_c=\sum_{t \in T_{ts}, p \in t \bullet}{m_{W(t,p)}(c)}, 0 \leq ts \leq k \}$ \newline per definition~\ref{def:pnpri:exec} (firing set execution)\label{pnpri:builda:tot_incr}
\item $A_{18}=\{ tot\_decr(p,q_c,c,ts): p \in P, c \in C, q_c=\sum_{t \in T_{ts}, p \in \bullet t}{m_{W(p,t)}(c)}+ \\ \sum_{t \in T_{ts}, p \in R(t) }{m_{M_{ts}(p)}(c)}, 0 \leq ts \leq k \}$ \newline per definition~\ref{def:pnpri:exec} (firing set execution)\label{pnpri:builda:tot_decr}
\item $A_{19}=\{ consumesmore(p,ts) : p \in P, c \in C, q_c=m_{M_{ts}(p)}(c), \\ q1_c=\sum_{t \in T_{ts}, p \in \bullet t}{m_{W(p,t)}(c)} + \sum_{t \in T_{ts}, p \in R(t)}{m_{M_{ts}(p)}(c)}, q1_c > q_c, 0 \leq ts \leq k \}$ \newline per definition~\ref{def:pnpri:conflict} (conflicting transitions)\label{pnpri:builda:consumesmore}
\item $A_{20}=\{ consumesmore : \exists p \in P, c \in C, q_c=m_{M_{ts}(p)}(c), \\ q1_c=\sum_{t \in T_{ts}, p \in \bullet t}{m_{W(p,t)}(c)} + \sum_{t \in T_{ts}, p \in R(t)}(m_{M_{ts}(p)}(c)), q1_c > q_c, 0 \leq ts \leq k \}$ \newline per definition~\ref{def:pnpri:conflict} (conflicting transitions)\label{pnpri:builda:consumesmore1}
\item $A_{21}=\{ holds(p,q_c,c,ts+1) : p \in P, c \in C, q_c=m_{M_{ts+1}(p)}(c), 0 \leq ts < k\}$, \newline where $M_{ts+1}(p) = M_j(p) - (\sum_{\substack{t \in T_{ts}, p \in \bullet t}}{W(p,t)} + \sum_{t \in T_{ts}, p \in R(t)}M_{ts}(p)) + \\ \sum_{\substack{t \in T_{ts}, p \in t \bullet}}{W(t,p)}$ \newline according to definition~\ref{def:pnpri:exec} (firing set execution)\label{pnpri:builda:holds}
\item $A_{22}=\{ transpr(t,pr) : t \in T, pr=Z(t) \}$
\item $A_{23}=\{ notprenabled(t,ts) : t \in T, enabled(t,ts) \in A_{13}, (\exists tt \in T, enabled(tt,ts) \in A_{13}, Z(tt) < Z(t)), 0 \leq ts \leq k \} \\ = \{ notprenabled(t,ts) : t \in T, (\forall p \in \bullet t, W(p,t) \leq M_{ts}(p)), (\forall p \in I(t), M_{ts}(p) = 0), (\forall (p,t) \in Q, M_0(p) \geq WQ(p,t)), \exists tt \in T, (\forall pp \in \bullet tt, W(pp,tt) \leq M_{ts}(pp)), \\ (\forall pp \in I(tt), M_{ts}(pp) = 0), (\forall (pp,tt) \in Q, M_{ts}(pp) \geq QW(pp,tt)), Z(tt) < Z(t), 0 \leq ts \leq k \}$
\item $A_{24}=\{ prenabled(t,ts) : t \in T, enabled(t,ts) \in A_{13}, (\nexists tt \in T: enabled(tt,ts) \in A_{13}, Z(tt) < Z(t)), 0 \leq ts \leq k \} = \{ prenabled(t,ts) : t \in T, (\forall p \in \bullet t, W(p,t) \leq M_{ts}(p)), (\forall p \in I(t), M_{ts}(p) = 0), (\forall (p,t) \in Q, M_0(p) \geq WQ(p,t)), \nexists tt \in T, (\forall pp \in \bullet tt, W(pp,tt) \leq M_{ts}(pp)), (\forall pp \in I(tt), M_{ts}(pp) = 0), (\forall (pp,tt) \in Q, M_{ts}(pp) \geq QW(pp,tt)), Z(tt) < Z(t), 0 \leq ts \leq k \}$
\item $A_{25}=\{ could\_not\_have(t,ts): t \in T, prenabled(t,ts) \in A_{24}, fires(t,ts) \notin A_{14}, (\exists p \in \bullet t \cup R(t): q > M_{ts}(p) - (\sum_{t' \in T_{ts}, p \in \bullet t'}{W(p,t')} + \sum_{t' \in T_{ts}, p \in R(t')}{M_{ts}(p)}), q=W(p,t) \text{~if~} (p,t) \in E^- \text{~or~} R(t) \text{~otherwise}), 0 \leq ts \leq k\}  =\{ could\_not\_have(t,ts) : t \in T, (\forall p \in \bullet t, W(p,t) \leq M_{ts}(p)), (\forall p \in I(t), M_{ts}(p) = 0), (\forall (p,t) \in Q, M_0(p) \geq QW(p,t)), (\nexists tt \in T, $ $(\forall pp \in \bullet tt, W(pp,tt) \leq M_{ts}(pp)), (\forall pp \in I(tt), M_{ts}(pp) = 0), (\forall (pp,tt) \in Q, M_{ts}(pp) \geq QW(pp,tt)), Z(tt) < Z(t)), t \notin T_{ts}, (\exists p \in \bullet t \cup R(t): q > M_{ts}(p) - (\sum_{t' \in T_{ts}, p \in \bullet t'}{W(p,t')} + \sum_{t' \in T_{ts}, p \in R(t')}{M_{ts}(p)}), q=W(p,t) \text{~if~} \\ (p,t) \in E^- \text{~or~} R(t) \text{~otherwise}), 0 \leq ts \leq k \}$ \newline
per the maximal firing set semantics
\end{enumerate}

\noindent
{\bf We show that $A$ satisfies \eqref{eqn:pnpri:fires} and \eqref{eqn:pnpri:holds}, and $A$ is an answer set of $\Pi^6$.}

$A$ satisfies \eqref{eqn:pnpri:fires} and \eqref{eqn:pnpri:holds} by its construction above. We show $A$ is an answer set of $\Pi^6$ by splitting. We split $lit(\Pi^6)$ into a sequence of $9k+11$ sets:

\begin{itemize}\renewcommand{\labelitemi}{$\bullet$}
\item $U_0= head(f\ref{f:c:place}) \cup head(f\ref{f:c:trans}) \cup head(f\ref{f:c:col}) \cup head(f\ref{f:c:time}) \cup head(f\ref{f:c:num}) \cup head(i\ref{i:c:init}) \cup head(f\ref{f:c:pr}) = \{place(p) : p \in P\} \cup \{ trans(t) : t \in T\} \cup \{ col(c) : c \in C \} \cup \{ time(0), \dots, time(k)\} \cup \{num(0), \dots, num(ntok)\} \cup \{ holds(p,q_c,c,0) : p \in P, c \in C, q_c=m_{M_0(p)}(c) \} \cup \{ transpr(t,pr) : t \in T, pr=Z(t) \} $
\item $U_{9k+1}=U_{9k+0} \cup head(f\ref{f:c:ptarc})^{ts=k} \cup head(f\ref{f:c:tparc})^{ts=k} \cup head(f\ref{f:c:rptarc})^{ts=k} \cup head(f\ref{f:c:iptarc})^{ts=k} \cup head(f\ref{f:c:tptarc})^{ts=k} = U_{10k+0} \cup \{ ptarc(p,t,n_c,c,k) : (p,t) \in E^-, c \in C, n_c=m_{W(p,t)}(c) \} \\ \cup \{  tparc(t,p,n_c,c,k) : (t,p) \in E^+, c \in C, n_c=m_{W(t,p)}(c) \} \cup \{ ptarc(p,t,n_c,c,k) : p \in R(t), c \in C, n_c=m_{M_{k}(p)}(c), n > 0 \} \cup \{ iptarc(p,t,1,c,k) : p \in I(t), c \in C \} \cup \{ tptarc(p,t,n_c,c,k) : (p,t) \in Q, c \in C, n_c=m_{QW(p,t)}(c) \}$
\item $U_{9k+2}=U_{9k+1} \cup head(e\ref{e:c:ne:ptarc})^{ts=k} \cup head(e\ref{e:c:ne:iptarc})^{ts=k} \cup head(e\ref{e:c:ne:tptarc})^{ts=k} = U_{9k+1} \\ \cup \{ notenabled(t,k) : t \in T \}$
\item $U_{9k+3}=U_{9k+2} \cup head(e\ref{e:c:enabled})^{ts=k} = U_{9k+2} \cup \{ enabled(t,k) : t \in T \}$
\item $U_{9k+4}=U_{9k+3} \cup head(a\ref{a:c:prne})^{ts=k} = U_{9k+3} \cup \{ notprenabled(t,k) : t \in T \}$
\item $U_{9k+5}=U_{9k+4} \cup head(a\ref{a:c:prenabled})^{ts=k} = U_{9k+4} \cup \{ prenabled(t,k) : t \in T \}$
\item $U_{9k+6}=U_{9k+5} \cup head(a\ref{a:c:prfires})^{ts=k} = U_{9k+5} \cup \{ fires(t,k) : t \in T \}$
\item $U_{9k+7}=U_{9k+6}  \cup head(r\ref{r:c:add})^{ts=k} \cup head(r\ref{r:c:del})^{ts=k} = U_{9k+6} \cup \{ add(p,q_c,t,c,k) : p \in P, t \in T, c \in C, q_c=m_{W(t,p)}(c) \} \cup \{ del(p,q_c,t,c,k) : p \in P, t \in T, c \in C, q_c=m_{W(p,t)}(c) \} \cup \{ del(p,q_c,t,c,k) : p \in P, t \in T, c \in C, q_c=m_{M_{k}(p)}(c) \}$
\item $U_{9k+8}=U_{9k+7} \cup head(r\ref{r:c:totincr})^{ts=k} \cup head(r\ref{r:c:totdecr})^{ts=k} = U_{9k+7} \cup \{ tot\_incr(p,q_c,c,k) : p \in P, c \in C, 0 \leq q_c \leq ntok \} \cup \{ tot\_decr(p,q_c,c,k) : p \in P, c \in C, 0 \leq q_c \leq ntok \}$
\item $U_{9k+9}=U_{9k+8} \cup head(a\ref{a:c:overc:place})^{ts=k} \cup head(r\ref{r:c:nextstate})^{ts=k} \cup head(a\ref{a:c:prmaxfire:cnh})^{ts=k} = U_{9k+8} \cup \\ \{ consumesmore(p,k) : p \in P\} \cup \{ holds(p,q,k+1) : p \in P, 0 \leq q \leq ntok \} \cup \{ could\_not\_have(t,k) : t \in T \}$
\item $U_{9k+10}=U_{9k+9} \cup head(a\ref{a:c:overc:gen})^{ts=k} = U_{9k+9} \cup \{ consumesmore \}$
\end{itemize}
where $head(r_i)^{ts=k}$ are head atoms of ground rule $r_i$ in which $ts=k$. We write $A_i^{ts=k} = \{ a(\dots,ts) : a(\dots,ts) \in A_i, ts=k \}$ as short hand for all atoms in $A_i$ with $ts=k$. $U_{\alpha}, 0 \leq \alpha \leq 9k+10$ form a splitting sequence, since each $U_i$ is a splitting set of $\Pi^6$, and $\langle U_{\alpha}\rangle_{\alpha < \mu}$ is a monotone continuous sequence, where $U_0 \subseteq U_1 \dots \subseteq U_{9k+10}$ and $\bigcup_{\alpha < \mu}{U_{\alpha}} = lit(\Pi^6)$. 

We compute the answer set of $\Pi^6$ using the splitting sets as follows:
\begin{enumerate}
\item $bot_{U_0}(\Pi^6) = f\ref{f:c:place} \cup f\ref{f:c:trans} \cup f\ref{f:c:col} \cup f\ref{f:c:time} \cup f\ref{f:c:num} \cup i\ref{i:c:init} \cup f\ref{f:c:pr}$ and $X_0 = A_1 \cup \dots \cup A_5 \cup A_8$ ($= U_0$) is its answer set -- using forced atom proposition

\item $eval_{U_0}(bot_{U_1}(\Pi^6) \setminus bot_{U_0}(\Pi^6), X_0) = \{ ptarc(p,t,q_c,c,0) \text{:-}. |  c \in C, $ $q_c=m_{W(p,t)}(c), q_c > 0 \} \cup \{ tparc(t,p,q_c,c,0) \text{:-}. | c \in C, q_c=m_{W(t,p)}(c), q_c > 0 \} \cup \{ ptarc(p,t,q_c,c,0) \text{:-}. | c \in C, $ $q_c=m_{M_0(p)}(c), q_c > 0 \} \cup \{ iptarc(p,t,1,c,0) \text{:-}. | c \in C \} \cup \{ tptarc(p,t,q_c,c,0) \text{:-}. | c \in C, q_c = m_{QW(p,t)}(c), q_c > 0 \} $. Its answer set $X_1=A_6^{ts=0} \cup A_7^{ts=0} \cup A_9^{ts=0} \cup A_{10}^{ts=0} \cup A_{11}^{ts=0}$ -- using forced atom proposition and construction of $A_6, A_7, A_9, A_{10}, A_{11}$.

\item $eval_{U_1}(bot_{U_2}(\Pi^6) \setminus bot_{U_1}(\Pi^6), X_0 \cup X_1) = \{ notenabled(t,0) \text{:-} . | (\{ trans(t), \\ ptarc(p,t,n_c,c,0), holds(p,q_c,c,0) \} \subseteq X_0 \cup X_1, \text{~where~}  q_c < n_c) \text{~or~} \\ (\{ notenabled(t,0) \text{:-} . | (\{ trans(t), iptarc(p,t,n2_c,c,0), holds(p,q_c,c,0) \} \subseteq X_0 \cup X_1, \\ \text{~where~}  q_c \geq n2_c \}) \text{~or~} (\{ trans(t), tptarc(p,t,n3_c,c,0), $ $holds(p,q_c,c,0) \} \subseteq X_0 \cup X_1, \text{~where~} $ $q_c < n3_c) \}$. Its answer set $X_2=A_{12}^{ts=0}$ -- using  forced atom proposition and construction of $A_{12}$.
\begin{enumerate}
\item where, $q_c=m_{M_0(p)}(c)$, and $n_c=m_{W(p,t)}(c)$ for an arc $(p,t) \in E^-$ -- by construction of $i\ref{i:c:init}$ and $f\ref{f:c:ptarc}$ in $\Pi^6$, and 
\item in an arc $(p,t) \in E^-$, $p \in \bullet t$ (by definition~\ref{def:pnpri:preset} of preset)
\item $n2_c=1$ -- by construction of $iptarc$ predicates in $\Pi^6$, meaning $q_c \geq n2_c \equiv q_c \geq 1 \equiv q_c > 0$,
\item $tptarc(p,t,n3_c,c,0)$ represents $n3_c=m_{QW(p,t)}(c)$, where $(p,t) \in Q$ 
\item thus, $notenabled(t,0) \in X_1$ represents $\exists c \in C, (\exists p \in \bullet t : m_{M_0(p)}(c) < m_{W(p,t)}(c)) \vee (\exists p \in I(t) : m_{M_0(p)}(c) > 0) \vee (\exists (p,t) \in Q : m_{M_0(p)}(c) < m_{QW(p,t)}(c))$.
\end{enumerate}

\item $eval_{U_2}(bot_{U_3}(\Pi^6) \setminus bot_{U_2}(\Pi^6), X_0 \cup \dots \cup X_2) = \{ enabled(t,0) \text{:-}. | trans(t) \in X_0 \cup \dots \cup X_2, notenabled(t,0) \notin X_0 \cup \dots \cup X_2 \}$. Its answer set is $X_3 = A_{13}^{ts=0}$ -- using forced atom proposition and construction of $A_{13}$. \label{pnpri:x2a:base:enabled}
\begin{enumerate}
\item since an $enabled(t,0) \in X_3$ if $\nexists ~notenabled(t,0) \in X_0 \cup \dots \cup X_2$; which is equivalent to $\nexists t, \forall c \in C, (\nexists p \in \bullet t, m_{M_0(p)}(c) < m_{W(p,t)}(c)), (\nexists p \in I(t), $ $m_{M_0(p)}(c) > 0), (\nexists (p,t) \in Q : m_{M_0(p)}(c) < m_{QW(p,t)}(c) ), \forall c \in C, (\forall p \in \bullet t: m_{M_0(p)}(c) \geq m_{W(p,t)}(c)), (\forall p \in I(t) : m_{M_0(p)}(c) = 0)$.
\end{enumerate}

\item $eval_{U_3}(bot_{U_4}(\Pi^6) \setminus bot_{U_3}(\Pi^6), X_0 \cup \dots \cup X_3) = \{ notprenabled(t,0) \text{:-}. | \\ \{ enabled(t,0), transpr(t,p), enabled(tt,0), transpr(tt,pp) \} \subseteq X_0 \cup \dots \cup X_3, pp < p \}$. Its answer set is $X_4 = A_{23}^{ts=k}$ -- using forced atom proposition and construction of $A_{23}$. \label{pnpri:x2a:base:notprenabled}
\begin{enumerate}
\item $enabled(t,0)$ represents $\exists t \in T, \forall c \in C, (\forall p \in \bullet t, m_{M_0(p)}(c) \geq m_{W(p,t)}(c)), $ $(\forall p \in I(t), m_{M_0(p)}(c) = 0), (\forall (p,t) \in Q, m_{M_0(p)}(c) \geq m_{QW(p,t)}(c))$
\item $enabled(tt,0)$ represents $\exists tt \in T, \forall c \in C, (\forall pp \in \bullet tt, m_{M_0(pp)}(c) \geq \\ m_{W(pp,tt)}(c)) \wedge (\forall pp \in I(tt), m_{M_0(pp)}(c) = 0), (\forall (pp,tt) \in Q, m_{M_0(pp)}(c) \geq m_{QW(pp,tt)}(c))$  
\item $transpr(t,p)$ represents $p=Z(t)$ -- by construction
\item $transpr(tt,pp)$ represents $pp=Z(tt)$ -- by construction
\item thus, $notprenabled(t,0)$ represents $\forall c \in C, (\forall p \in \bullet t, m_{M_0(p)}(c) \geq \\ m_{W(p,t)}(c)) \wedge (\forall p \in I(t), m_{M_0(p)}(c) = 0), \exists tt \in T, (\forall pp \in \bullet tt, m_{M_0(pp)}(c) \geq m_{W(pp,tt)}(c)) \wedge (\forall pp \in I(tt), m_{M_0(pp)}(c) = 0), Z(tt) < Z(t)$ 
\item which is equivalent to $(\forall p \in \bullet t: M_0(p) \geq W(p,t)) \wedge (\forall p \in I(t), M_0(p) = 0), \exists tt \in T, (\forall pp \in \bullet tt, M_0(pp) \geq W(pp,tt)), (\forall pp \in I(tt), M_0(pp) = 0), Z(tt) < Z(t)$ -- assuming multiset domain $C$ for all operations
\end{enumerate}

\item $eval_{U_4}(bot_{U_5}(\Pi^6) \setminus bot_{U_4}(\Pi^6), X_0 \cup \dots \cup X_4) = \{ prenabled(t,0) \text{:-}. | enabled(t,0) \in X_0 \cup \dots \cup X_4, notprenabled(t,0) \notin X_0 \cup \dots \cup X_4 \}$. Its answer set is $X_5 = A_{24}^{ts=k}$ -- using forced atom proposition and construction of $A_{24}$ \label{pnpri:x2a:base:prenabled}
\begin{enumerate}
\item $enabled(t,0)$ represents $\forall c \in C, (\forall p \in \bullet t, m_{M_0(p)}(c) \geq m_{W(p,t)}(c)), (\forall p \in I(t), m_{M_0(p)}(c) = 0), (\forall (p,t) \in Q, m_{M_0(p)}(c) \geq m_{QW(p,t)}(c))  \equiv (\forall p \in \bullet t, \\ M_0(p) \geq W(p,t)), (\forall p \in I(t), M_0(p) = 0), (\forall (p,t) \in Q, M_0(p) \geq QW(p,t))$ -- from \ref{pnpri:x2a:base:enabled} above and assuming multiset domain $C$ for all operations
\item $notprenabled(t,0)$ represents $(\forall p \in \bullet t, M_0(p) \geq W(p,t)), (\forall p \in I(t), \\ M_0(p) = 0), (\forall (p,t) \in Q, M_0(p) \geq QW(p,t)), \exists tt \in T, (\forall pp \in \bullet tt, M_0(pp) \geq W(pp,tt)),  (\forall pp \in I(tt), M_0(pp) = 0), (\forall (pp,tt) \in Q, M_0(pp) \geq W(pp,tt)), \\ Z(tt) < Z(t)$ -- from \ref{pnpri:x2a:base:notprenabled} above and assuming multiset domain $C$ for all operations
\item then, $prenabled(t,0)$ represents $(\forall p \in \bullet t, M_0(p) \geq W(p,t)), (\forall p \in I(t), \\ M_0(p) = 0), (\forall (p,t) \in Q, M_0(p) \geq QW(p,t)), \nexists tt \in T, ((\forall pp \in \bullet tt, M_0(pp) \geq W(pp,tt)),  (\forall pp \in I(tt), M_0(pp) = 0), (\forall (pp,tt) \in Q, M_0(pp) \geq W(pp,tt)), \\ Z(tt) < Z(t))$ -- from (a), (b) and $enabled(t,0) \in X_0 \cup \dots \cup X_4$ 
\end{enumerate}

\item $eval_{U_5}(bot_{U_6}(\Pi^6) \setminus bot_{U_5}(\Pi^6), X_0 \cup \dots \cup X_5) = \{\{fires(t,0)\} \text{:-}. | prenabled(t,0) \\ \text{~holds in~} X_0 \cup \dots \cup X_5 \}$. It has multiple answer sets $X_{6.1}, \dots, X_{6.n}$, corresponding to elements of power set of $fires(t,0)$ atoms in $eval_{U_5}(...)$ -- using supported rule proposition. Since we are showing that the union of answer sets of $\Pi^6$ determined using splitting is equal to $A$, we only consider the set that matches the $fires(t,0)$ elements in $A$ and call it $X_6$, ignoring the rest. Thus, $X_6 = A_{14}^{ts=0}$, representing $T_0$.
\begin{enumerate}
\item in addition, for every $t$ such that $prenabled(t,0) \in X_0 \cup \dots \cup X_5,  R(t) \neq \emptyset$; $fires(t,0) \in X_6$ -- per definition~\ref{def:pnpri:firing_set} (firing set); requiring that a reset transition is fired when enabled
\item thus, the firing set $T_0$ will not be eliminated by the constraint $f\ref{f:c:pr:rptarc:elim}$ 
\end{enumerate}

\item $eval_{U_6}(bot_{U_7}(\Pi^6) \setminus bot_{U_6}(\Pi^6), X_0 \cup \dots \cup X_6) = \{add(p,n_c,t,c,0) \text{:-}. | \{fires(t,0), \\ tparc(t,p,n_c,c,0) \} \subseteq X_0 \cup \dots \cup X_6 \} \cup \{ del(p,n_c,t,c,0) \text{:-}. | \{ fires(t,0), \\ ptarc(p,t,n_c,c,0) \} \subseteq X_0 \cup \dots \cup X_6 \}$. It's answer set is $X_7 = A_{15}^{ts=0} \cup A_{16}^{ts=0}$ -- using forced atom proposition and definitions of $A_{15}$ and $A_{16}$. 
\begin{enumerate}
\item where, each $add$ atom is equivalent to $n_c=m_{W(t,p)}(c),c \in C, p \in t \bullet$, 
\item and each $del$ atom is equivalent to $n_c=m_{W(p,t)}(c), c \in C, p \in \bullet t$; or $n_c=m_{M_0(p)}(c), c \in C, p \in R(t)$,
\item representing the effect of transitions in $T_0$ -- by construction
\end{enumerate}

\item $eval_{U_7}(bot_{U_8}(\Pi^6) \setminus bot_{U_7}(\Pi^6), X_0 \cup \dots \cup X_7) = \{tot\_incr(p,qq_c,c,0) \text{:-}. | qq_c=\sum_{add(p,q_c,t,c,0) \in X_0 \cup \dots \cup X_7}{q_c} \} \cup \{ tot\_decr(p,qq_c,c,0) \text{:-}. | qq_c=\sum_{del(p,q_c,t,c,0) \in X_0 \cup \dots \cup X_7}{q_c} \}$. It's answer set is $X_8 = A_{17}^{ts=0} \cup A_{18}^{ts=0}$ --  using forced atom proposition and definitions of $A_{17}$ and $A_{18}$.
\begin{enumerate}
\item where, each $tot\_incr(p,qq_c,c,0)$, $qq_c=\sum_{add(p,q_c,t,c,0) \in X_0 \cup \dots X_7}{q_c}$ \\ $\equiv qq_c=\sum_{t \in X_6, p \in t \bullet}{m_{W(p,t)}(c)}$,
\item where, each $tot\_decr(p,qq_c,c,0)$, $qq_c=\sum_{del(p,q_c,t,c,0) \in X_0 \cup \dots X_7}{q_c}$ \\ $\equiv qq=\sum_{t \in X_6, p \in \bullet t}{m_{W(t,p)}(c)} + \sum_{t \in X_6, p \in R(t)}{m_{M_0(p)}(c)}$, 
\item represent the net effect of transitions in $T_0$
\end{enumerate}
\item $eval_{U_8}(bot_{U_9}(\Pi^6) \setminus bot_{U_8}(\Pi^6), X_0 \cup \dots \cup X_8) = \{ consumesmore(p,0) \text{:-}. | \\ \{holds(p,q_c,c,0), tot\_decr(p,q1_c,c,0) \} \subseteq X_0 \cup \dots \cup X_8, q1_c > q_c \} \cup \\ \{ holds(p,q_c,c,1) \text{:-}. | \{ holds(p,q1_c,c,0), tot\_incr(p,q2_c,c,0), tot\_decr(p,q3_c,c,0) \} \\ \subseteq X_0 \cup \dots \cup X_6, q_c=q1_c+q2_c-q3_c \}  \cup \{ could\_not\_have(t,0) \text{:-}. | \{ prenabled(t,0), \\ ptarc(s,t,q,c,0), holds(s,qq,c,0), tot\_decr(s,qqq,c,0) \} \subseteq X_0 \cup \dots \cup X_8, \\ fires(t,0) \notin X_0 \cup \dots \cup X_8, q > qq - qqq \}$. It's answer set is $X_9 = A_{19}^{ts=0} \cup A_{21}^{ts=0} \cup A_{25}^{ts=0}$  -- using forced atom proposition and definitions of $A_{19}, A_{21}, A_{25}$.
\begin{enumerate}
\item where, $consumesmore(p,0)$ represents $\exists p : q_c=m_{M_0(p)}(c), q1_c= \\ \sum_{t \in T_0, p \in \bullet t}{m_{W(p,t)}(c)}+\sum_{t \in T_0, p \in R(t)}{m_{M_0(p)}(c)}, q1_c > q_c, c \in C$, indicating place $p$ will be over consumed if $T_0$ is fired, as defined in definition~\ref{def:pnpri:conflict} (conflicting transitions),
\item $holds(p,q_c,c,1)$ if $q_c=m_{M_0(p)}(c)+\sum_{t \in T_0, p \in t \bullet}{m_{W(t,p)}(c)}- \\ (\sum_{t \in T_0, p \in \bullet t}{m_{W(p,t)}(c)}+ \sum_{t \in T_0, p \in R(t)}{m_{M_0(p)}(c)})$ represented by $q_c=m_{M_1(p)}(c)$ for some $c \in C$ -- by construction of $\Pi^6$
\item $could\_not\_have(t,0)$ if
\begin{enumerate}
\item $(\forall p \in \bullet t, W(p,t) \leq M_0(p)), (\forall p \in I(t), M_0(p) = 0), (\forall (p,t) \in Q, \\ M_0(p) \geq WQ(p,t)), \nexists tt \in T, (\forall pp \in \bullet tt, W(pp,tt) \leq M_{ts}(pp)), (\forall pp \in I(tt), M_0(pp) = 0), (\forall (pp,tt) \in Q, M_0(pp) \geq QW(pp,tt)), Z(tt) < Z(t)$,
\item and $q_c > m_{M_0(s)}(c) - (\sum_{t' \in T_0, s \in \bullet t'}{m_{W(s,t')}(c)}+ \sum_{t' \in T_0, s \in R(t)}{m_{M_0(s)}(c)}), \\ q_c = m_{W(s,t)}(c) \text{~if~} s \in \bullet t \text{~or~} m_{M_0(s)}(c) \text{~otherwise}$ for some $c \in C$, which becomes $q > M_0(s) - (\sum_{t' \in T_0, s \in \bullet t'}{W(s,t')}+ \sum_{t' \in T_0, s \in R(t)}{M_0(s)}), q = W(s,t) \text{~if~} s \in \bullet t \text{~or~} M_0(s) \text{~otherwise}$ for all $c \in C$
\item (i), (ii) above combined match the definition of $A_{25}$
\end{enumerate}
\item $X_9$ does not contain $could\_not\_have(t,0)$, when $prenabled(t,0) \in X_0 \cup \dots \cup X_8$ and $fires(t,0) \notin X_0 \cup \dots \cup X_8$ due to construction of $A$, encoding of $a\ref{a:c:overc:place}$ and its body atoms. As a result, it is not eliminated by the constraint  $a\ref{a:c:maxfire:elim}$
\end{enumerate}

\[ \vdots \]

\item $eval_{U_{9k+0}}(bot_{U_{9k+1}}(\Pi^6) \setminus bot_{U_{9k+0}}(\Pi^6), X_0 \cup \dots \cup X_{9k+0}) = \{ ptarc(p,t,q_c,c,k) \text{:-}. |  c \in C, q_c=m_{W(p,t)}(c), q_c > 0 \} \cup \{ tparc(t,p,q_c,c,k) \text{:-}. | c \in C, q_c=m_{W(t,p)}(c), q_c > 0 \} \cup \{ ptarc(p,t,q_c,c,k) \text{:-}. | c \in C, q_c=m_{M_0(p)}(c), q_c > 0 \} \cup \{ iptarc(p,t,1,c,k) \text{:-}. | c \in C \} \cup \{ tptarc(p,t,q_c,c,k) \text{:-}. | c \in C, q_c = m_{QW(p,t)}(c), q_c > 0 \} $. Its answer set $X_{9k+1}=A_6^{ts=k} \cup A_7^{ts=k} \cup A_9^{ts=k} \cup A_{10}^{ts=k} \cup A_{11}^{ts=k}$ -- using forced atom proposition and construction of $A_6, A_7, A_9, A_{10}, A_{11}$.

\item $eval_{U_{9k+1}}(bot_{U_{9k+2}}(\Pi^6) \setminus bot_{U_{9k+1}}(\Pi^6), X_0 \cup \dots \cup X_{9k+1}) = $ $\{ notenabled(t,k) \text{:-} . | $ $(\{ trans(t), $ $ptarc(p,t,n_c,c,k), $ $holds(p,q_c,c,k) \} \subseteq X_0 \cup \dots \cup X_{10k+1}, \text{~where~}  $ $q_c < n_c) \text{~or~} $ $(\{ notenabled(t,k) \text{:-} . | $ $(\{ trans(t), $ $iptarc(p,t,n2_c,c,k), $ $holds(p,q_c,c,k) \} \subseteq X_0 \cup \dots \cup X_{10k+1}, \text{~where~}  $ $q_c \geq n2_c \}) \text{~or~} $ $(\{ trans(t), $ $tptarc(p,t,n3_c,c,k), \\ holds(p,q_c,c,k) \} \subseteq X_0 \cup X_{9k+1}, \text{~where~} q_c < n3_c) \}$. Its answer set $X_{9k+2}=A_{12}^{ts=k}$ -- using  forced atom proposition and construction of $A_{12}$.
\begin{enumerate}
\item where, $q_c=m_{M_0(p)}(c)$, and $n_c=m_{W(p,t)}(c)$ for an arc $(p,t) \in E^-$ -- by construction of $i\ref{i:c:init}$ and $f\ref{f:c:ptarc}$ predicates in $\Pi^6$, and 
\item in an arc $(p,t) \in E^-$, $p \in \bullet t$ (by definition~\ref{def:pnpri:preset} of preset)
\item $n2_c=1$ -- by construction of $iptarc$ predicates in $\Pi^6$, meaning $q_c \geq n2_c \equiv q_c \geq 1 \equiv q_c > 0$,
\item $tptarc(p,t,n3_c,c,k)$ represents $n3_c=m_{QW(p,t)}(c)$, where $(p,t) \in Q$ 
\item thus, $notenabled(t,k) \in X_{9k+1}$ represents $\exists c \in C, (\exists p \in \bullet t : m_{M_0(p)}(c) < m_{W(p,t)}(c)) \vee (\exists p \in I(t) : m_{M_0(p)}(c) > k) \vee (\exists (p,t) \in Q : m_{M_0(p)}(c) < m_{QW(p,t)}(c))$.
\end{enumerate}

\item $eval_{U_{9k+2}}(bot_{U_{9k+3}}(\Pi^6) \setminus bot_{U_{9k+2}}(\Pi^6), X_0 \cup \dots \cup X_{9k+2}) = \{ enabled(t,k) \text{:-}. | trans(t) \in X_0 \cup \dots \cup X_{9k+2}, notenabled(t,k) \notin X_0 \cup \dots \cup X_{9k+2} \}$. Its answer set is $X_{9k+3} = A_{13}^{ts=k}$ -- using forced atom proposition and construction of $A_{13}$. \label{pnpri:x2a:k:enabled}
\begin{enumerate}
\item since an $enabled(t,k) \in X_{9k+3}$ if $\nexists ~notenabled(t,k) \in X_0 \cup \dots \cup X_{9k+2}$; which is equivalent to $\nexists t, \forall c \in C, (\nexists p \in \bullet t, m_{M_0(p)}(c) < m_{W(p,t)}(c)), (\nexists p \in I(t), m_{M_0(p)}(c) > k), (\nexists (p,t) \in Q : m_{M_0(p)}(c) < m_{QW(p,t)}(c) ), \forall c \in C, (\forall p \in \bullet t: m_{M_0(p)}(c) \geq m_{W(p,t)}(c)), (\forall p \in I(t) : m_{M_0(p)}(c) = k)$.
\end{enumerate}

\item $eval_{U_{9k+3}}(bot_{U_{9k+4}}(\Pi^6) \setminus bot_{U_{9k+3}}(\Pi^6), X_0 \cup \dots \cup X_{9k+3}) = \{ notprenabled(t,k) \text{:-}. | \\ \{ enabled(t,k), transpr(t,p), enabled(tt,k), transpr(tt,pp) \} \subseteq X_0 \cup \dots \cup X_{9k+3}, \\ pp < p \}$. Its answer set is $X_{9k+4} = A_{23}^{ts=k}$ -- using forced atom proposition and construction of $A_{23}$. \label{pnpri:x2a:k:notprenabled}
\begin{enumerate}
\item $enabled(t,k)$ represents $\exists t \in T, \forall c \in C, (\forall p \in \bullet t, m_{M_0(p)}(c) \geq m_{W(p,t)}(c)), \\ (\forall p \in I(t), m_{M_0(p)}(c) = k), (\forall (p,t) \in Q, m_{M_0(p)}(c) \geq m_{QW(p,t)}(c))$
\item $enabled(tt,k)$ represents $\exists tt \in T, \forall c \in C, (\forall pp \in \bullet tt, m_{M_0(pp)}(c) \geq \\ m_{W(pp,tt)}(c)) \wedge (\forall pp \in I(tt), m_{M_0(pp)}(c) = k), (\forall (pp,tt) \in Q, m_{M_0(pp)}(c) \geq m_{QW(pp,tt)}(c))$  
\item $transpr(t,p)$ represents $p=Z(t)$ -- by construction
\item $transpr(tt,pp)$ represents $pp=Z(tt)$ -- by construction
\item thus, $notprenabled(t,k)$ represents $\forall c \in C, (\forall p \in \bullet t, m_{M_0(p)}(c) \geq m_{W(p,t)}(c)) \\ \wedge (\forall p \in I(t), m_{M_0(p)}(c) = k), \exists tt \in T, (\forall pp \in \bullet tt, m_{M_0(pp)}(c) \geq m_{W(pp,tt)}(c)) \wedge (\forall pp \in I(tt), m_{M_0(pp)}(c) = k), Z(tt) < Z(t)$ 
\item which is equivalent to $(\forall p \in \bullet t: M_0(p) \geq W(p,t)) \wedge (\forall p \in I(t), M_0(p) = k), \exists tt \in T, (\forall pp \in \bullet tt, M_0(pp) \geq W(pp,tt)), (\forall pp \in I(tt), M_0(pp) = k), Z(tt) < Z(t)$ -- assuming multiset domain $C$ for all operations
\end{enumerate}

\item $eval_{U_{9k+4}}(bot_{U_{9k+5}}(\Pi^6) \setminus bot_{U_{9k+4}}(\Pi^6), X_0 \cup \dots \cup X_{9k+4}) = \{ prenabled(t,k) \text{:-}. | $ $enabled(t,k) \in X_0 \cup \dots \cup X_{9k+4}, notprenabled(t,k) \notin X_0 \cup \dots \cup X_{9k+4} \}$. Its answer set is $X_{9k+5} = A_{24}^{ts=k}$ -- using forced atom proposition and construction of $A_{24}$ \label{pnpri:x2a:k:prenabled}
\begin{enumerate}
\item $enabled(t,k)$ represents $\forall c \in C, (\forall p \in \bullet t, m_{M_0(p)}(c) \geq m_{W(p,t)}(c)), $ $(\forall p \in I(t), m_{M_0(p)}(c) = k), (\forall (p,t) \in Q, m_{M_0(p)}(c) \geq m_{QW(p,t)}(c))  \equiv (\forall p \in \bullet t, M_0(p) \geq W(p,t)), $ $(\forall p \in I(t), M_0(p) = k), (\forall (p,t) \in Q, M_0(p) \geq QW(p,t))$ -- from \ref{pnpri:x2a:k:enabled} above and assuming multiset domain $C$ for all operations
\item $notprenabled(t,k)$ represents $(\forall p \in \bullet t, M_0(p) \geq W(p,t)), (\forall p \in I(t), \\ M_0(p) = k), (\forall (p,t) \in Q, M_0(p) \geq QW(p,t)), \exists tt \in T, $ $(\forall pp \in \bullet tt, M_0(pp) \geq W(pp,tt)), $ $(\forall pp \in I(tt), M_0(pp) = k), (\forall (pp,tt) \in Q, M_0(pp) \geq W(pp,tt)), \\ Z(tt) < Z(t)$ -- from \ref{pnpri:x2a:k:notprenabled} above and assuming multiset domain $C$ for all operations
\item then, $prenabled(t,k)$ represents $(\forall p \in \bullet t, M_0(p) \geq W(p,t)), (\forall p \in I(t), \\ M_0(p) = k), (\forall (p,t) \in Q, M_0(p) \geq QW(p,t)), \nexists tt \in T, $ $((\forall pp \in \bullet tt, M_0(pp) \geq W(pp,tt)), $ $(\forall pp \in I(tt), M_0(pp) = k), (\forall (pp,tt) \in Q, M_0(pp) \geq W(pp,tt)), \\ Z(tt) < Z(t))$ -- from (a), (b) and $enabled(t,k) \in X_0 \cup \dots \cup X_{9k+4}$ 
\end{enumerate}

\item $eval_{U_{9k+5}}(bot_{U_{9k+6}}(\Pi^6) \setminus bot_{U_{9k+5}}(\Pi^6), X_0 \cup \dots \cup X_{9k+5}) = \{\{fires(t,k)\} \text{:-}. | \\ prenabled(t,k) \text{~holds in~} X_0 \cup \dots \cup X_{9k+5} \}$. It has multiple answer sets $X_{9k+6.1}, \dots, \\ X_{9k+6.n}$, corresponding to elements of power set of $fires(t,k)$ atoms in $eval_{U_{9k+5}}(...)$ -- using supported rule proposition. Since we are showing that the union of answer sets of $\Pi^6$ determined using splitting is equal to $A$, we only consider the set that matches the $fires(t,k)$ elements in $A$ and call it $X_{9k+6}$, ignoring the rest. Thus, $X_{9k+6} = A_{14}^{ts=k}$, representing $T_k$.
\begin{enumerate}
\item in addition, for every $t$ such that $prenabled(t,k) \in X_0 \cup \dots \cup X_{9k+5},  R(t) \neq \emptyset$; $fires(t,k) \in X_{9k+6}$ -- per definition~\ref{def:pnpri:firing_set} (firing set); requiring that a reset transition is fired when enabled
\item thus, the firing set $T_k$ will not be eliminated by the constraint $f\ref{f:c:pr:rptarc:elim}$ 
\end{enumerate}

\item $eval_{U_{9k+6}}(bot_{U_{9k+7}}(\Pi^6) \setminus bot_{U_{9k+6}}(\Pi^6), X_0 \cup \dots \cup X_{9k+6}) = \{add(p,n_c,t,c,k) \text{:-}. | \\ \{fires(t,k), tparc(t,p,n_c,c,k) \} \subseteq X_0 \cup \dots \cup X_{9k+6} \} \cup \{ del(p,n_c,t,c,k) \text{:-}. | \\ \{ fires(t,k), ptarc(p,t,n_c,c,k) \} \subseteq X_0 \cup \dots \cup X_{9k+6} \}$. It's answer set is $X_{9k+7} = A_{15}^{ts=k} \cup A_{16}^{ts=k}$ -- using forced atom proposition and definitions of $A_{15}$ and $A_{16}$. 
\begin{enumerate}
\item where, each $add$ atom is equivalent to $n_c=m_{W(t,p)}(c),c \in C, p \in t \bullet$, 
\item and each $del$ atom is equivalent to $n_c=m_{W(p,t)}(c), c \in C, p \in \bullet t$; or $n_c=m_{M_0(p)}(c), c \in C, p \in R(t)$,
\item representing the effect of transitions in $T_k$.
\end{enumerate}

\item $eval_{U_{9k+7}}(bot_{U_{9k+8}}(\Pi^6) \setminus bot_{U_{9k+7}}(\Pi^6), X_0 \cup \dots \cup X_{9k+7}) = \{tot\_incr(p,qq_c,c,k) \text{:-}. | $ $qq_c=\sum_{add(p,q_c,t,c,k) \in X_0 \cup \dots \cup X_{10k+7}}{q_c} \} \cup \{ tot\_decr(p,qq_c,c,k) \text{:-}. | qq_c= \\ \sum_{del(p,q_c,t,c,k) \in X_0 \cup \dots \cup X_{9k+7}}{q_c} \}$. It's answer set is $X_{10k+8} = A_{17}^{ts=k} \cup A_{18}^{ts=k}$ --  using forced atom proposition and definitions of $A_{17}$ and $A_{18}$.
\begin{enumerate}
\item where, each $tot\_incr(p,qq_c,c,k)$, $qq_c=\sum_{add(p,q_c,t,c,k) \in X_0 \cup \dots X_{10k+7}}{q_c}$ \\ $\equiv qq_c=\sum_{t \in X_{9k+6}, p \in t \bullet}{m_{W(p,t)}(c)}$, 
\item and each $tot\_decr(p,qq_c,c,k)$, $qq_c=\sum_{del(p,q_c,t,c,k) \in X_0 \cup \dots X_{10k+7}}{q_c}$ \\ $\equiv qq=\sum_{t \in X_{10k+6}, p \in \bullet t}{m_{W(t,p)}(c)} + \sum_{t \in X_{9k+6}, p \in R(t)}{m_{M_0(p)}(c)}$, 
\item represent the net effect of transitions in $T_k$
\end{enumerate}
\item $eval_{U_{9k+8}}(bot_{U_{9k+9}}(\Pi^6) \setminus bot_{U_{9k+8}}(\Pi^6), X_0 \cup \dots \cup X_{9k+8}) = \{ consumesmore(p,k) \text{:-}. | \\ \{holds(p,q_c,c,k), tot\_decr(p,q1_c,c,k) \} \subseteq X_0 \cup \dots \cup X_{9k+8}, q1_c > q_c \} \cup \\ \{ holds(p,q_c,c,1) \text{:-}. | \{ holds(p,q1_c,c,k), tot\_incr(p,q2_c,c,k), tot\_decr(p,q3_c,c,k) \} \\ \subseteq $ $X_0 \cup \dots \cup X_{9k+6}, q_c=q1_c+q2_c-q3_c \}  \cup \{ could\_not\_have(t,k) \text{:-}. | \{ prenabled(t,k), \\ ptarc(s,t,q,c,k), holds(s,qq,c,k), tot\_decr(s,qqq,c,k) \} \subseteq X_0 \cup \dots \cup X_{9k+8}, \\ fires(t,k) \notin X_0 \cup \dots \cup X_{9k+8}, q > qq - qqq \}$. It's answer set is $X_{10k+9} = A_{19}^{ts=k} \cup A_{21}^{ts=k} \cup A_{25}^{ts=k}$  -- using forced atom proposition and definitions of $A_{19}, A_{21}, A_{25}$.
\begin{enumerate}
\item where, $consumesmore(p,k)$ represents $\exists p : q_c=m_{M_0(p)}(c), \\ q1_c=$ $\sum_{t \in T_k, p \in \bullet t}{m_{W(p,t)}(c)}+\sum_{t \in T_k, p \in R(t)}{m_{M_0(p)}(c)}, q1_c > q_c, c \in C$, indicating place $p$ that will be over consumed if $T_k$ is fired, as defined in definition~\ref{def:pnpri:conflict} (conflicting transition)
\item $holds(p,q_c,c,k+1)$ if $q_c=m_{M_0(p)}(c)+\sum_{t \in T_k, p \in t \bullet}{m_{W(t,p)}(c)}- \\ (\sum_{t \in T_k, p \in \bullet t}{m_{W(p,t)}(c)}+ \sum_{t \in T_k, p \in R(t)}{m_{M_0(p)}(c)})$ represented by $q_c=m_{M_1(p)}(c)$ for some $c \in C$ -- by construction of $\Pi^6$
\item $could\_not\_have(t,k)$ if
\begin{enumerate}
\item $(\forall p \in \bullet t, W(p,t) \leq M_0(p)), (\forall p \in I(t), M_0(p) = k), (\forall (p,t) \in Q, $ $M_0(p) \geq WQ(p,t)), \nexists tt \in T, (\forall pp \in \bullet tt, W(pp,tt) \leq M_{ts}(pp)), (\forall pp \in I(tt), M_0(pp) = k), (\forall (pp,tt) \in Q, $ $M_0(pp) \geq QW(pp,tt)), Z(tt) < Z(t)$,
\item and $q_c > m_{M_0(s)}(c) - (\sum_{t' \in T_k, s \in \bullet t'}{m_{W(s,t')}(c)}+ \sum_{t' \in T_k, s \in R(t)}{m_{M_0(s)}(c)}), $ $q_c = m_{W(s,t)}(c) \text{~if~} s \in \bullet t \text{~or~} m_{M_0(s)}(c) \text{~otherwise}$ for some $c \in C$, which becomes $q > M_0(s) - (\sum_{t' \in T_k, s \in \bullet t'}{W(s,t')}+ \sum_{t' \in T_k, s \in R(t)}{M_0(s)}), q = W(s,t) \text{~if~} s \in \bullet t \text{~or~} M_0(s) \text{~otherwise}$ for all $c \in C$
\item (i), (ii) above combined match the definition of $A_{25}$
\end{enumerate}
\item $X_{9k+9}$ does not contain $could\_not\_have(t,k)$, when $prenabled(t,k) \in X_0 \cup \dots \cup X_{9k+6}$ and $fires(t,k) \notin X_0 \cup \dots \cup X_{9k+5}$due to construction of $A$, encoding of $a\ref{a:c:maxfire:cnh}$ and its body atoms. As a result it is not eliminated by the constraint $a\ref{a:c:maxfire:elim}$
\end{enumerate}

\item $eval_{U_{9k+9}}(bot_{U_{9k+10}}(\Pi^6) \setminus bot_{U_{9k+9}}(\Pi^6), X_0 \cup \dots \cup X_{9k+9}) = \{ consumesmore \text{:-}. | \\ \{ consumesmore(p,k) \} \subseteq A \}$. It's answer set is $X_{9k+10} = A_{20}$ -- using forced atom proposition and the definition of $A_{20}$
\begin{enumerate}
\item $X_{9k+10}$ will be empty since none of $consumesmore(p,0),\dots, \\ consumesmore(p,k)$ hold in $X_0 \cup \dots \cup X_{9k+9}$ due to the construction of $A$, encoding of $a\ref{a:c:overc:place}$ and its body atoms. As a result, it is not eliminated by the constraint $a\ref{a:c:overc:elim}$
\end{enumerate}

\end{enumerate}

The set $X = X_0 \cup \dots \cup X_{9k+10}$ is the answer set of $\Pi^6$ by the splitting sequence theorem~\ref{def:split_seq_thm}. Each $X_i, 0 \leq i \leq 9k+10$ matches a distinct portion of $A$, and $X = A$, thus $A$ is an answer set of $\Pi^6$.

\vspace{30pt}
\noindent
{\bf Next we show (\ref{prove:a2x:pnpri}):} Given $\Pi^6$ be the encoding of a Petri Net $PN(P,T,E,C,W,R,I,$ $Q,QW,Z)$ with initial marking $M_0$, and $A$ be an answer set of $\Pi^6$ that satisfies (\ref{eqn:pnpri:fires}) and (\ref{eqn:pnpri:holds}), then we can construct $X=M_0,T_k,\dots,M_k,T_k,M_{k+1}$ from $A$, such that it is an execution sequence of $PN$.

We construct the $X$ as follows:
\begin{enumerate}
\item $M_i = (M_i(p_0), \dots, M_i(p_n))$, where $\{ holds(p_0,m_{M_i(p_0)}(c),c,i), \dots,\\ holds(p_n,m_{M_i(p_n)}(c),c,i) \} \subseteq A$, for $c \in C, 0 \leq i \leq k+1$
\item $T_i = \{ t : fires(t,i) \in A\}$, for $0 \leq i \leq k$ 
\end{enumerate}
and show that $X$ is indeed an execution sequence of $PN$. We show this by induction over $k$ (i.e. given $M_k$, $T_k$ is a valid firing set and its firing produces marking $M_{k+1}$).

\vspace{20pt}

\noindent
{\bf Base case:} Let $k=0$, and $M_0$ is a valid marking in $X$ for $PN$, show
\begin{inparaenum}[(1)]
\item $T_0$ is a valid firing set for $M_0$, and 
\item firing $T_0$ in $M_0$ produces marking $M_1$.
\end{inparaenum}

\begin{enumerate}
\item We show $T_0$ is a valid firing set for $M_0$. Let $\{ fires(t_0,0), \dots, fires(t_x,0) \}$ be the set of all $fires(\dots,0)$ atoms in $A$,\label{pnpri:prove:fires_t0}

\begin{enumerate}
\item Then for each $fires(t_i,0) \in A$

\begin{enumerate}
\item $prenabled(t_i,0) \in A$ -- from rule $a\ref{a:c:prfires}$ and supported rule proposition
\item Then $enabled(t_i,0) \in A$ -- from rule $a\ref{a:c:prenabled}$ and supported rule proposition
\item And $notprenabled(t_i,0) \notin A$ -- from rule $a\ref{a:c:prenabled}$ and supported rule proposition

\item For $enabled(t_i,0) \in A$
\begin{enumerate}
\item $notenabled(t_i,0) \notin A$ -- from rule $e\ref{e:c:enabled}$ and supported rule proposition
\item Then either of $body(e\ref{e:c:ne:ptarc})$, $body(e\ref{e:c:ne:iptarc})$, or $body(e\ref{e:c:ne:tptarc})$ must not hold in $A$ for $t_i$ -- from rules $e\ref{e:c:ne:ptarc},e\ref{e:c:ne:iptarc},e\ref{e:c:ne:tptarc}$ and forced atom proposition
\item Then $q_c \not< {n_i}_c \equiv q_c \geq {n_i}_c$ in $e\ref{e:c:ne:ptarc}$ for all $\{holds(p,q_c,c,0), \\ ptarc(p,t_i,{n_i}_c,c,0)\} \subseteq A$ -- from $e:\ref{e:c:ne:ptarc}$, forced atom proposition, and given facts ($holds(p,q_c,c,0) \in A, ptarc(p,t_i,{n_i}_c,0) \in A$)
\item And $q_c \not\geq {n_i}_c \equiv q_c < {n_i}_c$ in $e\ref{e:c:ne:iptarc}$ for all $\{ holds(p,q_c,c,0), \\ iptarc(p,t_i,{n_i}_c,c,0) \} \subseteq A, {n_i}_c=1$; $q_c > {n_i}_c \equiv q_c = 0$ -- from $e\ref{e:c:ne:iptarc}$, forced atom proposition, given facts ($holds(p,q_c,c,0) \in A, \\  iptarc(p,t_i,1,c,0) \in A$), and $q_c$ is a positive integer
\item And $q_c \not< {n_i}_c \equiv q_c \geq {n_i}_c$ in $e\ref{e:c:ne:tptarc}$ for all $\{ holds(p,q_c,c,0), \\ tptarc(p,t_i,{n_i}_c,c,0) \} \subseteq A$ -- from $e\ref{e:c:ne:tptarc}$, forced atom proposition, and given facts
\item Then $\forall c \in C, (\forall p \in \bullet t_i, m_{M_0(p)}(c) \geq m_{W(p,t_i)}(c)) \wedge (\forall p \in I(t_i), \\ m_{M_0(p)}(c) = 0) \wedge (\forall (p,t_i) \in Q, m_{M_0(p)}(c) \geq m_{QW(p,t_i)}(c))$ -- from $i\ref{i:c:init},f\ref{f:c:ptarc},f\ref{f:c:iptarc},f\ref{f:c:tptarc}$ construction, definition~\ref{def:pnpri:preset} of preset $\bullet t_i$ in $PN$, definition~\ref{def:pnpri:enable} of enabled transition in $PN$, and that the construction of reset arcs by $f\ref{f:c:rptarc}$ ensures $notenabled(t,0)$ is never true for a reset arc, where $holds(p,q_c,c,0) \in A$ represents $q_c=m_{M_0(p)}(c)$, $ptarc(p,t_i,{n_i}_c,0) \in A$ represents ${n_i}_c=m_{W(p,t_i)}(c)$, ${n_i}_c = m_{M_0(p)}(c)$.
\item Which is equivalent to $(\forall p \in \bullet t_i, M_0(p) \geq W(p,t_i)) \wedge (\forall p \in I(t_i), M_0(p) = 0) \wedge (\forall (p,t_i) \in Q, M_0(p) \geq QW(p,t_i))$ -- assuming multiset domain $C$
\end{enumerate}

\item For $notprenabled(t_i,0) \notin A$
\begin{enumerate}
\item Either $(\nexists enabled(tt,0) \in A : pp < p_i)$ or $(\forall enabled(tt,0) \in A : pp \not< p_i)$ where $pp = Z(tt), p_i = Z(t_i)$ -- from rule $a\ref{a:c:prne},f\ref{f:c:pr}$ and forced atom proposition
\item This matches the definition of an enabled priority transition
\end{enumerate}

\item Then $t_i$ is enabled and can fire in $PN$, as a result it can belong to $T_0$ -- from definition~\ref{def:pnpri:enable} of enabled transition

\end{enumerate}
\item And $consumesmore \notin A$, since $A$ is an answer set of $\Pi^6$ -- from rule $a\ref{a:c:overc:gen}$ and supported rule proposition
\begin{enumerate}
\item Then $\nexists consumesmore(p,0) \in A$ -- from rule $a\ref{a:c:overc:place}$ and supported rule proposition
\item  Then $\nexists \{ holds(p,q_c,c,0), tot\_decr(p,q1_c,c,0) \} \subseteq A, q1_c>q_c$ in $body(a\ref{a:c:overc:place})$ -- from $a\ref{a:c:overc:place}$ and forced atom proposition
\item Then $\nexists c \in C \nexists p \in P, (\sum_{t_i \in \{t_0,\dots,t_x\}, p \in \bullet t_i}{m_{W(p,t_i)}(c)}+ \\ \sum_{t_i \in \{t_0,\dots,t_x\}, p \in R(t_i)}{m_{M_0(p)}(c)}) > m_{M_0(p)}(c)$ -- from the following
\begin{enumerate}
\item $holds(p,q_c,c,0)$ represents $q_c=m_{M_0(p)}(c)$ -- from rule $i\ref{i:c:init}$ encoding, given
\item $tot\_decr(p,q1_c,c,0) \in A$ if $\{ del(p,{q1_0}_c,t_0,c,0), \dots, \\ del(p,{q1_x}_c,t_x,c,0) \} \subseteq A$, where $q1_c = {q1_0}_c+\dots+{q1_x}_c$ -- from $r\ref{r:c:totdecr}$ and forced atom proposition
\item $del(p,{q1_i}_c,t_i,c,0) \in A$ if $\{ fires(t_i,0), ptarc(p,t_i,{q1_i}_c,c,0) \} \subseteq A$ -- from $r\ref{r:c:del}$ and supported rule proposition
\item $del(p,{q1_i}_c,t_i,c,0)$ either represents removal of ${q1_i}_c = m_{W(p,t_i)}(c)$ tokens from $p \in \bullet t_i$; or it represents removal of ${q1_i}_c = m_{M_0(p)}(c)$ tokens from $p \in R(t_i)$-- from rules $r\ref{r:c:del},f\ref{f:c:ptarc},f\ref{f:c:rptarc}$, supported rule proposition, and definition~\ref{def:pnpri:texec} of transition execution in $PN$
\end{enumerate}
\item Then the set of transitions in $T_0$ do not conflict -- by the definition~\ref{def:pnpri:conflict} of conflicting transitions
\end{enumerate}

\item And for each $prenabled(t_j,0) \in A$ and $fires(t_j,0) \notin A$, $could\_not\_have(t_j,0) \in A$, since $A$ is an answer set of $\Pi^6$ - from rule $a\ref{a:c:prmaxfire:elim}$ and supported rule proposition
\begin{enumerate}
\item Then $\{ prenabled(t_j,0), holds(s,qq_c,c,0), ptarc(s,t_j,q_c,c,0), \\ tot\_decr(s,qqq_c,c,0) \} \subseteq A$, such that $q_c > qq_c - qqq_c$ and $fires(t_j,0) \notin A$ - from rule $a\ref{a:c:prmaxfire:cnh}$ and supported rule proposition
\item Then for an $s \in \bullet t_j \cup R(t_j)$, $q_c > m_{M_0(s)}(c) - (\sum_{t_i \in T_0, s \in \bullet t_i}{m_{W(s,t_i)}(c)} + \sum_{t_i \in T_0, s \in R(t_i)}{m_{M_0(s)}(c))}$, where $q_c=m_{W(s,t_j)}(c) \text{~if~} s \in \bullet t_j, \text{~or~} m_{M_0(s)}(c)$ $ \text{~otherwise}$ -- from the following %
\begin{enumerate}
\item $ptarc(s,t_i,q_c,c,0)$ represents $q_c=m_{W(s,t_i)}(c)$ if $(s,t_i) \in E^-$ or $q_c=m_{M_0(s)}(c)$ if $s \in R(t_i)$ -- from rule $f\ref{f:c:ptarc},f\ref{f:c:rptarc}$ construction
\item $holds(s,qq_c,c,0)$ represents $qq_c=m_{M_0(s)}(c)$ -- from $i\ref{i:c:init}$ construction
\item $tot\_decr(s,qqq_c,c,0) \in A$ if $\{ del(s,{qqq_0}_c,t_0,c,0), \dots, \\ del(s,{qqq_x}_c,t_x,c,0) \} \subseteq A$ -- from rule $r\ref{r:c:totdecr}$ construction and supported rule proposition
\item $del(s,{qqq_i}_c,t_i,c,0) \in A$ if $\{ fires(t_i,0), ptarc(s,t_i,{qqq_i}_c,c,0) \} \subseteq A$ -- from rule $r\ref{r:c:del}$ and supported rule proposition
\item $del(s,{qqq_i}_c,t_i,c,0)$ either represents ${qqq_i}_c = m_{W(s,t_i)}(c) : t_i \in T_0, (s,t_i) \in E^-$, or ${qqq_i}_c = m_{M_0(t_i)}(c) : t_i \in T_0, s \in R(t_i)$ -- from rule $f\ref{f:c:ptarc},f\ref{f:c:rptarc}$ construction 
\item $tot\_decr(q,qqq_c,c,0)$ represents $\sum_{t_i \in T_0, s \in \bullet t_i}{m_{W(s,t_i)}(c)} + \\ \sum_{t_i \in T_0, s \in R(t_i)}{m_{M_0(s)}(c)}$ -- from (C,D,E) above 
\end{enumerate}

\item Then firing $T_0 \cup \{ t_j \}$ would have required more tokens than are present at its source place $s \in \bullet t_j \cup R(t_j)$. Thus, $T_0$ is a maximal set of transitions that can simultaneously fire.
\end{enumerate}

\item And for each reset transition $t_r$ with $prenabled(t_r,0) \in A$, $fires(t_r,0) \in A$, since $A$ is an answer set of $\Pi^6$ - from rule $f\ref{f:c:pr:rptarc:elim}$ and supported rule proposition
\begin{enumerate}
\item Then, the firing set $T_0$ satisfies the reset-transition requirement of definition~\ref{def:pnpri:firing_set} (firing set)
\end{enumerate}

\item Then $\{t_0, \dots, t_x\} = T_0$ -- using 1(a),1(b),1(d) above; and using 1(c) it is a maximal firing set  
\end{enumerate}

\item We show $M_1$ is produced by firing $T_0$ in $M_0$. Let $holds(p,q_c,c,1) \in A$
\begin{enumerate}
\item Then $\{ holds(p,q1_c,c,0), tot\_incr(p,q2_c,c,0), tot\_decr(p,q3_c,c,0) \} \subseteq A : q_c=q1_c+q2_c-q3_c$ -- from rule $r\ref{r:c:nextstate}$ and supported rule proposition \label{pnpri:x:1:base}
\item \label{pnpri:x:2:base} Then, $holds(p,q1_c,c,0) \in A$ represents $q1_c=m_{M_0(p)}(c)$ -- given, construction;  

and $\{add(p,{q2_0}_c,t_0,c,0), \dots, $ $add(p,{q2_j}_c,t_j,c,0)\} \subseteq A : {q2_0}_c + \dots + {q2_j}_c = q2_c$  \label{pnpri:stmt:add:base} and $\{del(p,{q3_0}_c,t_0,c,0), \dots, $ $del(p,{q3_l}_c,t_l,c,0)\} \subseteq A : {q3_0}_c + \dots + {q3_l}_c = q3_c$ \label{pnpri:stmt:del:base}  -- rules $r\ref{r:c:totincr},r\ref{r:c:totdecr}$ and supported rule proposition, respectively
\item Then $\{ fires(t_0,0), \dots, fires(t_j,0) \} \subseteq A$ and $\{ fires(t_0,0), \dots, fires(t_l,0) \} \subseteq A$ -- rules $r\ref{r:c:add},r\ref{r:c:del}$ and supported rule proposition, respectively
\item Then $\{ fires(t_0,0), \dots, $ $fires(t_j,0) \} \cup $ $\{ fires(t_0,0), \dots, $ $fires(t_l,0) \} \subseteq $ $A = \{ fires(t_0,0), \dots, $ $fires(t_x,0) \} \subseteq A$ -- set union of subsets
\item Then for each $fires(t_x,0) \in A$ we have $t_x \in T_0$ -- already shown in item~\ref{pnpri:prove:fires_t0} above
\item Then $q_c = m_{M_0(p)}(c) + \sum_{t_x \in T_0 \wedge p \in t_x \bullet}{m_{W(t_x,p)}(c)} - (\sum_{t_x \in T_0 \wedge p \in \bullet t_x}{m_{W(p,t_x)}(c)} + \sum_{t_x \in T_0 \wedge p \in R(t_x)}{m_{M_0(p)}(c)})$ -- from 
\eqref{pnpri:x:2:base} above and the following
\begin{enumerate}
\item Each $add(p,{q_j}_c,t_j,c,0) \in A$ represents ${q_j}_c=m_{W(t_j,p)}(c)$ for $p \in t_j \bullet$ -- rule $r\ref{r:c:add},f\ref{f:c:tparc}$ encoding, and definition~\ref{def:pnpri:texec} of transition execution in $PN$ %
\item Each $del(p,t_y,{q_y}_c,c,0) \in A$ represents either ${q_y}_c=m_{W(p,t_y)}(c)$ for $p \in \bullet t_y$, or ${q_y}_c=m_{M_0(p)}(c)$ for $p \in R(t_y)$ -- from rule $r\ref{r:c:del},f\ref{f:c:ptarc}$ encoding and definition~\ref{def:pnpri:texec} of transition execution in $PN$; or from rule $r\ref{r:c:del},f\ref{f:c:rptarc}$ encoding and definition of reset arc in $PN$
\item Each $tot\_incr(p,q2_c,c,0) \in A$ represents $q2_c=\sum_{t_x \in T_0 \wedge p \in t_x  \bullet}{m_{W(t_x,p)}(c)}$ -- aggregate assignment atom semantics in rule $r\ref{r:c:totincr}$
\item Each $tot\_decr(p,q3_c,c,0) \in A$ represents $q3_c=\sum_{t_x \in T_0 \wedge p \in \bullet t_x}{m_{W(p,t_x)}(c)} + $ $\sum_{t_x \in T_0 \wedge p \in R(t_x)}{m_{M_0(p)}(c)}$ -- aggregate assignment atom semantics in rule $r\ref{r:c:totdecr}$
\end{enumerate}
\item Then, $m_{M_1(p)}(c) = q_c$ -- since $holds(p,q_c,c,1) \in A$ encodes $q_c=m_{M_1(p)}(c)$ -- from construction
\end{enumerate}
\end{enumerate}

\noindent
{\bf Inductive Step:} Let $k > 0$, and $M_k$ is a valid marking in $X$ for $PN$, show 
\begin{inparaenum}[(1)]
\item $T_k$ is a valid firing set for $M_k$, and 
\item firing $T_k$ in $M_k$ produces marking $M_{k+1}$.
\end{inparaenum}

\begin{enumerate}
\item We show that $T_k$ is a valid firing set in $M_k$. Let $\{ fires(T_k,k), \dots, fires(t_x,k) \}$ be the set of all $fires(\dots,k)$ atoms in $A$,\label{pnpri:prove:fires_tk}

\begin{enumerate}
\item Then for each $fires(t_i,k) \in A$

\begin{enumerate}
\item $prenabled(t_i,k) \in A$ -- from rule $a\ref{a:c:prfires}$ and supported rule proposition
\item Then $enabled(t_i,k) \in A$ -- from rule $a\ref{a:c:prenabled}$ and supported rule proposition
\item And $notprenabled(t_i,k) \notin A$ -- from rule $a\ref{a:c:prenabled}$ and supported rule proposition

\item For $enabled(t_i,k) \in A$
\begin{enumerate}
\item $notenabled(t_i,k) \notin A$ -- from rule $e\ref{e:c:enabled}$ and supported rule proposition
\item Then either of $body(e\ref{e:c:ne:ptarc})$, $body(e\ref{e:c:ne:iptarc})$, or $body(e\ref{e:c:ne:tptarc})$ must not hold in $A$ for $t_i$ -- from rules $e\ref{e:c:ne:ptarc},e\ref{e:c:ne:iptarc},e\ref{e:c:ne:tptarc}$ and forced atom proposition
\item Then $q_c \not< {n_i}_c \equiv q_c \geq {n_i}_c$ in $e\ref{e:c:ne:ptarc}$ for all $\{holds(p,q_c,c,k), \\ ptarc(p,t_i,{n_i}_c,c,k)\} \subseteq A$ -- from $e\ref{e:c:ne:ptarc}$, forced atom proposition, and given facts ($holds(p,q_c,c,k) \in A, ptarc(p,t_i,{n_i}_c,k) \in A$)
\item And $q_c \not\geq {n_i}_c \equiv q_c < {n_i}_c$ in $e\ref{e:c:ne:iptarc}$ for all $\{ holds(p,q_c,c,k), \\ iptarc(p,t_i,{n_i}_c,c,k) \} \subseteq A, {n_i}_c=1$; $q_c > {n_i}_c \equiv q_c = 0$ -- from $e\ref{e:c:ne:iptarc}$, forced atom proposition, given facts ($holds(p,q_c,c,k) \in A, \\ iptarc(p,t_i,1,c,k) \in A$), and $q_c$ is a positive integer
\item And $q_c \not< {n_i}_c \equiv q_c \geq {n_i}_c$ in $e\ref{e:c:ne:tptarc}$ for all $\{ holds(p,q_c,c,k), \\ tptarc(p,t_i,{n_i}_c,c,k) \} \subseteq A$ -- from $e\ref{e:c:ne:tptarc}$, forced atom proposition, and given facts
\item Then $\forall c \in C, (\forall p \in \bullet t_i, m_{M_k(p)}(c) \geq m_{W(p,t_i)}(c)) \wedge (\forall p \in I(t_i), $ $m_{M_k(p)}(c) = k) \wedge (\forall (p,t_i) \in Q, m_{M_k(p)}(c) \geq m_{QW(p,t_i)}(c))$ -- from $f\ref{f:c:ptarc},f\ref{f:c:iptarc},f\ref{f:c:tptarc}$ construction, inductive assumption, definition~\ref{def:pnpri:preset} of preset $\bullet t_i$ in $PN$, definition~\ref{def:pnpri:enable} of enabled transition in $PN$, and that the construction of reset arcs by $f\ref{f:c:rptarc}$ ensures $notenabled(t,k)$ is never true for a reset arc, where $holds(p,q_c,c,k) \in A$ represents $q_c=m_{M_k(p)}(c)$, $ptarc(p,t_i,{n_i}_c,k) \in A$ represents ${n_i}_c=m_{W(p,t_i)}(c)$, ${n_i}_c = m_{M_k(p)}(c)$.
\item Which is equivalent to $(\forall p \in \bullet t_i, M_k(p) \geq W(p,t_i)) \wedge (\forall p \in I(t_i), M_k(p) = k) \wedge (\forall (p,t_i) \in Q, M_k(p) \geq QW(p,t_i))$ -- assuming multiset domain $C$
\end{enumerate}

\item For $notprenabled(t_i,k) \notin A$
\begin{enumerate}
\item Either $(\nexists enabled(tt,k) \in A : pp < p_i)$ or $(\forall enabled(tt,k) \in A : pp \not< p_i)$ where $pp = Z(tt), p_i = Z(t_i)$ -- from rule $a\ref{a:c:prne}, f\ref{f:c:pr}$ and forced atom proposition
\item This matches the definition of an enabled priority transition
\end{enumerate}

\item Then $t_i$ is enabled and can fire in $PN$, as a result it can belong to $T_k$ -- from definition~\ref{def:pnpri:enable} of enabled transition

\end{enumerate}
\item And $consumesmore \notin A$, since $A$ is an answer set of $\Pi^6$ -- from rule $a\ref{a:c:overc:elim}$ and supported rule proposition
\begin{enumerate}
\item Then $\nexists consumesmore(p,k) \in A$ -- from rule $a\ref{a:c:overc:elim}$ and supported rule proposition
\item  Then $\nexists \{ holds(p,q_c,c,k), tot\_decr(p,q1_c,c,k) \} \subseteq A, q1_c>q_c$ in $body(a\ref{a:c:overc:place})$ -- from $a\ref{a:c:overc:place}$ and forced atom proposition
\item Then $\nexists c \in C \nexists p \in P, (\sum_{t_i \in \{T_k,\dots,t_x\}, p \in \bullet t_i}{m_{W(p,t_i)}(c)}+ \\ \sum_{t_i \in \{T_k,\dots,t_x\}, p \in R(t_i)}{m_{M_k(p)}(c)}) > m_{M_k(p)}(c)$ -- from the following
\begin{enumerate}
\item $holds(p,q_c,c,k)$ represents $q_c=m_{M_k(p)}(c)$ -- from inductive assumption and construction, given
\item $tot\_decr(p,q1_c,c,k) \in A$ if $\{ del(p,{q1_0}_c,T_k,c,k), \dots, \\ del(p,{q1_x}_c,t_x,c,k) \} \subseteq A$, where $q1_c = {q1_0}_c+\dots+{q1_x}_c$ -- from $r\ref{r:c:totdecr}$ and forced atom proposition
\item $del(p,{q1_i}_c,t_i,c,k) \in A$ if $\{ fires(t_i,k), ptarc(p,t_i,{q1_i}_c,c,k) \} \subseteq A$ -- from $r\ref{r:c:del}$ and supported rule proposition
\item $del(p,{q1_i}_c,t_i,c,k)$ either represents removal of ${q1_i}_c = m_{W(p,t_i)}(c)$ tokens from $p \in \bullet t_i$; or it represents removal of ${q1_i}_c = m_{M_k(p)}(c)$ tokens from $p \in R(t_i)$-- from rules $r\ref{r:c:del},f\ref{f:c:ptarc},f\ref{f:c:rptarc}$, supported rule proposition, and definition~\ref{def:pnpri:texec} of transition execution in $PN$
\end{enumerate}
\item Then the set of transitions in $T_k$ do not conflict -- by the definition~\ref{def:pnpri:conflict} of conflicting transitions
\end{enumerate}

\item And for each $prenabled(t_j,k) \in A$ and $fires(t_j,k) \notin A$, $could\_not\_have(t_j,k) \in A$, since $A$ is an answer set of $\Pi^6$ - from rule $a\ref{a:c:prmaxfire:elim}$ and supported rule proposition
\begin{enumerate}
\item Then $\{ prenabled(t_j,k), holds(s,qq_c,c,k), ptarc(s,t_j,q_c,c,k), \\ tot\_decr(s,qqq_c,c,k) \} \subseteq A$, such that $q_c > qq_c - qqq_c$ and $fires(t_j,k) \notin A$ - from rule $a\ref{a:c:prmaxfire:cnh}$ and supported rule proposition
\item Then for an $s \in \bullet t_j \cup R(t_j)$, $q_c > m_{M_k(s)}(c) - (\sum_{t_i \in T_k, s \in \bullet t_i}{m_{W(s,t_i)}(c)} + \sum_{t_i \in T_k, s \in R(t_i)}{m_{M_k(s)}(c))}$, where $q_c=m_{W(s,t_j)}(c) \text{~if~} s \in \bullet t_j, \text{~or~} m_{M_k(s)}(c) $ $\text{~otherwise}$. %
\begin{enumerate}
\item $ptarc(s,t_i,q_c,c,k)$ represents $q_c=m_{W(s,t_i)}(c)$ if $(s,t_i) \in E^-$ or $q_c=m_{M_k(s)}(c)$ if $s \in R(t_i)$ -- from rule $f\ref{f:c:ptarc},f\ref{f:c:rptarc}$ construction
\item $holds(s,qq_c,c,k)$ represents $qq_c=m_{M_k(s)}(c)$ -- from inductive assumption and construction
\item $tot\_decr(s,qqq_c,c,k) \in A$ if $\{ del(s,{qqq_0}_c,T_k,c,k), \dots, \\ del(s,{qqq_x}_c,t_x,c,k) \} \subseteq A$ -- from rule $r\ref{r:c:totdecr}$ construction and supported rule proposition
\item $del(s,{qqq_i}_c,t_i,c,k) \in A$ if $\{ fires(t_i,k), ptarc(s,t_i,{qqq_i}_c,c,k) \} \subseteq A$ -- from rule $r\ref{r:c:del}$ and supported rule proposition
\item $del(s,{qqq_i}_c,t_i,c,k)$ either represents ${qqq_i}_c = m_{W(s,t_i)}(c) : t_i \in T_k, (s,t_i) \in E^-$, or ${qqq_i}_c = m_{M_k(t_i)}(c) : t_i \in T_k, s \in R(t_i)$ -- from rule $f\ref{f:c:ptarc},f\ref{f:c:rptarc}$ construction 
\item $tot\_decr(q,qqq_c,c,k)$ represents $\sum_{t_i \in T_k, s \in \bullet t_i}{m_{W(s,t_i)}(c)} + \\ \sum_{t_i \in T_k, s \in R(t_i)}{m_{M_k(s)}(c)}$ -- from (C,D,E) above 
\end{enumerate}

\item Then firing $T_k \cup \{ t_j \}$ would have required more tokens than are present at its source place $s \in \bullet t_j \cup R(t_j)$. Thus, $T_k$ is a maximal set of transitions that can simultaneously fire.
\end{enumerate}

\item And for each reset transition $t_r$ with $prenabled(t_r,k) \in A$, $fires(t_r,k) \in A$, since $A$ is an answer set of $\Pi^6$ - from rule $f\ref{f:c:pr:rptarc:elim}$ and supported rule proposition
\begin{enumerate}
\item Then the firing set $T_k$ satisfies the reset transition requirement of definition~\ref{def:pnpri:firing_set} (firing set)
\end{enumerate}

\item Then $\{t_0, \dots, t_x\} = T_k$ -- using 1(a),1(b), 1(d) above; and using 1(c) it is a maximal firing set  
\end{enumerate}

\item We show that $M_{k+1}$ is produced by firing $T_k$ in $M_k$. Let $holds(p,q_c,c,k+1) \in A$
\begin{enumerate}
\item Then $\{ holds(p,q1_c,c,k), tot\_incr(p,q2_c,c,k), tot\_decr(p,q3_c,c,k) \} \subseteq A : q_c=q1_c+q2_c-q3_c$ -- from rule $r\ref{r:c:nextstate}$ and supported rule proposition \label{pnpri:x:1:induction}
\item \label{pnpri:x:2:induction} Then, $holds(p,q1_c,c,k) \in A$ represents $q1_c=m_{M_k(p)}(c)$ -- inductive assumption; 

and $\{add(p,{q2_0}_c,T_k,c,k), \dots, $ $add(p,{q2_j}_c,t_j,c,k)\} \subseteq A : {q2_0}_c + \dots + {q2_j}_c = q2_c$  \label{pnpri:stmt:add:induction} and $\{del(p,{q3_0}_c,T_k,c,k), \dots, $ $del(p,{q3_l}_c,t_l,c,k)\} \subseteq A : {q3_0}_c + \dots + {q3_l}_c = q3_c$ \label{pnpri:stmt:del:induction}  -- rules $r\ref{r:c:totincr},r\ref{r:c:totdecr}$ and supported rule proposition, respectively
\item Then $\{ fires(T_k,k), \dots, fires(t_j,k) \} \subseteq A$ and $\{ fires(T_k,k), \dots, \\ fires(t_l,k) \} \subseteq A$ -- rules $r\ref{r:c:add},r\ref{r:c:del}$ and supported rule proposition, respectively
\item Then $\{ fires(T_k,k), \dots, $ $fires(t_j,k) \} \cup $ $\{ fires(T_k,k), \dots, $ $fires(t_l,k) \} \subseteq $ $A = \{ fires(T_k,k), \dots, fires(t_x,k) \} \subseteq A$ -- set union of subsets
\item Then for each $fires(t_x,k) \in A$ we have $t_x \in T_k$ -- already shown in item~\ref{pnpri:prove:fires_tk} above
\item Then $q_c = m_{M_k(p)}(c) + \sum_{t_x \in T_k \wedge p \in t_x \bullet}{m_{W(t_x,p)}(c)} - (\sum_{t_x \in T_k \wedge p \in \bullet t_x}{m_{W(p,t_x)}(c)} + \sum_{t_x \in T_k \wedge p \in R(t_x)}{m_{M_k(p)}(c)})$ -- from 
\eqref{pnpri:x:2:induction} above and the following
\begin{enumerate}
\item Each $add(p,{q_j}_c,t_j,c,k) \in A$ represents ${q_j}_c=m_{W(t_j,p)}(c)$ for $p \in t_j \bullet$ -- rule $r\ref{r:c:add},f\ref{f:c:tparc}$ encoding, and definition~\ref{def:pnpri:texec} of transition execution in $PN$ %
\item Each $del(p,t_y,{q_y}_c,c,k) \in A$ represents either ${q_y}_c=m_{W(p,t_y)}(c)$ for $p \in \bullet t_y$, or ${q_y}_c=m_{M_k(p)}(c)$ for $p \in R(t_y)$ -- from rule $r\ref{r:c:del},f\ref{f:c:ptarc}$ encoding and definition~\ref{def:pnpri:texec} of transition execution in $PN$; or from rule $r\ref{r:c:del},f\ref{f:c:rptarc}$ encoding and definition of reset arc in $PN$
\item Each $tot\_incr(p,q2_c,c,k) \in A$ represents $q2_c=\sum_{t_x \in T_k \wedge p \in t_x  \bullet}{m_{W(t_x,p)}(c)}$ -- aggregate assignment atom semantics in rule $r\ref{r:c:totincr}$
\item Each $tot\_decr(p,q3_c,c,k) \in A$ represents $q3_c=\sum_{t_x \in T_k \wedge p \in \bullet t_x}{m_{W(p,t_x)}(c)} + \sum_{t_x \in T_k \wedge p \in R(t_x)}{m_{M_k(p)}(c)}$ -- aggregate assignment atom semantics in rule $r\ref{r:c:totdecr}$
\end{enumerate}
\item Then, $m_{M_{k+1}(p)}(c) = q_c$ -- since $holds(p,q_c,c,k+1) \in A$ encodes $q_c=m_{M_{k+1}(p)}(c)$ -- from construction
\end{enumerate}
\end{enumerate}

\noindent
As a result, for any $n > k$, $T_n$ will be a valid firing set for $M_n$ and $M_{n+1}$ will be its target marking. 

\noindent
{\bf Conclusion:} Since both \eqref{prove:x2a:pnpri} and \eqref{prove:a2x:pnpri} hold, $X=M_0,T_k,M_1,\dots,M_k,T_{k+1}$ is an execution sequence of $PN(P,T,E,C,W,R,I,Q,QW,Z)$ (w.r.t $M_0$) iff there is an answer set $A$ of $\Pi^6(PN,M_0,k,ntok)$ such that \eqref{eqn:pnpri:fires} and \eqref{eqn:pnpri:holds} hold.

\section{Proof of Proposition~\ref{prop:dur}}

Let $PN=(P,T,E,C,W,R,I,Q,QW,Z,D)$ be a Petri Net, $M_0$ be its initial marking and let $\Pi^7(PN,M_0,k,ntok)$ be the ASP encoding of $PN$ and $M_0$ over a simulation length $k$, with maximum $ntok$ tokens on any place node, as defined in section~\ref{sec:enc_dur}. Then $X=M_0,T_k,M_1,\dots,M_k,T_k,M_{k+1}$ is an execution sequence of $PN$ (w.r.t. $M_0$) iff there is an answer set $A$ of $\Pi^7(PN,M_0,k,ntok)$ such that: 
\begin{equation}
\{ fires(t,ts) : t \in T_{ts}, 0 \leq ts \leq k\} = \{ fires(t,ts) : fires(t,ts) \in A \} \label{eqn:pndur:fires}
\end{equation}
\begin{equation}
\begin{split}
\{ holds(p,q,c,ts) &: p \in P, c/q = M_{ts}(p), 0 \leq ts \leq k+1 \} \\
&= \{ holds(p,q,c,ts) : holds(p,q,c,ts) \in A \} \label{eqn:pndur:holds}
\end{split}
\end{equation}

We prove this by showing that:
\begin{enumerate}[(I)]
\item Given an execution sequence $X$, we create a set $A$ such that it satisfies \eqref{eqn:pndur:fires} and \eqref{eqn:pndur:holds} and show that $A$ is an answer set of $\Pi^7$ \label{prove:x2a:pndur}
\item Given an answer set $A$ of $\Pi^7$, we create an execution sequence $X$ such that \eqref{eqn:pndur:fires} and \eqref{eqn:pndur:holds} are satisfied. \label{prove:a2x:pndur}
\end{enumerate}

\noindent
{\bf First we show (\ref{prove:x2a:pndur})}: We create a set $A$ as a union of the following sets:
\begin{enumerate}
\item $A_1=\{ num(n) : 0 \leq n \leq ntok \}$\label{pndur:builda:num}
\item $A_2=\{ time(ts) : 0 \leq ts \leq k\}$\label{pndur:builda:time}
\item $A_3=\{ place(p) : p \in P \}$\label{pndur:builda:place}
\item $A_4=\{ trans(t) : t \in T \}$\label{pndur:builda:trans}
\item $A_5=\{ color(c) : c \in C \}$
\item $A_6=\{ ptarc(p,t,n_c,c,ts) : (p,t) \in E^-, c \in C, n_c=m_{W(p,t)}(c), n_c > 0, 0 \leq ts \leq k \}$, where $E^- \subseteq E$\label{pndur:builda:ptarc}
\item $A_7=\{ tparc(t,p,n_c,c,ts,d) : (t,p) \in E^+, c \in C, n_c=m_{W(t,p)}(c), n_c > 0, d=D(t), 0 \leq ts \leq k \}$, where $E^+ \subseteq E$\label{pndur:builda:tparc}
\item $A_8=\{ holds(p,q_c,c,0) : p \in P, c \in C, q_c=m_{M_{0}(p)}(c) \}$\label{pndur:builda:holds0}
\item $A_9=\{ ptarc(p,t,n_c,c,ts) : p \in R(t), c \in C, n_c = m_{M_{ts}(p)}, n_c > 0, 0 \leq ts \leq k \}$
\item $A_{10}=\{ iptarc(p,t,1,c,ts) : p \in I(t), c \in C, 0 \leq ts < k \}$
\item $A_{11}=\{ tptarc(p,t,n_c,c,ts) : (p,t) \in Q, c \in C, n_c=m_{QW(p,t)}(c), n_c > 0, 0 \leq ts \leq k \}$
\item $A_{12}=\{ notenabled(t,ts) : t \in T, 0 \leq ts \leq k, \exists c \in C, (\exists p \in \bullet t, m_{M_{ts}(p)}(c) < m_{W(p,t)}(c)) \vee (\exists p \in I(t), m_{M_{ts}(p)}(c) > 0) \vee (\exists (p,t) \in Q, m_{M_{ts}(p)}(c) < m_{QW(p,t)}(c)) \}$ \newline per definition~\ref{def:pndur:enable} (enabled transition)\label{pndur:builda:notenabled} %
\item $A_{13}=\{ enabled(t,ts) : t \in T, 0 \leq ts \leq k, \forall c \in C, (\forall p \in \bullet t, m_{W(p,t)}(c) \leq m_{M_{ts}(p)}(c)) \wedge (\forall p \in I(t), m_{M_{ts}(p)}(c) = 0) \wedge (\forall (p,t) \in Q, m_{M_{ts}(p)}(c) \geq m_{QW(p,t)}(c)) \}$ \newline per definition~\ref{def:pndur:enable} (enabled transition)\label{pndur:builda:enabled} %
\item $A_{14}=\{ fires(t,ts) : t \in T_{ts}, 0 \leq ts \leq k \}$ \newline per definition~\ref{def:pndur:enable} (enabled transitions), only an enabled transition may fire\label{pndur:builda:fires}
\item $A_{15}=\{ add(p,q_c,t,c,ts+d-1) : t \in T_{ts}, p \in t \bullet, c \in C, q_c=m_{W(t,p)}(c), \\d = D(t), 0 \leq ts \leq k \}$ \newline per definition~\ref{def:pndur:texec} (transition execution)\label{pndur:builda:add}
\item $A_{16}=\{ del(p,q_c,t,c,ts) : t \in T_{ts}, p \in \bullet t, c \in C, q_c=m_{W(p,t)}(c), 0 \leq ts \leq k \} \cup \{ del(p,q_c,t,c,ts) : t \in T_{ts}, p \in R(t), c \in C, q_c=m_{M_{ts}(p)}(c), 0 \leq ts \leq k \}$ \newline per definition~\ref{def:pndur:texec} (transition execution) %
\item $A_{17}=\{ tot\_incr(p,q_c,c,ts) : p \in P, c \in C, \\ q_c=\sum_{t \in T_{l}, p \in t \bullet, l \leq ts, l+D(t)=ts+1}{m_{W(t,p)}(c)}, 0 \leq ts \leq k \}$ \newline per definition~\ref{def:pndur:exec} (firing set execution)\label{pndur:builda:tot_incr}
\item $A_{18}=\{ tot\_decr(p,q_c,c,ts): p \in P, c \in C, q_c=\sum_{t \in T_{ts}, p \in \bullet t}{m_{W(p,t)}(c)}+ \\ \sum_{t \in T_{ts}, p \in R(t) }{m_{M_{ts}(p)}(c)}, 0 \leq ts \leq k \}$ \newline per definition~\ref{def:pndur:exec} (firing set execution)\label{pndur:builda:tot_decr}
\item $A_{19}=\{ consumesmore(p,ts) : p \in P, c \in C, q_c=m_{M_{ts}(p)}(c), \\ q1_c=\sum_{t \in T_{ts}, p \in \bullet t}{m_{W(p,t)}(c)} + \sum_{t \in T_{ts}, p \in R(t)}{m_{M_{ts}(p)}(c)}, q1_c > q_c, 0 \leq ts \leq k \}$ \newline per definition~\ref{def:pndur:conflict} (conflicting transitions)\label{pndur:builda:consumesmore}
\item $A_{20}=\{ consumesmore : \exists p \in P, c \in C, q_c=m_{M_{ts}(p)}(c), \\ q1_c=\sum_{t \in T_{ts}, p \in \bullet t}{m_{W(p,t)}(c)} + \sum_{t \in T_{ts}, p \in R(t)}(m_{M_{ts}(p)}(c)), q1_c > q_c, 0 \leq ts \leq k \}$ \newline per definition~\ref{def:pndur:conflict} (conflicting transitions)\label{pndur:builda:consumesmore1}
\item $A_{21}=\{ holds(p,q_c,c,ts+1) : p \in P, c \in C, q_c=m_{M_{ts+1}(p)}(c), 0 \leq ts < k\}$, \newline where $M_{ts+1}(p) = M_{ts}(p) - (\sum_{\substack{t \in T_{ts}, p \in \bullet t}}{W(p,t)} + \sum_{t \in T_{ts}, p \in R(t)}M_{ts}(p)) + \\ \sum_{\substack{t \in T_l, p \in t \bullet, l \leq ts, l+D(t)-1=ts}}{W(t,p)}$ \newline according to definition~\ref{def:pndur:exec} (firing set execution)\label{pndur:builda:holds}
\item $A_{22}=\{ transpr(t,pr) : t \in T, pr=Z(t) \}$
\item $A_{23}=\{ notprenabled(t,ts) : t \in T, enabled(t,ts) \in A_{13}, (\exists tt \in T, enabled(tt,ts) \in A_{13}, Z(tt) < Z(t)), 0 \leq ts \leq k \} \\ = \{ notprenabled(t,ts) : t \in T, (\forall p \in \bullet t, W(p,t) \leq M_{ts}(p)), (\forall p \in I(t), M_{ts}(p) = 0), (\forall (p,t) \in Q, M_0(p) \geq WQ(p,t)), \exists tt \in T, $ $(\forall pp \in \bullet tt, W(pp,tt) \leq M_{ts}(pp)), $ $(\forall pp \in I(tt), M_{ts}(pp) = 0), (\forall (pp,tt) \in Q, M_{ts}(pp) \geq QW(pp,tt)), Z(tt) < Z(t), 0 \leq ts \leq k \}$
\item $A_{24}=\{ prenabled(t,ts) : t \in T, enabled(t,ts) \in A_{13}, (\nexists tt \in T: enabled(tt,ts) \in A_{13}, Z(tt) < Z(t)), 0 \leq ts \leq k \} \\ = \{ prenabled(t,ts) : t \in T, (\forall p \in \bullet t, W(p,t) \leq M_{ts}(p)), (\forall p \in I(t), M_{ts}(p) = 0), (\forall (p,t) \in Q, M_0(p) \geq WQ(p,t)), \nexists tt \in T, $ $(\forall pp \in \bullet tt, W(pp,tt) \leq M_{ts}(pp)), $ $(\forall pp \in I(tt), M_{ts}(pp) = 0), (\forall (pp,tt) \in Q, M_{ts}(pp) \geq QW(pp,tt)), Z(tt) < Z(t), 0 \leq ts \leq k \}$
\item $A_{25}=\{ could\_not\_have(t,ts): t \in T, prenabled(t,ts) \in A_{24}, fires(t,ts) \notin A_{14}, (\exists p \in \bullet t \cup R(t): q > M_{ts}(p) - (\sum_{t' \in T_{ts}, p \in \bullet t'}{W(p,t')} + \sum_{t' \in T_{ts}, p \in R(t')}{M_{ts}(p)}), q=W(p,t) \text{~if~} (p,t) \in E^- \text{~or~} R(t) \text{~otherwise}), 0 \leq ts \leq k\} \\ =\{ could\_not\_have(t,ts) : t \in T, (\forall p \in \bullet t, W(p,t) \leq M_{ts}(p)), (\forall p \in I(t), M_{ts}(p) = 0), (\forall (p,t) \in Q, M_0(p) \geq QW(p,t)), (\nexists tt \in T, $ $(\forall pp \in \bullet tt, W(pp,tt) \leq M_{ts}(pp)), $ $(\forall pp \in I(tt), M_{ts}(pp) = 0), (\forall (pp,tt) \in Q, M_{ts}(pp) \geq QW(pp,tt)), Z(tt) < Z(t)), t \notin T_{ts}, (\exists p \in \bullet t \cup R(t): q > M_{ts}(p) - (\sum_{t' \in T_{ts}, p \in \bullet t'}{W(p,t')} + \\ \sum_{t' \in T_{ts}, p \in R(t')}{M_{ts}(p)}), q=W(p,t) \text{~if~} (p,t) \in E^- \text{~or~} R(t) \text{~otherwise}), 0 \leq ts \leq k \}$ \newline
per the maximal firing set semantics
\end{enumerate}

\noindent
{\bf We show that $A$ satisfies \eqref{eqn:pndur:fires} and \eqref{eqn:pndur:holds}, and $A$ is an answer set of $\Pi^7$.}

$A$ satisfies \eqref{eqn:pndur:fires} and \eqref{eqn:pndur:holds} by its construction above. We show $A$ is an answer set of $\Pi^7$ by splitting. We split $lit(\Pi^7)$ into a sequence of $9k+11$ sets:

\begin{itemize}\renewcommand{\labelitemi}{$\bullet$}
\item $U_0= head(f\ref{f:c:place}) \cup head(f\ref{f:c:trans}) \cup head(f\ref{f:c:col}) \cup head(f\ref{f:c:time}) \cup head(f\ref{f:c:num}) \cup head(i\ref{i:c:init}) \cup head(f\ref{f:c:pr}) = \{place(p) : p \in P\} \cup \{ trans(t) : t \in T\} \cup \{ col(c) : c \in C \} \cup \{ time(0), \dots, time(k)\} \cup \{num(0), \dots, num(ntok)\} \cup \{ holds(p,q_c,c,0) : p \in P, c \in C, q_c=m_{M_0(p)}(c) \} \cup \{ transpr(t,pr) : t \in T, pr=Z(t) \} $
\item $U_{9k+1}=U_{9k+0} \cup head(f\ref{f:c:ptarc})^{ts=k} \cup head(f\ref{f:c:tparc})^{ts=k} \cup head(f\ref{f:c:rptarc})^{ts=k} \cup head(f\ref{f:c:iptarc})^{ts=k} \cup head(f\ref{f:c:tptarc})^{ts=k} = U_{9k+0} \cup $ $\{ ptarc(p,t,n_c,c,k) : (p,t) \in E^-, c \in C, n_c=m_{W(p,t)}(c) \} \\ \cup \{  tparc(t,p,n_c,c,k,d) : (t,p) \in E^+, c \in C, n_c=m_{W(t,p)}(c), d=D(t) \} \cup $ $\{ ptarc(p,t,n_c,c,k) : p \in R(t), c \in C, n_c=m_{M_{k}(p)}(c), n > 0 \}  \cup \{ iptarc(p,t,1,c,k) : \\ p \in I(t), c \in C \} \cup $ $\{ tptarc(p,t,n_c,c,k) : (p,t) \in Q, c \in C, n_c=m_{QW(p,t)}(c) \}$
\item $U_{9k+2}=U_{9k+1} \cup head(e\ref{e:c:ne:ptarc})^{ts=k} \cup head(e\ref{e:c:ne:iptarc})^{ts=k} \cup head(e\ref{e:c:ne:tptarc})^{ts=k} = U_{10k+1} \cup \\ \{ notenabled(t,k) : t \in T \}$
\item $U_{9k+3}=U_{9k+2} \cup head(e\ref{e:c:enabled})^{ts=k} = U_{9k+2} \cup \{ enabled(t,k) : t \in T \}$
\item $U_{9k+4}=U_{9k+3} \cup head(a\ref{a:c:prne})^{ts=k} = U_{9k+3} \cup \{ notprenabled(t,k) : t \in T \}$
\item $U_{9k+5}=U_{9k+4} \cup head(a\ref{a:c:prenabled})^{ts=k} = U_{9k+4} \cup \{ prenabled(t,k) : t \in T \}$
\item $U_{9k+6}=U_{9k+5} \cup head(a\ref{a:c:prfires})^{ts=k} = U_{9k+5} \cup \{ fires(t,k) : t \in T \}$
\item $U_{9k+7}=U_{9k+6}  \cup head(r\ref{r:c:dur:add})^{ts=k} \cup head(r\ref{r:c:del})^{ts=k} = U_{9k+6} \cup \{ add(p,q_c,t,c,k) : p \in P, t \in T, c \in C, q_c=m_{W(t,p)}(c) \} \cup \{ del(p,q_c,t,c,k) : p \in P, t \in T, c \in C, q_c=m_{W(p,t)}(c) \} \cup \{ del(p,q_c,t,c,k) : p \in P, t \in T, c \in C, q_c=m_{M_{k}(p)}(c) \}$
\item $U_{9k+8}=U_{9k+7} \cup head(r\ref{r:c:totincr})^{ts=k} \cup head(r\ref{r:c:totdecr})^{ts=k} = U_{9k+7} \cup \{ tot\_incr(p,q_c,c,k) : p \in P, c \in C, 0 \leq q_c \leq ntok \} \cup \{ tot\_decr(p,q_c,c,k) : p \in P, c \in C, 0 \leq q_c \leq ntok \}$
\item $U_{9k+9}=U_{9k+8} \cup head(a\ref{a:c:overc:place})^{ts=k} \cup head(r\ref{r:c:nextstate})^{ts=k} \cup head(a\ref{a:c:prmaxfire:cnh})^{ts=k} = U_{9k+8} \cup \\ \{ consumesmore(p,k) : p \in P\} \cup \{ holds(p,q,k+1) : p \in P, 0 \leq q \leq ntok \} \cup \{ could\_not\_have(t,k) : t \in T \}$
\item $U_{9k+10}=U_{9k+9} \cup head(a\ref{a:c:overc:gen}) = U_{9k+9} \cup \{ consumesmore \}$
\end{itemize}
where $head(r_i)^{ts=k}$ are head atoms of ground rule $r_i$ in which $ts=k$. We write $A_i^{ts=k} = \{ a(\dots,ts) : a(\dots,ts) \in A_i, ts=k \}$ as short hand for all atoms in $A_i$ with $ts=k$. $U_{\alpha}, 0 \leq \alpha \leq 9k+10$ form a splitting sequence, since each $U_i$ is a splitting set of $\Pi^7$, and $\langle U_{\alpha}\rangle_{\alpha < \mu}$ is a monotone continuous sequence, where $U_0 \subseteq U_1 \dots \subseteq U_{9k+10}$ and $\bigcup_{\alpha < \mu}{U_{\alpha}} = lit(\Pi^7)$. 

We compute the answer set of $\Pi^7$ using the splitting sets as follows:
\begin{enumerate}
\item $bot_{U_0}(\Pi^7) = f\ref{f:c:place} \cup f\ref{f:c:trans} \cup f\ref{f:c:col} \cup f\ref{f:c:time} \cup f\ref{f:c:num} \cup i\ref{i:c:init} \cup f\ref{f:c:pr}$ and $X_0 = A_1 \cup \dots \cup A_5 \cup A_8$ ($= U_0$) is its answer set -- using forced atom proposition

\item $eval_{U_0}(bot_{U_1}(\Pi^7) \setminus bot_{U_0}(\Pi^7), X_0) = \{ ptarc(p,t,q_c,c,0) \text{:-}. |  c \in C, q_c=m_{W(p,t)}(c), q_c > 0 \} \cup $ $\{ tparc(t,p,q_c,c,0,d) \text{:-}. | c \in C, q_c=m_{W(t,p)}(c), q_c > 0, d=D(t) \} \cup \\ \{ ptarc(p,t,q_c,c,0) \text{:-}. | c \in C, q_c=m_{M_0(p)}(c), q_c > 0 \} \cup $ $\{ iptarc(p,t,1,c,0) \text{:-}. | c \in C \} \cup $ $\{ tptarc(p,t,q_c,c,0) \text{:-}. | c \in C, q_c = m_{QW(p,t)}(c), q_c > 0 \} $. Its answer set $X_1=A_6^{ts=0} \cup A_7^{ts=0} \cup A_9^{ts=0} \cup A_{10}^{ts=0} \cup A_{11}^{ts=0}$ -- using forced atom proposition and construction of $A_6, A_7, A_9, A_{10}, A_{11}$.

\item $eval_{U_1}(bot_{U_2}(\Pi^7) \setminus bot_{U_1}(\Pi^7), X_0 \cup X_1) = \{ notenabled(t,0) \text{:-} . | (\{ trans(t), \\ ptarc(p,t,n_c,c,0), $ $holds(p,q_c,c,0) \} \subseteq X_0 \cup X_1, \text{~where~}  q_c < n_c) \text{~or~} \\ (\{ notenabled(t,0) \text{:-} . | $ $(\{ trans(t), $ $iptarc(p,t,n2_c,c,0), $ $ holds(p,q_c,c,0) \} \subseteq X_0 \cup X_1, \text{~where~}  q_c \geq n2_c \}) \text{~or~} $ $(\{ trans(t), $ $tptarc(p,t,n3_c,c,0), $ $holds(p,q_c,c,0) \} \subseteq X_0 \cup X_1, \text{~where~} q_c < n3_c) \}$. Its answer set $X_2=A_{12}^{ts=0}$ -- using  forced atom proposition and construction of $A_{12}$.
\begin{enumerate}
\item where, $q_c=m_{M_0(p)}(c)$, and $n_c=m_{W(p,t)}(c)$ for an arc $(p,t) \in E^-$ -- by construction of $i\ref{i:c:init}$ and $f\ref{f:c:ptarc}$  in $\Pi^7$, and 
\item in an arc $(p,t) \in E^-$, $p \in \bullet t$ (by definition~\ref{def:pnpri:preset} of preset)
\item $n2_c=1$ -- by construction of $iptarc$ predicates in $\Pi^7$, meaning $q_c \geq n2_c \equiv q_c \geq 1 \equiv q_c > 0$,
\item $tptarc(p,t,n3_c,c,0)$ represents $n3_c=m_{QW(p,t)}(c)$, where $(p,t) \in Q$ 
\item thus, $notenabled(t,0) \in X_1$ means $\exists c \in C, (\exists p \in \bullet t : m_{M_0(p)}(c) < m_{W(p,t)}(c)) \\ \vee (\exists p \in I(t) : m_{M_0(p)}(c) > 0) \vee (\exists (p,t) \in Q : m_{M_0(p)}(c) < m_{QW(p,t)}(c))$.
\end{enumerate}

\item $eval_{U_2}(bot_{U_3}(\Pi^7) \setminus bot_{U_2}(\Pi^7), X_0 \cup \dots \cup X_2) = \{ enabled(t,0) \text{:-}. | trans(t) \in X_0 \cup \dots \cup X_2, notenabled(t,0) \notin X_0 \cup \dots \cup X_2 \}$. Its answer set is $X_3 = A_{13}^{ts=0}$ -- using forced atom proposition and construction of $A_{13}$. \label{pndur:x2a:base:enabled}
\begin{enumerate}
\item since an $enabled(t,0) \in X_3$ if $\nexists ~notenabled(t,0) \in X_0 \cup \dots \cup X_2$; which is equivalent to $\nexists t, \forall c \in C, (\nexists p \in \bullet t, m_{M_0(p)}(c) < m_{W(p,t)}(c)), (\nexists p \in I(t), \\ m_{M_0(p)}(c) > 0), (\nexists (p,t) \in Q : m_{M_0(p)}(c) < m_{QW(p,t)}(c) ), \forall c \in C, (\forall p \in \bullet t: m_{M_0(p)}(c) \geq m_{W(p,t)}(c)), (\forall p \in I(t) : m_{M_0(p)}(c) = 0)$.
\end{enumerate}

\item $eval_{U_3}(bot_{U_4}(\Pi^7) \setminus bot_{U_3}(\Pi^7), X_0 \cup \dots \cup X_3) = \{ notprenabled(t,0) \text{:-}. | \\ \{ enabled(t,0), transpr(t,p), enabled(tt,0), transpr(tt,pp) \} \subseteq X_0 \cup \dots \cup X_3, pp < p \}$. Its answer set is $X_4 = A_{23}^{ts=k}$ -- using forced atom proposition and construction of $A_{23}$. \label{pndur:x2a:base:notprenabled}
\begin{enumerate}
\item $enabled(t,0)$ represents $\exists t \in T, \forall c \in C, (\forall p \in \bullet t, m_{M_0(p)}(c) \geq m_{W(p,t)}(c)), $ $(\forall p \in I(t), m_{M_0(p)}(c) = 0), (\forall (p,t) \in Q, m_{M_0(p)}(c) \geq m_{QW(p,t)}(c))$
\item $enabled(tt,0)$ represents $\exists tt \in T, \forall c \in C, $ $(\forall pp \in \bullet tt, m_{M_0(pp)}(c) \geq \\ m_{W(pp,tt)}(c)) \wedge $ $(\forall pp \in I(tt), m_{M_0(pp)}(c) = 0), (\forall (pp,tt) \in Q, m_{M_0(pp)}(c) \geq m_{QW(pp,tt)}(c))$  
\item $transpr(t,p)$ represents $p=Z(t)$ -- by construction
\item $transpr(tt,pp)$ represents $pp=Z(tt)$ -- by construction
\item thus, $notprenabled(t,0)$ represents $\forall c \in C, $ $(\forall p \in \bullet t, m_{M_0(p)}(c) \geq m_{W(p,t)}(c)), $ $(\forall p \in I(t), m_{M_0(p)}(c) = 0), (\forall (p,t) \in Q, m_{M_0(p)}(c) \geq m_{QW(p,t)}(c)), \exists tt \in T, (\forall pp \in \bullet tt, m_{M_0(pp)}(c) \geq m_{W(pp,tt)}(c)), $ $(\forall pp \in I(tt), m_{M_0(pp)}(c) = 0), \\ (\forall (pp,tt) \in Q, m_{M_0(pp)}(c) \geq m_{QW(pp,tt)}(c)), Z(tt) < Z(t)$ 
\item which is equivalent to $(\forall p \in \bullet t: M_0(p) \geq W(p,t)), $ $(\forall p \in I(t), M_0(p) = 0), (\forall (p,t) \in Q, M_0(p) \geq QW(p,t)), \exists tt \in T, $ $(\forall pp \in \bullet tt, M_0(pp) \geq \\ W(pp,tt)), $ $(\forall pp \in I(tt), M_0(pp) = 0), (\forall (pp,tt) \in Q, M_0 \geq QW(pp,tt)), \\ Z(tt) < Z(t)$ -- assuming multiset domain $C$ for all operations
\end{enumerate}

\item $eval_{U_4}(bot_{U_5}(\Pi^7) \setminus bot_{U_4}(\Pi^7), X_0 \cup \dots \cup X_4) = \{ prenabled(t,0) \text{:-}. | enabled(t,0) \in X_0 \cup \dots \cup X_4, notprenabled(t,0) \notin X_0 \cup \dots \cup X_4 \}$. Its answer set is $X_5 = A_{24}^{ts=k}$ -- using forced atom proposition and construction of $A_{24}$ \label{pndur:x2a:base:prenabled}
\begin{enumerate}
\item $enabled(t,0)$ represents $\forall c \in C, (\forall p \in \bullet t, m_{M_0(p)}(c) \geq m_{W(p,t)}(c)), (\forall p \in I(t), m_{M_0(p)}(c) = 0), (\forall (p,t) \in Q, m_{M_0(p)}(c) \geq m_{QW(p,t)}(c))  \equiv $ $(\forall p \in \bullet t, \\ M_0(p) \geq W(p,t)), $ $(\forall p \in I(t), M_0(p) = 0), (\forall (p,t) \in Q, M_0(p) \geq QW(p,t))$ -- from \ref{pndur:x2a:base:enabled} above and assuming multiset domain $C$ for all operations
\item $notprenabled(t,0)$ represents $(\forall p \in \bullet t, M_0(p) \geq W(p,t)), $ $(\forall p \in I(t), \\ M_0(p) = 0), (\forall (p,t) \in Q, M_0(p) \geq QW(p,t)), \exists tt \in T, $ $(\forall pp \in \bullet tt, M_0(pp) \geq W(pp,tt)), $ $(\forall pp \in I(tt), M_0(pp) = 0), (\forall (pp,tt) \in Q, M_0(pp) \geq QW(pp,tt)), \\ Z(tt) < Z(t)$ -- from \ref{pndur:x2a:base:notprenabled} above and assuming multiset domain $C$ for all operations
\item then, $prenabled(t,0)$ represents $(\forall p \in \bullet t, M_0(p) \geq W(p,t)), $ $(\forall p \in I(t), \\ M_0(p) = 0), (\forall (p,t) \in Q, M_0(p) \geq QW(p,t)), \nexists tt \in T, $ $((\forall pp \in \bullet tt, \\ M_0(pp) \geq W(pp,tt)), $ $(\forall pp \in I(tt), M_0(pp) = 0), (\forall (pp,tt) \in Q, M_0(pp) \geq QW(pp,tt)), Z(tt) < Z(t))$ -- from (a), (b) and $enabled(t,0) \in X_0 \cup \dots \cup X_4$ 
\end{enumerate}

\item $eval_{U_5}(bot_{U_6}(\Pi^7) \setminus bot_{U_5}(\Pi^7), X_0 \cup \dots \cup X_5) = \{\{fires(t,0)\} \text{:-}. | prenabled(t,0) \\ \text{~holds in~} X_0 \cup \dots \cup X_5 \}$. It has multiple answer sets $X_{6.1}, \dots, X_{6.n}$, corresponding to elements of power set of $fires(t,0)$ atoms in $eval_{U_5}(...)$ -- using supported rule proposition. Since we are showing that the union of answer sets of $\Pi^7$ determined using splitting is equal to $A$, we only consider the set that matches the $fires(t,0)$ elements in $A$ and call it $X_6$, ignoring the rest. Thus, $X_6 = A_{14}^{ts=0}$, representing $T_0$.
\begin{enumerate}
\item in addition, for every $t$ such that $prenabled(t,0) \in X_0 \cup \dots \cup X_5,  R(t) \neq \emptyset$; $fires(t,0) \in X_6$ -- per definition~\ref{def:pndur:firing_set} (firing set); requiring that a reset transition is fired when enabled
\item thus, the firing set $T_0$ will not be eliminated by the constraint $f\ref{f:c:pr:rptarc:elim}$ 
\end{enumerate}

\item $eval_{U_6}(bot_{U_7}(\Pi^7) \setminus bot_{U_6}(\Pi^7), X_0 \cup \dots \cup X_6) = \{add(p,n_c,t,c,0) \text{:-}. | $ $\{fires(t,0-d+1), $ $tparc(t,p,n_c,c,0,d) \} \subseteq X_0 \cup \dots \cup X_6 \} \cup \{ del(p,n_c,t,c,0) \text{:-}. | $ $\{ fires(t,0), $ $ptarc(p,t,n_c,c,0) \} \subseteq X_0 \cup \dots \cup X_6 \}$. It's answer set is $X_7 = A_{15}^{ts=0} \cup A_{16}^{ts=0}$ -- using forced atom proposition and definitions of $A_{15}$ and $A_{16}$. 
\begin{enumerate}
\item where, each $add$ atom is equivalent to $n_c=m_{W(t,p)}(c),c \in C, p \in t \bullet$, 
\item and each $del$ atom is equivalent to $n_c=m_{W(p,t)}(c), c \in C, p \in \bullet t$; or $n_c=m_{M_0(p)}(c), c \in C, p \in R(t)$,
\item representing the effect of transitions in $T_0$
\end{enumerate}

\item $eval_{U_7}(bot_{U_8}(\Pi^7) \setminus bot_{U_7}(\Pi^7), X_0 \cup \dots \cup X_7) =  \\ \{tot\_incr(p,qq_c,c,0) \text{:-}. | qq_c=\sum_{add(p,q_c,t,c,0) \in X_0 \cup \dots \cup X_7}{q_c} \} \cup \\ \{ tot\_decr(p,qq_c,c,0) \text{:-}. |  qq_c=\sum_{del(p,q_c,t,c,0) \in X_0 \cup \dots \cup X_7}{q_c} \}$. It's answer set is $X_8 = A_{17}^{ts=0} \cup A_{18}^{ts=0}$ --  using forced atom proposition and definitions of $A_{17}$ and $A_{18}$.
\begin{enumerate}
\item where, each $tot\_incr(p,qq_c,c,0)$, $qq_c=\sum_{add(p,q_c,t,c,0) \in X_0 \cup \dots X_7}{q_c}$ \\ $\equiv qq_c=\sum_{t \in X_6, p \in t \bullet, 0+D(t)-1=0}{m_{W(p,t)}(c)}$, 
\item $tot\_decr(p,qq_c,c,0)$, $qq_c=\sum_{del(p,q_c,t,c,0) \in X_0 \cup \dots X_7}{q_c}$ \\ $\equiv qq=\sum_{t \in X_6, p \in \bullet t}{m_{W(t,p)}(c)} + \sum_{t \in X_6, p \in R(t)}{m_{M_0(p)}(c)}$, 
\item represent the net effect of transitions in $T_0$
\end{enumerate}
\item $eval_{U_8}(bot_{U_9}(\Pi^7) \setminus bot_{U_8}(\Pi^7), X_0 \cup \dots \cup X_8) = \{ consumesmore(p,0) \text{:-}. | \\ \{holds(p,q_c,c,0), tot\_decr(p,q1_c,c,0) \} \subseteq X_0 \cup \dots \cup X_8, q1_c > q_c \} \cup $ \\$\{ holds(p,q_c,c,1) \text{:-}. | \{ holds(p,q1_c,c,0), tot\_incr(p,q2_c,c,0), \\ tot\_decr(p,q3_c,c,0) \} \subseteq X_0 \cup \dots \cup X_6, q_c=q1_c+q2_c-q3_c \}  \cup \{ could\_not\_have(t,0) \text{:-}. | \\  \{ prenabled(t,0), ptarc(s,t,q,c,0), holds(s,qq,c,0), tot\_decr(s,qqq,c,0) \} \subseteq X_0 \cup \dots \cup X_8, fires(t,0) \notin X_0 \cup \dots \cup X_8, q > qq - qqq \}$. It's answer set is $X_9 = A_{19}^{ts=0} \cup A_{21}^{ts=0} \cup A_{25}^{ts=0}$  -- using forced atom proposition and definitions of $A_{19}, A_{21}, A_{25}$.
\begin{enumerate}
\item where $consumesmore(p,0)$ represents $\exists p : q_c=m_{M_0(p)}(c), q1_c= \\ \sum_{t \in T_0, p \in \bullet t}{m_{W(p,t)}(c)}+\sum_{t \in T_0, p \in R(t)}{m_{M_0(p)}(c)}, q1_c > q_c, c \in C$, indicating place $p$ will be over consumed if $T_0$ is fired, as defined in definition~\ref{def:pndur:conflict} (conflicting transitions),
\item $holds(p,q_c,c,1)$ if $q_c=m_{M_0(p)}(c)+\sum_{t \in T_0, p \in t \bullet, 0+D(t)-1=0}{m_{W(t,p)}(c)}- \\ (\sum_{t \in T_0, p \in \bullet t}{m_{W(p,t)}(c)}+ $ $\sum_{t \in T_0, p \in R(t)}{m_{M_0(p)}(c)})$ represented by $q_c=m_{M_1(p)}(c)$ for some $c \in C$ -- by construction of $\Pi^7$
\item and $consumesmore(p,0)$ if $\sum_{t \in T_0, p \in \bullet t}{m_{W(p,t)}(c)}+ $ $\sum_{t \in T_0, p \in R(t)}{m_{M_0(p)}(c)} > m_{M_0(p)}(c)$ for any $c \in C$
\item $could\_not\_have(t,0)$ if
\begin{enumerate}
\item $(\forall p \in \bullet t, W(p,t) \leq M_0(p)), (\forall p \in I(t),  M_0(p) = 0), (\forall (p,t) \in Q, \\ M_0(p) \geq WQ(p,t)), \nexists tt \in T, (\forall pp \in \bullet tt, W(pp,tt) \leq M_{ts}(pp)), (\forall pp \in I(tt), M_0(pp) = 0), (\forall (pp,tt) \in Q, M_0(pp) \geq QW(pp,tt)), Z(tt) < Z(t)$,
\item and $q_c > m_{M_0(s)}(c) - (\sum_{t' \in T_0, s \in \bullet t'}{m_{W(s,t')}(c)}+ \sum_{t' \in T_0, s \in R(t)}{m_{M_0(s)}(c)}), \\ q_c = m_{W(s,t)}(c) \text{~if~} s \in \bullet t \text{~or~} m_{M_0(s)}(c) \text{~otherwise}$ for some $c \in C$, which becomes $q > M_0(s) - (\sum_{t' \in T_0, s \in \bullet t'}{W(s,t')}+ \sum_{t' \in T_0, s \in R(t)}{M_0(s)}), q = W(s,t) \text{~if~} s \in \bullet t \text{~or~} M_0(s) \text{~otherwise}$ for all $c \in C$
\item (i), (ii) above combined match the definition of $A_{25}$
\end{enumerate}
\item $X_9$ does not contain $could\_not\_have(t,0)$, when $prenabled(t,0) \in X_0 \cup \dots \cup X_6$ and $fires(t,0) \notin X_0 \cup \dots \cup X_5$ due to construction of $A$, encoding of $a\ref{a:c:maxfire:cnh}$ and its body atoms. As a result, it is not eliminated by the constraint $a\ref{a:c:prmaxfire:elim}$
\end{enumerate}

\[ \vdots \]

\item $eval_{U_{9k+0}}(bot_{U_{9k+1}}(\Pi^7) \setminus bot_{U_{9k+0}}(\Pi^7), X_0 \cup \dots \cup X_{9k+0}) = \{ ptarc(p,t,q_c,c,k) \text{:-}. |  c \in C, q_c=m_{W(p,t)}(c), q_c > 0 \} \cup \{ tparc(t,p,q_c,c,k,d) \text{:-}. | c \in C, q_c=m_{W(t,p)}(c), q_c > 0, d=D(t) \} \cup $ $\{ ptarc(p,t,q_c,c,k) \text{:-}. | c \in C, q_c=m_{M_k(p)}(c), q_c > 0 \} \cup \\ \{ iptarc(p,t,1,c,k) \text{:-}. | c \in C \} \cup $ $\{ tptarc(p,t,q_c,c,k) \text{:-}. | c \in C, q_c = m_{QW(p,t)}(c), q_c > 0 \} $. Its answer set $X_{9k+1}=A_6^{ts=k} \cup A_7^{ts=k} \cup A_9^{ts=k} \cup A_{10}^{ts=k} \cup A_{11}^{ts=k}$ -- using forced atom proposition and construction of $A_6, A_7, A_9, A_{10}, A_{11}$.

\item $eval_{U_{9k+1}}(bot_{U_{10k+2}}(\Pi^7) \setminus bot_{U_{9k+1}}(\Pi^7), X_0 \cup \dots \cup X_{9k+1}) = \{ notenabled(t,k) \text{:-} . | \\ (\{ trans(t), ptarc(p,t,n_c,c,k), holds(p,q_c,c,k) \} \subseteq X_0 \cup \dots \cup X_{9k+1}, \text{~where~}  q_c < n_c) \text{~or~}  (\{ notenabled(t,k) \text{:-} . |  (\{ trans(t), iptarc(p,t,n2_c,c,k), holds(p,q_c,c,k) \} \subseteq X_0 \cup \dots \cup X_{9k+1}, $ $\text{~where~}  q_c \geq n2_c \}) \text{~or~} $ $(\{ trans(t), tptarc(p,t,n3_c,c,k), \\ holds(p,q_c,c,k) \} \subseteq X_0 \cup \dots \cup X_{9k+1}, $ $\text{~where~} q_c < n3_c) \}$. Its answer set $X_{9k+2}=A_{12}^{ts=k}$ -- using  forced atom proposition and construction of $A_{12}$.
\begin{enumerate}
\item where, $q_c=m_{M_k(p)}(c)$, and $n_c=m_{W(p,t)}(c)$ for an arc $(p,t) \in E^-$ -- by construction of $i\ref{i:c:init}$ and $f\ref{f:c:ptarc}$ predicates in $\Pi^7$, and 
\item in an arc $(p,t) \in E^-$, $p \in \bullet t$ (by definition~\ref{def:pnpri:preset} of preset)
\item $n2_c=1$ -- by construction of $iptarc$ predicates in $\Pi^7$, meaning $q_c \geq n2_c \equiv q_c \geq 1 \equiv q_c > 0$,
\item $tptarc(p,t,n3_c,c,k)$ represents $n3_c=m_{QW(p,t)}(c)$, where $(p,t) \in Q$ 
\item thus, $notenabled(t,k) \in X_{9k+1}$ represents $\exists c \in C, (\exists p \in \bullet t : m_{M_k(p)}(c) < m_{W(p,t)}(c)) \vee (\exists p \in I(t) : m_{M_k(p)}(c) > 0) \vee (\exists (p,t) \in Q : m_{M_k(p)}(c) < m_{QW(p,t)}(c))$.
\end{enumerate}

\item $eval_{U_{9k+2}}(bot_{U_{9k+3}}(\Pi^7) \setminus bot_{U_{9k+2}}(\Pi^7), X_0 \cup \dots \cup X_{9k+2}) = \{ enabled(t,k) \text{:-}. | trans(t) \in X_0 \cup \dots \cup X_{9k+2}, notenabled(t,k) \notin X_0 \cup \dots \cup X_{9k+2} \}$. Its answer set is $X_{9k+3} = A_{13}^{ts=k}$ -- using forced atom proposition and construction of $A_{13}$. \label{pndur:x2a:k:enabled}
\begin{enumerate}
\item Since an $enabled(t,k) \in X_{9k+3}$ if $\nexists ~notenabled(t,k) \in X_0 \cup \dots \cup X_{9k+2}$; which is equivalent to $\nexists t, \forall c \in C, (\nexists p \in \bullet t, m_{M_k(p)}(c) < m_{W(p,t)}(c)), (\nexists p \in I(t), m_{M_k(p)}(c) > 0), (\nexists (p,t) \in Q : m_{M_k(p)}(c) < m_{QW(p,t)}(c) ), \forall c \in C, (\forall p \in \bullet t: m_{M_k(p)}(c) \geq m_{W(p,t)}(c)), (\forall p \in I(t) : m_{M_k(p)}(c) = 0)$.
\end{enumerate}

\item $eval_{U_{9k+3}}(bot_{U_{9k+4}}(\Pi^7) \setminus bot_{U_{9k+3}}(\Pi^7), X_0 \cup \dots \cup X_{9k+3}) = \{ notprenabled(t,k) \text{:-}. | \\ \{ enabled(t,k), transpr(t,p), enabled(tt,k), transpr(tt,pp) \} \subseteq X_0 \cup \dots \cup X_{9k+3}, \\ pp < p \}$. Its answer set is $X_{9k+4} = A_{23}^{ts=k}$ -- using forced atom proposition and construction of $A_{23}$. \label{pndur:x2a:k:notprenabled}
\begin{enumerate}
\item $enabled(t,k)$ represents $\exists t \in T, \forall c \in C, (\forall p \in \bullet t, m_{M_k(p)}(c) \geq m_{W(p,t)}(c)), $ $(\forall p \in I(t), m_{M_k(p)}(c) = 0), (\forall (p,t) \in Q, m_{M_k(p)}(c) \geq m_{QW(p,t)}(c))$
\item $enabled(tt,k)$ represents $\exists tt \in T, \forall c \in C, (\forall pp \in \bullet tt, m_{M_k(pp)}(c) \geq \\ m_{W(pp,tt)}(c)) \wedge (\forall pp \in I(tt), m_{M_k(pp)}(c) = 0), (\forall (pp,tt) \in Q, m_{M_k(pp)}(c) \geq m_{QW(pp,tt)}(c))$  
\item $transpr(t,p)$ represents $p=Z(t)$ -- by construction
\item $transpr(tt,pp)$ represents $pp=Z(tt)$ -- by construction
\item thus, $notprenabled(t,k)$ represents $\forall c \in C, (\forall p \in \bullet t, m_{M_k(p)}(c) \geq \\ m_{W(p,t)}(c)), $ $(\forall p \in I(t), m_{M_k(p)}(c) = 0), (\forall (p,t) \in Q, m_{M_k(p)}(c) \geq \\ m_{QW(p,t)}(c)), \exists tt \in T, $ $(\forall pp \in \bullet tt, m_{M_k(pp)}(c) \geq m_{W(pp,tt)}(c)), $ $(\forall pp \in I(tt), \\ m_{M_k(pp)}(c) = 0), (\forall (pp,tt) \in Q, m_{M_k(pp)}(c) \geq m_{QW(pp,tt)}(c)), Z(tt) < Z(t)$ 
\item which is equivalent to $(\forall p \in \bullet t: M_k(p) \geq W(p,t)), $ $(\forall p \in I(t), \\ M_k(p) = 0), (\forall (p,t) \in Q, M_k(p) \geq QW(p,t)), \exists tt \in T, $ $(\forall pp \in \bullet tt, \\ M_k(pp) \geq W(pp,tt)), $ $(\forall pp \in I(tt), M_k(pp) = 0), (\forall (pp,tt) \in Q, M_k(pp) \geq QW(pp,tt)), Z(tt) < Z(t)$ -- assuming multiset domain $C$ for all operations
\end{enumerate}

\item $eval_{U_{9k+4}}(bot_{U_{9k+5}}(\Pi^7) \setminus bot_{U_{9k+4}}(\Pi^7), X_0 \cup \dots \cup X_{9k+4}) = \{ prenabled(t,k) \text{:-}. | \\ enabled(t,k) \in X_0 \cup \dots \cup X_{9k+4}, notprenabled(t,k) \notin X_0 \cup \dots \cup X_{9k+4} \}$. Its answer set is $X_{9k+5} = A_{24}^{ts=k}$ -- using forced atom proposition and construction of $A_{24}$ \label{pndur:x2a:k:prenabled}
\begin{enumerate}
\item $enabled(t,k)$ represents $\forall c \in C, (\forall p \in \bullet t, m_{M_k(p)}(c) \geq m_{W(p,t)}(c)), $ $(\forall p \in I(t), m_{M_k(p)}(c) = 0), (\forall (p,t) \in Q, m_{M_k(p)}(c) \geq m_{QW(p,t)}(c))  \equiv (\forall p \in \bullet t, M_k(p) \geq W(p,t)), $ $(\forall p \in I(t), M_k(p) = 0), (\forall (p,t) \in Q, M_k(p) \geq QW(p,t))$ -- from \ref{pndur:x2a:k:enabled} above and assuming multiset domain $C$ for all operations
\item $notprenabled(t,k)$ represents $(\forall p \in \bullet t, M_k(p) \geq W(p,t)), (\forall p \in I(t), \\ M_k(p) = 0), (\forall (p,t) \in Q, M_k(p) \geq QW(p,t)), \exists tt \in T, $ $(\forall pp \in \bullet tt, M_k(pp) \geq W(pp,tt)), (\forall pp \in I(tt), M_k(pp) = 0), (\forall (pp,tt) \in Q, M_k(pp) \geq QW(pp,tt)), \\ Z(tt) < Z(t)$ -- from \ref{pndur:x2a:k:notprenabled} above and assuming multiset domain $C$ for all operations
\item then, $prenabled(t,k)$ represents $(\forall p \in \bullet t, M_k(p) \geq W(p,t)), (\forall p \in I(t), \\ M_k(p) = 0), (\forall (p,t) \in Q, M_k(p) \geq QW(p,t)), \nexists tt \in T, $ $((\forall pp \in \bullet tt, M_k(pp) \geq W(pp,tt)), $ $(\forall pp \in I(tt), M_k(pp) = 0), (\forall (pp,tt) \in Q, \\ M_k(pp) \geq QW(pp,tt)), Z(tt) < Z(t))$ -- from (a), (b) and $enabled(t,k) \in X_0 \cup \dots \cup X_{9k+4}$ 
\end{enumerate}

\item $eval_{U_{9k+5}}(bot_{U_{9k+6}}(\Pi^7) \setminus bot_{U_{9k+5}}(\Pi^7), X_0 \cup \dots \cup X_{9k+5}) = \{\{fires(t,k)\} \text{:-}. | \\ prenabled(t,k)  \text{~holds in~} X_0 \cup \dots \cup X_{9k+5} \}$. It has multiple answer sets \\ $X_{9k+6.1}, \dots, X_{9k+6.n}$, corresponding to elements of power set of $fires(t,k)$ atoms in $eval_{U_{9k+5}}(...)$ -- using supported rule proposition. Since we are showing that the union of answer sets of $\Pi^7$ determined using splitting is equal to $A$, we only consider the set that matches the $fires(t,k)$ elements in $A$ and call it $X_{9k+6}$, ignoring the rest. Thus, $X_{9k+6} = A_{14}^{ts=k}$, representing $T_k$.
\begin{enumerate}
\item in addition, for every $t$ such that $prenabled(t,k) \in X_0 \cup \dots \cup X_{9k+5},  R(t) \neq \emptyset$; $fires(t,k) \in X_{9k+6}$ -- per definition~\ref{def:pndur:firing_set} (firing set); requiring that a reset transition is fired when enabled
\item thus, the firing set $T_k$ will not be eliminated by the constraint $f\ref{f:c:pr:rptarc:elim}$ 
\end{enumerate}

\item $eval_{U_{9k+6}}(bot_{U_{9k+7}}(\Pi^7) \setminus bot_{U_{9k+6}}(\Pi^7), X_0 \cup \dots \cup X_{9k+6}) = \{add(p,n_c,t,c,k) \text{:-}. | $ $\{fires(t,k-d+1), tparc(t,p,n_c,c,0,d) \} \subseteq X_0 \cup \dots \cup X_{9k+6} \} \cup \\ \{ del(p,n_c,t,c,k) \text{:-}. | \{ fires(t,k), ptarc(p,t,n_c,c,k) \} \subseteq X_0 \cup \dots \cup X_{9k+6} \}$. It's answer set is $X_{9k+7} = A_{15}^{ts=k} \cup A_{16}^{ts=k}$ -- using forced atom proposition and definitions of $A_{15}$ and $A_{16}$.
\begin{enumerate}
\item where, each $add$ atom is equivalent to $n_c=m_{W(t,p)}(c),c \in C, p \in t \bullet$, 
\item and each $del$ atom is equivalent to $n_c=m_{W(p,t)}(c), c \in C, p \in \bullet t$; or $n_c=m_{M_k(p)}(c), c \in C, p \in R(t)$,
\item representing the effect of transitions in $T_k$
\end{enumerate}

\item $eval_{U_{9k+7}}(bot_{U_{9k+8}}(\Pi^7) \setminus bot_{U_{9k+7}}(\Pi^7), X_0 \cup \dots \cup X_{9k+7}) = \{tot\_incr(p,qq_c,c,k) \text{:-}. | $ \\$ qq_c=\sum_{add(p,q_c,t,c,k) \in X_0 \cup \dots \cup X_{9k+7}}{q_c} \} \cup \{ tot\_decr(p,qq_c,c,k) \text{:-}. | $ \\$ qq_c=\sum_{del(p,q_c,t,c,k) \in X_0 \cup \dots \cup X_{9k+7}}{q_c} \}$. It's answer set is $X_{9k+8} = A_{17}^{ts=k} \cup A_{18}^{ts=k}$ --  using forced atom proposition and definitions of $A_{17}$ and $A_{18}$.
\begin{enumerate}
\item where, each $tot\_incr(p,qq_c,c,k)$, $qq_c=\sum_{add(p,q_c,t,c,k) \in X_0 \cup \dots X_{9k+7}}{q_c}$ \\ $\equiv qq_c=\sum_{t \in X_{9k+6}, p \in t \bullet, 0 \leq l \leq k, l+D(t)-1=k }{m_{W(p,t)}(c)}$,
\item and each $tot\_decr(p,qq_c,c,k)$, $qq_c=\sum_{del(p,q_c,t,c,k) \in X_0 \cup \dots X_{9k+7}}{q_c}$ \\ $\equiv qq=\sum_{t \in X_{9k+6}, p \in \bullet t}{m_{W(t,p)}(c)} + \sum_{t \in X_{9k+6}, p \in R(t)}{m_{M_k(p)}(c)}$,
\item represent the net effect of transitions in $T_k$
\end{enumerate}
\item $eval_{U_{9k+8}}(bot_{U_{9k+9}}(\Pi^7) \setminus bot_{U_{9k+8}}(\Pi^7), X_0 \cup \dots \cup X_{9k+8}) = $ $\{ consumesmore(p,k) \text{:-}. | \\ \{holds(p,q_c,c,k), tot\_decr(p,q1_c,c,k) \} \subseteq X_0 \cup \dots \cup X_{9k+8}, q1_c > q_c \} \cup \\ \{ holds(p,q_c,c,k+1) \text{:-}., |  \{ holds(p,q1_c,c,k), tot\_incr(p,q2_c,c,k), \\ tot\_decr(p,q3_c,c,k) \} \subseteq X_0 \cup \dots \cup X_{9k+6}, q_c=q1_c+q2_c-q3_c \}  \cup \\ \{ could\_not\_have(t,k) \text{:-}. | \{ prenabled(t,k), \\ ptarc(s,t,q,c,k), holds(s,qq,c,k), tot\_decr(s,qqq,c,k) \} \subseteq X_0 \cup \dots \cup X_{9k+8}, \\ fires(t,k) \notin X_0 \cup \dots \cup X_{10k+8}, q > qq - qqq \}$. It's answer set is $X_{9k+9} = A_{19}^{ts=k} \cup A_{21}^{ts=k} \cup A_{25}^{ts=k}$  -- using forced atom proposition and definitions of $A_{19}, A_{21}, A_{25}$.
\begin{enumerate}
\item where, $consumesmore(p,k)$ represents $\exists p : q_c=m_{M_k(p)}(c), \\ q1_c=\sum_{t \in T_k, p \in \bullet t}{m_{W(p,t)}(c)}+\sum_{t \in T_k, p \in R(t)}{m_{M_k(p)}(c)}, q1_c > q_c, c \in C$, indicating place $p$ that will be over consumed if $T_k$ is fired, as defined in definition~\ref{def:pndur:conflict} (conflicting transitions),
\item $holds(p,q_c,c,k+1)$ if $q_c=m_{M_k(p)}(c)+\sum_{t \in T_l, p \in t \bullet, 0 \leq l \leq k, l+D(t)-1=k}{m_{W(t,p)}(c)}-(\sum_{t \in T_k, p \in \bullet t}{m_{W(p,t)}(c)}+ \sum_{t \in T_k, p \in R(t)}{m_{M_k(p)}(c)})$ represented by $q_c=m_{M_1(p)}(c)$ for some $c \in C$ -- by construction of $\Pi^7$,
\item and $could\_not\_have(t,k)$ if
\begin{enumerate}
\item $(\forall p \in \bullet t, W(p,t) \leq M_k(p)), (\forall p \in I(t), M_k(p) = 0), (\forall (p,t) \in Q, \\ M_k(p) \geq WQ(p,t)), \nexists tt \in T, (\forall pp \in \bullet tt, W(pp,tt) \leq M_{ts}(pp)), (\forall pp \in I(tt), M_k(pp) = 0), (\forall (pp,tt) \in Q, M_k(pp) \geq QW(pp,tt)), Z(tt) < Z(t)$,
\item and $q_c > m_{M_k(s)}(c) - (\sum_{t' \in T_k, s \in \bullet t'}{m_{W(s,t')}(c)}+ \sum_{t' \in T_k, s \in R(t)}{m_{M_k(s)}(c)}), \\ q_c = m_{W(s,t)}(c) \text{~if~} s \in \bullet t \text{~or~} m_{M_k(s)}(c) \text{~otherwise}$ for some $c \in C$, which becomes $q > M_k(s) - (\sum_{t' \in T_k, s \in \bullet t'}{W(s,t')}+ \sum_{t' \in T_k, s \in R(t)}{M_k(s)}), q = W(s,t) \text{~if~} s \in \bullet t \text{~or~} M_k(s) \text{~otherwise}$ for all $c \in C$
\item (i), (ii) above combined match the definition of $A_{25}$
\end{enumerate}
\item $X_{9k+9}$ does not contain $could\_not\_have(t,k)$, when $prenabled(t,k) \in X_0 \cup \dots \cup X_{9k+6}$ and $fires(t,k) \notin X_0 \cup \dots \cup X_{9k+5}$ due to construction of $A$, encoding of $a\ref{a:c:maxfire:cnh}$ and its body atoms. As a result it is not eliminated by  the constraint $a\ref{a:c:prmaxfire:elim}$
\end{enumerate}

\item $eval_{U_{9k+9}}(bot_{U_{9k+10}}(\Pi^7) \setminus bot_{U_{9k+9}}(\Pi^7), X_0 \cup \dots \cup X_{9k+9}) = \{ consumesmore \text{:-}. | \\ \{ consumesmore(p,0),\dots,$ $consumesmore(p,k) \} \cap (X_0 \cup \dots \cup X_{9k+9}) \neq \emptyset \}$. It's answer set is $X_{9k+10} = A_{20}$ -- using forced atom proposition and the definition of $A_{20}$
\begin{enumerate}
\item $X_{9k+10}$ will be empty since none of $consumesmore(p,0),\dots, \\ consumesmore(p,k)$ hold in $X_0 \cup \dots \cup X_{9k+10}$ due to the construction of $A$, encoding of $a\ref{a:overc:place}$ and its body atoms. As a result, it is not eliminated by the constraint $a\ref{a:overc:elim}$
\end{enumerate}

\end{enumerate}

The set $X = X_0 \cup \dots \cup X_{9k+10}$ is the answer set of $\Pi^7$ by the splitting sequence theorem~\ref{def:split_seq_thm}. Each $X_i, 0 \leq i \leq 9k+10$ matches a distinct portion of $A$, and $X = A$, thus $A$ is an answer set of $\Pi^7$.

\vspace{30pt}
\noindent
{\bf Next we show (\ref{prove:a2x:pndur}):} Given $\Pi^7$ be the encoding of a Petri Net $PN(P,T,E,C,W,R,I,$ $Q,QW,Z,D)$ with initial marking $M_0$, and $A$ be an answer set of $\Pi^7$ that satisfies (\ref{eqn:pndur:fires}) and (\ref{eqn:pndur:holds}), then we can construct $X=M_0,T_k,\dots,M_k,T_k,M_{k+1}$ from $A$, such that it is an execution sequence of $PN$.

We construct the $X$ as follows:
\begin{enumerate}
\item $M_i = (M_i(p_0), \dots, M_i(p_n))$, where $\{ holds(p_0,m_{M_i(p_0)}(c),c,i), \dots,\\ holds(p_n,m_{M_i(p_n)}(c),c,i) \} \subseteq A$, for $c \in C, 0 \leq i \leq k+1$
\item $T_i = \{ t : fires(t,i) \in A\}$, for $0 \leq i \leq k$ 
\end{enumerate}
and show that $X$ is indeed an execution sequence of $PN$. We show this by induction over $k$ (i.e. given $M_k$, $T_k$ is a valid firing set and its firing produces marking $M_{k+1}$).

\vspace{20pt}
\noindent
{\bf Base case:} Let $k=0$, and $M_0$ is a valid marking in $X$ for $PN$, show
\begin{inparaenum}[(1)]
\item $T_0$ is a valid firing set for $M_0$, and 
\item firing $T_0$ in $M_0$ produces marking $M_1$.
\end{inparaenum} 

\begin{enumerate}
\item We show $T_0$ is a valid firing set for $M_0$. Let $\{ fires(t_0,0), \dots, fires(t_x,0) \}$ be the set of all $fires(\dots,0)$ atoms in $A$,\label{pndur:prove:fires_t0}

\begin{enumerate}
\item Then for each $fires(t_i,0) \in A$

\begin{enumerate}
\item $prenabled(t_i,0) \in A$ -- from rule $a\ref{a:c:prfires}$ and supported rule proposition
\item Then $enabled(t_i,0) \in A$ -- from rule $a\ref{a:c:prenabled}$ and supported rule proposition
\item And $notprenabled(t_i,0) \notin A$ -- from rule $a\ref{a:c:prenabled}$ and supported rule proposition

\item For $enabled(t_i,0) \in A$
\begin{enumerate}
\item $notenabled(t_i,0) \notin A$ -- from rule $e\ref{e:c:enabled}$ and supported rule proposition
\item Then either of $body(e\ref{e:c:ne:ptarc})$, $body(e\ref{e:c:ne:iptarc})$, or $body(e\ref{e:c:ne:tptarc})$ must not hold in $A$ for $t_i$ -- from rules $body(e\ref{e:c:ne:ptarc}), body(e\ref{e:c:ne:iptarc}), body(e\ref{e:c:ne:tptarc})$ and forced atom proposition
\item Then $q_c \not< {n_i}_c \equiv q_c \geq {n_i}_c$ in $e\ref{e:c:ne:ptarc}$ for all $\{holds(p,q_c,c,0), \\ ptarc(p,t_i,{n_i}_c,c,0)\} \subseteq A$ -- from $e\ref{e:c:ne:ptarc}$, forced atom proposition, and given facts ($holds(p,q_c,c,0) \in A, ptarc(p,t_i,{n_i}_c,0) \in A$)
\item And $q_c \not\geq {n_i}_c \equiv q_c < {n_i}_c$ in $e\ref{e:c:ne:iptarc}$ for all $\{ holds(p,q_c,c,0), \\ iptarc(p,t_i,{n_i}_c,c,0) \} \subseteq A, {n_i}_c=1$; $q_c > {n_i}_c \equiv q_c = 0$ -- from $e\ref{e:c:ne:iptarc}$, forced atom proposition, given facts ($holds(p,q_c,c,0) \in A, \\ iptarc(p,t_i,1,c,0) \in A$), and $q_c$ is a positive integer
\item And $q_c \not< {n_i}_c \equiv q_c \geq {n_i}_c$ in $e\ref{e:c:ne:tptarc}$ for all $\{ holds(p,q_c,c,0), \\ tptarc(p,t_i,{n_i}_c,c,0) \} \subseteq A$ -- from $e\ref{e:c:ne:tptarc}$, forced atom proposition, and given facts
\item Then $\forall c \in C, (\forall p \in \bullet t_i, m_{M_0(p)}(c) \geq m_{W(p,t_i)}(c)) \wedge (\forall p \in I(t_i), \\ m_{M_0(p)}(c) = 0) \wedge (\forall (p,t_i) \in Q, m_{M_0(p)}(c) \geq m_{QW(p,t_i)}(c))$ -- from $i\ref{i:c:init},f\ref{f:c:ptarc},f\ref{f:c:rptarc}$ construction, definition~\ref{def:pndur:preset} of preset $\bullet t_i$ in $PN$, definition~\ref{def:pndur:enable} of enabled transition in $PN$, and that the construction of reset arcs by $f\ref{f:c:rptarc}$ ensures $notenabled(t,0)$ is never true for a reset arc, where $holds(p,q_c,c,0) \in A$ represents $q_c=m_{M_0(p)}(c)$, $ptarc(p,t_i,{n_i}_c,0) \in A$ represents ${n_i}_c=m_{W(p,t_i)}(c)$, ${n_i}_c = m_{M_0(p)}(c)$.
\item Which is equivalent to $(\forall p \in \bullet t_i, M_0(p) \geq W(p,t_i)) \wedge (\forall p \in I(t_i), M_0(p) = 0) \wedge (\forall (p,t_i) \in Q, M_0(p) \geq QW(p,t_i))$ -- assuming multiset domain $C$
\end{enumerate}

\item For $notprenabled(t_i,0) \notin A$
\begin{enumerate}
\item Either $(\nexists enabled(tt,0) \in A : pp < p_i)$ or $(\forall enabled(tt,0) \in A : pp \not< p_i)$ where $pp = Z(tt), p_i = Z(t_i)$ -- from rule $a\ref{a:c:prne}, f\ref{f:c:pr}$ and forced atom proposition
\item This matches the definition of an enabled priority transition
\end{enumerate}

\item Then $t_i$ is enabled and can fire in $PN$, as a result it can belong to $T_0$ -- from definition~\ref{def:pndur:enable} of enabled transition

\end{enumerate}
\item And $consumesmore \notin A$, since $A$ is an answer set of $\Pi^7$ -- from rule $a\ref{a:c:overc:elim}$ and supported rule proposition
\begin{enumerate}
\item Then $\nexists consumesmore(p,0) \in A$ -- from rule $a\ref{a:c:overc:gen}$ and supported rule proposition
\item  Then $\nexists \{ holds(p,q_c,c,0), tot\_decr(p,q1_c,c,0) \} \subseteq A, q1_c>q_c$ in $body(a\ref{a:c:overc:place})$ -- from $a\ref{a:c:overc:place}$ and forced atom proposition
\item Then $\nexists c \in C \nexists p \in P, (\sum_{t_i \in \{t_0,\dots,t_x\}, p \in \bullet t_i}{m_{W(p,t_i)}(c)}+ \\ \sum_{t_i \in \{t_0,\dots,t_x\}, p \in R(t_i)}{m_{M_0(p)}(c)}) > m_{M_0(p)}(c)$ -- from the following
\begin{enumerate}
\item $holds(p,q_c,c,0)$ represents $q_c=m_{M_0(p)}(c)$ -- from rule $i\ref{i:c:init}$ encoding, given
\item $tot\_decr(p,q1_c,c,0) \in A$ if $\{ del(p,{q1_0}_c,t_0,c,0), \dots, \\ del(p,{q1_x}_c,t_x,c,0) \} \subseteq A$, where $q1_c = {q1_0}_c+\dots+{q1_x}_c$ -- from $r\ref{r:c:totdecr}$ and forced atom proposition
\item $del(p,{q1_i}_c,t_i,c,0) \in A$ if $\{ fires(t_i,0), ptarc(p,t_i,{q1_i}_c,c,0) \} \subseteq A$ -- from $r\ref{r:c:del}$ and supported rule proposition
\item $del(p,{q1_i}_c,t_i,c,0)$ represents removal of ${q1_i}_c = m_{W(p,t_i)}(c)$ tokens from $p \in \bullet t_i$; or it represents removal of ${q1_i}_c = m_{M_0(p)}(c)$ tokens from $p \in R(t_i)$-- from rules $r\ref{r:c:del},f\ref{f:c:ptarc},f\ref{f:c:rptarc}$, supported rule proposition, and definition~\ref{def:pndur:preset} of preset in $PN$
\end{enumerate}
\item Then the set of transitions in $T_0$ do not conflict -- by the definition~\ref{def:pndur:conflict} of conflicting transitions
\end{enumerate}

\item And for each $prenabled(t_j,0) \in A$ and $fires(t_j,0) \notin A$, \\ $could\_not\_have(t_j,0) \in A$, since $A$ is an answer set of $\Pi^7$ - from rule $a\ref{a:c:prmaxfire:elim}$ and supported rule proposition
\begin{enumerate}
\item Then $\{ prenabled(t_j,0), holds(s,qq_c,c,0), ptarc(s,t_j,q_c,c,0), \\ tot\_decr(s,qqq_c,c,0) \} \subseteq A$, such that $q_c > qq_c - qqq_c$ and $fires(t_j,0) \notin A$ - from rule $a\ref{a:c:prmaxfire:cnh}$ and supported rule proposition
\item Then for an $s \in \bullet t_j \cup R(t_j)$, $q_c > m_{M_0(s)}(c) - (\sum_{t_i \in T_0, s \in \bullet t_i}{m_{W(s,t_i)}(c)} + \sum_{t_i \in T_0, s \in R(t_i)}{m_{M_0(s)}(c))}$, where $q_c=m_{W(s,t_j)}(c) \text{~if~} s \in \bullet t_j, \text{~or~} m_{M_0(s)}(c)$ $ \text{~otherwise}$. %
\begin{enumerate}
\item $ptarc(s,t_i,q_c,c,0)$ represents $q_c=m_{W(s,t_i)}(c)$ if $(s,t_i) \in E^-$ or $q_c=m_{M_0(s)}(c)$ if $s \in R(t_i)$ -- from rule $f\ref{f:c:ptarc},f\ref{f:c:rptarc}$ construction
\item $holds(s,qq_c,c,0)$ represents $qq_c=m_{M_0(s)}(c)$ -- from $i\ref{i:c:init}$ construction
\item $tot\_decr(s,qqq_c,c,0) \in A$ if $\{ del(s,{qqq_0}_c,t_0,c,0), \dots, \\ del(s,{qqq_x}_c,t_x,c,0) \} \subseteq A$ -- from rule $r\ref{r:c:totdecr}$ construction and supported rule proposition
\item $del(s,{qqq_i}_c,t_i,c,0) \in A$ if $\{ fires(t_i,0), ptarc(s,t_i,{qqq_i}_c,c,0) \} \subseteq A$ -- from rule $r\ref{r:c:del}$ and supported rule proposition
\item $del(s,{qqq_i}_c,t_i,c,0)$ either represents ${qqq_i}_c = m_{W(s,t_i)}(c) : t_i \in T_0, (s,t_i) \in E^-$, or ${qqq_i}_c = m_{M_0(t_i)}(c) : t_i \in T_0, s \in R(t_i)$ -- from rule $f\ref{f:c:ptarc},f\ref{f:c:rptarc}$ construction 
\item $tot\_decr(q,qqq_c,c,0)$ represents $\sum_{t_i \in T_0, s \in \bullet t_i}{m_{W(s,t_i)}(c)} + \\ \sum_{t_i \in T_0, s \in R(t_i)}{m_{M_0(s)}(c)}$ -- from (C,D,E) above 
\end{enumerate}

\item Then firing $T_0 \cup \{ t_j \}$ would have required more tokens than are present at its source place $s \in \bullet t_j \cup R(t_j)$. Thus, $T_0$ is a maximal set of transitions that can simultaneously fire.
\end{enumerate}

\item And for each reset transition $t_r$ with $prenabled(t_r,0) \in A$, $fires(t_r,0) \in A$, since $A$ is an answer set of $\Pi^7$ - from rule $f\ref{f:c:pr:rptarc:elim}$ and supported rule proposition
\begin{enumerate}
\item Then, the firing set $T_0$ satisfies the reset-transition requirement of definition~\ref{def:pndur:firing_set} (firing set)
\end{enumerate}

\item Then $\{t_0, \dots, t_x\} = T_0$ -- using 1(a),1(b),1(d) above; and using 1(c) it is a maximal firing set  
\end{enumerate}

\item Let $holds(p,q_c,c,1) \in A$
\begin{enumerate}
\item Then $\{ holds(p,q1_c,c,0), tot\_incr(p,q2_c,c,0), tot\_decr(p,q3_c,c,0) \} \subseteq A : q_c=q1_c+q2_c-q3_c$ -- from rule $r\ref{r:c:nextstate}$ and supported rule proposition \label{pndur:x:1:base}
\item \label{pndur:x:2:base} Then, $holds(p,q1_c,c,0) \in A$ represents $q1_c=m_{M_0(p)}(c)$ -- given ;

and $\{add(p,{q2_0}_c,t_0,c,0), \dots, $ $add(p,{q2_j}_c,t_j,c,0)\} \subseteq A : {q2_0}_c + \dots + {q2_j}_c = q2_c$  \label{pndur:stmt:add:base} and $\{del(p,{q3_0}_c,t_0,c,0), \dots, $ $del(p,{q3_l}_c,t_l,c,0)\} \subseteq A : {q3_0}_c + \dots + {q3_l}_c = q3_c$ \label{pndur:stmt:del:base}  -- rules $r\ref{r:c:totincr},r\ref{r:c:totdecr}$ and supported rule proposition, respectively
\item Then $\{ fires(t_0,0), \dots, fires(t_j,0) \} \subseteq A$ and $\{ fires(t_0,0), \dots, fires(t_l,0) \} \\ \subseteq A$ -- rules $r\ref{r:c:dur:add},r\ref{r:c:del}$ and supported rule proposition, respectively
\item Then $\{ fires(t_0,0), \dots, $ $fires(t_j,0) \} \cup \{ fires(t_0,0), \dots, $ $fires(t_l,0) \} \subseteq $ $A = \{ fires(t_0,0), \dots, $ $fires(t_x,0) \} \subseteq A$ -- set union of subsets
\item Then for each $fires(t_x,0) \in A$ we have $t_x \in T_0$ -- already shown in item~\ref{pndur:prove:fires_t0} above
\item Then $q_c = m_{M_0(p)}(c) + \sum_{t_x \in T_0, p \in t_x \bullet, 0+D(t_x)-1=0}{m_{W(t_x,p)}(c)} - \\ (\sum_{t_x \in T_0 \wedge p \in \bullet t_x}{m_{W(p,t_x)}(c)} + $ $\sum_{t_x \in T_0 \wedge p \in R(t_x)}{m_{M_0(p)}(c)})$ -- from 
\eqref{pndur:x:2:base} above and the following
\begin{enumerate}
\item Each $add(p,{q_j}_c,t_j,c,0) \in A$ represents ${q_j}_c=m_{W(t_j,p)}(c)$ for $p \in t_j \bullet$ -- rule $r\ref{r:c:dur:add},f\ref{f:c:dur:tparc}$ encoding, and definition~\ref{def:pndur:texec} of transition execution in $PN$ %
\item Each $del(p,t_y,{q_y}_c,c,0) \in A$ represents either ${q_y}_c=m_{W(p,t_y)}(c)$ for $p \in \bullet t_y$, or ${q_y}_c=m_{M_0(p)}(c)$ for $p \in R(t_y)$ -- from rule $r\ref{r:c:del},f\ref{f:c:ptarc}$ encoding and definition~\ref{def:pndur:texec} of transition execution in $PN$; or from rule $r\ref{r:c:del},f\ref{f:c:rptarc}$ encoding and definition of reset arc in $PN$
\item Each $tot\_incr(p,q2_c,c,0) \in A$ represents \\ $q2_c=\sum_{t_x \in T_0 \wedge p \in t_x  \bullet, 0+D(t_x)-1=0}{m_{W(t_x,p)}(c)}$ -- aggregate assignment atom semantics in rule $r\ref{r:c:totincr}$
\item Each $tot\_decr(p,q3_c,c,0) \in A$ represents $q3_c=\sum_{t_x \in T_0 \wedge p \in \bullet t_x}{m_{W(p,t_x)}(c)} \\ + \sum_{t_x \in T_0 \wedge p \in R(t_x)}{m_{M_0(p)}(c)}$ -- aggregate assignment atom semantics in rule $r\ref{r:c:totdecr}$
\end{enumerate}
\item Then, $m_{M_1(p)}(c) = q_c$ -- since $holds(p,q_c,c,1) \in A$ encodes $q_c=m_{M_1(p)}(c)$ -- from construction
\end{enumerate}
\end{enumerate}

\noindent
{\bf Inductive Step:} Let $k > 0$, and $M_k$ is a valid marking in $X$ for $PN$, show 
\begin{inparaenum}[(1)]
\item $T_k$ is a valid firing set for $M_k$, and 
\item firing $T_k$ in $M_k$ produces marking $M_{k+1}$.
\end{inparaenum}

\begin{enumerate}
\item We show that $T_k$ is a valid firing set in $M_k$. Let $\{ fires(t_0,k), \dots, fires(t_x,k) \}$ be the set of all $fires(\dots,k)$ atoms in $A$,\label{pndur:prove:fires_tk}

\begin{enumerate}
\item We show that $T_k$ is a valid firing set in $M_k$. Then for each $fires(t_i,k) \in A$

\begin{enumerate}
\item $prenabled(t_i,k) \in A$ -- from rule $a\ref{a:c:prfires}$ and supported rule proposition
\item Then $enabled(t_i,k) \in A$ -- from rule $a\ref{a:c:prenabled}$ and supported rule proposition
\item And $notprenabled(t_i,k) \notin A$ -- from rule $a\ref{a:c:prenabled}$ and supported rule proposition

\item For $enabled(t_i,k) \in A$
\begin{enumerate}
\item $notenabled(t_i,k) \notin A$ -- from rule $e\ref{e:c:enabled}$ and supported rule proposition
\item Then either of $body(e\ref{e:c:ne:ptarc})$, $body(e\ref{e:c:ne:iptarc})$, or $body(e\ref{e:c:ne:tptarc})$ must not hold in $A$ for $t_i$ -- from rules $body(e\ref{e:c:ne:ptarc}), body(e\ref{e:c:ne:iptarc}), body(e\ref{e:c:ne:tptarc})$ and forced atom proposition
\item Then $q_c \not< {n_i}_c \equiv q_c \geq {n_i}_c$ in $e\ref{e:c:ne:ptarc}$ for all $\{holds(p,q_c,c,k), \\ ptarc(p,t_i,{n_i}_c,c,k)\} \subseteq A$ -- from $e\ref{e:c:ne:ptarc}$, forced atom proposition, and given facts ($holds(p,q_c,c,k) \in A, ptarc(p,t_i,{n_i}_c,k) \in A$)
\item And $q_c \not\geq {n_i}_c \equiv q_c < {n_i}_c$ in $e\ref{e:c:ne:iptarc}$ for all $\{ holds(p,q_c,c,k), \\ iptarc(p,t_i,{n_i}_c,c,k) \} \subseteq A, {n_i}_c=1$; $q_c > {n_i}_c \equiv q_c = 0$ -- from $e\ref{e:c:ne:iptarc}$, forced atom proposition, given facts ($holds(p,q_c,c,k) \in A, \\ iptarc(p,t_i,1,c,k) \in A$), and $q_c$ is a positive integer
\item And $q_c \not< {n_i}_c \equiv q_c \geq {n_i}_c$ in $e\ref{e:c:ne:tptarc}$ for all $\{ holds(p,q_c,c,k), \\ tptarc(p,t_i,{n_i}_c,c,k) \} \subseteq A$ -- from $e\ref{e:c:ne:tptarc}$, forced atom proposition, and given facts
\item Then $\forall c \in C, (\forall p \in \bullet t_i, m_{M_k(p)}(c) \geq m_{W(p,t_i)}(c)) \wedge (\forall p \in I(t_i), \\ m_{M_k(p)}(c) = 0) \wedge (\forall (p,t_i) \in Q, m_{M_k(p)}(c) \geq m_{QW(p,t_i)}(c))$ -- from the inductive assumption, $f\ref{f:c:ptarc},f\ref{f:c:rptarc}$ construction, definition~\ref{def:pndur:preset} of preset $\bullet t_i$ in $PN$, definition~\ref{def:pndur:enable} of enabled transition in $PN$, and that the construction of reset arcs by $f\ref{f:c:rptarc}$ ensures $notenabled(t,k)$ is never true for a reset arc, where $holds(p,q_c,c,k) \in A$ represents $q_c=m_{M_k(p)}(c)$, $ptarc(p,t_i,{n_i}_c,k) \in A$ represents ${n_i}_c=m_{W(p,t_i)}(c)$, ${n_i}_c = m_{M_k(p)}(c)$.
\item Which is equivalent to $(\forall p \in \bullet t_i, M_k(p) \geq W(p,t_i)) \wedge (\forall p \in I(t_i), M_k(p) = 0) \wedge (\forall (p,t_i) \in Q, M_k(p) \geq QW(p,t_i))$ -- assuming multiset domain $C$
\end{enumerate}

\item For $notprenabled(t_i,k) \notin A$
\begin{enumerate}
\item Either $(\nexists enabled(tt,k) \in A : pp < p_i)$ or $(\forall enabled(tt,k) \in A : pp \not< p_i)$ where $pp = Z(tt), p_i = Z(t_i)$ -- from rule $a\ref{a:c:prne}, f\ref{f:c:pr}$ and forced atom proposition
\item This matches the definition of an enabled priority transition
\end{enumerate}

\item Then $t_i$ is enabled and can fire in $PN$, as a result it can belong to $T_k$ -- from definition~\ref{def:pndur:enable} of enabled transition

\end{enumerate}
\item And $consumesmore \notin A$, since $A$ is an answer set of $\Pi^7$ -- from rule $a\ref{a:c:overc:elim}$ and supported rule proposition
\begin{enumerate}
\item Then $\nexists consumesmore(p,k) \in A$ -- from rule $a\ref{a:c:overc:gen}$ and supported rule proposition
\item  Then $\nexists \{ holds(p,q_c,c,k), tot\_decr(p,q1_c,c,k) \} \subseteq A, q1_c>q_c$ in $body(a\ref{a:c:overc:place})$ -- from $a\ref{a:c:overc:place}$ and forced atom proposition
\item Then $\nexists c \in C \nexists p \in P, (\sum_{t_i \in \{t_0,\dots,t_x\}, p \in \bullet t_i}{m_{W(p,t_i)}(c)}+ \\ \sum_{t_i \in \{t_0,\dots,t_x\}, p \in R(t_i)}{m_{M_k(p)}(c)}) > m_{M_k(p)}(c)$ -- from the following
\begin{enumerate}
\item $holds(p,q_c,c,k)$ represents $q_c=m_{M_k(p)}(c)$ -- from rule $PN$ encoding, given
\item $tot\_decr(p,q1_c,c,k) \in A$ if $\{ del(p,{q1_0}_c,t_0,c,k), \dots, \\ del(p,{q1_x}_c,t_x,c,k) \} \subseteq A$, where $q1_c = {q1_0}_c+\dots+{q1_x}_c$ -- from $r\ref{r:c:totdecr}$ and forced atom proposition
\item $del(p,{q1_i}_c,t_i,c,k) \in A$ if $\{ fires(t_i,k), ptarc(p,t_i,{q1_i}_c,c,k) \} \subseteq A$ -- from $r\ref{r:c:del}$ and supported rule proposition
\item $del(p,{q1_i}_c,t_i,c,k)$ either represents removal of ${q1_i}_c = m_{W(p,t_i)}(c)$ tokens from $p \in \bullet t_i$; or it represents removal of ${q1_i}_c = m_{M_k(p)}(c)$ tokens from $p \in R(t_i)$-- from rules $r\ref{r:c:del},f\ref{f:c:ptarc},f\ref{f:c:rptarc}$, supported rule proposition, and definition~\ref{def:pndur:texec} of transition execution in $PN$
\end{enumerate}
\item Then the set of transitions in $T_k$ do not conflict -- by the definition~\ref{def:pndur:conflict} of conflicting transitions
\end{enumerate}

\item And for each $prenabled(t_j,k) \in A$ and $fires(t_j,k) \notin A$, \\ $could\_not\_have(t_j,k) \in A$, since $A$ is an answer set of $\Pi^7$ - from rule $a\ref{a:c:prmaxfire:elim}$ and supported rule proposition
\begin{enumerate}
\item Then $\{ prenabled(t_j,k), holds(s,qq_c,c,k), ptarc(s,t_j,q_c,c,k), \\ tot\_decr(s,qqq_c,c,k) \} \subseteq A$, such that $q_c > qq_c - qqq_c$ and $fires(t_j,k) \notin A$ - from rule $a\ref{a:c:prmaxfire:cnh}$ and supported rule proposition
\item Then for an $s \in \bullet t_j \cup R(t_j)$, $q_c > m_{M_k(s)}(c) - (\sum_{t_i \in T_k, s \in \bullet t_i}{m_{W(s,t_i)}(c)} + \sum_{t_i \in T_k, s \in R(t_i)}{m_{M_k(s)}(c))}$, where $q_c=m_{W(s,t_j)}(c) \text{~if~} s \in \bullet t_j, \text{~or~} m_{M_k(s)}(c) $ $\text{~otherwise}$. %
\begin{enumerate}
\item $ptarc(s,t_i,q_c,c,k)$ represents $q_c=m_{W(s,t_i)}(c)$ if $(s,t_i) \in E^-$ or $q_c=m_{M_k(s)}(c)$ if $s \in R(t_i)$ -- from rule $f\ref{f:c:ptarc},f\ref{f:c:rptarc}$ construction
\item $holds(s,qq_c,c,k)$ represents $qq_c=m_{M_k(s)}(c)$ -- from $i\ref{i:c:init}$ construction
\item $tot\_decr(s,qqq_c,c,k) \in A$ if $\{ del(s,{qqq_0}_c,t_0,c,k), \dots, \\ del(s,{qqq_x}_c,t_x,c,k) \} \subseteq A$ -- from rule $r\ref{r:c:totdecr}$ construction and supported rule proposition
\item $del(s,{qqq_i}_c,t_i,c,k) \in A$ if $\{ fires(t_i,k), ptarc(s,t_i,{qqq_i}_c,c,k) \} \subseteq A$ -- from rule $r\ref{r:c:del}$ and supported rule proposition
\item $del(s,{qqq_i}_c,t_i,c,k)$ represents ${qqq_i}_c = m_{W(s,t_i)}(c) : t_i \in T_k, (s,t_i) \in E^-$, or ${qqq_i}_c = m_{M_k(t_i)}(c) : t_i \in T_k, s \in R(t_i)$ -- from rule $f\ref{f:c:ptarc},f\ref{f:c:rptarc}$ construction 
\item $tot\_decr(q,qqq_c,c,k)$ represents $\sum_{t_i \in T_k, s \in \bullet t_i}{m_{W(s,t_i)}(c)} + \\ \sum_{t_i \in T_k, s \in R(t_i)}{m_{M_k(s)}(c)}$ -- from (C,D,E) above 
\end{enumerate}

\item Then firing $T_k \cup \{ t_j \}$ would have required more tokens than are present at its source place $s \in \bullet t_j \cup R(t_j)$. Thus, $T_k$ is a maximal set of transitions that can simultaneously fire.
\end{enumerate}

\item And for each reset transition $t_r$ with $enabled(t_r,k) \in A$, $fires(t_r,k) \in A$, since $A$ is an answer set of $\Pi^7$ - from rule $f\ref{f:c:pr:rptarc:elim}$ and supported rule proposition
\begin{enumerate}
\item Then the firing set $T_k$ satisfies the reset transition requirement of definition~\ref{def:pndur:firing_set} (firing set)
\end{enumerate}

\item Then $\{t_0, \dots, t_x\} = T_k$ -- using 1(a),1(b), 1(d) above; and using 1(c) it is a maximal firing set  
\end{enumerate}

\item We show that $M_{k+1}$ is produced by firing $T_k$ in $M_k$. Let $holds(p,q_c,c,k+1) \in A$
\begin{enumerate}
\item Then $\{ holds(p,q1_c,c,k), tot\_incr(p,q2_c,c,k), tot\_decr(p,q3_c,c,k) \} \subseteq A : q_c=q1_c+q2_c-q3_c$ -- from rule $r\ref{r:c:nextstate}$ and supported rule proposition \label{pndur:x:1:induction}
\item \label{pndur:x:2:induction} Then $holds(p,q1_c,c,k) \in A$ represents $q1_c=m_{M_k(p)}(c)$ -- inductive assumption;  
and $\{add(p,{q2_0}_c,t_0,c,k), \dots, $ $add(p,{q2_j}_c,t_j,c,k)\} \subseteq A : {q2_0}_c + \dots + {q2_j}_c = q2_c$  \label{pndur:stmt:add:induction} and $\{del(p,{q3_0}_c,t_0,c,k), \dots, $ $del(p,{q3_l}_c,t_l,c,k)\} \subseteq A : {q3_0}_c + \dots + {q3_l}_c = q3_c$ \label{pndur:stmt:del:induction}  -- rules $r\ref{r:c:totincr},r\ref{r:c:totdecr}$ and supported rule proposition, respectively
\item Then $\{ fires(t_0,k), \dots, fires(t_j,k) \} \subseteq A$ and $\{ fires(t_0,k), \dots, fires(t_l,k) \} \\ \subseteq A$ -- rules $r\ref{r:c:dur:add},r\ref{r:c:del}$ and supported rule proposition, respectively
\item Then $\{ fires(t_0,k), \dots, fires(t_j,k) \} \cup \{ fires(t_0,k), \dots, fires(t_l,k) \} \subseteq A = \{ fires(t_0,k), \dots, fires(t_x,k) \} \subseteq A$ -- set union of subsets
\item Then for each $fires(t_x,k) \in A$ we have $t_x \in T_k$ -- already shown in item~\ref{pnpri:prove:fires_t0} above
\item Then $q_c = m_{M_k(p)}(c) + \sum_{t_x \in T_l, p \in t_x \bullet, 0 \leq l \leq k, l+D(t_x)-1=k}{m_{W(t_x,p)}(c)} - \\ (\sum_{t_x \in T_k \wedge p \in \bullet t_x}{m_{W(p,t_x)}(c)} + \sum_{t_x \in T_k \wedge p \in R(t_x)}{m_{M_k(p)}(c)})$ -- from 
\eqref{pnpri:x:2:induction} above and the following
\begin{enumerate}
\item Each $add(p,{q_j}_c,t_j,c,k) \in A$ represents ${q_j}_c=m_{W(t_j,p)}(c)$ for $p \in t_j \bullet$ -- rule $r\ref{r:c:dur:add},f\ref{f:c:dur:tparc}$ encoding, and definition~\ref{def:pndur:preset} of postset in $PN$ %
\item Each $del(p,t_y,{q_y}_c,c,k) \in A$ represents either ${q_y}_c=m_{W(p,t_y)}(c)$ for $p \in \bullet t_y$, or ${q_y}_c=m_{M_k(p)}(c)$ for $p \in R(t_y)$ -- from rule $r\ref{r:c:del},f\ref{f:c:ptarc}$ encoding and definition~\ref{def:pndur:preset} of preset in $PN$; or from rule $r\ref{r:c:del},f\ref{f:c:rptarc}$ encoding and definition of reset arc in $PN$
\item Each $tot\_incr(p,q2_c,c,k) \in A$ represents \\ $q2_c=\sum_{t_x \in T_l, p \in t_x  \bullet, 0 \leq l \leq k, l+D(t_x)-1=k}{m_{W(t_x,p)}(c)}$ -- aggregate assignment atom semantics in rule $r\ref{r:c:totincr}$
\item Each $tot\_decr(p,q3_c,c,k) \in A$ represents $q3_c=\sum_{t_x \in T_k \wedge p \in \bullet t_x}{m_{W(p,t_x)}(c)} \\ + \sum_{t_x \in T_k \wedge p \in R(t_x)}{m_{M_k(p)}(c)}$ -- aggregate assignment atom semantics in rule $r\ref{r:c:totdecr}$
\end{enumerate}
\item Then, $m_{M_{k+1}(p)}(c) = q_c$ -- since $holds(p,q_c,c,k+1) \in A$ encodes $q_c=m_{M_{k+1}(p)}(c)$ -- from construction 
\end{enumerate}
\end{enumerate}

\noindent
As a result, for any $n > k$, $T_n$ will be a valid firing set for $M_n$ and $M_{n+1}$ will be its target marking. 

\noindent
{\bf Conclusion:} Since both \eqref{prove:x2a:pndur} and \eqref{prove:a2x:pndur} hold, $X=M_0,T_k,M_1,\dots,M_k,T_{k+1}$ is an execution sequence of $PN(P,T,E,C,W,R,I,Q,QW,Z,D)$ (w.r.t $M_0$) iff there is an answer set $A$ of $\Pi^7(PN,M_0,k,ntok)$ such that \eqref{eqn:pndur:fires} and \eqref{eqn:pndur:holds} hold.

\chapter[Drug-Drug Interaction Queries]{Complete Set of Queries Used for Drug-Drug Interaction}

\section{Drug Activates Gene}
\begin{lstlisting}[breaklines,numbers=left,basicstyle=\footnotesize]
//S{/NP{//?[Value='activation'](kw1)=>//?[Value='of'](kw2)=>//?[Tag='GENE'](kw0)=>//?[Value='by'](kw4)=>//?[Tag='DRUG'](kw3)}} ::: distinct sent.cid, kw3.value, kw1.value, kw0.value, sent.value
//NP{/NP{/?[Tag='DRUG'](kw3)=>/?[Value='activation'](kw1)}=>/PP{/?[Value='of'](kw2)=>//?[Tag='GENE'](kw0)}} ::: distinct sent.cid, kw3.value, kw1.value, kw0.value, sent.value
//NP{/?[Value='activation'](kw1)=>/PP{/?[Value='of'](kw2)=>//?[Tag='GENE'](kw0)=>//?[Value='by'](kw4)=>//?[Tag='DRUG'](kw3)}} ::: distinct sent.cid, kw3.value, kw1.value, kw0.value, sent.value
//S{/NP{//?[Tag='GENE'](kw2)}=>/VP{//?[Value='activation'](kw1)=>//?[Tag='DRUG'](kw0)}} ::: distinct sent.cid, kw0.value, kw1.value, kw2.value, sent.value
//S{/NP{//?[Tag='DRUG'](kw0)}=>/VP{//?[Value='activation'](kw1)=>//?[Tag='GENE'](kw2)}} ::: distinct sent.cid, kw0.value, kw1.value, kw2.value, sent.value
//NP{/NP{/?[Tag='DRUG'](kw0)=>/?[Value='activation'](kw1)}=>/PP{//?[Tag='GENE'](kw2)}} ::: distinct sent.cid, kw0.value, kw1.value, kw2.value, sent.value
//S{/NP{//?[Value='activation'](kw1)=>//?[Tag='GENE'](kw2)}=>/VP{//?[Tag='DRUG'](kw0)}} ::: distinct sent.cid, kw0.value, kw1.value, kw2.value, sent.value
//S{/NP{//?[Value='activation'](kw1)=>//?[Value='of'](kw2)=>//?[Tag='GENE'](kw0)=>//?[Value='by'](kw4)=>//?[Tag='DRUG'](kw3)}} ::: distinct sent.cid, kw3.value, kw1.value, kw0.value, sent.value
//NP{/NP{/?[Tag='DRUG'](kw3)=>/?[Value='activation'](kw1)}=>/PP{/?[Value='of'](kw2)=>//?[Tag='GENE'](kw0)}} ::: distinct sent.cid, kw3.value, kw1.value, kw0.value, sent.value
//NP{/?[Value='activation'](kw1)=>/PP{/?[Value='of'](kw2)=>//?[Tag='GENE'](kw0)=>//?[Value='by'](kw4)=>//?[Tag='DRUG'](kw3)}} ::: distinct sent.cid, kw3.value, kw1.value, kw0.value, sent.value
\end{lstlisting}

\section{Gene Induces Gene}
\begin{lstlisting}[breaklines,numbers=left,basicstyle=\footnotesize]
//S{/NP{//?[Tag='GENE'](kw0)}=>/VP{//?[Value='induced'](kw1)=>//?[Value='by'](kw3)=>//?[Tag='GENE'](kw2)}} ::: distinct sent.cid, kw0.value, kw1.value, kw2.value, sent.value
//S{/NP{/?[Tag='GENE'](kw0)}=>/VP{//?[Value='induced'](kw1)=>//?[Value='by'](kw3)=>//?[Tag='GENE'](kw2)}} ::: distinct sent.cid, kw0.value, kw1.value, kw2.value, sent.value
//S{/NP{//?[Tag='GENE'](kw0)}=>/VP{//?[Value IN {'increase','increased'}](kw1)=>//?[Tag='GENE'](kw2)=>//?[Value='activity'](kw3)}} ::: distinct sent.cid, kw0.value, kw1.value, kw2.value, sent.value
//S{/NP{//?[Tag='GENE'](kw0)=>//?[Value='activity'](kw3)}=>/VP{//?[Value='increase'](kw1)=>//?[Tag='GENE'](kw2)}} ::: distinct sent.cid, kw0.value, kw1.value, kw2.value, sent.value
//S{/NP{//?[Tag='GENE'](kw0)}=>/VP{//?[Value IN {'stimulates','stimulate', 'stimulated'}](kw1)=>//?[Tag='GENE'](kw2)}} ::: distinct sent.cid, kw0.value, kw1.value, kw2.value, sent.value
//S{/NP{//?[Tag='GENE'](kw0)}=>/VP{/?[Value IN {'stimulates','stimulate'}](kw1)=>//?[Tag='GENE'](kw2)}} ::: distinct sent.cid, kw0.value, kw1.value, kw2.value, sent.value
//S{/NP{/?[Tag='GENE'](kw0)}=>/VP{/?[Value='stimulates'](kw1)=>//?[Tag='GENE'](kw2)}} ::: distinct sent.cid, kw0.value, kw1.value, kw2.value, sent.value
//S{/NP{//?[Tag='GENE'](kw0)}=>/VP{/?[Value='activated'](kw1)=>//?[Tag='GENE'](kw2)}} ::: distinct sent.cid, kw0.value, kw1.value, kw2.value, sent.value
//NP{/NP{/?[Value='induction'](kw1)}=>/PP{/?[Value='of'](kw3)=>/NP{//?[Tag='GENE'](kw0)=>//?[Value='by'](kw4)=>//?[Tag='GENE'](kw2)}}} ::: distinct sent.cid, kw0.value, kw1.value, kw2.value, sent.value
//NP{/NP{/?[Value='stimulation'](kw1)}=>/PP{/?[Value='of'](kw3)=>/NP{//?[Tag='GENE'](kw0)=>//?[Value='by'](kw4)=>//?[Tag='GENE'](kw2)}}} ::: distinct sent.cid, kw0.value, kw1.value, kw2.value, sent.value
//NP{/NP{/?[Value='activation'](kw1)}=>/PP{/?[Value IN {'of'}](kw3)=>/NP{//?[Tag='GENE'](kw0)=>//?[Value='by'](kw4)=>//?[Tag='GENE'](kw2)}}} ::: distinct sent.cid, kw0.value, kw1.value, kw2.value, sent.value
//S{/NP{//?[Tag='GENE'](kw0)}=>/VP{//?[Value='inducible'](kw1)=>//?[Tag='GENE'](kw2)}} ::: distinct sent.cid, kw0.value, kw1.value, kw2.value, sent.value
//S{/NP{//?[Tag='GENE'](kw0)}=>/VP{//?[Value='induced'](kw1)=>//?[Value='by'](kw3)=>//?[Tag='GENE'](kw2)}} ::: distinct sent.cid, kw0.value, kw1.value, kw2.value, sent.value
//S{/NP{/?[Tag='GENE'](kw0)}=>/VP{//?[Value='induced'](kw1)=>//?[Value='by'](kw3)=>//?[Tag='GENE'](kw2)}} ::: distinct sent.cid, kw0.value, kw1.value, kw2.value, sent.value
//S{/NP{//?[Tag='GENE'](kw0)}=>/VP{//?[Value IN {'increase','increased'}](kw1)=>//?[Tag='GENE'](kw2)=>//?[Value='activity'](kw3)}} ::: distinct sent.cid, kw0.value, kw1.value, kw2.value, sent.value
//S{/NP{//?[Tag='GENE'](kw0)=>//?[Value='activity'](kw3)}=>/VP{//?[Value='increase'](kw1)=>//?[Tag='GENE'](kw2)}} ::: distinct sent.cid, kw0.value, kw1.value, kw2.value, sent.value
//S{/NP{//?[Tag='GENE'](kw0)}=>/VP{//?[Value='stimulated'](kw1)=>//?[Tag='GENE'](kw2)}} ::: distinct sent.cid, kw0.value, kw1.value, kw2.value, sent.value
//S{/NP{//?[Tag='GENE'](kw0)}=>/VP{/?[Value IN {'stimulates','stimulate'}](kw1)=>//?[Tag='GENE'](kw2)}} ::: distinct sent.cid, kw0.value, kw1.value, kw2.value, sent.value
//S{/NP{/?[Tag='GENE'](kw0)}=>/VP{/?[Value='stimulates'](kw1)=>//?[Tag='GENE'](kw2)}} ::: distinct sent.cid, kw0.value, kw1.value, kw2.value, sent.value
//S{/NP{//?[Tag='GENE'](kw0)}=>/VP{/?[Value='activated'](kw1)=>//?[Tag='GENE'](kw2)}} ::: distinct sent.cid, kw0.value, kw1.value, kw2.value, sent.value
//VP{/?[Value='activated'](kw1)=>/PP{//?[Tag='GENE'](kw0)=>//?[Tag='GENE'](kw2)}} ::: distinct sent.cid, kw0.value, kw1.value, kw2.value, sent.value
//NP{/NP{/?[Value='induction'](kw1)}=>/PP{/?[Value='of'](kw3)=>/NP{//?[Tag='GENE'](kw0)=>//?[Value='by'](kw4)=>//?[Tag='GENE'](kw2)}}} ::: distinct sent.cid, kw0.value, kw1.value, kw2.value, sent.value
//NP{/NP{/?[Value='stimulation'](kw1)}=>/PP{/?[Value='of'](kw3)=>/NP{//?[Tag='GENE'](kw0)=>//?[Value='by'](kw4)=>//?[Tag='GENE'](kw2)}}} ::: distinct sent.cid, kw0.value, kw1.value, kw2.value, sent.value
//NP{/NP{/?[Value='activation'](kw1)}=>/PP{/?[Value IN {'of'}](kw3)=>/NP{//?[Tag='GENE'](kw0)=>//?[Value='by'](kw4)=>//?[Tag='GENE'](kw2)}}} ::: distinct sent.cid, kw0.value, kw1.value, kw2.value, sent.value
//S{/NP{//?[Tag='GENE'](kw0)}=>/VP{//?[Value='inducible'](kw1)=>//?[Tag='GENE'](kw2)}} ::: distinct sent.cid, kw0.value, kw1.value, kw2.value, sent.value
\end{lstlisting}

\section{Gene Inhibits Gene}
\begin{lstlisting}[breaklines,numbers=left,basicstyle=\footnotesize]
//S{/NP{//?[Tag='GENE'](kw0)}=>/VP{//?[Value='induced'](kw1)=>//?[Value='by'](kw3)=>//?[Tag='GENE'](kw2)}} ::: distinct sent.cid, kw0.value, kw1.value, kw2.value, sent.value
//S{/NP{/?[Tag='GENE'](kw0)}=>/VP{//?[Value='induced'](kw1)=>//?[Value='by'](kw3)=>//?[Tag='GENE'](kw2)}} ::: distinct sent.cid, kw0.value, kw1.value, kw2.value, sent.value
//S{/NP{//?[Tag='GENE'](kw0)}=>/VP{//?[Value IN {'increase','increased'}](kw1)=>//?[Tag='GENE'](kw2)=>//?[Value='activity'](kw3)}} ::: distinct sent.cid, kw0.value, kw1.value, kw2.value, sent.value
//S{/NP{//?[Tag='GENE'](kw0)=>//?[Value='activity'](kw3)}=>/VP{//?[Value='increase'](kw1)=>//?[Tag='GENE'](kw2)}} ::: distinct sent.cid, kw0.value, kw1.value, kw2.value, sent.value
//S{/NP{//?[Tag='GENE'](kw0)}=>/VP{//?[Value IN {'stimulates','stimulate', 'stimulated'}](kw1)=>//?[Tag='GENE'](kw2)}} ::: distinct sent.cid, kw0.value, kw1.value, kw2.value, sent.value
//S{/NP{//?[Tag='GENE'](kw0)}=>/VP{/?[Value IN {'stimulates','stimulate'}](kw1)=>//?[Tag='GENE'](kw2)}} ::: distinct sent.cid, kw0.value, kw1.value, kw2.value, sent.value
//S{/NP{/?[Tag='GENE'](kw0)}=>/VP{/?[Value='stimulates'](kw1)=>//?[Tag='GENE'](kw2)}} ::: distinct sent.cid, kw0.value, kw1.value, kw2.value, sent.value
//S{/NP{//?[Tag='GENE'](kw0)}=>/VP{/?[Value='activated'](kw1)=>//?[Tag='GENE'](kw2)}} ::: distinct sent.cid, kw0.value, kw1.value, kw2.value, sent.value
//NP{/NP{/?[Value='induction'](kw1)}=>/PP{/?[Value='of'](kw3)=>/NP{//?[Tag='GENE'](kw0)=>//?[Value='by'](kw4)=>//?[Tag='GENE'](kw2)}}} ::: distinct sent.cid, kw0.value, kw1.value, kw2.value, sent.value
//NP{/NP{/?[Value='stimulation'](kw1)}=>/PP{/?[Value='of'](kw3)=>/NP{//?[Tag='GENE'](kw0)=>//?[Value='by'](kw4)=>//?[Tag='GENE'](kw2)}}} ::: distinct sent.cid, kw0.value, kw1.value, kw2.value, sent.value
//NP{/NP{/?[Value='activation'](kw1)}=>/PP{/?[Value IN {'of'}](kw3)=>/NP{//?[Tag='GENE'](kw0)=>//?[Value='by'](kw4)=>//?[Tag='GENE'](kw2)}}} ::: distinct sent.cid, kw0.value, kw1.value, kw2.value, sent.value
//S{/NP{//?[Tag='GENE'](kw0)}=>/VP{//?[Value='inducible'](kw1)=>//?[Tag='GENE'](kw2)}} ::: distinct sent.cid, kw0.value, kw1.value, kw2.value, sent.value
\end{lstlisting}

\section{Drug Changes Gene Expression/Activity}
\begin{lstlisting}[breaklines,numbers=left,basicstyle=\footnotesize]
//S{/?[Tag='DRUG'](kw2)=>/?[Value IN {'increased','increase','increases'}](kw1)=>/?[Value IN {'levels','level'}](kw3)=>/?[Tag='GENE'](kw0)} ::: distinct sent.cid, kw0.value, kw1.value, kw2.value, sent.value
//S{/NP{/?[Tag='DRUG'](kw2)}=>/VP{/?[Value IN {'increased','increases'}](kw1)=>/NP{//?[Tag='GENE'](kw0)=>//?[Value IN {'expression','level','activity','activities','levels'}](kw3)}}} ::: distinct sent.cid, kw0.value, kw1.value, kw2.value, sent.value
//S{/?[Tag='DRUG'](kw2)=>/VP{/NP{//?[Value IN {'increases','increase'}](kw1)=>//?[Tag='GENE'](kw0)=>//?[Value IN {'levels','activity'}](kw3)}}} ::: distinct sent.cid, kw0.value, kw1.value, kw2.value, sent.value
//S{/?[Tag='DRUG'](kw2)=>/VP{/?[Value IN {'increased','increases'}](kw1)=>/NP{//?[Tag='GENE'](kw0)=>//?[Value IN {'activity','activities','levels'}](kw3)}}} ::: distinct sent.cid, kw0.value, kw1.value, kw2.value, sent.value
//S{/NP{/?[Tag='DRUG'](kw2)}=>/VP{/?[Value='increased'](kw1)=>/NP{//?[Value IN {'activity','levels','expression'}](kw3)=>//?[Tag='GENE'](kw0)}}} ::: distinct sent.cid, kw0.value, kw1.value, kw2.value, sent.value
//S{/?[Tag='DRUG'](kw2)=>/VP{/VP{//?[Value IN {'increased','increases','increase'}](kw1)=>//?[Tag='GENE'](kw0)=>//?[Value IN {'activity','levels'}](kw3)}}} ::: distinct sent.cid, kw0.value, kw1.value, kw2.value, sent.value
//S{/?[Tag='DRUG'](kw2)=>/VP{/VP{/?[Value='increased'](kw1)=>//?[Tag='GENE'](kw0)=>//?[Value IN {'expression','activity'}](kw3)}}} ::: distinct sent.cid, kw0.value, kw1.value, kw2.value, sent.value
//S{/?[Tag='DRUG'](kw2)=>/VP{/?[Value IN {'increased','increases'}](kw1)=>/NP{//?[Value IN {'expression','activity','activities','level','levels'}](kw3)=>//?[Tag='GENE'](kw0)}}} ::: distinct sent.cid, kw0.value, kw1.value, kw2.value, sent.value
//S{/?[Tag='DRUG'](kw2)=>/VP{/?[Value IN {'decreased','decreases'}](kw1)=>/NP{//?[Tag='GENE'](kw0)=>//?[Value IN {'activity','expression'}](kw3)}}} ::: distinct sent.cid, kw2.value, kw1,value, kw0.value, sent.value
//S{/?[Tag='DRUG'](kw2)=>/?[Value IN {'increased','increase','increases'}](kw1)=>/?[Value IN {'levels','level'}](kw3)=>/?[Tag='GENE'](kw0)} ::: distinct sent.cid, kw0.value, kw1.value, kw2.value, kw3.value, sent.value
//S{/NP{/?[Tag='DRUG'](kw2)}=>/VP{/?[Value IN {'increased','increases'}](kw1)=>/NP{//?[Tag='GENE'](kw0)=>//?[Value IN {'expression','level','activity','activities','levels'}](kw3)}}} ::: distinct sent.cid, kw0.value, kw1.value, kw2.value, kw3.value, sent.value
//S{/?[Tag='DRUG'](kw2)=>/VP{/NP{//?[Value IN {'increases','increase'}](kw1)=>//?[Tag='GENE'](kw0)=>//?[Value IN {'levels','activity'}](kw3)}}} ::: distinct sent.cid, kw0.value, kw1.value, kw2.value, kw3.value, sent.value
//S{/?[Tag='DRUG'](kw2)=>/VP{/?[Value IN {'increased','increases'}](kw1)=>/NP{//?[Tag='GENE'](kw0)=>//?[Value IN {'activity','activities','levels'}](kw3)}}} ::: distinct sent.cid, kw0.value, kw1.value, kw2.value, kw3.value, sent.value
//S{/NP{/?[Tag='DRUG'](kw2)}=>/VP{/?[Value='increased'](kw1)=>/NP{//?[Value IN {'activity','levels','expression'}](kw3)=>//?[Tag='GENE'](kw0)}}} ::: distinct sent.cid, kw0.value, kw1.value, kw2.value, kw3.value, sent.value
//S{/?[Tag='DRUG'](kw2)=>/VP{/VP{//?[Value IN {'increased','increases','increase'}](kw1)=>//?[Tag='GENE'](kw0)=>//?[Value IN {'activity','levels'}](kw3)}}} ::: distinct sent.cid, kw0.value, kw1.value, kw2.value, kw3.value, sent.value
//S{/?[Tag='DRUG'](kw2)=>/VP{/VP{/?[Value='increased'](kw1)=>//?[Tag='GENE'](kw0)=>//?[Value IN {'expression','activity'}](kw3)}}} ::: distinct sent.cid, kw0.value, kw1.value, kw2.value, kw3.value, sent.value
//S{/?[Tag='DRUG'](kw2)=>/VP{/?[Value IN {'increased','increases'}](kw1)=>/NP{//?[Value IN {'expression','activity','activities','level','levels'}](kw3)=>//?[Tag='GENE'](kw0)}}} ::: distinct sent.cid, kw0.value, kw1.value, kw2.value, kw3.value, sent.value
//S{/?[Tag='DRUG'](kw2)=>/VP{/?[Value IN {'decreased','decreases'}](kw1)=>/NP{//?[Tag='GENE'](kw0)=>//?[Value IN {'activity','expression'}](kw3)}}} ::: distinct sent.cid, kw0.value, kw1.value, kw2.value, kw3.value, sent.value
\end{lstlisting}

\section{Drug Induces/Stimulates Gene}
\begin{lstlisting}[breaklines,numbers=left,basicstyle=\footnotesize]
//S{/?[Tag='DRUG'](kw2)=>/VP{/?[Value IN {'stimulated','induced'}](kw1)=>/NP{//?[Tag='GENE'](kw0)}}} ::: distinct sent.cid, kw2.value, kw1.value, kw0.value, sent.value
//VP{/?[Value IN {'stimulated','induced'}](kw1)=>/PP{/NP{//?[Tag='DRUG'](kw2)=>//?[Tag='GENE'](kw0)}}} ::: distinct sent.cid, kw2.value, kw1.value, kw0.value, sent.value
//S{/NP{/?[Tag='DRUG'](kw2)}=>/VP{/?[Value IN {'stimulated','induced'}](kw1)=>/NP{//?[Tag='GENE'](kw0)}}} ::: distinct sent.cid, kw2.value, kw1.value, kw0.value, sent.value
//S{/NP{/?[Tag='GENE'](kw0)}=>/VP{/?[Value IN {'stimulated','induced'}](kw1)=>/PP{//?[Tag='DRUG'](kw2)}}} ::: distinct sent.cid, kw2.value, kw1.value, kw0.value, sent.value
//S{/?[Tag='DRUG'](kw2)=>/VP{/NP{//?[Tag='GENE'](kw0)=>//?[Value IN {'stimulated','induced'}](kw1)}}} ::: distinct sent.cid, kw2.value, kw1.value, kw0.value, sent.value
//S{/NP{/PP{//?[Tag='DRUG'](kw2)}}=>/VP{/?[Value IN {'stimulated','induced'}](kw1)=>/NP{//?[Tag='GENE'](kw0)}}} ::: distinct sent.cid, kw2.value, kw1.value, kw0.value, sent.value
//S{/NP{/?[Tag='GENE'](kw0)}=>/VP{/VP{/?[Value IN {'stimulated','induced'}](kw1)=>//?[Tag='DRUG'](kw2)}}} ::: distinct sent.cid, kw2.value, kw1.value, kw0.value, sent.value
//S{/NP{/?[Tag='GENE'](kw0)}=>/VP{/VP{//?[Value='induced'](kw1)=>//?[Tag='DRUG'](kw2)}}} ::: distinct sent.cid, kw2.value, kw1.value, kw0.value, sent.value
//S{/?[Tag='DRUG'](kw2)=>/VP{/?[Value IN {'stimulated','induced'}](kw1)=>/NP{/?[Tag='GENE'](kw0)}}} ::: distinct sent.cid, kw2.value, kw1.value, kw0.value, sent.value
//S{/?[Tag='DRUG'](kw2)=>/VP{/VP{//?[Value IN {'stimulate','induce'}](kw1)=>//?[Tag='GENE'](kw0)}}} ::: distinct sent.cid, kw2.value, kw1.value, kw0.value, sent.value
//NP{/NP{/?[Value IN {'induction','stimulation'}](kw1)}=>/PP{/NP{//?[Tag='GENE'](kw0)=>//?[Tag='DRUG'](kw2)}}} ::: distinct sent.cid, kw2.value, kw1.value, kw0.value, sent.value
//S{/NP{/?[Value IN {'stimulation','induction'}](kw1)=>/PP{//?[Tag='GENE'](kw0)}}=>/VP{/VP{//?[Tag='DRUG'](kw2)}}} ::: distinct sent.cid, kw2.value, kw1.value, kw0.value, sent.value
//S{/NP{/?[Tag='DRUG'](kw2)}=>/VP{/NP{//?[Value IN {'stimulation','induction'}](kw1)=>//?[Tag='GENE'](kw0)}}} ::: distinct sent.cid, kw2.value, kw1.value, kw0.value, sent.value
//NP{/NP{/NP{/?[Value='induction'](kw1)}=>/PP{//?[Tag='GENE'](kw0)}}=>/PP{/NP{/?[Tag='DRUG'](kw2)}}} ::: distinct sent.cid, kw2.value, kw1.value, kw0.value, sent.value
//NP{/NP{/?[Tag='GENE'](kw0)=>/?[Value IN {'stimulation','induction'}](kw1)}=>/PP{/NP{/?[Tag='DRUG'](kw2)}}} ::: distinct sent.cid, kw2.value, kw1.value, kw0.value, sent.value
//NP{/?[Value IN {'stimulation','induction'}](kw1)=>/PP{/NP{//?[Tag='GENE'](kw0)=>//?[Tag='DRUG'](kw2)}}} ::: distinct sent.cid, kw2.value, kw1.value, kw0.value, sent.value
//S{/NP{/?[Tag='DRUG'](kw2)}=>/VP{/?[Value IN {'stimulates','induces'}](kw1)=>/NP{//?[Tag='GENE'](kw0)}}} ::: distinct sent.cid, kw2.value, kw1.value, kw0.value, sent.value
//S{/?[Tag='DRUG'](kw2)=>/VP{/?[Value IN {'stimulates','induces'}](kw1)=>/NP{//?[Tag='GENE'](kw0)}}} ::: distinct sent.cid, kw2.value, kw1.value, kw0.value, sent.value
//S{/?[Tag='DRUG'](kw2)=>/VP{/?[Value IN {'stimulated','induced'}](kw1)=>/NP{//?[Tag='GENE'](kw0)}}} ::: distinct sent.cid, kw0.value, kw1.value, kw2.value, sent.value
//VP{/?[Value IN {'stimulated','induced'}](kw1)=>/PP{/NP{//?[Tag='DRUG'](kw2)=>//?[Tag='GENE'](kw0)}}} ::: distinct sent.cid, kw0.value, kw1.value, kw2.value, sent.value
//S{/NP{/?[Tag='DRUG'](kw2)}=>/VP{/?[Value IN {'stimulated','induced'}](kw1)=>/NP{//?[Tag='GENE'](kw0)}}} ::: distinct sent.cid, kw0.value, kw1.value, kw2.value, sent.value
//S{/NP{/?[Tag='GENE'](kw0)}=>/VP{/?[Value IN {'stimulated','induced'}](kw1)=>/PP{//?[Tag='DRUG'](kw2)}}} ::: distinct sent.cid, kw0.value, kw1.value, kw2.value, sent.value
//S{/?[Tag='DRUG'](kw2)=>/VP{/NP{//?[Tag='GENE'](kw0)=>//?[Value IN {'stimulated','induced'}](kw1)}}} ::: distinct sent.cid, kw0.value, kw1.value, kw2.value, sent.value
//S{/NP{/PP{//?[Tag='DRUG'](kw2)}}=>/VP{/?[Value IN {'stimulated','induced'}](kw1)=>/NP{//?[Tag='GENE'](kw0)}}} ::: distinct sent.cid, kw0.value, kw1.value, kw2.value, sent.value
//S{/NP{/?[Tag='GENE'](kw0)}=>/VP{/VP{/?[Value IN {'stimulated','induced'}](kw1)=>//?[Tag='DRUG'](kw2)}}} ::: distinct sent.cid, kw0.value, kw1.value, kw2.value, sent.value
//S{/NP{/?[Tag='GENE'](kw0)}=>/VP{/VP{//?[Value='induced'](kw1)=>//?[Tag='DRUG'](kw2)}}} ::: distinct sent.cid, kw0.value, kw1.value, kw2.value, sent.value
//S{/?[Tag='DRUG'](kw2)=>/VP{/?[Value IN {'stimulated','induced'}](kw1)=>/NP{/?[Tag='GENE'](kw0)}}} ::: distinct sent.cid, kw0.value, kw1.value, kw2.value, sent.value
//S{/?[Tag='DRUG'](kw2)=>/VP{/VP{//?[Value IN {'stimulate','induce'}](kw1)=>//?[Tag='GENE'](kw0)}}} ::: distinct sent.cid, kw0.value, kw1.value, kw2.value, sent.value
//NP{/NP{/?[Value IN {'induction','stimulation'}](kw1)}=>/PP{/NP{//?[Tag='GENE'](kw0)=>//?[Tag='DRUG'](kw2)}}} ::: distinct sent.cid, kw0.value, kw1.value, kw2.value, sent.value
//S{/NP{/?[Value IN {'stimulation','induction'}](kw1)=>/PP{//?[Tag='GENE'](kw0)}}=>/VP{/VP{//?[Tag='DRUG'](kw2)}}} ::: distinct sent.cid, kw0.value, kw1.value, kw2.value, sent.value
//S{/NP{/?[Tag='DRUG'](kw2)}=>/VP{/NP{//?[Value IN {'stimulation','induction'}](kw1)=>//?[Tag='GENE'](kw0)}}} ::: distinct sent.cid, kw0.value, kw1.value, kw2.value, sent.value
//NP{/NP{/NP{/?[Value='induction'](kw1)}=>/PP{//?[Tag='GENE'](kw0)}}=>/PP{/NP{/?[Tag='DRUG'](kw2)}}} ::: distinct sent.cid, kw0.value, kw1.value, kw2.value, sent.value
//NP{/NP{/?[Tag='GENE'](kw0)=>/?[Value IN {'stimulation','induction'}](kw1)}=>/PP{/NP{/?[Tag='DRUG'](kw2)}}} ::: distinct sent.cid, kw0.value, kw1.value, kw2.value, sent.value
//NP{/?[Value IN {'stimulation','induction'}](kw1)=>/PP{/NP{//?[Tag='GENE'](kw0)=>//?[Tag='DRUG'](kw2)}}} ::: distinct sent.cid, kw0.value, kw1.value, kw2.value, sent.value
//S{/NP{/?[Tag='DRUG'](kw2)}=>/VP{/?[Value IN {'stimulates','induces'}](kw1)=>/NP{//?[Tag='GENE'](kw0)}}} ::: distinct sent.cid, kw0.value, kw1.value, kw2.value, sent.value
//S{/?[Tag='DRUG'](kw2)=>/VP{/?[Value IN {'stimulates','induces'}](kw1)=>/NP{//?[Tag='GENE'](kw0)}}} ::: distinct sent.cid, kw0.value, kw1.value, kw2.value, sent.value
\end{lstlisting}

\section{Drug Inhibits Gene}
\begin{lstlisting}[breaklines,numbers=left,basicstyle=\footnotesize]
//NP{/NP{/NP{/?[Value='inhibitory'](kw1)}=>/PP{//?[Tag='DRUG'](kw2)}}=>/PP{/NP{//?[Tag='GENE'](kw0)}}} ::: distinct sent.cid, kw0.value, kw1.value, kw2.value, sent.value
//S{/?[Tag='DRUG'](kw2)=>/VP{/NP{/?[Tag='GENE'](kw0)=>/?[Value='inhibitor'](kw1)}}} ::: distinct sent.cid, kw0.value, kw1.value, kw2.value, sent.value
//NP{/?[Tag='DRUG'](kw2)=>/NP{/?[Tag='GENE'](kw0)=>/?[Value='inhibitor'](kw1)}} ::: distinct sent.cid, kw0.value, kw1.value, kw2.value, sent.value
//NP{/NP{/?[Tag='GENE'](kw0)=>/?[Value='inhibitor'](kw1)}=>/?[Tag='DRUG'](kw2)} ::: distinct sent.cid, kw0.value, kw1.value, kw2.value, sent.value
//NP{/NP{/NP{/?[Tag='GENE'](kw0)=>/?[Tag='DRUG'](kw2)}}=>/?[Value='inhibitor'](kw1)} ::: distinct sent.cid, kw0.value, kw1.value, kw2.value, sent.value
//NP{/?[Tag='GENE'](kw0)=>/?[Value='inhibitor'](kw1)=>/?[Tag='DRUG'](kw2)} ::: distinct sent.cid, kw0.value, kw1.value, kw2.value, sent.value
//NP{/NP{/?[Value='inhibitor'](kw1)}=>/PP{/NP{//?[Tag='GENE'](kw0)=>//?[Tag='DRUG'](kw2)}}} ::: distinct sent.cid, kw0.value, kw1.value, kw2.value, sent.value
//S{/?[Tag='DRUG'](kw2)=>/VP{/VP{//?[Tag='GENE'](kw0)=>//?[Value='inhibitor'](kw1)}}} ::: distinct sent.cid, kw0.value, kw1.value, kw2.value, sent.value
//S{/?[Tag='DRUG'](kw2)=>/VP{/NP{//?[Value='inhibitor'](kw1)=>//?[Tag='GENE'](kw0)}}} ::: distinct sent.cid, kw0.value, kw1.value, kw2.value, sent.value
//NP{/?[Tag='DRUG'](kw2)=>/NP{/NP{/?[Tag='GENE'](kw0)}}=>/?[Value='inhibitor'](kw1)} ::: distinct sent.cid, kw0.value, kw1.value, kw2.value, sent.value
//S{/?[Tag='DRUG'](kw2)=>/?[Value='inhibitor'](kw1)=>/?[Tag='GENE'](kw0)} ::: distinct sent.cid, kw0.value, kw1.value, kw2.value, sent.value
//NP{/NP{/?[Tag='DRUG'](kw2)=>/NP{/?[Value='inhibitor'](kw1)}=>/PP{//?[Tag='GENE'](kw0)}}} ::: distinct sent.cid, kw0.value, kw1.value, kw2.value, sent.value
//S{/?[Tag='GENE'](kw0)=>/?[Value='inhibitor'](kw1)=>/?[Tag='DRUG'](kw2)} ::: distinct sent.cid, kw0.value, kw1.value, kw2.value, sent.value
//NP{/?[Tag='DRUG'](kw2)=>/NP{/NP{/?[Value='inhibitor'](kw1)}=>/PP{//?[Tag='GENE'](kw0)}}} ::: distinct sent.cid, kw0.value, kw1.value, kw2.value, sent.value
//S{/?[Tag='DRUG'](kw2)=>/VP{/VP{//?[Value='inhibitor'](kw1)=>//?[Tag='GENE'](kw0)}}} ::: distinct sent.cid, kw0.value, kw1.value, kw2.value, sent.value
//S{/NP{/?[Tag='DRUG'](kw2)}=>/VP{/NP{//?[Value='inhibitor'](kw1)=>//?[Tag='GENE'](kw0)}}} ::: distinct sent.cid, kw0.value, kw1.value, kw2.value, sent.value
//NP{/?[Tag='DRUG'](kw2)=>/NP{/NP{/?[Tag='GENE'](kw0)=>/?[Value='inhibitor'](kw1)}}} ::: distinct sent.cid, kw0.value, kw1.value, kw2.value, sent.value
//NP{/NP{/PP{//?[Tag='DRUG'](kw2)=>//?[Tag='GENE'](kw0)}=>/NP{/?[Value='inhibitor'](kw1)}}} ::: distinct sent.cid, kw0.value, kw1.value, kw2.value, sent.value
//NP{/NP{/?[Tag='GENE'](kw0)=>/?[Value='inhibitor'](kw1)=>/?[Tag='DRUG'](kw2)}} ::: distinct sent.cid, kw0.value, kw1.value, kw2.value, sent.value
//S{/?[Tag='DRUG'](kw2)=>/?[Tag='GENE'](kw0)=>/?[Value='inhibitor'](kw1)} ::: distinct sent.cid, kw0.value, kw1.value, kw2.value, sent.value
//NP{/NP{/?[Tag='DRUG'](kw2)=>/?[Tag='GENE'](kw0)=>/?[Value='inhibitor'](kw1)}} ::: distinct sent.cid, kw0.value, kw1.value, kw2.value, sent.value
//NP{/NP{/?[Tag='DRUG'](kw2)=>/NP{/?[Tag='GENE'](kw0)=>/?[Value='inhibitor'](kw1)}}} ::: distinct sent.cid, kw0.value, kw1.value, kw2.value, sent.value
//NP{/NP{/PP{//?[Tag='DRUG'](kw2)}=>/?[Tag='GENE'](kw0)=>/?[Value='inhibitor'](kw1)}} ::: distinct sent.cid, kw0.value, kw1.value, kw2.value, sent.value
//S{/?[Tag='DRUG'](kw2)=>/VP{/?[Value='inhibited'](kw1)=>/NP{//?[Tag='GENE'](kw0)}}} ::: distinct sent.cid, kw2.value, kw1.value, kw0.value, sent.value
//S{/NP{/NP{/?[Tag='DRUG'](kw2)}}=>/VP{/?[Value='inhibited'](kw1)=>/NP{//?[Tag='GENE'](kw0)}}} ::: distinct sent.cid, kw2.value, kw1.value, kw0.value, sent.value
//S{/?[Tag='DRUG'](kw2)=>/?[Value='inhibited'](kw1)=>/?[Tag='GENE'](kw0)} ::: distinct sent.cid, kw2.value, kw1.value, kw0.value, sent.value
//S{/NP{/?[Tag='DRUG'](kw2)}=>/VP{/?[Value='inhibited'](kw1)=>/NP{//?[Tag='GENE'](kw0)}}} ::: distinct sent.cid, kw2.value, kw1.value, kw0.value, sent.value
//S{/NP{/NP{//?[Tag='DRUG'](kw2)}}=>/VP{/?[Value='inhibited'](kw1)=>/NP{/?[Tag='GENE'](kw0)}}} ::: distinct sent.cid, kw2.value, kw1.value, kw0.value, sent.value
//S{/NP{/PP{//?[Tag='GENE'](kw0)}}=>/VP{/VP{/?[Value='inhibited'](kw1)=>//?[Tag='DRUG'](kw2)}}} ::: distinct sent.cid, kw2.value, kw1.value, kw0.value, sent.value
//S{/NP{/NP{//?[Tag='DRUG'](kw2)}}=>/VP{/?[Value='inhibited'](kw1)=>/NP{//?[Tag='GENE'](kw0)}}} ::: distinct sent.cid, kw2.value, kw1.value, kw0.value, sent.value
//S{/?[Tag='DRUG'](kw2)=>/VP{/VP{/?[Value='inhibited'](kw1)=>//?[Tag='GENE'](kw0)}}} ::: distinct sent.cid, kw2.value, kw1.value, kw0.value, sent.value
//S{/NP{/PP{//?[Tag='DRUG'](kw2)}}=>/VP{/?[Value='inhibited'](kw1)=>/NP{/?[Tag='GENE'](kw0)}}} ::: distinct sent.cid, kw2.value, kw1.value, kw0.value, sent.value
//S{/NP{/NP{/?[Tag='GENE'](kw0)}}=>/VP{/VP{/?[Value='inhibited'](kw1)=>//?[Tag='DRUG'](kw2)}}} ::: distinct sent.cid, kw2.value, kw1.value, kw0.value, sent.value
//S{/NP{/NP{/?[Tag='GENE'](kw0)}}=>/VP{/?[Value='inhibited'](kw1)=>/PP{//?[Tag='DRUG'](kw2)}}} ::: distinct sent.cid, kw2.value, kw1.value, kw0.value, sent.value
//S{/?[Tag='DRUG'](kw2)=>/VP{/?[Value='inhibited'](kw1)=>/PP{//?[Tag='GENE'](kw0)}}} ::: distinct sent.cid, kw2.value, kw1.value, kw0.value, sent.value
//S{/NP{/?[Tag='DRUG'](kw2)}=>/VP{/VP{/?[Value='inhibited'](kw1)=>//?[Tag='GENE'](kw0)}}} ::: distinct sent.cid, kw2.value, kw1.value, kw0.value, sent.value
//S{/NP{/NP{/?[Tag='DRUG'](kw2)}}=>/VP{/?[Value='inhibited'](kw1)=>/NP{/?[Tag='GENE'](kw0)}}} ::: distinct sent.cid, kw2.value, kw1.value, kw0.value, sent.value
//S{/NP{/?[Tag='GENE'](kw0)}=>/VP{/VP{/?[Value='inhibited'](kw1)=>//?[Tag='DRUG'](kw2)}}} ::: distinct sent.cid, kw2.value, kw1.value, kw0.value, sent.value
//S{/NP{/?[Tag='GENE'](kw0)}=>/VP{/?[Value='inhibited'](kw1)=>/NP{/?[Tag='DRUG'](kw2)}}} ::: distinct sent.cid, kw2.value, kw1.value, kw0.value, sent.value
//S{/NP{/?[Tag='GENE'](kw0)}=>/VP{/VP{//?[Value='inhibited'](kw1)=>//?[Tag='DRUG'](kw2)}}} ::: distinct sent.cid, kw2.value, kw1.value, kw0.value, sent.value
//S{/?[Tag='DRUG'](kw2)=>/VP{/?[Value='inhibited'](kw1)=>/NP{/?[Tag='GENE'](kw0)}}} ::: distinct sent.cid, kw2.value, kw1.value, kw0.value, sent.value
//S{/NP{/NP{/?[Tag='GENE'](kw0)=>/?[Tag='DRUG'](kw2)}}=>/VP{/?[Value='inhibit'](kw1)}} ::: distinct sent.cid, kw2.value, kw1.value, kw0.value, sent.value
//S{/NP{/?[Tag='DRUG'](kw2)}=>/VP{/VP{//?[Value='inhibit'](kw1)=>//?[Tag='GENE'](kw0)}}} ::: distinct sent.cid, kw2.value, kw1.value, kw0.value, sent.value
//S{/?[Tag='DRUG'](kw2)=>/VP{/VP{//?[Value='inhibit'](kw1)=>//?[Tag='GENE'](kw0)}}} ::: distinct sent.cid, kw2.value, kw1.value, kw0.value, sent.value
//NP{/NP{/?[Tag='GENE'](kw0)=>/?[Value='inhibition'](kw1)}=>/PP{/NP{/?[Tag='DRUG'](kw2)}}} ::: distinct sent.cid, kw2.value, kw1.value, kw0.value, sent.value
//S{/?[Tag='DRUG'](kw2)=>/VP{/PP{//?[Value='inhibition'](kw1)=>//?[Tag='GENE'](kw0)}}} ::: distinct sent.cid, kw2.value, kw1.value, kw0.value, sent.value
//S{/NP{/PP{//?[Tag='DRUG'](kw2)}}=>/VP{/NP{//?[Value='inhibition'](kw1)=>//?[Tag='GENE'](kw0)}}} ::: distinct sent.cid, kw2.value, kw1.value, kw0.value, sent.value
//NP{/?[Value='inhibition'](kw1)=>/PP{/NP{//?[Tag='GENE'](kw0)=>//?[Tag='DRUG'](kw2)}}} ::: distinct sent.cid, kw2.value, kw1.value, kw0.value, sent.value
//S{/?[Value='inhibition'](kw1)=>/PP{/NP{//?[Tag='GENE'](kw0)=>//?[Tag='DRUG'](kw2)}}} ::: distinct sent.cid, kw2.value, kw1.value, kw0.value, sent.value
//NP{/NP{/?[Tag='DRUG'](kw2)=>/?[Value='inhibition'](kw1)}=>/PP{/NP{/?[Tag='GENE'](kw0)}}} ::: distinct sent.cid, kw2.value, kw1.value, kw0.value, sent.value
//NP{/NP{/?[Value='inhibition'](kw1)}=>/PP{/NP{/?[Tag='GENE'](kw0)=>/?[Tag='DRUG'](kw2)}}} ::: distinct sent.cid, kw2.value, kw1.value, kw0.value, sent.value
//S{/NP{/?[Value='inhibition'](kw1)=>/PP{//?[Tag='GENE'](kw0)}}=>/VP{/PP{//?[Tag='DRUG'](kw2)}}} ::: distinct sent.cid, kw2.value, kw1.value, kw0.value, sent.value
//S{/NP{/NP{/?[Value='inhibition'](kw1)}=>/PP{//?[Tag='GENE'](kw0)}}=>/VP{/NP{//?[Tag='DRUG'](kw2)}}} ::: distinct sent.cid, kw2.value, kw1.value, kw0.value, sent.value
//S{/?[Value IN {'inhibition','Inhibition'}](kw1)=>/?[Tag='GENE'](kw0)=>/?[Tag='DRUG'](kw2)} ::: distinct sent.cid, kw2.value, kw1.value, kw0.value, sent.value
//NP{/NP{/NP{/?[Value='inhibition'](kw1)}=>/PP{//?[Tag='GENE'](kw0)}}=>/PP{/NP{//?[Tag='DRUG'](kw2)}}} ::: distinct sent.cid, kw2.value, kw1.value, kw0.value, sent.value
//NP{/?[Tag='DRUG'](kw2)=>/NP{/?[Value='inhibition'](kw1)}=>/PP{/NP{/?[Tag='GENE'](kw0)}}} ::: distinct sent.cid, kw2.value, kw1.value, kw0.value, sent.value
//NP{/NP{/?[Value='inhibition'](kw1)}=>/PP{/NP{//?[Tag='GENE'](kw0)=>//?[Tag='DRUG'](kw2)}}} ::: distinct sent.cid, kw2.value, kw1.value, kw0.value, sent.value
//S{/?[Tag='DRUG'](kw2)=>/VP{/NP{//?[Value='inhibition'](kw1)=>//?[Tag='GENE'](kw0)}}} ::: distinct sent.cid, kw2.value, kw1.value, kw0.value, sent.value
//S{/NP{/NP{/?[Value='inhibition'](kw1)}=>/PP{//?[Tag='DRUG'](kw2)}}=>/VP{/NP{//?[Tag='GENE'](kw0)}}} ::: distinct sent.cid, kw2.value, kw1.value, kw0.value, sent.value
//NP{/NP{/?[Value='inhibition'](kw1)=>/PP{//?[Tag='GENE'](kw0)}}=>/PP{/NP{//?[Tag='DRUG'](kw2)}}} ::: distinct sent.cid, kw2.value, kw1.value, kw0.value, sent.value
//NP{/NP{/?[Tag='DRUG'](kw2)=>/?[Value='inhibition'](kw1)}=>/PP{/NP{//?[Tag='GENE'](kw0)}}} ::: distinct sent.cid, kw2.value, kw1.value, kw0.value, sent.value
//NP{/NP{/?[Value='inhibition'](kw1)=>/PP{//?[Tag='GENE'](kw0)}}=>/PP{/NP{/?[Tag='DRUG'](kw2)}}} ::: distinct sent.cid, kw2.value, kw1.value, kw0.value, sent.value
//NP{/NP{/NP{/?[Value='inhibition'](kw1)}=>/PP{//?[Tag='GENE'](kw0)}}=>/PP{/NP{/?[Tag='DRUG'](kw2)}}} ::: distinct sent.cid, kw2.value, kw1.value, kw0.value, sent.value
//S{/NP{/?[Tag='DRUG'](kw2)}=>/VP{/?[Value='inhibits'](kw1)=>/NP{//?[Tag='GENE'](kw0)}}} ::: distinct sent.cid, kw2.value, kw1.value, kw0.value, sent.value
/S{/?[Tag='DRUG'](kw2)=>/VP{/?[Value='inhibits'](kw1)=>/NP{/?[Tag='GENE'](kw0)}}} ::: distinct sent.cid, kw2.value, kw1.value, kw0.value, sent.value
//S{/?[Tag='DRUG'](kw2)=>/VP{/VP{/?[Value='inhibits'](kw1)=>//?[Tag='GENE'](kw0)}}} ::: distinct sent.cid, kw2.value, kw1.value, kw0.value, sent.value
//S{/?[Tag='DRUG'](kw2)=>/VP{/?[Value='inhibits'](kw1)=>/NP{//?[Tag='GENE'](kw0)}}} ::: distinct sent.cid, kw2.value, kw1.value, kw0.value, sent.value
//NP{/NP{/NP{/?[Value='inhibitory'](kw1)}=>/PP{//?[Tag='DRUG'](kw2)}}=>/PP{/NP{//?[Tag='GENE'](kw0)}}} ::: distinct sent.cid, kw0.value, kw1.value, kw2.value, sent.value
//S{/?[Tag='DRUG'](kw2)=>/VP{/NP{/?[Tag='GENE'](kw0)=>/?[Value='inhibitor'](kw1)}}} ::: distinct sent.cid, kw0.value, kw1.value, kw2.value, sent.value
//NP{/?[Tag='DRUG'](kw2)=>/NP{/?[Tag='GENE'](kw0)=>/?[Value='inhibitor'](kw1)}} ::: distinct sent.cid, kw0.value, kw1.value, kw2.value, sent.value
//NP{/NP{/?[Tag='GENE'](kw0)=>/?[Value='inhibitor'](kw1)}=>/?[Tag='DRUG'](kw2)} ::: distinct sent.cid, kw0.value, kw1.value, kw2.value, sent.value
//NP{/NP{/NP{/?[Tag='GENE'](kw0)=>/?[Tag='DRUG'](kw2)}}=>/?[Value='inhibitor'](kw1)} ::: distinct sent.cid, kw0.value, kw1.value, kw2.value, sent.value
//NP{/?[Tag='GENE'](kw0)=>/?[Value='inhibitor'](kw1)=>/?[Tag='DRUG'](kw2)} ::: distinct sent.cid, kw0.value, kw1.value, kw2.value, sent.value
//NP{/NP{/?[Value='inhibitor'](kw1)}=>/PP{/NP{//?[Tag='GENE'](kw0)=>//?[Tag='DRUG'](kw2)}}} ::: distinct sent.cid, kw0.value, kw1.value, kw2.value, sent.value
//S{/?[Tag='DRUG'](kw2)=>/VP{/VP{//?[Tag='GENE'](kw0)=>//?[Value='inhibitor'](kw1)}}} ::: distinct sent.cid, kw0.value, kw1.value, kw2.value, sent.value
//S{/?[Tag='DRUG'](kw2)=>/VP{/NP{//?[Value='inhibitor'](kw1)=>//?[Tag='GENE'](kw0)}}} ::: distinct sent.cid, kw0.value, kw1.value, kw2.value, sent.value
//NP{/?[Tag='DRUG'](kw2)=>/NP{/NP{/?[Tag='GENE'](kw0)}}=>/?[Value='inhibitor'](kw1)} ::: distinct sent.cid, kw0.value, kw1.value, kw2.value, sent.value
//S{/?[Tag='DRUG'](kw2)=>/?[Value='inhibitor'](kw1)=>/?[Tag='GENE'](kw0)} ::: distinct sent.cid, kw0.value, kw1.value, kw2.value, sent.value
//NP{/NP{/?[Tag='DRUG'](kw2)=>/NP{/?[Value='inhibitor'](kw1)}=>/PP{//?[Tag='GENE'](kw0)}}} ::: distinct sent.cid, kw0.value, kw1.value, kw2.value, sent.value
//S{/?[Tag='GENE'](kw0)=>/?[Value='inhibitor'](kw1)=>/?[Tag='DRUG'](kw2)} ::: distinct sent.cid, kw0.value, kw1.value, kw2.value, sent.value
//NP{/?[Tag='DRUG'](kw2)=>/NP{/NP{/?[Value='inhibitor'](kw1)}=>/PP{//?[Tag='GENE'](kw0)}}} ::: distinct sent.cid, kw0.value, kw1.value, kw2.value, sent.value
//S{/?[Tag='DRUG'](kw2)=>/VP{/VP{//?[Value='inhibitor'](kw1)=>//?[Tag='GENE'](kw0)}}} ::: distinct sent.cid, kw0.value, kw1.value, kw2.value, sent.value
//S{/NP{/?[Tag='DRUG'](kw2)}=>/VP{/NP{//?[Value='inhibitor'](kw1)=>//?[Tag='GENE'](kw0)}}} ::: distinct sent.cid, kw0.value, kw1.value, kw2.value, sent.value
//NP{/?[Tag='DRUG'](kw2)=>/NP{/NP{/?[Tag='GENE'](kw0)=>/?[Value='inhibitor'](kw1)}}} ::: distinct sent.cid, kw0.value, kw1.value, kw2.value, sent.value
//NP{/NP{/PP{//?[Tag='DRUG'](kw2)=>//?[Tag='GENE'](kw0)}=>/NP{/?[Value='inhibitor'](kw1)}}} ::: distinct sent.cid, kw0.value, kw1.value, kw2.value, sent.value
//NP{/NP{/?[Tag='GENE'](kw0)=>/?[Value='inhibitor'](kw1)=>/?[Tag='DRUG'](kw2)}} ::: distinct sent.cid, kw0.value, kw1.value, kw2.value, sent.value
//S{/?[Tag='DRUG'](kw2)=>/?[Tag='GENE'](kw0)=>/?[Value='inhibitor'](kw1)} ::: distinct sent.cid, kw0.value, kw1.value, kw2.value, sent.value
//NP{/NP{/?[Tag='DRUG'](kw2)=>/?[Tag='GENE'](kw0)=>/?[Value='inhibitor'](kw1)}} ::: distinct sent.cid, kw0.value, kw1.value, kw2.value, sent.value
//NP{/NP{/?[Tag='DRUG'](kw2)=>/NP{/?[Tag='GENE'](kw0)=>/?[Value='inhibitor'](kw1)}}} ::: distinct sent.cid, kw0.value, kw1.value, kw2.value, sent.value
//NP{/NP{/PP{//?[Tag='DRUG'](kw2)}=>/?[Tag='GENE'](kw0)=>/?[Value='inhibitor'](kw1)}} ::: distinct sent.cid, kw0.value, kw1.value, kw2.value, sent.value
//S{/?[Tag='DRUG'](kw2)=>/VP{/?[Value='inhibited'](kw1)=>/NP{//?[Tag='GENE'](kw0)}}} ::: distinct sent.cid, kw0.value, kw1.value, kw2.value, sent.value
//S{/NP{/NP{/?[Tag='DRUG'](kw2)}}=>/VP{/?[Value='inhibited'](kw1)=>/NP{//?[Tag='GENE'](kw0)}}} ::: distinct sent.cid, kw0.value, kw1.value, kw2.value, sent.value
//S{/?[Tag='DRUG'](kw2)=>/?[Value='inhibited'](kw1)=>/?[Tag='GENE'](kw0)} ::: distinct sent.cid, kw0.value, kw1.value, kw2.value, sent.value
//S{/NP{/?[Tag='DRUG'](kw2)}=>/VP{/?[Value='inhibited'](kw1)=>/NP{//?[Tag='GENE'](kw0)}}} ::: distinct sent.cid, kw0.value, kw1.value, kw2.value, sent.value
//S{/NP{/NP{//?[Tag='DRUG'](kw2)}}=>/VP{/?[Value='inhibited'](kw1)=>/NP{/?[Tag='GENE'](kw0)}}} ::: distinct sent.cid, kw0.value, kw1.value, kw2.value, sent.value
//S{/NP{/PP{//?[Tag='GENE'](kw0)}}=>/VP{/VP{/?[Value='inhibited'](kw1)=>//?[Tag='DRUG'](kw2)}}} ::: distinct sent.cid, kw0.value, kw1.value, kw2.value, sent.value
//S{/NP{/NP{//?[Tag='DRUG'](kw2)}}=>/VP{/?[Value='inhibited'](kw1)=>/NP{//?[Tag='GENE'](kw0)}}} ::: distinct sent.cid, kw0.value, kw1.value, kw2.value, sent.value
//S{/?[Tag='DRUG'](kw2)=>/VP{/VP{/?[Value='inhibited'](kw1)=>//?[Tag='GENE'](kw0)}}} ::: distinct sent.cid, kw0.value, kw1.value, kw2.value, sent.value
//S{/NP{/PP{//?[Tag='DRUG'](kw2)}}=>/VP{/?[Value='inhibited'](kw1)=>/NP{/?[Tag='GENE'](kw0)}}} ::: distinct sent.cid, kw0.value, kw1.value, kw2.value, sent.value
//S{/NP{/NP{/?[Tag='GENE'](kw0)}}=>/VP{/VP{/?[Value='inhibited'](kw1)=>//?[Tag='DRUG'](kw2)}}} ::: distinct sent.cid, kw0.value, kw1.value, kw2.value, sent.value
//S{/NP{/NP{/?[Tag='GENE'](kw0)}}=>/VP{/?[Value='inhibited'](kw1)=>/PP{//?[Tag='DRUG'](kw2)}}} ::: distinct sent.cid, kw0.value, kw1.value, kw2.value, sent.value
//S{/?[Tag='DRUG'](kw2)=>/VP{/?[Value='inhibited'](kw1)=>/PP{//?[Tag='GENE'](kw0)}}} ::: distinct sent.cid, kw0.value, kw1.value, kw2.value, sent.value
//S{/NP{/?[Tag='DRUG'](kw2)}=>/VP{/VP{/?[Value='inhibited'](kw1)=>//?[Tag='GENE'](kw0)}}} ::: distinct sent.cid, kw0.value, kw1.value, kw2.value, sent.value
//S{/NP{/NP{/?[Tag='DRUG'](kw2)}}=>/VP{/?[Value='inhibited'](kw1)=>/NP{/?[Tag='GENE'](kw0)}}} ::: distinct sent.cid, kw0.value, kw1.value, kw2.value, sent.value
//S{/NP{/?[Tag='GENE'](kw0)}=>/VP{/VP{/?[Value='inhibited'](kw1)=>//?[Tag='DRUG'](kw2)}}} ::: distinct sent.cid, kw0.value, kw1.value, kw2.value, sent.value
//S{/NP{/?[Tag='GENE'](kw0)}=>/VP{/?[Value='inhibited'](kw1)=>/NP{/?[Tag='DRUG'](kw2)}}} ::: distinct sent.cid, kw0.value, kw1.value, kw2.value, sent.value
//S{/NP{/?[Tag='GENE'](kw0)}=>/VP{/VP{//?[Value='inhibited'](kw1)=>//?[Tag='DRUG'](kw2)}}} ::: distinct sent.cid, kw0.value, kw1.value, kw2.value, sent.value
//S{/?[Tag='DRUG'](kw2)=>/VP{/?[Value='inhibited'](kw1)=>/NP{/?[Tag='GENE'](kw0)}}} ::: distinct sent.cid, kw0.value, kw1.value, kw2.value, sent.value
//S{/NP{/NP{/?[Tag='GENE'](kw0)=>/?[Tag='DRUG'](kw2)}}=>/VP{/?[Value='inhibit'](kw1)}} ::: distinct sent.cid, kw0.value, kw1.value, kw2.value, sent.value
//S{/NP{/?[Tag='DRUG'](kw2)}=>/VP{/VP{//?[Value='inhibit'](kw1)=>//?[Tag='GENE'](kw0)}}} ::: distinct sent.cid, kw0.value, kw1.value, kw2.value, sent.value
//S{/?[Tag='DRUG'](kw2)=>/VP{/VP{//?[Value='inhibit'](kw1)=>//?[Tag='GENE'](kw0)}}} ::: distinct sent.cid, kw0.value, kw1.value, kw2.value, sent.value
//NP{/NP{/?[Tag='GENE'](kw0)=>/?[Value='inhibition'](kw1)}=>/PP{/NP{/?[Tag='DRUG'](kw2)}}} ::: distinct sent.cid, kw0.value, kw1.value, kw2.value, sent.value
//S{/?[Tag='DRUG'](kw2)=>/VP{/PP{//?[Value='inhibition'](kw1)=>//?[Tag='GENE'](kw0)}}} ::: distinct sent.cid, kw0.value, kw1.value, kw2.value, sent.value
//S{/NP{/PP{//?[Tag='DRUG'](kw2)}}=>/VP{/NP{//?[Value='inhibition'](kw1)=>//?[Tag='GENE'](kw0)}}} ::: distinct sent.cid, kw0.value, kw1.value, kw2.value, sent.value
//NP{/?[Value='inhibition'](kw1)=>/PP{/NP{//?[Tag='GENE'](kw0)=>//?[Tag='DRUG'](kw2)}}} ::: distinct sent.cid, kw0.value, kw1.value, kw2.value, sent.value
//S{/?[Value='inhibition'](kw1)=>/PP{/NP{//?[Tag='GENE'](kw0)=>//?[Tag='DRUG'](kw2)}}} ::: distinct sent.cid, kw0.value, kw1.value, kw2.value, sent.value
//NP{/NP{/?[Tag='DRUG'](kw2)=>/?[Value='inhibition'](kw1)}=>/PP{/NP{/?[Tag='GENE'](kw0)}}} ::: distinct sent.cid, kw0.value, kw1.value, kw2.value, sent.value
//NP{/NP{/?[Value='inhibition'](kw1)}=>/PP{/NP{/?[Tag='GENE'](kw0)=>/?[Tag='DRUG'](kw2)}}} ::: distinct sent.cid, kw0.value, kw1.value, kw2.value, sent.value
//S{/NP{/?[Value='inhibition'](kw1)=>/PP{//?[Tag='GENE'](kw0)}}=>/VP{/PP{//?[Tag='DRUG'](kw2)}}} ::: distinct sent.cid, kw0.value, kw1.value, kw2.value, sent.value
//S{/NP{/NP{/?[Value='inhibition'](kw1)}=>/PP{//?[Tag='GENE'](kw0)}}=>/VP{/NP{//?[Tag='DRUG'](kw2)}}} ::: distinct sent.cid, kw0.value, kw1.value, kw2.value, sent.value
//S{/?[Value IN {'inhibition','Inhibition'}](kw1)=>/?[Tag='GENE'](kw0)=>/?[Tag='DRUG'](kw2)} ::: distinct sent.cid, kw0.value, kw1.value, kw2.value, sent.value
//NP{/NP{/NP{/?[Value='inhibition'](kw1)}=>/PP{//?[Tag='GENE'](kw0)}}=>/PP{/NP{//?[Tag='DRUG'](kw2)}}} ::: distinct sent.cid, kw0.value, kw1.value, kw2.value, sent.value
//NP{/?[Tag='DRUG'](kw2)=>/NP{/?[Value='inhibition'](kw1)}=>/PP{/NP{/?[Tag='GENE'](kw0)}}} ::: distinct sent.cid, kw0.value, kw1.value, kw2.value, sent.value
//NP{/NP{/?[Value='inhibition'](kw1)}=>/PP{/NP{//?[Tag='GENE'](kw0)=>//?[Tag='DRUG'](kw2)}}} ::: distinct sent.cid, kw0.value, kw1.value, kw2.value, sent.value
//S{/?[Tag='DRUG'](kw2)=>/VP{/NP{//?[Value='inhibition'](kw1)=>//?[Tag='GENE'](kw0)}}} ::: distinct sent.cid, kw0.value, kw1.value, kw2.value, sent.value
//S{/NP{/NP{/?[Value='inhibition'](kw1)}=>/PP{//?[Tag='DRUG'](kw2)}}=>/VP{/NP{//?[Tag='GENE'](kw0)}}} ::: distinct sent.cid, kw0.value, kw1.value, kw2.value, sent.value
//NP{/NP{/?[Value='inhibition'](kw1)=>/PP{//?[Tag='GENE'](kw0)}}=>/PP{/NP{//?[Tag='DRUG'](kw2)}}} ::: distinct sent.cid, kw0.value, kw1.value, kw2.value, sent.value
//NP{/NP{/?[Tag='DRUG'](kw2)=>/?[Value='inhibition'](kw1)}=>/PP{/NP{//?[Tag='GENE'](kw0)}}} ::: distinct sent.cid, kw0.value, kw1.value, kw2.value, sent.value
//NP{/NP{/?[Value='inhibition'](kw1)=>/PP{//?[Tag='GENE'](kw0)}}=>/PP{/NP{/?[Tag='DRUG'](kw2)}}} ::: distinct sent.cid, kw0.value, kw1.value, kw2.value, sent.value
//NP{/NP{/NP{/?[Value='inhibition'](kw1)}=>/PP{//?[Tag='GENE'](kw0)}}=>/PP{/NP{/?[Tag='DRUG'](kw2)}}} ::: distinct sent.cid, kw0.value, kw1.value, kw2.value, sent.value
//S{/NP{/?[Tag='DRUG'](kw2)}=>/VP{/?[Value='inhibits'](kw1)=>/NP{//?[Tag='GENE'](kw0)}}} ::: distinct sent.cid, kw0.value, kw1.value, kw2.value, sent.value
/S{/?[Tag='DRUG'](kw2)=>/VP{/?[Value='inhibits'](kw1)=>/NP{/?[Tag='GENE'](kw0)}}} ::: distinct sent.cid, kw0.value, kw1.value, kw2.value, sent.value
//S{/?[Tag='DRUG'](kw2)=>/VP{/VP{/?[Value='inhibits'](kw1)=>//?[Tag='GENE'](kw0)}}} ::: distinct sent.cid, kw0.value, kw1.value, kw2.value, sent.value
//S{/?[Tag='DRUG'](kw2)=>/VP{/?[Value='inhibits'](kw1)=>/NP{//?[Tag='GENE'](kw0)}}} ::: distinct sent.cid, kw0.value, kw1.value, kw2.value, sent.value
\end{lstlisting}

\section{Gene Metabolized Drug}
\begin{lstlisting}[breaklines,numbers=left,basicstyle=\footnotesize]
//S{/NP{/NP{/?[Tag='DRUG'](kw2)}}=>/VP{/VP{/?[Value IN {'metabolised','metabolized'}](kw1)=>//?[Tag='GENE'](kw0)}}} ::: distinct sent.cid, kw0.value, kw1.value, kw2.value, sent.value
//S{/?[Tag='DRUG'](kw2)=>/VP{/VP{//?[Value IN {'metabolised','metabolized'}](kw1)=>//?[Tag='GENE'](kw0)}}} ::: distinct sent.cid, kw0.value, kw1.value, kw2.value, sent.value
//S{/NP{/SBAR{//?[Tag='DRUG'](kw2)}}=>/VP{/?[Value IN {'metabolised','metabolized'}](kw1)=>/PP{//?[Tag='GENE'](kw0)}}} ::: distinct sent.cid, kw0.value, kw1.value, kw2.value, sent.value
//S{/NP{/?[Tag='DRUG'](kw2)}=>/VP{/VP{/?[Value IN {'metabolised','metabolized'}](kw1)=>//?[Tag='GENE'](kw0)}}} ::: distinct sent.cid, kw0.value, kw1.value, kw2.value, sent.value
//S{/NP{/?[Tag='DRUG'](kw2)}=>/VP{/VP{//?[Value IN {'metabolized','metabolised'}](kw1)=>//?[Tag='GENE'](kw0)}}} ::: distinct sent.cid, kw0.value, kw1.value, kw2.value, sent.value
//S{/?[Tag='DRUG'](kw2)=>/VP{/VP{/?[Value IN {'metabolised','metabolized'}](kw1)=>//?[Tag='GENE'](kw0)}}} ::: distinct sent.cid, kw0.value, kw1.value, kw2.value, sent.value
//S{/?[Tag='DRUG'](kw2)=>/VP{/?[Value='metabolized'](kw1)=>/PP{//?[Tag='GENE'](kw0)}}} ::: distinct sent.cid, kw0.value, kw1.value, kw2.value, sent.value
//S{/?[Tag='DRUG'](kw2)=>/NP{/?[Value='metabolism'](kw1)}=>/VP{/VP{//?[Tag='GENE'](kw0)}}} ::: distinct sent.cid, kw0.value, kw1.value, kw2.value, sent.value
//S{/NP{/?[Tag='GENE'](kw0)}=>/VP{/NP{//?[Tag='DRUG'](kw2)=>//?[Value='metabolism'](kw1)}}} ::: distinct sent.cid, kw0.value, kw1.value, kw2.value, sent.value
//NP{/NP{/?[Tag='GENE'](kw0)=>/?[Value='metabolism'](kw1)}=>/PP{/NP{/?[Tag='DRUG'](kw2)}}} ::: distinct sent.cid, kw0.value, kw1.value, kw2.value, sent.value
//NP{/?[Value='metabolism'](kw1)=>/PP{/NP{//?[Tag='DRUG'](kw2)=>//?[Tag='GENE'](kw0)}}} ::: distinct sent.cid, kw0.value, kw1.value, kw2.value, sent.value
//S{/NP{/?[Tag='GENE'](kw0)}=>/VP{/VP{//?[Value='metabolism'](kw1)=>//?[Tag='DRUG'](kw2)}}} ::: distinct sent.cid, kw0.value, kw1.value, kw2.value, sent.value
//S{/NP{/NP{/?[Tag='GENE'](kw0)}}=>/VP{/NP{//?[Value='metabolism'](kw1)=>//?[Tag='DRUG'](kw2)}}} ::: distinct sent.cid, kw0.value, kw1.value, kw2.value, sent.value
//NP{/NP{/PP{//?[Tag='GENE'](kw0)}}=>/PP{/NP{/?[Tag='DRUG'](kw2)=>/?[Value='metabolism'](kw1)}}} ::: distinct sent.cid, kw0.value, kw1.value, kw2.value, sent.value
//S{/NP{/?[Tag='GENE'](kw0)}=>/VP{/PP{//?[Value='metabolism'](kw1)=>//?[Tag='DRUG'](kw2)}}} ::: distinct sent.cid, kw0.value, kw1.value, kw2.value, sent.value
//NP{/NP{/?[Tag='DRUG'](kw2)=>/?[Value='metabolism'](kw1)}=>/PP{/NP{//?[Tag='GENE'](kw0)}}} ::: distinct sent.cid, kw0.value, kw1.value, kw2.value, sent.value
//NP{/NP{/?[Tag='GENE'](kw0)=>/?[Value='metabolism'](kw1)}=>/PP{/NP{//?[Tag='DRUG'](kw2)}}} ::: distinct sent.cid, kw0.value, kw1.value, kw2.value, sent.value
//S{/NP{/NP{/?[Value='metabolism'](kw1)}=>/PP{//?[Tag='DRUG'](kw2)}}=>/VP{/VP{//?[Tag='GENE'](kw0)}}} ::: distinct sent.cid, kw0.value, kw1.value, kw2.value, sent.value
//S{/NP{/?[Tag='GENE'](kw0)}=>/VP{/PP{//?[Tag='DRUG'](kw2)=>//?[Value='metabolism'](kw1)}}} ::: distinct sent.cid, kw0.value, kw1.value, kw2.value, sent.value
//NP{/NP{/PP{//?[Tag='GENE'](kw0)}}=>/PP{/NP{//?[Value='metabolism'](kw1)=>//?[Tag='DRUG'](kw2)}}} ::: distinct sent.cid, kw0.value, kw1.value, kw2.value, sent.value
//S{/NP{/?[Tag='DRUG'](kw2)}=>/VP{/NP{//?[Value='substrate'](kw1)=>//?[Tag='GENE'](kw0)}}} ::: distinct sent.cid, kw0.value, kw1.value, kw2.value, sent.value
//NP{/NP{/?[Tag='DRUG'](kw2)=>/NP{/?[Value='substrate'](kw1)}=>/PP{//?[Tag='GENE'](kw0)}}} ::: distinct sent.cid, kw0.value, kw1.value, kw2.value, sent.value
//S{/?[Tag='DRUG'](kw2)=>/VP{/VP{//?[Value='substrate'](kw1)=>//?[Tag='GENE'](kw0)}}} ::: distinct sent.cid, kw0.value, kw1.value, kw2.value, sent.value
//NP{/?[Tag='GENE'](kw0)=>/?[Value='substrate'](kw1)=>/?[Tag='DRUG'](kw2)} ::: distinct sent.cid, kw0.value, kw1.value, kw2.value, sent.value
//NP{/?[Tag='DRUG'](kw2)=>/NP{/NP{/?[Value='substrate'](kw1)}=>/PP{//?[Tag='GENE'](kw0)}}} ::: distinct sent.cid, kw0.value, kw1.value, kw2.value, sent.value
//S{/NP{/NP{/?[Tag='DRUG'](kw2)}}=>/VP{/VP{/?[Value IN {'metabolised','metabolized'}](kw1)=>//?[Tag='GENE'](kw0)}}} ::: distinct sent.cid, kw0.value, kw1.value, kw2.value, sent.value
//S{/?[Tag='DRUG'](kw2)=>/VP{/VP{//?[Value IN {'metabolised','metabolized'}](kw1)=>//?[Tag='GENE'](kw0)}}} ::: distinct sent.cid, kw0.value, kw1.value, kw2.value, sent.value
//S{/NP{/SBAR{//?[Tag='DRUG'](kw2)}}=>/VP{/?[Value IN {'metabolised','metabolized'}](kw1)=>/PP{//?[Tag='GENE'](kw0)}}} ::: distinct sent.cid, kw0.value, kw1.value, kw2.value, sent.value
//S{/NP{/?[Tag='DRUG'](kw2)}=>/VP{/VP{/?[Value IN {'metabolised','metabolized'}](kw1)=>//?[Tag='GENE'](kw0)}}} ::: distinct sent.cid, kw0.value, kw1.value, kw2.value, sent.value
//S{/NP{/?[Tag='DRUG'](kw2)}=>/VP{/VP{//?[Value IN {'metabolized','metabolised'}](kw1)=>//?[Tag='GENE'](kw0)}}} ::: distinct sent.cid, kw0.value, kw1.value, kw2.value, sent.value
//S{/?[Tag='DRUG'](kw2)=>/VP{/VP{/?[Value IN {'metabolised','metabolized'}](kw1)=>//?[Tag='GENE'](kw0)}}} ::: distinct sent.cid, kw0.value, kw1.value, kw2.value, sent.value
//S{/?[Tag='DRUG'](kw2)=>/VP{/?[Value='metabolized'](kw1)=>/PP{//?[Tag='GENE'](kw0)}}} ::: distinct sent.cid, kw0.value, kw1.value, kw2.value, sent.value
//S{/?[Tag='DRUG'](kw2)=>/NP{/?[Value='metabolism'](kw1)}=>/VP{/VP{//?[Tag='GENE'](kw0)}}} ::: distinct sent.cid, kw0.value, kw1.value, kw2.value, sent.value
//S{/NP{/?[Tag='GENE'](kw0)}=>/VP{/NP{//?[Tag='DRUG'](kw2)=>//?[Value='metabolism'](kw1)}}} ::: distinct sent.cid, kw0.value, kw1.value, kw2.value, sent.value
//NP{/NP{/?[Tag='GENE'](kw0)=>/?[Value='metabolism'](kw1)}=>/PP{/NP{/?[Tag='DRUG'](kw2)}}} ::: distinct sent.cid, kw0.value, kw1.value, kw2.value, sent.value
//NP{/?[Value='metabolism'](kw1)=>/PP{/NP{//?[Tag='DRUG'](kw2)=>//?[Tag='GENE'](kw0)}}} ::: distinct sent.cid, kw0.value, kw1.value, kw2.value, sent.value
//S{/NP{/?[Tag='GENE'](kw0)}=>/VP{/VP{//?[Value='metabolism'](kw1)=>//?[Tag='DRUG'](kw2)}}} ::: distinct sent.cid, kw0.value, kw1.value, kw2.value, sent.value
//S{/NP{/NP{/?[Tag='GENE'](kw0)}}=>/VP{/NP{//?[Value='metabolism'](kw1)=>//?[Tag='DRUG'](kw2)}}} ::: distinct sent.cid, kw0.value, kw1.value, kw2.value, sent.value
//NP{/NP{/PP{//?[Tag='GENE'](kw0)}}=>/PP{/NP{/?[Tag='DRUG'](kw2)=>/?[Value='metabolism'](kw1)}}} ::: distinct sent.cid, kw0.value, kw1.value, kw2.value, sent.value
//S{/NP{/?[Tag='GENE'](kw0)}=>/VP{/PP{//?[Value='metabolism'](kw1)=>//?[Tag='DRUG'](kw2)}}} ::: distinct sent.cid, kw0.value, kw1.value, kw2.value, sent.value
//NP{/NP{/?[Tag='DRUG'](kw2)=>/?[Value='metabolism'](kw1)}=>/PP{/NP{//?[Tag='GENE'](kw0)}}} ::: distinct sent.cid, kw0.value, kw1.value, kw2.value, sent.value
//NP{/NP{/?[Tag='GENE'](kw0)=>/?[Value='metabolism'](kw1)}=>/PP{/NP{//?[Tag='DRUG'](kw2)}}} ::: distinct sent.cid, kw0.value, kw1.value, kw2.value, sent.value
//S{/NP{/NP{/?[Value='metabolism'](kw1)}=>/PP{//?[Tag='DRUG'](kw2)}}=>/VP{/VP{//?[Tag='GENE'](kw0)}}} ::: distinct sent.cid, kw0.value, kw1.value, kw2.value, sent.value
//S{/NP{/?[Tag='GENE'](kw0)}=>/VP{/PP{//?[Tag='DRUG'](kw2)=>//?[Value='metabolism'](kw1)}}} ::: distinct sent.cid, kw0.value, kw1.value, kw2.value, sent.value
//NP{/NP{/PP{//?[Tag='GENE'](kw0)}}=>/PP{/NP{//?[Value='metabolism'](kw1)=>//?[Tag='DRUG'](kw2)}}} ::: distinct sent.cid, kw0.value, kw1.value, kw2.value, sent.value
//S{/NP{/?[Tag='DRUG'](kw2)}=>/VP{/NP{//?[Value='substrate'](kw1)=>//?[Tag='GENE'](kw0)}}} ::: distinct sent.cid, kw0.value, kw1.value, kw2.value, sent.value
//NP{/NP{/?[Tag='DRUG'](kw2)=>/NP{/?[Value='substrate'](kw1)}=>/PP{//?[Tag='GENE'](kw0)}}} ::: distinct sent.cid, kw0.value, kw1.value, kw2.value, sent.value
//S{/?[Tag='DRUG'](kw2)=>/VP{/VP{//?[Value='substrate'](kw1)=>//?[Tag='GENE'](kw0)}}} ::: distinct sent.cid, kw0.value, kw1.value, kw2.value, sent.value
//NP{/?[Tag='GENE'](kw0)=>/?[Value='substrate'](kw1)=>/?[Tag='DRUG'](kw2)} ::: distinct sent.cid, kw0.value, kw1.value, kw2.value, sent.value
//NP{/?[Tag='DRUG'](kw2)=>/NP{/NP{/?[Value='substrate'](kw1)}=>/PP{//?[Tag='GENE'](kw0)}}} ::: distinct sent.cid, kw0.value, kw1.value, kw2.value, sent.value
\end{lstlisting}

\section{Gene Regulates Gene}
\begin{lstlisting}[breaklines,numbers=left,basicstyle=\footnotesize]
//S{/NP{/?[Tag='GENE'](kw0)}=>/VP{/VP{//?[Value IN {'regulated', 'upregulated', 'downregulated', 'up-regulated', 'down-regulated'}](kw1)=>//?[Value='by'](kw3)=>//?[Tag='GENE'](kw2)}}} ::: distinct sent.cid, kw0.value, kw1.value, kw2.value, kw3.value, sent.value
//S{/NP{/?[Tag='GENE'](kw0)}=>/VP{/VP{/?[Value IN {'regulated', 'upregulated', 'downregulated', 'up-regulated', 'down-regulated'}](kw1)=>//?[Value='by'](kw3)=>//?[Tag='GENE'](kw2)}}} ::: distinct sent.cid, kw0.value, kw1.value, kw2.value, kw3.value, sent.value
//S{/NP{/?[Tag='GENE'](kw0)}=>/VP{/?[Value IN {'regulated','down-regulated'}](kw1)=>/NP{//?[Tag='GENE'](kw2)}}} ::: distinct sent.cid, kw0.value, kw1.value, kw2.value, sent.value
//NP{/NP{/?[Value IN {'regulation', 'upregulation', 'downregulation', 'up-regulation', 'down-regulation'}](kw1)}=>/PP{/NP{//?[Tag='GENE'](kw0)=>//?[Value='by'](kw3)=>//?[Tag='GENE'](kw2)}}} ::: distinct sent.cid, kw0.value, kw1.value, kw2.value, kw3.value, sent.value
//S{/NP{/?[Tag='GENE'](kw0)}=>/VP{/?[Value IN {'regulates', 'upregulates', 'downregulates', 'up-regulates', 'down-regulates'}](kw1)=>/NP{//?[Tag='GENE'](kw2)}}} ::: distinct sent.cid, kw0.value, kw1.value, kw2.value, sent.value
//S{/NP{/?[Tag='GENE'](kw0)}=>/VP{/VP{//?[Value='in'](kw3)=>//?[Value='regulating'](kw1)=>//?[Tag='GENE'](kw2)}}} ::: distinct sent.cid, kw0.value, kw1.value, kw2.value, kw3.value, sent.value
\end{lstlisting}

\section{Gene Regulate Gene (Xenobiotic Metabolism)}
\begin{lstlisting}[breaklines,numbers=left,basicstyle=\footnotesize]
//S{//?[Tag='GENE' AND Canonical LIKE 'CYP\%'](kw0)<=>//?[Tag='GENE' AND Canonical IN {'AhR', 'CASR', 'CAR', 'PXR', 'NR1I2', 'NR1I3'}](kw1)} ::: distinct sent.cid, kw0.value, kw1.value, sent.value
//S{//?[Tag='GENE' AND Value LIKE 'cytochrome\%'](kw0)<=>//?[Tag='GENE' AND Canonical IN {'AhR', 'CASR', 'CAR', 'PXR', 'NR1I2', 'NR1I3'}](kw1)} ::: distinct sent.cid, kw0.value, kw1.value, sent.value
\end{lstlisting}

\section{Negative Drug Induces/Metabolizes/Inhibits Gene}
\begin{lstlisting}[breaklines,numbers=left,basicstyle=\footnotesize]
//S{/?[Tag='DRUG'](kw0)=>/VP{/VP{/?[Value IN {'induced', 'inhibited', 'metabolized', 'metabolised'}](kw1)=>//?[Tag='GENE'](kw2)=>//?[Value='not'](kw3)=>//?[Tag='GENE'](kw4)}}} ::: distinct sent.cid, kw0.value, kw1.value, kw4.value, kw3.value, sent.value
//S{/?[Tag='DRUG'](kw0)=>/?[Value='not'](kw3)=>/?[Tag='DRUG'](kw4)=>/?[Value='inhibited'](kw1)=>/?[Tag='GENE'](kw2)} ::: distinct sent.cid, kw4.value, kw1.value, kw2.value, kw3.value, sent.value
//S{/SBAR{/S{//?[Tag='DRUG'](kw0)}}=>/S{/S{//?[Value='metabolized'](kw1)=>//?[Tag='GENE'](kw2)=>//?[Value='not'](kw3)=>//?[Tag='GENE'](kw4)}}} ::: distinct sent.cid, kw0.value, kw1.value, kw4.value, kw3.value, sent.value
//S{/NP{/PP{//?[Tag='GENE'](kw2)}}=>/VP{/VP{/?[Value IN {'induced', 'inhibited', 'metabolized', 'metabolised'}](kw1)=>//?[Tag='DRUG'](kw0)=>//?[Value='not'](kw3)=>//?[Tag='DRUG'](kw4)}}} ::: distinct sent.cid, kw4.value, kw1.value, kw2.value, kw3.value, sent.value
//S{/NP{/PP{//?[Tag='DRUG'](kw0)}}=>/VP{/VP{/?[Value IN {'induced', 'inhibited', 'metabolized', 'metabolised'}](kw1)=>//?[Tag='GENE'](kw2)=>//?[Value IN {'not','no'}](kw3)=>//?[Tag='GENE'](kw4)}}} ::: distinct sent.cid, kw0.value, kw1.value, kw4.value, kw3.value, sent.value
//S{/?[Tag='DRUG'](kw0)=>/VP{/?[Value='not'](kw3)=>/VP{//?[Value IN {'induced', 'inhibited', 'metabolized', 'metabolised'}](kw1)=>//?[Tag='GENE'](kw2)}}} ::: distinct sent.cid, kw0.value, kw1.value, kw2.value, kw3.value, sent.value
//S{/NP{/?[Tag='GENE'](kw2)}=>/VP{/VP{/?[Value IN {'induced', 'inhibited', 'metabolized', 'metabolised'}](kw1)=>//?[Tag='DRUG'](kw0)=>//?[Value='not'](kw3)=>//?[Tag='DRUG'](kw4)}}} ::: distinct sent.cid, kw4.value, kw1.value, kw2.value, kw3.value, sent.value
//S{/NP{/NP{/?[Value='not'](kw3)=>/?[Tag='DRUG'](kw0)}}=>/VP{/?[Value IN {'inhibited','induced'}](kw1)=>/NP{//?[Tag='GENE'](kw2)}}} ::: distinct sent.cid, kw0.value, kw1.value, kw2.value, kw3.value, sent.value
//S{/NP{/NP{/?[Value IN {'no','not'}](kw3)=>/?[Tag='DRUG'](kw0)}}=>/VP{/?[Value IN {'inhibited','induced'}](kw1)=>/NP{/?[Tag='GENE'](kw2)}}} ::: distinct sent.cid, kw0.value, kw1.value, kw2.value, kw3.value, sent.value
//S{/?[Tag='DRUG'](kw0)=>/S{/S{//?[Value='not'](kw3)=>//?[Value IN {'induce', 'inhibit'}](kw1)=>//?[Tag='GENE'](kw2)}}} ::: distinct sent.cid, kw0.value, kw1.value, kw2.value, kw3.value, sent.value
//S{/?[Tag='DRUG'](kw0)=>/?[Tag='GENE'](kw2)=>/?[Value IN {'not'}](kw3)=>/?[Value IN {'inhibit','induce'}](kw1)} ::: distinct sent.cid, kw0.value, kw1.value, kw2.value, kw3.value, sent.value
//S{/?[Tag='DRUG'](kw0)=>/VP{/?[Value='not'](kw3)=>/VP{/?[Value IN {'inhibit','induce'}](kw1)=>//?[Tag='GENE'](kw2)}}} ::: distinct sent.cid, kw0.value, kw1.value, kw2.value, kw3.value, sent.value
//S{/NP{/?[Tag='DRUG'](kw0)}=>/VP{/VP{/?[Value='not'](kw3)=>//?[Value='inhibit'](kw1)=>//?[Tag='GENE'](kw2)}}} ::: distinct sent.cid, kw0.value, kw1.value, kw2.value, kw3.value, sent.value
//S{/NP{/NP{/?[Tag='GENE'](kw2)}}=>/VP{/?[Value='not'](kw3)=>/VP{/?[Value='metabolize'](kw1)=>//?[Tag='DRUG'](kw0)}}} ::: distinct sent.cid, kw0.value, kw1.value, kw2.value, kw3.value, sent.value
//S{/NP{/NP{//?[Tag='DRUG'](kw0)}}=>/VP{/?[Value='not'](kw3)=>/VP{/?[Value IN {'inhibit','induce'}](kw1)=>//?[Tag='GENE'](kw2)}}} ::: distinct sent.cid, kw0.value, kw1.value, kw2.value, kw3.value, sent.value
//S{/NP{/?[Tag='DRUG'](kw0)}=>/VP{/?[Value='not'](kw3)=>/VP{/?[Value='inhibit'](kw1)=>//?[Tag='GENE'](kw2)}}} ::: distinct sent.cid, kw0.value, kw1.value, kw2.value, kw3.value, sent.value
//S{/NP{/?[Tag='DRUG'](kw0)}=>/VP{/?[Value IN {'induces', 'inhibits'}](kw1)=>/NP{//?[Tag='GENE'](kw2)=>//?[Value='not'](kw3)=>//?[Tag='GENE'](kw4)}}} ::: distinct sent.cid, kw0.value, kw1.value, kw4.value, kw3.value, sent.value
//S{/?[Tag='DRUG'](kw0)=>/VP{/?[Value='inhibits'](kw1)=>/NP{//?[Tag='GENE'](kw2)=>//?[Value='not'](kw3)=>//?[Tag='GENE'](kw4)}}} ::: distinct sent.cid, kw0.value, kw1.value, kw4.value, kw3.value, sent.value
//S{/?[Tag='DRUG'](kw0)=>/VP{/NP{//?[Value='no'](kw3)=>//?[Value IN {'induction','metabolism','inhibition'}](kw1)=>//?[Tag='GENE'](kw2)}}} ::: distinct sent.cid, kw0.value, kw1.value, kw2.value, kw3.value, sent.value
//S{/NP{/NP{//?[Tag='DRUG'](kw0)=>//?[Tag='GENE'](kw2)=>//?[Value IN {'induction','metabolism','inhibition'}](kw1)}}=>/VP{/?[Value='not'](kw3)}} ::: distinct sent.cid, kw0.value, kw1.value, kw2.value, kw3.value, sent.value
//S{/?[Tag='DRUG'](kw0)=>/VP{/?[Value='not'](kw3)=>/VP{//?[Value IN {'induction','metabolism','inhibition'}](kw1)=>//?[Tag='GENE'](kw2)}}} ::: distinct sent.cid, kw0.value, kw1.value, kw2.value, kw3.value, sent.value
//S{/NP{/PP{//?[Tag='DRUG'](kw0)}}=>/VP{/?[Value='not'](kw3)=>/VP{//?[Value IN {'induction','metabolism','inhibition'}](kw1)=>//?[Tag='GENE'](kw2)}}} ::: distinct sent.cid, kw0.value, kw1.value, kw2.value, kw3.value, sent.value
\end{lstlisting}

\section{Negative Drug Induces Gene}
\begin{lstlisting}[breaklines,numbers=left,basicstyle=\footnotesize]
//S{//?[Tag='GENE'](kw0)=>//?[Value IN {'induced','increased','stimulated'}](kw1)=>//?[Value='by'](kw3)=>//?[Tag='DRUG'](kw2)=>//?[Value='not'](kw4)=>//?[Tag='DRUG'](kw5)} ::: distinct sent.cid, kw5.value, kw1.value, kw0.value, kw4.value, sent.value
//S{//?[Tag='GENE'](kw0)=>//?[Value='not'](kw4)=>//?[Tag='GENE'](kw5)=>//?[Value IN {'induced','increased','stimulated'}](kw1)=>//?[Value='by'](kw3)=>//?[Tag='DRUG'](kw2)} ::: distinct sent.cid, kw2.value, kw1.value, kw5.value, kw4.value, sent.value
//S{//?[Tag='DRUG'](kw0)=>//?[Value='not'](kw4)=>//?[Value IN {'induce','induced','increase','increased','stimulate','stimulated'}](kw1)=>//?[Tag='GENE'](kw2)} ::: distinct sent.cid, kw0.value, kw1.value, kw2.value, kw4.value, sent.value
//S{//?[Tag='DRUG'](kw0)=>//?[Value='not'](kw4)=>//?[Tag='DRUG'](kw5)=>//?[Value IN {'induces','increases','stimulates','induced','increased','stimulated'}](kw1)=>//?[Tag='GENE'](kw2)} ::: distinct sent.cid, kw5.value, kw1.value, kw2.value, kw4.value, sent.value
//S{//?[Value='no'](kw5)=>//?[value IN {'induction','stimulation'}](kw1)=>//?[Value IN {'of'}](kw2)=>//?[Tag='GENE'](kw0)=>//?[value IN {'by'}](kw3)=>//?[Tag='DRUG'](kw4)} ::: distinct sent.cid, kw4.value, kw1.value, kw0.value, kw5.value, sent.value
\end{lstlisting}

\section{Negative Drug Inhibits Gene}
\begin{lstlisting}[breaklines,numbers=left,basicstyle=\footnotesize]
//S{//?[Tag='GENE'](kw0)=>//?[Value IN {'inhibited','decreased'}](kw1)=>//?[Value='by'](kw3)=>//?[Tag='DRUG'](kw2)=>//?[Value='not'](kw4)=>//?[Tag='DRUG'](kw5)} ::: distinct sent.cid, kw5.value, kw1.value, kw0.value, kw4.value, sent.value
//S{//?[Tag='GENE'](kw0)=>//?[Value='not'](kw4)=>//?[Tag='GENE'](kw5)=>//?[Value IN {'inhibited','decreased'}](kw1)=>//?[Value='by'](kw3)=>//?[Tag='DRUG'](kw2)} ::: distinct sent.cid, kw2.value, kw1.value, kw5.value, kw4.value, sent.value
//S{//?[Tag='DRUG'](kw0)=>//?[Value='not'](kw4)=>//?[Value IN {'inhibit','inhibited','decrease','decreased'}](kw1)=>//?[Tag='GENE'](kw2)} ::: distinct sent.cid, kw0.value, kw1.value, kw2.value, kw4.value, sent.value
//S{//?[Tag='DRUG'](kw0)=>//?[Value='not'](kw4)=>//?[Tag='DRUG'](kw5)=>//?[Value IN {'inhibits','decreases','inhibited','decreased'}](kw1)=>//?[Tag='GENE'](kw2)} ::: distinct sent.cid, kw5.value, kw1.value, kw2.value, kw4.value, sent.value
//S{//?[Value='no'](kw5)=>//?[value IN {'inhibition'}](kw1)=>//?[Value IN {'of'}](kw2)=>//?[Tag='GENE'](kw0)=>//?[value IN {'by'}](kw3)=>//?[Tag='DRUG'](kw4)} ::: distinct sent.cid, kw4.value, kw1.value, kw0.value, kw5.value, sent.value
\end{lstlisting}

\section{Negative Gene Metabolizes Drug}
\begin{lstlisting}[breaklines,numbers=left,basicstyle=\footnotesize]
//S{//?[Tag='DRUG'](kw0)=>//?[Value IN {'metabolized','metabolised'}](kw1)=>//?[Value='by'](kw3)=>//?[Tag='GENE'](kw2)=>//?[Value='not'](kw4)=>//?[Tag='GENE'](kw5)} ::: distinct sent.cid, kw0.value, kw1.value, kw5.value, kw4.value, sent.value
//S{//?[Tag='DRUG'](kw0)=>//?[Value='not'](kw4)=>//?[Tag='DRUG'](kw5)=>//?[Value IN {'metabolized','metabolised'}](kw1)=>//?[Value='by'](kw3)=>//?[Tag='GENE'](kw2)} ::: distinct sent.cid, kw5.value, kw1.value, kw2.value, kw4.value, sent.value
//S{//?[Tag='GENE'](kw0)=>//?[Value='not'](kw4)=>//?[Value IN {'metabolize','metabolise'}](kw1)=>//?[Tag='DRUG'](kw2)} ::: distinct sent.cid, kw2.value, kw1.value, kw0.value, kw4.value, sent.value
//S{//?[Tag='GENE'](kw0)=>//?[Value='not'](kw4)=>//?[Tag='GENE'](kw5)=>//?[Value IN {'metabolize','metabolise','metabolizes','metabolises'}](kw1)=>//?[Tag='DRUG'](kw2)} ::: distinct sent.cid, kw2.value, kw1.value, kw5.value, kw4.value, sent.value
\end{lstlisting}

\section{Negative Gene Downregulates Gene}
\begin{lstlisting}[breaklines,numbers=left,basicstyle=\footnotesize]
//S{//?[Tag='GENE'](kw0)=>//?[Value IN {'suppressed','downregulated','inhibited','down-regulated','repressed','disrupted'}](kw1)=>//?[Value='by'](kw3)=>//?[Tag='GENE'](kw2)=>//?[Value='not'](kw4)=>//?[Tag='GENE'](kw5)} ::: distinct sent.cid, kw5.value, kw1.value, kw0.value, kw4.value, sent.value
//S{//?[Tag='GENE'](kw0)=>//?[Value='not'](kw4)=>//?[Tag='GENE'](kw5)=>//?[Value IN {'suppressed','downregulated','inhibited','down-regulated','repressed','disrupted'}](kw1)=>//?[Value='by'](kw3)=>//?[Tag='GENE'](kw2)} ::: distinct sent.cid, kw2.value, kw1.value, kw5.value, kw4.value, sent.value
//S{//?[Tag='GENE'](kw0)=>//?[Value='not'](kw4)=>//?[Value IN {'suppressed','suppress','downregulated','downregulate','inhibited','inhibit','down-regulated','down-regulate','repressed','repress','disrupted','disrupt'}](kw1)=>//?[Tag='GENE'](kw2)} ::: distinct sent.cid, kw0.value, kw1.value, kw2.value, kw4.value, sent.value
//S{//?[Tag='GENE'](kw0)=>//?[Value='not'](kw4)=>//?[Tag='GENE'](kw5)=>//?[Value IN {'suppresses','downregulates','inhibits','down-regulates','represses','disrupts','suppressed','downregulated','inhibited','down-regulated','repressed','disrupted'}](kw1)=>//?[Tag='GENE'](kw2)} ::: distinct sent.cid, kw5.value, kw1.value, kw2.value, kw4.value, sent.value
//S{//?[Value='no'](kw5)=>//?[value IN {'inhibition','downregulation','down-regulation'}](kw1)=>//?[Value IN {'of'}](kw2)=>//?[Tag='GENE'](kw0)=>//?[value IN {'on'}](kw3)=>//?[Tag='GENE'](kw4)} ::: distinct sent.cid, kw0.value, kw1.value, kw4.value, kw5.value, sent.value
\end{lstlisting}

\section{Negative Gene Upregulates Gene}
\begin{lstlisting}[breaklines,numbers=left,basicstyle=\footnotesize]
//S{//?[Tag='GENE'](kw0)=>//?[Value IN {'activated','induced','stimulated','regulated','upregulated','up-regulated'}](kw1)=>//?[Value='by'](kw3)=>//?[Tag='GENE'](kw2)=>//?[Value='not'](kw4)=>//?[Tag='GENE'](kw5)} ::: distinct sent.cid, kw5.value, kw1.value, kw0.value, kw4.value, sent.value
//S{//?[Tag='GENE'](kw0)=>//?[Value='not'](kw4)=>//?[Tag='GENE'](kw5)=>//?[Value IN {'activated','induced','stimulated','regulated','upregulated','up-regulated'}](kw1) ::: distinct sent.cid, kw2.value, kw1.value, kw5.value, kw4.value, sent.value
//S{//?[Tag='GENE'](kw0)=>//?[Value='not'](kw4)=>//?[Value IN {'activated','induced','stimulated','regulated','upregulated','up-regulated','activate','induce','stimulate','regulate','upregulate','up-regulate'}](kw1)=>//?[Tag='GENE'](kw2)} ::: distinct sent.cid, kw0.value, kw1.value, kw2.value, kw4.value, sent.value
//S{//?[Tag='GENE'](kw0)=>//?[Value='not'](kw4)=>//?[Tag='GENE'](kw5)=>//?[Value IN {'activates','induces','stimulates','regulates','upregulates','up-regulates','activate','induce','stimulate','regulate','upregulate','up-regulate'}](kw1)=>//?[Tag='GENE'](kw2)} ::: distinct sent.cid, kw5.value, kw1.value, kw2.value, kw4.value, sent.value
//S{//?[Value='no'](kw5)=>//?[value IN {'induction','activation','stimulation','regulation','upregulation','up-regulation'}](kw1)=>//?[Value IN {'of'}](kw2)=>//?[Tag='GENE'](kw0)=>//?[value IN {'by'}](kw3)=>//?[Tag='GENE'](kw4)} ::: distinct sent.cid, kw4.value, kw1.value, kw0.value, kw5.value, sent.value
\end{lstlisting}

\section{Drug Gene Co-Occurrence}
\begin{lstlisting}[breaklines,numbers=left,basicstyle=\footnotesize]
//S{//?[Tag='DRUG'](kw0)<=>//?[Tag='GENE'](kw1)} ::: distinct sent.cid, kw0.value, kw1.value, kw0.type, kw1.type, sent.value
\end{lstlisting}

\bibliographystyle{asudis}
\bibliography{thesis}

\end{document}